\pdfoutput=1
\documentclass{article}

\PassOptionsToPackage{numbers, compress}{natbib}

\usepackage[preprint]{neurips_2024}

\usepackage[utf8]{inputenc} %
\usepackage[T1]{fontenc}    %
\usepackage{hyperref}       %
\usepackage{url}            %
\usepackage{booktabs}       %
\usepackage{amsfonts}       %
\usepackage{nicefrac}       %
\usepackage{microtype}      %
\usepackage{xcolor}         %

\usepackage{subcaption}
\usepackage{graphicx}
\usepackage{float}
\usepackage{natbib}

\usepackage{amsmath,amsfonts,bm}

\def\eqref#1{equation~\ref{#1}}

\def\1{\bm{1}}

\DeclareMathAlphabet{\mathsfit}{\encodingdefault}{\sfdefault}{m}{sl}
\SetMathAlphabet{\mathsfit}{bold}{\encodingdefault}{\sfdefault}{bx}{n}

\def\gM{{\mathcal{M}}}

\DeclareMathOperator*{\argmax}{arg\,max}

\input{Definitions.sty}

\newcommand{\GuessPhase}{\emph{Guessing-Phase}\xspace}
\newcommand{\MemPhase}{\emph{Memorisation-Phase}\xspace}
\newcommand{\random}{\texttt{Random}}
\newcommand{\constant}{\texttt{Constant}}

\newcommand{\allModelsWidth}{0.23\textwidth}

\newcommand{\smallThirdWidth}{0.3\textwidth}
\newcommand{\thirdWidth}{0.33\textwidth}
\newcommand{\capthead}[2]{\textbf{[#1 (#2)]}}

\title{Understanding Memorisation in LLMs:\\ Dynamics, Influencing Factors, and Implications}

\author{%
    Till Speicher \\
    MPI-SWS \\
    \And
    Mohammad Aflah Khan \\
    MPI-SWS, IIIT Delhi \\
    \And
    Qinyuan Wu \\
    MPI-SWS \\
    \And
    Vedant Nanda \\
    MPI-SWS \\
    \And
    Soumi Das \\
    MPI-SWS \\
    \And
    Bishwamittra Ghosh \\
    MPI-SWS \\
    \And
    Krishna P. Gummadi \\
    MPI-SWS \\
    \And
    Evimaria Terzi \\
    Boston University \\
}

\begin{document}

\maketitle

\begin{abstract}
Understanding whether and to what extent large language models (LLMs) have memorised training data has important implications for the reliability of their output and the privacy of their training data. In order to
cleanly measure and disentangle memorisation from other phenomena ({\eg} in-context learning), we create an experimental framework that is based on \emph{repeatedly exposing LLMs to random strings}.
Our framework allows us to better understand the \emph{dynamics}, {\ie}, the behaviour of the model, when repeatedly exposing it to  random strings.
Using our framework, we make several striking observations:
(a) we find consistent phases of the dynamics across families of models (Pythia, Phi and Llama2),
(b) we identify factors that make some strings easier to memorise than others,
 and
(c) we identify the role of local prefixes and global context in memorisation.
We also show that sequential exposition to different random strings has a significant effect on memorisation. Our results, often surprising, have significant downstream implications in the study and usage of LLMs.
\end{abstract}

\section{Introduction}
The potential for large language models (LLMs) to memorise training data has many important implications. %
Our goal in this paper is to create a framework that enables us to better understand, measure and distinguish memorisation from other phenomena ({\eg}, in-context learning)
that influence the generation (recollection) of the next token in LLMs.
Such distinction is particularly challenging in an era where models are trained on internet-scale corpora and even open-weight models often have no documentation about the training data they used.

We study memorisation by creating a ``clean" experimental setting, where we train LLMs to memorise \emph{random strings}.
Random strings allow us to study memorisation in isolation from other phenomena.
Random strings i) guarantee that models have not seen the data during pre-training, ii) ensure that models have to memorize the data in order to achieve low loss ({\ie} there is no other way to predict tokens better than a distribution-based guess), and iii) give us precise control over all aspects of the data, such as string length, alphabet size, and entropy.
Achieving all of these properties with natural language would not be possible.
Note that privacy sensitive data in the real-world looks often similar to random strings (e.g., private keys, phone numbers, etc.).
Thus, our results are  relevant to the memorisation abilities of LLMs on such data.

The goal of our experiments is to unveil phenomena that happen when LLMs memorise data.
Therefore, in our experiments, we \emph{repeatedly} expose models to random strings since, intuitively, repetition is associated with memorisation.
Through this process, we can understand the \emph{dynamics} of memorisation, {\ie} when memorisation ``kicks-in" over the period of repeated exposition to the same or different random strings and how the model behaves.

In order to obtain a comprehensive understanding of the memorisation process in LLMs, we perform an extensive set of experiments over both pretrained and untrained models from different families (Pythia, Phi, Llama, GPT and OPT) with parameter counts spanning more than two orders of magnitude.
We use random data generated from different distributions, vocabularies and string structures.
We train models on random strings appearing in isolation as well as in the context of natural language data. 

In all cases we make the same observations: with repeated exposure to the same random string, models memorise the random strings %
and we consistently observe two phases in this process: the {\GuessPhase} and the {\MemPhase} phase.
In the {\GuessPhase} a model learns the probability distribution of the tokens in the string; in the {\MemPhase}  it memorises the next token based on prefixes and this is when memorisation actually happens.
We further analyse the memorisation process, motivated by the following questions:

\noindent{\bf Q1:}
\emph{Are some strings easier to memorise than others?} 
To address this question, we experiment with strings of different alphabet sizes and different entropy over the distribution of their tokens. 
We find that the entropy of the distribution of tokens in the random strings affects the two-phase dynamics:
in the {\GuessPhase}, strings with lower entropy are predicted better, however, in the subsequent {\MemPhase} strings with higher entropy are memorised faster.

\noindent{\bf Q2:} \emph{What information do models need in order to recall memorised tokens?} 
We address this question by investigating the role of \emph{exact prefixes} in next-token recollection.  We find that as the number of repeated exposures to the random string increases, the shorter the prefix needed for next-token recollection become.
Somewhat surprisingly, we also find that the exact prefix alone is not sufficient.
\emph{Global context}, {\ie}, knowledge of the probability distribution of the tokens in the string, significantly increases next-token recollection accuracy.

\noindent{\bf Q3:} \emph{ How do models behave when asked to  memorise different random strings sequentially?}
We also train models to \emph{sequentially} memorise different random strings -- one at a time.
Our results indicate that models forget old random strings when they get repeatedly exposed to new ones. However, as models get exposed to more random strings they forget less, and they memorise new strings faster.

Contrary to related work on memorisation that aims to quantify privacy risks,
we focus on  understanding and revealing the intricacies of the memorisation process in LLMs.
Some of the phenomena we unveil as we tackle the above questions are surprising and unexpected. Many have significant implications for quantifying memorisation, understanding how memorisation works, and estimating the risks of memorising different types of training data, including uncovering new threats with priming models for memorising data. 
For many of the phenomena we do not have clear explanations as of why they happen but we feel it is important to report them to the community 
as they rule out certain theories related to memorisation and give rise to new ones. We also think that our findings can
motivate further studies that will increase our understanding of memorisation.

\smallskip\noindent\textbf{Related work:} 
The topic of memorisation has received great attention in the context of LLMs that are trained on large ``internet-scale'' data~\citep{song2019auditing, carlini2019secret,Zhang2021CounterfactualMI,biderman2023emergent,mattern23membership,lukas2023analyzing,mccoy2023much}.
Most of these works propose a definition of memorisation to test whether the model can generate a given 
string (present in the training data) using particular prompts or prefixes. While they subtly differ in how exactly they operationalize a measure of memorisation, at a higher level, all these works are concerned with answering the ``why'' question around memorisation, \eg why should memorisation be a practical concern? To this end, these works show compelling examples of cases where memorisation can hurt (\eg privacy leaks via reconstruction~\citep{carlini2021extracting} or membership inference~\citep{mattern23membership}). Similarly, there is also a case to be made for memorisation being desirable in cases where the goal is to generate facts and reduce LLM hallucinations. Grounding the generation by LLMs in some verified training data sources can be an effective way to generate trustworthy information~\citep{li-etal-2023-large,borgeaud2022improving,khandelwal2019generalization,tay2022transformer,alkhamissi2022review,petroni2019language,guu2020retrieval,haviv2022understanding}.

We differ from existing works in two key aspects. Firstly, our key goal is to build a foundational understanding of \textit{how} these models memorise. Thus, we do not engage with the  question of memorisation being desirable or undesirable and rather provide observations on \textit{how} memorisation happens at an input-output level.
Secondly, prior works are motivated by applications and thus simulate scenarios where memorisation happens \textit{unintentionally}, \ie, these works typically do not repeat token sequences during training or finetuning~\citep{tirumala2022memorization,carlini2022quantifying}. We instead force the model to memorise random strings by training on the same tokens multiple times, until the model can generate these random strings verbatim. Our work adds to the nascent literature focused on building a better scientific understanding of memorisation in LLMs (\eg~\citep{tirumala2022memorization,jagielski2022measuring,carlini2022quantifying,kharitonov2021bpe}).

\section{Preliminaries and experimental setup}
\label{sec:preliminaries}

\noindent{\textbf{Approach:}}
Throughout, we use random strings in order to train and test LLMs.
To create a random string, we first choose an \emph{alphabet} $A$ that consists of $|A|=\ell$ unique tokens; we call $\ell$ the \emph{size of the alphabet}.
The alphabet we use for string generation is a subset of a much larger \emph{vocabulary of all tokens} $V$, $A \subset V$.
Tokens in an LLM's vocabulary can range from single characters to entire words and its size
spans from tens of thousands to a few hundred thousand tokens.
In most experiments, we use tokens corresponding to lowercase characters in the Latin alphabet. %

We use
$P_A$ to denote a probability distribution over the unique tokens of the alphabet $A$.
We can compute the \emph{entropy} of $P_A$ using the standard definition of entropy: $H(P_A) = -\sum_{a\in A} P_A(a)\log P_A(a)$.  For any given alphabet of size $\ell$ we use
$H_\ell$ to denote the entropy of the uniform probability distribution over the alphabet's tokens.
We generate a random string $s = (s_1,\ldots , s_n)$ of length $n$ 
by sampling every token $s_i$ independently from $P_A$.
Unless otherwise noted, we assume that $P_A$ is 
the uniform probability distribution over the tokens in $A$.
Given a string $s$ of length $n$ we use $s_{[i,j]}$, with $i\leq j$, to denote the 
substring of $s$ that consists of positions $(s_i,s_{i+1},\ldots ,s_{j-1},s_j)$.

Given a string $s$, we train a causal, \ie~\emph{autoregressive}, language model $\gM$ to memorise $s$.
We do so, by minimizing 
the cross-entropy loss
over all positions of $s$. 
We denote by $P_{\gM}(s_i=t | s_{[1,i-1]})$ the probability that  $\mathcal{M}$ assigns to token $t$ at the $i$-th position of string $s$.
We define the prediction of $\mathcal{M}$ for position $i$ as the token with the largest $P_\mathcal{M}(s_i=t\mid s_{[1,i-1]}))$.
Then, we define the \emph{recollection accuracy} of $\mathcal{M}$ with respect to $s$ based on greedy decoding as:

\begin{equation}\label{eq:accuracy}
\text{Accuracy}(\gM,  s) = \frac{1}{n}\sum_{i=1}^n \delta(s_i = \argmax_{t\in V} P_\mathcal{M}(s_i = t |  s_{[1,i-1]})).
\end{equation}

$\delta(\text{condition})$ is an indicator variable; it takes value $1$ (resp.\ $0$) when the condition is true (resp. false).
We report accuracy values in the main paper, and loss values in the appendix.

\noindent{\bf Data generation process:}
In our experiments, we focus on alphabets $A$ with $\ell \in \{2, 4, 7, 13, 26\}$.
We primarily use tokens corresponding to the first $\ell$ lowercase characters from the Latin alphabet, \ie~$A \subseteq \{a, \dots, z\}$, but also report results for non-Latin alphabets in Appendix~\ref{app:dynamics_untrained}.
We therefore often refer to the elements of the alphabet as characters, even though they are technically tokens.
We generate random strings of lengths $n\in \{16, 32, 64, 128,\dots, 1024\}$ by sampling tokens uniformly at random from $A$.
All our results are aggregates over 5 runs with different random string samples; we highlight one standard deviation in the plots.
We show examples of random strings in Appendix~\ref{app:data_examples}.
We also show that our observations on random string memorisation are robust when random strings appear in the context of natural language data from the \href{https://huggingface.co/datasets/wikitext}{wikitext}~\cite{merity2016pointer} dataset in Section\ref{sec:memorability} and Appendix~\ref{app:real_world_validation}.

\textbf{LLM models:}
We make use of the Pythia \citep{biderman2023pythia}, Phi~\citep{li2023textbooks2} and Llama-2 model~\citep{touvron2023llama2} families. 
For the Pythia family, we use variants with 70M, 1B and 12B parameters, for the Phi family we use 1.3B and 2.7B parameter variants, and for Llama-2 we use 7B and 13B parameter variants.
We refer to each model by its parameter count, \eg,~Pythia-1B or Llama2-13B.
We choose these models, since they represent popular, modern architectures, and span a wide spectrum of parameter counts (more than two orders of magnitude).
We also report results for older GPT-2 and OPT models in Appendix~\ref{app:abs_pos_prefix_mappings}.

\textbf{Training and fine-tuning:} 
We experimented with both untrained and pretrained models to capture the memorisation dynamics early during training and later during continual training or fine-tuning of a pretrained model. All our findings largely hold in both cases, with the primary difference being pretrained models memorise faster. We report the results for both pretrained models (in the main paper) and untrained models (in the appendix).  
A detailed description of the training setup can be found in Appendix~\ref{app:dynamics_setup}.

\section{The dynamics of repeated exposure to random strings}
\label{sec:phases}

\begin{figure}[t]
    \centering
    \subfloat[Pythia-1B]{
        \includegraphics[width=0.32\linewidth]{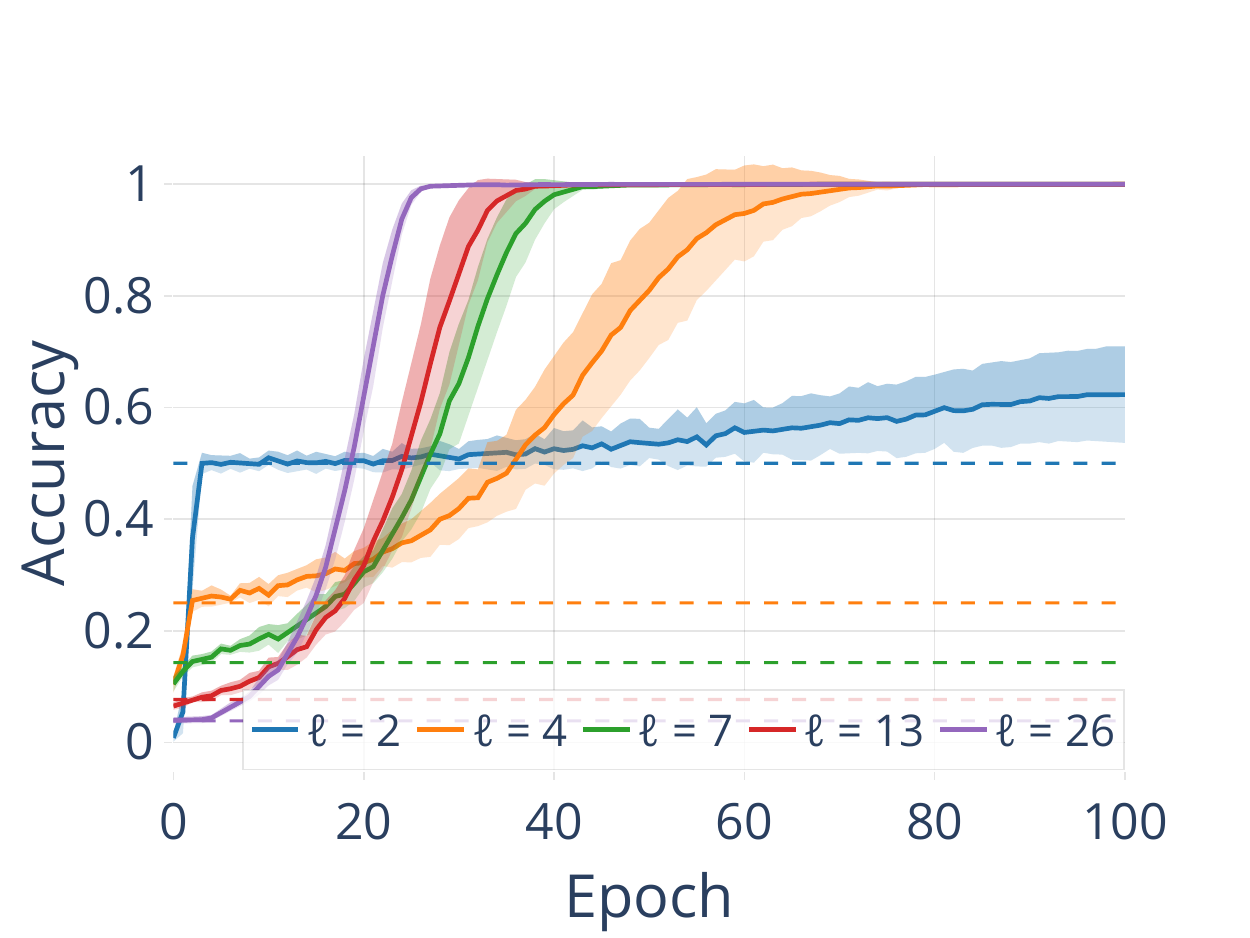}
        \label{fig:accuracy_pythia_1b_alphabet_size}
    }
    \subfloat[Phi-2.7B]{
        \includegraphics[width=0.32\linewidth]{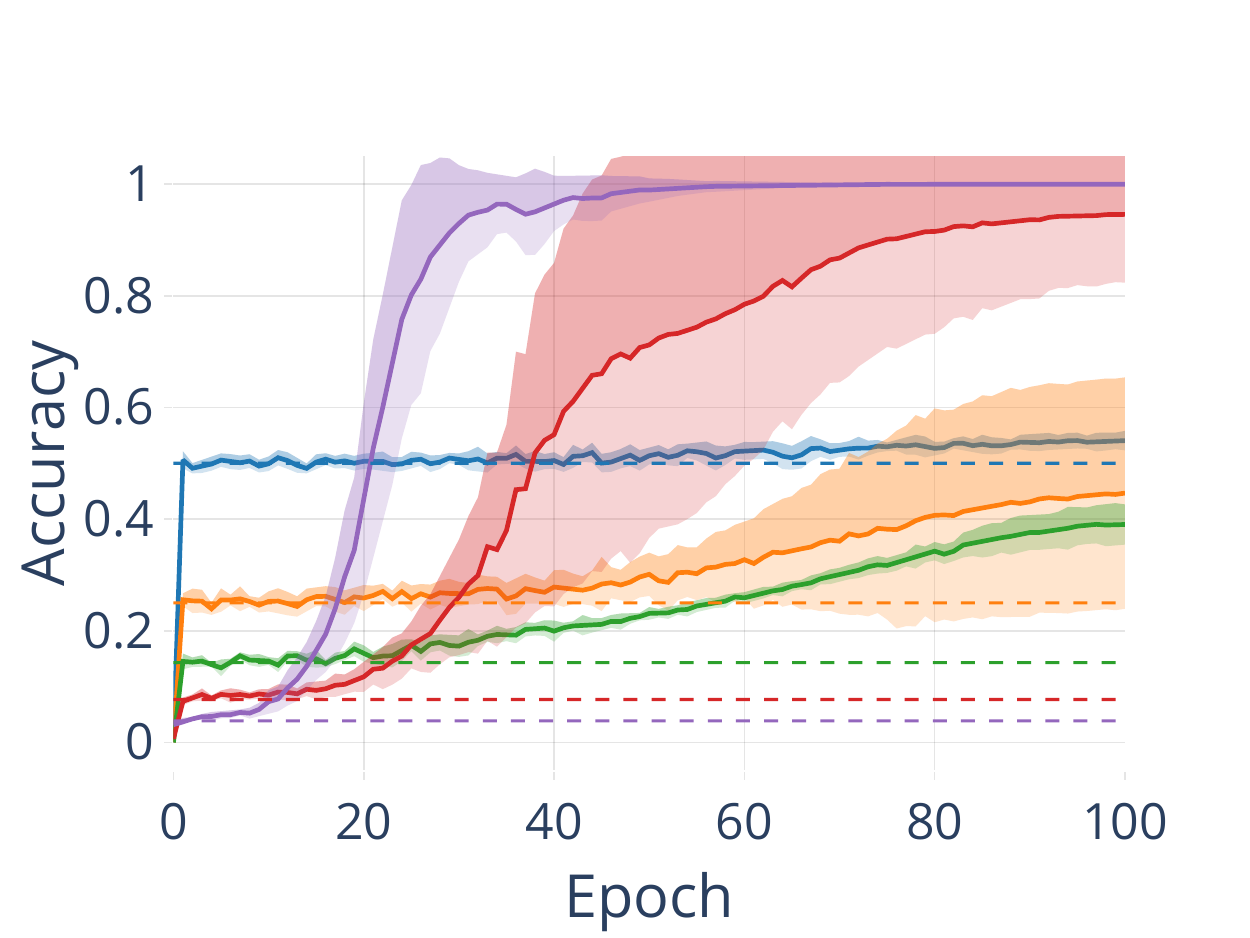}
    }
    \subfloat[Llama2-13B]{
        \includegraphics[width=0.32\linewidth]{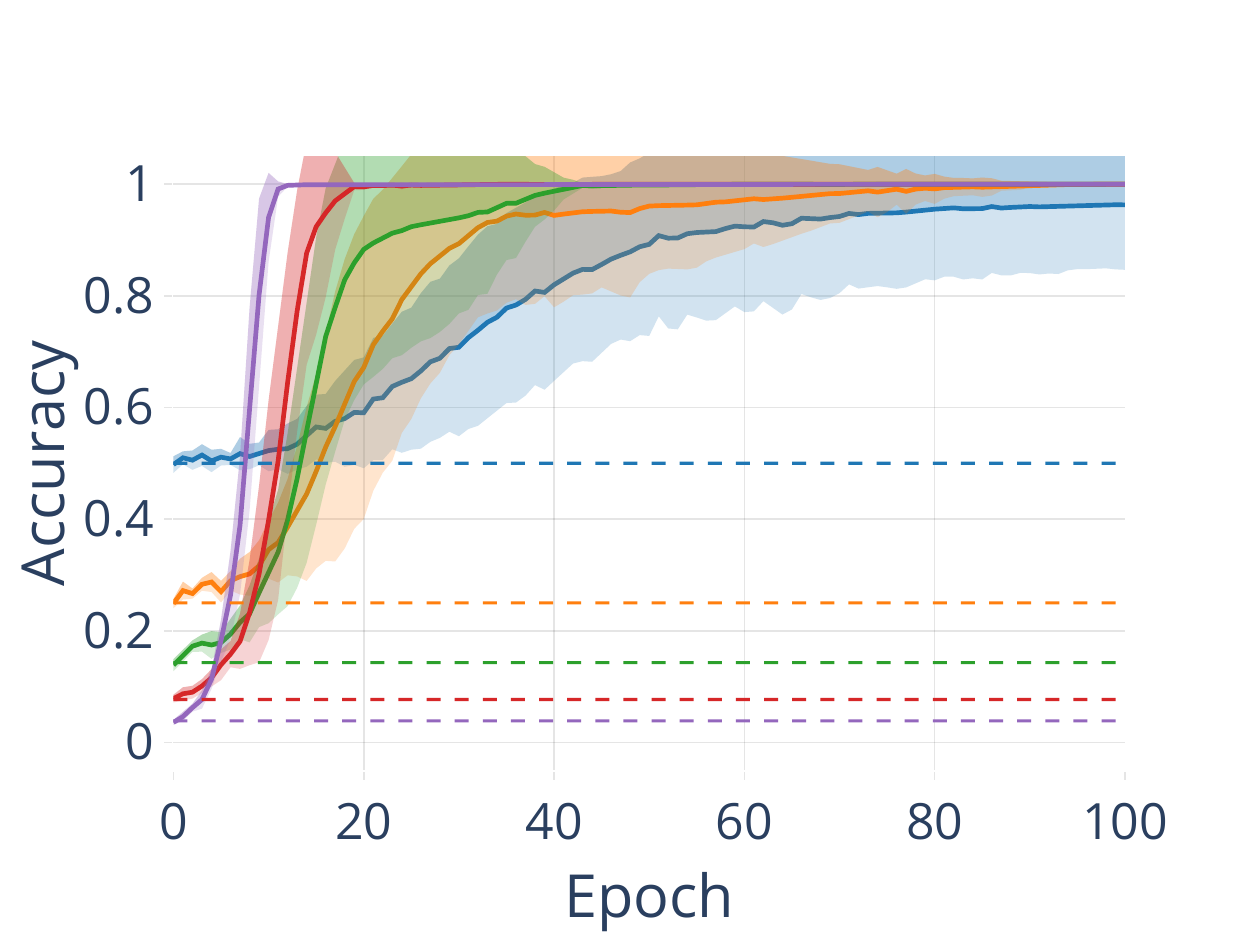}
        \label{fig:accuracy_llama2-13b_alphabet_size}
    }
\caption{\capthead{Recollection accuracy for different alphabet sizes $\ell$ and models $\gM$.}{$n = 1024$}
For all models, the accuracy initially increases quickly before stagnating at the random guess level during the \GuessPhase.
Afterwards, the accuracy converges more slowly towards $1$ during the \MemPhase.
The accuracy of randomly guessing tokens from $A$ is shown with dashed lines.
}
\vspace{-1mm}
\label{fig:accuracy_loss_alphabet_size}
\end{figure}

We start by providing some basic observations we make as we repeatedly expose models to the same random string for $100$ epochs. Figure~\ref{fig:accuracy_loss_alphabet_size} shows the recollection accuracy of three models for random strings with $\ell=2,4,7,13,26$.
The dotted lines in the two plots show the accuracy of a model that performs random guessing over alphabet $A$.

\textbf{Two phases:}
At a high level, Figure~\ref{fig:accuracy_loss_alphabet_size} shows that for all models $\gM$ the accuracy converges towards $1$ as training progresses.
The convergence is not uniform, however, since after an initial rise, the accuracy reaches a plateau at the random guessing baseline (\eg for Pythia-1B and Phi-2.7B after about 8 epochs), before increasing towards towards $1$ more slowly afterwards.
We observe this pattern consistently across all $\ell$ and $\gM$.
As models are repeatedly exposed to the same random string, they go through two phases:
During the first phase, the accuracy of the model reaches that of a random guess and then plateaus; thus, we call this the {\GuessPhase}.
During the second phase, the accuracy of the model exceeds the guessing plateau and converges to $1$.
We call this phase the {\MemPhase}.

\begin{figure}[ht]
    \centering
    \subfloat[Pythia-1B, Aggregate Prob.]{
        \includegraphics[width=0.32\linewidth]{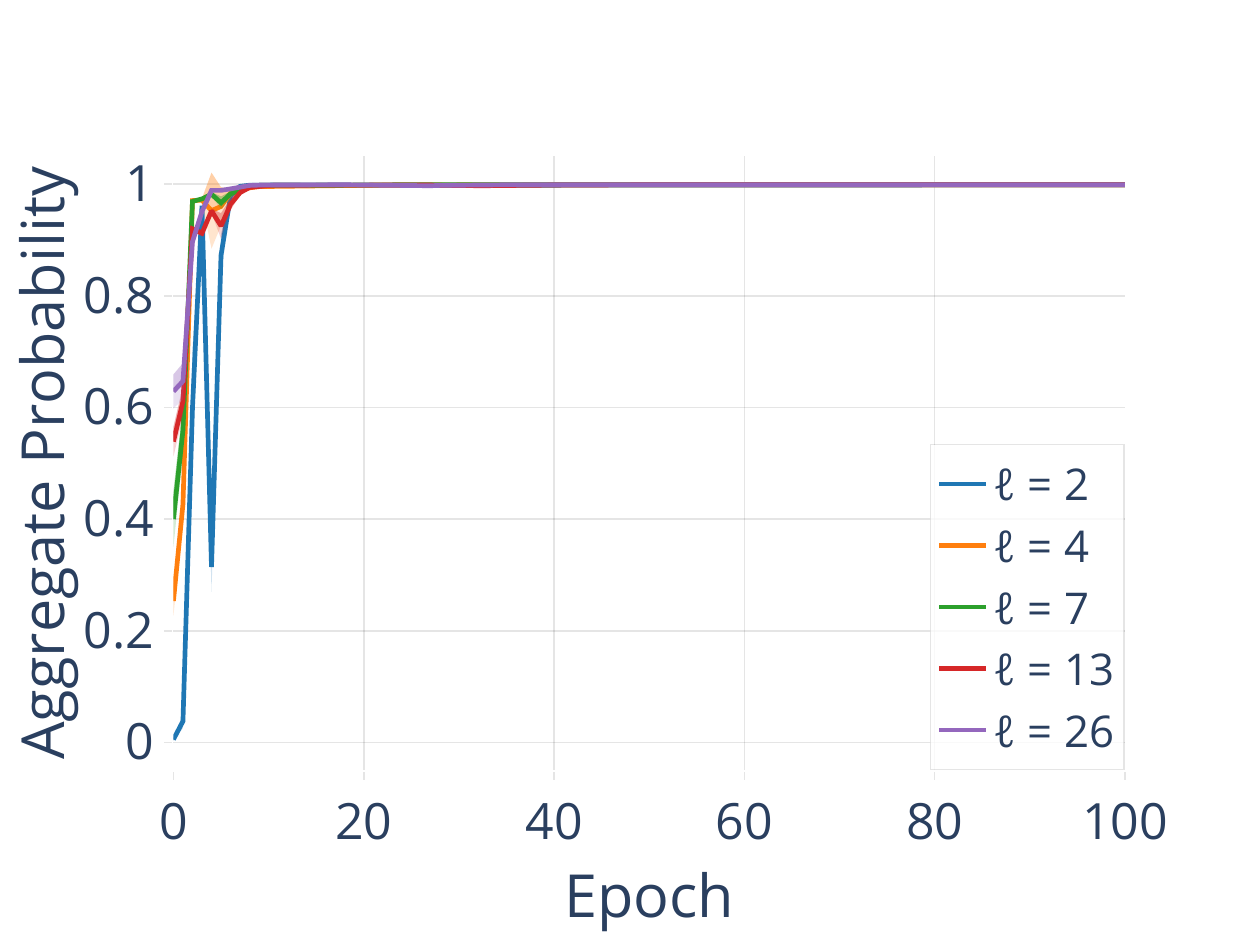}
    }
    \subfloat[Phi-2.7B, Aggregate Prob.]{
        \includegraphics[width=0.32\linewidth]{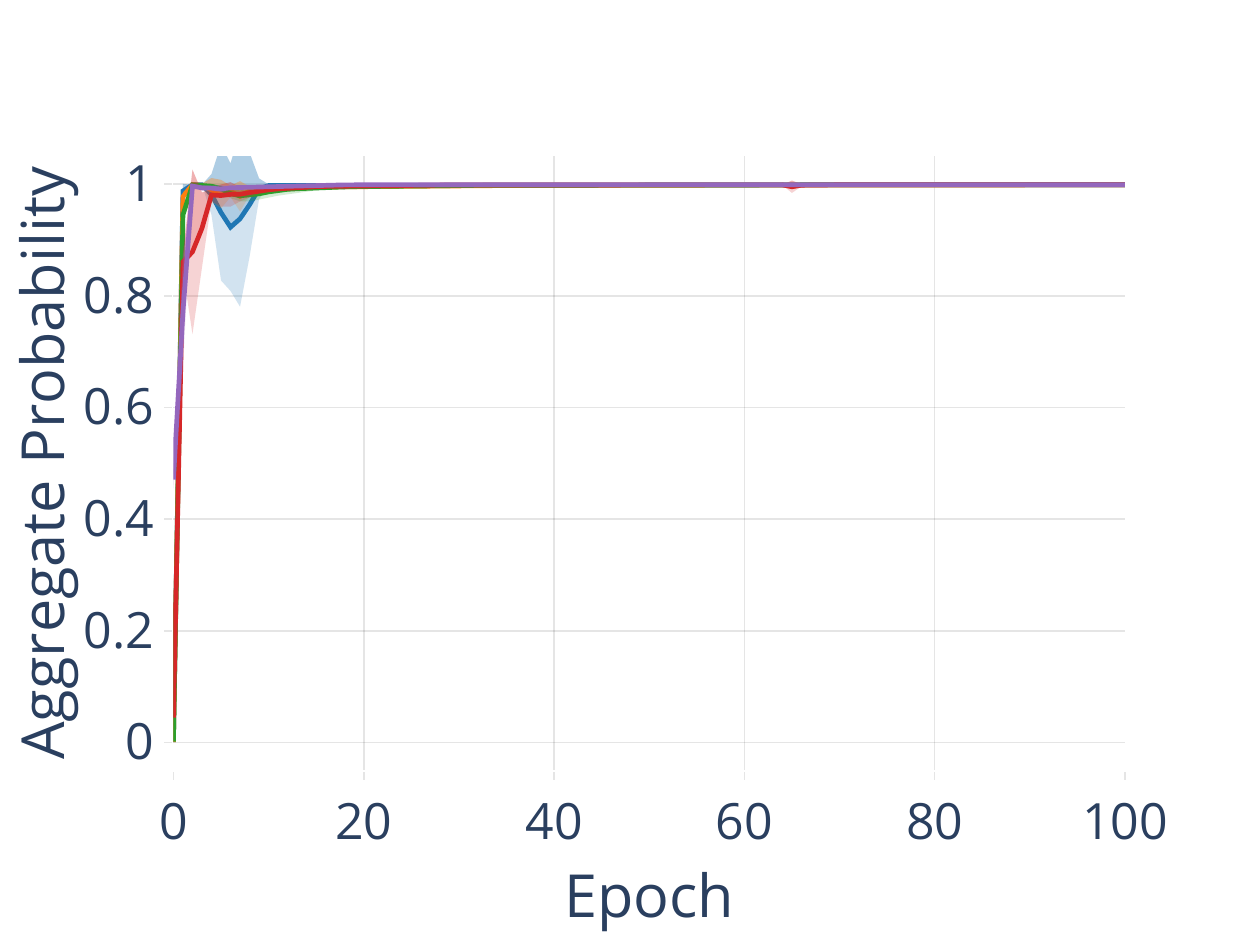}
    }
    \subfloat[Llama2-13B, Aggregate Prob.]{
        \includegraphics[width=0.32\linewidth]{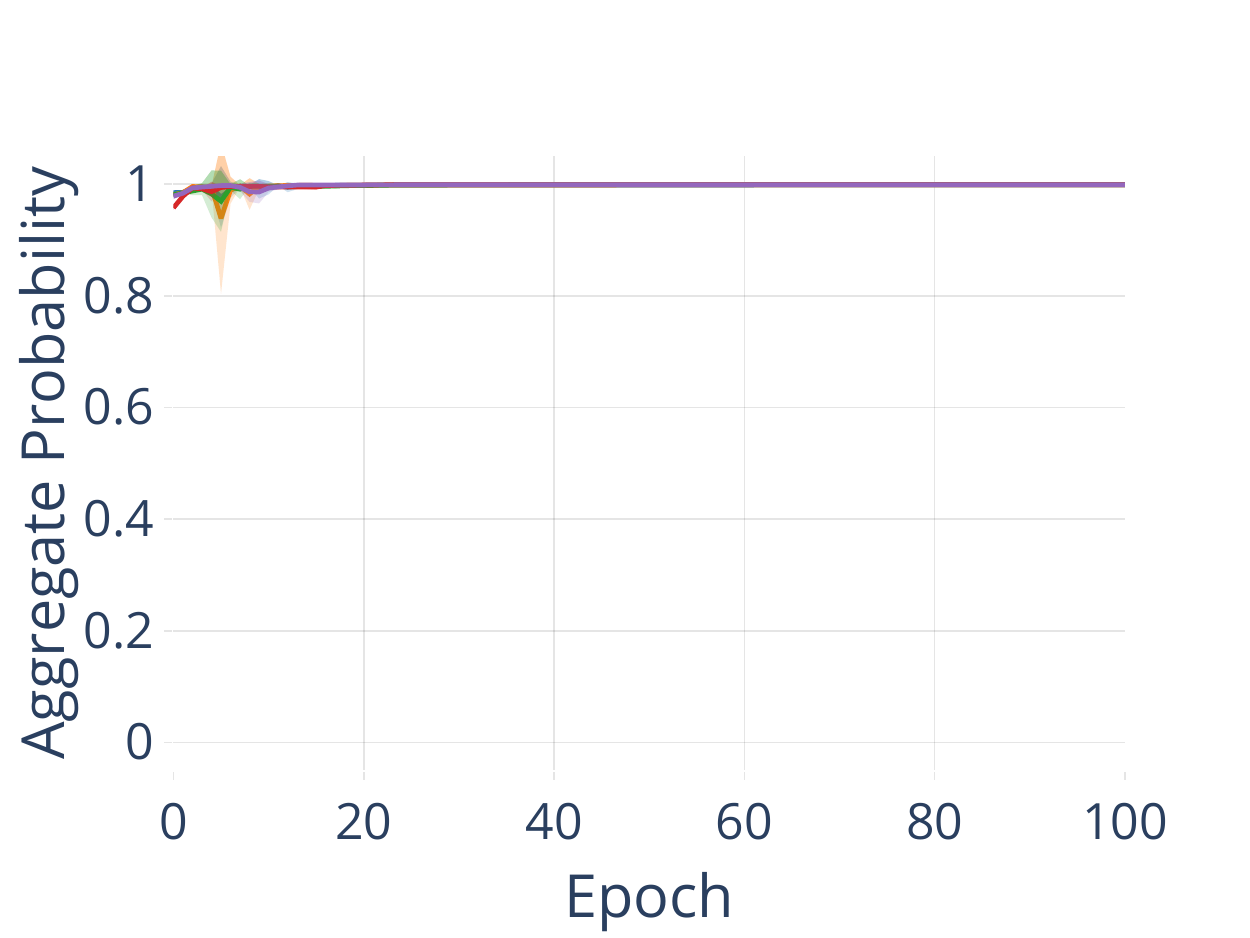}
        \label{fig:cdf_llama2_13b_alphabet_size}
    }
    \\
    \subfloat[Pythia-1B, Entropy]{
        \includegraphics[width=0.32\linewidth]{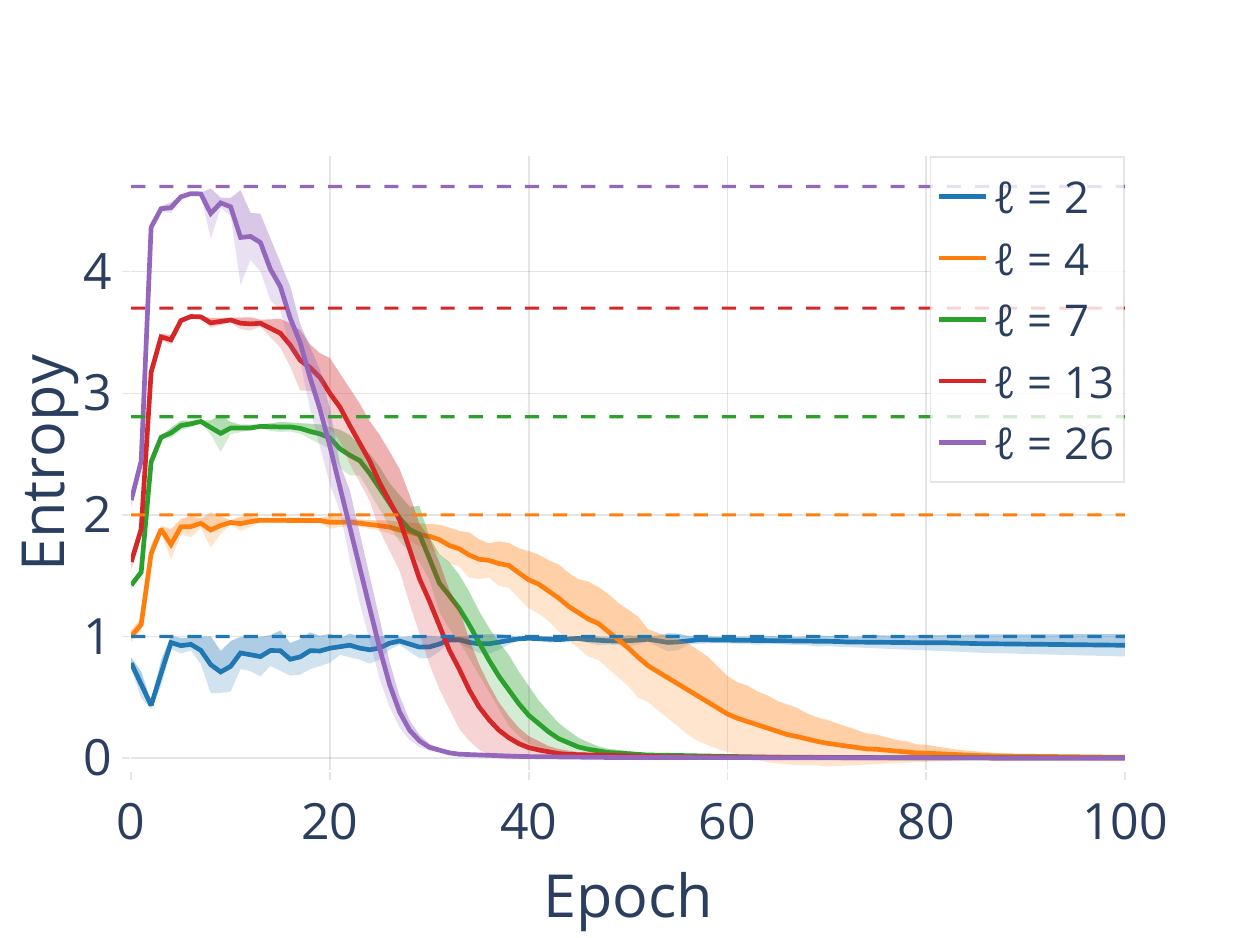}
    }
    \subfloat[Phi-2.7B, Entropy]{
        \includegraphics[width=0.32\linewidth]{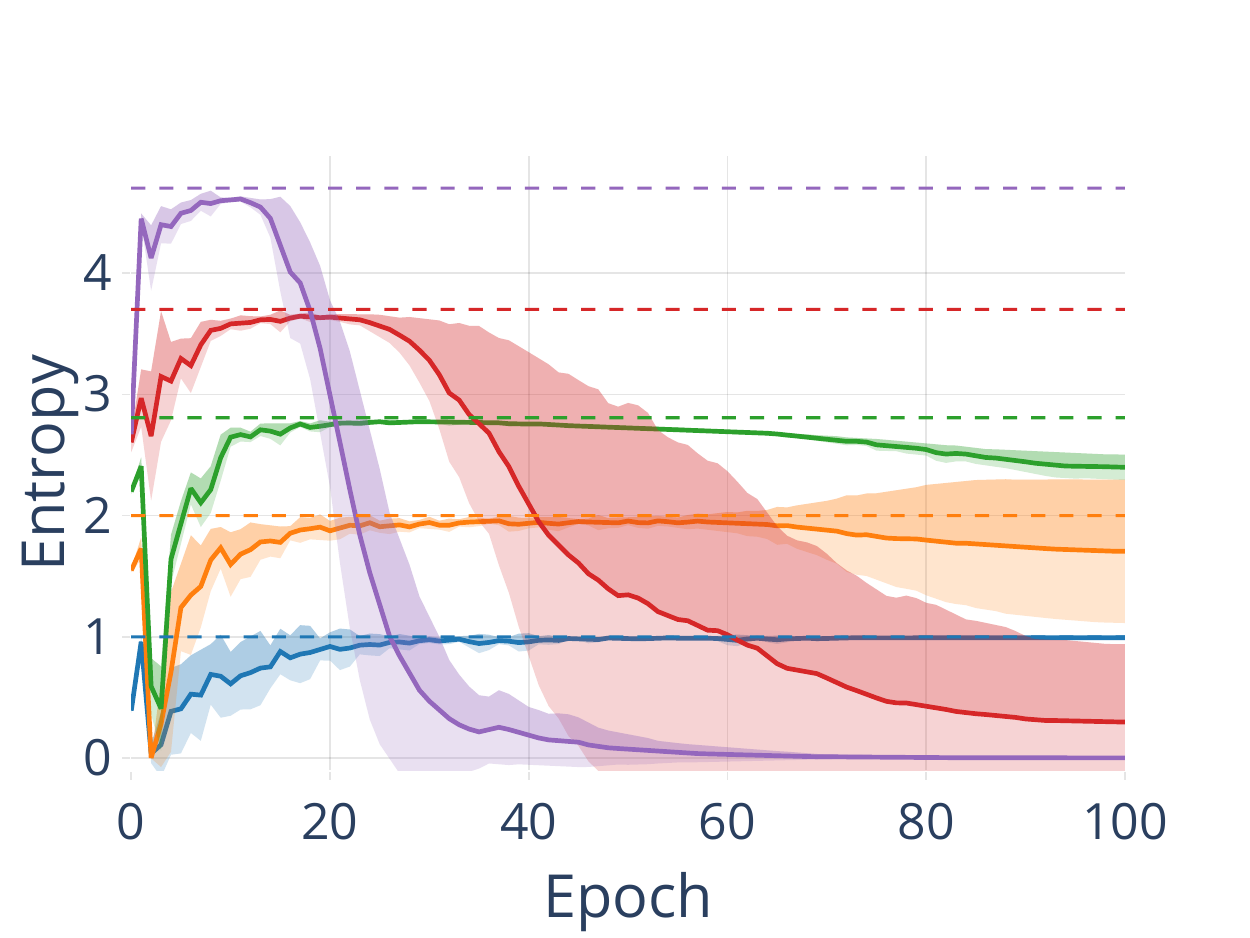}
    }
    \subfloat[Llama2-13B, Entropy]{
        \includegraphics[width=0.32\linewidth]{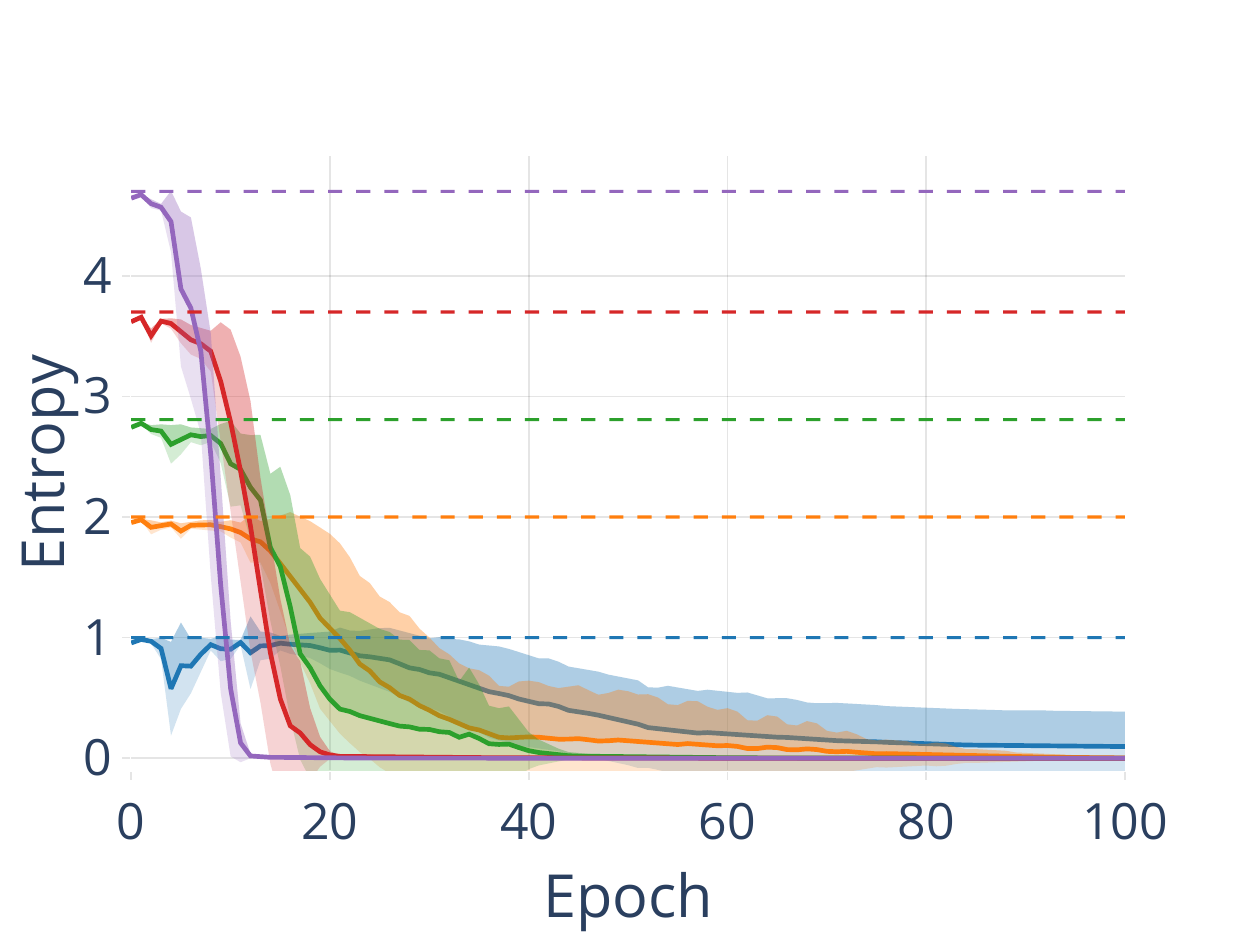}
    }
\caption{\capthead{Aggregate probability mass and entropy for different $\ell$.}{$n = 1024$}
i) Plots on the top show the probability mass that $\gM$ assigns to tokens in $A$. %
In all cases, models quickly learn to allocate the maximum possible probability mass to the tokens within the alphabet $A$, \ie~they only predict tokens from $A$ after a few training epochs.
ii) We show the average entropy of the probability distribution of model $\gM$ over $A$. %
The entropy initially rises to its maximum value, before decreasing to 0.
The maximum attainable entropy (for different $\ell$) is shown with dashed lines.
}
\vspace{-1mm}
\label{fig:cdf_entropy_alphabet_size}
\end{figure}

\textbf{Understanding the transition into the \MemPhase:}
 We now take a closer look at the token-level probability distributions produced by the models.
Figure~\ref{fig:cdf_entropy_alphabet_size}, upper row, shows the aggregate probability mass that models assign to the tokens inside the alphabet $A$, i.e., $\sum_{t\in A}P_\gM (s_i=t\mid s_{[1,i-1]})$, averaged over all positions $i$.
Analogously to the quick initial increase in accuracy (Figure~\ref{fig:accuracy_loss_alphabet_size}), we see that models quickly learn to assign all the probability mass to tokens within the alphabet $A$ and separate those from the whole vocabulary of tokens $V$. 

We also compute the token-level entropy over the course of training; i.e., the entropy of $P_\gM (s_i=t\mid s_{[1,i-1]})$ for $t\in A$, averaged over $i$.
The results in Figure~\ref{fig:cdf_entropy_alphabet_size}, lower row, show a sharp increase in entropy to the maximum possible value in the initial stages of training, that coincides with the rise in aggregate probability mass and the initial rise of the accuracy curves.
After the entropy peaks (at around epoch 8 for Pythia-1B and Phi-2.7B), it drops to 0, matching the second increase of the accuracy values.

Thus, in the initial \GuessPhase, the models are learning which tokens are in $A$ and separate those from $V$ (rise in aggregate probability).
In that phase, they do not know \textit{which specific tokens} to predict at each position in the string and thus \textit{guess} tokens from $A$ randomly (high entropy).
In the subsequent \MemPhase, the models actually start to memorise the specific tokens at each position (decrease in entropy) and become more accurate.
In Appendix~\ref{app:dynamics_additional_results} (Figures~\ref{fig:loss_alphabet_size_all},~\ref{fig:accuracy_alphabet_size_all},~\ref{fig:cum-prob_alphabet_size_all}, and~\ref{fig:entropy_alphabet_size_all}) we show that other models exhibit the same two-phase dynamics, and in Figures~\ref{fig:kld_alphabet_size_all} and~\ref{fig:kld_entropy_level_all} that during the \GuessPhase, $\gM$ actually approximates the distribution over $A$ (KL-Divergence).

It is worth noting that Llama2-13B assigns a total probability mass of $1$ to the tokens in $A$ almost immediately (Figure~\ref{fig:cdf_llama2_13b_alphabet_size}) and that its accuracy exhibits only a single ascend phase (Figure~\ref{fig:accuracy_llama2-13b_alphabet_size}).
Furthermore, its initial accuracy values before any training match the random guess baseline that the other models only reach after completing the \GuessPhase.
We investigate this phenomenon further in Appendix~\ref{app:dynamics_in_context} and show in Figure~\ref{fig:dynamics_in_context} that --- in contrast to the other models --- Llama2 models exhibit strong in-context learning abilities that enable them to infer the alphabet distribution $P_A$ after about only $100$ tokens of context.
These models effectively shorten the \GuessPhase to zero. %
However, when using non-Latin alphabets (see Appendix~\ref{app:dynamics_non_latin}, Figure~\ref{fig:non_latin_alphabet_all}), Llama2-13B also exhibits a \GuessPhase, indicating that its ability to learn the distribution from the context is limited.

\textbf{Memorisation order:}
We also analyse the order in which models memorise the tokens in the string.
In Appendix~\ref{app:memorization_order}, we show that tokens are memorised in random order, \ie~that there is no connection between the position of a token in the string, and the epoch at which it is memorised.
This observation is based on Spearman rank correlation values between the position of a token in the string and the epoch at which it is memorised, which are between $-0.1$ and $0.1$ in most cases, with moderate position correlation only for some untrained models on higher entropy strings.

\textbf{Implications for studying and quantifying memorisation:} Our discovery of the two phases in memorisation dynamics underscores the importance of our experimental setup, which  enables us to distinguish between token recollection due to in-context learning (\GuessPhase) and memorisation (\MemPhase). Our findings also call for fundamentally rethinking existing approaches to quantify memorisation~\citep{carlini2022quantifying}. 
On one hand, current measures risk \emph{overestimating} the degree of memorisation by not discounting for token recollection due to in-context learning (guessing) --  see Figures~\ref{fig:unique_substrings_a2_all} and~\ref{fig:unique_substrings_a26_all} in Appendix~\ref{app:dynamics_string_length}. 
On the other hand, the measures risk \emph{underestimating} the degree of memorisation by focusing on the recollection of contiguous token sequences, while tokens are memorized in random order -- see Figure~\ref{fig:string_mem_measure} in Appendix~\ref{app:string_measure_underestimation}. To be robust, future measures of memorisation need to account for these dynamics.

\section{Q1: Are some strings easier to memorise than others?}
\label{sec:memorability}

\begin{figure}[t]
    \centering
    \subfloat[Pythia-1B]{
        \includegraphics[width=0.32\linewidth]{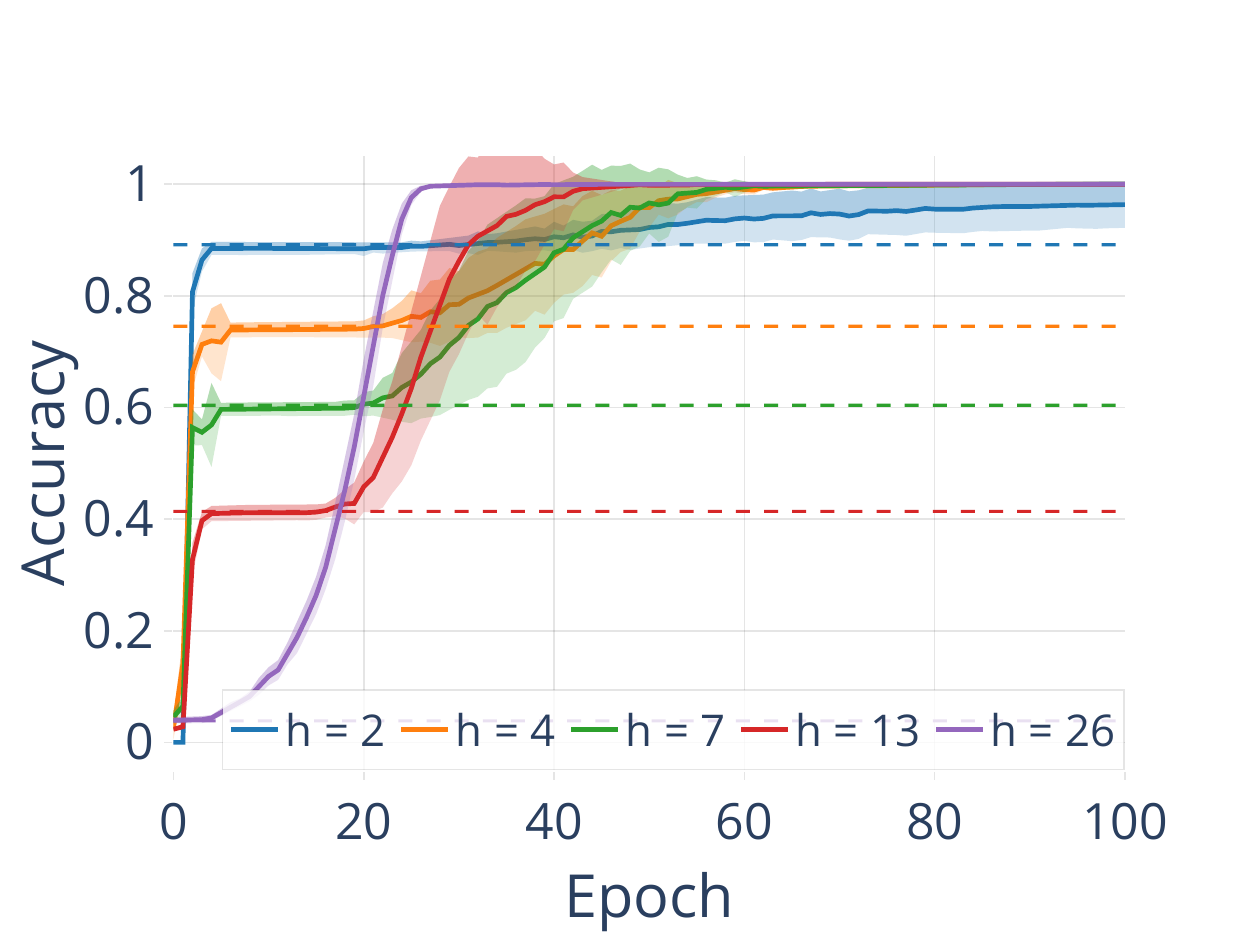}
    }
    \subfloat[Phi-2.7B]{
        \includegraphics[width=0.32\linewidth]{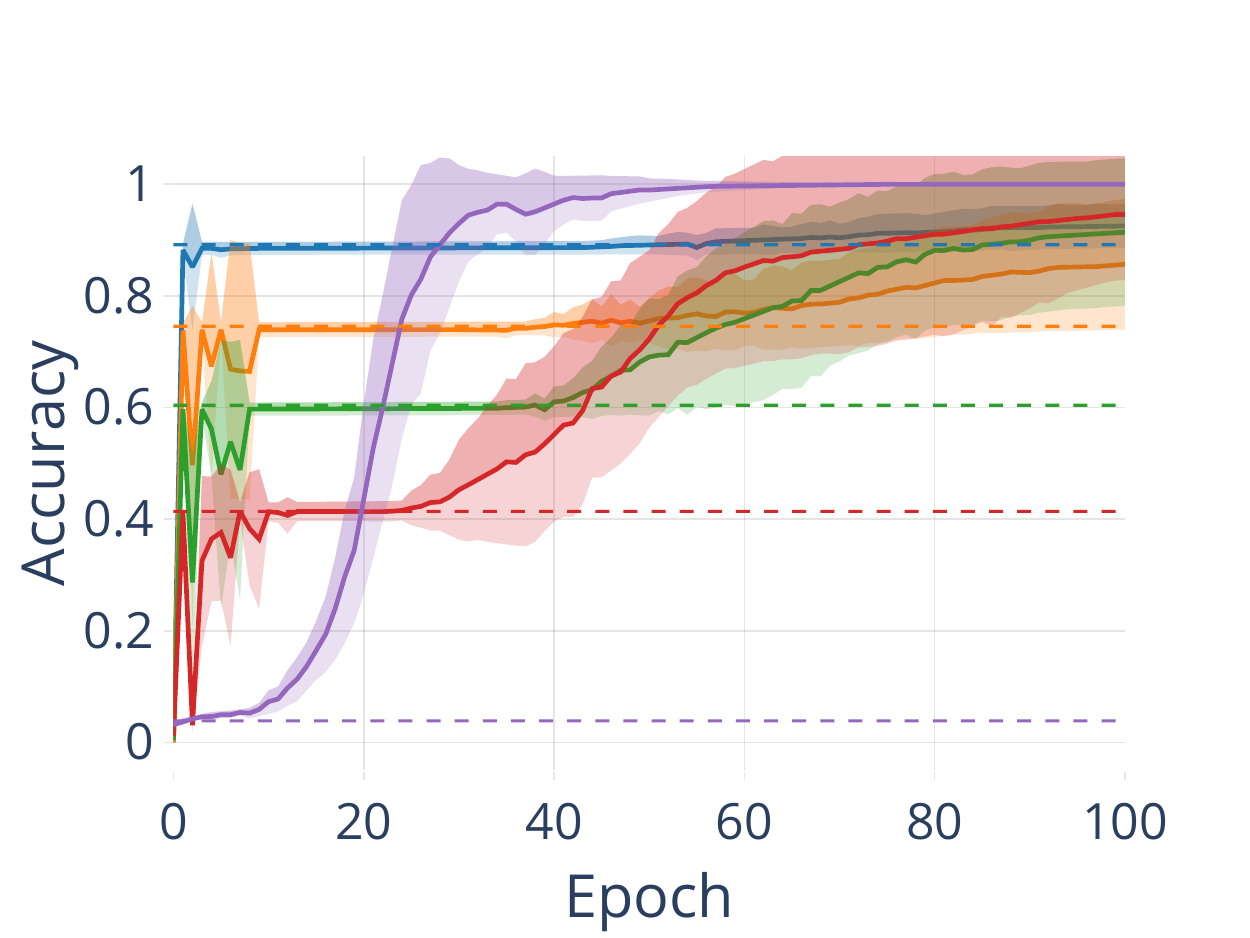}
    }
    \subfloat[Llama2-13B]{
        \includegraphics[width=0.32\linewidth]{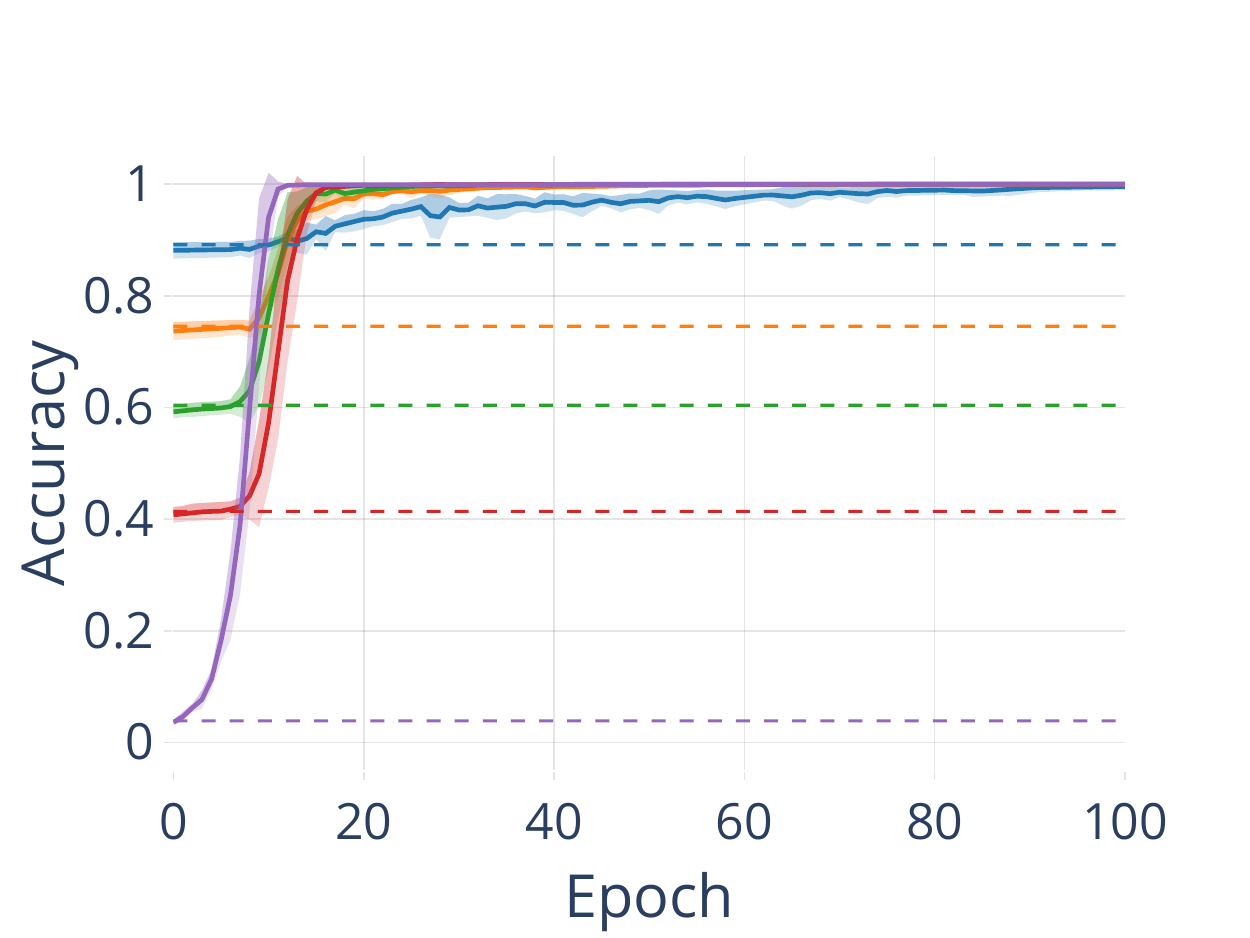}
    }
\caption{\capthead{Recollection accuracy for different entropy levels $h$.}{$n = 1024$}
Analogously to strings with different $\ell$, strings with lower $h$ are easier to guess, but harder to memorise.
Dashed lines indicate the performance of a random guess, equivalent to always guessing ``a''.
}
\vspace{-1mm}
\label{fig:accuracy_loss_entropy_level}
\end{figure}

In the previous section, we saw that during repeated exposure to the same random string, models undergo two phases.
Figure~\ref{fig:accuracy_loss_alphabet_size} shows that for the same model $\gM$, the \GuessPhase has the same length for different $\ell$, but that the length of the \MemPhase varies significantly, depending on $\ell$.
Even though strings with smaller $\ell$ end the \GuessPhase with higher accuracy values, they take longer to memorize in the subsequent \MemPhase, and are eventually ``overtaken'' by strings with larger $\ell$.
For all models, the strings with $\ell = 26$ and $\ell = 13$ are memorised first, \ie~reach accuracy $1$, whereas the $\ell = 2$ and $\ell = 4$ strings are memorised last (for Llama2-13b) or not at all within 100 epochs (Pythia-1B, Phi-2.7B).
The question therefore is: What property makes the random strings harder or easier to memorise?

\textbf{Effect of entropy on memorisation:}
The strings we use in Section~\ref{sec:phases} have different alphabet sizes $\ell \in \{2, 4, 7, 13, 26\}$, but they also have different levels of entropy $H_\ell$, with $H_\ell < H_{\ell'}$ for $\ell < \ell'$.
To test whether $\ell$ or $H_\ell$ is responsible for the differences in memorisation dynamics, we create strings with the same $\ell$, but with different entropy levels.
Specifically, we create random strings with an $\ell = 26$-letter alphabet $A$ with uniform $P_A$, except for the first letter (``a''), which is oversampled to match the entropy of the previous strings ({\ie}, $H_2, H_4, H_7, H_{13}, H_{26}$)\footnote{The more we oversample a letter in $A$, relative to the other letters, the lower the entropy of the string.}.

Accuracy curves for repeated exposure to such strings for different models are shown in Figure~\ref{fig:accuracy_loss_entropy_level}.
We see strikingly similar patterns to those shown in Figure~\ref{fig:accuracy_loss_alphabet_size}, with lower entropy strings achieving higher accuracy during the \GuessPhase, but subsequently being memorised more slowly than higher entropy strings.
These results indicate that it is the entropy of the probability distribution of the tokens that affects the memorability of random strings.
In Appendix~\ref{app:dynamics_additional_results} (Figures~\ref{fig:loss_entropy_level_all},~\ref{fig:accuracy_entropy_level_all},~\ref{fig:cum-prob_entropy_level_all} and~\ref{fig:entropy_entropy_level_all}) we show that the same observations hold for additional models.
For the rest of the paper we will only consider strings with different $\ell$ and uniform $P_A$, keeping in mind that the $H_\ell$ of the strings is the important part.

\textbf{Other factors affecting memorability}

\textit{Model size:}
In addition to entropy, we find that the \emph{size of the model} affects memorisation, and show in Appendix~\ref{app:dynamics_additional_results}, Figures~\ref{fig:accuracy_alphabet_size_all} and~\ref{fig:accuracy_entropy_level_all} that larger models within the same family tend to memorise strings faster.
This observation is congruent with findings by other work on memorisation~\cite{carlini2021extracting, carlini2022quantifying, biderman2023emergent}.

\textit{String length and structure:}
Further, we find in Appendix~\ref{app:dynamics_string_length} that the length of the random string plays a role (Figures~\ref{fig:string_length_a2_all},~\ref{fig:string_length_a26_all}), with longer strings being harder to memorise than shorter ones, but only if all tokens are sampled independently.
If we create longer strings by repeating shorter random strings (Figures~\ref{fig:unique_substrings_a2_all},~\ref{fig:unique_substrings_a26_all}), only the length of the unique base string matters, since the models can predict the remaining tokens without memorisation from the context.
Additionally, memorising a random string as one long piece or in multiple shorter partitions in the same batch does not affect memorisation speed (Figures~\ref{fig:partitions_a2_all},~\ref{fig:partitions_a26_all}), which indicates that a main factor for memorisation difficulty --- other than entropy --- is the total number of independently sampled tokens in the string.

\textit{Conditional entropy:}
In Appendix~\ref{app:conditional_probability_strings} we change the n-gram conditional entropy of strings, while keeping their conditional entropy fixed, and show that unconditional entropy affects the \GuessPhase, but that it does not significantly impact the length of the \MemPhase.

\textit{Alphabet type:}
In Appendix~\ref{app:dynamics_non_latin} we conduct ablations with non-Latin alphabets (Figure~\ref{fig:non_latin_alphabet_all}) and non-pretrained versions of the models shown earlier (Figure~\ref{fig:untrained_all}), and observe the same memorisation dynamics as before, although non-pretrained models memorise more slowly.

\textit{Real-world training setups:}
In practice, memorised strings appear embedded into a context of other data and not in isolation, \eg~an email address or phone number embedded inside other text.
To validate whether memorization follows the same patterns when it happens in the context of other data, we train models to memorise random strings under conditions that closely resemble real-world settings.
We present random strings to the model in the context of natural language data from wikitext~\cite{merity2016pointer}, in two different ways: 1) by presenting random strings to the model as elements inside larger batches of natural language strings, and 2) by embedding random strings inside longer natural language strings as substrings.
We show in Appendix~\ref{app:real_world_validation} that -- while random strings are memorized more slowly, the more natural language context there is -- the same dynamics as for memorization in isolation hold, and that we thus can expect our findings to generalize to practical training scenarios.

\textbf{Implications for risks and ways of memorising training data:} Our findings demonstrate that {\it not all strings are equally memorable} -- in fact, we establish that memorability of random strings is intrinsically related to their entropy. But, generalising our findings to natural language strings remains an open challenge as it is far from clear how one would estimate entropy of such strings. Intuitively, however, our findings imply that the less guessable a string is (higher entropy), the easier it is for an LLM to memorise it. Put differently, rather ironically, the strings (e.g., cryptographic secret keys) that are harder to guess and offer more security are the ones that are at greater risk of being memorised by LLMs. Finally, as the more guessable strings (lower entropy) are also more compressible, our findings rule out compression as a potential (latent) model for how strings are stored by LLMs during memorisation. A detailed exploration of memorisation techniques used by LLMs is beyond the scope of this work, but we take the first step towards exploring prefix associations used by LLMs to recollect tokens in the next Section.

\section{Q2: What information do models need to recall memorised tokens?}
\label{sec:recall_information}

We setup the following experiment:
Given a token at position $i$ in random string $s$ sampled from $P_A$ as described in Section~\ref{sec:preliminaries}, we split $i$'s full prefix $s_{[1,i-1]}$ into two parts:
a \emph{local prefix} of size $k$, \ie~$s_{[i - k, i - 1]}$, and the \emph{global context}, given by $s_{[1,i-k-1]}$.
In our analysis, we keep the local prefix fixed while modifying the global context.
Our goal is to study the role that the prefix and the global context play in the recollection accuracy of the model.

\begin{figure}%
    \centering
    \subfloat[Prefix perf. $\ell = 7$]{
        \includegraphics[width=0.24\linewidth]{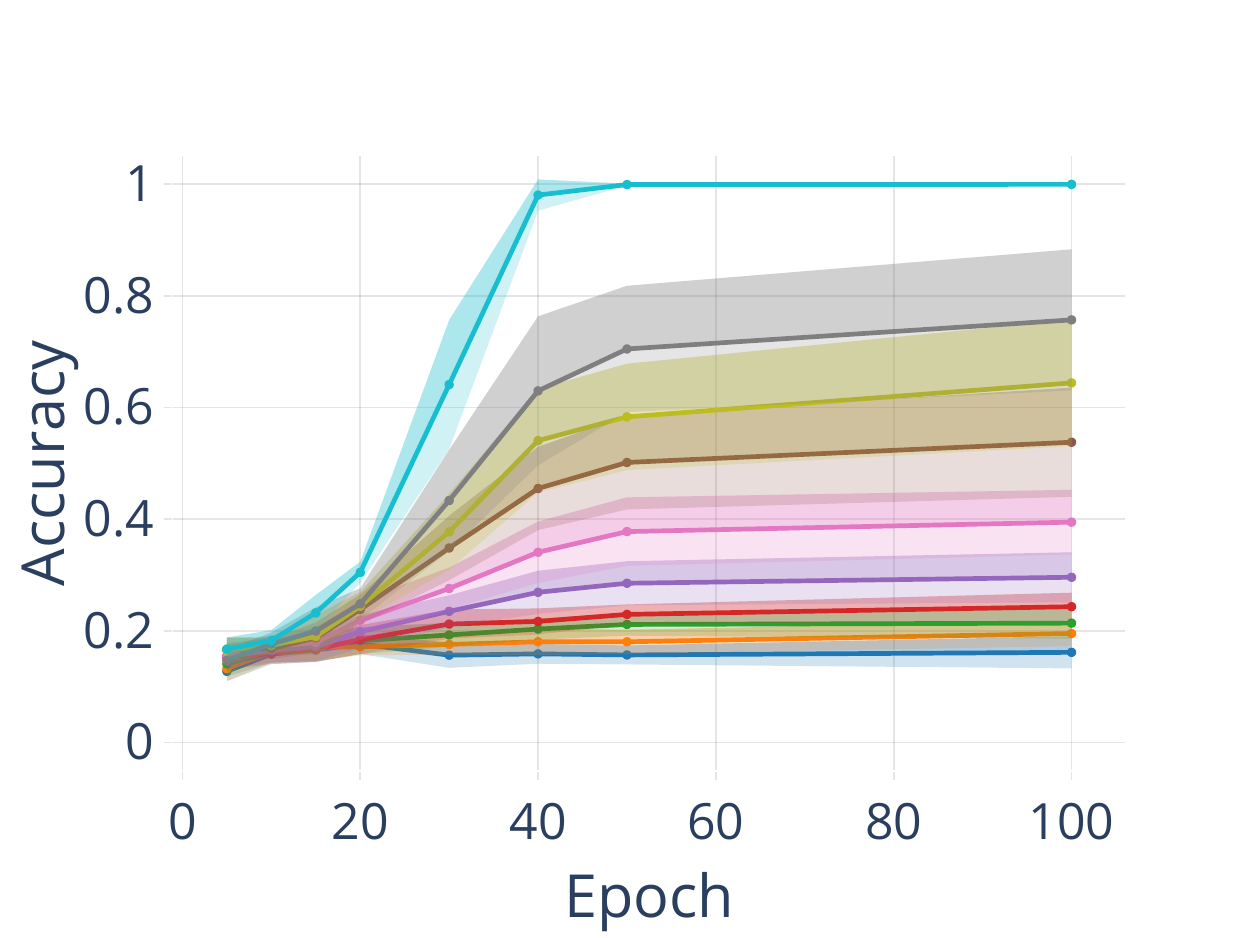}
        \label{fig:prefix_len_a-7}
    }
    \subfloat[Prefix perf. $\ell = 26$]{
        \includegraphics[width=0.24\linewidth]{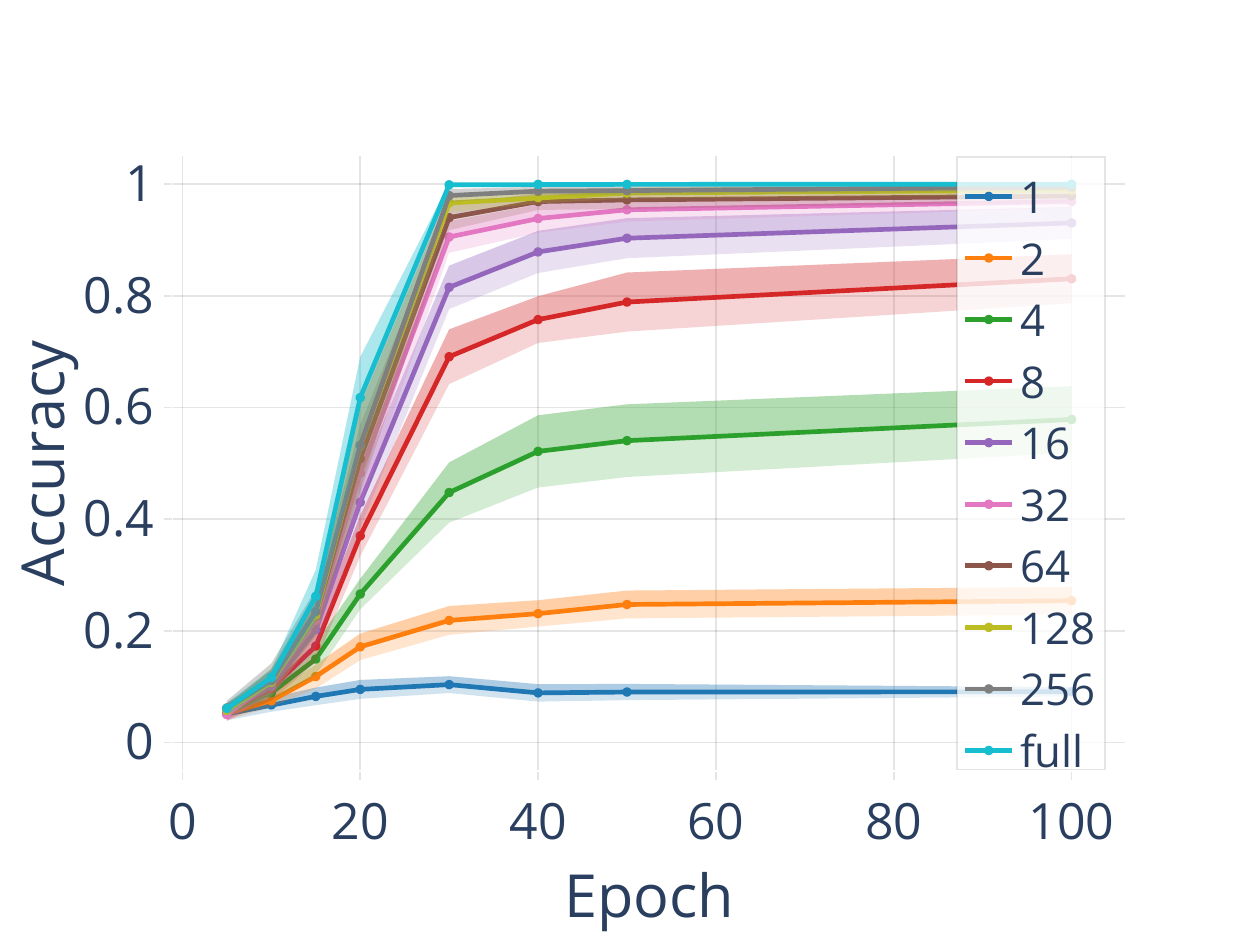}
        \label{fig:prefix_len_a-26}
    }
    \subfloat[GC distribution]{
        \includegraphics[width=0.24\linewidth]{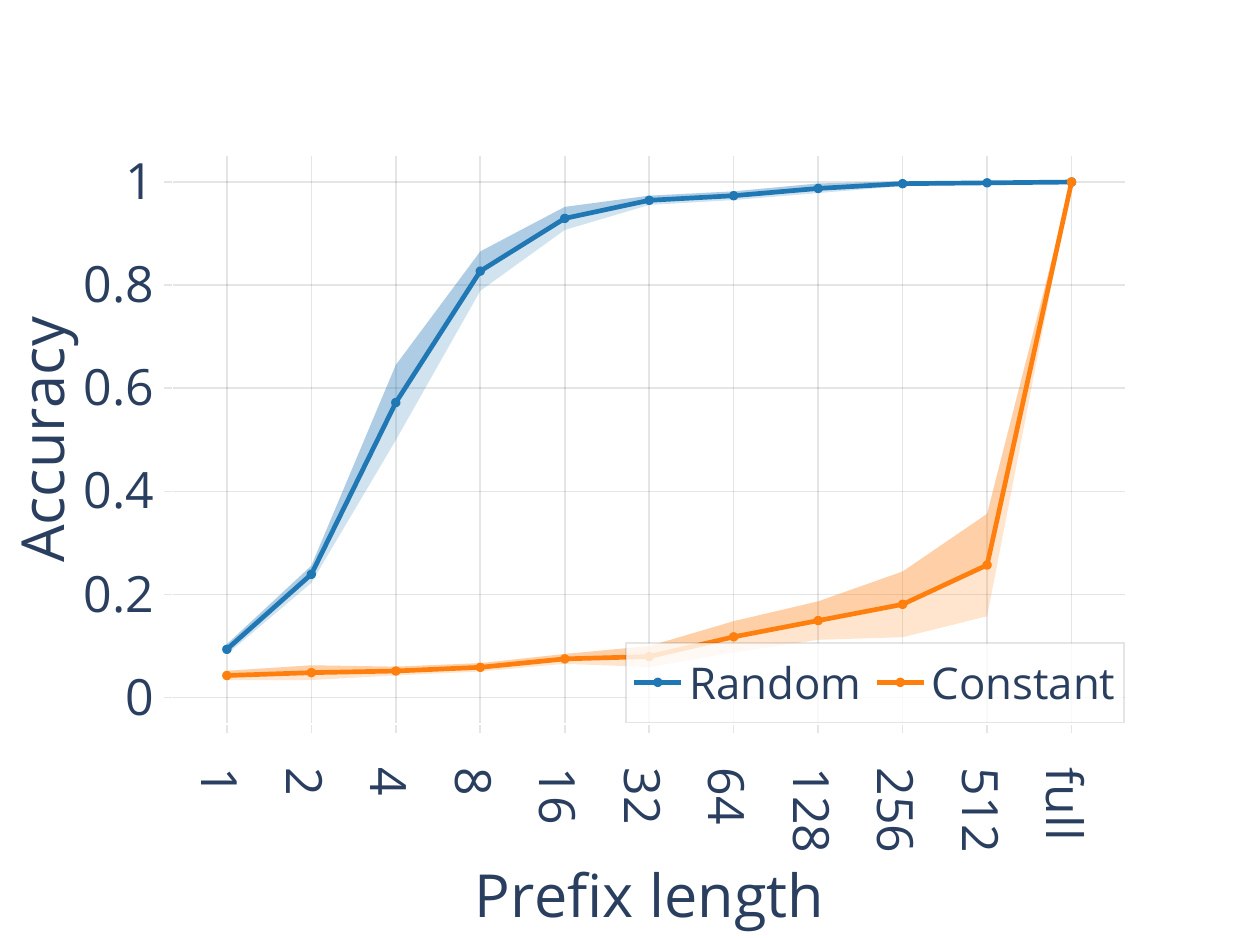}
        \label{fig:prefix_len_replacement_strategy}
    }
    \subfloat[Amount of GC]{
        \includegraphics[width=0.24\linewidth]{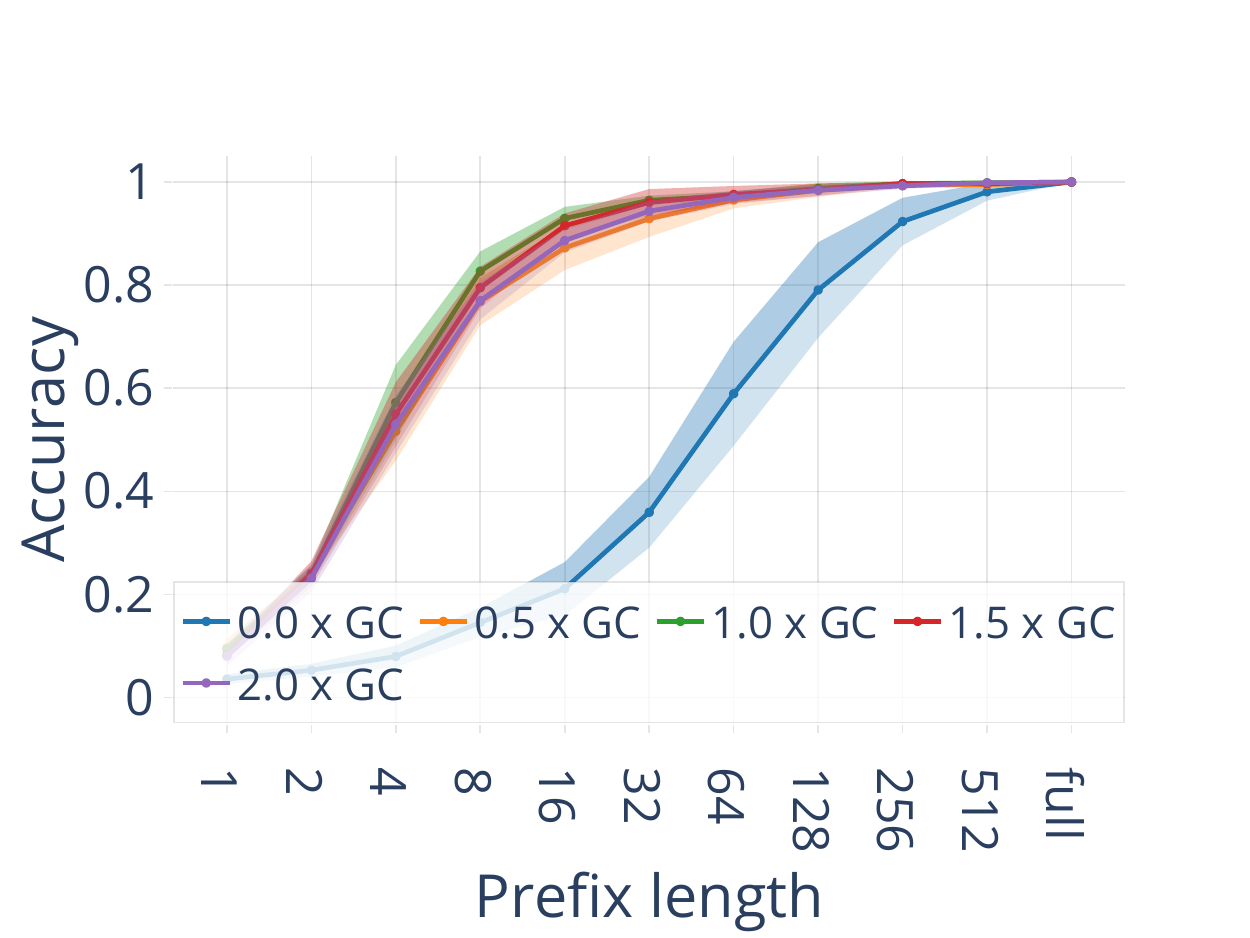}
        \label{fig:prefix_len_rel_pos}
    }
\caption{\capthead{Recollection accuracy for different prefix lengths and for changes in the global context (GC) during training.}{$n = 1024, \gM = \text{Pythia-1B}$}
(a) and (b) show what fraction of tokens can be recollected correctly with different prefix lengths, at different points during training.
In many cases, prefixes much shorter than the full string are sufficient to predict most of the tokens accurately.
(c) shows the performance of a randomly re-sampled vs a constant global context with only one repeated token, and (d) shows the impact of changing the size of the global context, where the numbers indicate multiples of the GC size.
}
\vspace{-2mm}
\label{fig:epoch_prefix_len}
\end{figure}

\textbf{The role of local prefixes:}
To determine how much information the model needs to recall tokens, we test the accuracy of the model for different prefix lengths.
For this, we keep the local prefix fixed, while changing the global context using a {\random} replacement policy.
\random~replaces every token of the global context with one sampled randomly from $P_A$.
For each position $i$, we sample $10$ global context replacements and count $i$ as predicted correctly if $s_i$ is predicted most frequently by the model (using greedy decoding) among the samples.

We show the recollection accuracy for different prefix lengths $k=1,2,4\ldots , 1024$ over the first $100$ epochs of training for $\ell=7$ in Figure~\ref{fig:prefix_len_a-7} and for $\ell=26$ in Figure~\ref{fig:prefix_len_a-26}.
We observe that in the initial \GuessPhase (at epoch 5 and 10), all prefix lengths  achieve very low accuracy.
Starting from epoch 15 (after transitioning into the \MemPhase), the accuracy starts to increase substantially, even for short prefixes.
Especially at epoch $30$ and later, for the larger $\ell$, the accuracy of short prefixes (less than $5 - 10$\% of the total string length) is close $100$\%. %
As $\ell$ gets smaller, shorter prefixes perform worse, relative to the full prefix.
Overall, small local prefixes --- much shorter than the entire string --- are very effective at correctly recalling tokens.
Moreover, as memorisation progresses, the same level of accuracy can be achieved with shorter prefixes.
In Appendix~\ref{app:local_prefix_results} we show results for additional models and values of $\ell$. %
There, we observe that for untrained models, short prefixes are not sufficient for recalling tokens, which suggests that during pretraining models might acquire a recency bias.

\textbf{The role of global context:}
To further investigate the role of global context, we also consider --  {\constant} replacement policy, 
which
replaces every token in the global context with a random %
token from $A$.
We also change the size of the global context by sampling more or less tokens than the original string. %

\textit{Importance of the replacement policy:}
Figure~\ref{fig:prefix_len_replacement_strategy} shows the accuracy of  prefixes of different lengths with global contexts being generated using {\random} and {\constant} policies.
Our results show that when trying to recollect tokens from memorised strings, the global context is important.
Maintaining the original distribution $P_A$ of the tokens in the global context ({\random}), even though the substring itself changes, leads to high recollection accuracy.
Biasing ({\constant}) the global context leads the model to assume a wrong distribution, lowering its recollection accuracy.

\textit{Importance of the length of global context:}
Is it enough to maintain $P_A$ when generating the global context, or does the length of the context also matter?
To determine this, we increase (by $50\%$ or $100\%$) or decrease (by $50\%$ or $100\%$) the number of tokens in the global context while applying the {\random} replacement strategy.
We show in Figure~\ref{fig:prefix_len_rel_pos} that adding or removing tokens from the global context does not impact the recollection accuracy for fixed local prefix length, as long as some global context is preserved.
When fully eliminating the global context (0 x GC), recollection accuracy drops drastically.
Thus, the amount of global context, and the position of $i$ within $s$ is not important, as long as some global context information, and the local prefix, are present.

In summary, while only keeping the tokens in the local prefix constant is mostly sufficient to recollect the target token, this only holds when the token distribution in the global context is preserved.
We provide additional details in Appendix~\ref{app:local_prefix_setup} and show in Appendix~\ref{app:global_context_results} results for additional models in Figure~\ref{fig:replacement_strategy_all} for replacement strategy, and in Figure~\ref{fig:size_change_all} for changes in global context size, with the same takeaways.

\textit{Models with absolute position encodings:}
We repeat the above experiments for models with older architectures that use absolute position encoding, such as GPT-2 and OPT.  The results are shown in Figures~\ref{fig:replacement_strategy_all} and~
\ref{fig:size_change_all} in Appendix~\ref{app:abs_pos_prefix_mappings}. Those results show that these models are relying much more on position-based memorisation and therefore the global context does not matter as long as it does not affect the position of the tokens in the string.

\textbf{Implications for how LLMs memorise strings:} Our investigation here reveals the subtle but important differences in the role played by the prefix tokens close to (local prefix) and far away (global context) from the token being recollected from memory. Our finding that the greater the degree of memorisation, the smaller the length of local prefix needed for recollection, suggests a potential (proxy) measure to estimate and quantify memorisation of a string. More importantly, our findings offer a starting point for a potential explanation for why higher entropy strings are more easily memorised -- in higher entropy strings, the same length of local prefix is more predictive of the next token. However, we are far from having a comprehensive theory explaining all our experiments. For example, we observed that creating strings where small prefixes are more predictive alone does not make them more memorable (Appendix~\ref{app:conditional_probability_strings}),
and introducing regularities into strings can also lead to interference with in-context learning (Appendix~\ref{app:dynamics_string_length}).

\section{Q3: How do models behave when sequentially memorising random strings?}
\label{sec:repeated_memorization}

\begin{figure}%
    \centering
    \subfloat[$\ell = 26, K = 16$]{
        \includegraphics[width=0.49\linewidth]{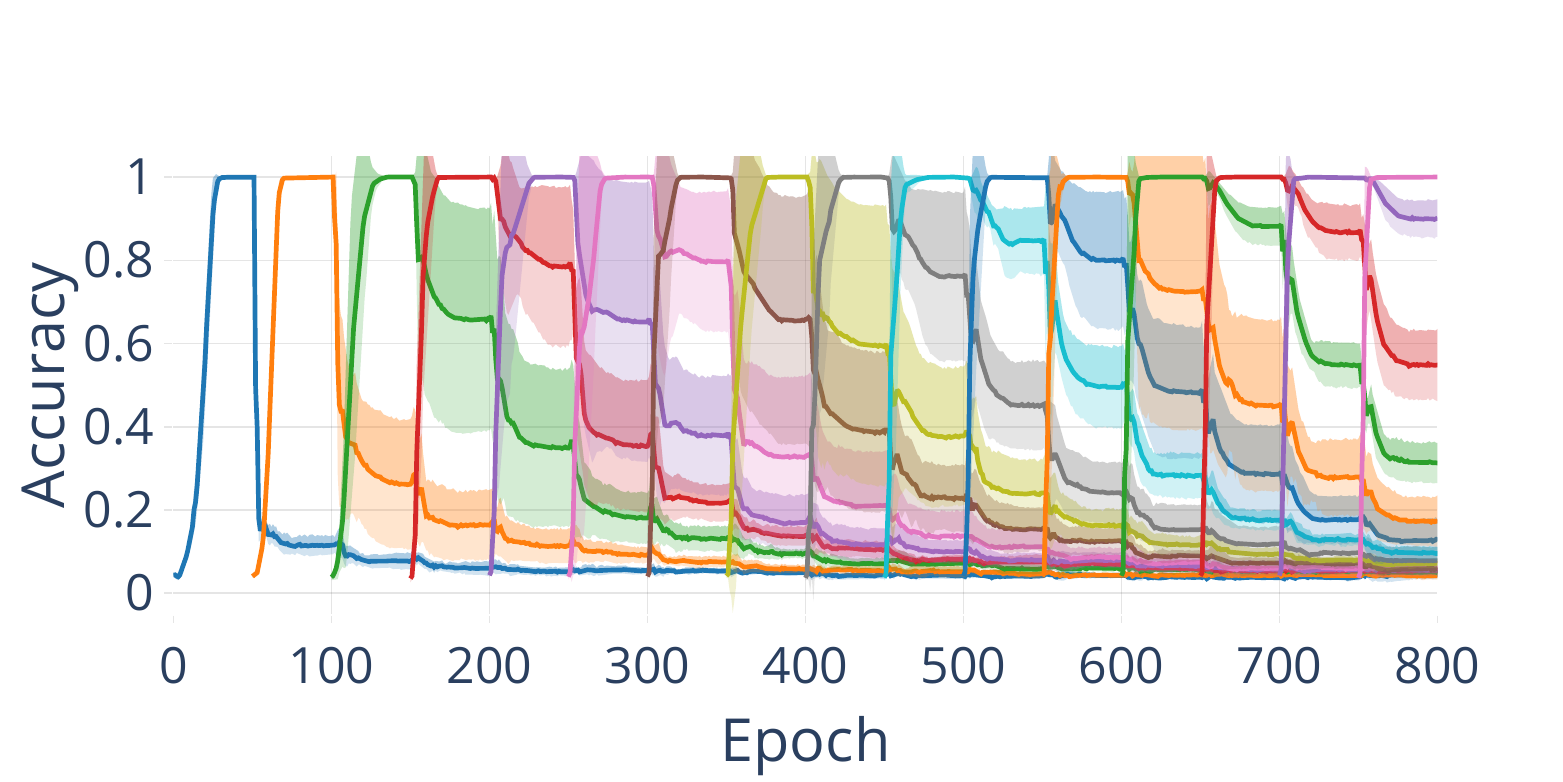}
    }
    \subfloat[$\ell = 2, K = 32$]{
        \includegraphics[width=0.49\linewidth]{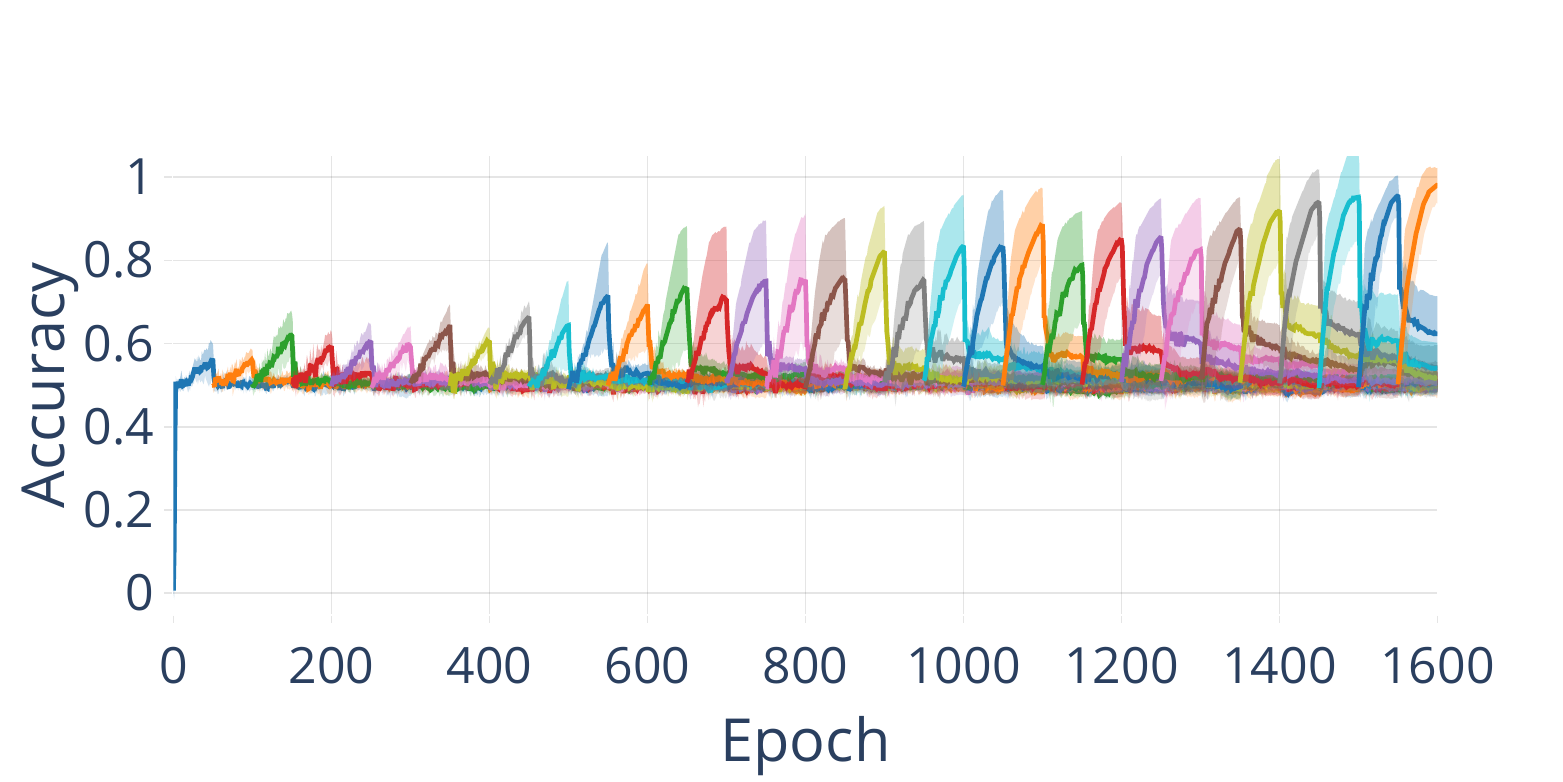}
    }
\caption{\capthead{Accuracy on different strings during sequential memorisation.}{$n = 1024, \gM = \text{Pythia-1B}$}
Each curve denotes a new string.
As the model memorises new strings, it forgets old ones, shown by the drop in accuracy after the first 50 epochs per string.
}
\label{fig:repeated_mem}
\end{figure}

In order to address this question we perform the following experiment: given an alphabet $A$ of size $\ell$ we generate $K \in \{16, 32\}$ random strings from $A$: $\{s_1,s_2,\ldots , s_K\}$.
Then, we train model $\gM$  to sequentially memorise each of the $s_i$'s, for $50$ epochs per string.

Figure~\ref{fig:repeated_mem} shows the accuracy of a Pythia-1B model $\gM$ during sequential memorisation.
Accuracy for different strings is indicated by different colours, and we show the accuracy for each $s_i$ during the initial 50 epochs when $\gM$ trains on it, as well as during the following epochs, when $\gM$ trains on other strings.
We make two key observations:
(i) As models memorise subsequent strings, they forget the strings they previously memorised, as seen by the decreasing accuracy curves after the first 50 epochs in Figure~\ref{fig:repeated_mem}.
However, forgetting happens more slowly for strings memorised later in the sequence.
(ii) With sequential memorisation, models become better at memorising new strings, since strings memorised later in the sequence are memorised faster. %
For instance, without previous memorisation, Pythia-1B takes $29$ epochs (iterations) to achieve $99\%$ accuracy in memorising strings with $\ell = 26$, 
but after memorising $15$ other strings, it can achieve the same accuracy within $8$ epochs, i.e., nearly four times fewer iterations.
This speedup happens during the \MemPhase.
We show similar results for more models in Appendix~\ref{app:repeated_memorization}.

\textbf{Implications for conditioning models to memorise or forget:}
Our findings here show that we can both trigger forgetting of previously-memorised information, and make models better at memorising new information.
Specifically, it suggests that one way of ensuring that a trained model forgets cryptographic keys of a certain format might be to memorise new randomly generated keys with a similar format. 
On the other hand, it is worrisome that models can be primed to better memorise specific types of strings, which raises the spectre of new types of memorisation risks and attacks. 
We are not aware of any prior works that identified such risks.

Additionally, we considered how memorisation affects the model's performance on other tasks. For this, we experimented with how memorising random strings impacts the model’s loss on the wikitext testset~\cite{merity2016pointer}.
Our results in Appendix~\ref{app:mem_perf_impact} show that memorising a single random string in isolation can negatively affect the model’s performance.
However, when memorizing random strings in the context of natural language data, models can both improve their natural language modelling abilities, while also memorizing the random strings. %

\section{Conclusions and Limitations}\label{sec:implications}

\textbf{Conclusions:}
In this paper, we study the phenomenon of memorisation at a foundational level.
We do so using random strings, which provide us with controlled ``laboratory'' conditions that ensure that our observations meet three important validity criteria:
1) \emph{Isolation from other memorized data}. %
2) \emph{Isolation from non-memorisation phenomena} such as in-context learning.
3) \emph{Targeted intervention on string properties without confounders.}
We make a number of intriguing observations, including that models exhibit a {\GuessPhase} and a {\MemPhase}, that strings with higher entropy are easier to memorise, that models can often recall memorised tokens using small subsets of the entire context, and that sequentially memorising strings changes the memorisation dynamics.

Our findings have significant implications for studies focusing on quantifying memorisation, understanding how memorisation works, and estimating privacy risks with memorisation. Furthermore, many of our empirical findings cannot be easily explained and the quest for a comprehensive explanatory theory of all our findings raises many open and challenging questions.

\textbf{Limitations:}
Our insights on memorisation heavily rely on random data, and it is possible that some of the observations might change for real-world data.
We conduct extensive validation experiments to ensure that our findings are robust, but there might be additional factors that impact memorisation behaviour in more complex scenarios.
We also focus on observing \emph{what} happens during memorisation, and leave it up to future work to develop a deeper understanding for \emph{why} models behave this way during memorisation, \eg~why memorising multiple strings in sequence makes memorisation faster.

\bibliographystyle{abbrvnat}
\bibliography{main}

\newpage
\appendix
\onecolumn

\section{Additional details on the experimental setup}
\label{app:setup}

\subsection{Technical details on the training setup}
\label{app:dynamics_setup}

\textbf{Models:}
In this paper, we use pretrained models of the \href{https://huggingface.co/EleutherAI/pythia-1b}{Pythia}~\citep{biderman2023pythia}, \href{https://huggingface.co/microsoft/phi-2}{Phi}~\citep{li2023textbooks2} and \href{https://huggingface.co/meta-llama/Llama-2-13b-hf}{Llama2}~\citep{touvron2023llama2} families. 
For the Pythia family, we use variants with 70M, 1B and 12B parameters, for the Phi family we use 1.3B and 2.7B parameter variants, and for Llama-2 we use 7B and 13B parameter variants.
We choose these models, since they represent popular, modern architectures, and span a wide spectrum of parameter counts (more than two orders of magnitude).

In addition to the above, we also use \href{https://huggingface.co/openai-community/gpt2}{GPT-2}~\citep{radford2019language} and \href{https://huggingface.co/facebook/opt-350m}{OPT}~\cite{zhang2022opt} models for some experiments, to study the effect of absolute position encodings.
In particular, we use GPT2-140M (GPT-2) and GPT2-1.5B (GPT-2-XL) parameter variants of GPT-2 and the OPT-350M model.
Pretrained versions for all models are publicly available on the \href{https://huggingface.co/docs/hub/models-the-hub}{Huggingface Model Hub}.

\textbf{Training:}
We train models to minimise the cross-entropy loss over string $s$.
We define the cross-entropy loss of a model $\gM$ on string $s$
as follows:
\begin{equation}\label{eq:cross-entropy}
\text{Loss}(\gM, s) = - \frac{1}{n}\sum_{i=1}^n \sum_{t \in V} \delta(s_i = t) \log P_{\gM}(s_i=t | s_{[1,i-1]} ).
\end{equation}

In most experiments on pretrained models we train models on random strings for 100 epochs (for single strings, each step is an epoch), with a linearly decaying learning rate schedule.
Untrained models memorise more slowly, so we train them for 300 epochs in most cases.
For Pythia-70M, Phi-1.3B and Phi-2.7B we use an initial learning rate of $5 * 10^{-5}$, for OPT-350M
$10^{-4}$, for GPT2-124M $5 * 10^{-4}$ and for all other models $10^{-5}$.
These learning rate values resulted in the fastest convergence during a grid search over values from $10^{-3}$ to $10^{-6}$.

\subsection{Examples of random strings used in the paper}
\label{app:data_examples}

\begin{table}[h]
    \centering
    \begin{tabular}{c|c}
        \hline
        \textbf{Alphabet and distribution} & \textbf{Tokens} \\
        \hline
        2 characters, uniform & bbabbabbababbabaaabbababaaaababb \\
        4 characters, uniform & cdbccbddbcaddbcabaccbcbcabaacadd \\
        7 characters, uniform & efceecffdeaggdebbbffddbdabaafaff \\
        13 characters, uniform & hleijdjkfibllfhcdcjjghdgbdaajakk \\
        26 characters, uniform & pwjqshtulrcxxlpegessmognchaatauv \\
        \hline
        26 characters, H2 & aaaaaaaaaagaaaaaaaaaaaaaaaaaaaaa \\ 
        26 characters, H4 & alaaaabfaaaroaaaaaaaaaaaaaaaaadj \\
        26 characters, H7 & bqadhakmagausabaaaiiaaaaaaaajalp \\
        26 characters, H13 & taknapqbmawvbjaaaoodgafaaaaoaqsa \\
    \end{tabular}
    \caption{\textbf{[Examples of random strings used in the paper.]}
    We show the first 32 tokens/characters.
    }
    \label{tab:random_string_examples}
\end{table}

Table~\ref{tab:random_string_examples} shows examples of random token strings used in the paper.
Each character is tokenized individually.

We also use non-Latin alphabets in Appendix~\ref{app:dynamics_non_latin}.
An example of such a random chosen alphabet for $\ell = 26$, from the Pythia tokenizer, is the following:
``Ġecosystem'',
``281'',
``Ġredistribute'',
``ĠEurope'',
``eni'',
``ricted'',
``Meanwhile'',
``Ġpropensity'',
``.""'',
``Du'',
``ĠAlice'',
``ortical'',
``Ġultrasonic'',
``Ġinclud'',
``Blocks'',
``thur'',
``Ġyears'',
``ramento'',
``ashion'',
``)\}\$\$'',
``onical'',
``Beck'',
``].)'',
``Ġpendant'',
``uma'',
``ynote''

\subsection{Computational resources}
\label{app:compute}

All experiments were conducted on machines in an in-house cluster with 2 x NVIDIA A40 GPUs with 48GB of memory, and with NVIDIA 2 x or 8 x A100 GPUs with 80GB of memory.
We use the A40 machines for training smaller models, such as Pythia-1B, and also Phi-2.7B with main memory offloading.
We use the A100 machines for training larger models, such as Pythia-12B, and Llama2-7B and Llama2-13B, and for some Phi-2.7B training runs.
It is possible to run most experiments, except the ones using larger batch sizes in Appendix~\ref{app:rw_val_batches}, on a single GPU, possibly using main memory offloading for larger models.

The experiments on the memorisation dynamics in Sections~\ref{sec:phases} and~\ref{sec:memorability}, including the corresponding results reported in the appendix, used around 1100 GPU hours.
The experiments on the role of local prefixes and global context in Section~\ref{sec:recall_information} used around 900 GPU hours.
The experiments on sequential memorisation in Section~\ref{sec:repeated_memorization} used around 500 GPU hours.
In total, the experiments in the paper used around 2500 GPU hours, distributed over the different GPU types.

\section{Additional results for the memorisation dynamics}
\label{app:memorability}

\subsection{Additional models and metrics}
\label{app:dynamics_additional_results}

\begin{figure}[H]
    \centering
    \subfloat[Pythia-70M]{
        \includegraphics[width=\allModelsWidth]{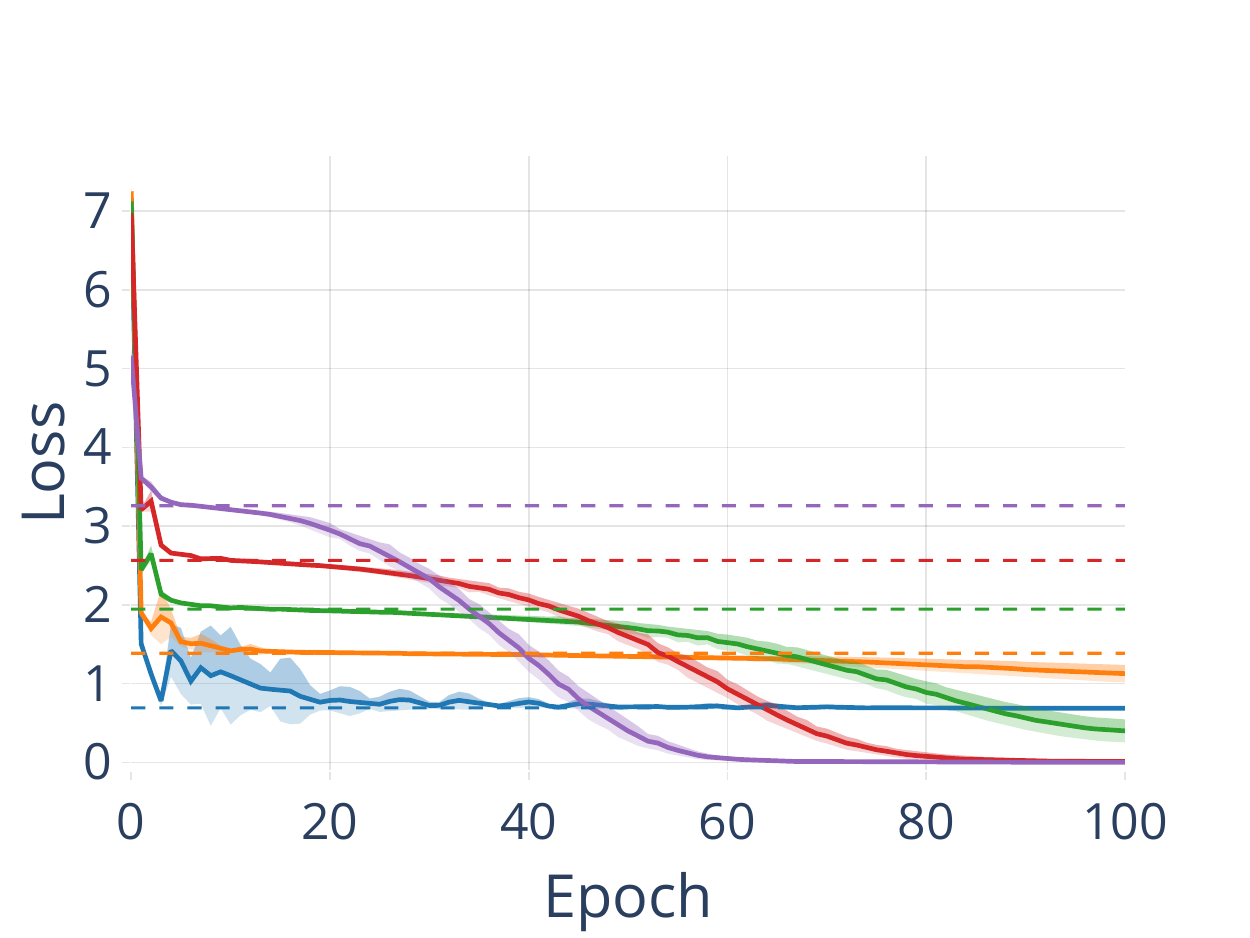}
    }
    \subfloat[Pythia-1B]{
        \includegraphics[width=\allModelsWidth]{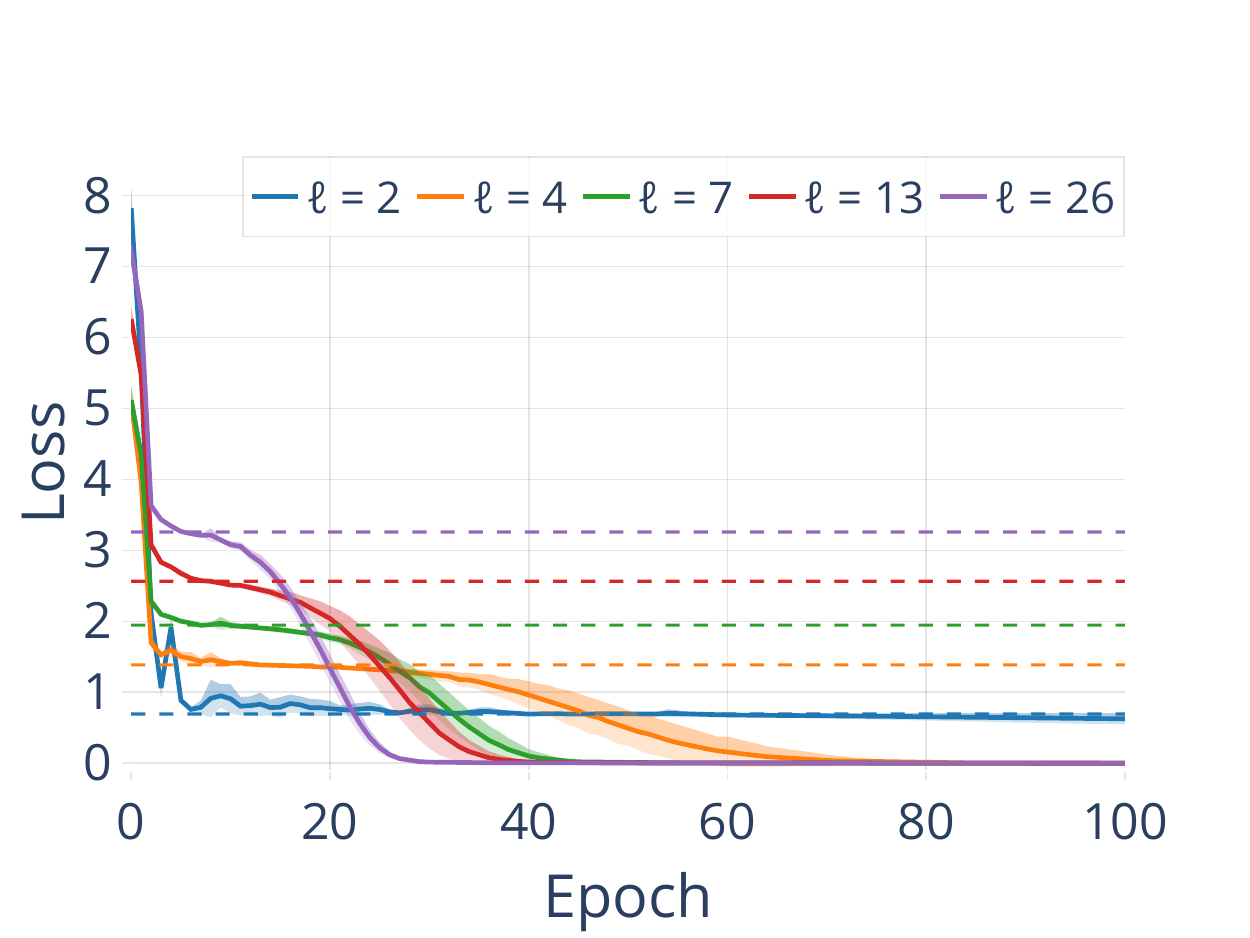}
    }
    \subfloat[Pythia-12B]{
        \includegraphics[width=\allModelsWidth]{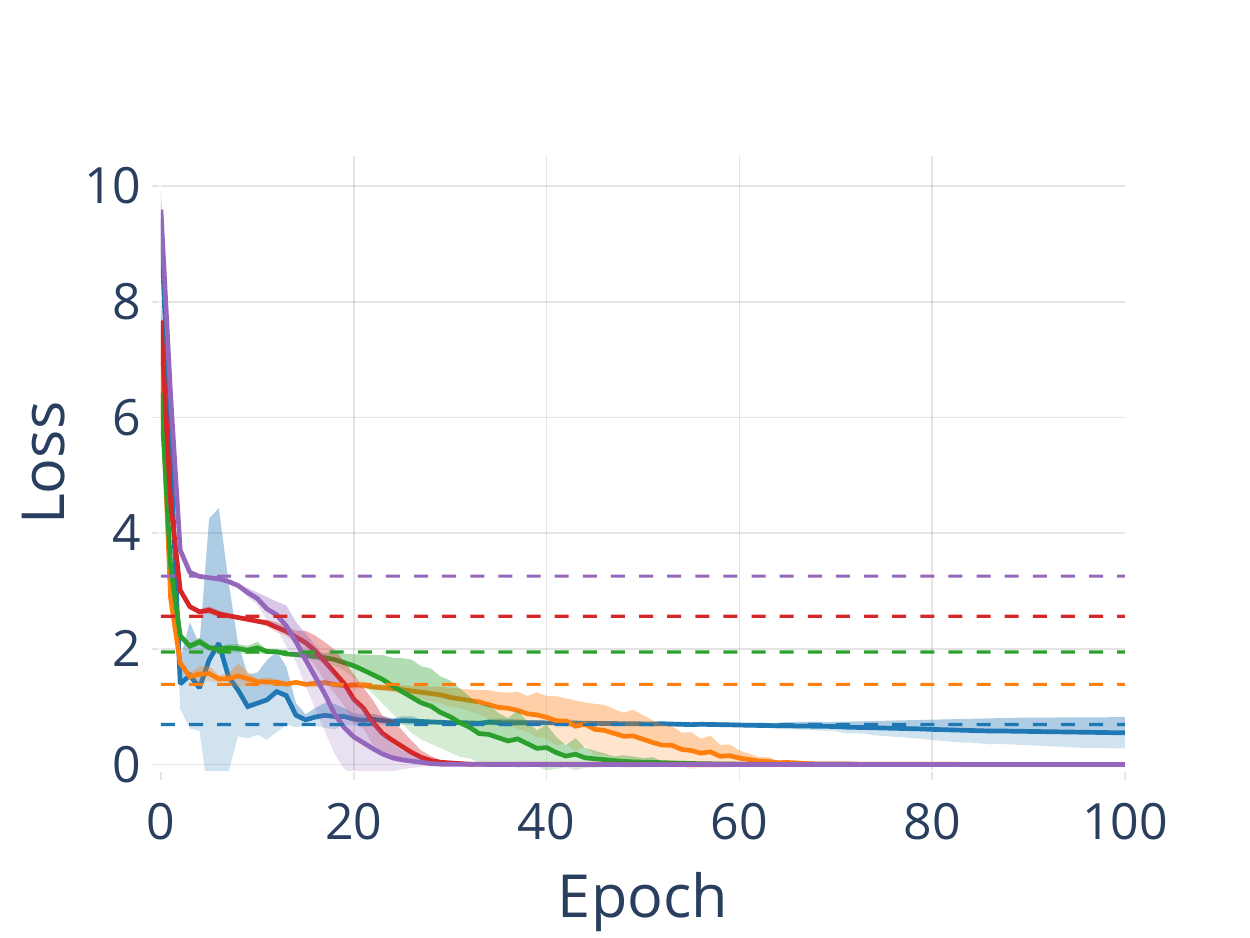}
    }
    \\
    \subfloat[Phi-1.3B]{
        \includegraphics[width=\allModelsWidth]{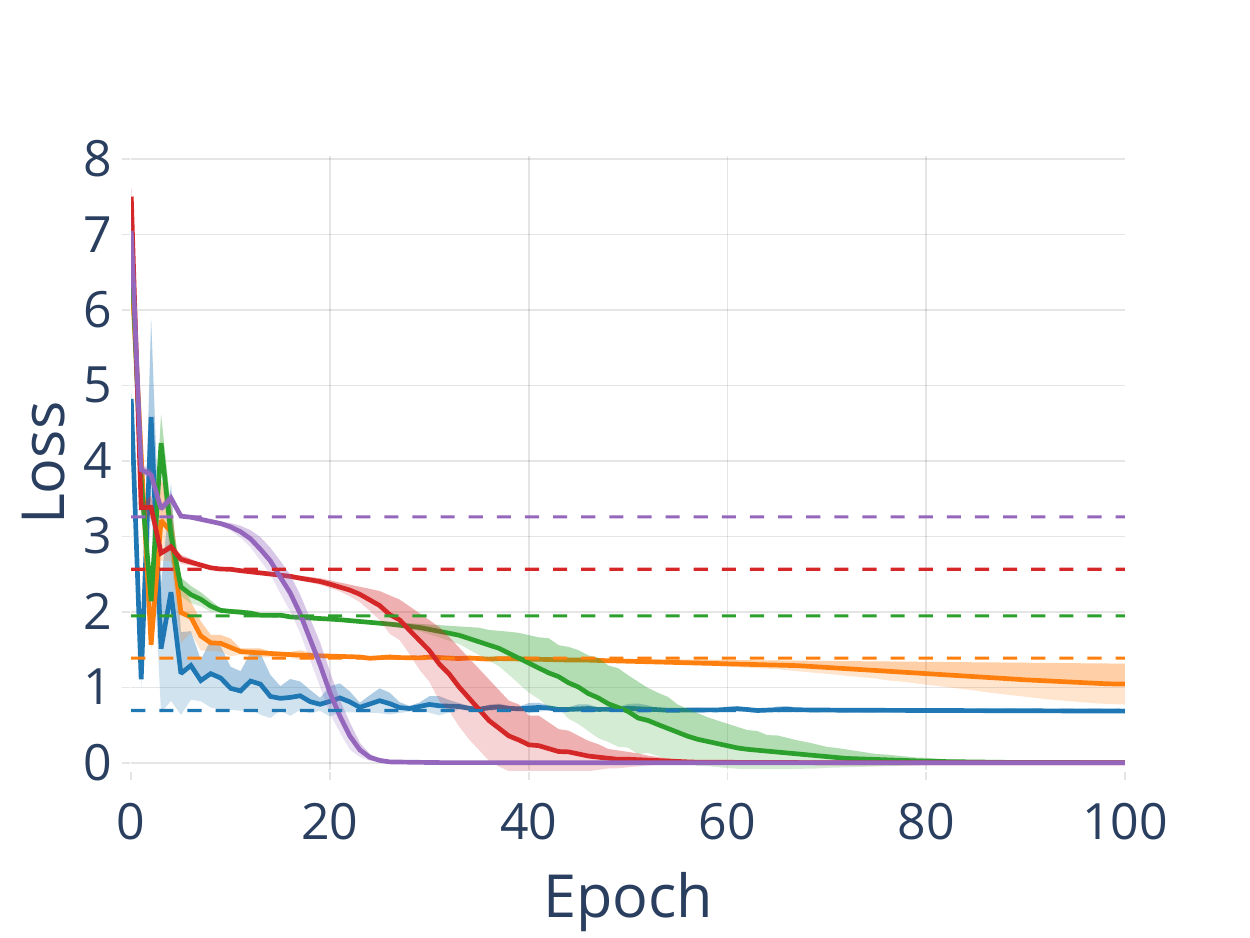}
    }
    \subfloat[Phi-2.7B]{
        \includegraphics[width=\allModelsWidth]{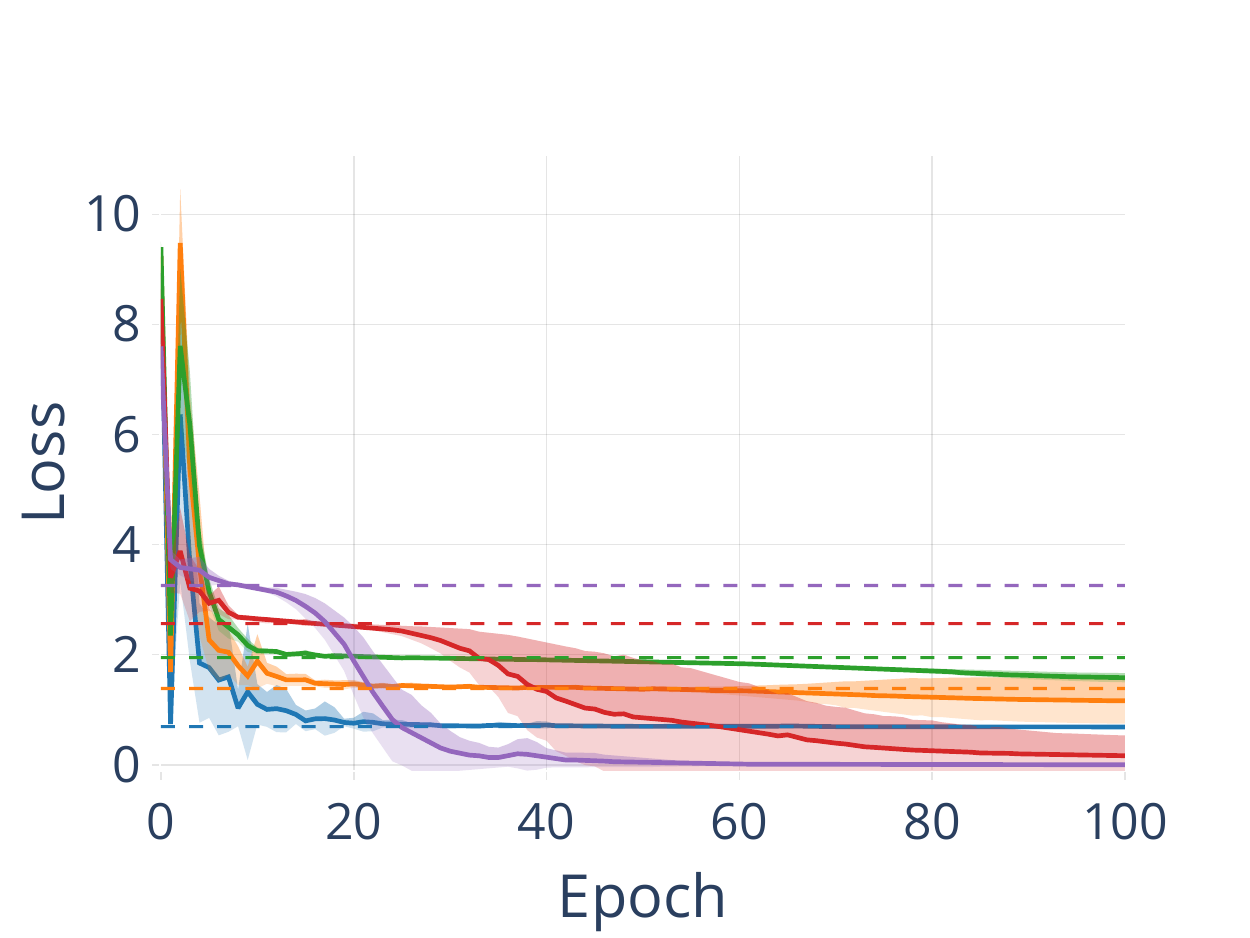}
    }
    \subfloat[Llama2-7B]{
        \includegraphics[width=\allModelsWidth]{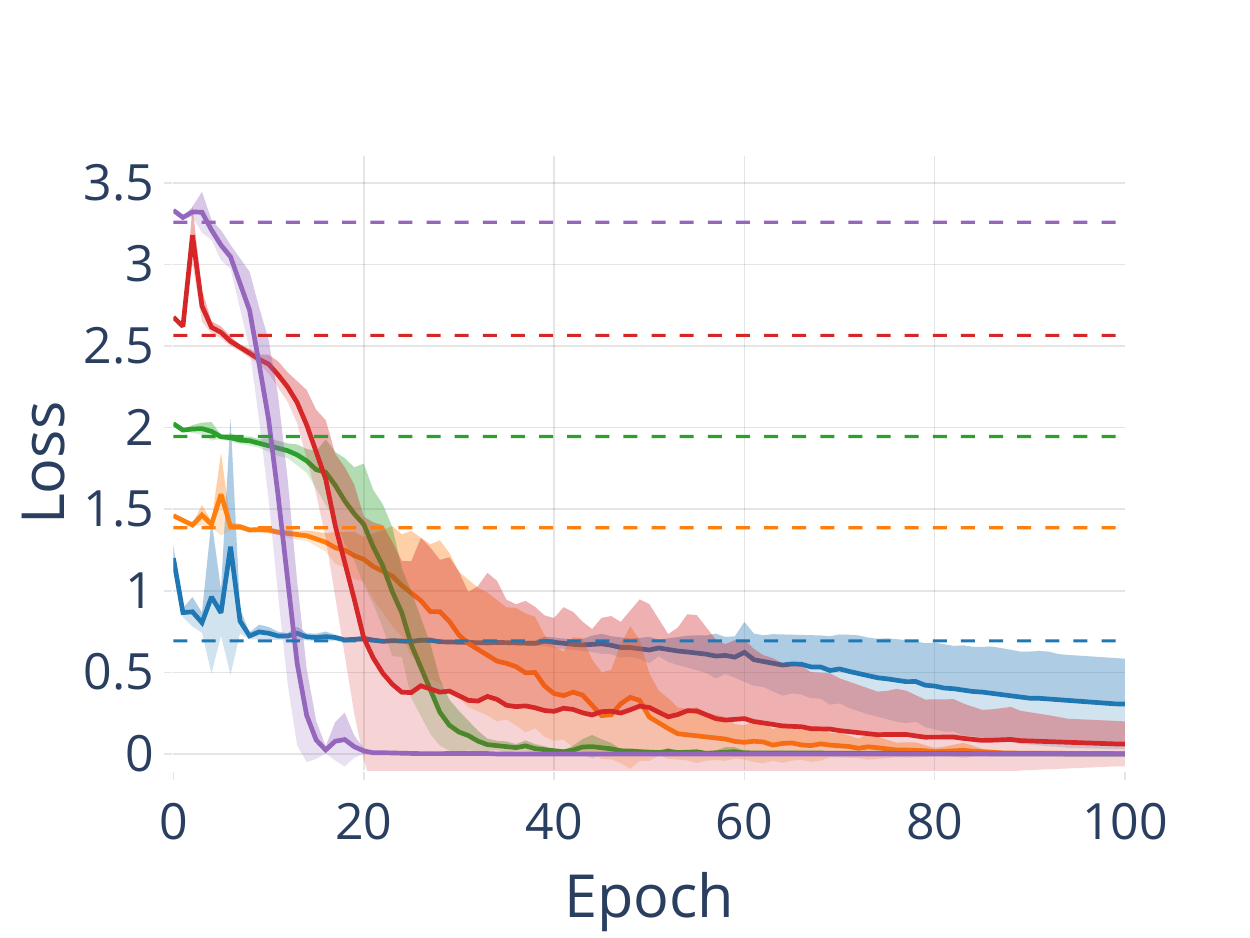}
    }
    \subfloat[Llama2-13B]{
        \includegraphics[width=\allModelsWidth]{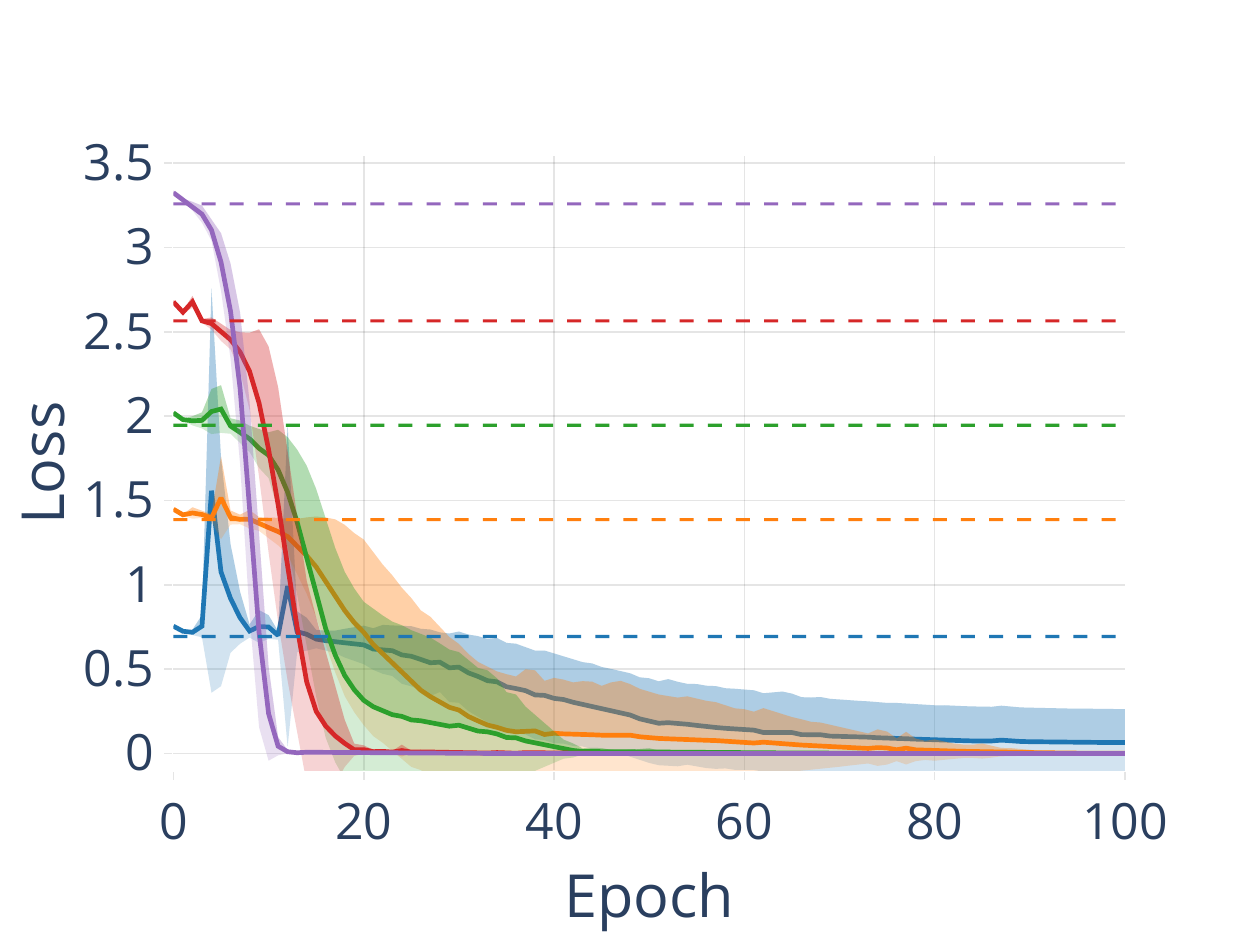}
    }
\caption{\capthead{Loss for all models for different $\ell$.}{$n = 1024$}
}
\label{fig:loss_alphabet_size_all}
\end{figure}

\begin{figure}[H]
    \centering
    \subfloat[Pythia-70M]{
        \includegraphics[width=\allModelsWidth]{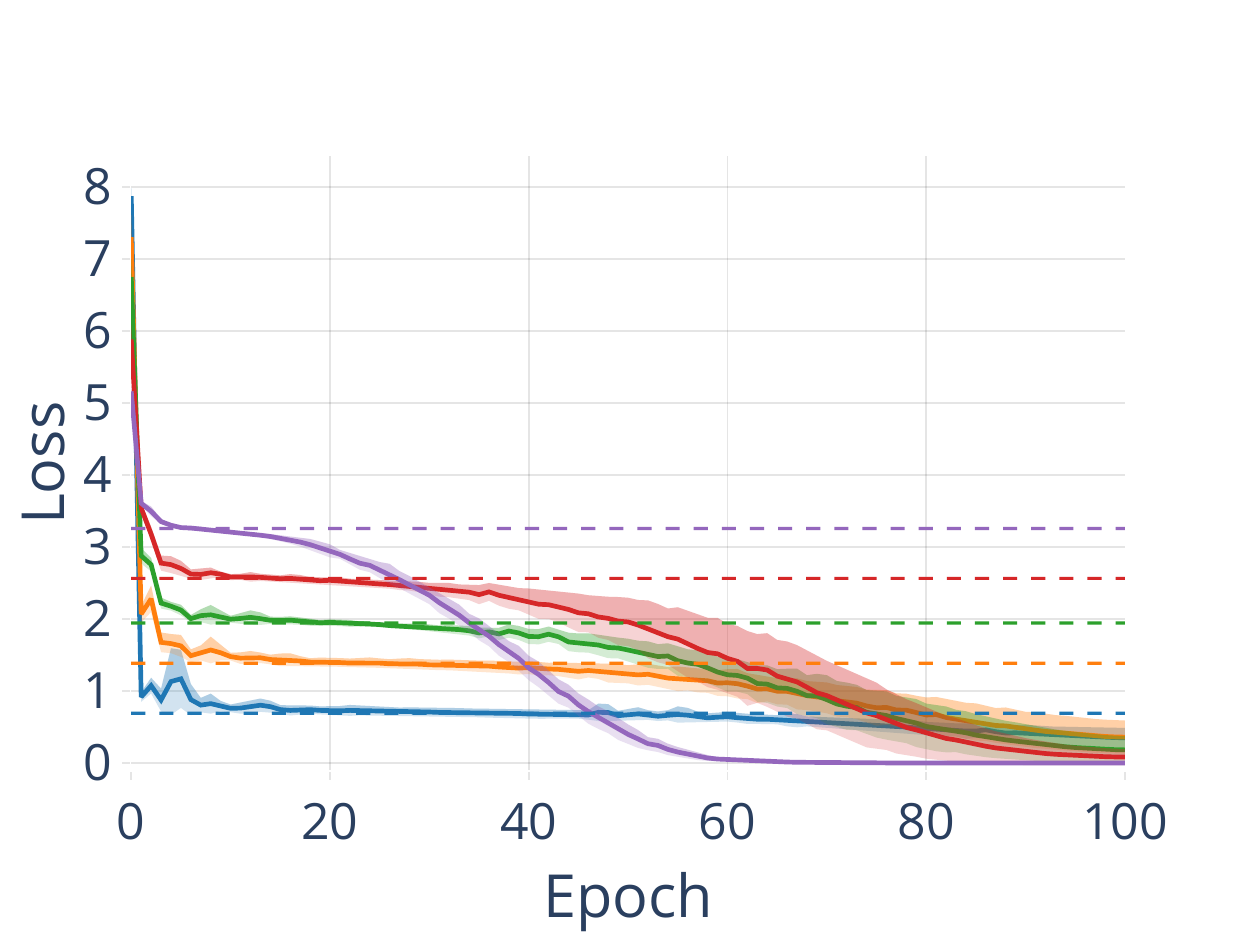}
    }
    \subfloat[Pythia-1B]{
        \includegraphics[width=\allModelsWidth]{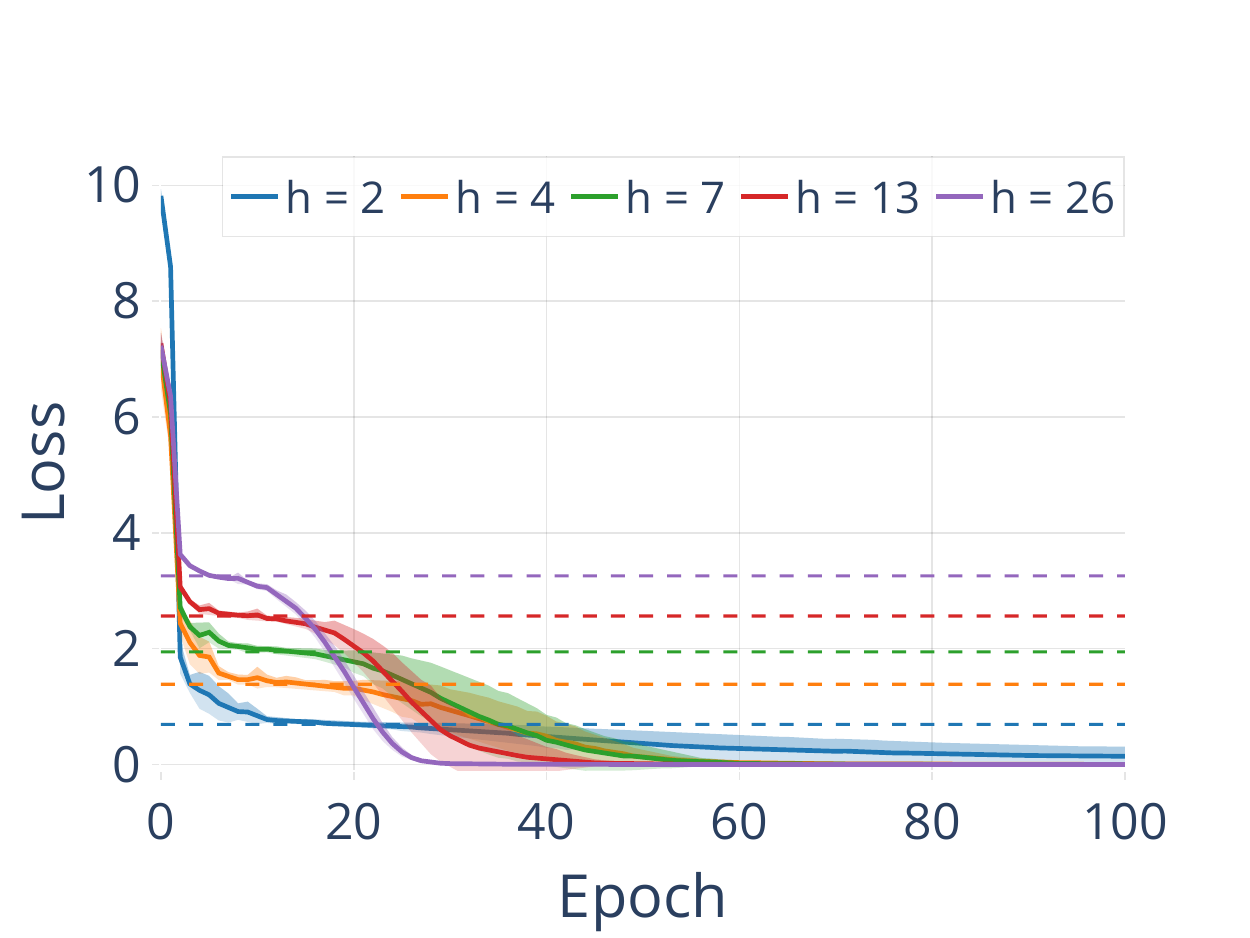}
    }
    \subfloat[Pythia-12B]{
        \includegraphics[width=\allModelsWidth]{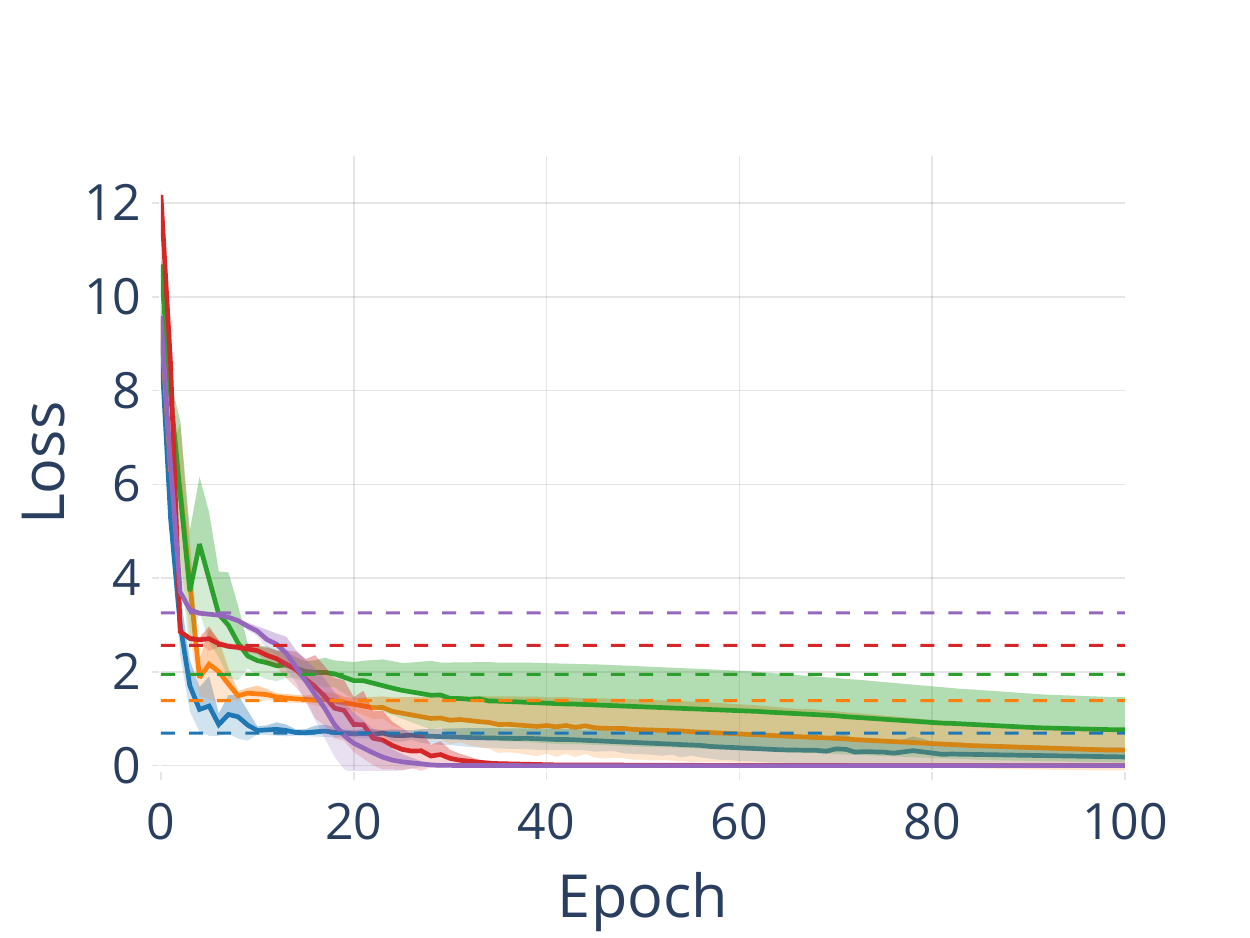}
    }
    \\
    \subfloat[Phi-1.3B]{
        \includegraphics[width=\allModelsWidth]{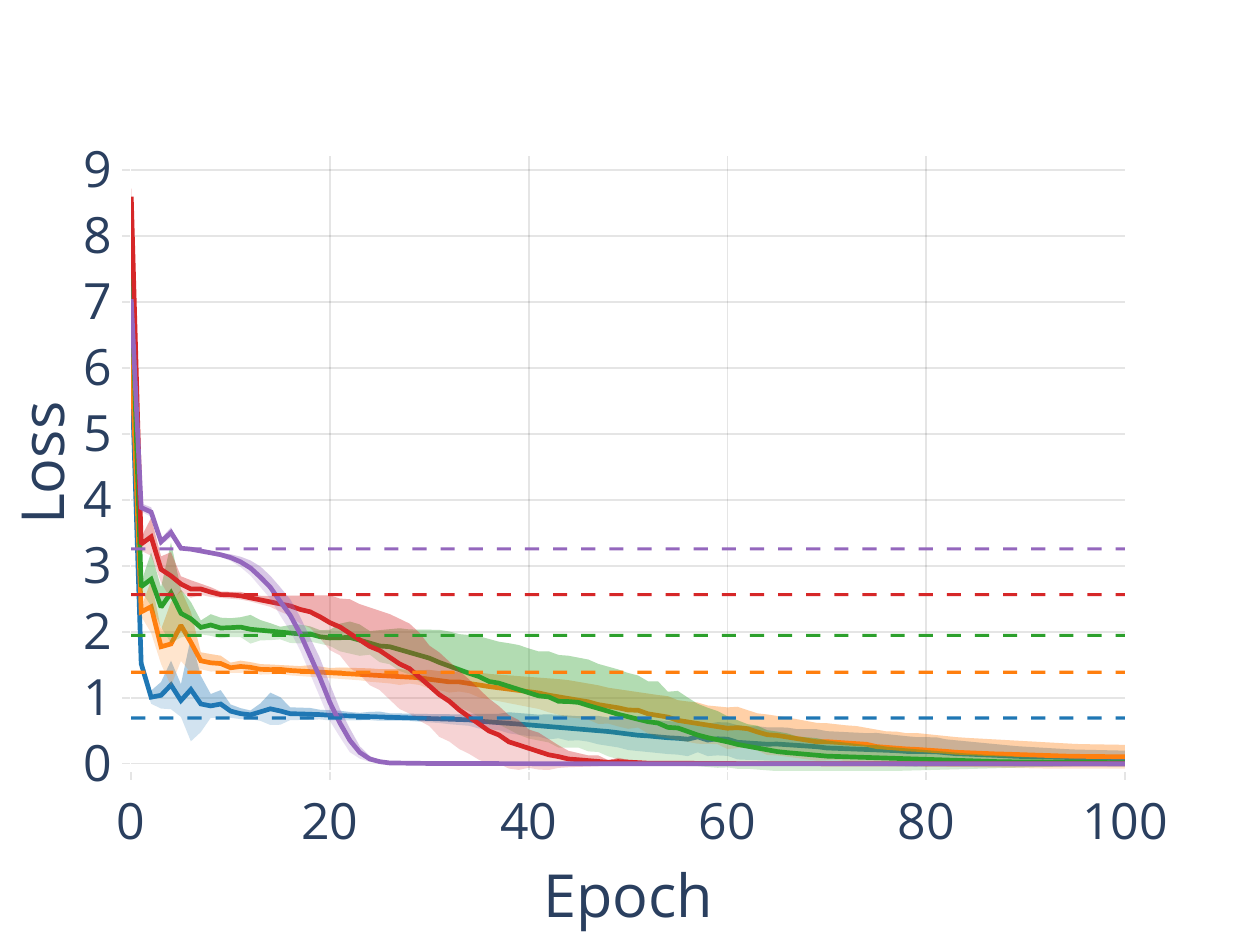}
    }
    \subfloat[Phi-2.7B]{
        \includegraphics[width=\allModelsWidth]{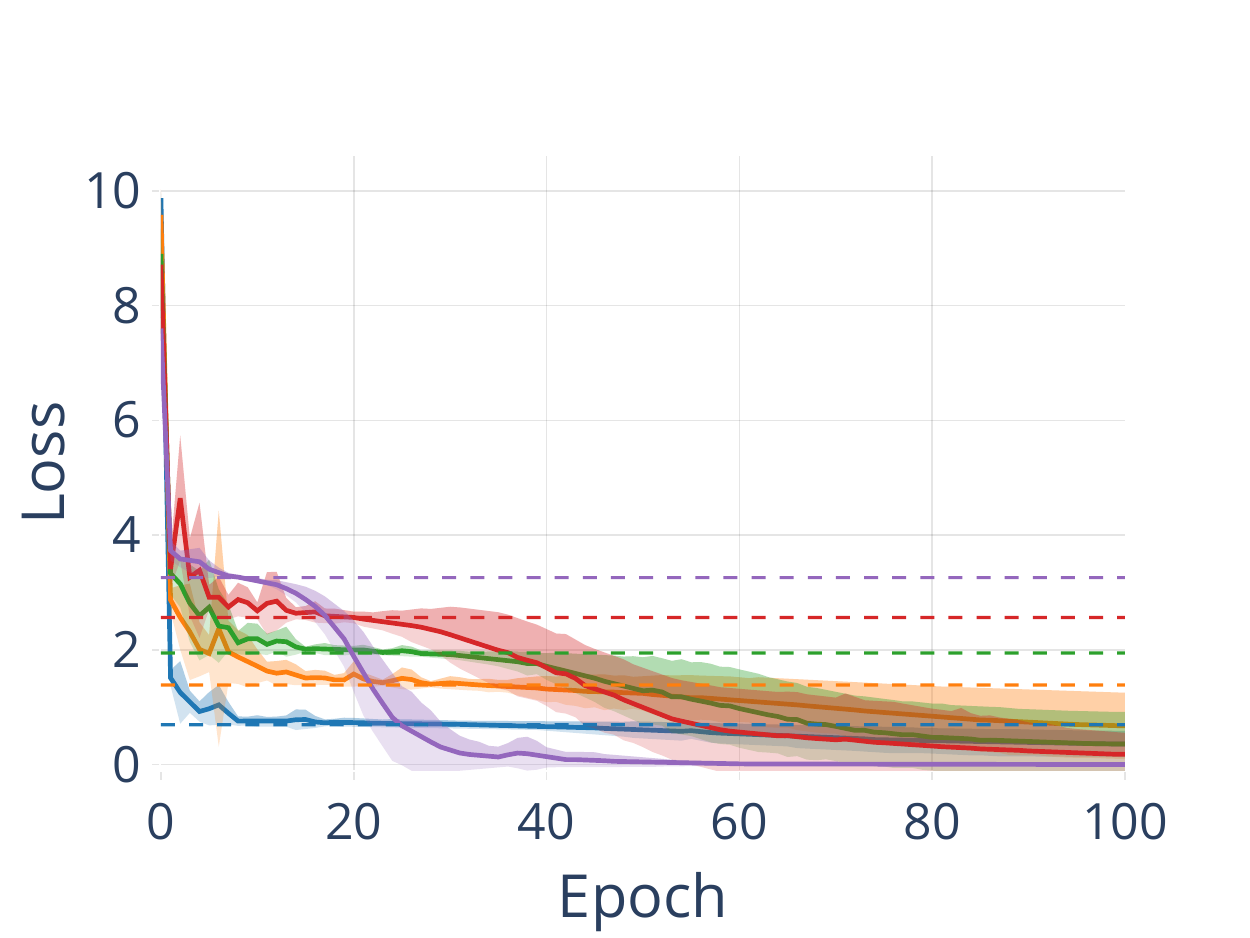}
    }
    \subfloat[Llama2-7B]{
        \includegraphics[width=\allModelsWidth]{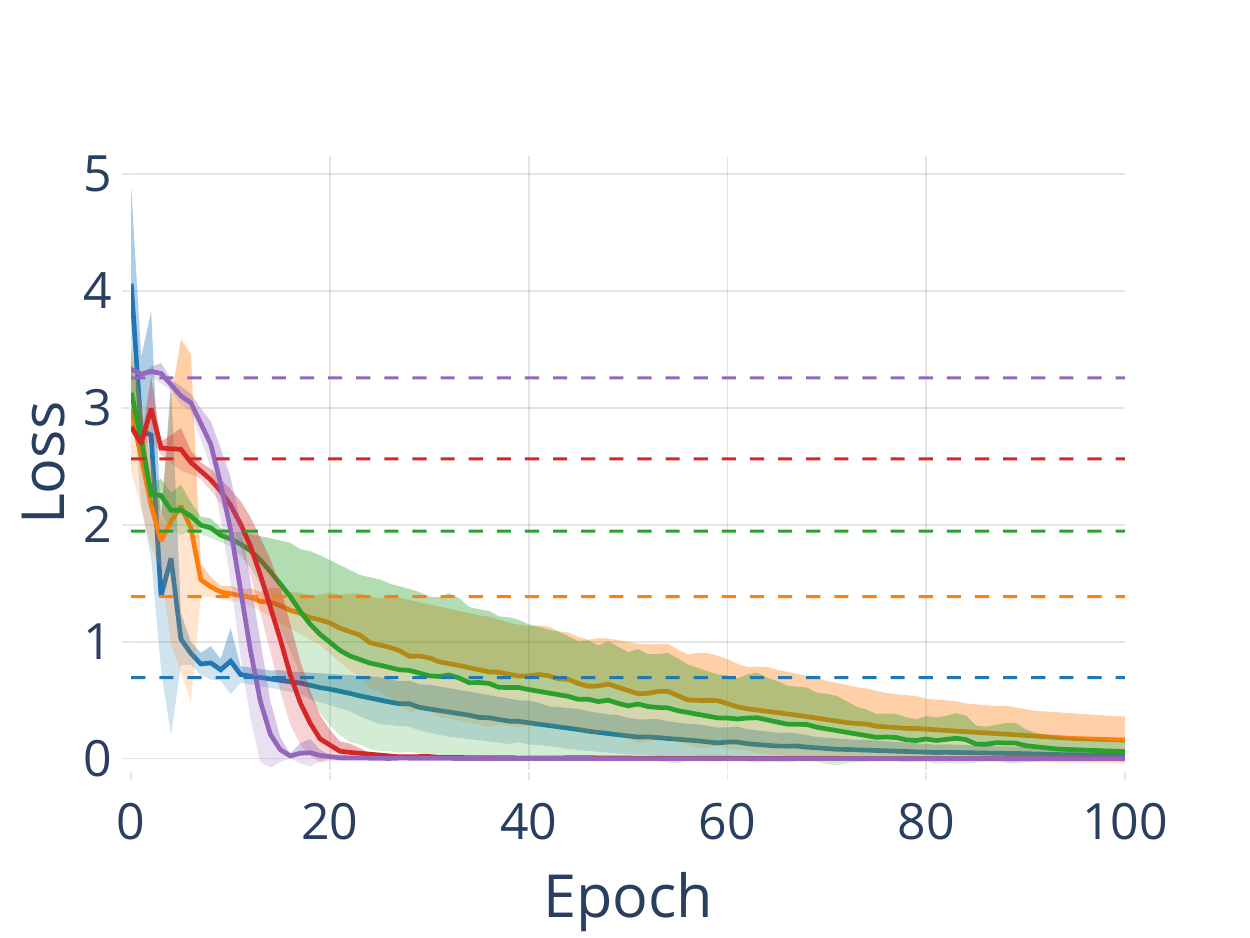}
    }
    \subfloat[Llama2-13B]{
        \includegraphics[width=\allModelsWidth]{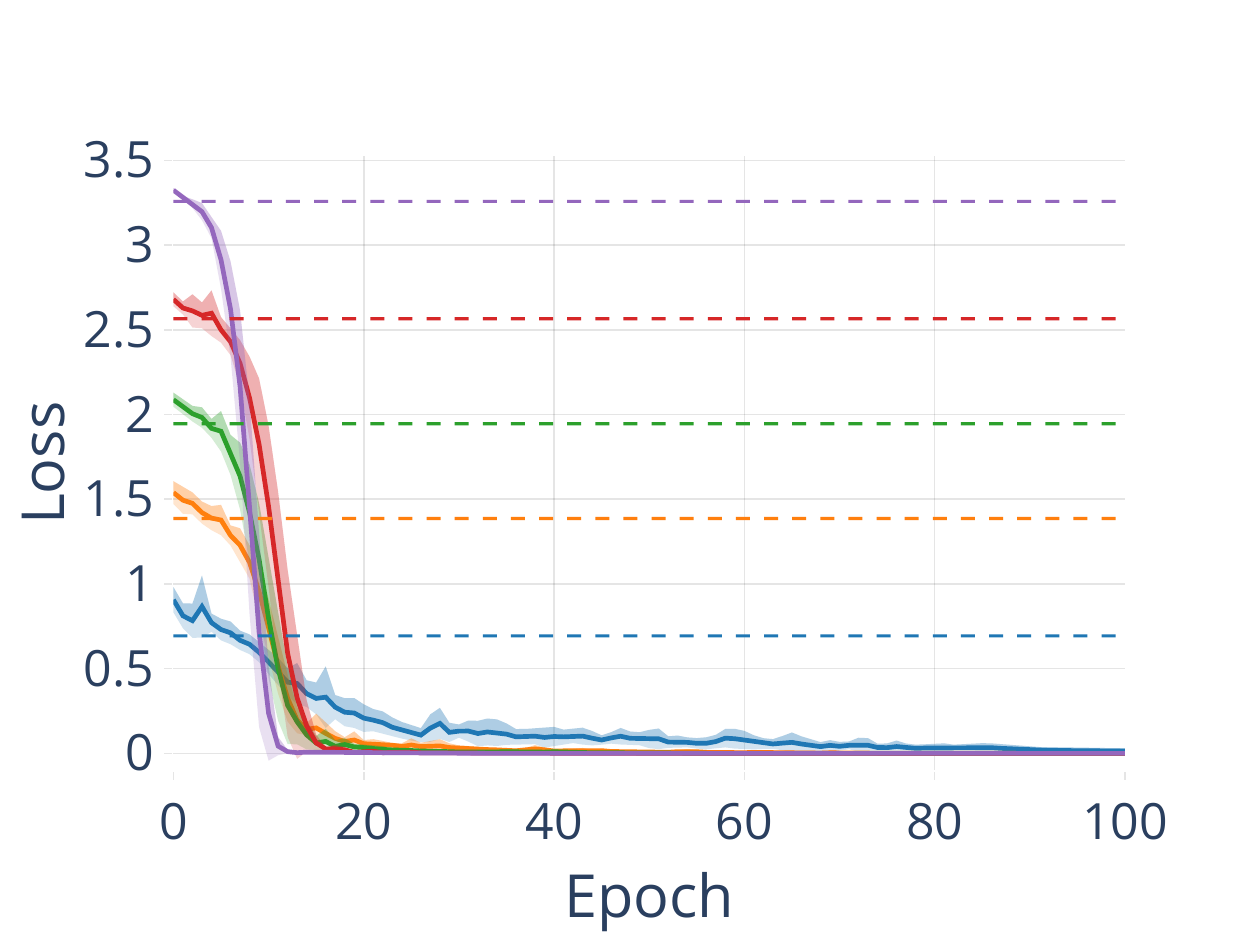}
    }
\caption{\capthead{Loss for all models for different $h$.}{$n = 1024, \ell = 26$}
}
\label{fig:loss_entropy_level_all}
\end{figure}

\begin{figure}[H]
    \centering
    \subfloat[Pythia-70M]{
        \includegraphics[width=\allModelsWidth]{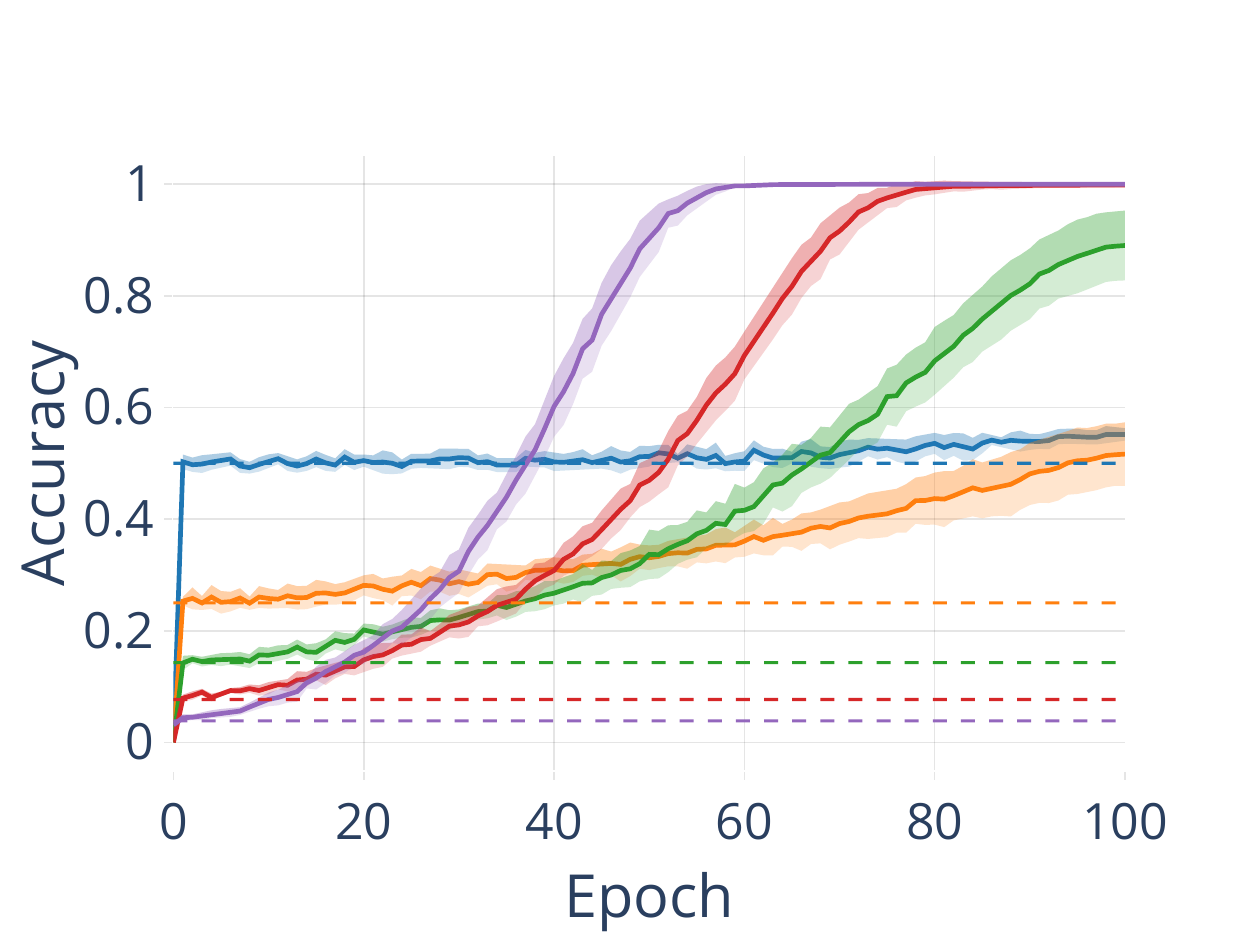}
    }
    \subfloat[Pythia-1B]{
        \includegraphics[width=\allModelsWidth]{figures/memorability/alphabet_size/accuracy_alphabet-size_pythia-1b.pdf}
    }
    \subfloat[Pythia-12B]{
        \includegraphics[width=\allModelsWidth]{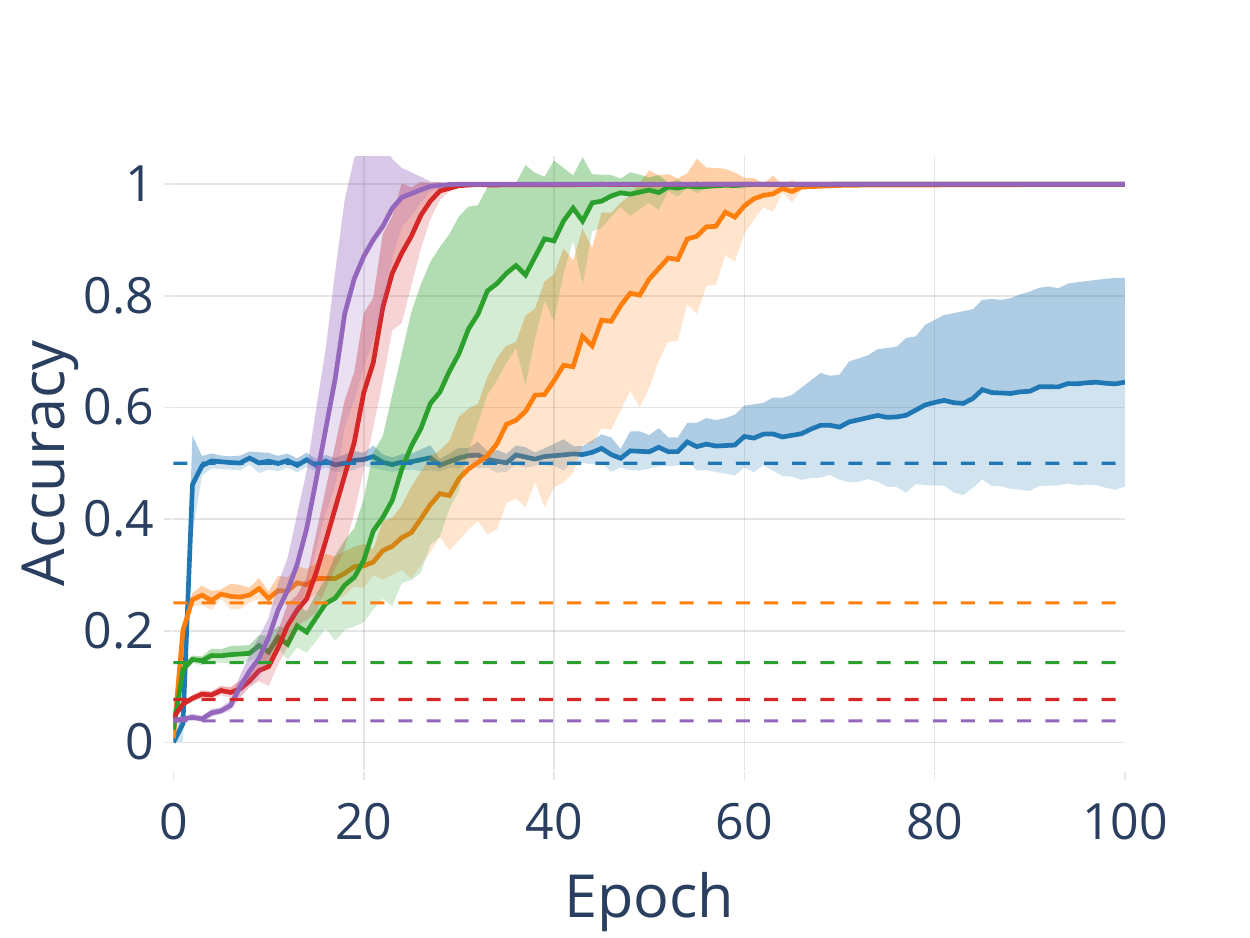}
    }
    \\
    \subfloat[Phi-1.3B]{
        \includegraphics[width=\allModelsWidth]{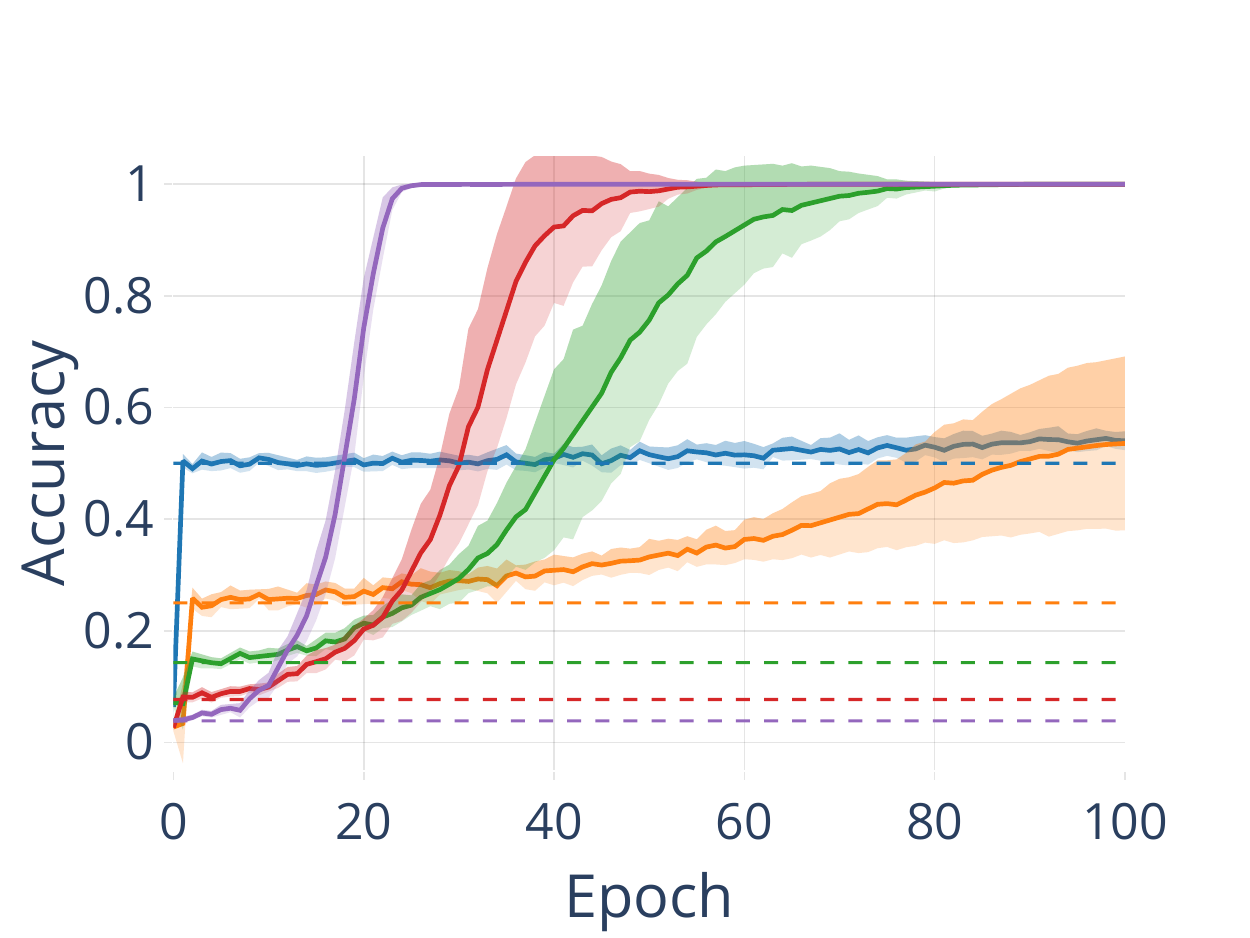}
    }
    \subfloat[Phi-2.7B]{
        \includegraphics[width=\allModelsWidth]{figures/memorability/alphabet_size/accuracy_alphabet-size_phi-2.7b.pdf}
    }
    \subfloat[Llama2-7B]{
        \includegraphics[width=\allModelsWidth]{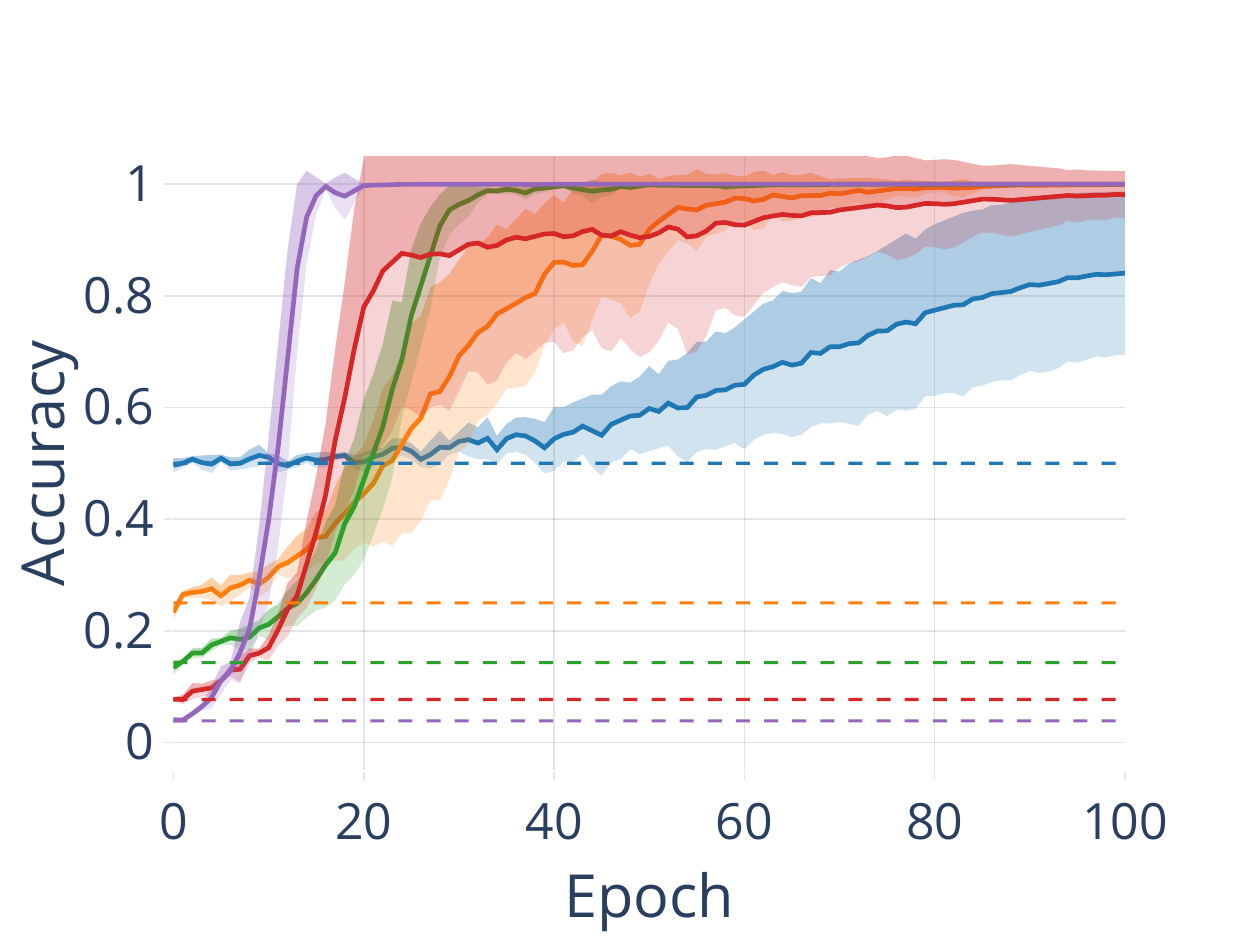}
    }
    \subfloat[Llama2-13B]{
        \includegraphics[width=\allModelsWidth]{figures/memorability/alphabet_size/accuracy_alphabet-size_llama2-13b.pdf}
    }
\caption{\capthead{Accuracy for all models for different $\ell$.}{$n = 1024$}
}
\label{fig:accuracy_alphabet_size_all}
\end{figure}

\begin{figure}[H]
    \centering
    \subfloat[Pythia-70M]{
        \includegraphics[width=\allModelsWidth]{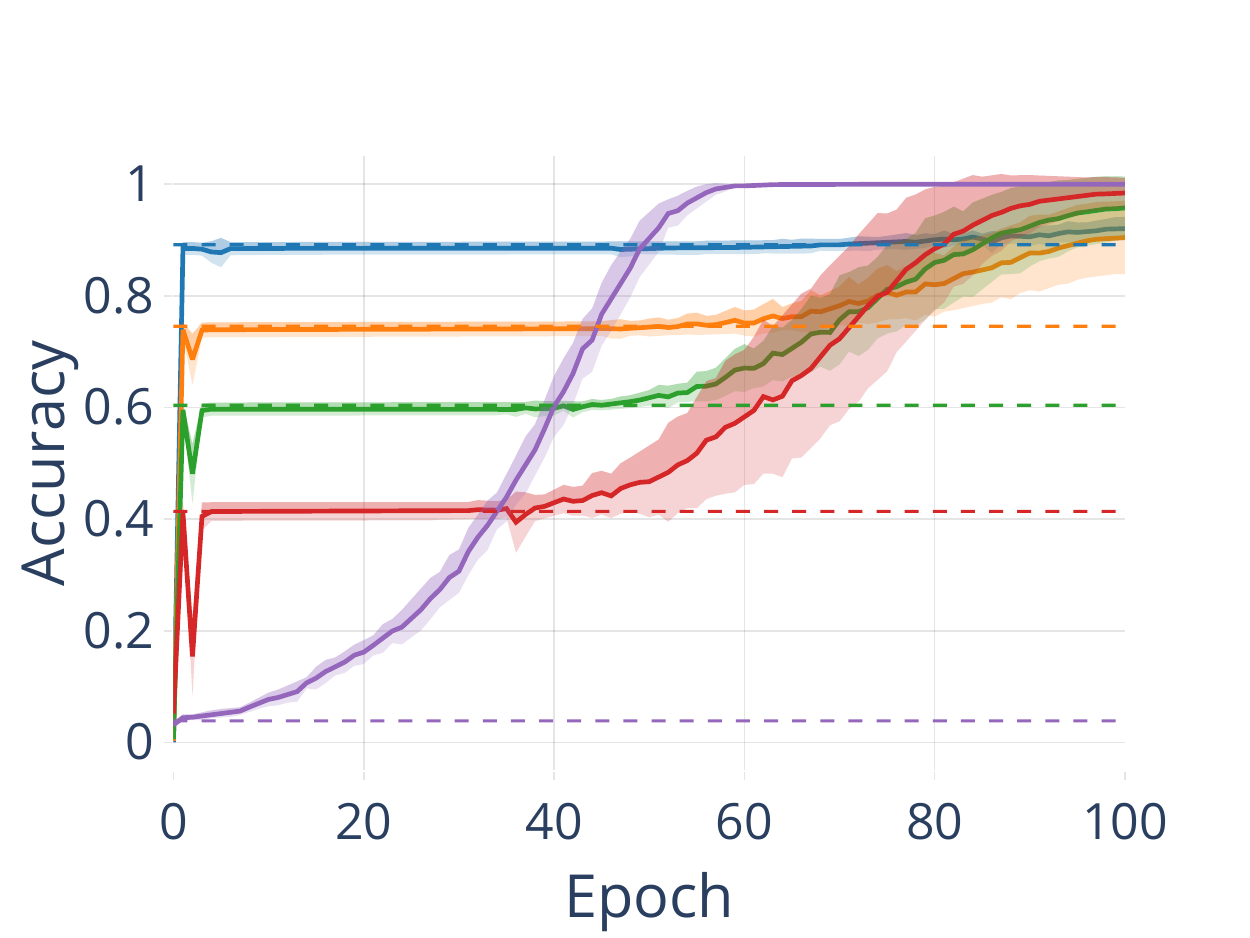}
    }
    \subfloat[Pythia-1B]{
        \includegraphics[width=\allModelsWidth]{figures/memorability/entropy_level/accuracy_entropy-level_pythia-1b.pdf}
    }
    \subfloat[Pythia-12B]{
        \includegraphics[width=\allModelsWidth]{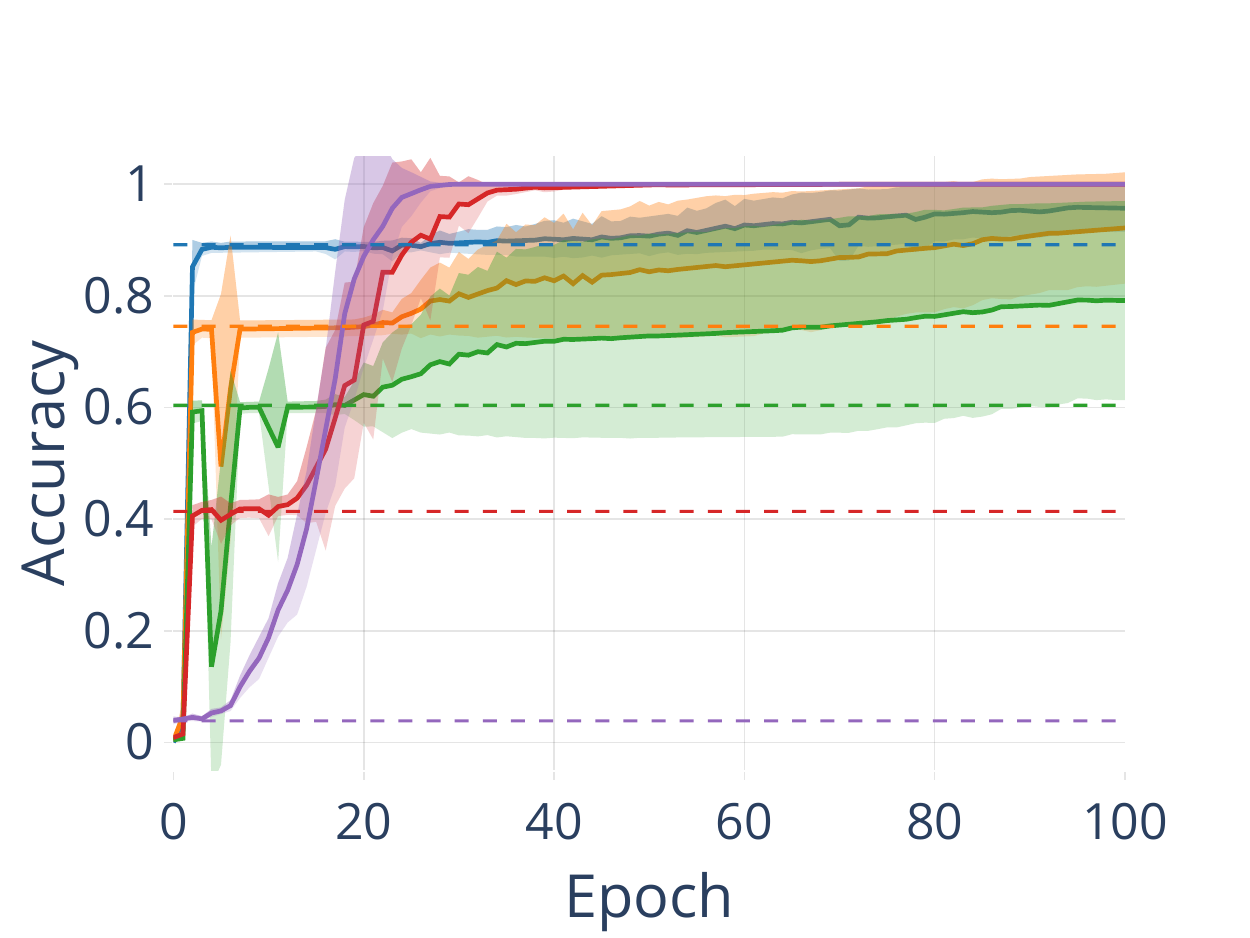}
    }
    \\
    \subfloat[Phi-1.3B]{
        \includegraphics[width=\allModelsWidth]{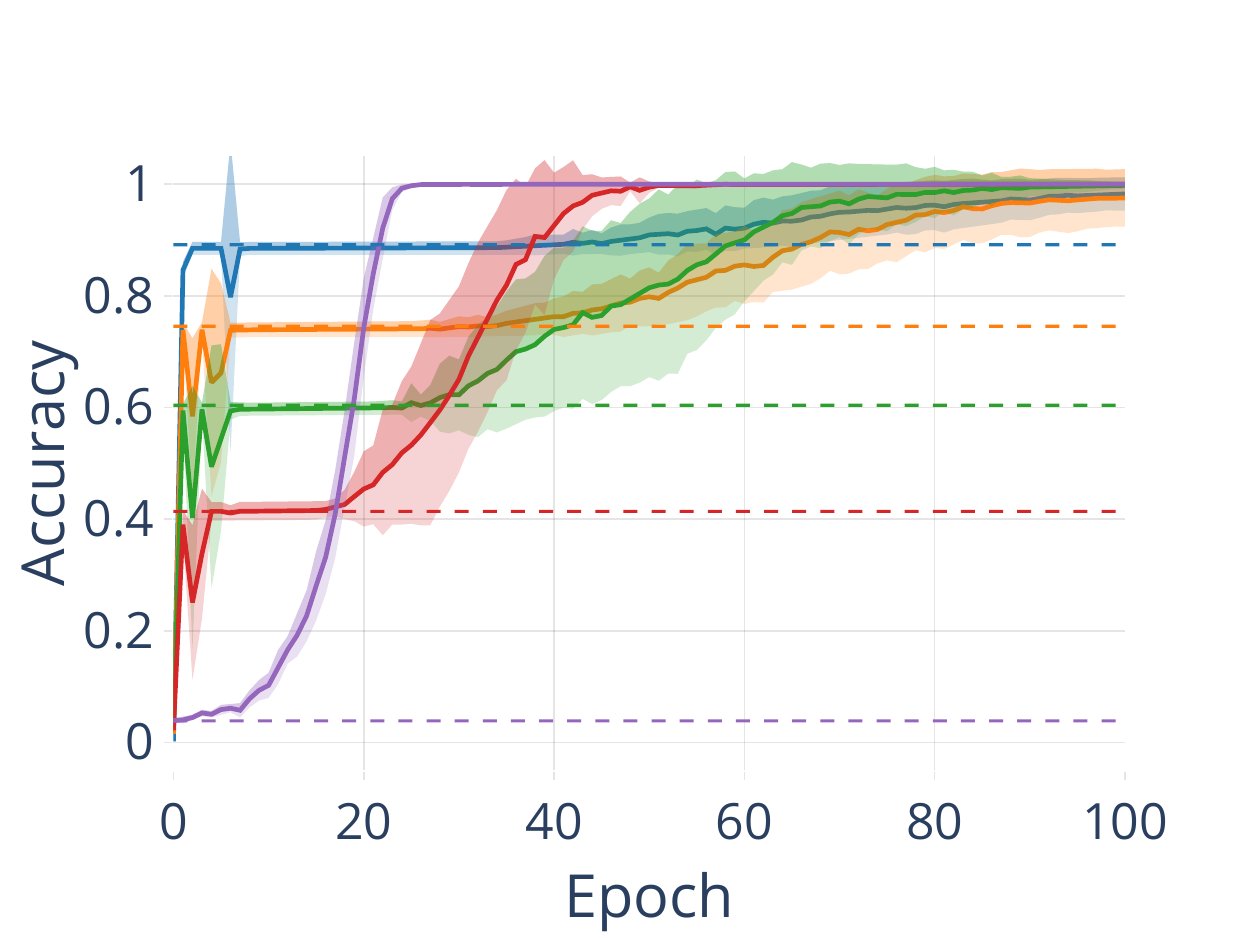}
    }
    \subfloat[Phi-2.7B]{
        \includegraphics[width=\allModelsWidth]{figures/memorability/entropy_level/accuracy_entropy-level_phi-2.7b.pdf}
    }
    \subfloat[Llama2-7B]{
        \includegraphics[width=\allModelsWidth]{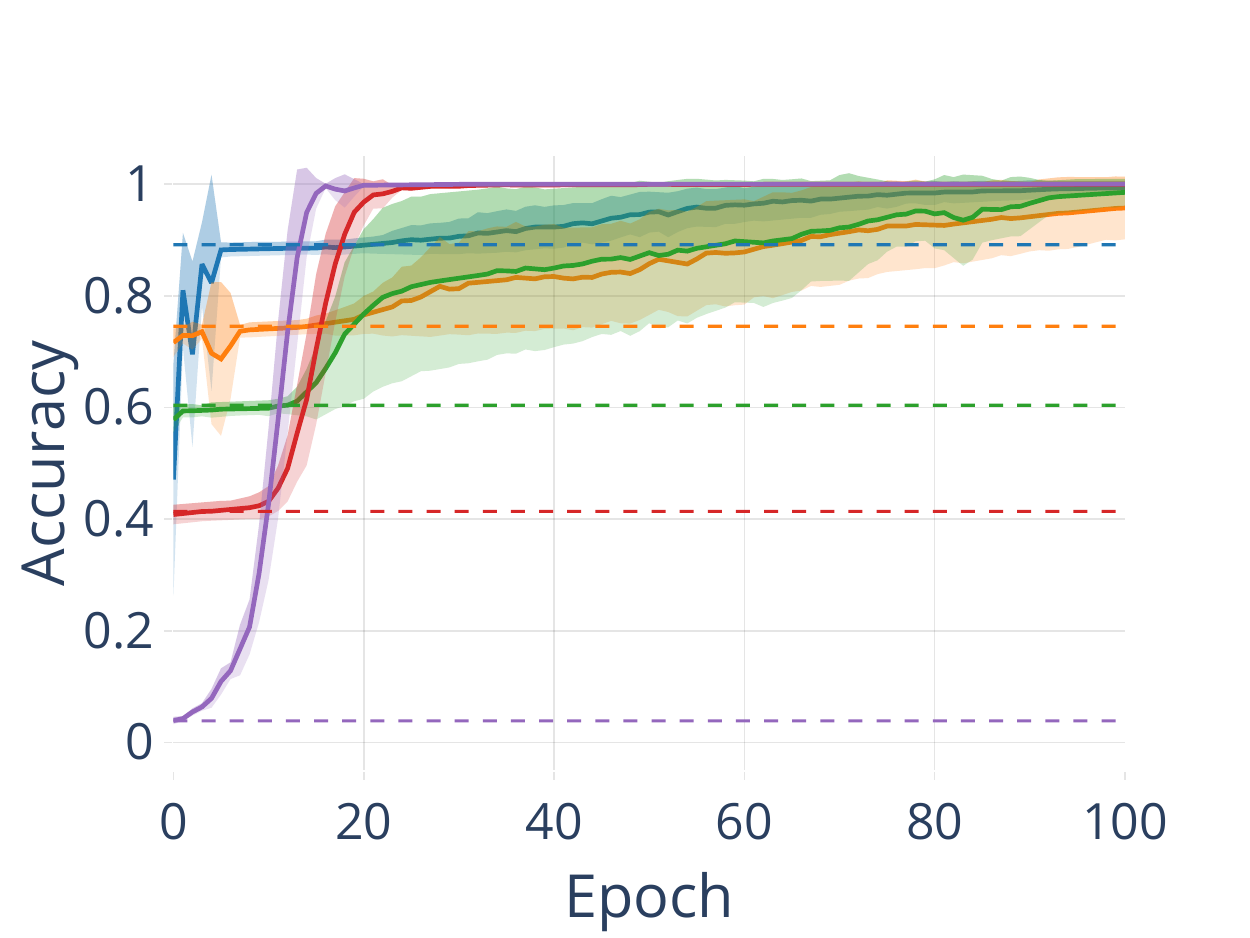}
    }
    \subfloat[Llama2-13B]{
        \includegraphics[width=\allModelsWidth]{figures/memorability/entropy_level/accuracy_entropy-level_llama2-13b.pdf}
    }
\caption{\capthead{Accuracy for all models for different $h$.}{$n = 1024, \ell = 26$}
}
\label{fig:accuracy_entropy_level_all}
\end{figure}

\begin{figure}[H]
    \centering
    \subfloat[Pythia-70M]{
        \includegraphics[width=\allModelsWidth]{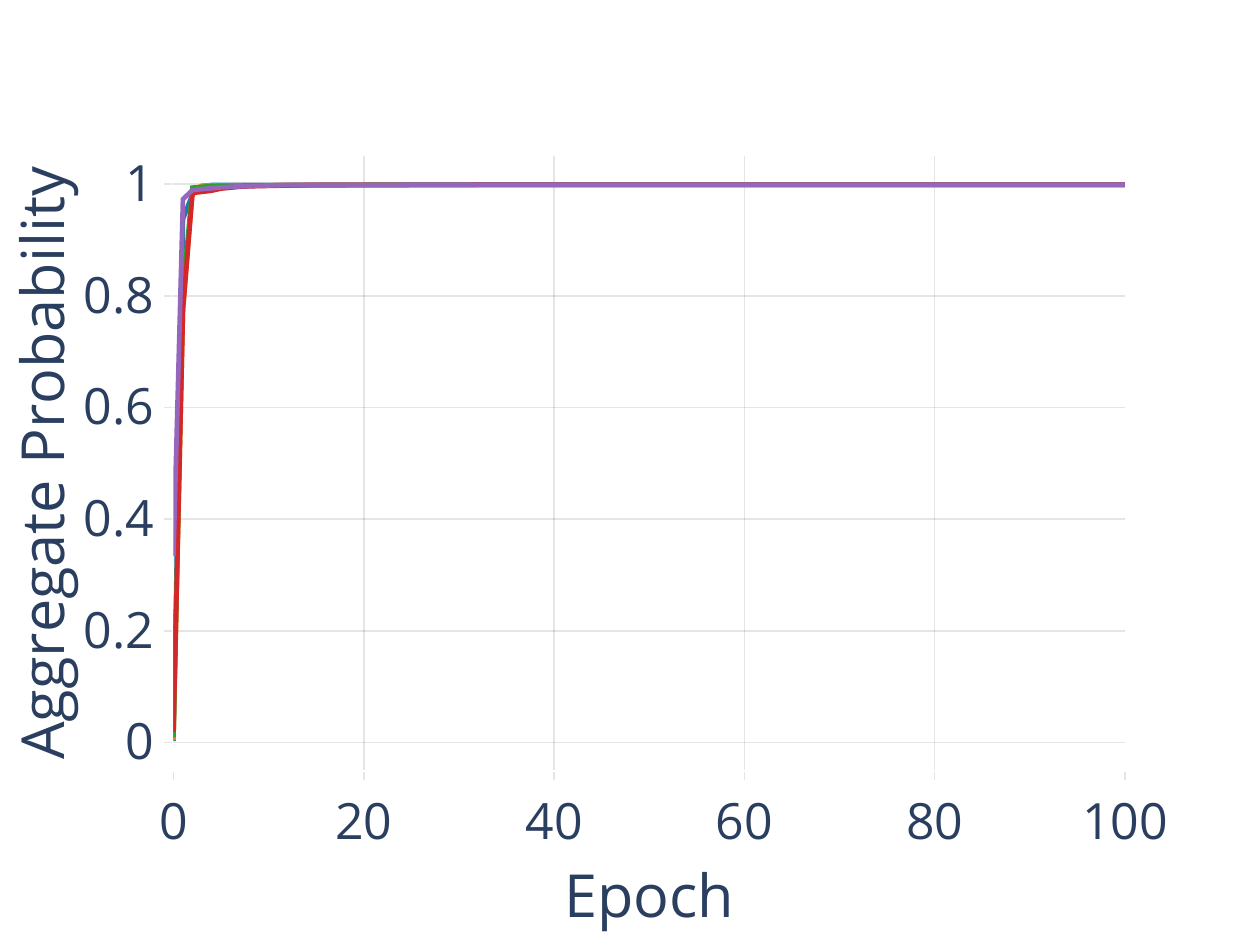}
    }
    \subfloat[Pythia-1B]{
        \includegraphics[width=\allModelsWidth]{figures/memorability/alphabet_size/cum-prob_alphabet-size_pythia-1b.pdf}
    }
    \subfloat[Pythia-12B]{
        \includegraphics[width=\allModelsWidth]{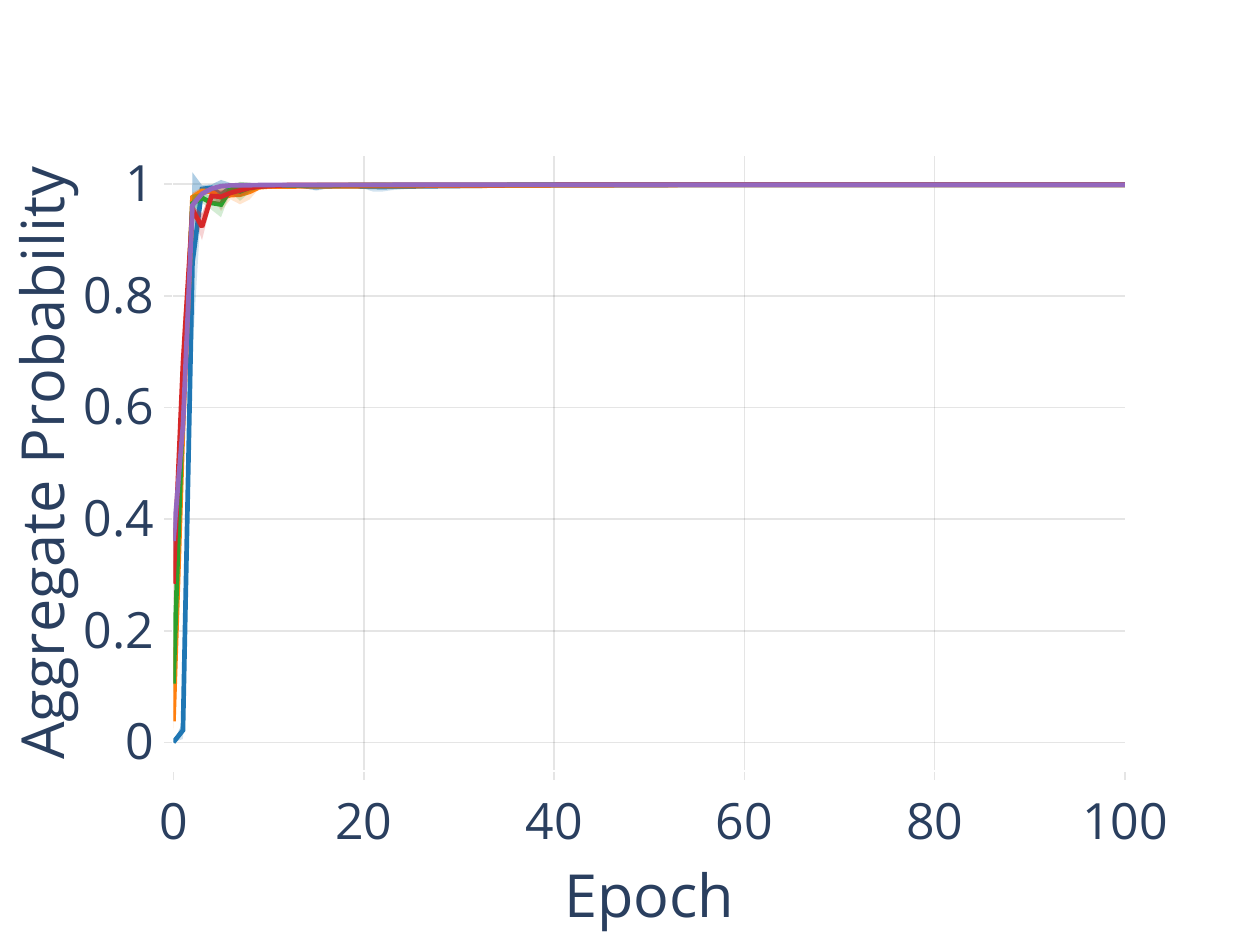}
    }
    \\
    \subfloat[Phi-1.3B]{
        \includegraphics[width=\allModelsWidth]{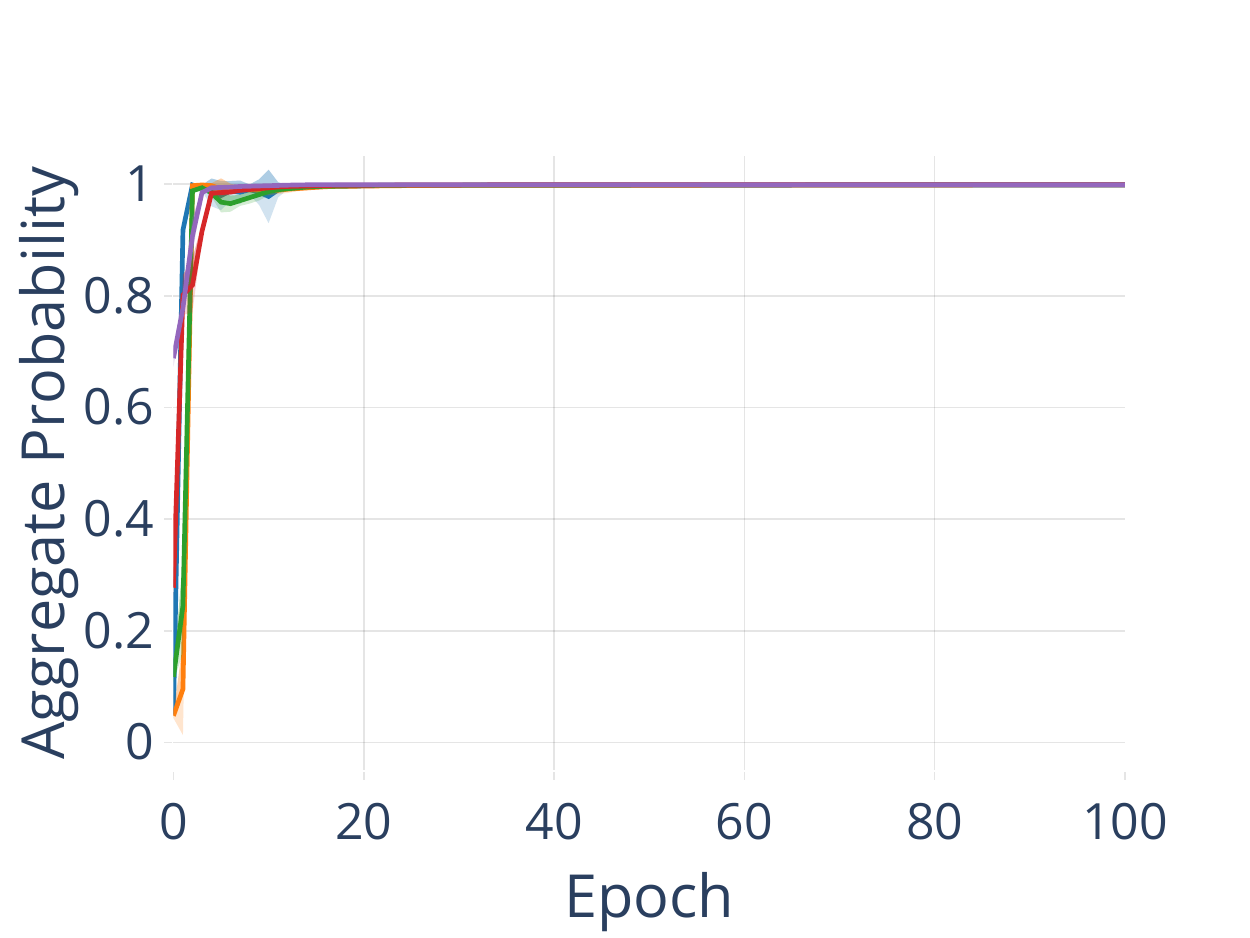}
    }
    \subfloat[Phi-2.7B]{
        \includegraphics[width=\allModelsWidth]{figures/memorability/alphabet_size/cum-prob_alphabet-size_phi-2.7b.pdf}
    }
    \subfloat[Llama2-7B]{
        \includegraphics[width=\allModelsWidth]{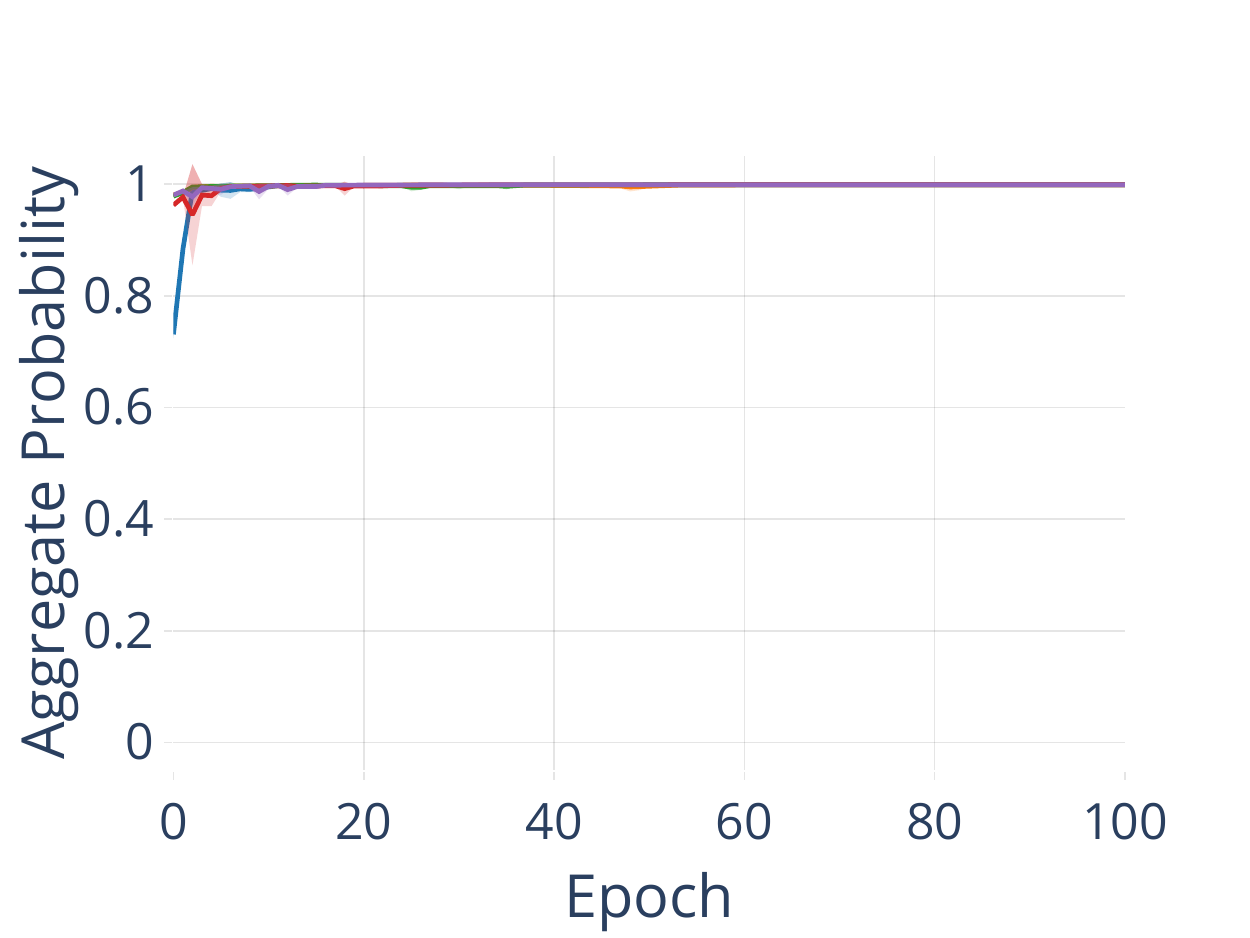}
    }
    \subfloat[Llama2-13B]{
        \includegraphics[width=\allModelsWidth]{figures/memorability/alphabet_size/cum-prob_alphabet-size_llama2-13b.pdf}
    }
\caption{\capthead{Aggregate Probability over $A$ for all models for different $\ell$.}{$n = 1024$}
}
\label{fig:cum-prob_alphabet_size_all}
\end{figure}

\begin{figure}[H]
    \centering
    \subfloat[Pythia-70M]{
        \includegraphics[width=\allModelsWidth]{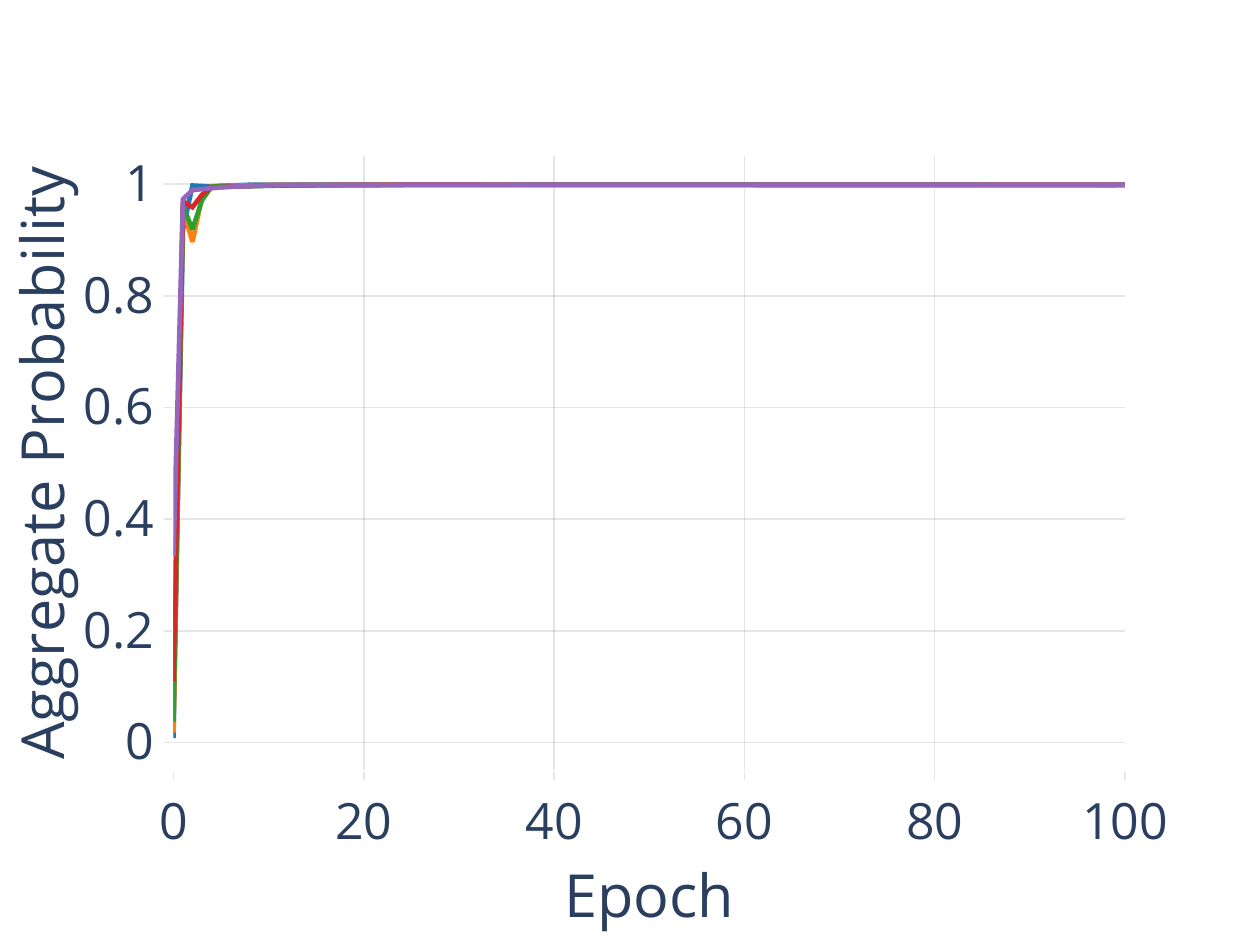}
    }
    \subfloat[Pythia-1B]{
        \includegraphics[width=\allModelsWidth]{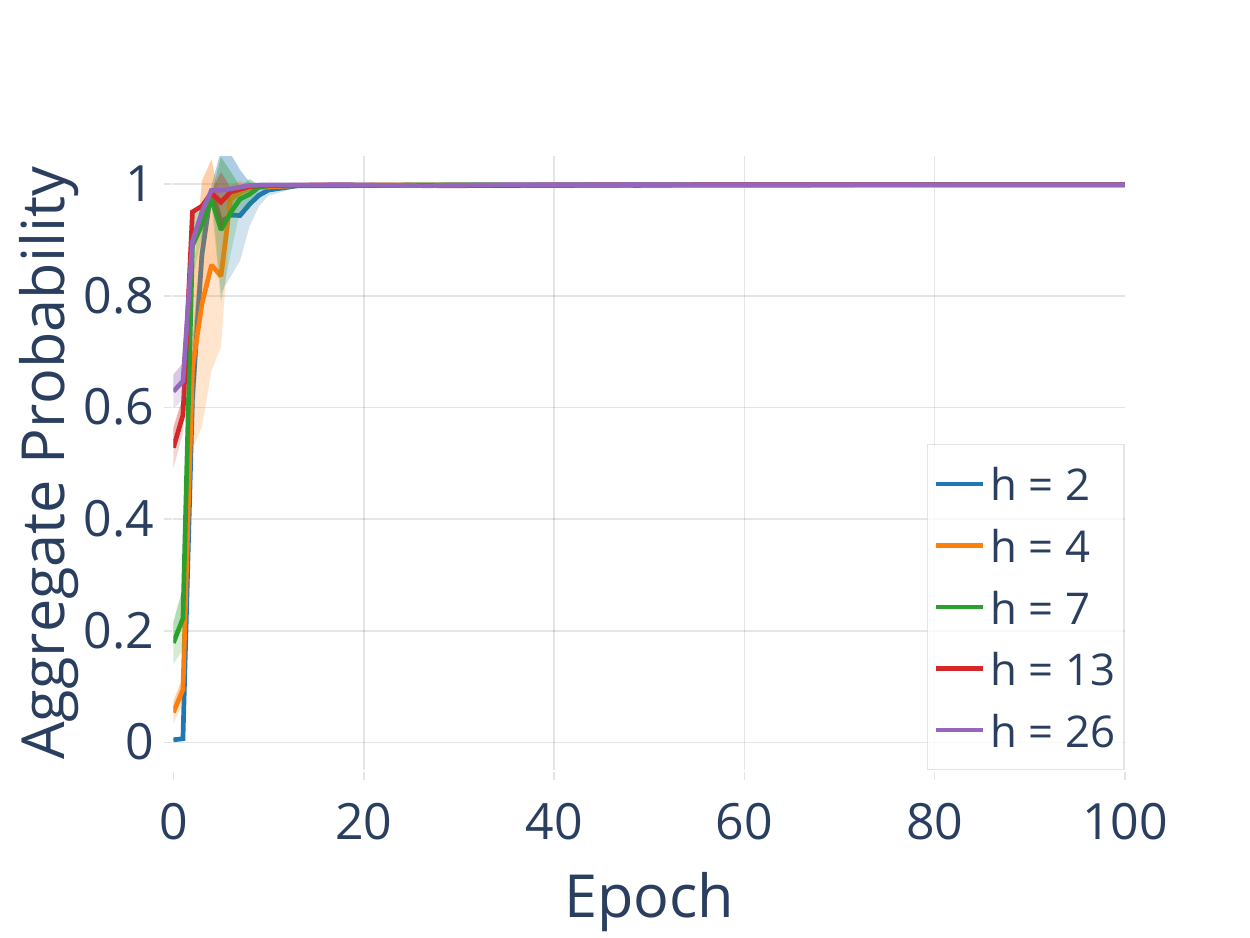}
    }
    \subfloat[Pythia-12B]{
        \includegraphics[width=\allModelsWidth]{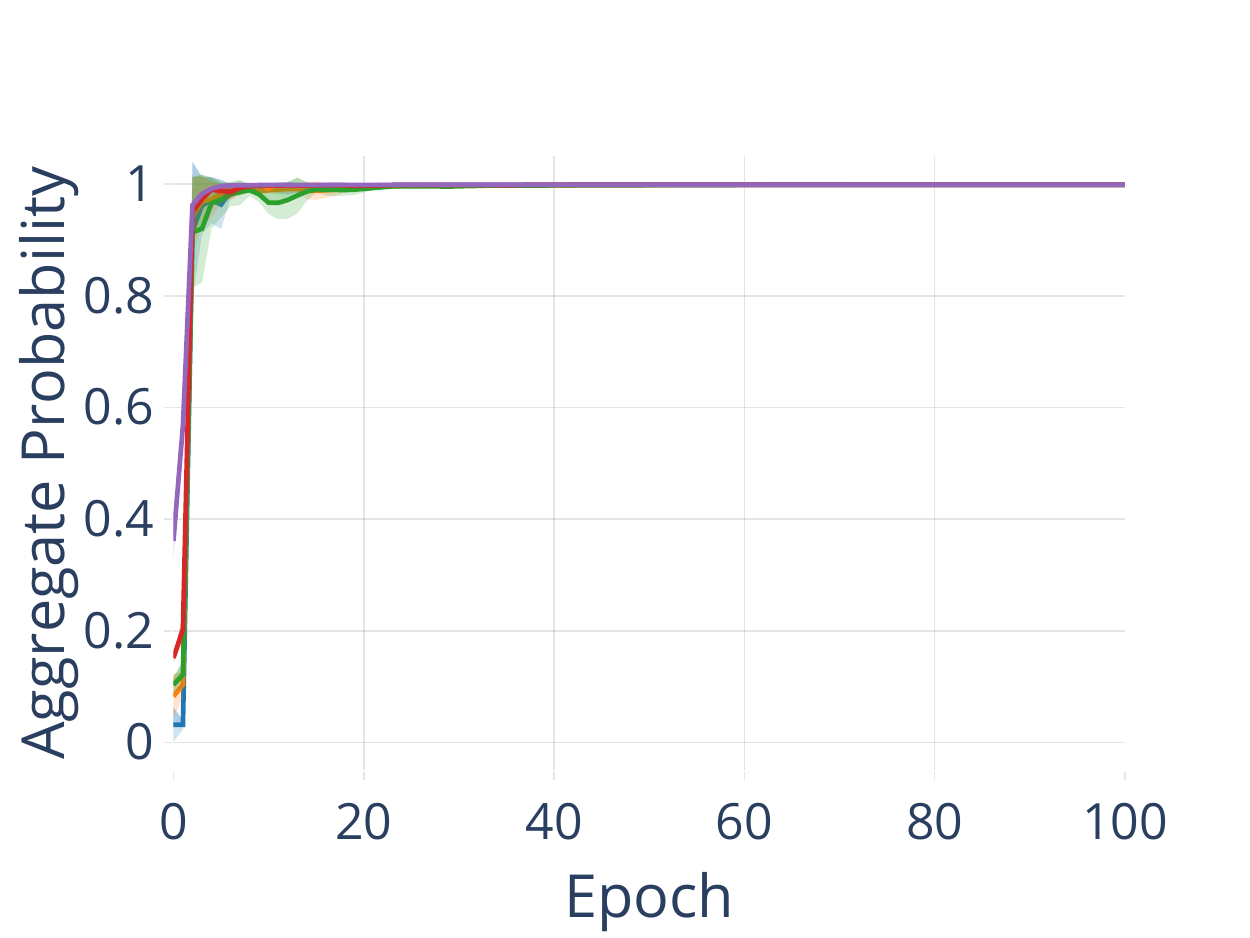}
    }
    \\
    \subfloat[Phi-1.3B]{
        \includegraphics[width=\allModelsWidth]{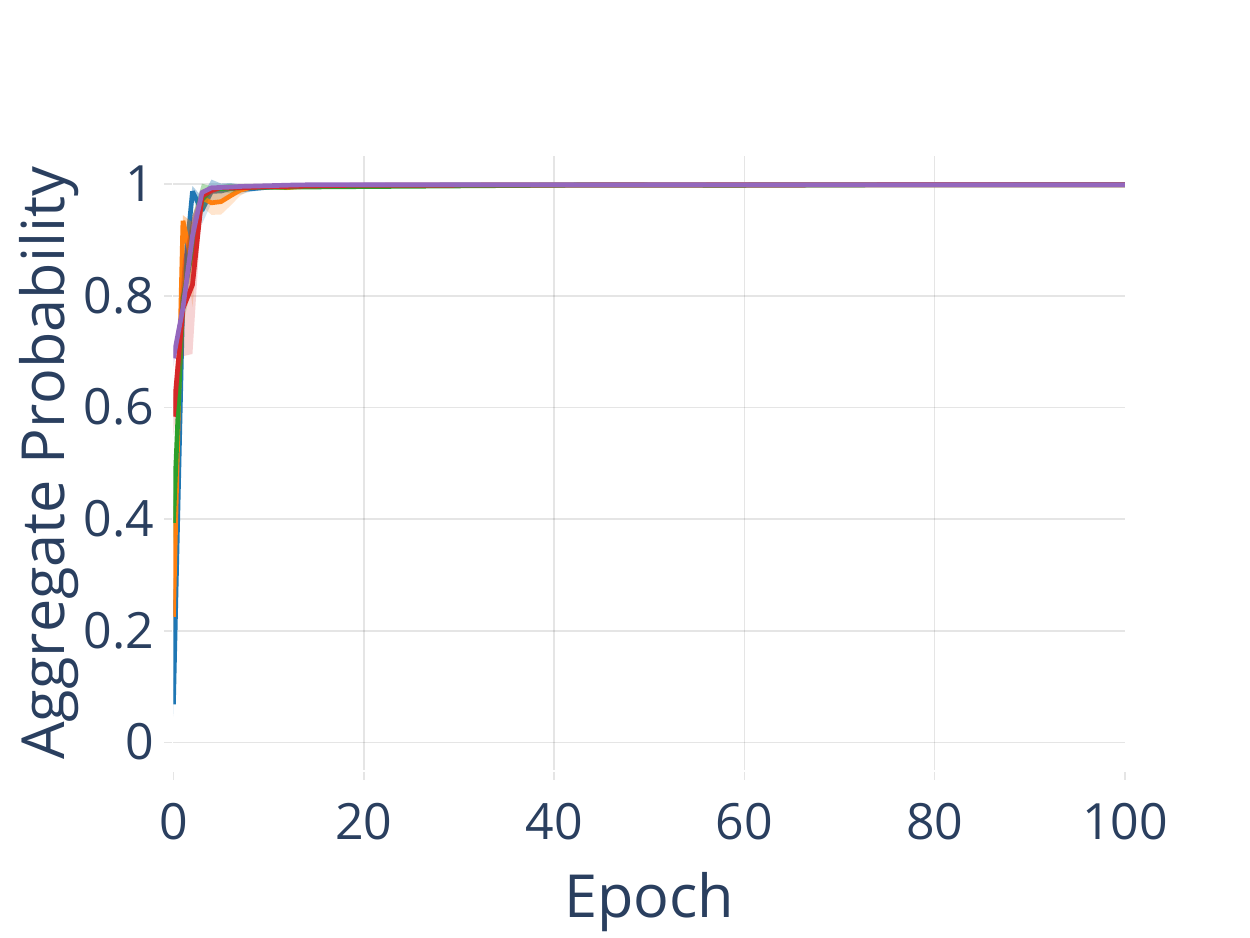}
    }
    \subfloat[Phi-2.7B]{
        \includegraphics[width=\allModelsWidth]{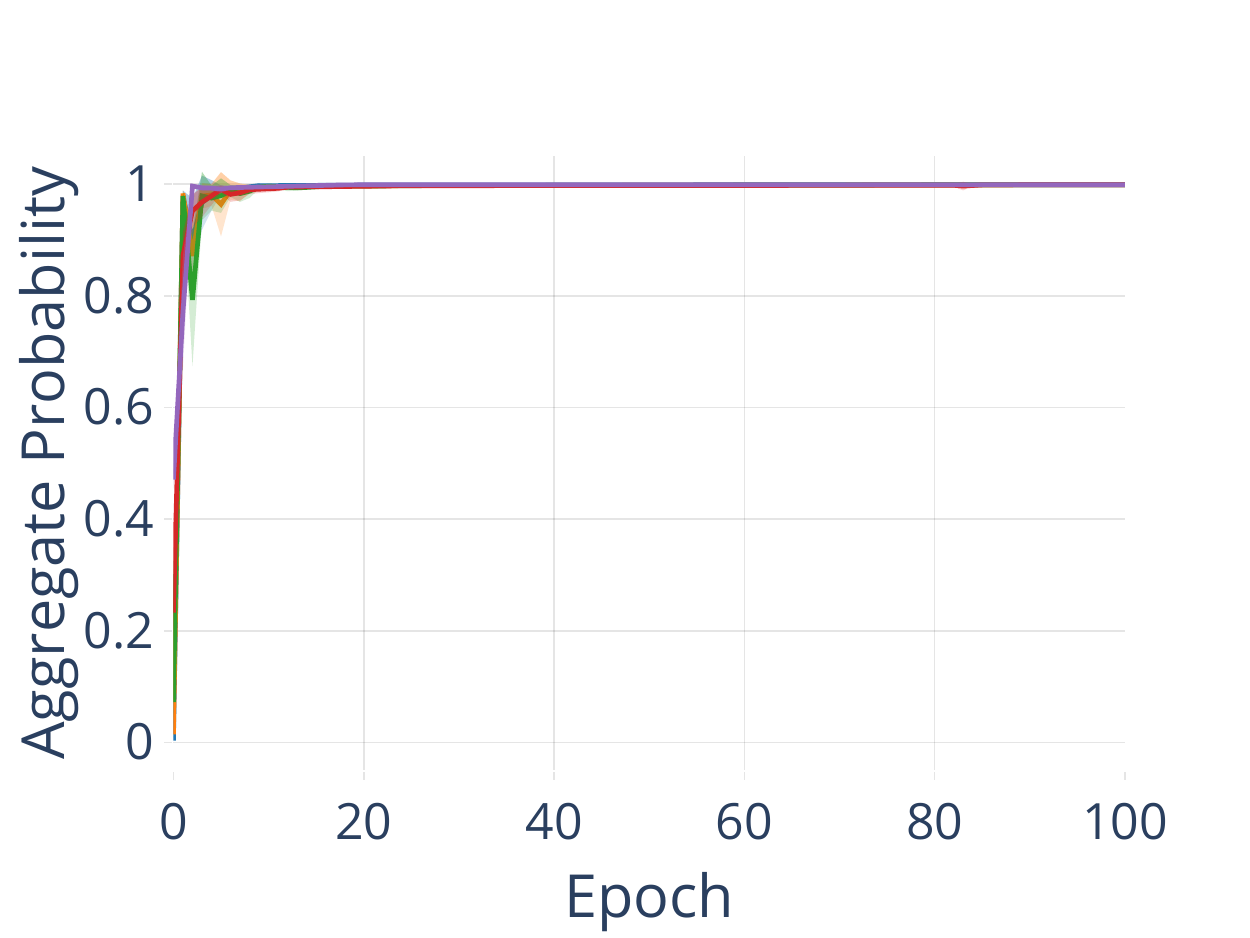}
    }
    \subfloat[Llama2-7B]{
        \includegraphics[width=\allModelsWidth]{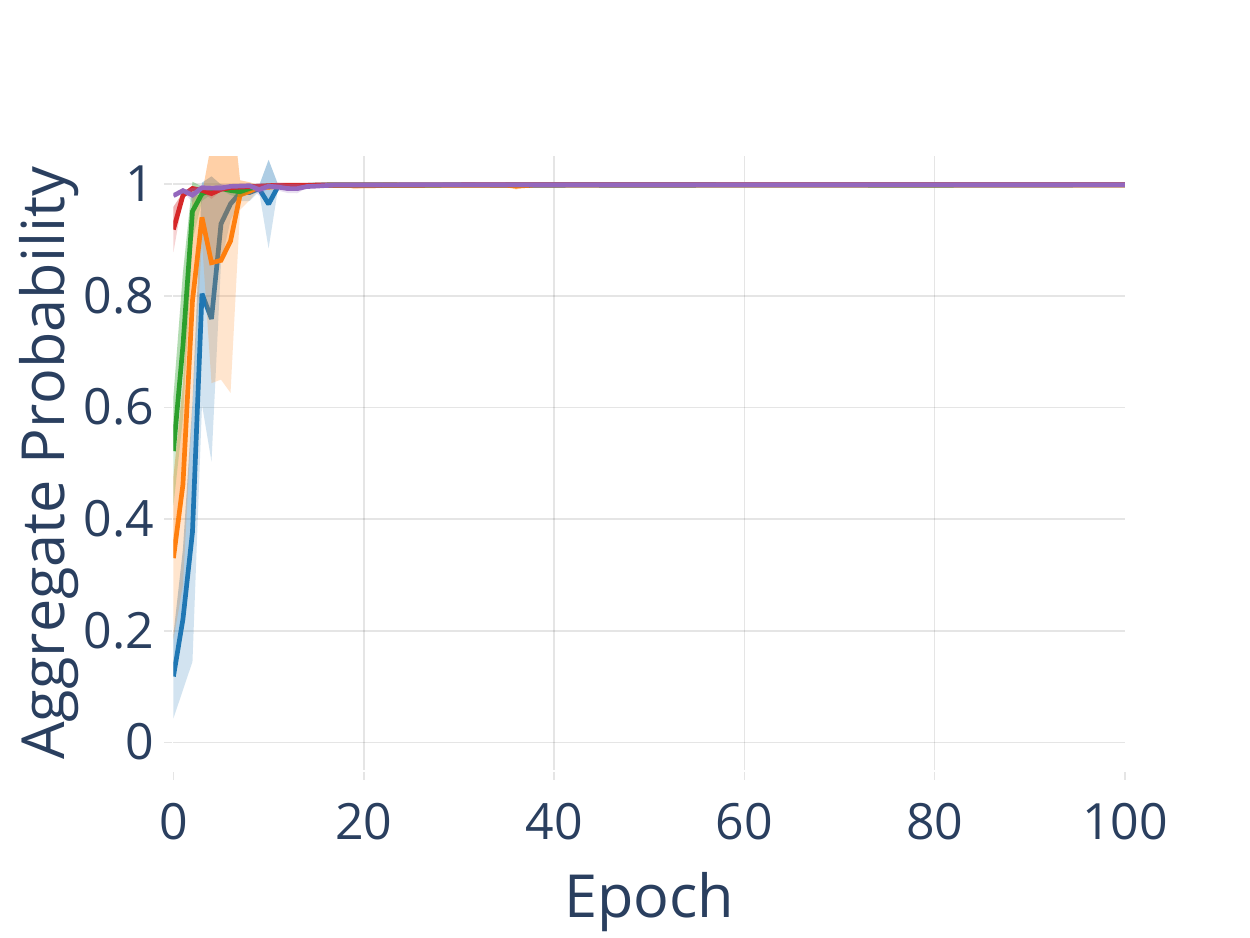}
    }
    \subfloat[Llama2-13B]{
        \includegraphics[width=\allModelsWidth]{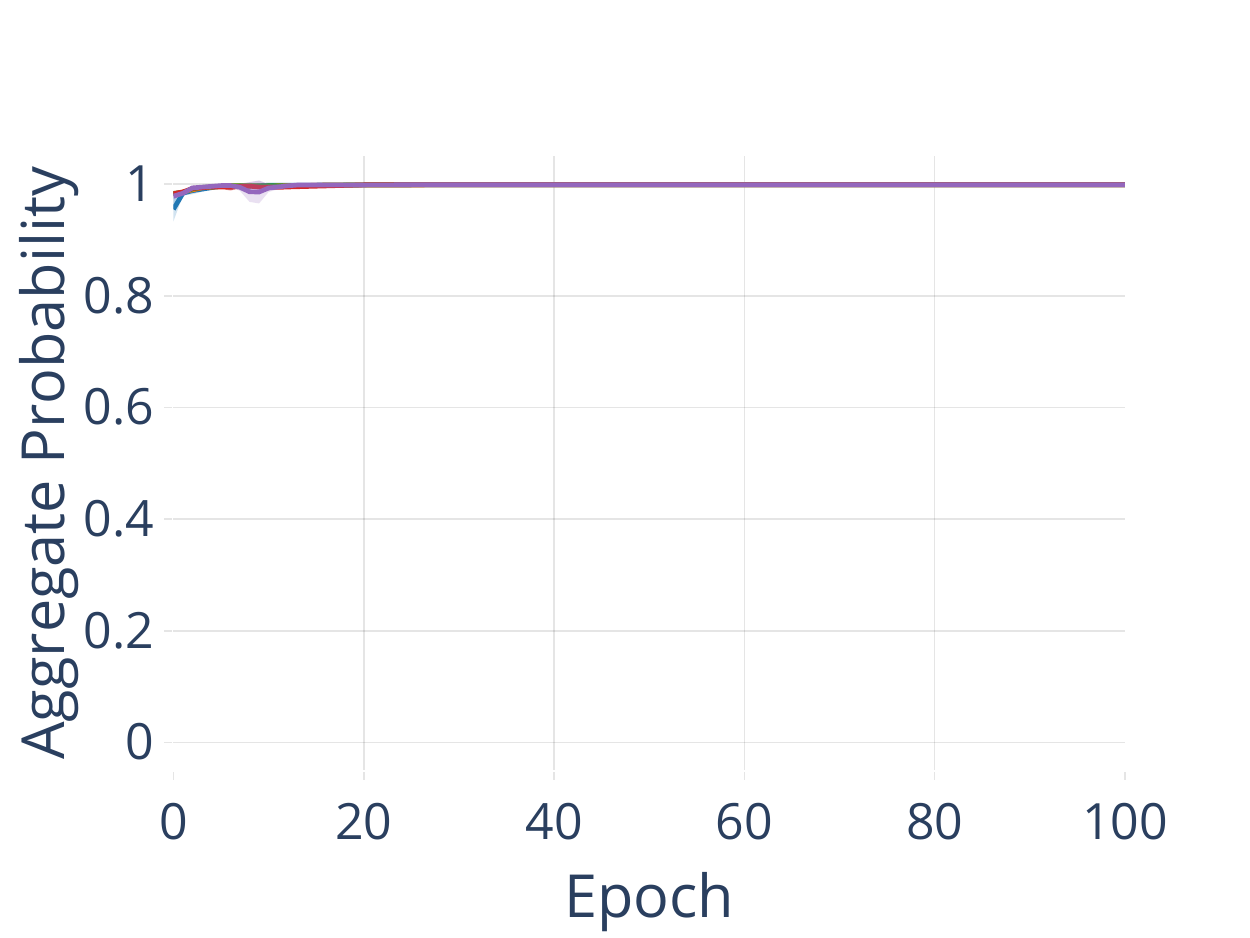}
    }
\caption{\capthead{Aggregate Probability over $A$ for all models for different $h$.}{$n = 1024, \ell = 26$}
}
\label{fig:cum-prob_entropy_level_all}
\end{figure}

\begin{figure}[H]
    \centering
    \subfloat[Pythia-70M]{
        \includegraphics[width=\allModelsWidth]{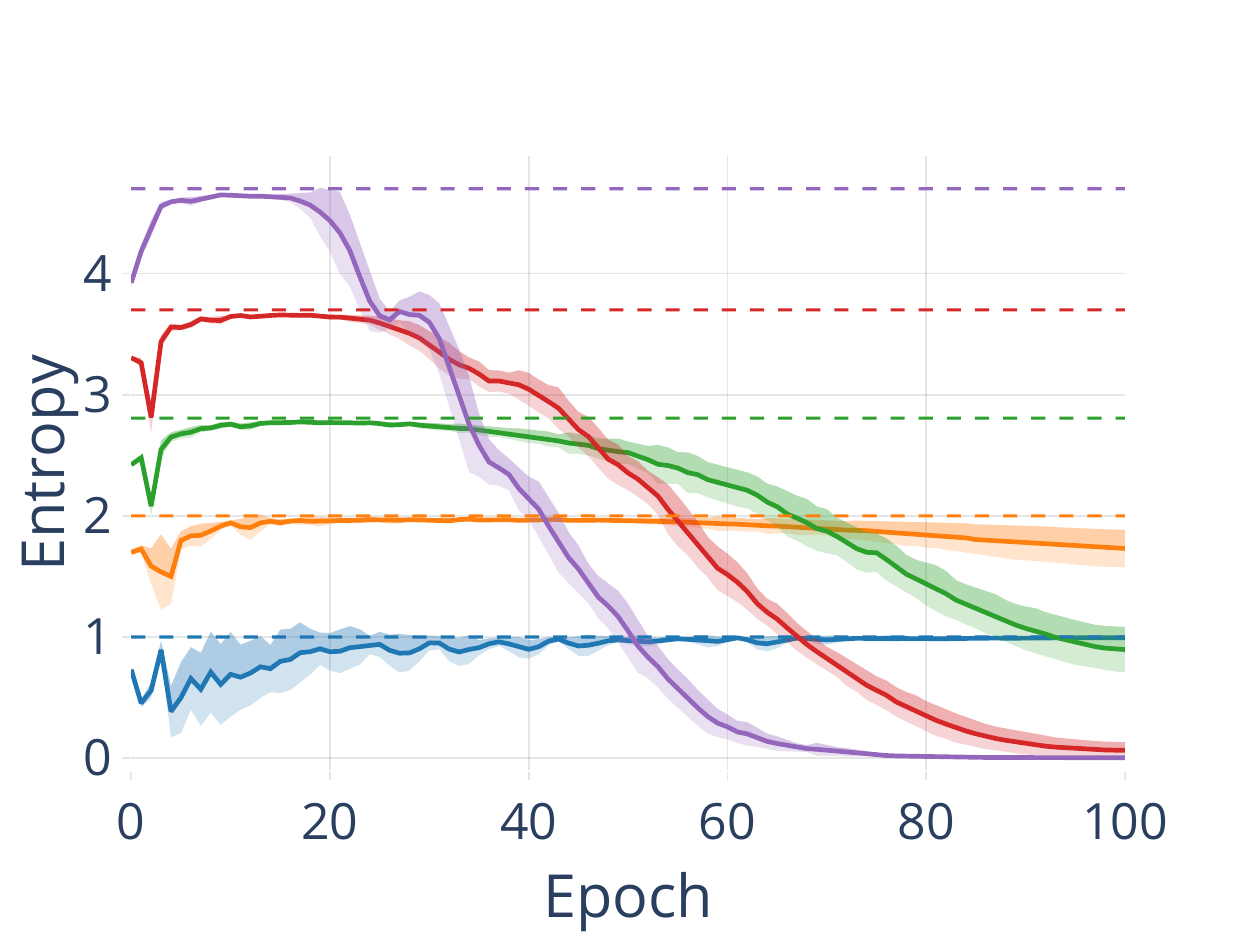}
    }
    \subfloat[Pythia-1B]{
        \includegraphics[width=\allModelsWidth]{figures/memorability/alphabet_size/entropy_alphabet-size_pythia-1b.pdf}
    }
    \subfloat[Pythia-12B]{
        \includegraphics[width=\allModelsWidth]{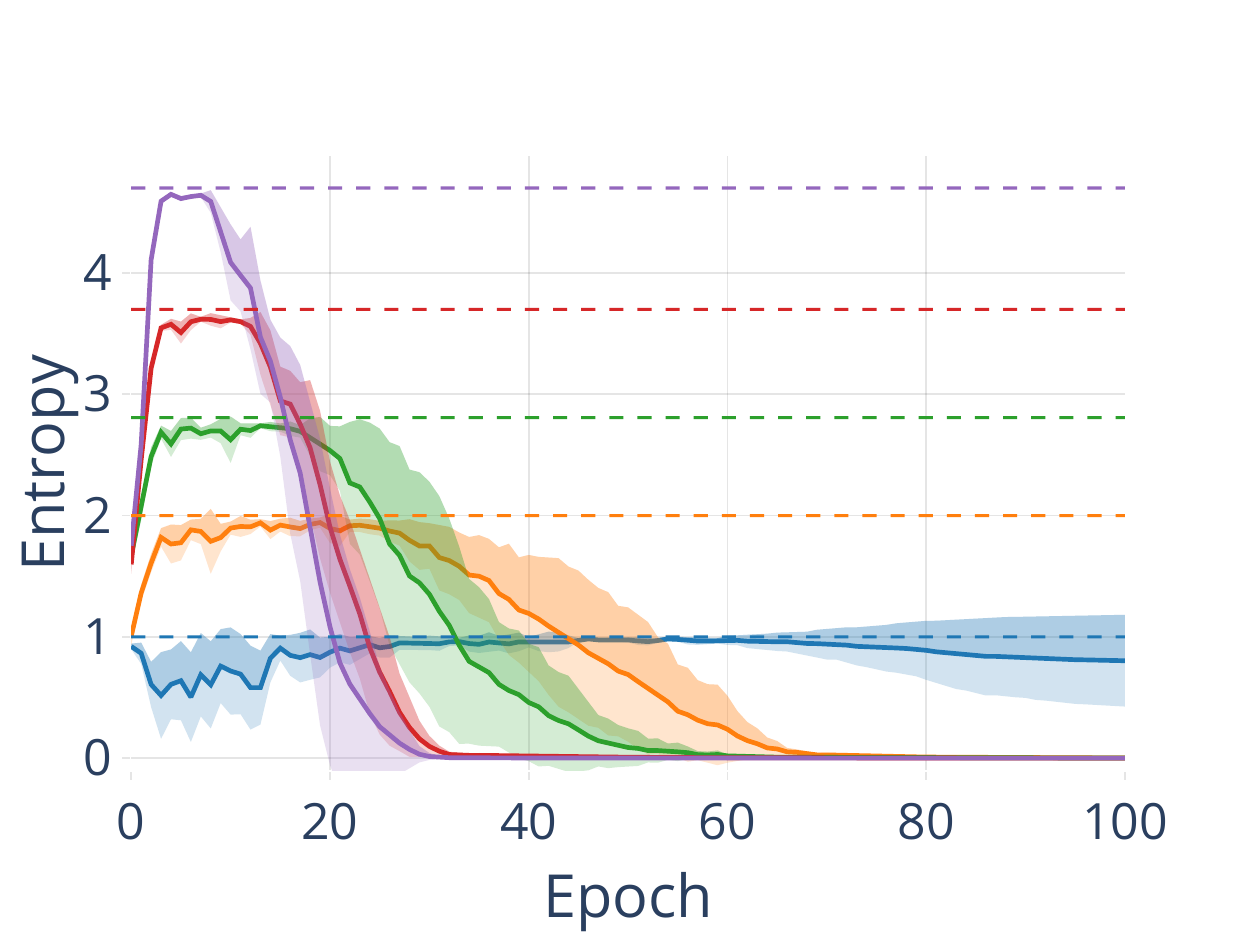}
    }
    \\
    \subfloat[Phi-1.3B]{
        \includegraphics[width=\allModelsWidth]{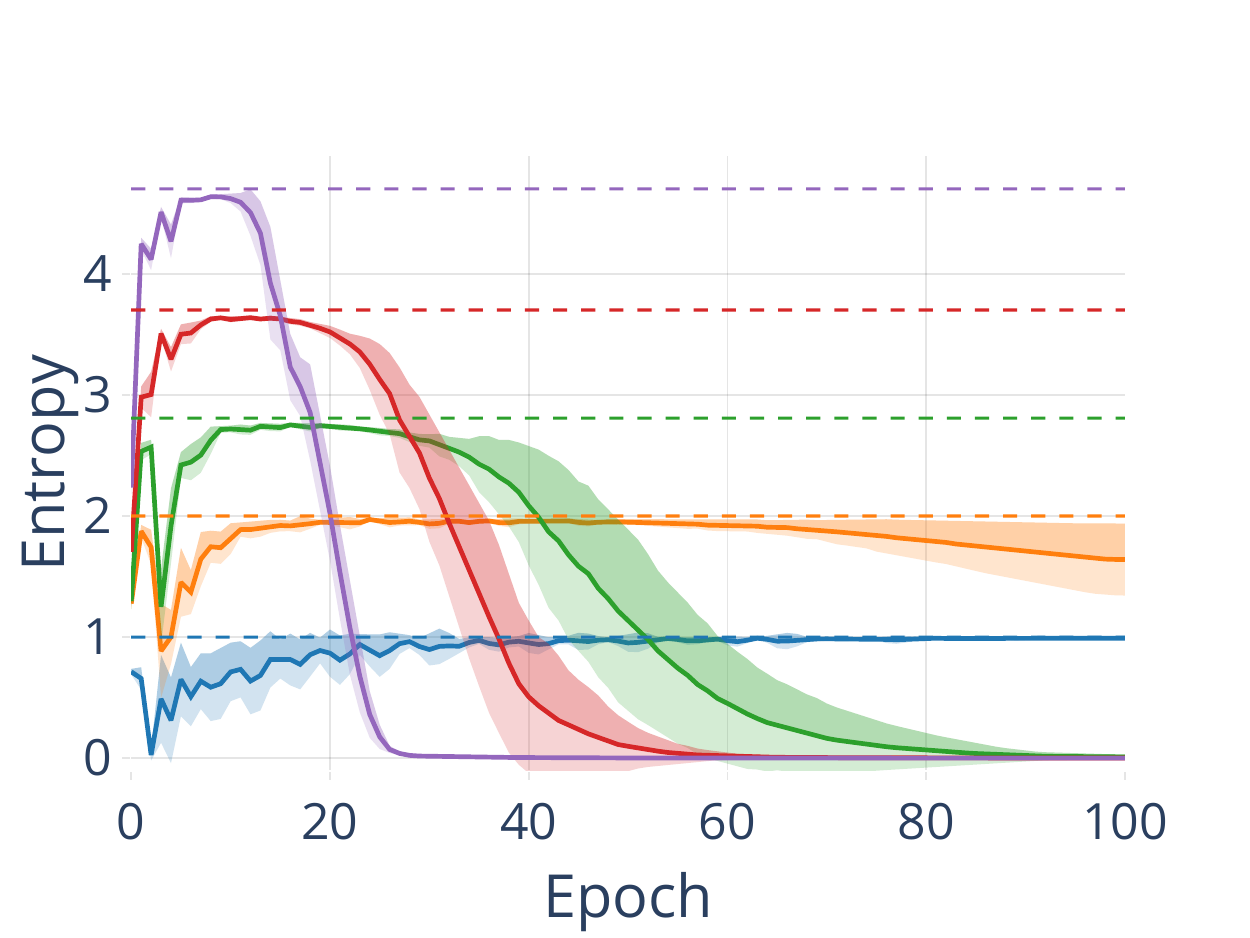}
    }
    \subfloat[Phi-2.7B]{
        \includegraphics[width=\allModelsWidth]{figures/memorability/alphabet_size/entropy_alphabet-size_phi-2.7b.pdf}
    }
    \subfloat[Llama2-7B]{
        \includegraphics[width=\allModelsWidth]{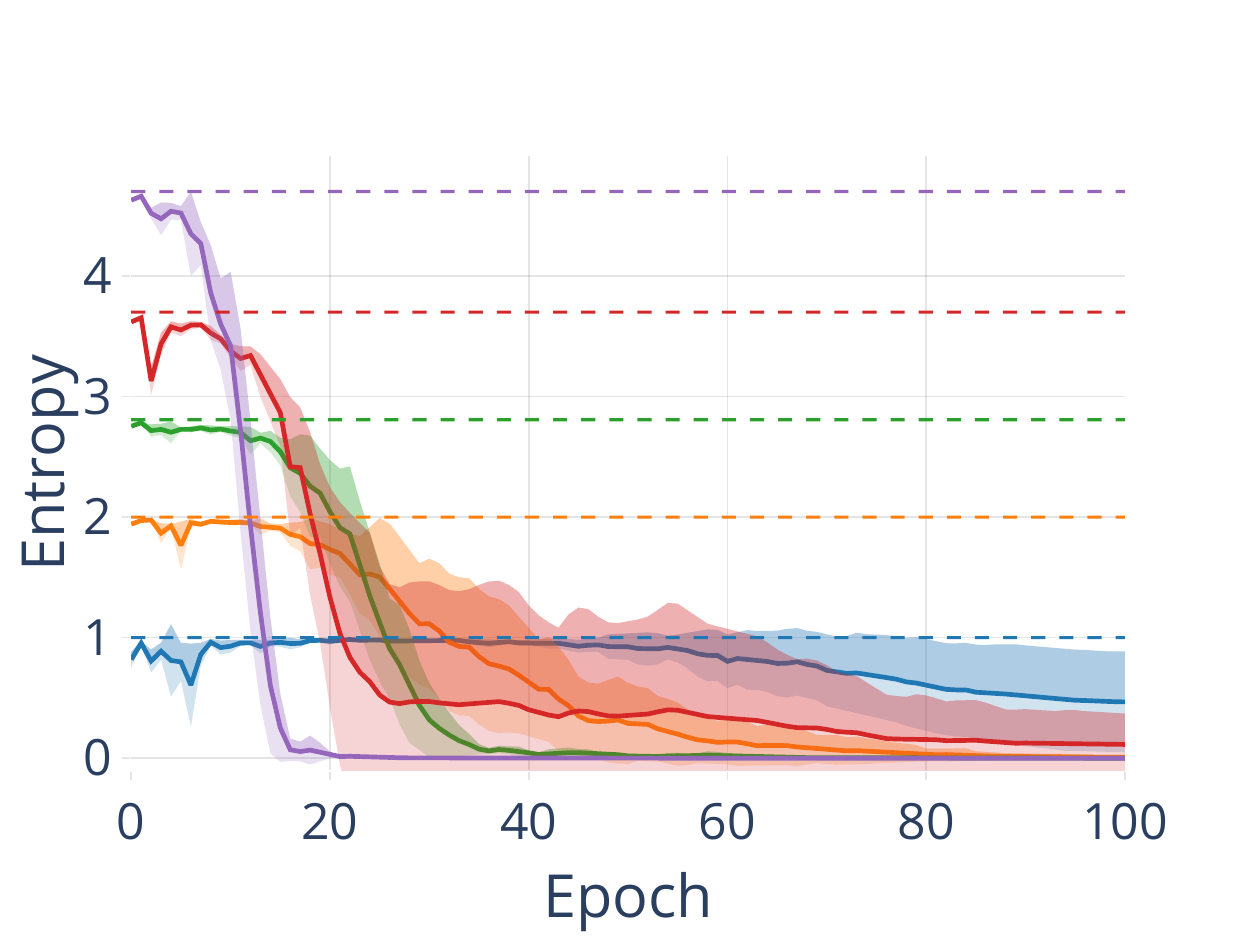}
    }
    \subfloat[Llama2-13B]{
        \includegraphics[width=\allModelsWidth]{figures/memorability/alphabet_size/entropy_alphabet-size_llama2-13b.pdf}
    }
\caption{\capthead{Entropy for all models for different $\ell$.}{$n = 1024$}
}
\label{fig:entropy_alphabet_size_all}
\end{figure}

\begin{figure}[H]
    \centering
    \subfloat[Pythia-70M]{
        \includegraphics[width=\allModelsWidth]{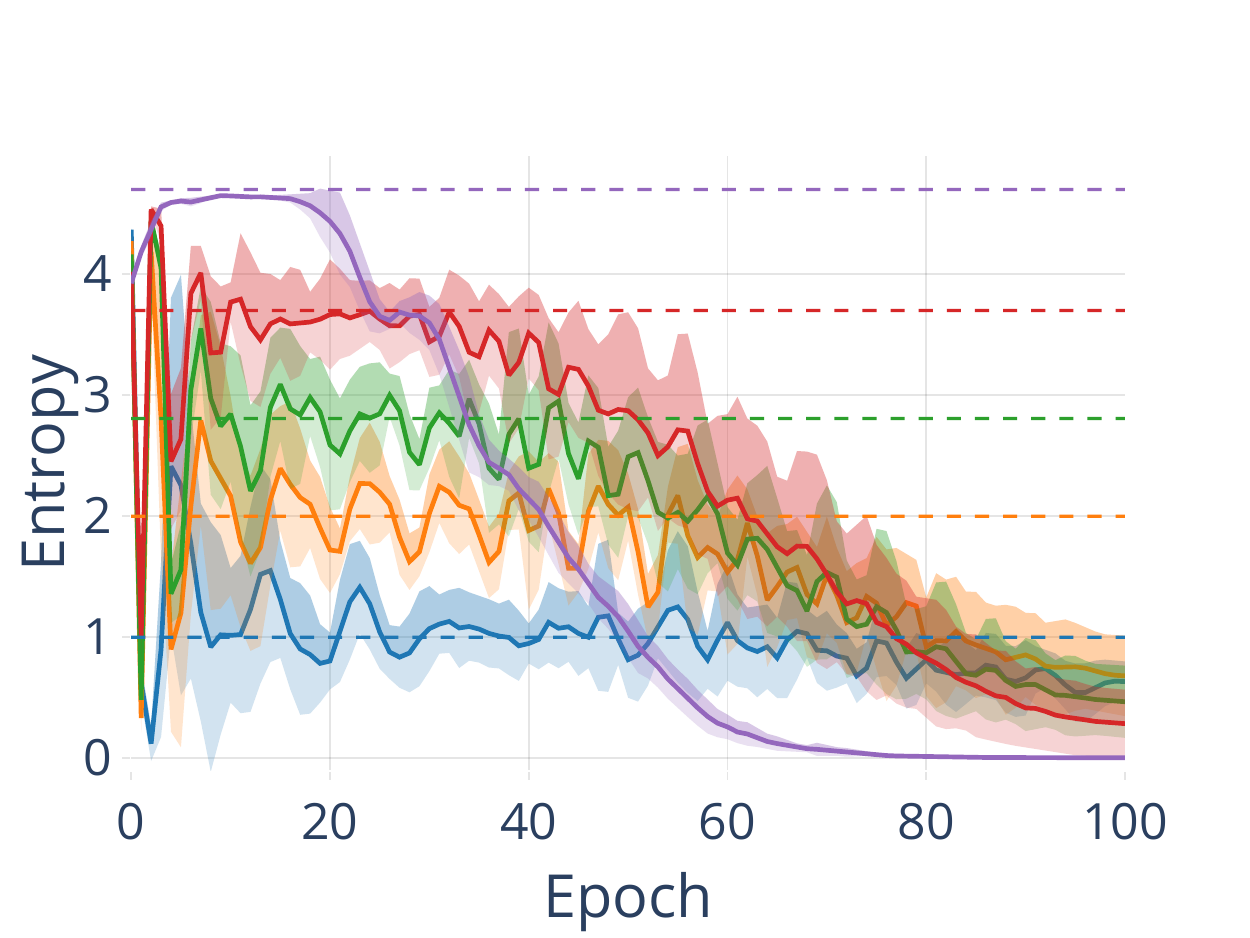}
    }
    \subfloat[Pythia-1B]{
        \includegraphics[width=\allModelsWidth]{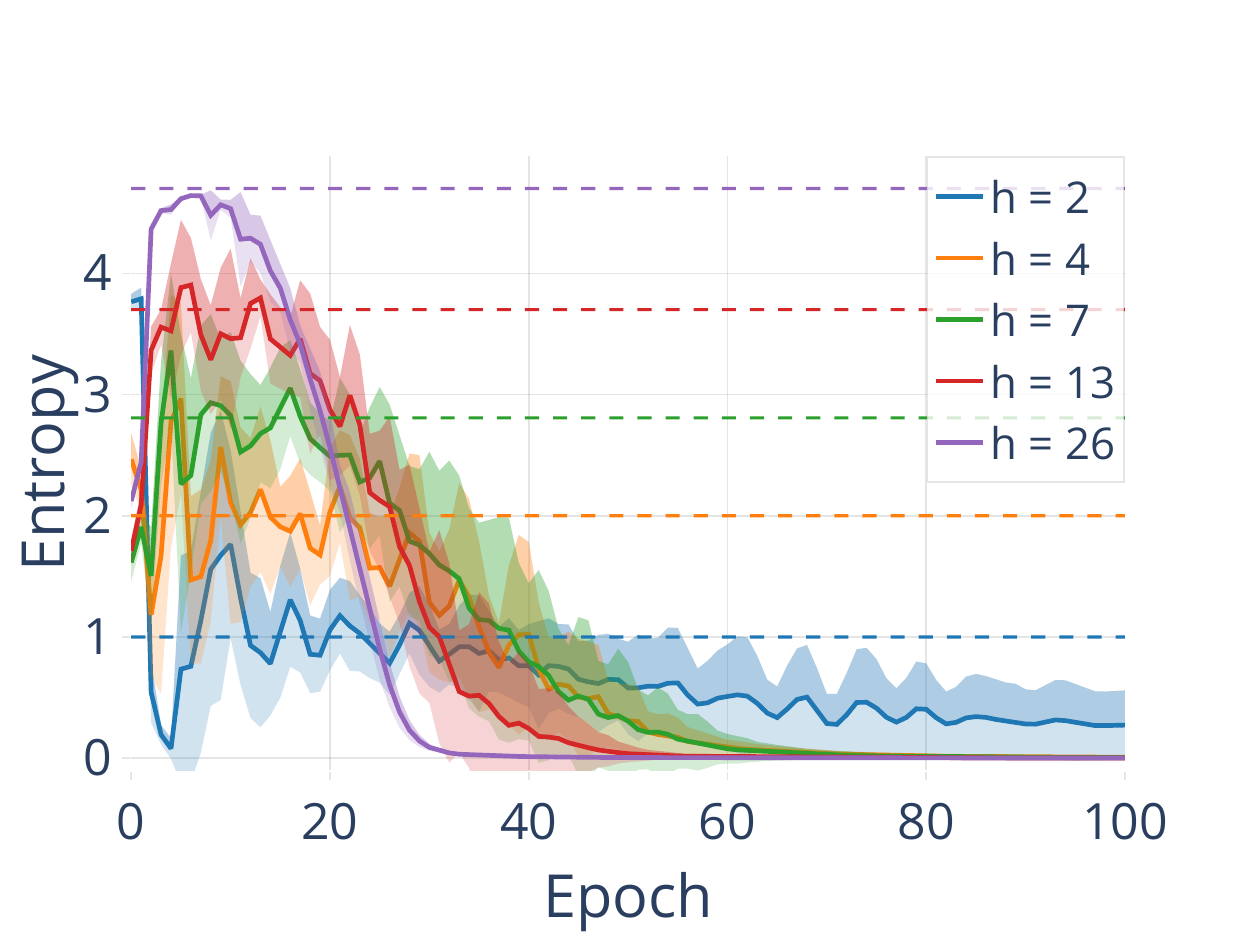}
    }
    \subfloat[Pythia-12B]{
        \includegraphics[width=\allModelsWidth]{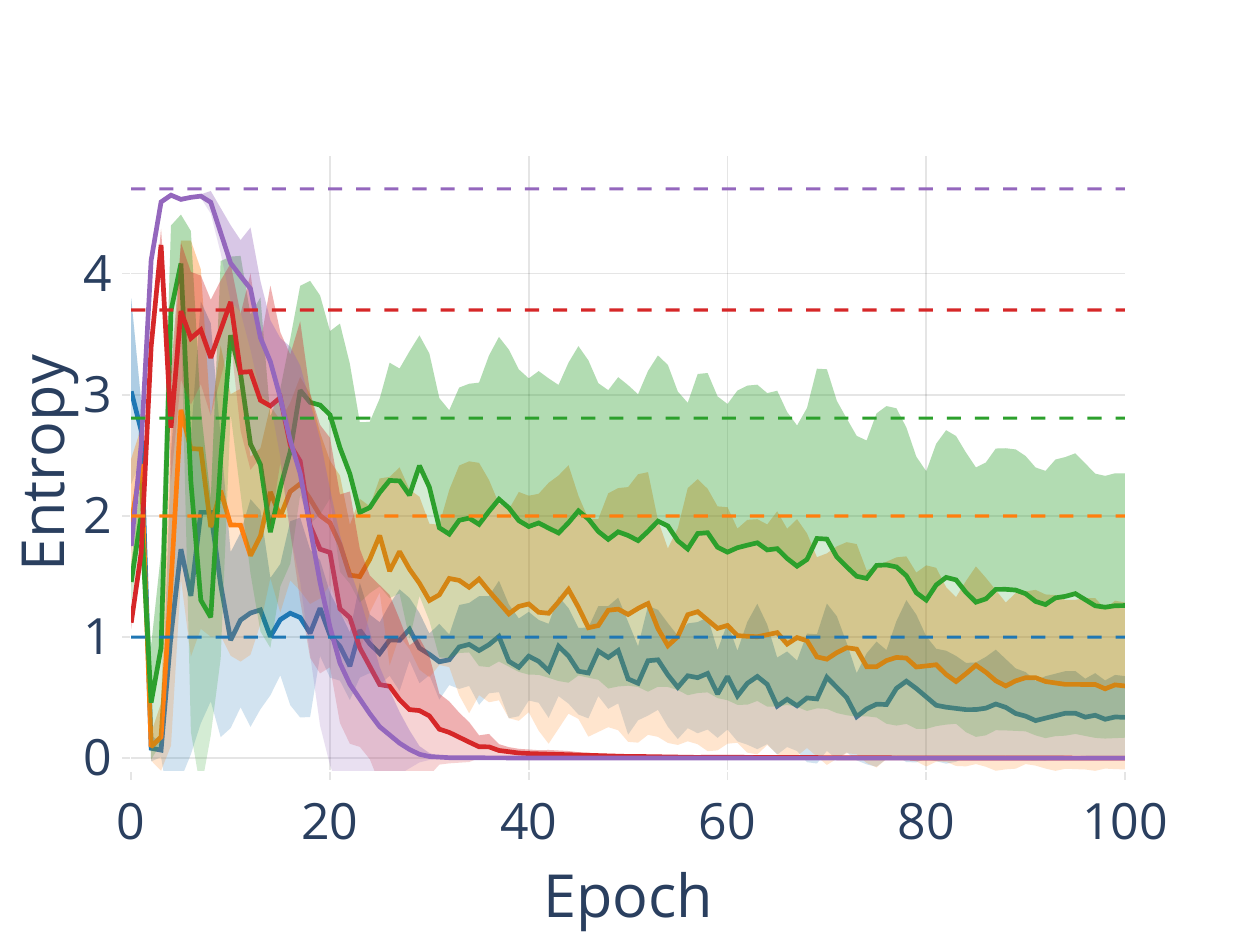}
    }
    \\
    \subfloat[Phi-1.3B]{
        \includegraphics[width=\allModelsWidth]{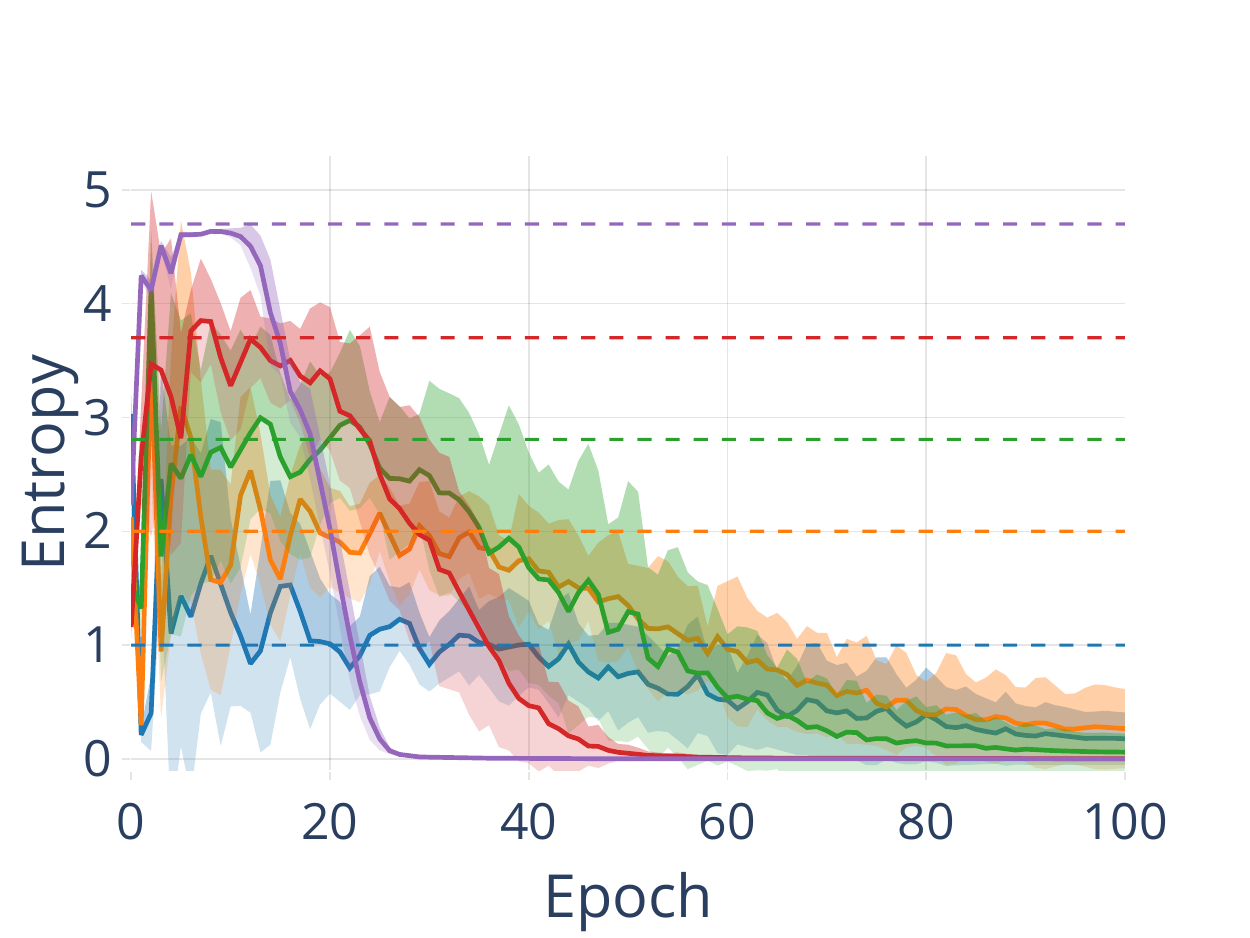}
    }
    \subfloat[Phi-2.7B]{
        \includegraphics[width=\allModelsWidth]{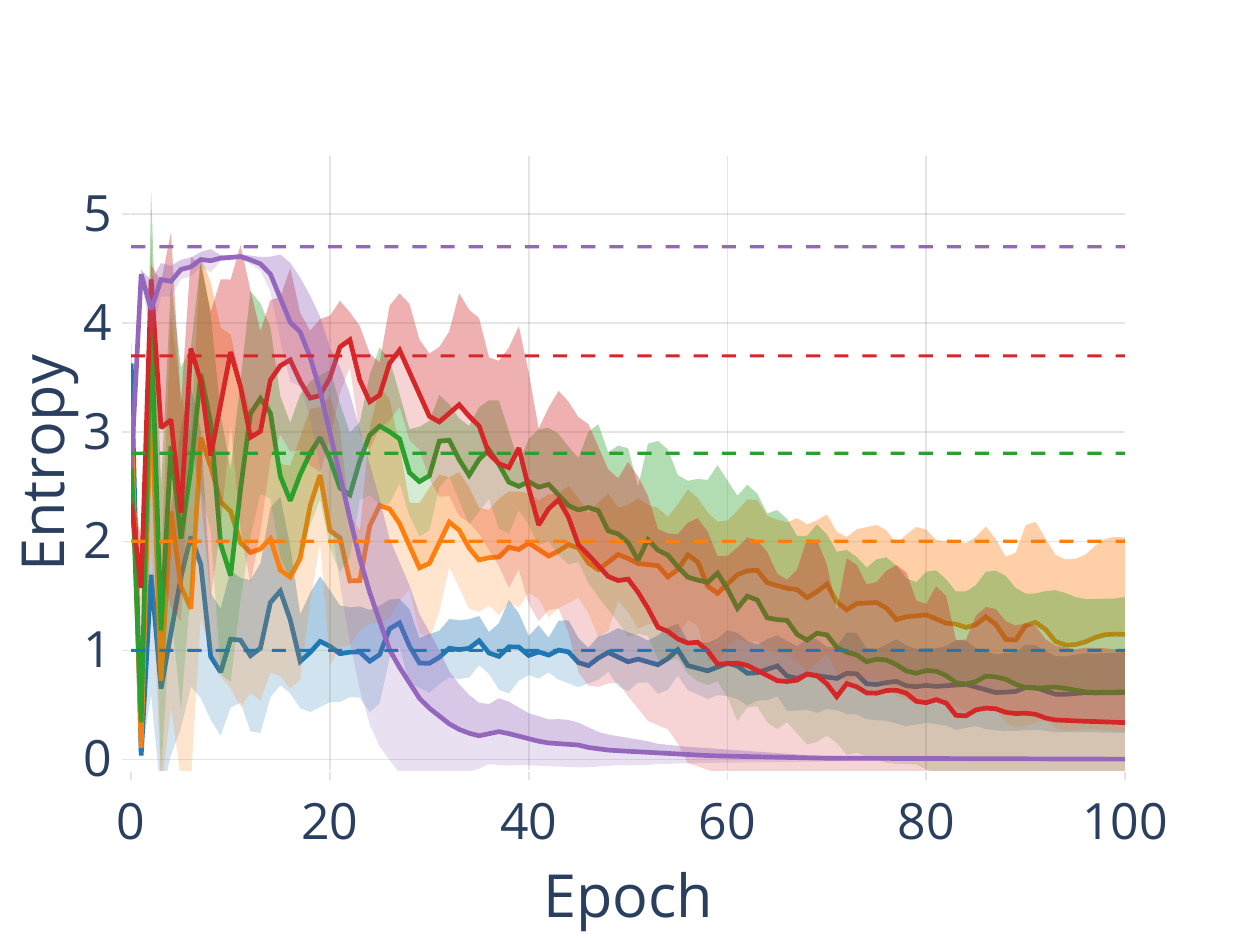}
    }
    \subfloat[Llama2-7B]{
        \includegraphics[width=\allModelsWidth]{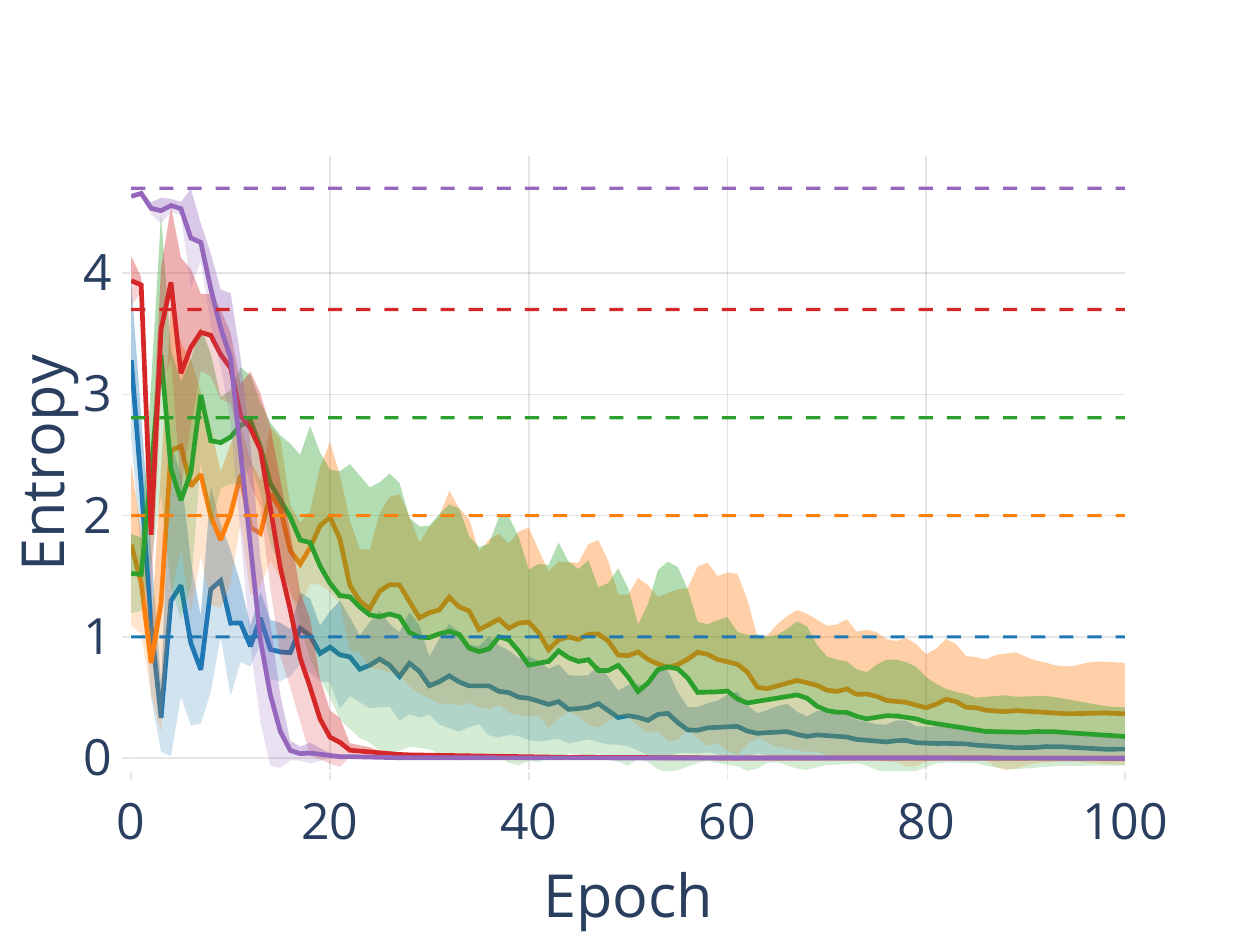}
    }
    \subfloat[Llama2-13B]{
        \includegraphics[width=\allModelsWidth]{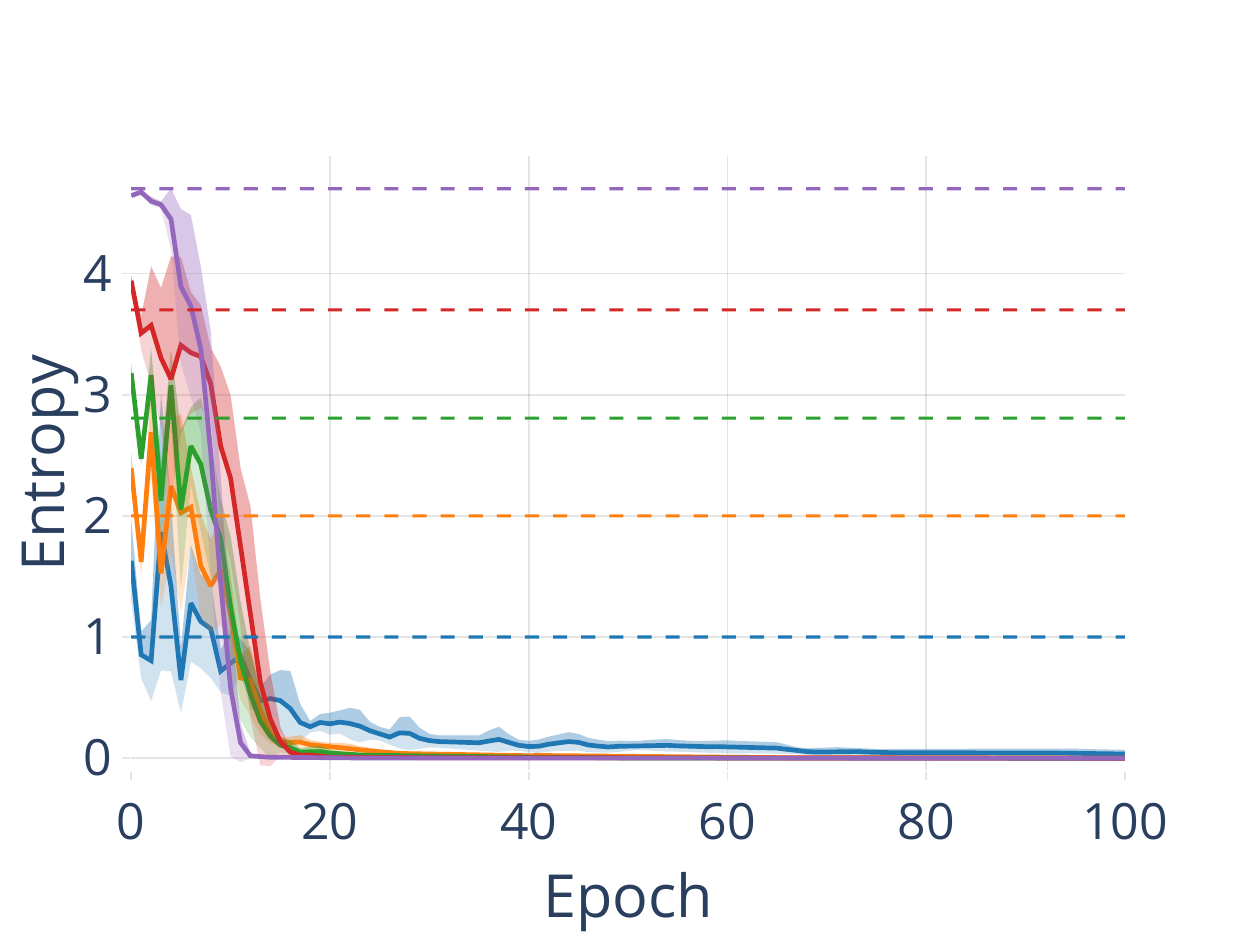}
    }
\caption{\capthead{Entropy for all models for different $h$.}{$n = 1024, \ell = 26$}
}
\label{fig:entropy_entropy_level_all}
\end{figure}

\begin{figure}[H]
    \centering
    \subfloat[Pythia-70M]{
        \includegraphics[width=\allModelsWidth]{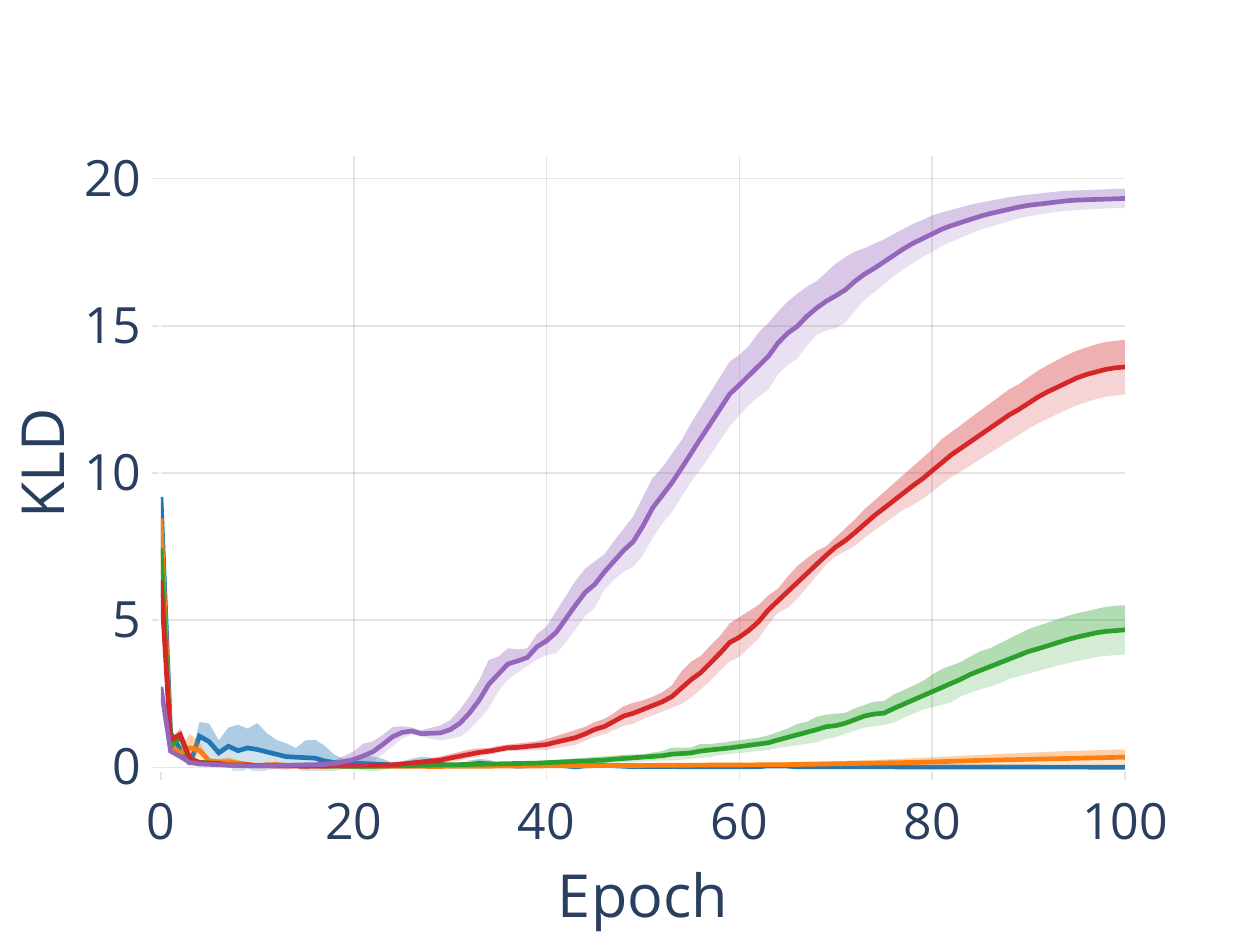}
    }
    \subfloat[Pythia-1B]{
        \includegraphics[width=\allModelsWidth]{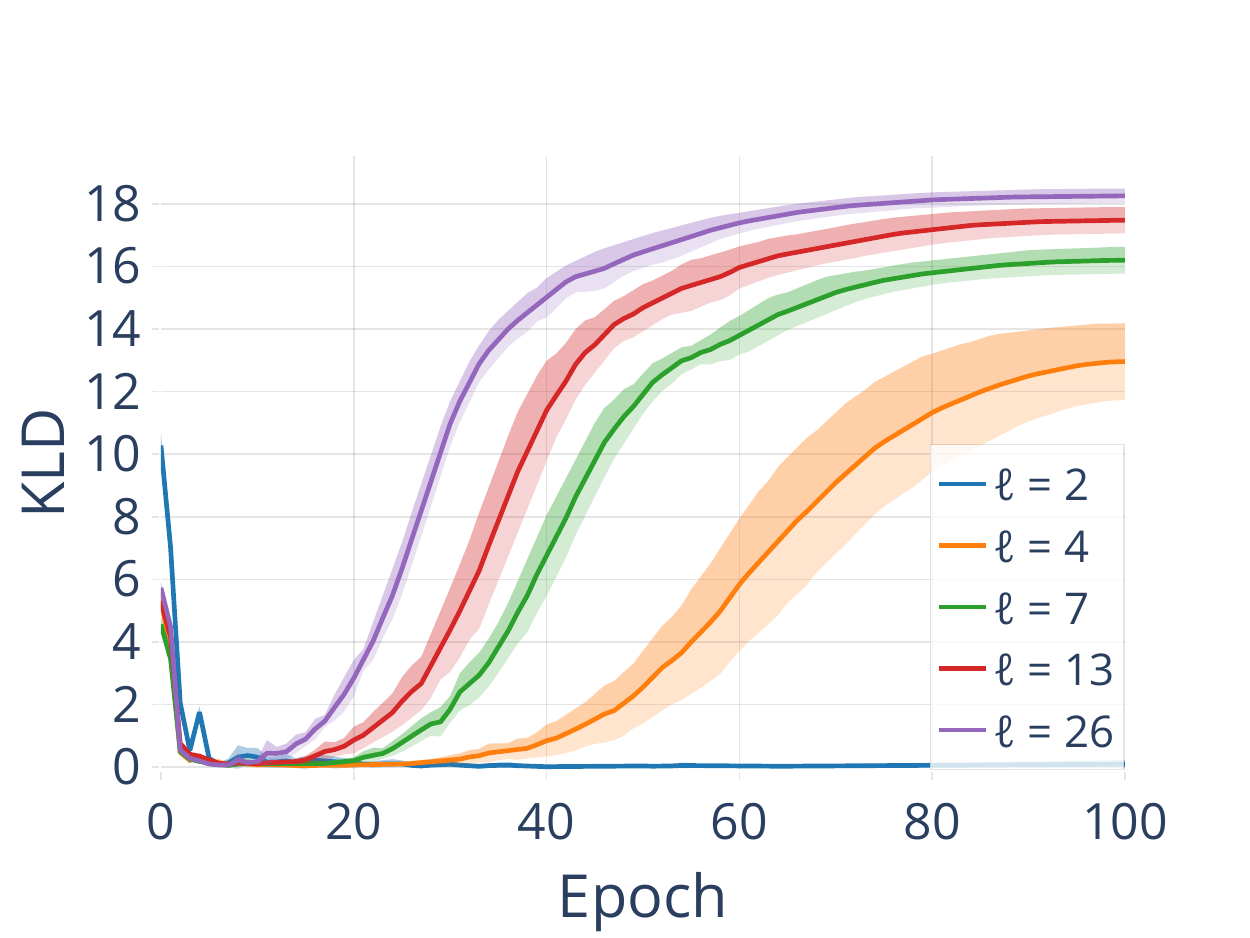}
    }
    \subfloat[Pythia-12B]{
        \includegraphics[width=\allModelsWidth]{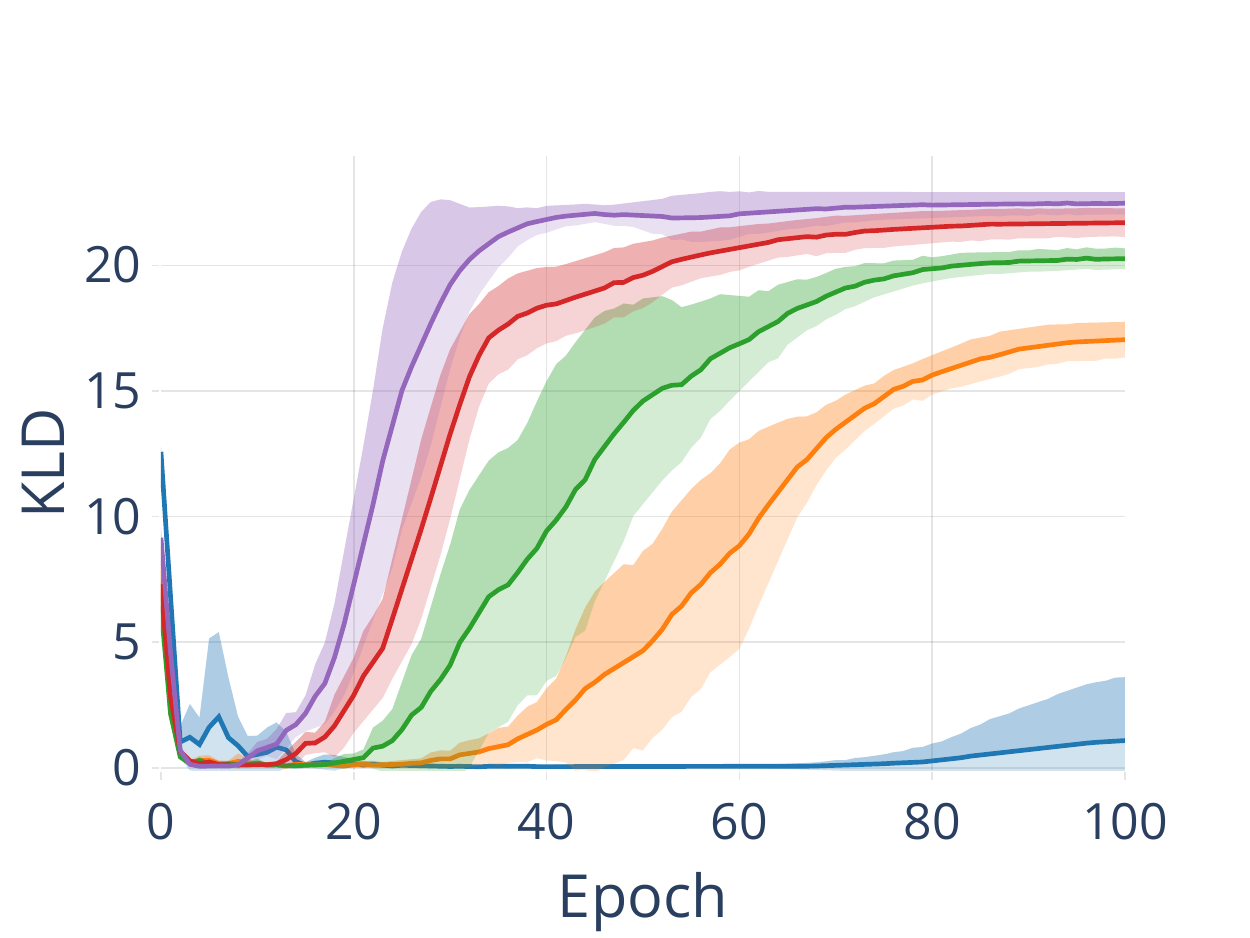}
    }
    \\
    \subfloat[Phi-1.3B]{
        \includegraphics[width=\allModelsWidth]{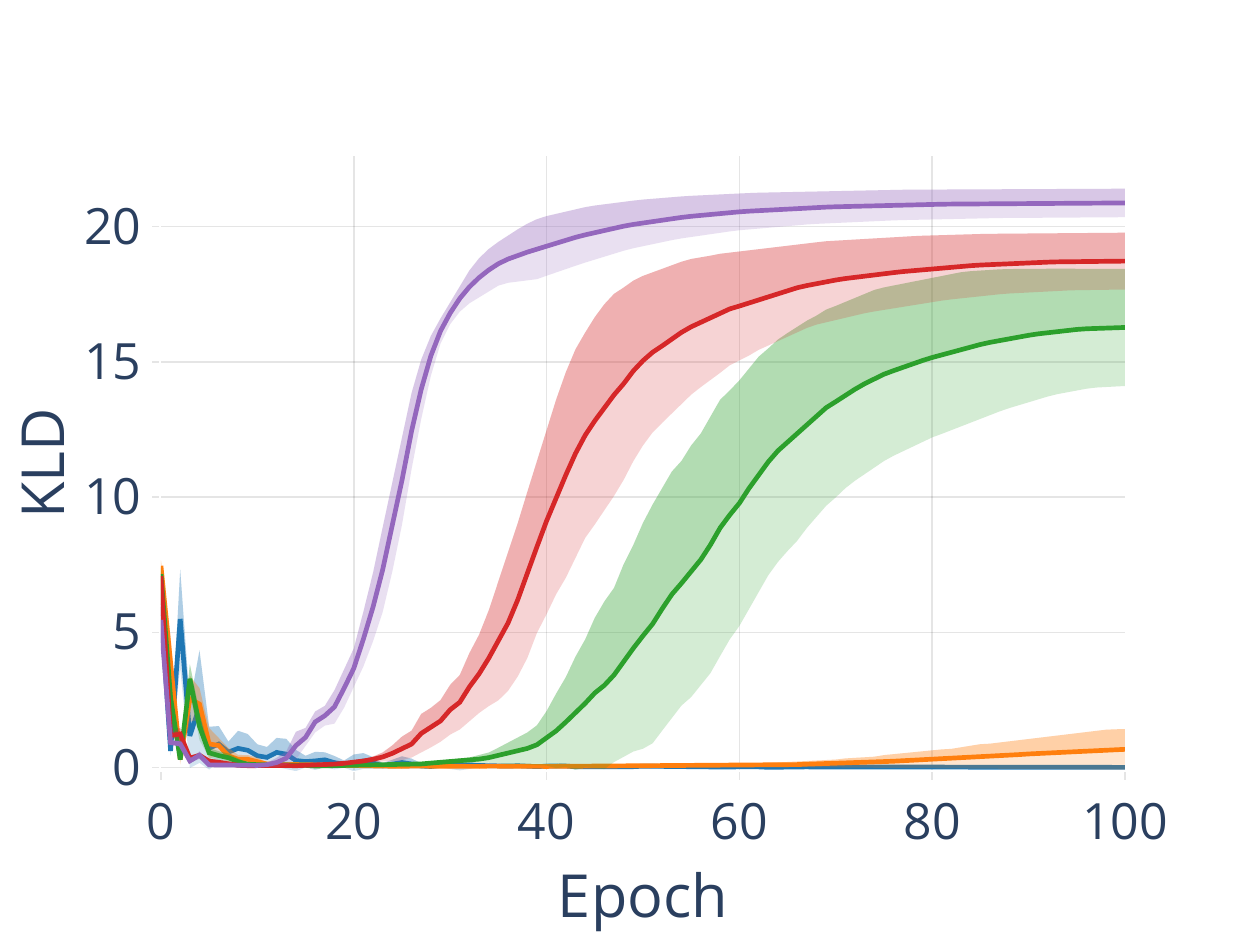}
    }
    \subfloat[Phi-2.7B]{
        \includegraphics[width=\allModelsWidth]{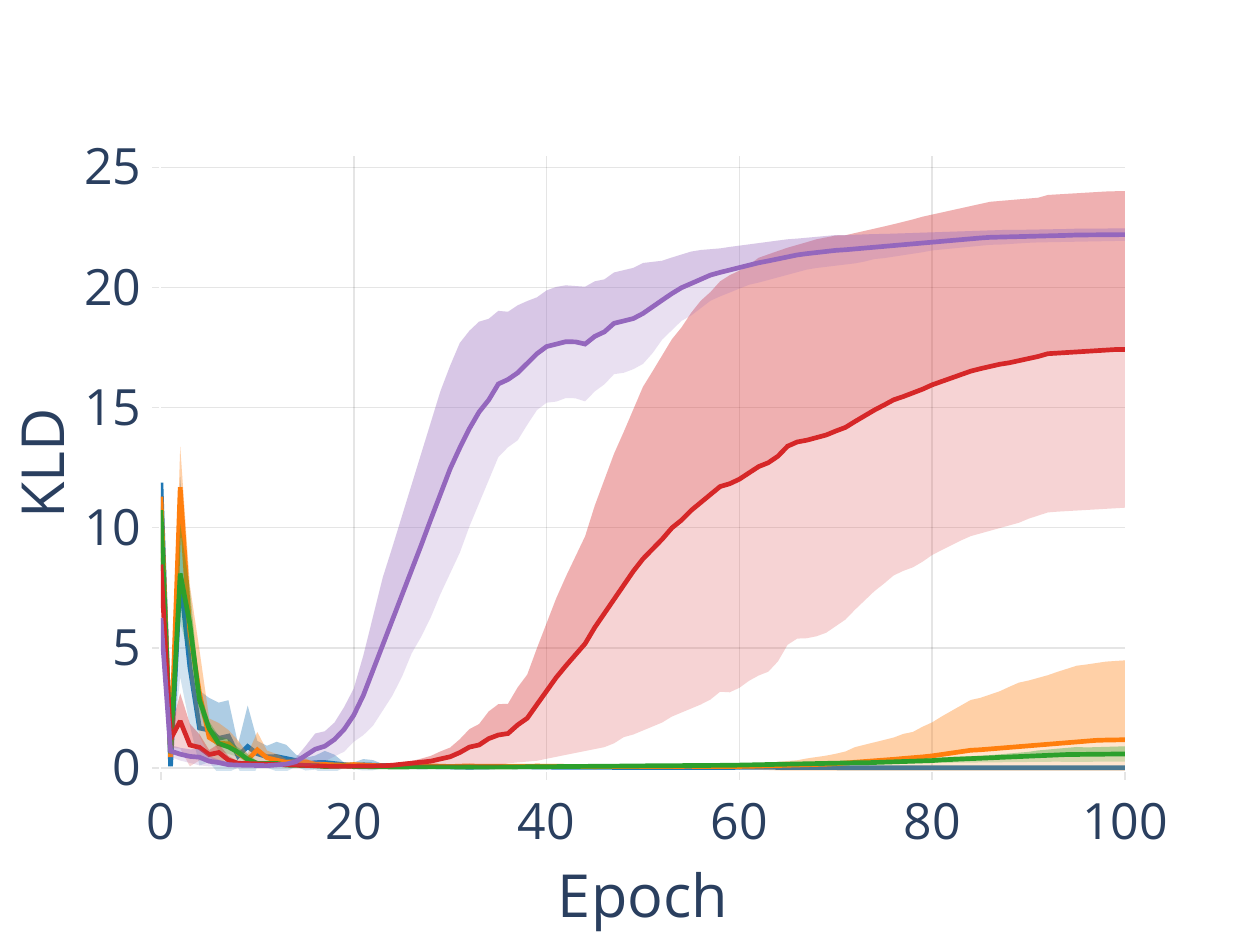}
    }
    \subfloat[Llama2-7B]{
        \includegraphics[width=\allModelsWidth]{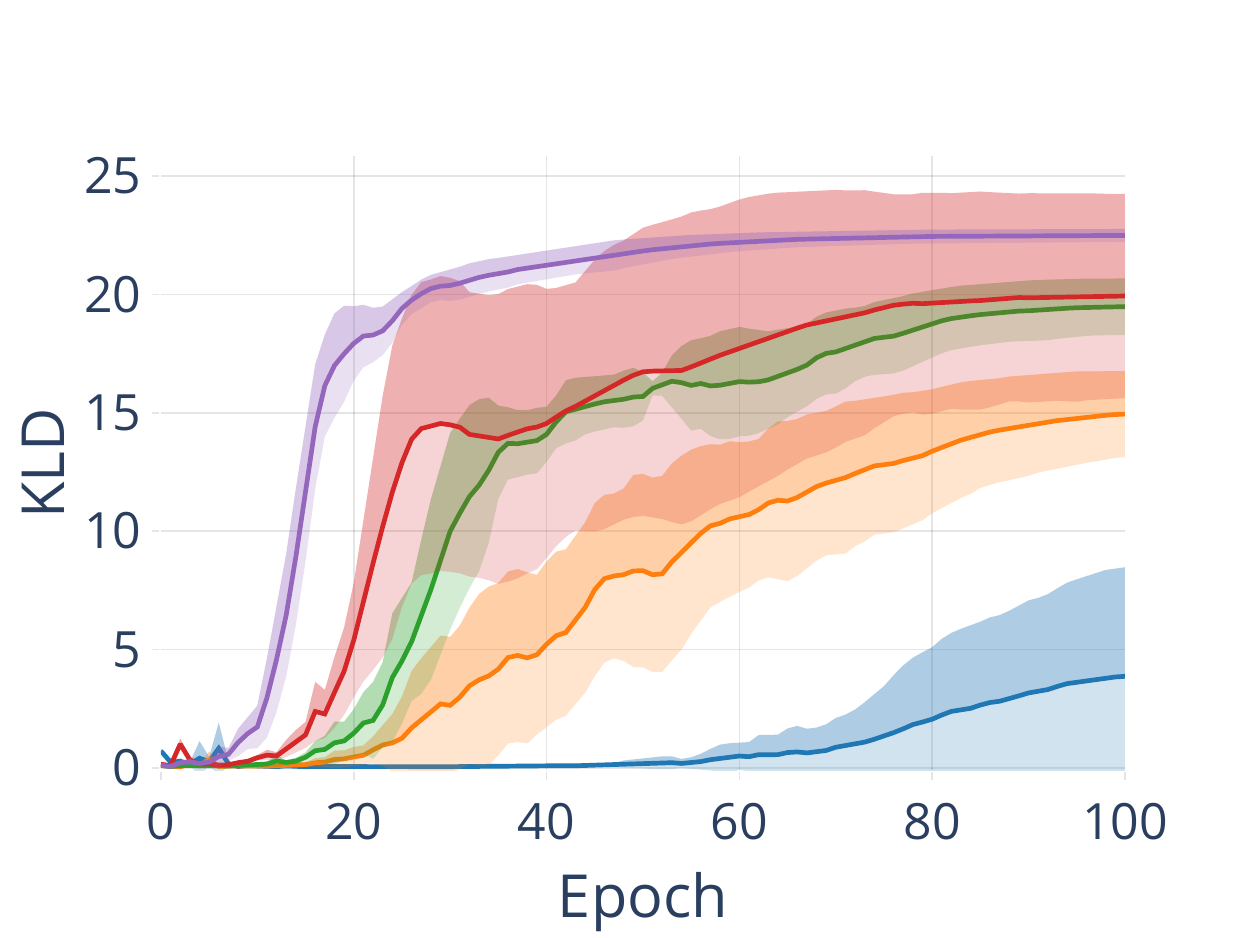}
    }
    \subfloat[Llama2-13B]{
        \includegraphics[width=\allModelsWidth]{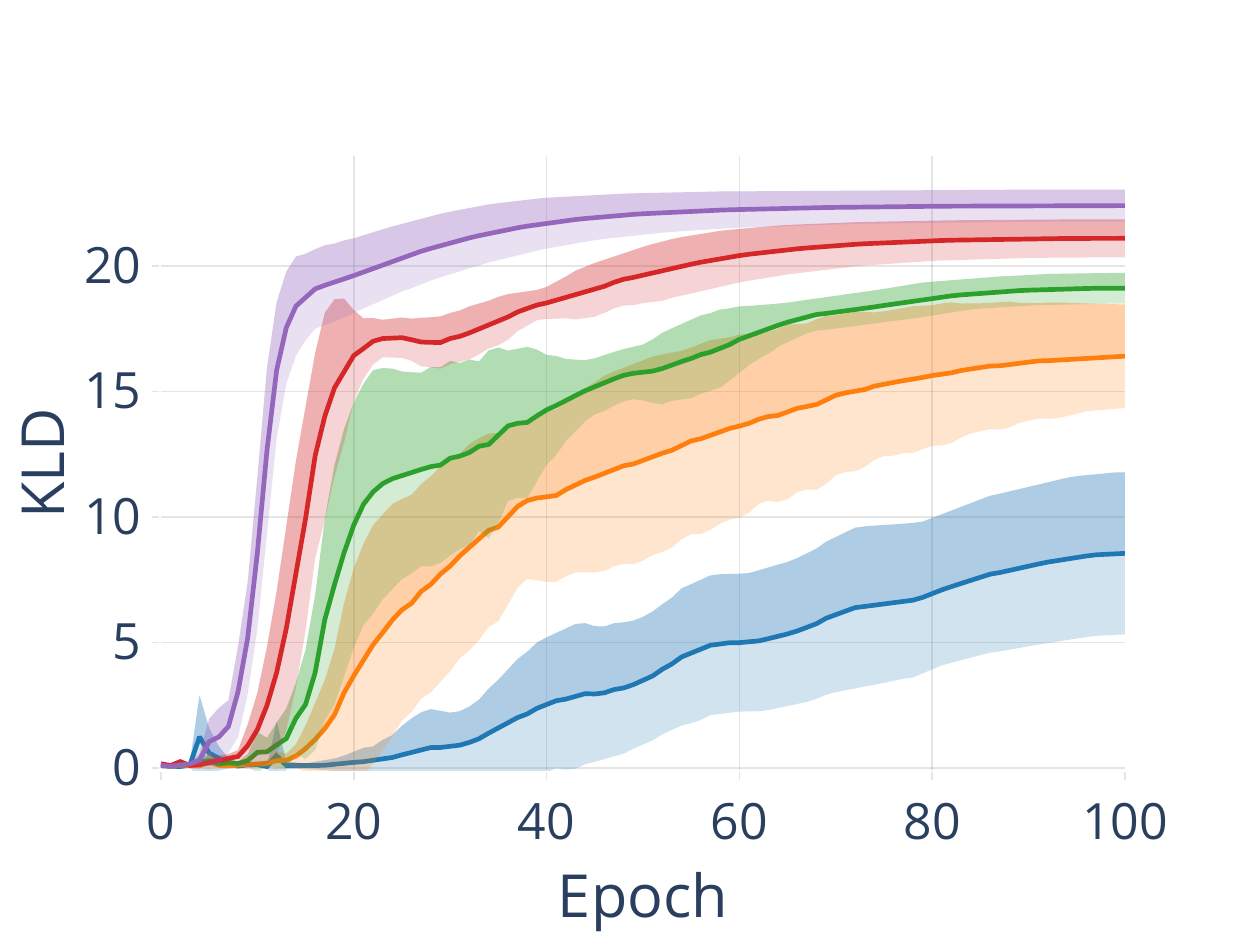}
    }
\caption{\capthead{KL-Divergence from the true distribution for all models for different $\ell$.}{$n = 1024$}
We compute $D_{KL}(P_A || P_\gM)$.
The dip during the \GuessPhase to $0$ shows that the models, in fact, approximate the string's true distribution $P_A$.
}
\label{fig:kld_alphabet_size_all}
\end{figure}

\begin{figure}[H]
    \centering
    \subfloat[Pythia-70M]{
        \includegraphics[width=\allModelsWidth]{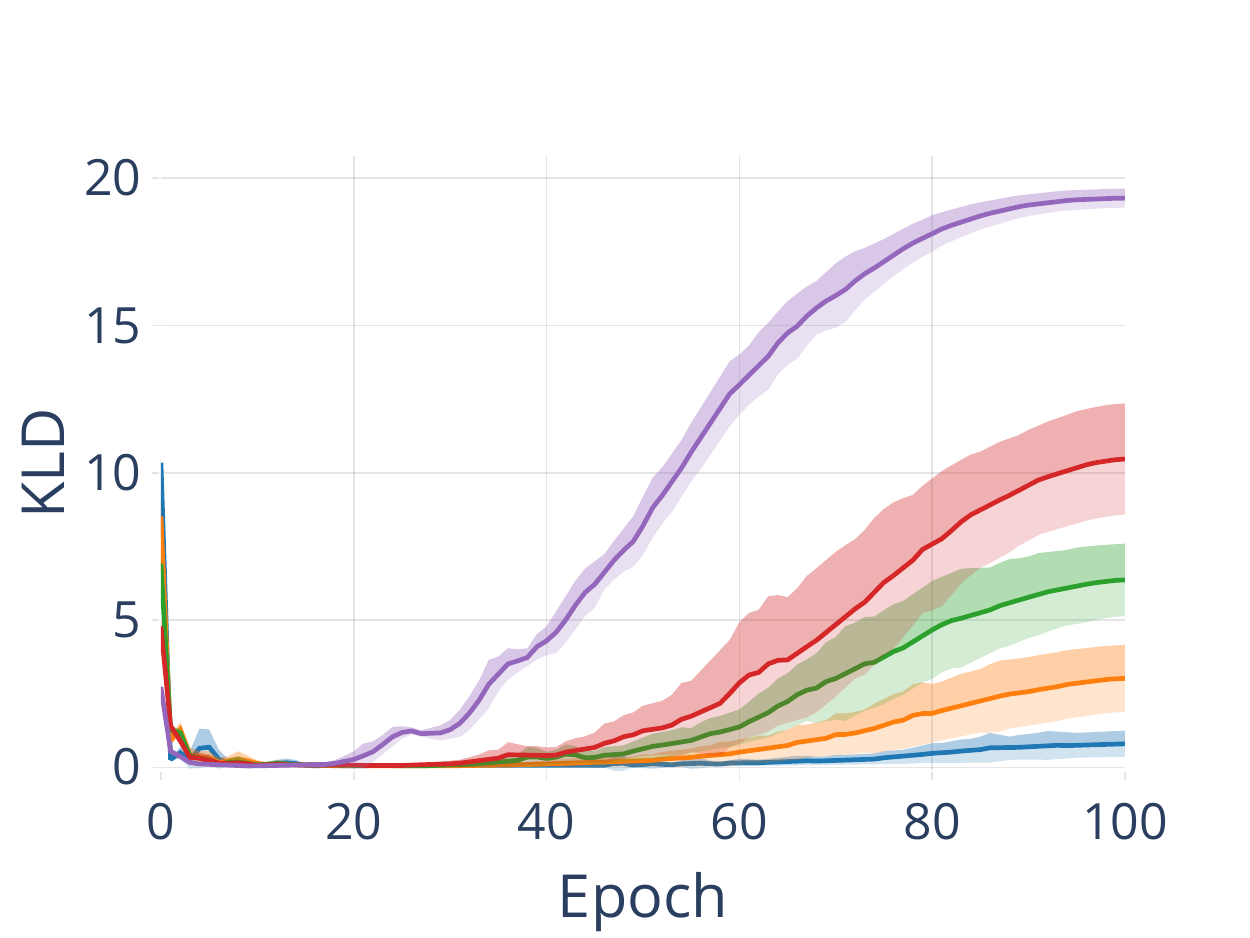}
    }
    \subfloat[Pythia-1B]{
        \includegraphics[width=\allModelsWidth]{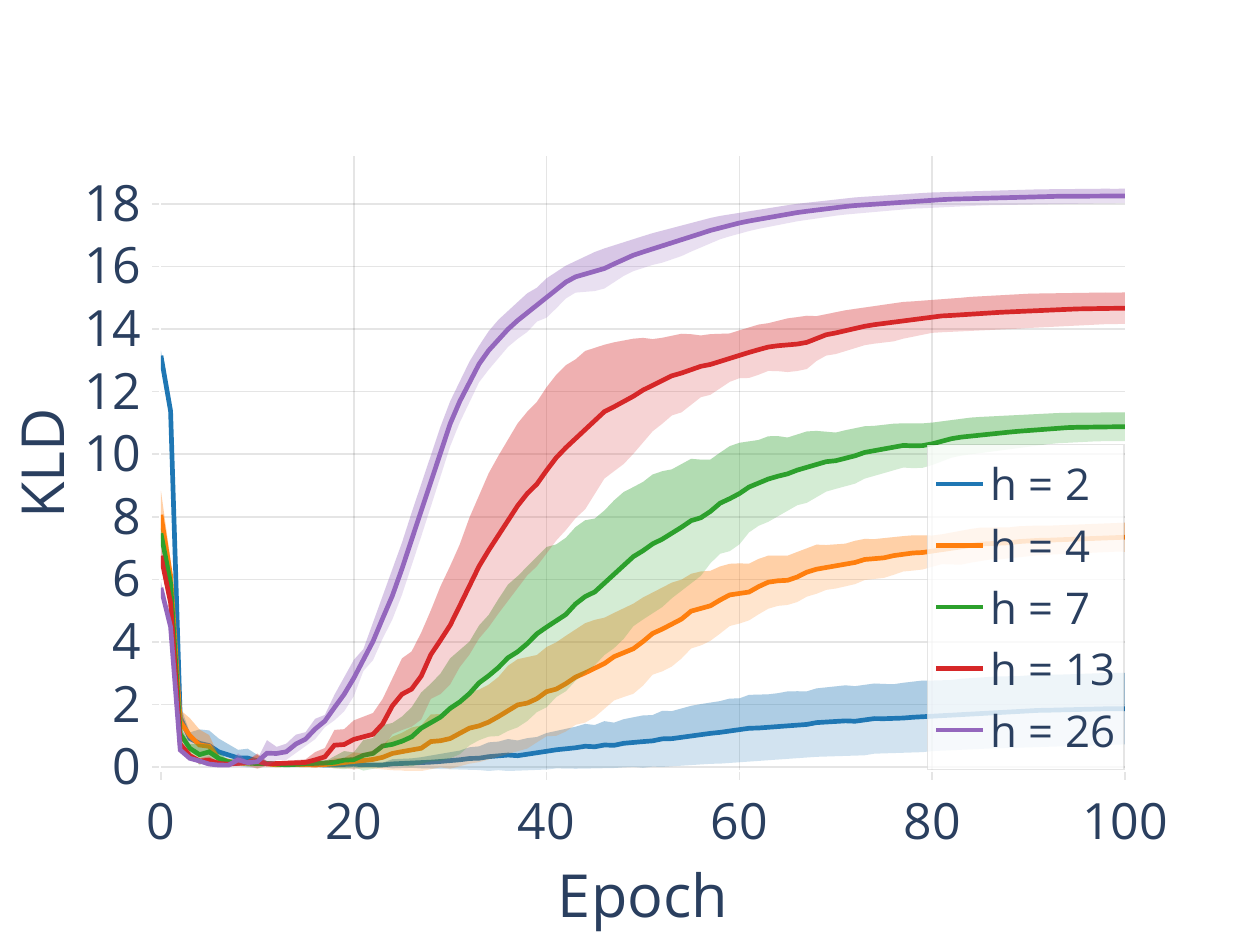}
    }
    \subfloat[Pythia-12B]{
        \includegraphics[width=\allModelsWidth]{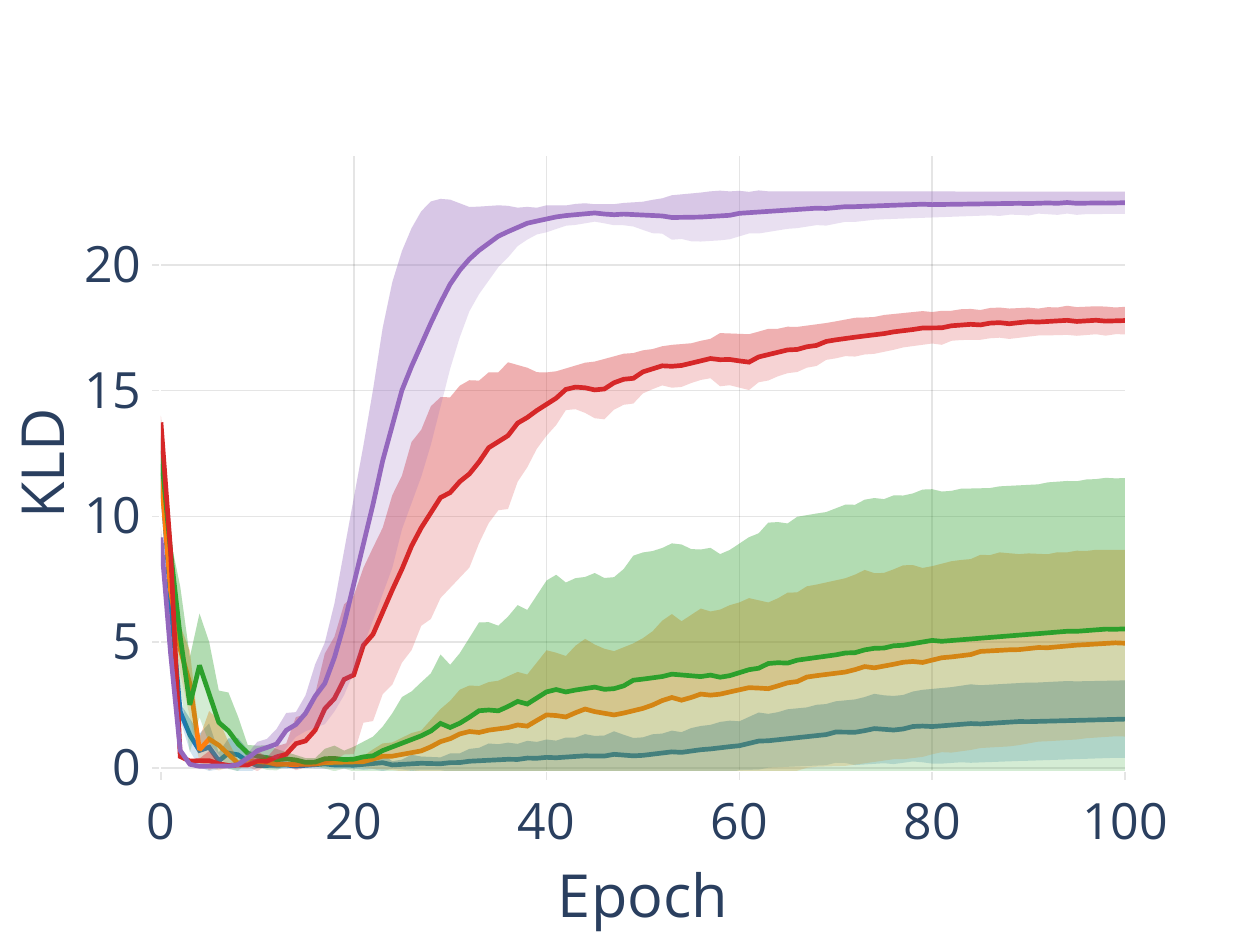}
    }
    \\
    \subfloat[Phi-1.3B]{
        \includegraphics[width=\allModelsWidth]{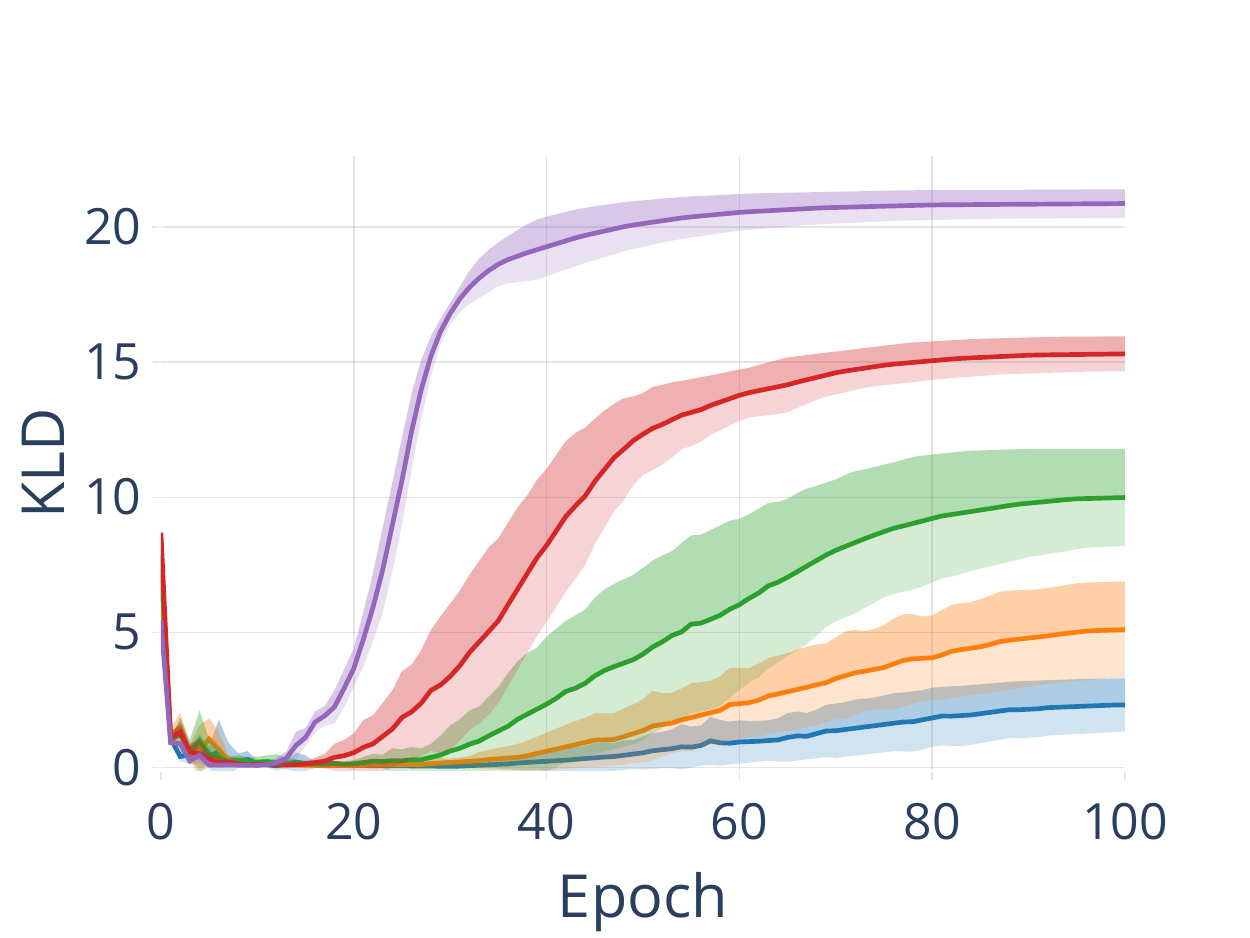}
    }
    \subfloat[Phi-2.7B]{
        \includegraphics[width=\allModelsWidth]{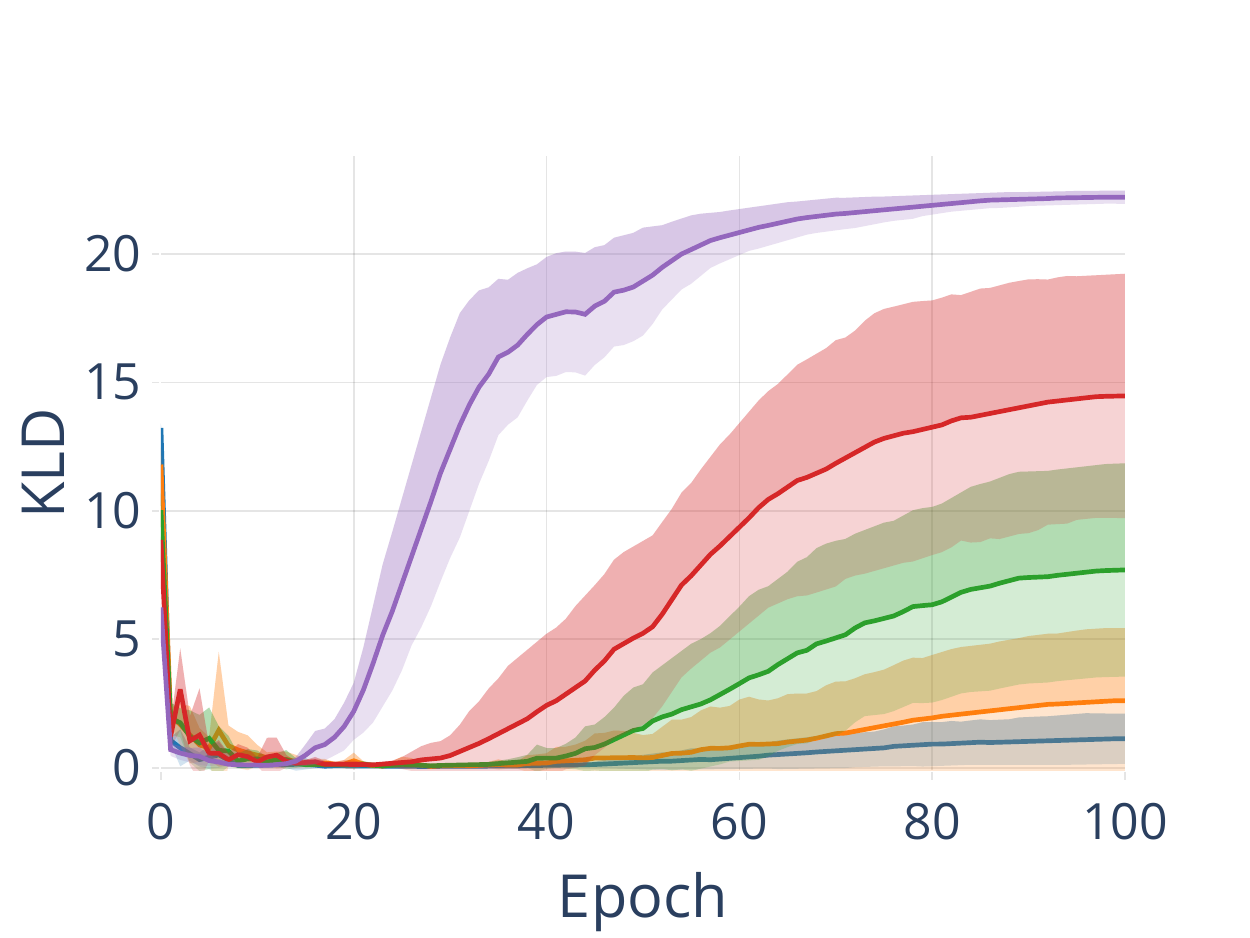}
    }
    \subfloat[Llama2-7B]{
        \includegraphics[width=\allModelsWidth]{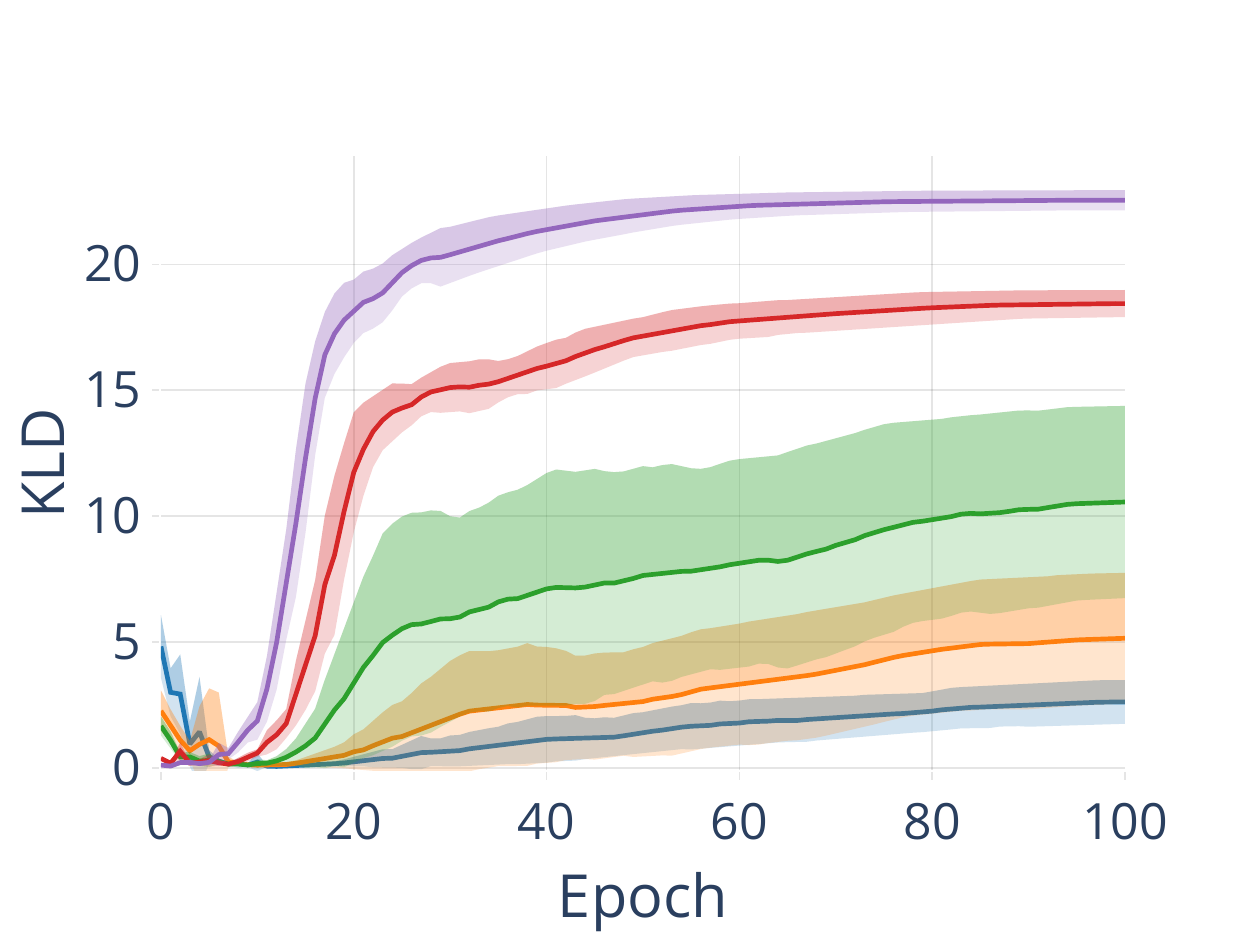}
    }
    \subfloat[Llama2-13B]{
        \includegraphics[width=\allModelsWidth]{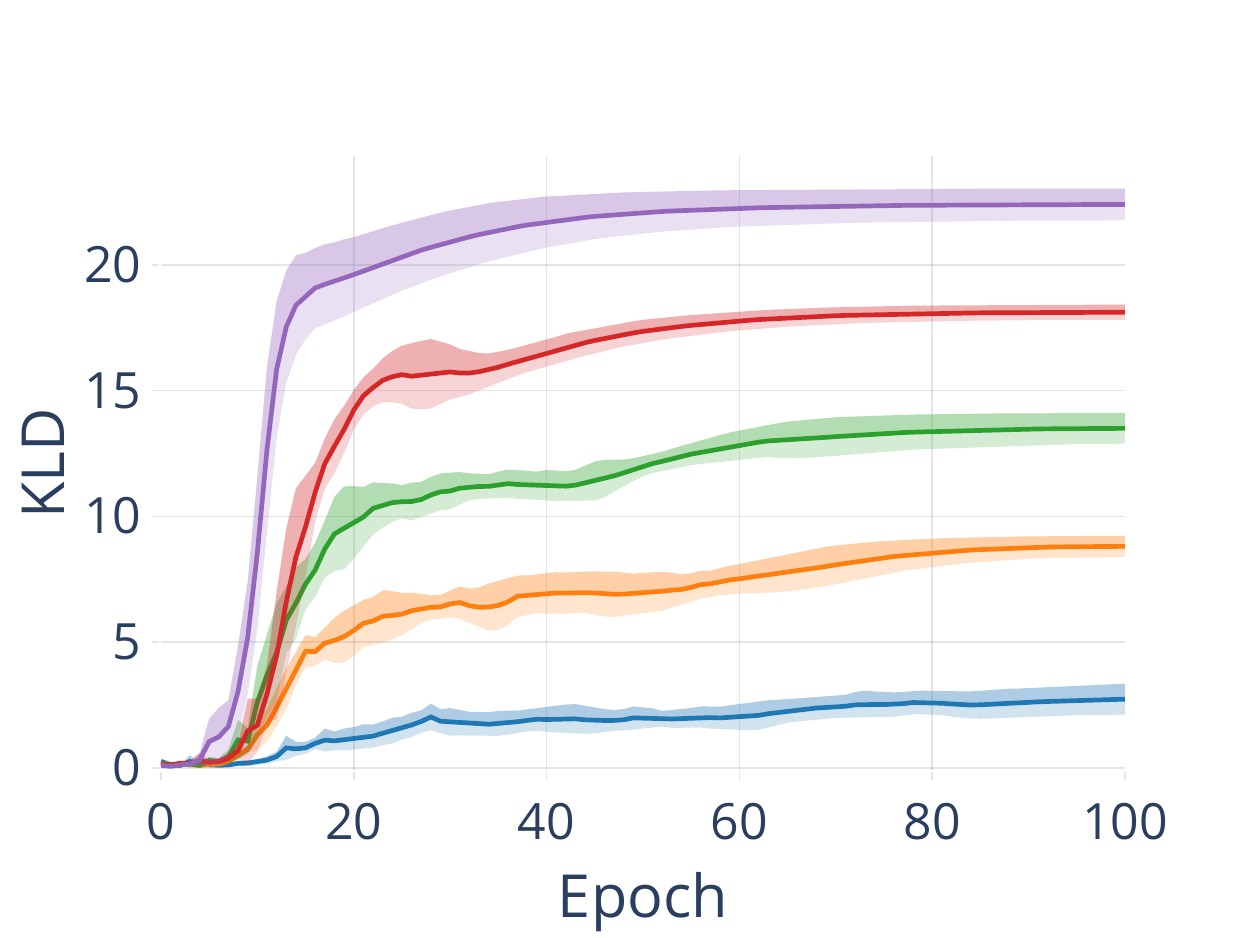}
    }
\caption{\capthead{KL-Divergence from the true distribution for all models for different $h$.}{$n = 1024, \ell = 26$}
We compute $D_{KL}(P_A || P_\gM)$.
The dip during the \GuessPhase to $0$ shows that the models, in fact, approximate the string's true distribution $P_A$.
}
\label{fig:kld_entropy_level_all}
\end{figure}

We show results for additional models for the experiments in Sections~\ref{sec:phases} and~\ref{sec:memorability}.
Memorisation dynamics are shown for different alphabet sizes $\ell$ and entropy levels $h$.
In Figures~\ref{fig:loss_alphabet_size_all} and~\ref{fig:loss_entropy_level_all} we show training loss, in Figures~\ref{fig:accuracy_alphabet_size_all} and~\ref{fig:accuracy_entropy_level_all} we show accuracy, in Figures~\ref{fig:cum-prob_alphabet_size_all} and~\ref{fig:cum-prob_entropy_level_all} we show aggregate probabilities over $A$, in Figures~\ref{fig:entropy_alphabet_size_all} and~\ref{fig:entropy_entropy_level_all} we show entropy over $A$, and in Figures~\ref{fig:kld_alphabet_size_all} and~\ref{fig:kld_entropy_level_all} we show the KLD of $\gM$'s distribution $P_\gM$ over $A$ from the true distribution $P_A$.

\subsection{Results for non-Latin alphabets}
\label{app:dynamics_non_latin}

\begin{figure}[H]
    \centering
    \subfloat[Pythia-1B, Loss]{
        \includegraphics[width=\smallThirdWidth]{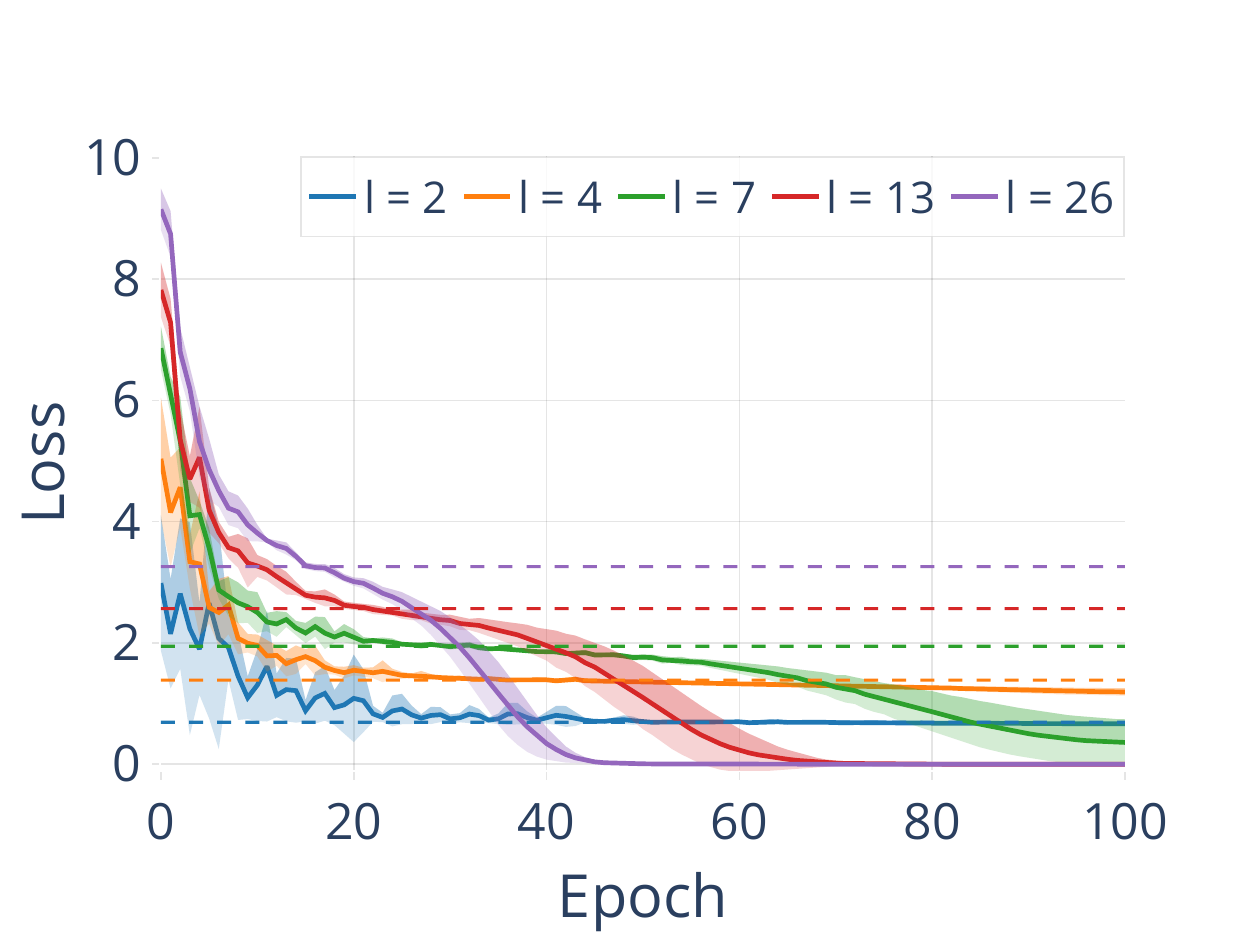}
    }
    \subfloat[Phi-2.7B, Loss]{
        \includegraphics[width=\smallThirdWidth]{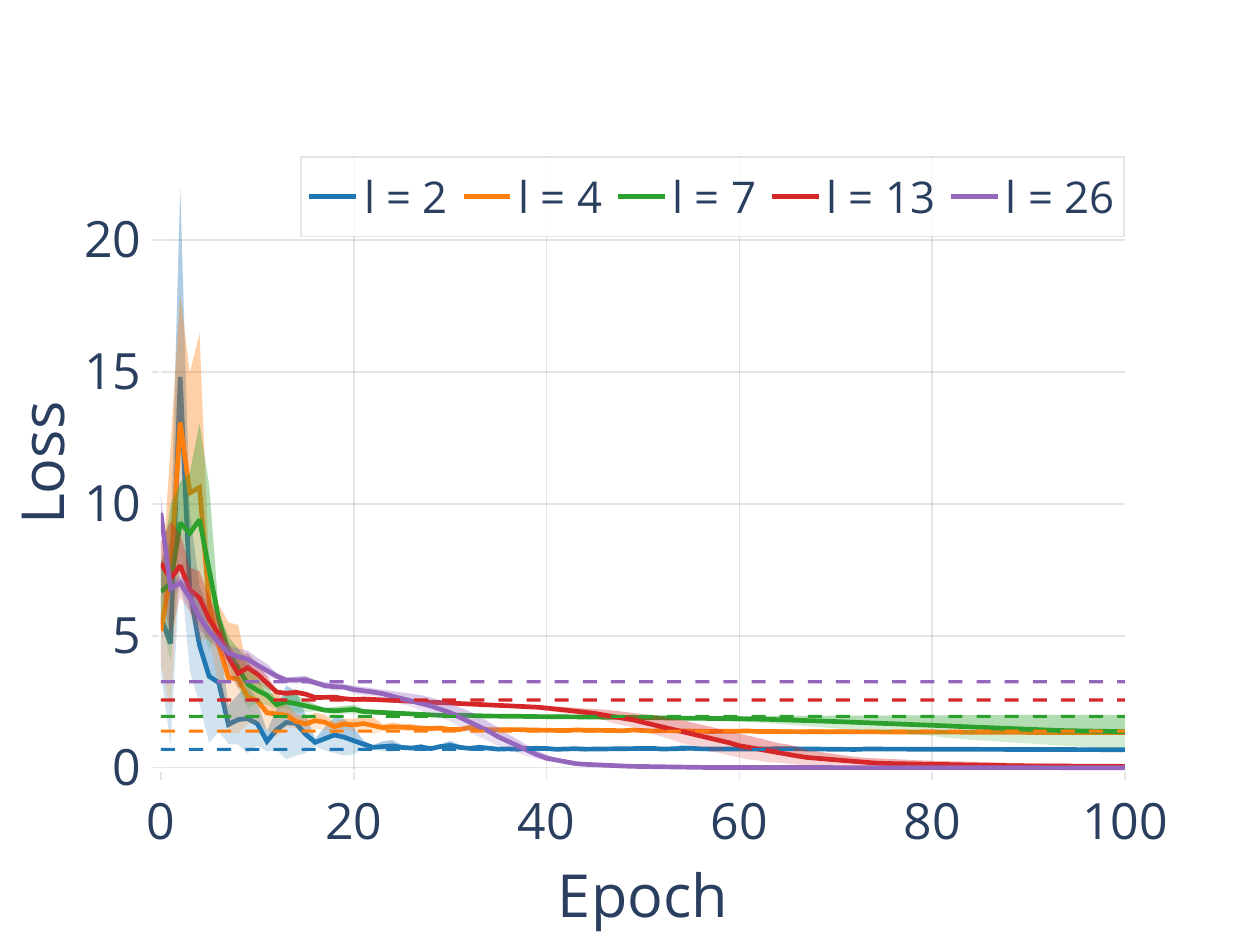}
    }
    \subfloat[Llama2-13B, Loss]{
        \includegraphics[width=\smallThirdWidth]{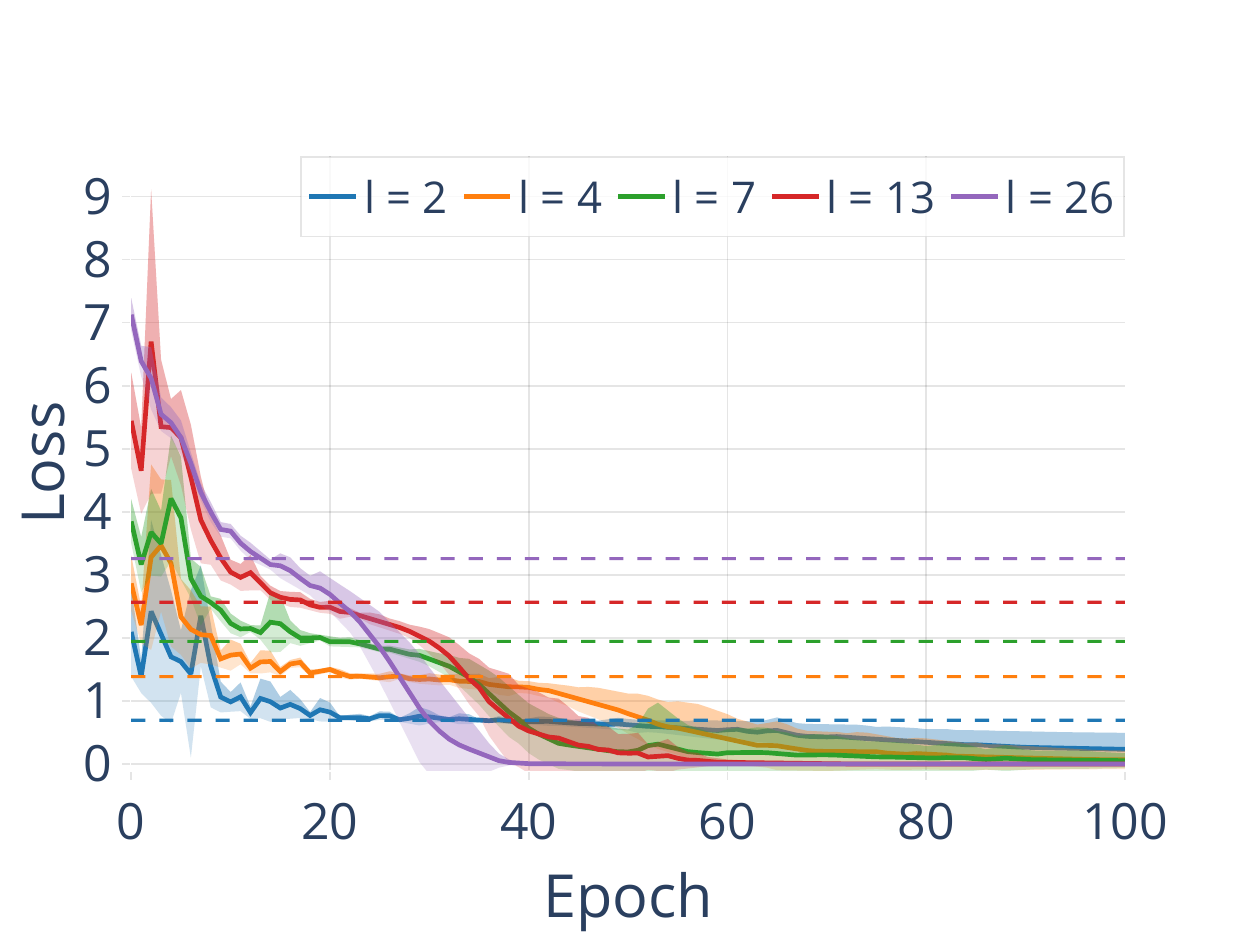}
    }
    \\
    \subfloat[Pythia-1B, Accuracy]{
        \includegraphics[width=\smallThirdWidth]{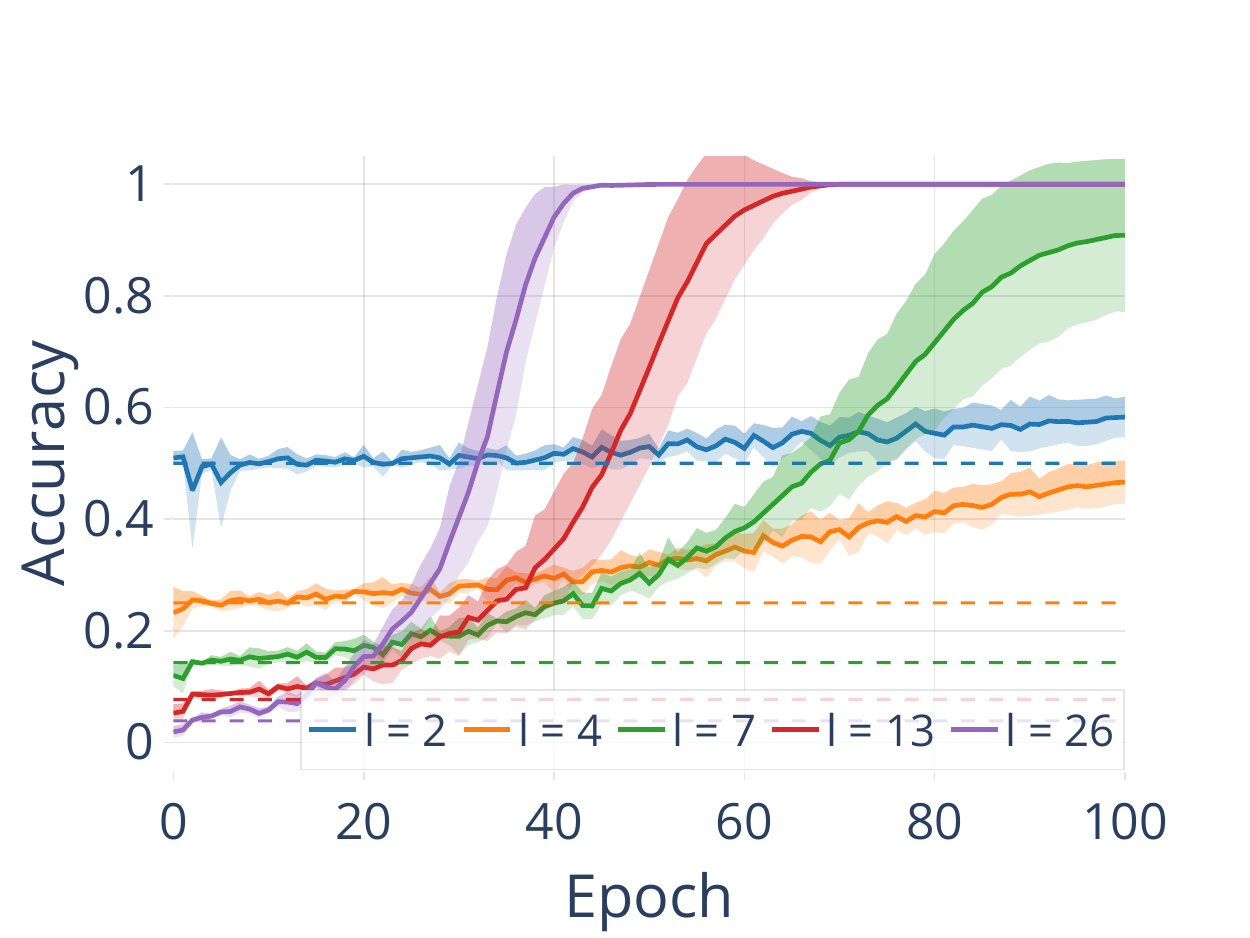}
    }
    \subfloat[Phi-2.7B, Accuracy]{
        \includegraphics[width=\smallThirdWidth]{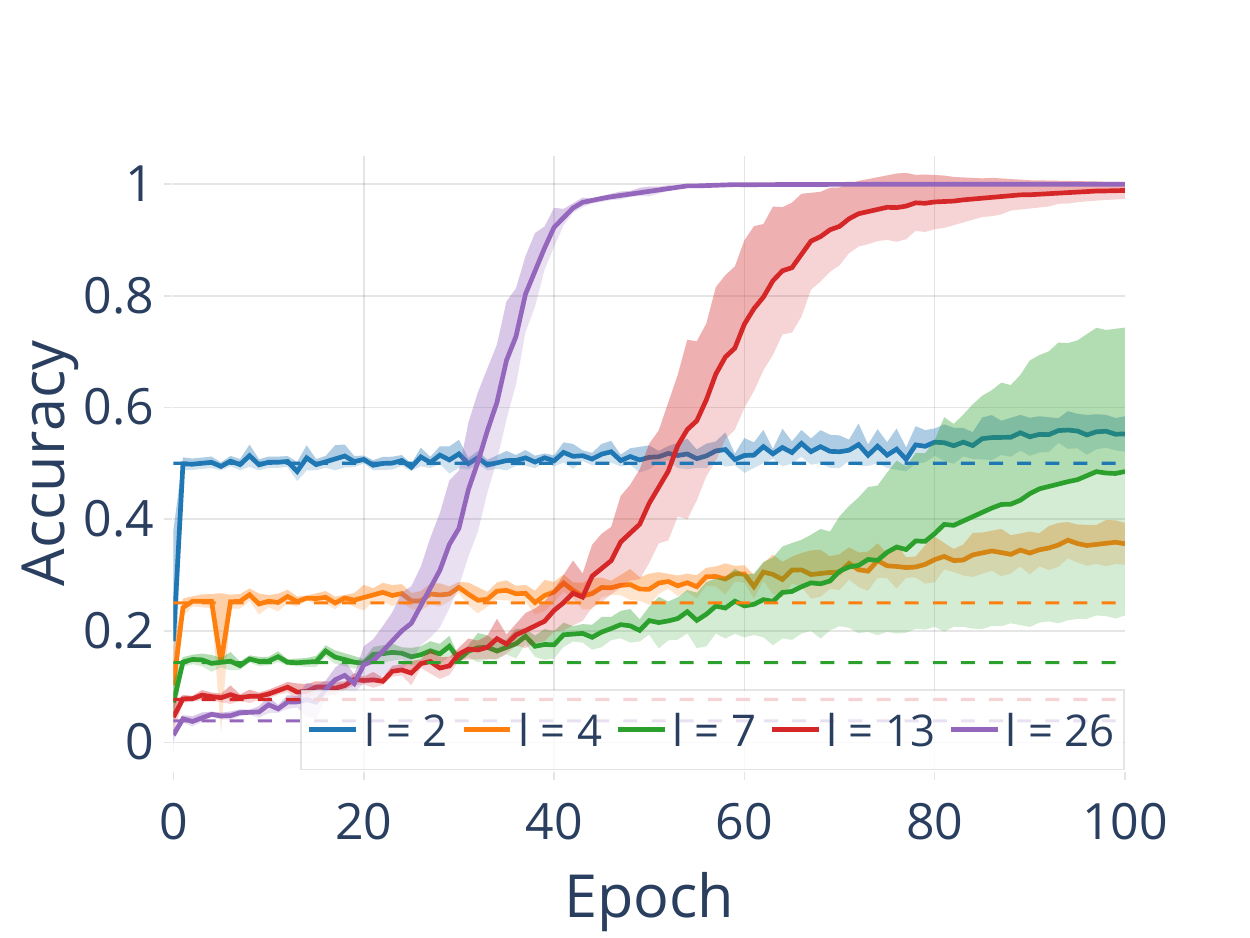}
    }
    \subfloat[Llama2-13B, Accuracy]{
        \includegraphics[width=\smallThirdWidth]{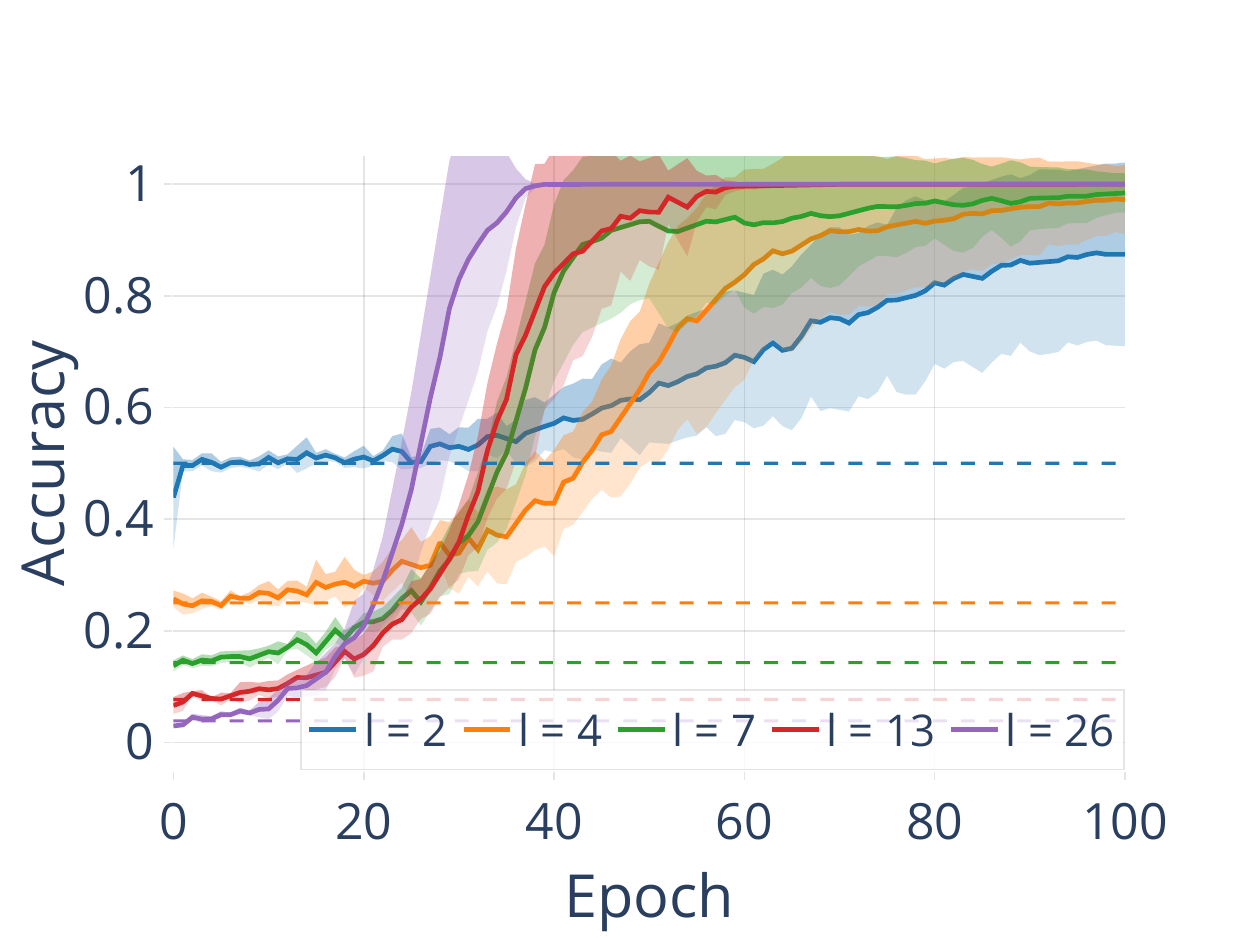}
    }
    \\
    \subfloat[Pythia-1B, Cumulative Probability]{
        \includegraphics[width=\smallThirdWidth]{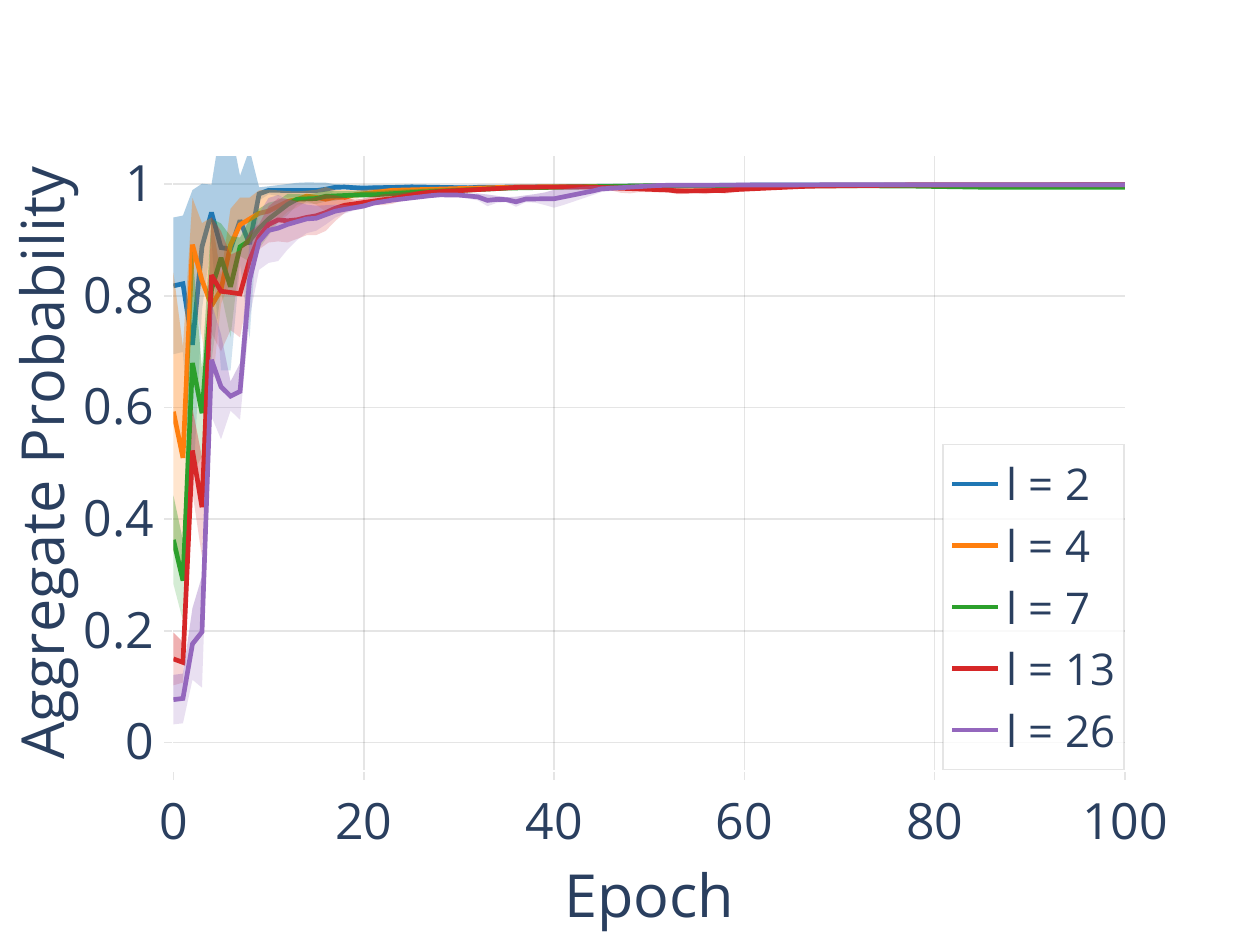}
    }
    \subfloat[Phi-2.7B, Cumulative Probability]{
        \includegraphics[width=\smallThirdWidth]{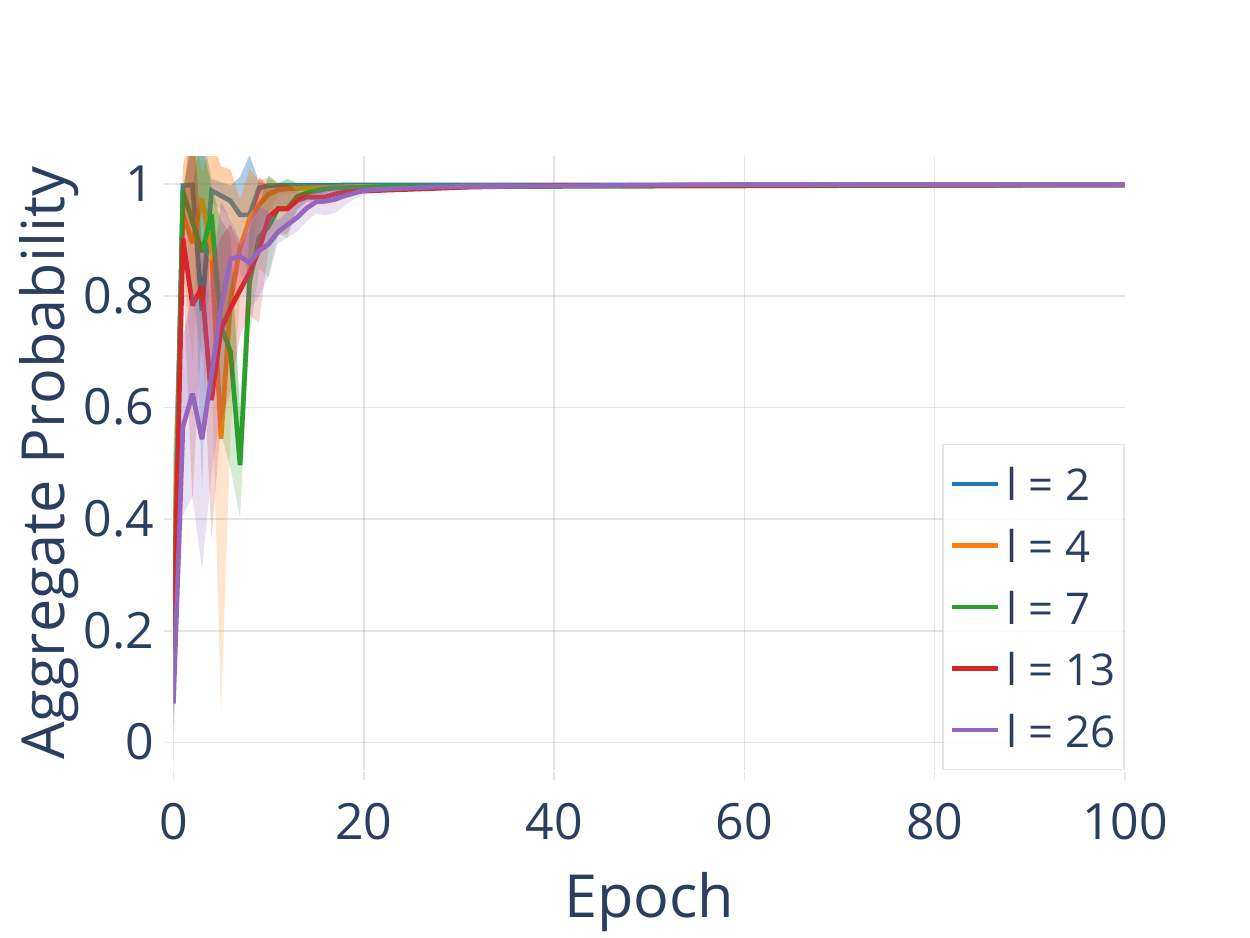}
    }
    \subfloat[Llama2-13B, Cumulative Probability]{
        \includegraphics[width=\smallThirdWidth]{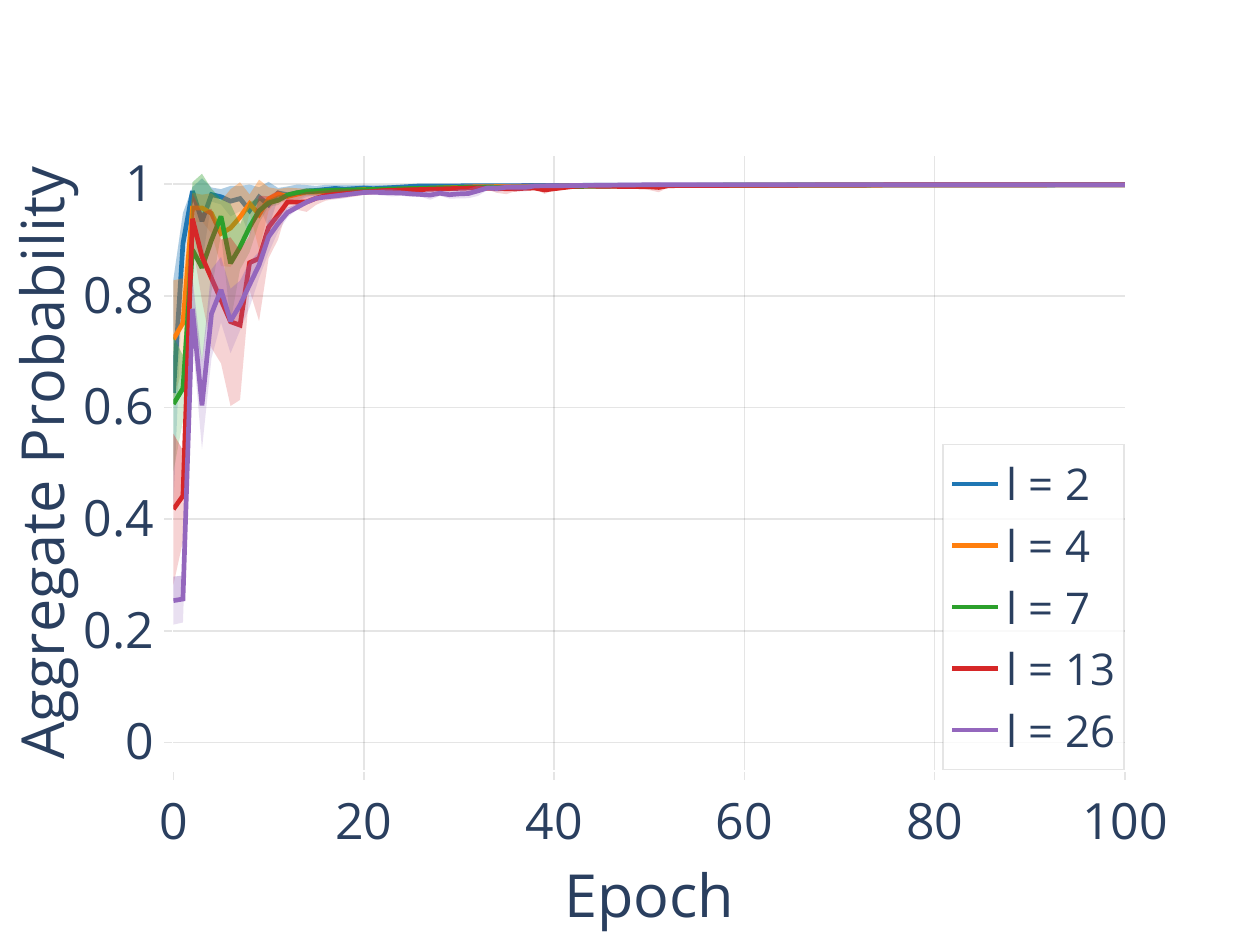}
    }
    \\
    \subfloat[Pythia-1B, Entropy]{
        \includegraphics[width=\smallThirdWidth]{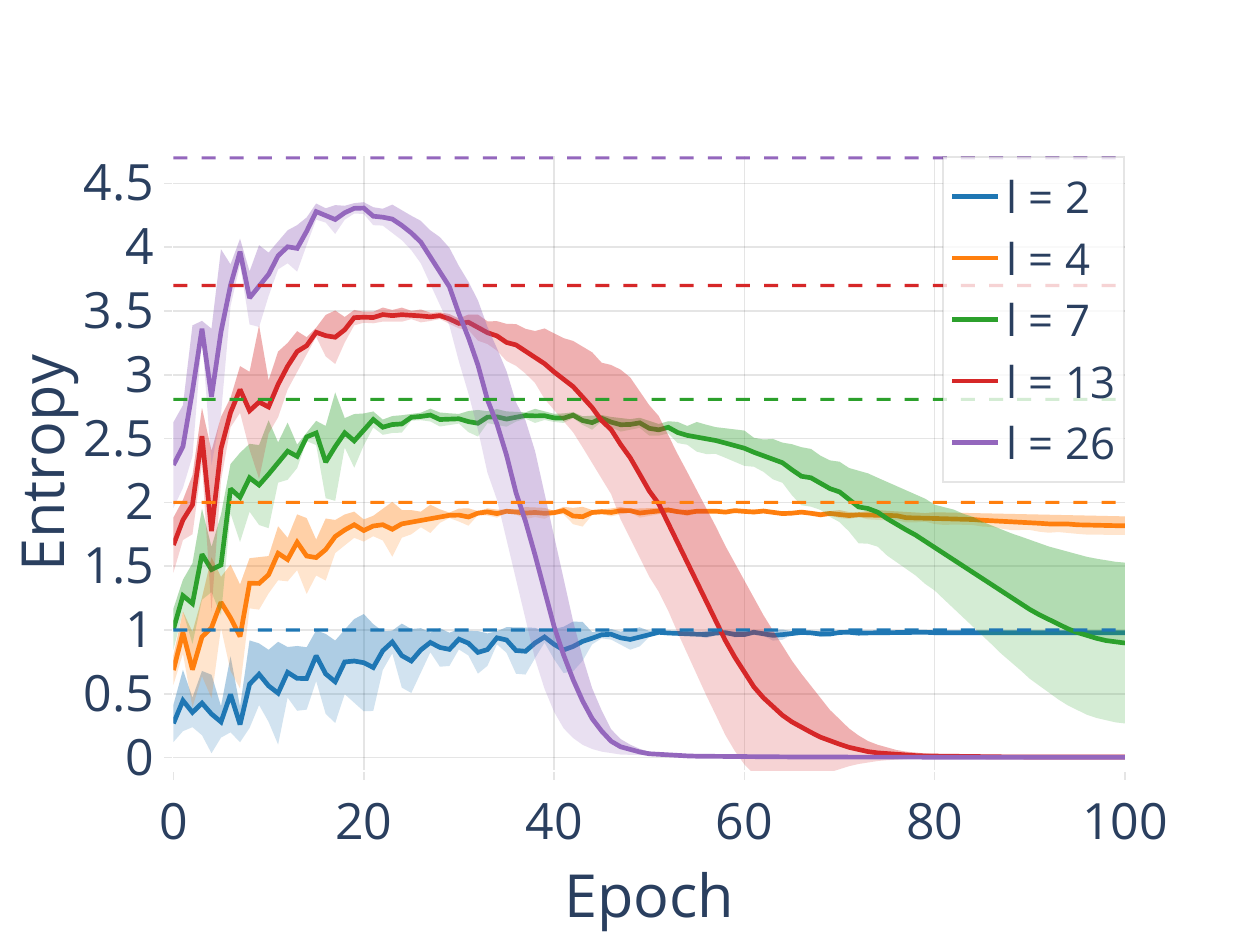}
    }
    \subfloat[Phi-2.7B, Entropy]{
        \includegraphics[width=\smallThirdWidth]{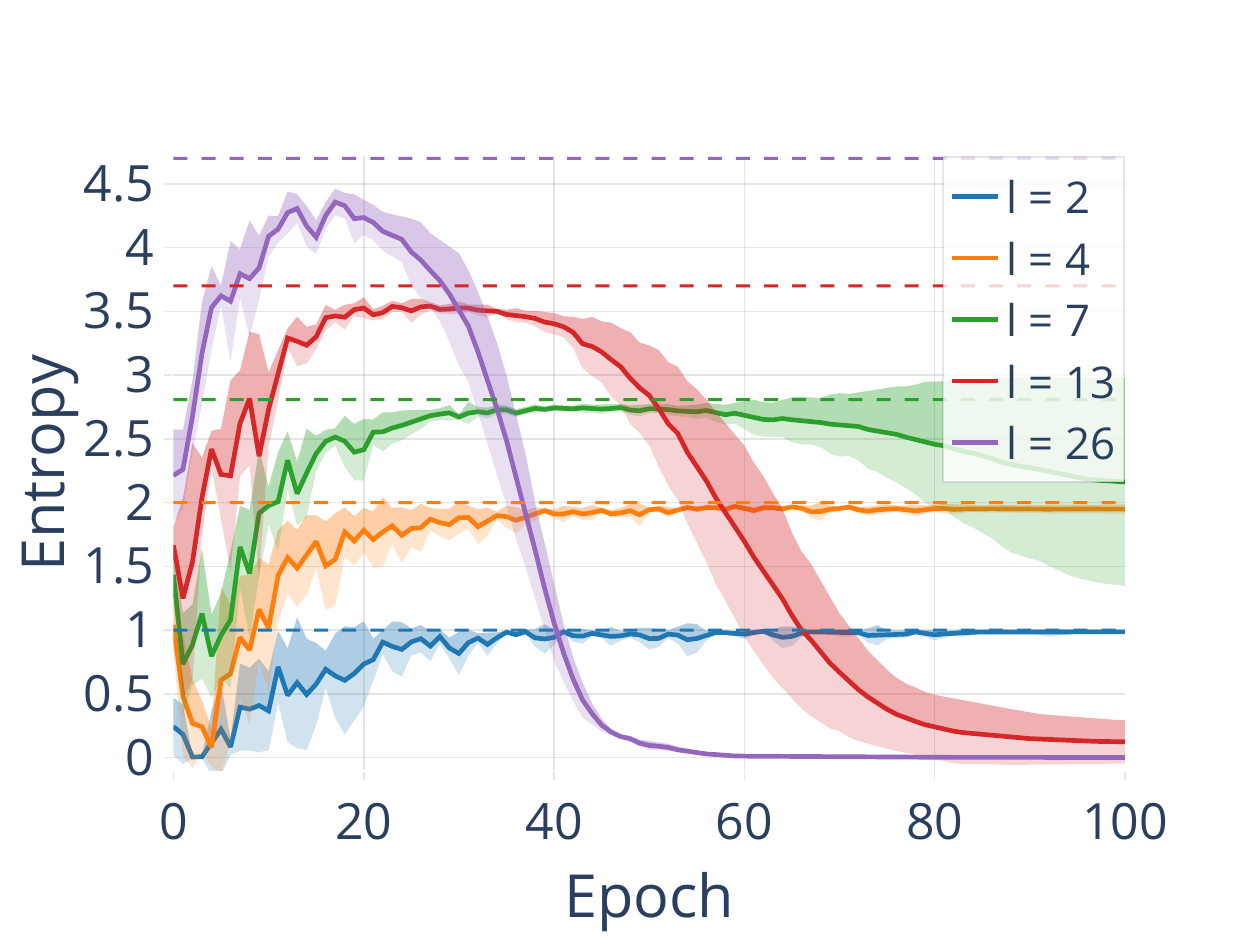}
    }
    \subfloat[Llama2-13B, Entropy]{
        \includegraphics[width=\smallThirdWidth]{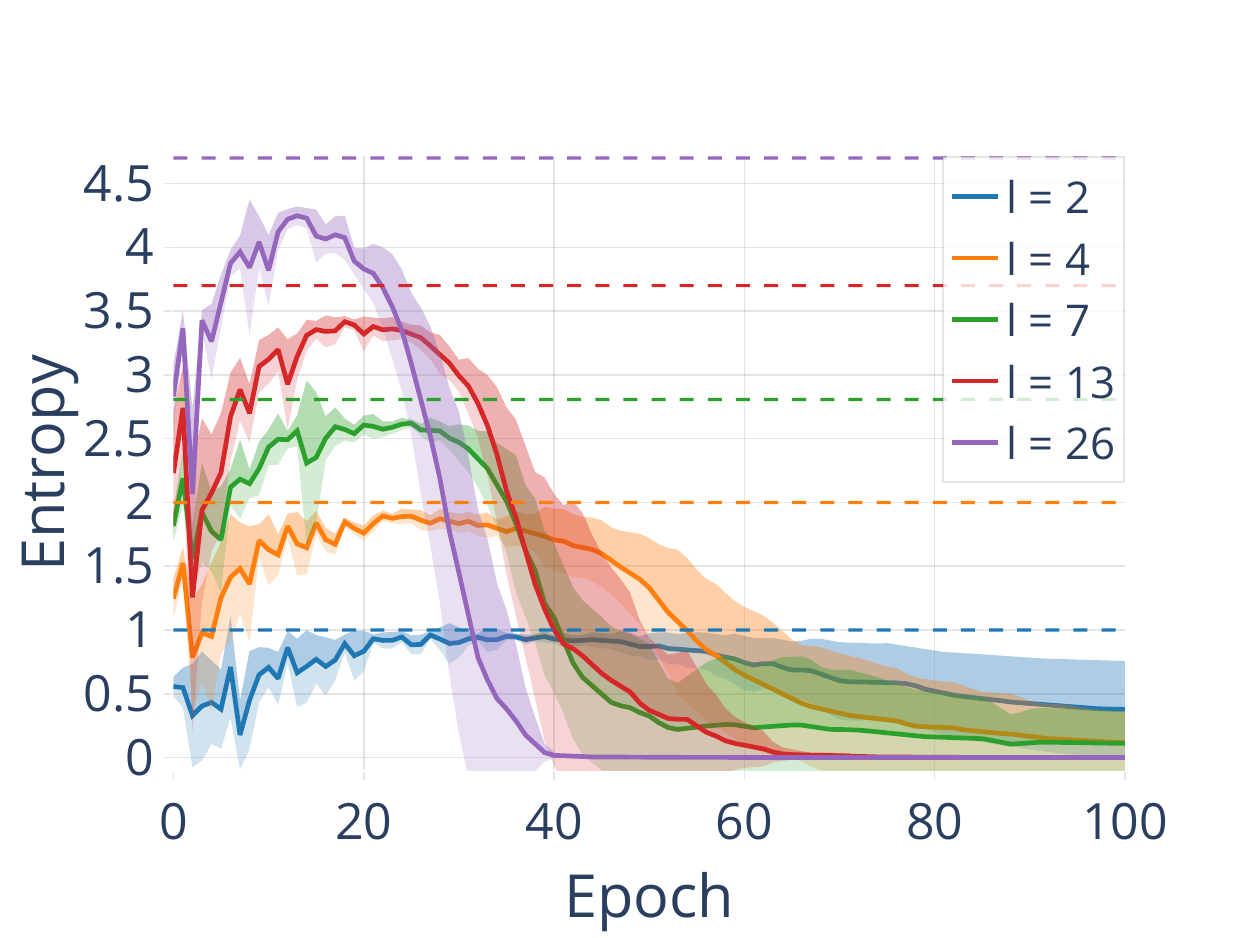}
    }
    \\
    \subfloat[Pythia-1B, KL-Divergence]{
        \includegraphics[width=\smallThirdWidth]{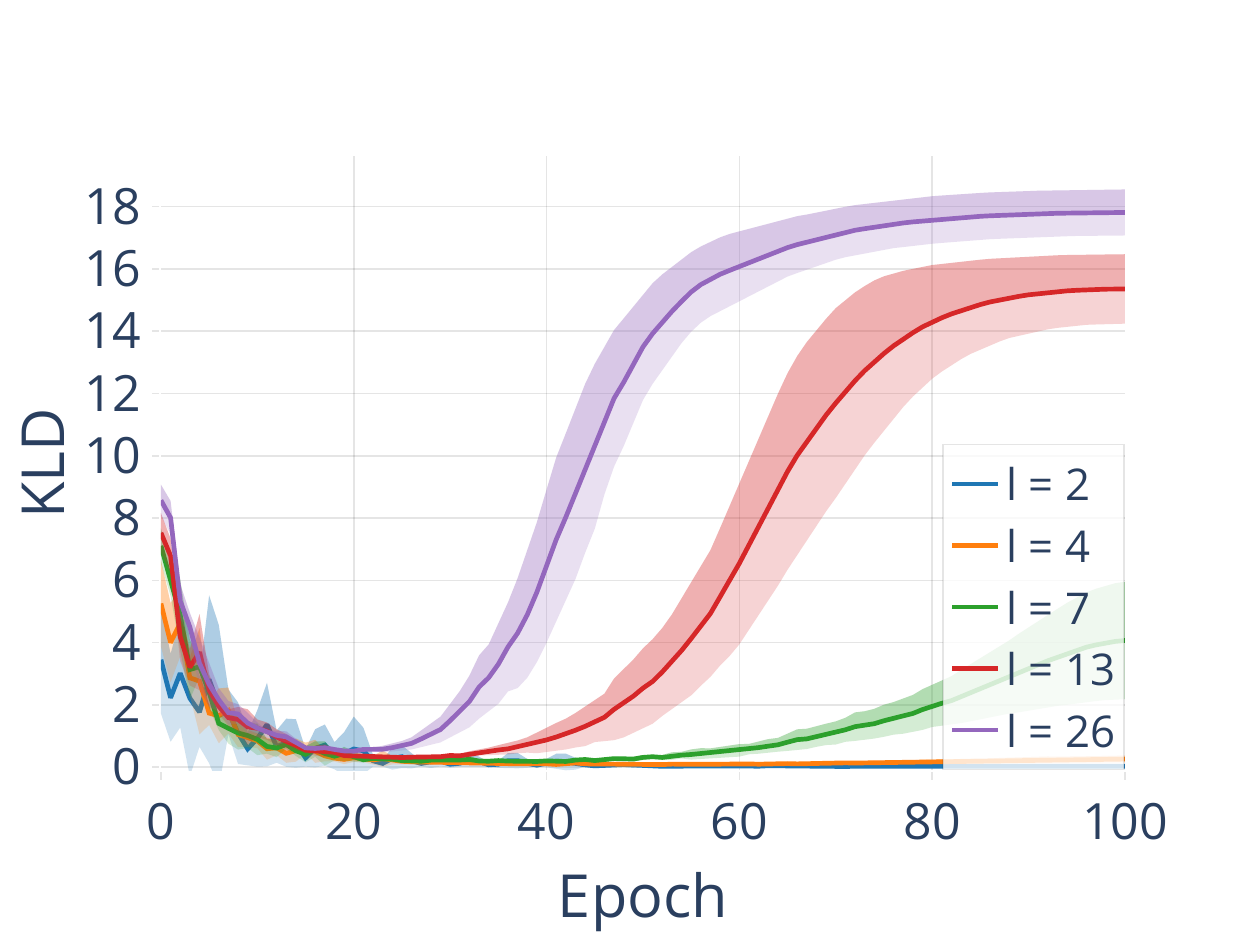}
    }
    \subfloat[Phi-2.7B, KL-Divergence]{
        \includegraphics[width=\smallThirdWidth]{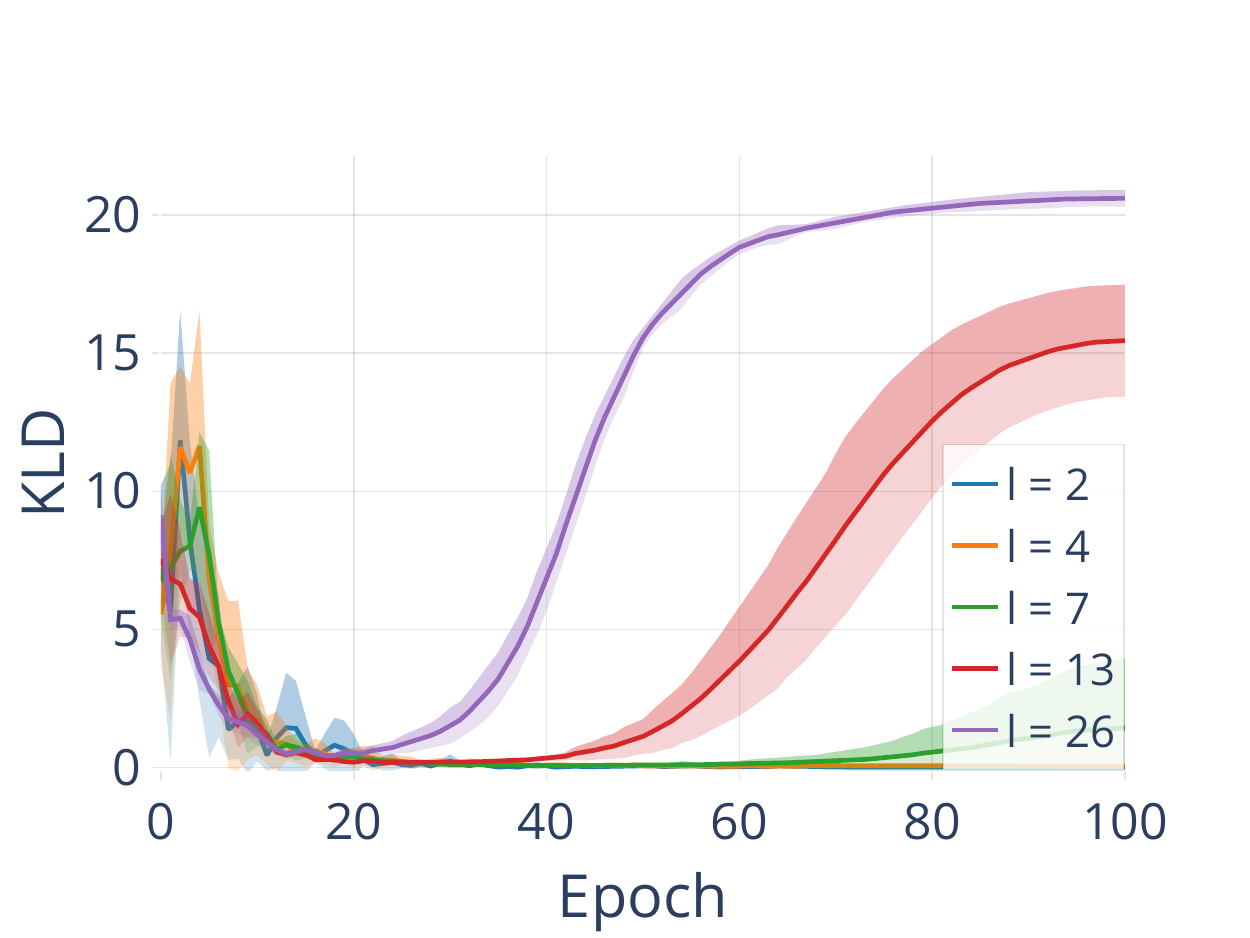}
    }
    \subfloat[Llama2-13B, KL-Divergence]{
        \includegraphics[width=\smallThirdWidth]{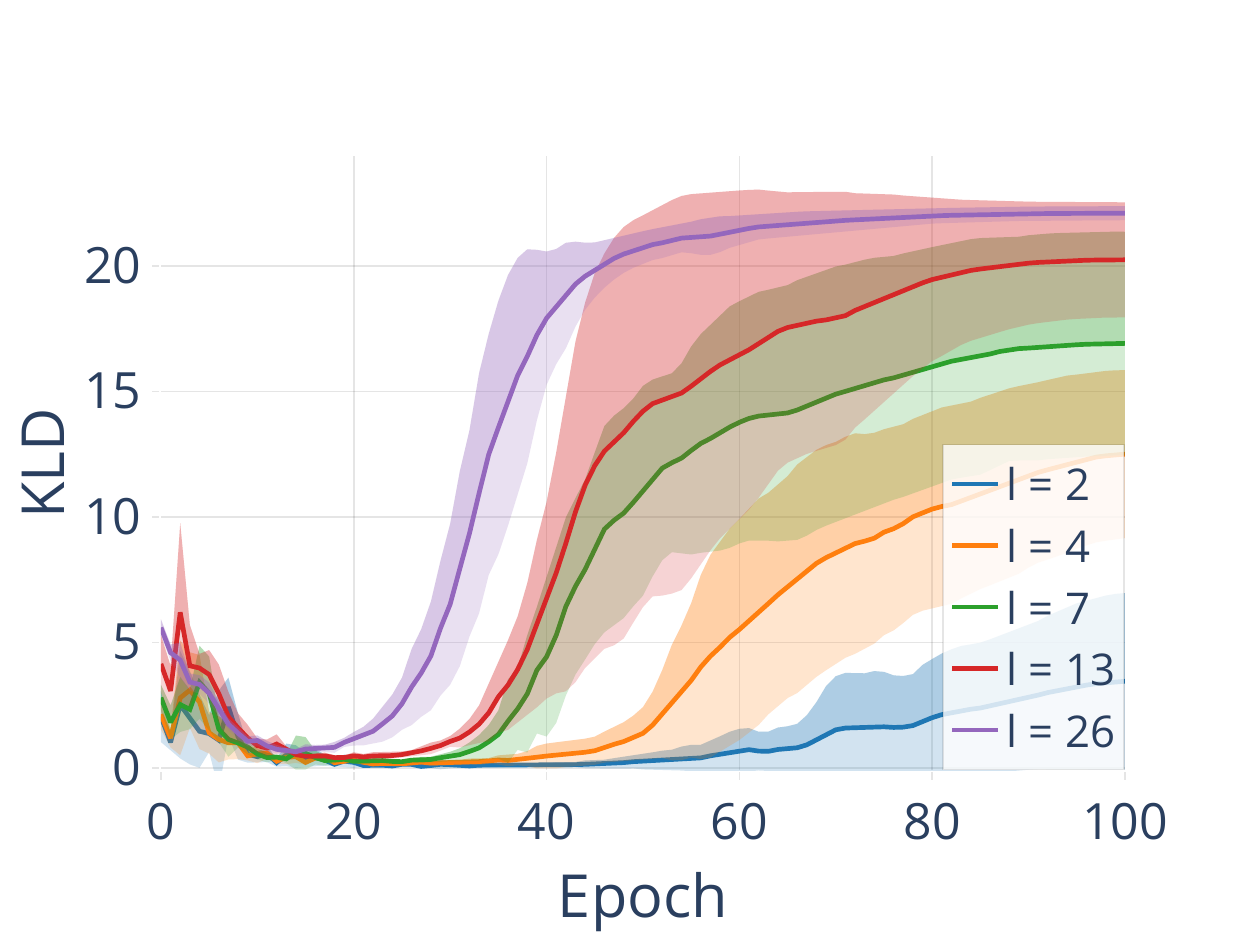}
    }
    \\
\caption{\capthead{Accuracy, loss, cumulative probability, entropy and KLD for different $\ell$ with non-latin alphabets.}{$n = 1024$}
We choose $\ell$ tokens randomly from entire token vocabulary $V$ instead of using lowercase Latin letters.
We observe the same patterns as for the Latin alphabets shown in Sections~\ref{sec:phases} and~\ref{sec:memorability} and in Appendix~\ref{app:dynamics_additional_results}.
However, Llama2-13B also exhibits a \GuessPhase in this context.
}
\label{fig:non_latin_alphabet_all}
\end{figure}

In Figure~\ref{fig:non_latin_alphabet_all} we show ablation results for different models and $\ell$ for non-Latin alphabets, i.e. with $\ell$ tokens chosen randomly from entire token vocabulary $V$.
We observe the same patterns as for the Latin alphabets shown in Sections~\ref{sec:phases} and~\ref{sec:memorability} and in Appendix~\ref{app:dynamics_additional_results}.
The only difference is that Llama2-13B also exhibits a \GuessPhase in this context.

\subsection{Results for untrained models}
\label{app:dynamics_untrained}

\begin{figure}[H]
    \centering
    \subfloat[Pythia-1B, Loss]{
        \includegraphics[width=\smallThirdWidth]{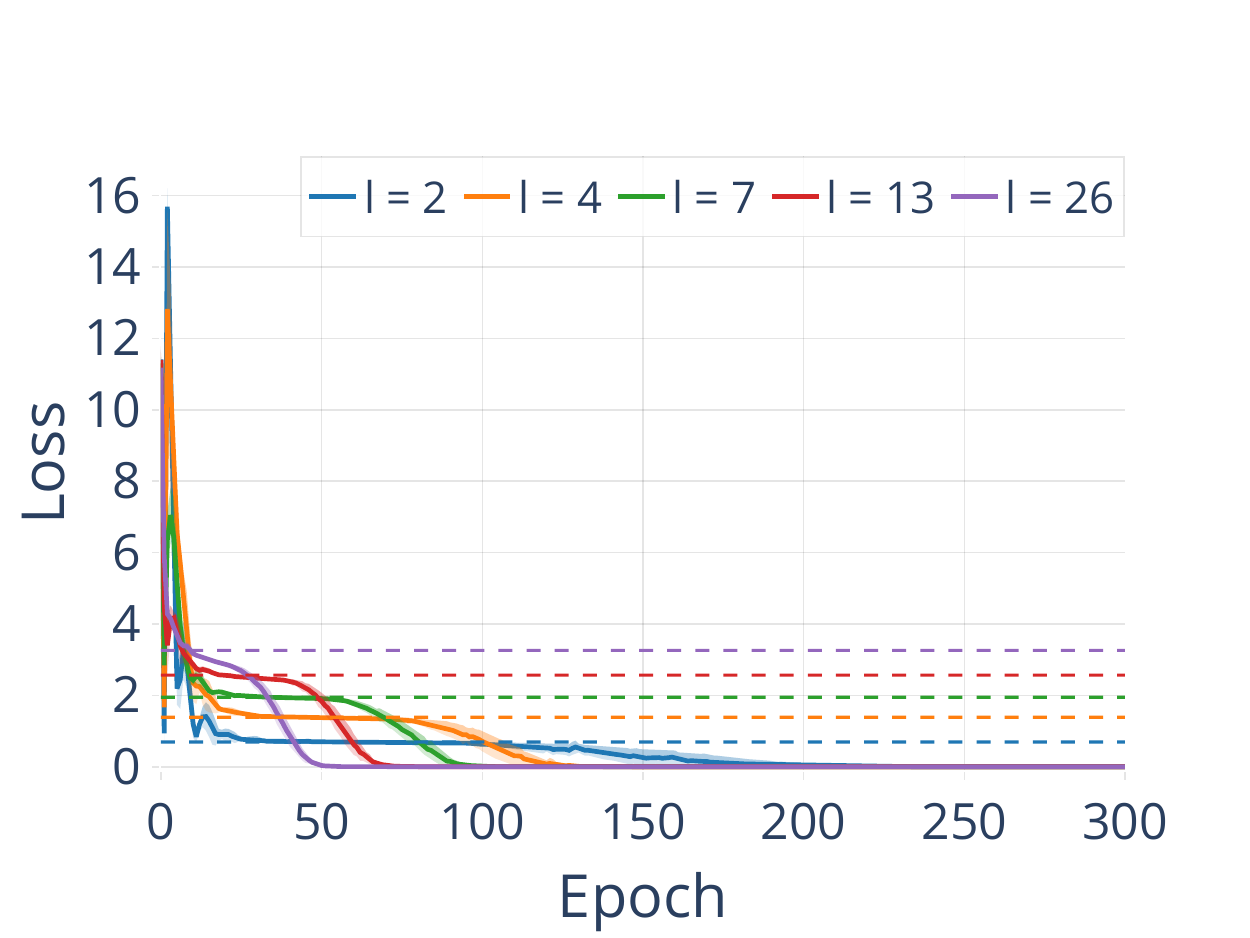}
    }
    \subfloat[Phi-2.7B, Loss]{
        \includegraphics[width=\smallThirdWidth]{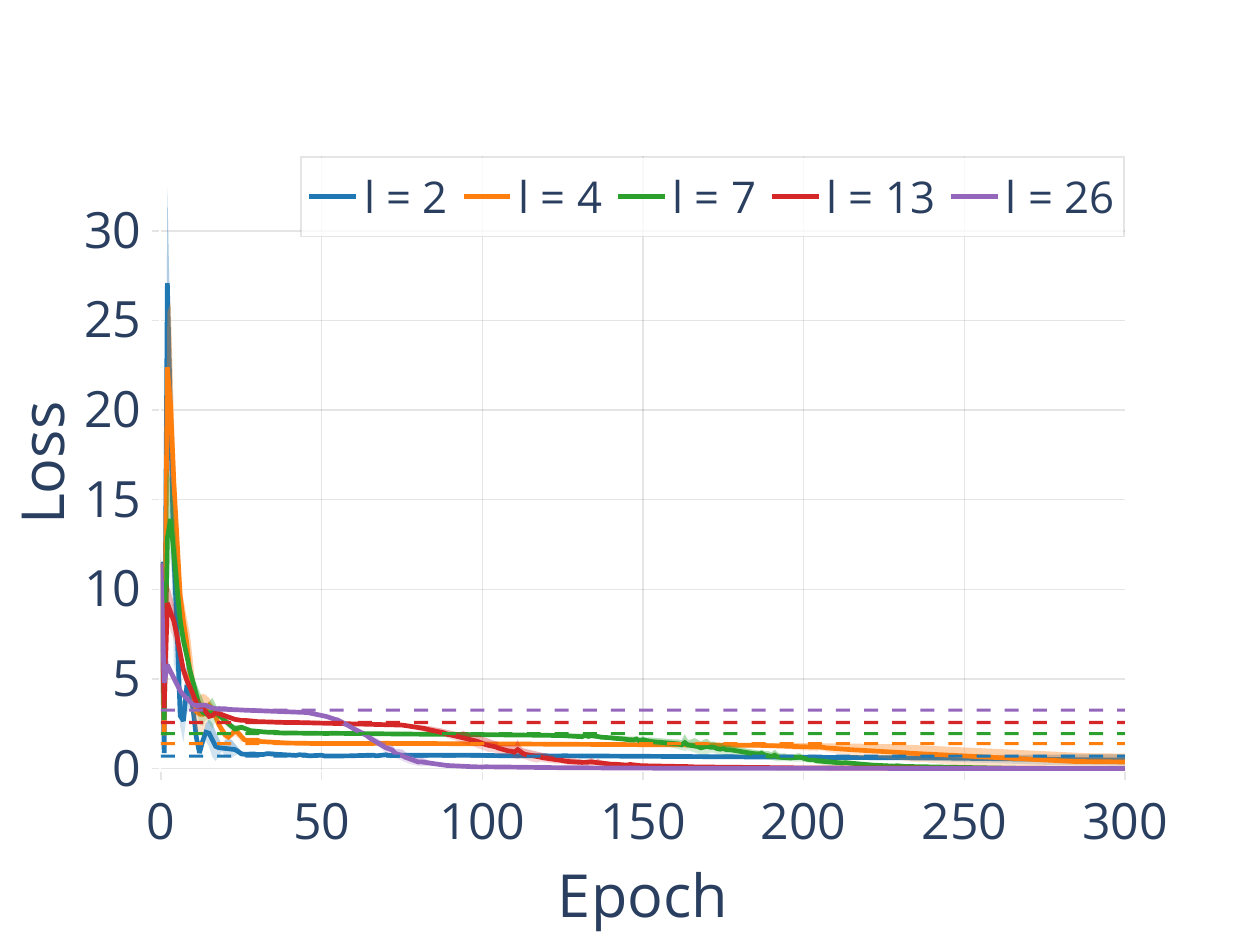}
    }
    \subfloat[Llama2-13B, Loss]{
        \includegraphics[width=\smallThirdWidth]{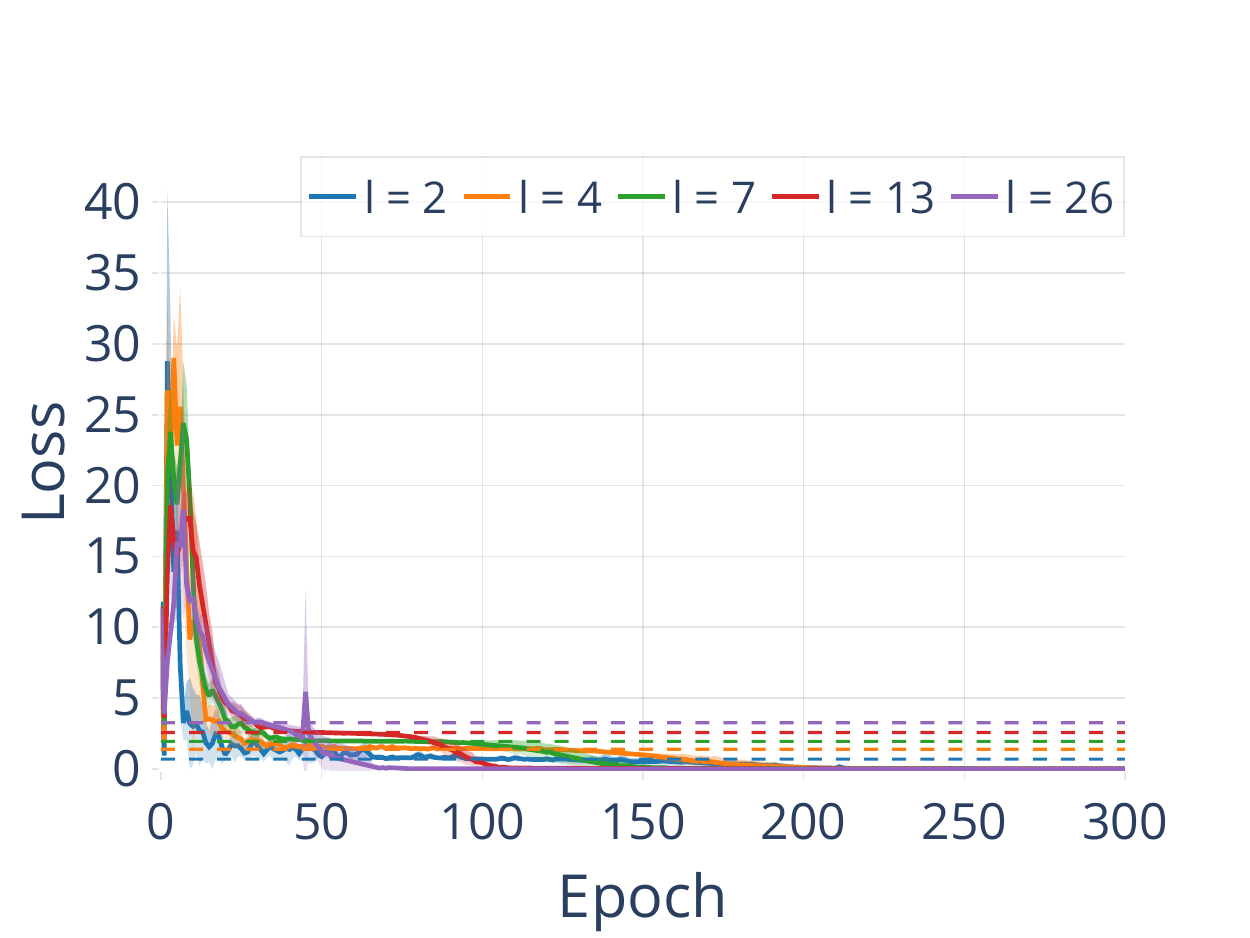}
    }
    \\
    \subfloat[Pythia-1B, Accuracy]{
        \includegraphics[width=\smallThirdWidth]{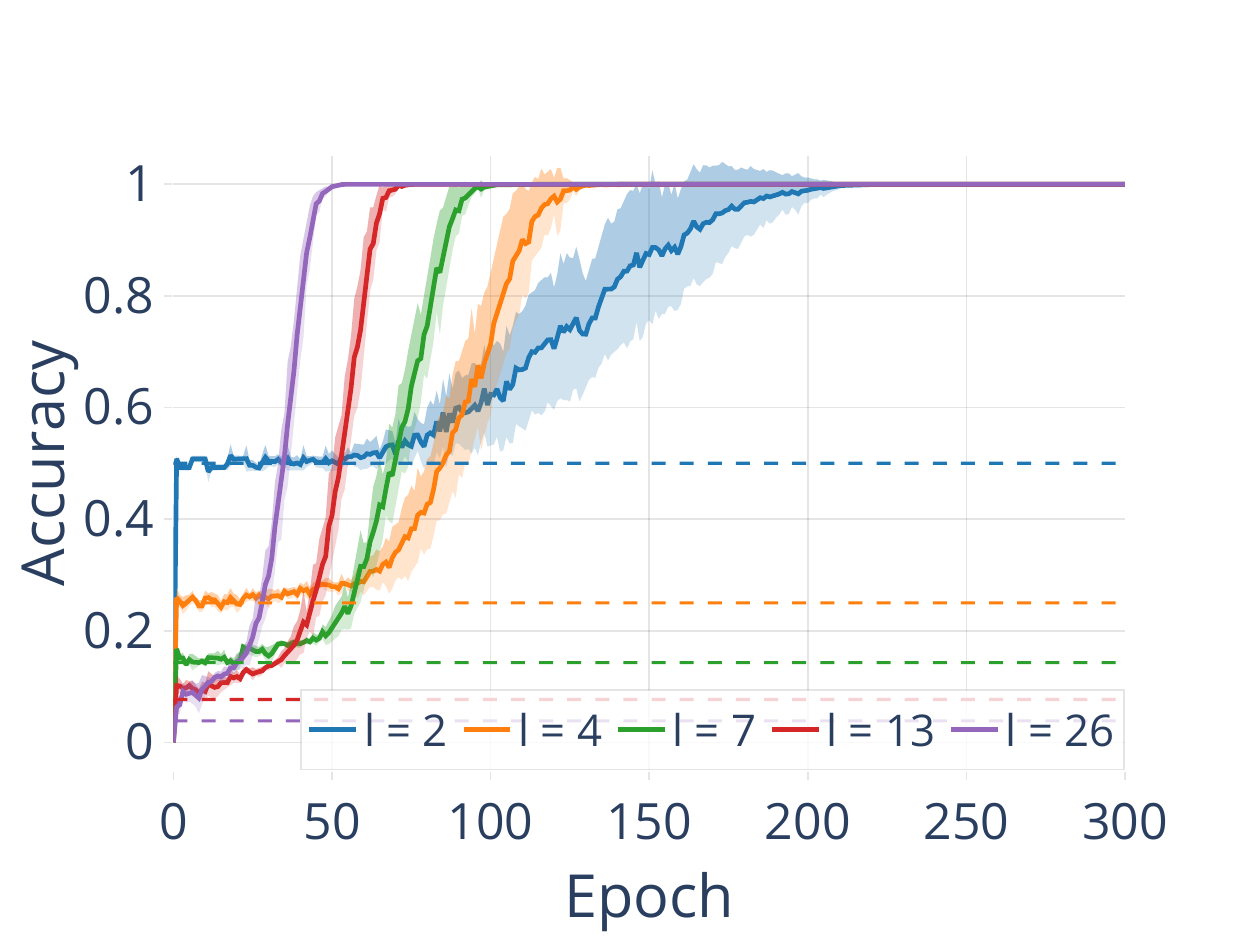}
    }
    \subfloat[Phi-2.7B, Accuracy]{
        \includegraphics[width=\smallThirdWidth]{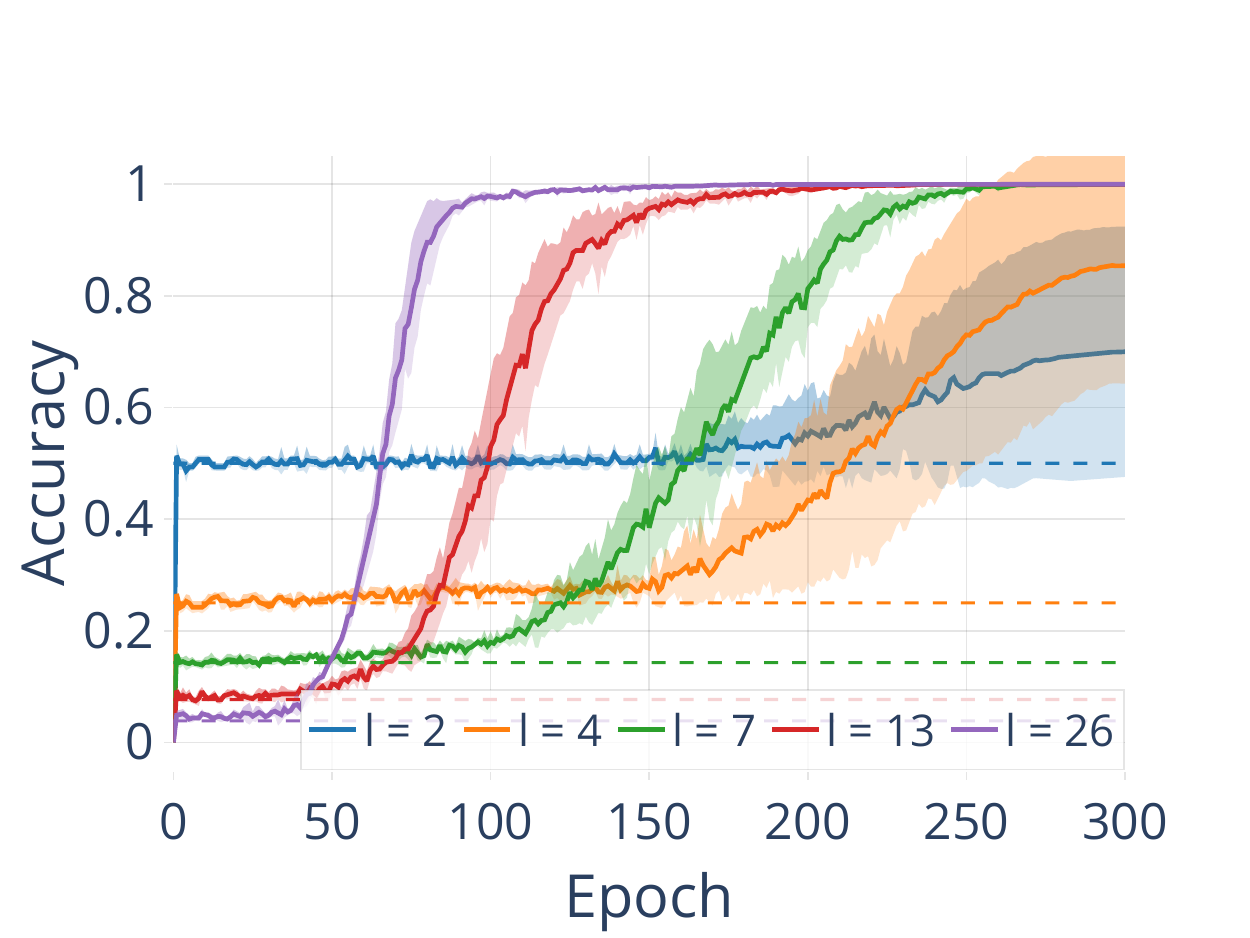}
    }
    \subfloat[Llama2-13B, Accuracy]{
        \includegraphics[width=\smallThirdWidth]{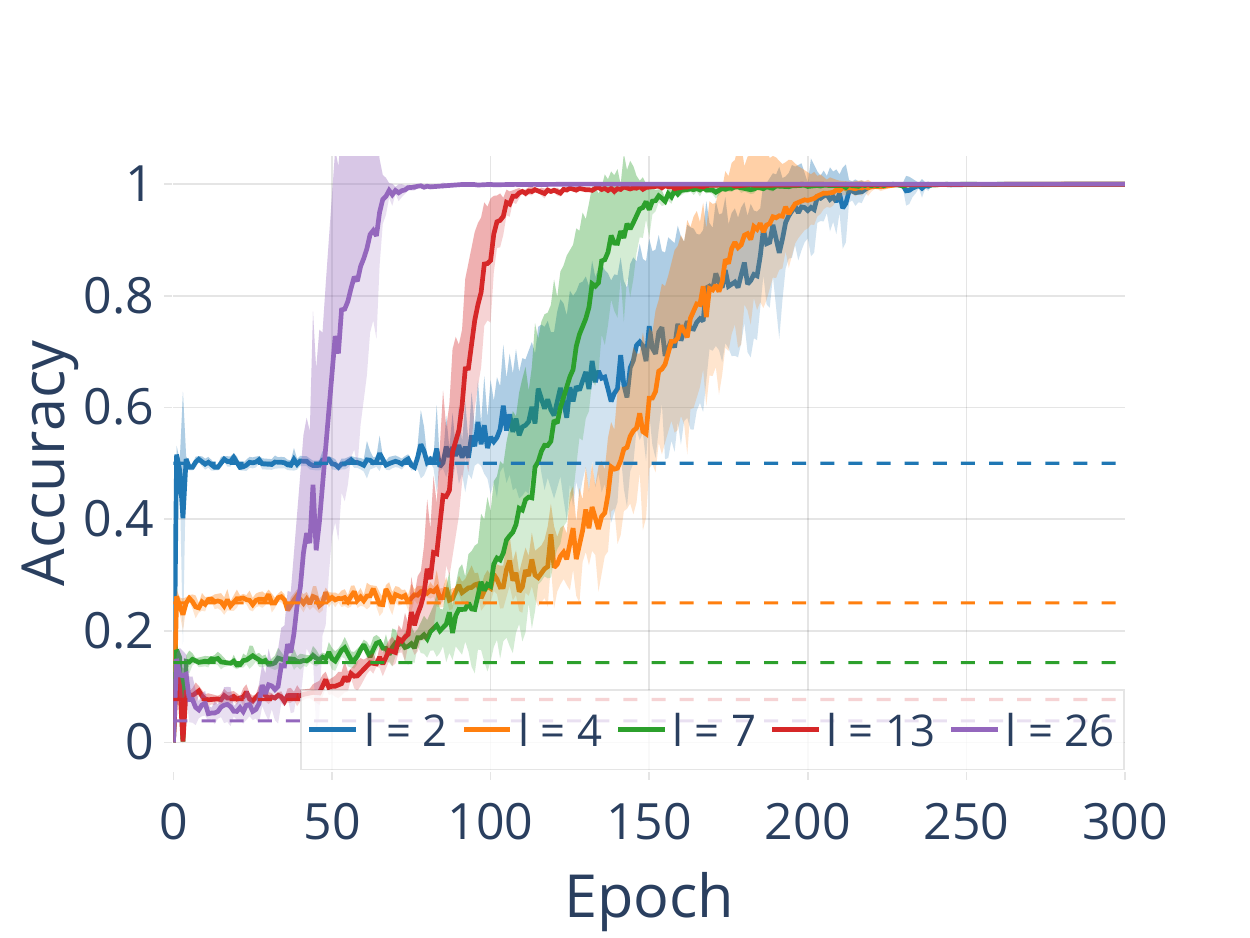}
    }
    \\
    \subfloat[Pythia-1B, Cumulative Probability]{
        \includegraphics[width=\smallThirdWidth]{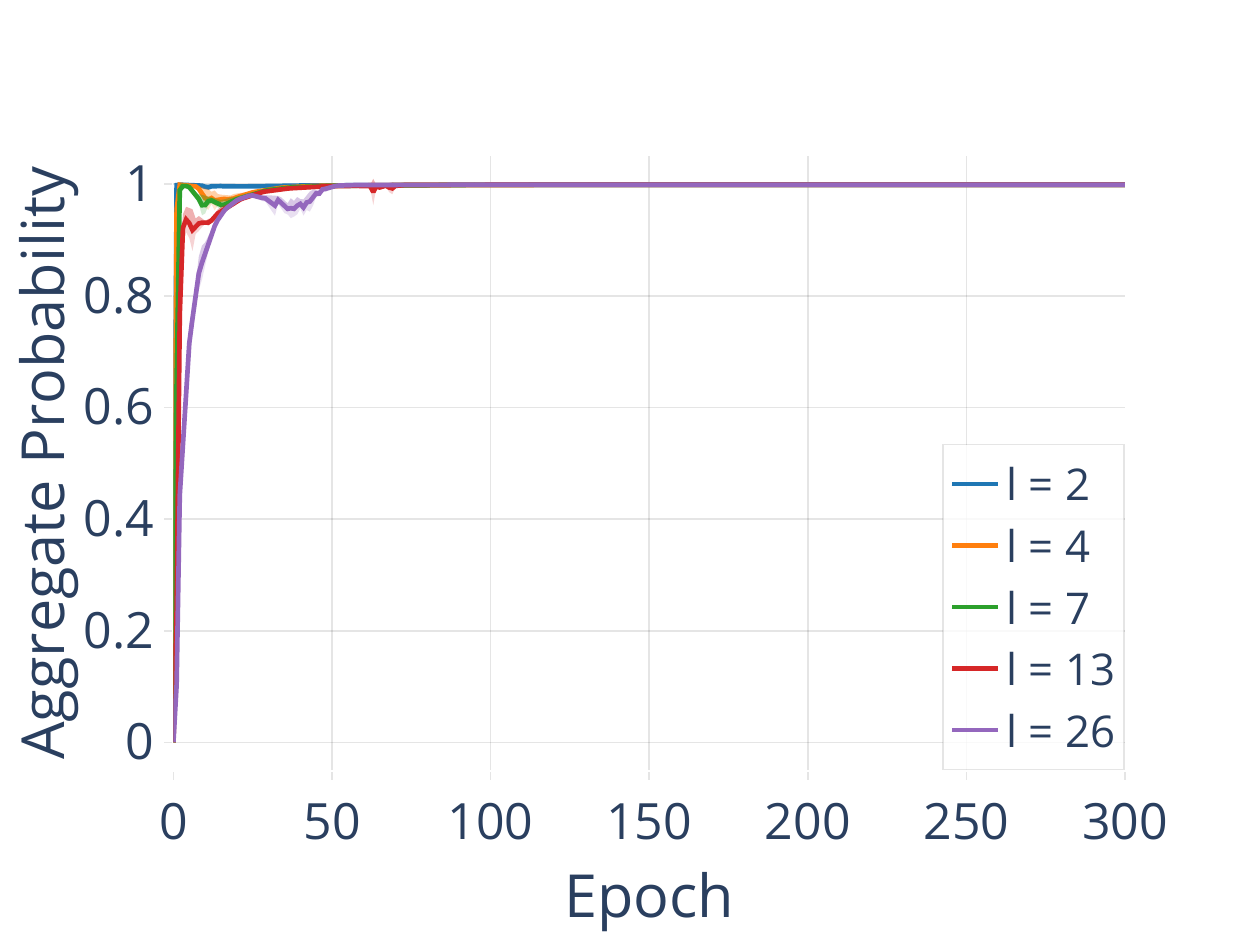}
    }
    \subfloat[Phi-2.7B, Cumulative Probability]{
        \includegraphics[width=\smallThirdWidth]{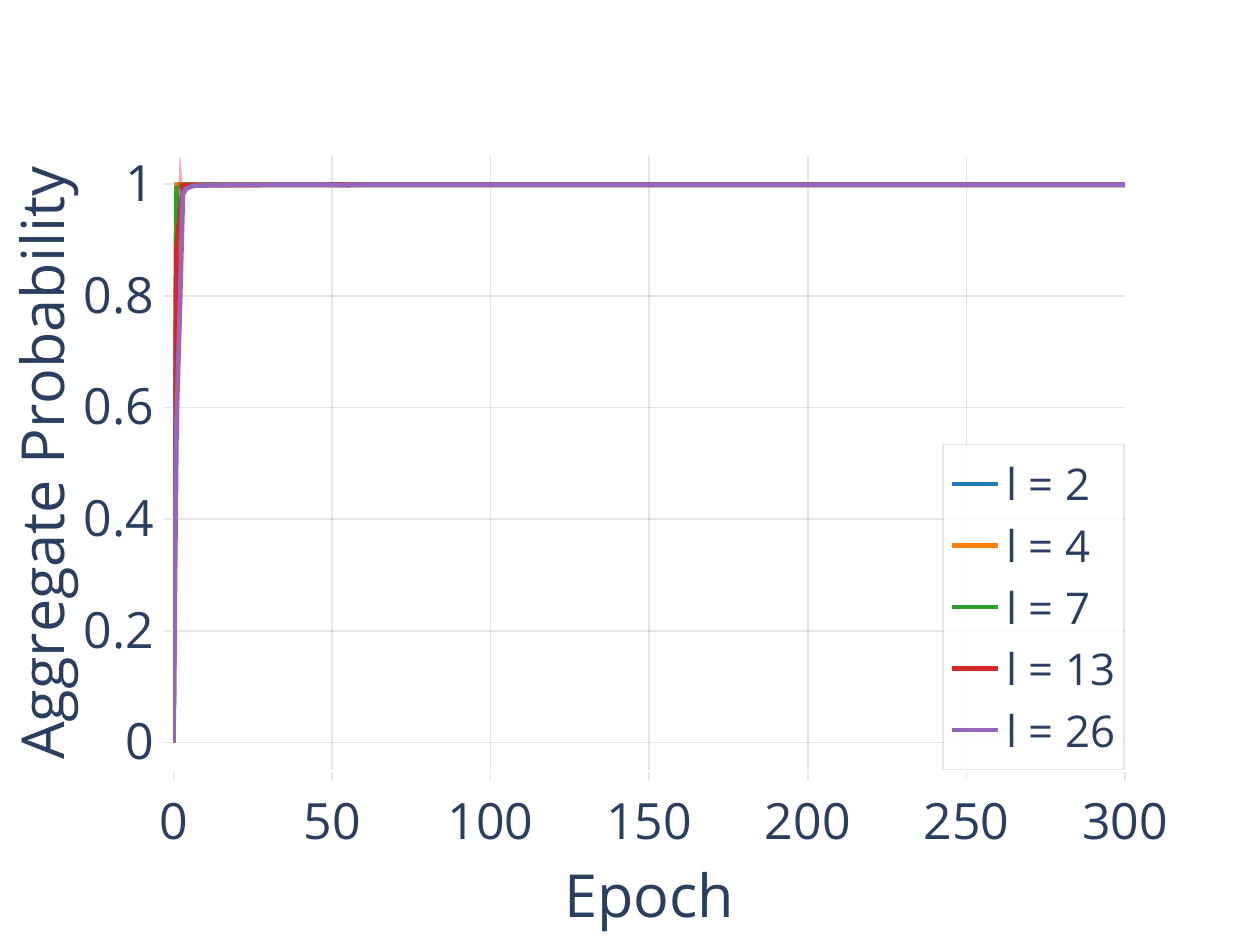}
    }
    \subfloat[Llama2-13B, Cumulative Probability]{
        \includegraphics[width=\smallThirdWidth]{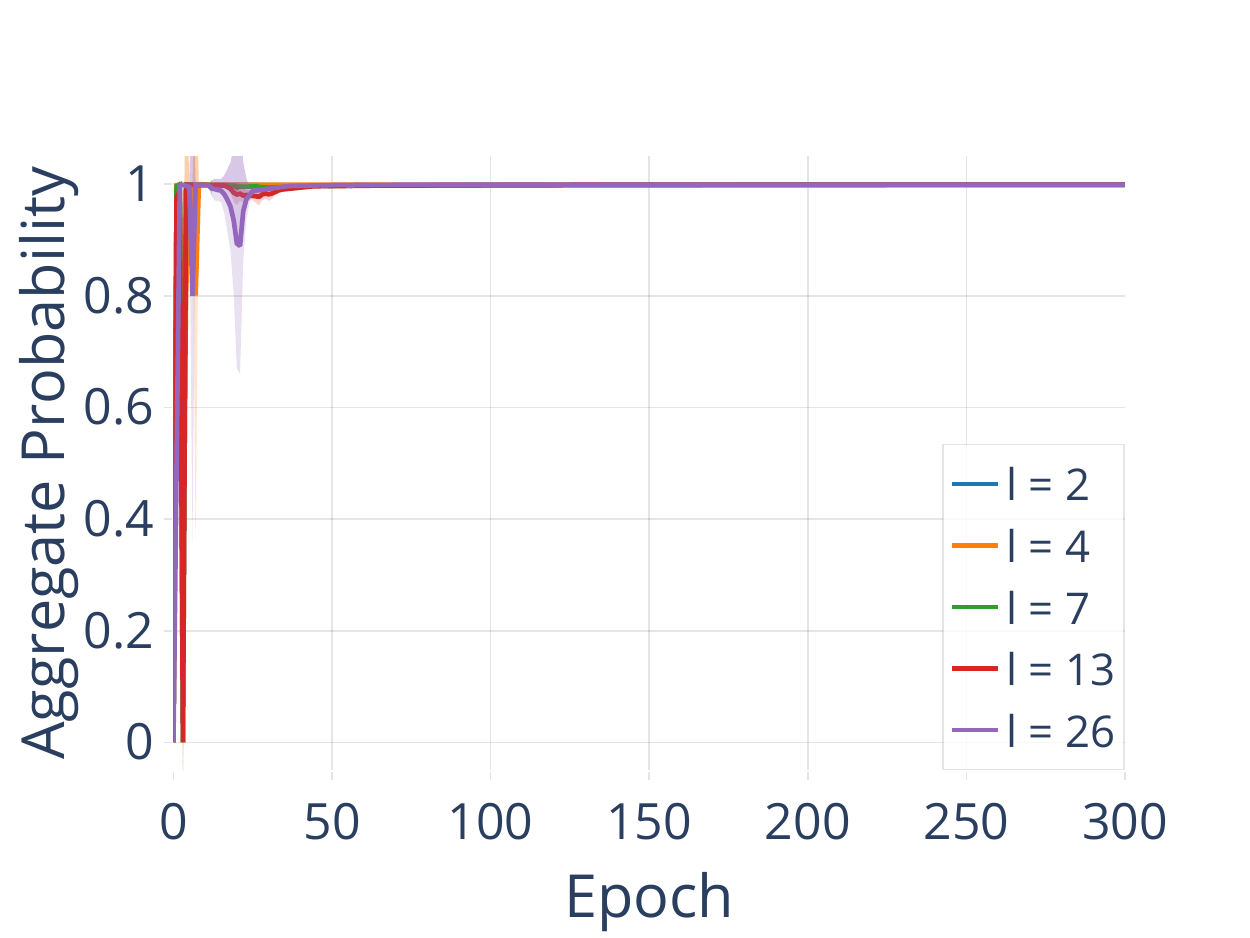}
    }
    \\
    \subfloat[Pythia-1B, Entropy]{
        \includegraphics[width=\smallThirdWidth]{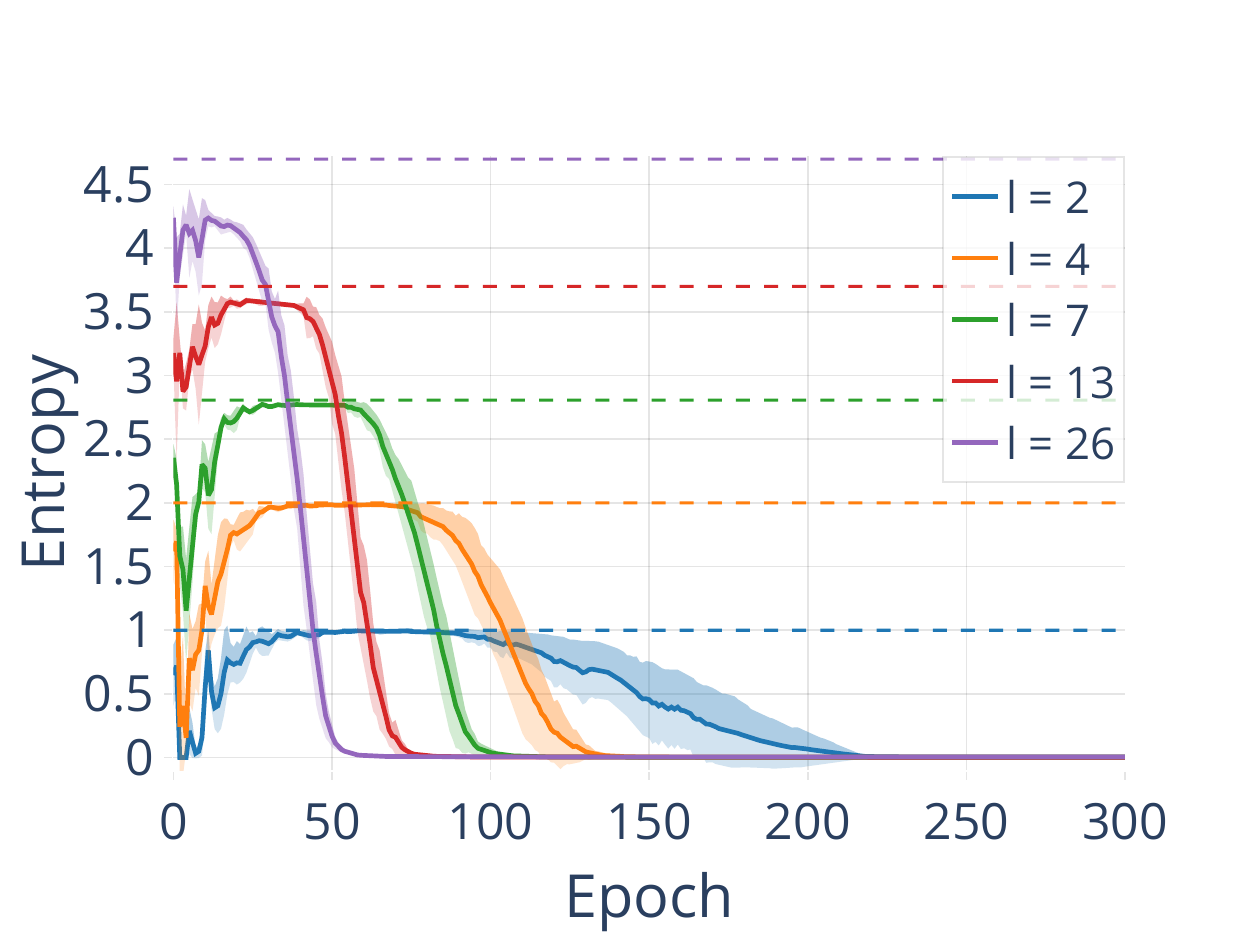}
    }
    \subfloat[Phi-2.7B, Entropy]{
        \includegraphics[width=\smallThirdWidth]{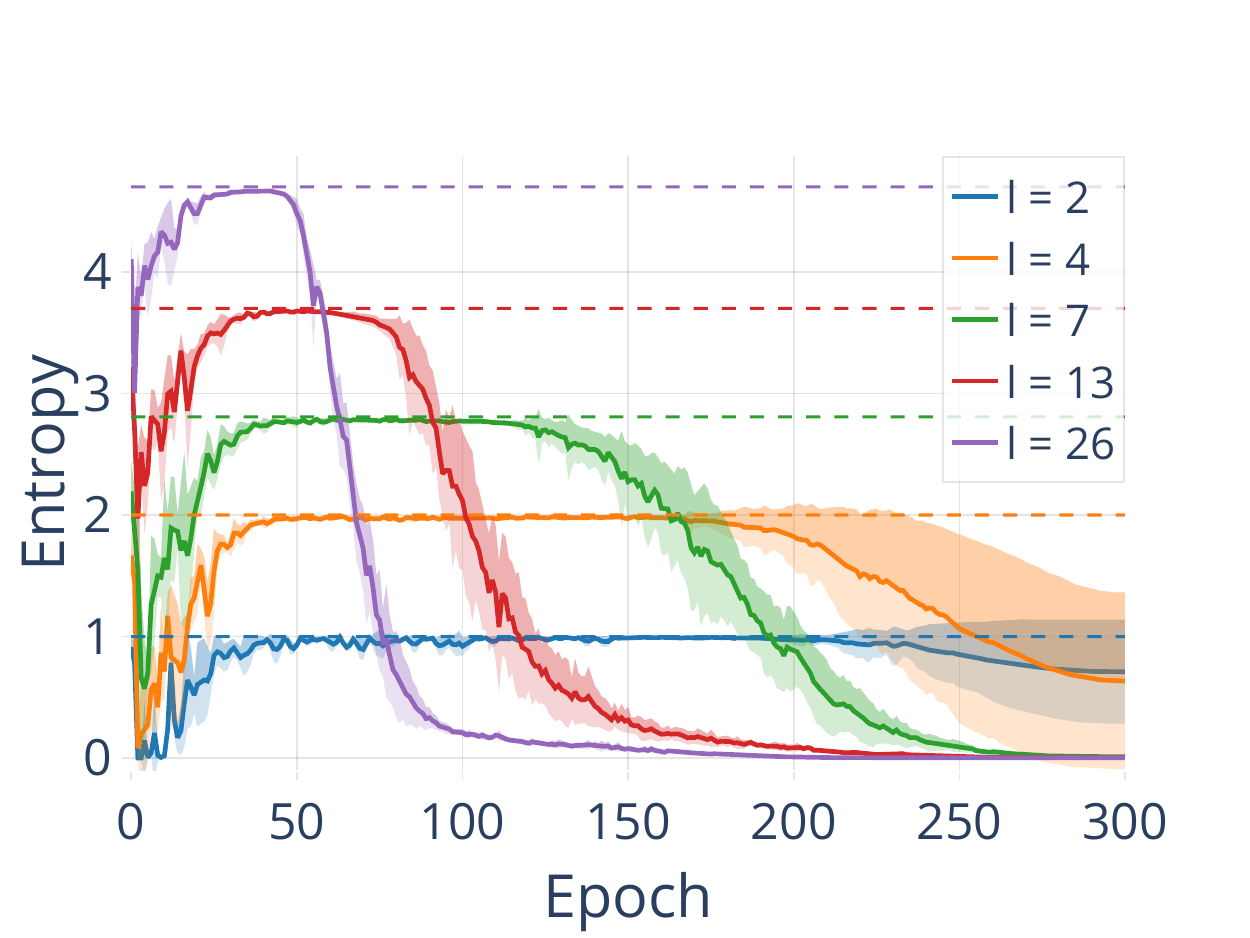}
    }
    \subfloat[Llama2-13B, Entropy]{
        \includegraphics[width=\smallThirdWidth]{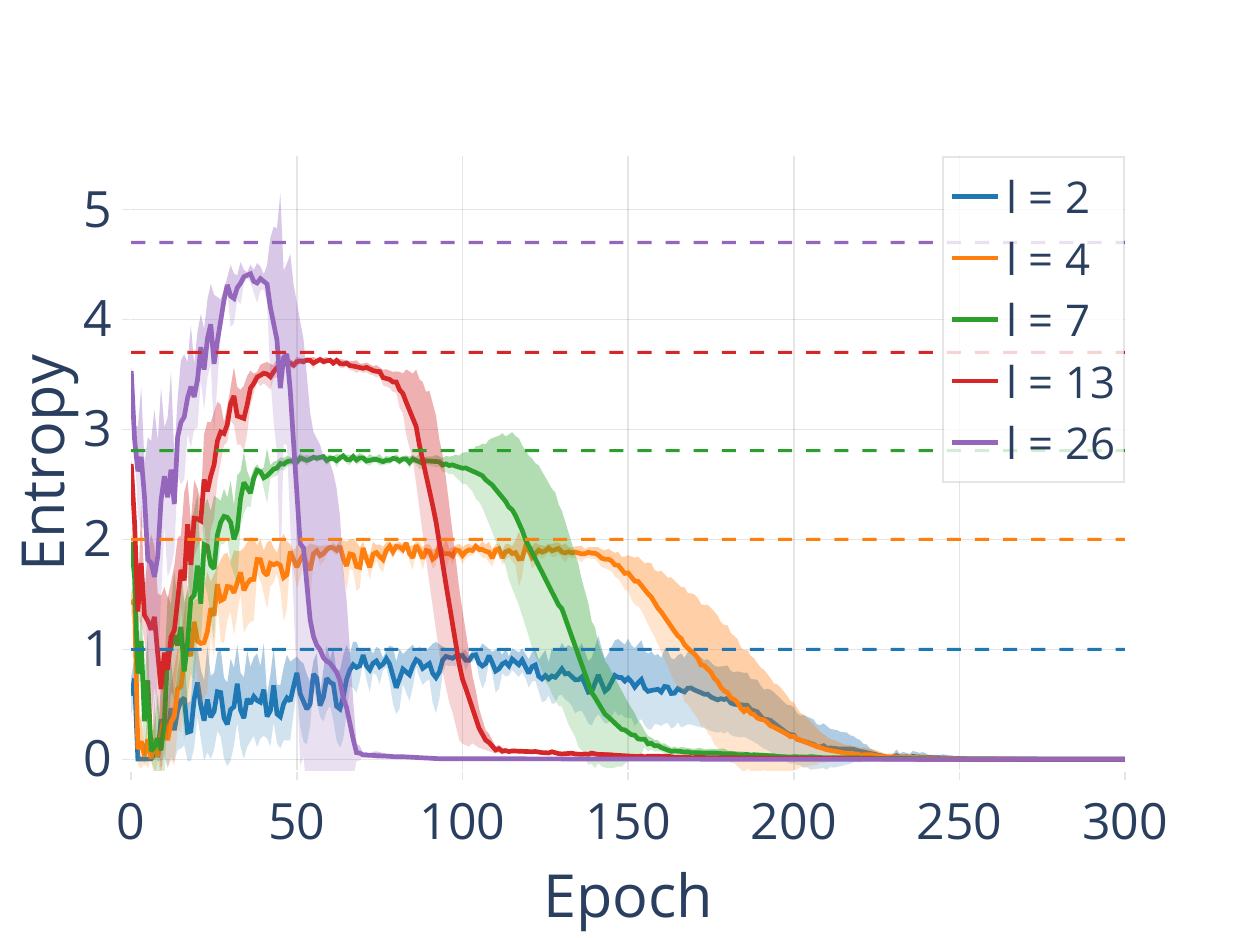}
    }
    \\
    \subfloat[Pythia-1B, KL-Divergence]{
        \includegraphics[width=\smallThirdWidth]{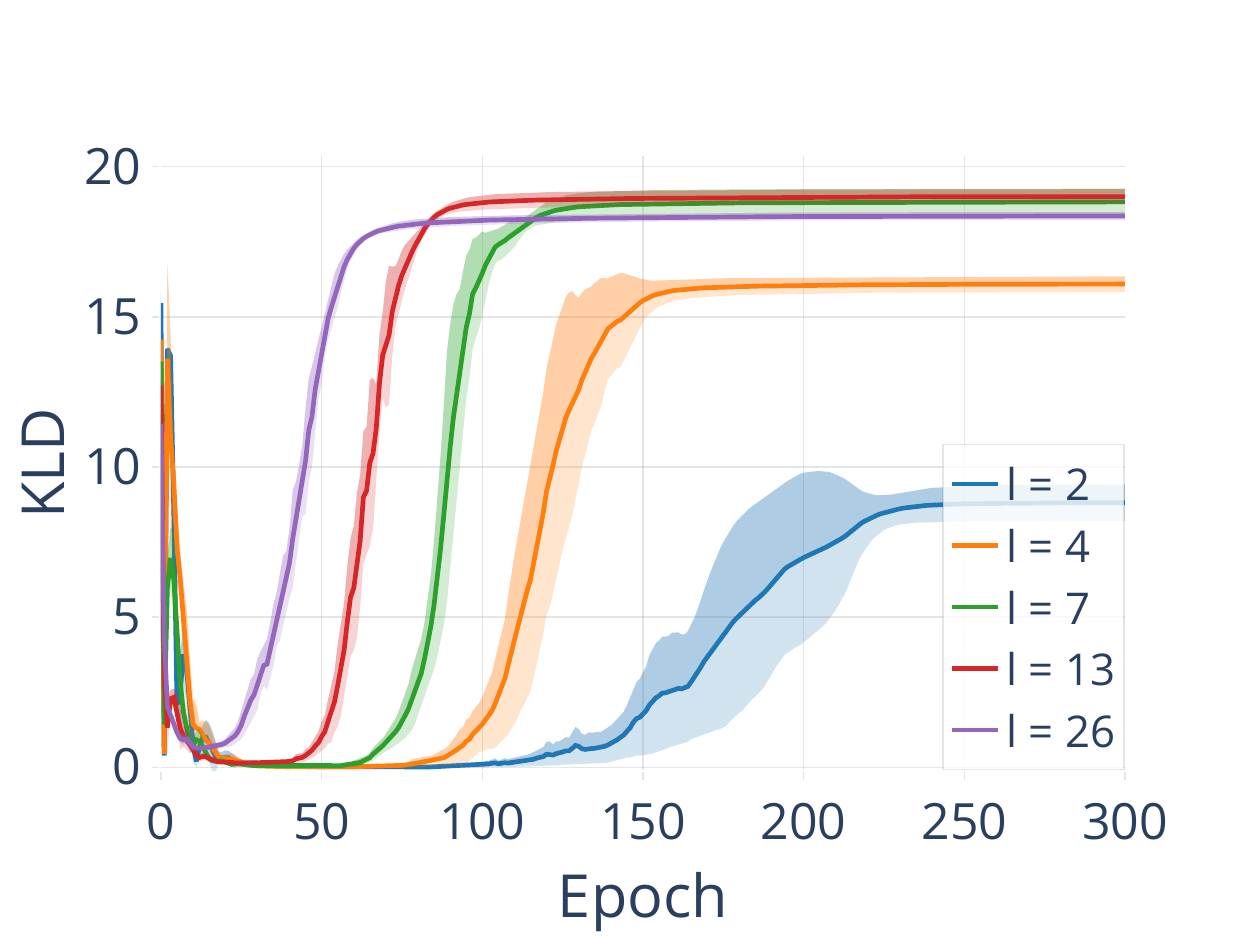}
    }
    \subfloat[Phi-2.7B, KL-Divergence]{
        \includegraphics[width=\smallThirdWidth]{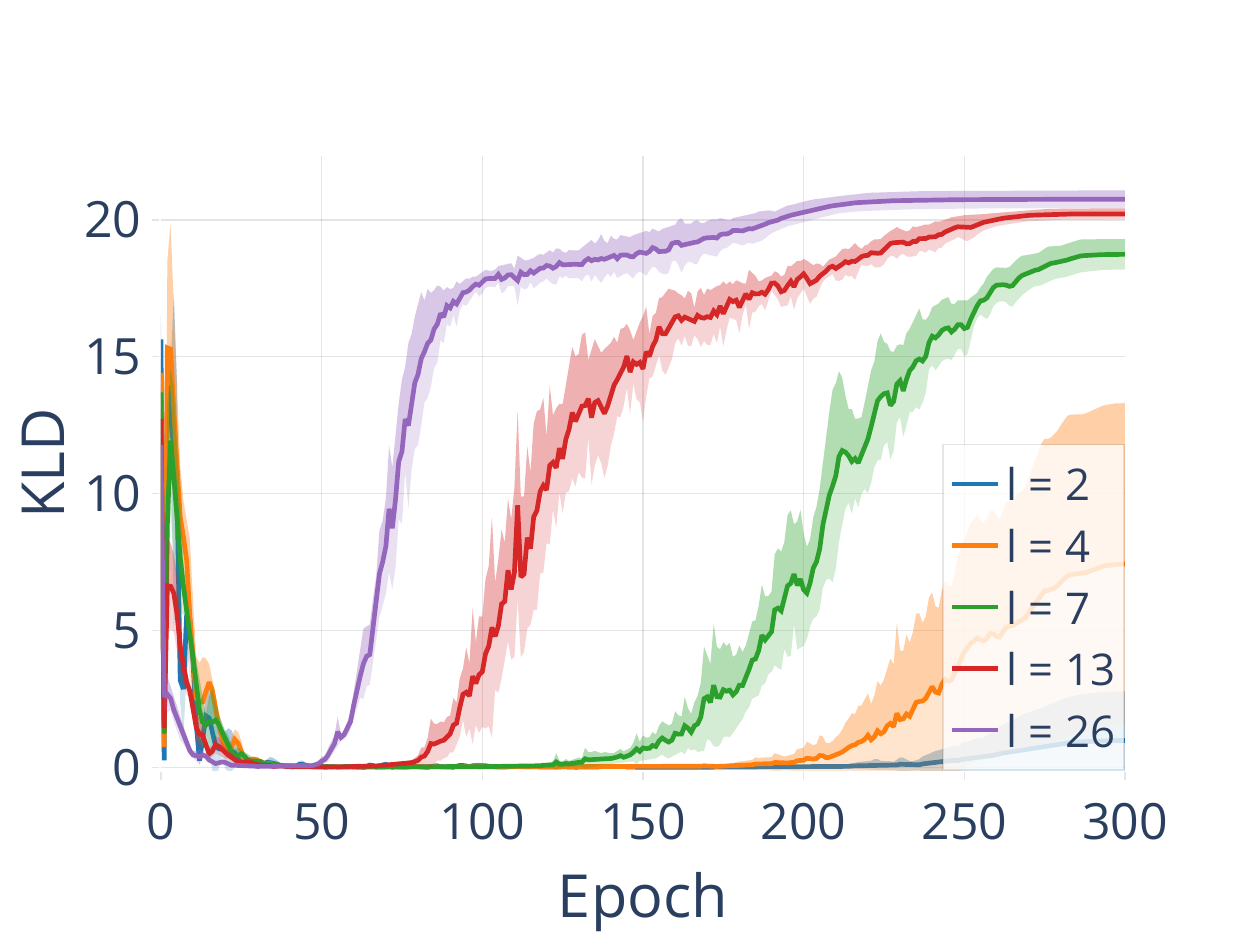}
    }
    \subfloat[Llama2-13B, KL-Divergence]{
        \includegraphics[width=\smallThirdWidth]{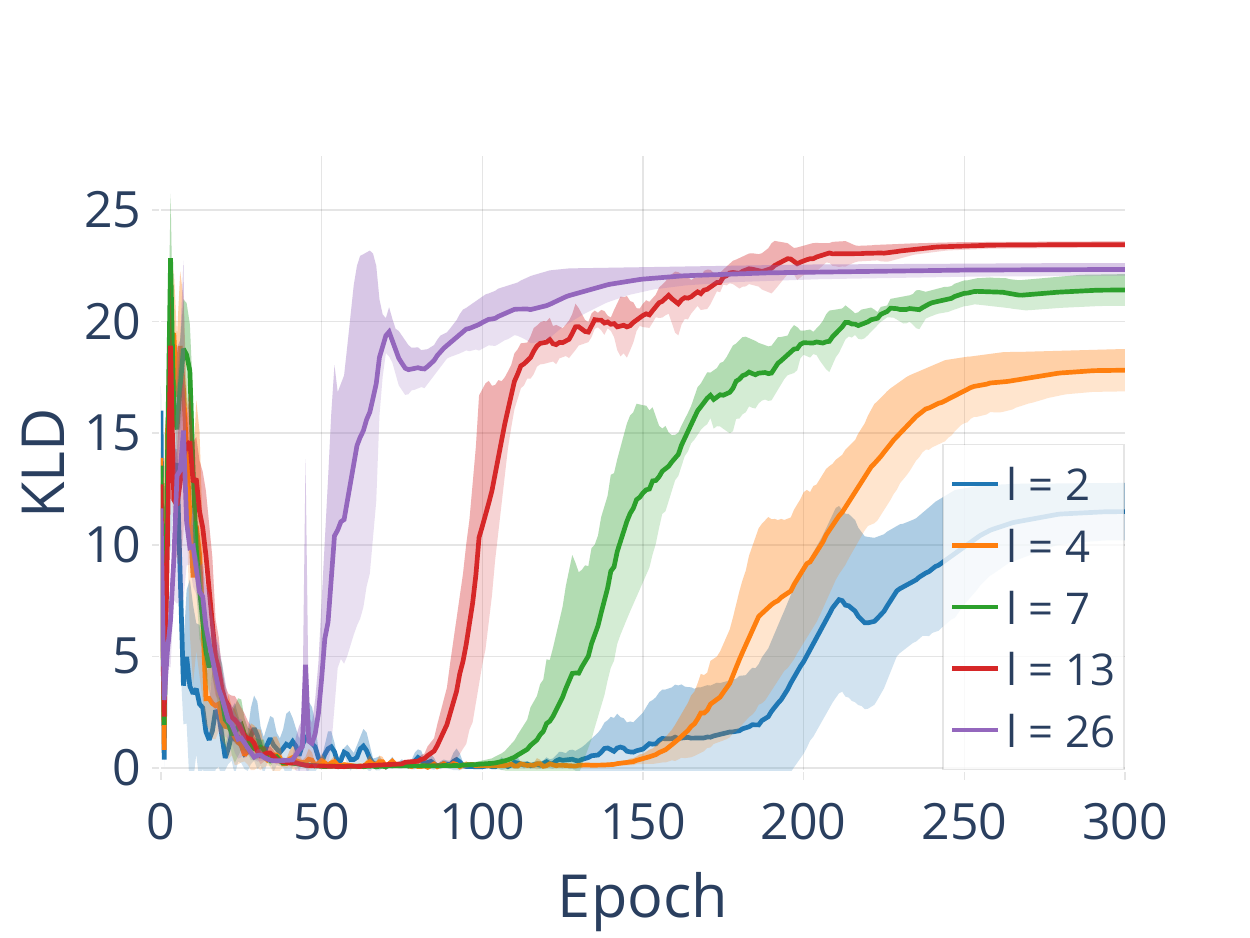}
    }
\caption{\capthead{Accuracy, loss, cumulative probability, entropy and KLD for different $\ell$ with non-pretrained models.}{$n = 1024$}
We observe the same patterns for untrained models as for the pretrained models shown in Sections~\ref{sec:phases} and~\ref{sec:memorability} and in Appendix~\ref{app:dynamics_additional_results}, although convergence is generally slower.
Note that we show memorisation dynamics over 300, instead of 100 epochs here.
}
\label{fig:untrained_all}
\end{figure}

In Figure~\ref{fig:untrained_all} we show ablations for untrained, \ie~non-pretrained models.
Again, we see the same patterns as for the pretrained models shown in Sections~\ref{sec:phases} and~\ref{sec:memorability} and in Appendix~\ref{app:dynamics_additional_results}, although convergence is generally slower.
Note that we show memorisation dynamics over 300, instead of 100 epochs.

\subsection{Additional results on string length, string partitions, and repeated substrings}
\label{app:dynamics_string_length}

\begin{figure}[H]
    \centering
    \subfloat[Pythia-1B, Loss]{
        \includegraphics[width=\thirdWidth]{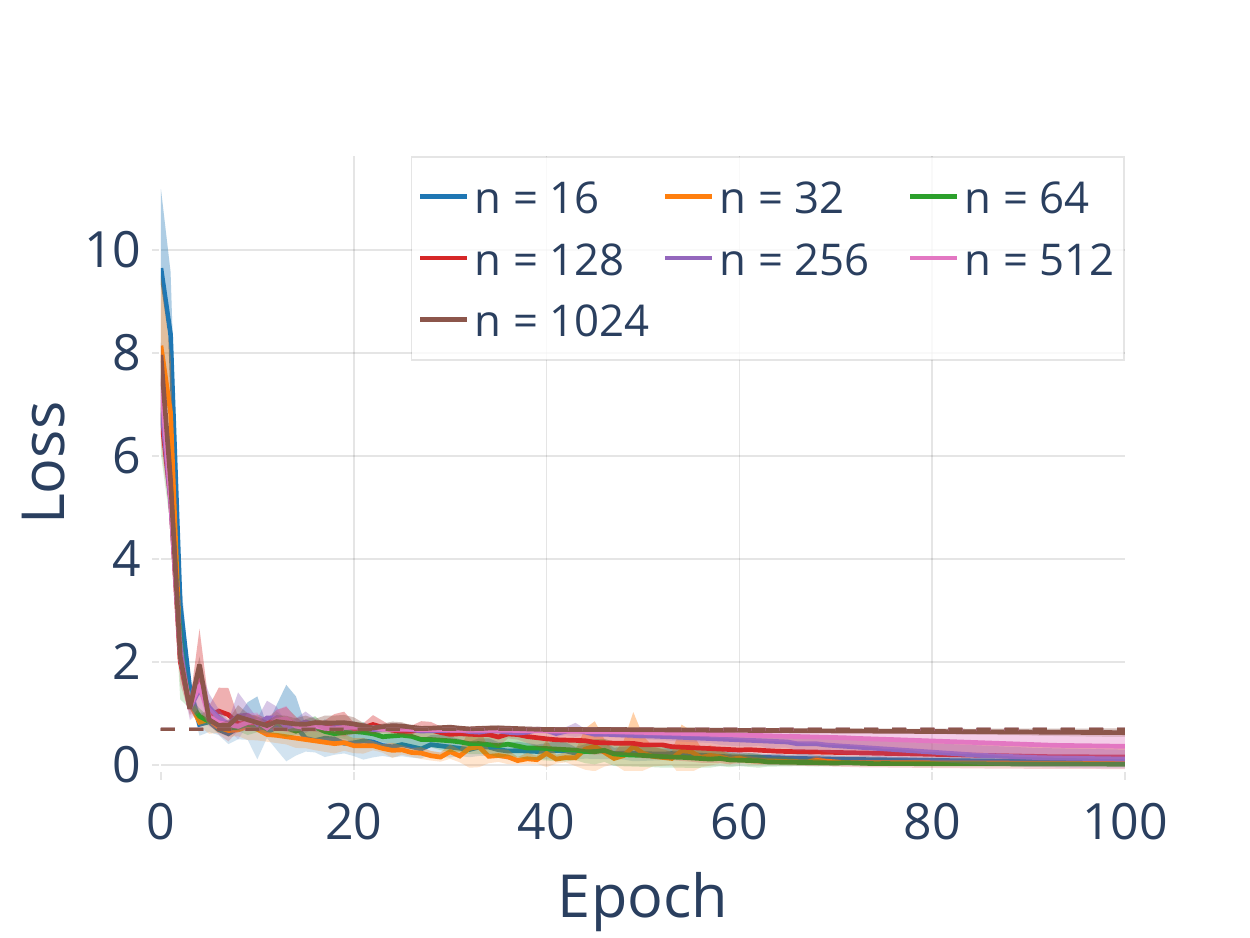}
    }
    \subfloat[Phi-2.7B, Loss]{
        \includegraphics[width=\thirdWidth]{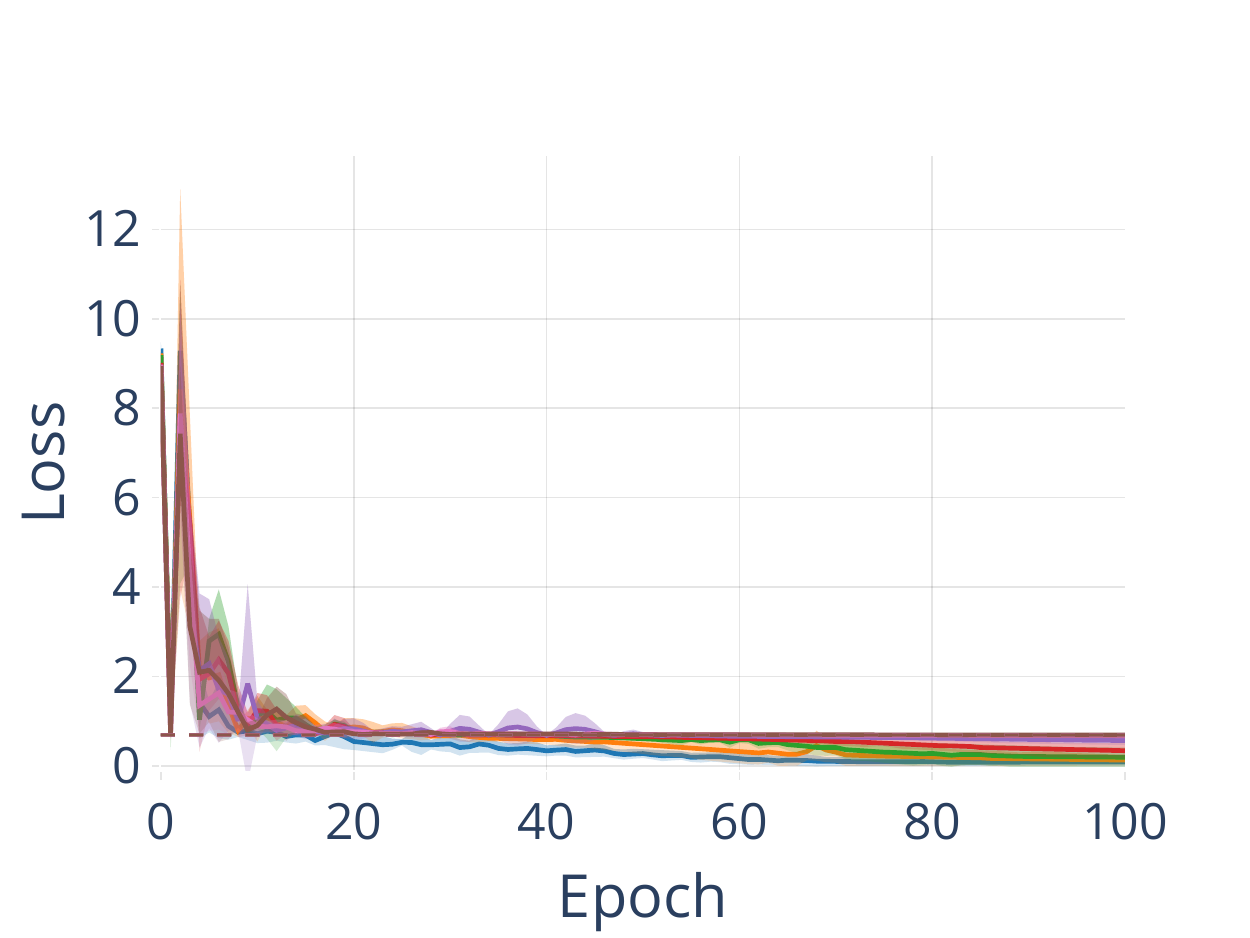}
    }
    \subfloat[Llama2-13B, Loss]{
        \includegraphics[width=\thirdWidth]{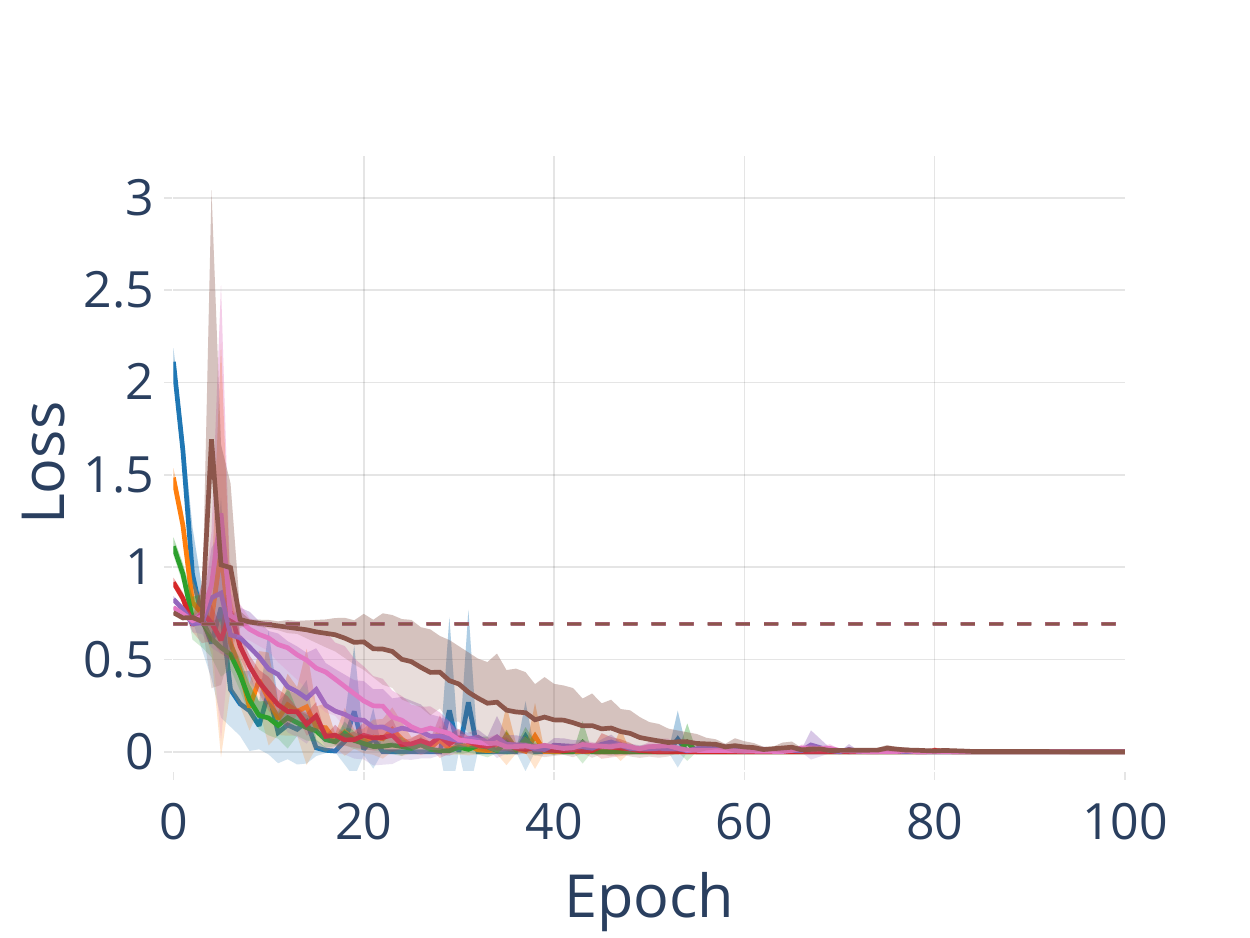}
    }
    \\
    \subfloat[Pythia-1B, Accuracy]{
        \includegraphics[width=\thirdWidth]{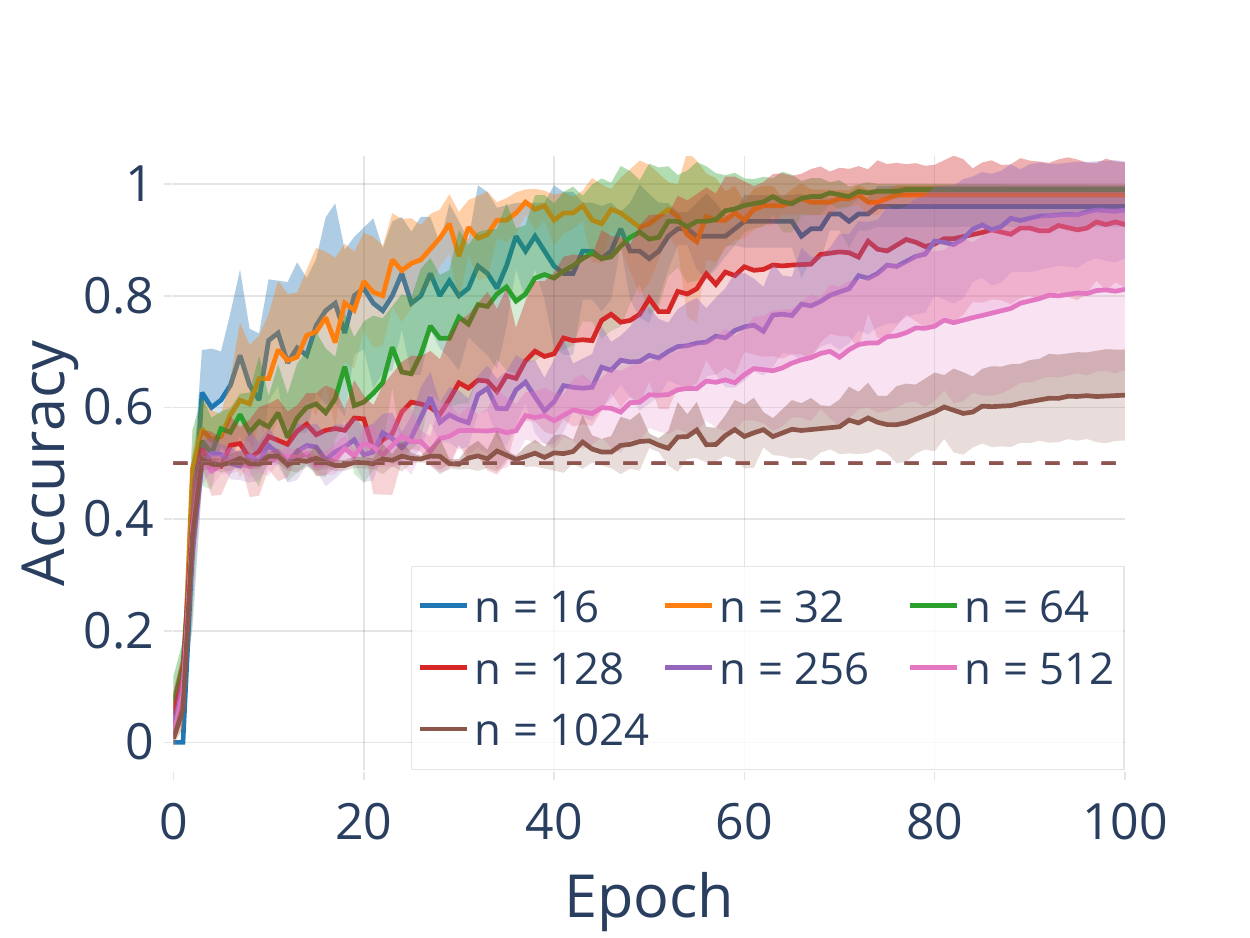}
    }
    \subfloat[Phi-2.7B, Accuracy]{
        \includegraphics[width=\thirdWidth]{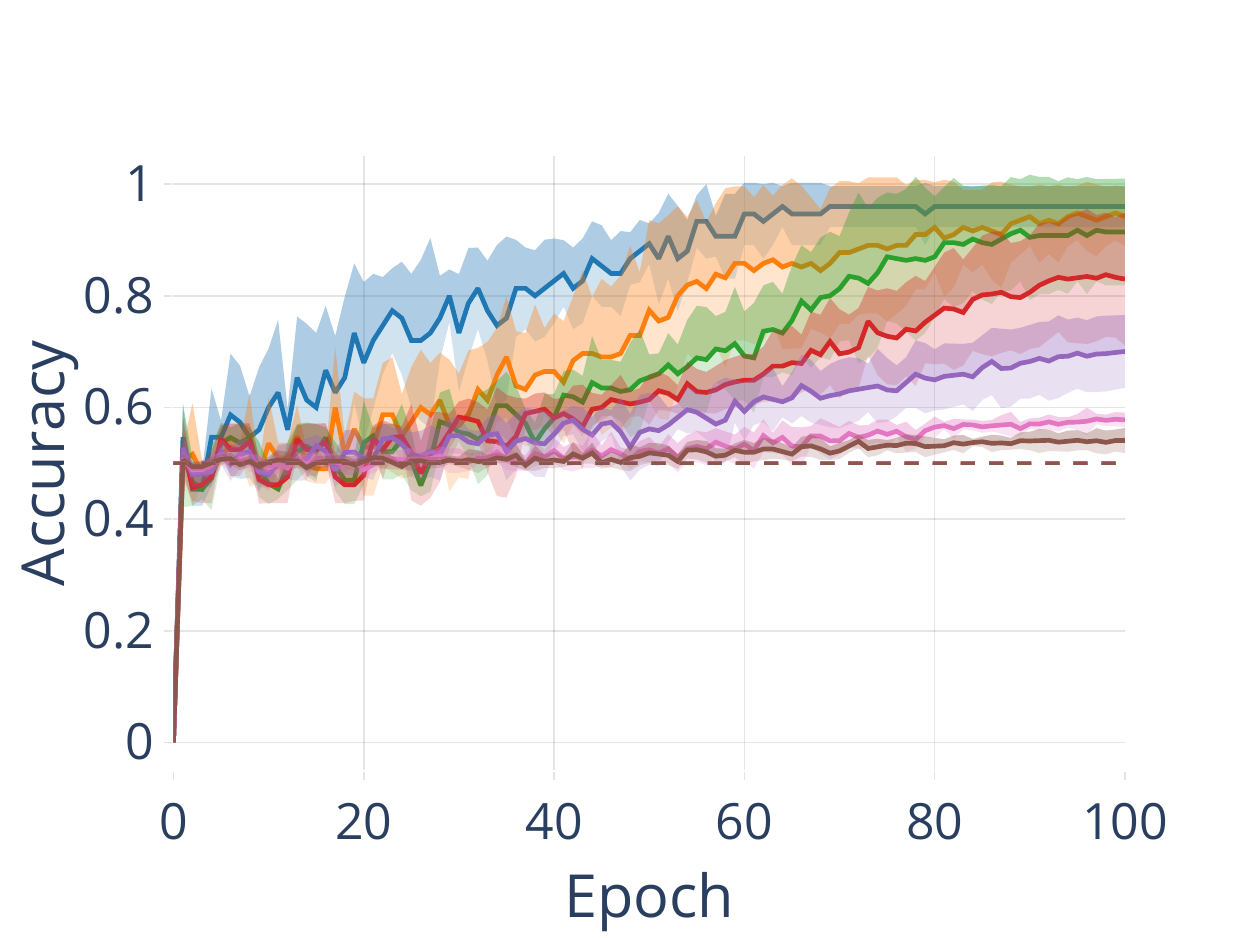}
    }
    \subfloat[Llama2-13B, Accuracy]{
        \includegraphics[width=\thirdWidth]{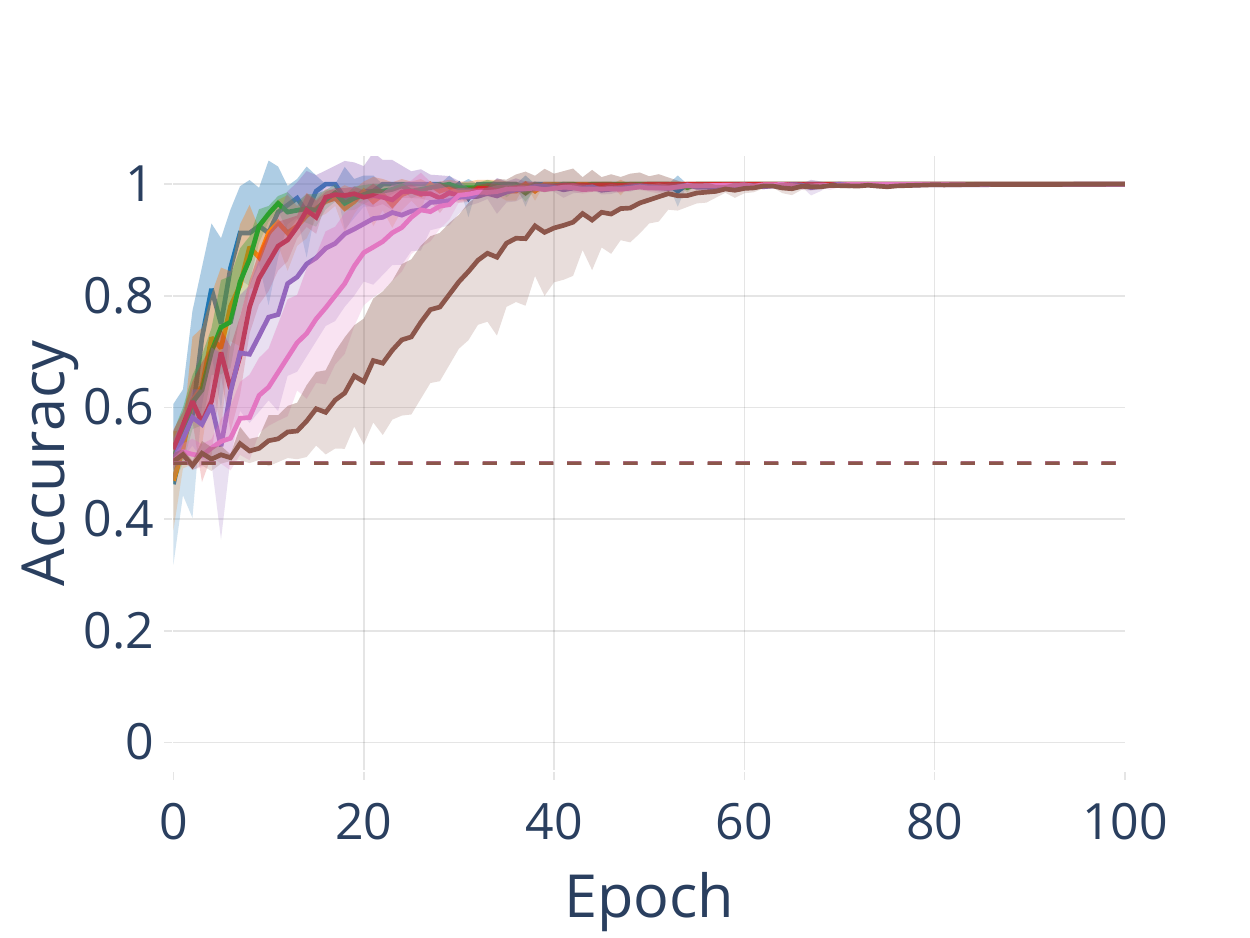}
    }
\caption{\capthead{Accuracy and loss for different string lengths $n$.}{$\ell = 2$}
}
\label{fig:string_length_a2_all}
\end{figure}

\begin{figure}[H]
    \centering
    \subfloat[Pythia-1B, Loss]{
        \includegraphics[width=\thirdWidth]{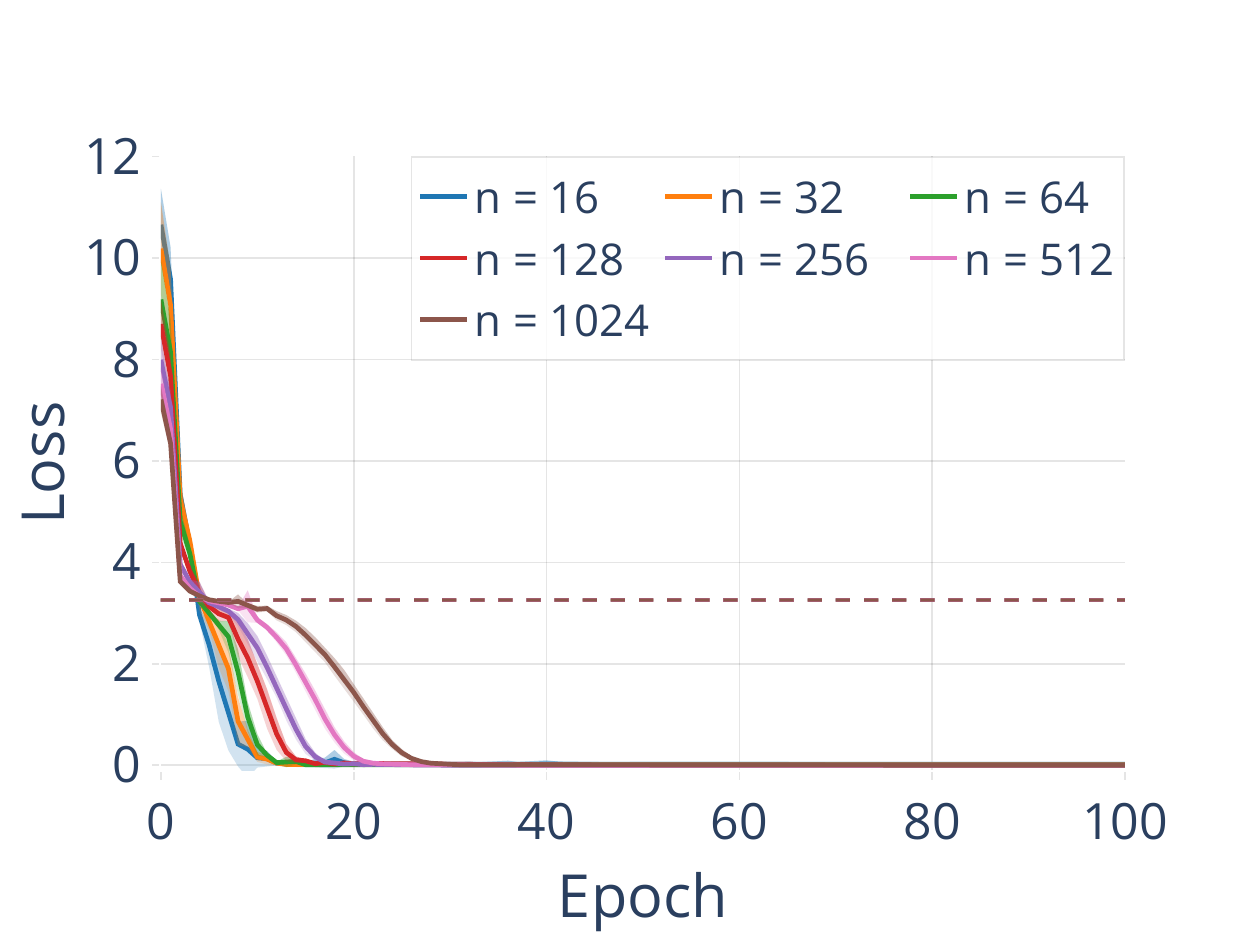}
    }
    \subfloat[Phi-2.7B, Loss]{
        \includegraphics[width=\thirdWidth]{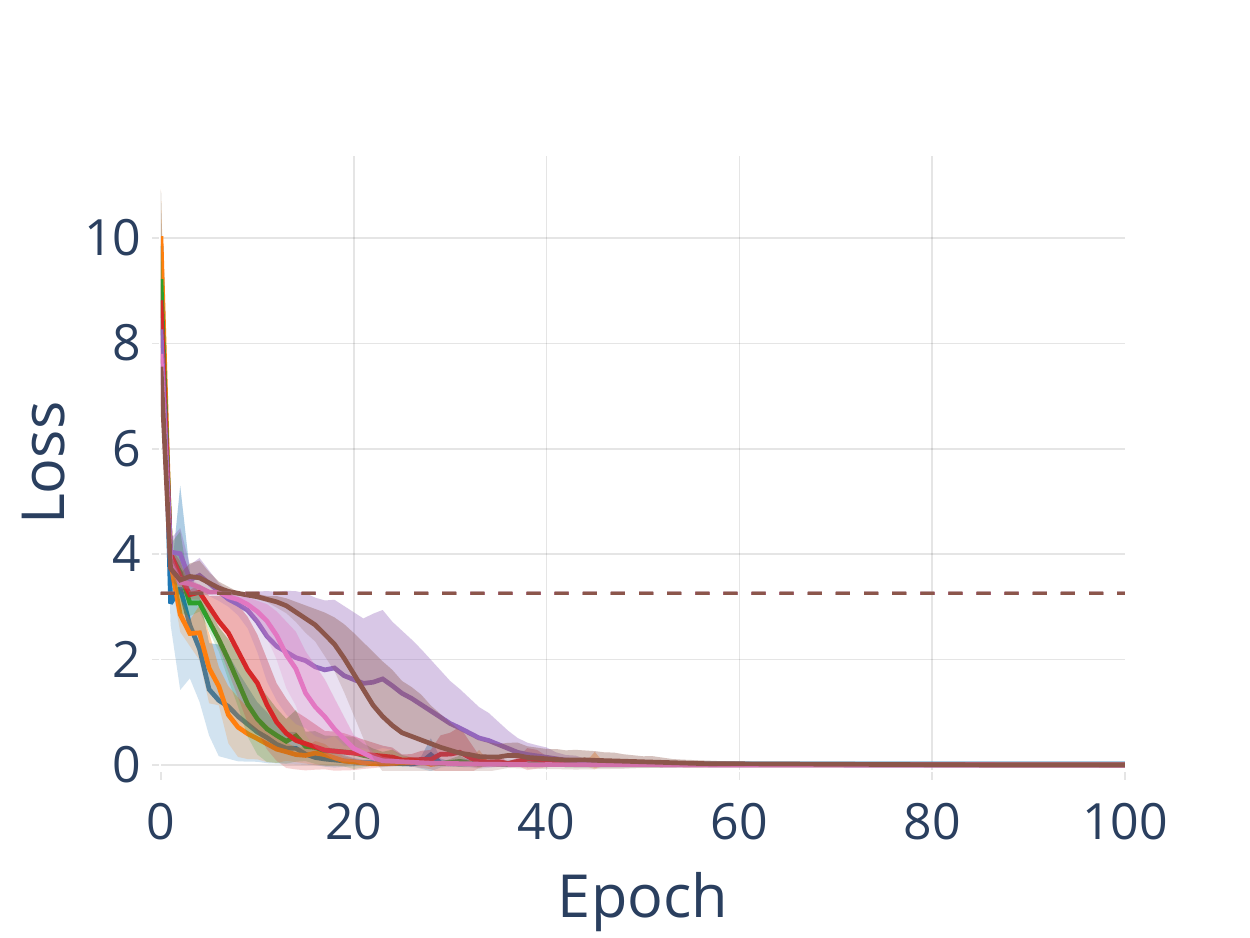}
    }
    \subfloat[Llama2-13B, Loss]{
        \includegraphics[width=\thirdWidth]{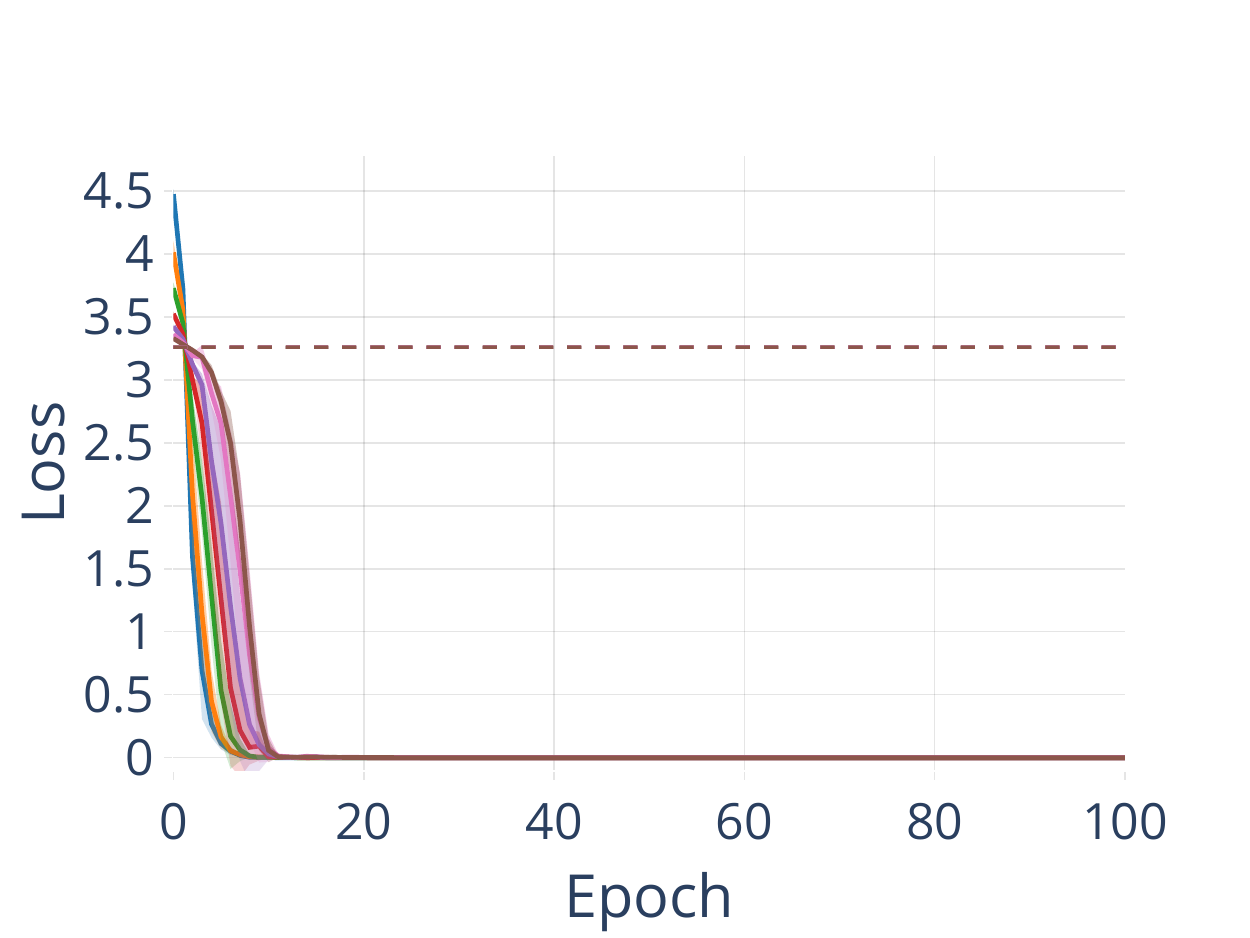}
    }
    \\
    \subfloat[Pythia-1B, Accuracy]{
        \includegraphics[width=\thirdWidth]{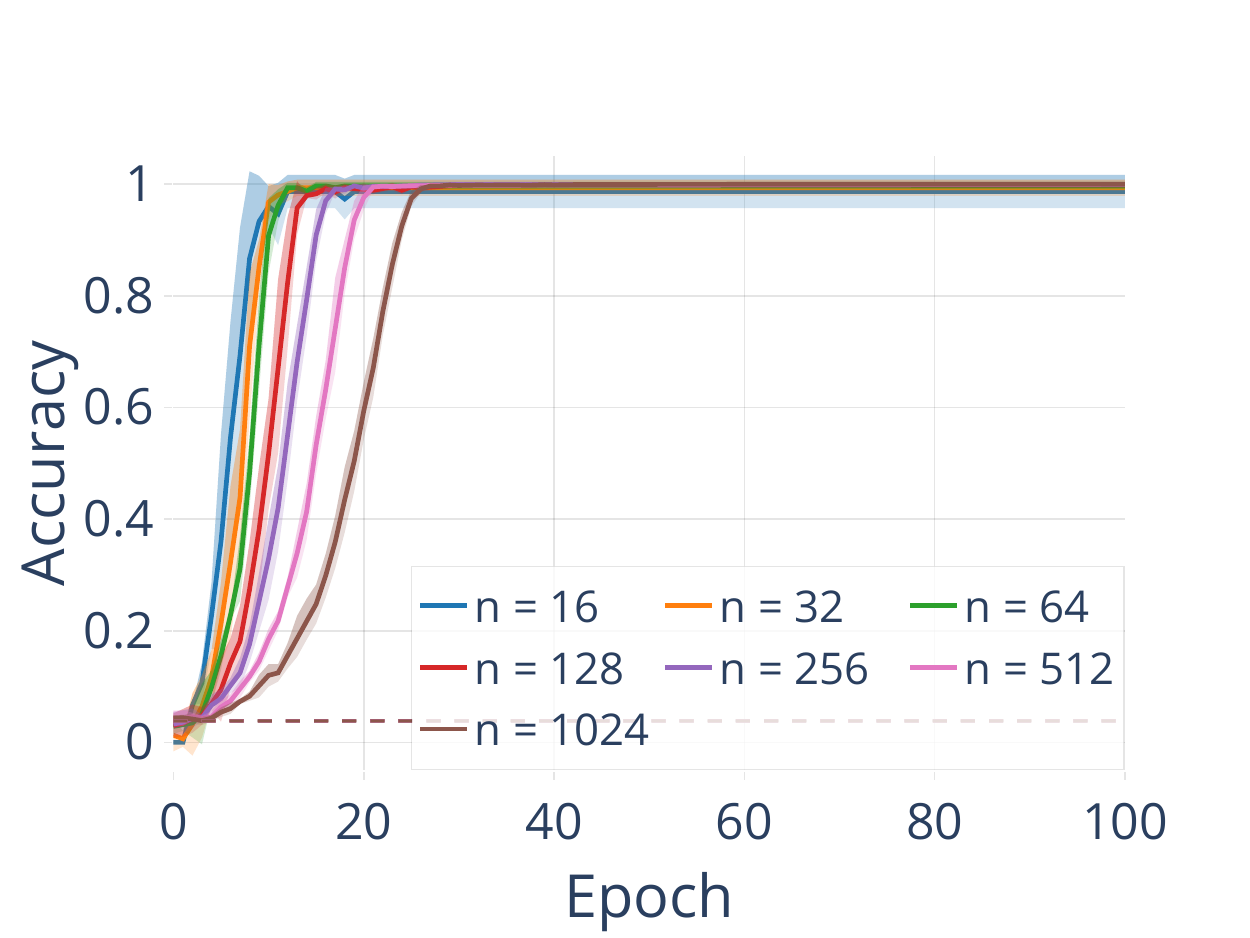}
    }
    \subfloat[Phi-2.7B, Accuracy]{
        \includegraphics[width=\thirdWidth]{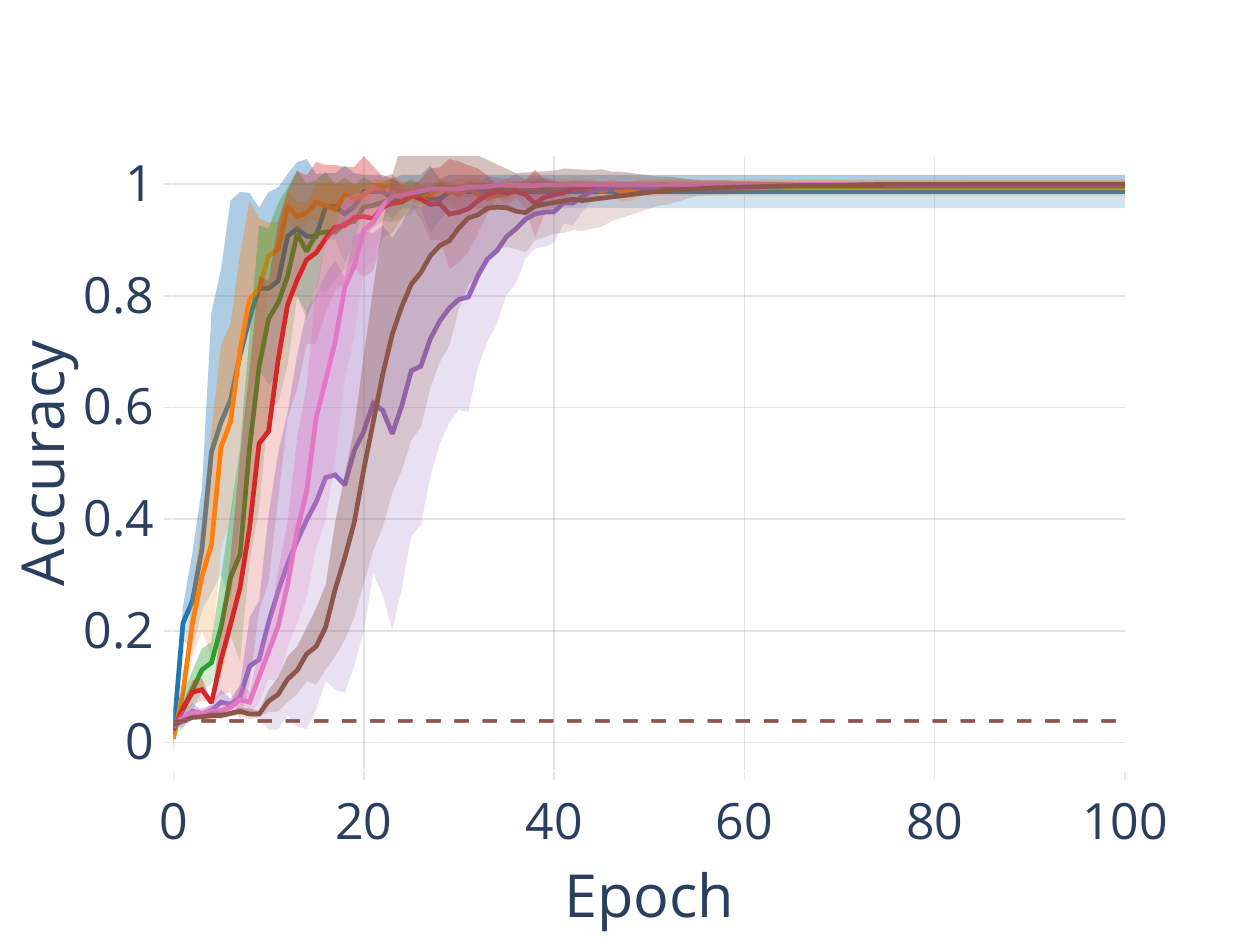}
    }
    \subfloat[Llama2-13B, Accuracy]{
        \includegraphics[width=\thirdWidth]{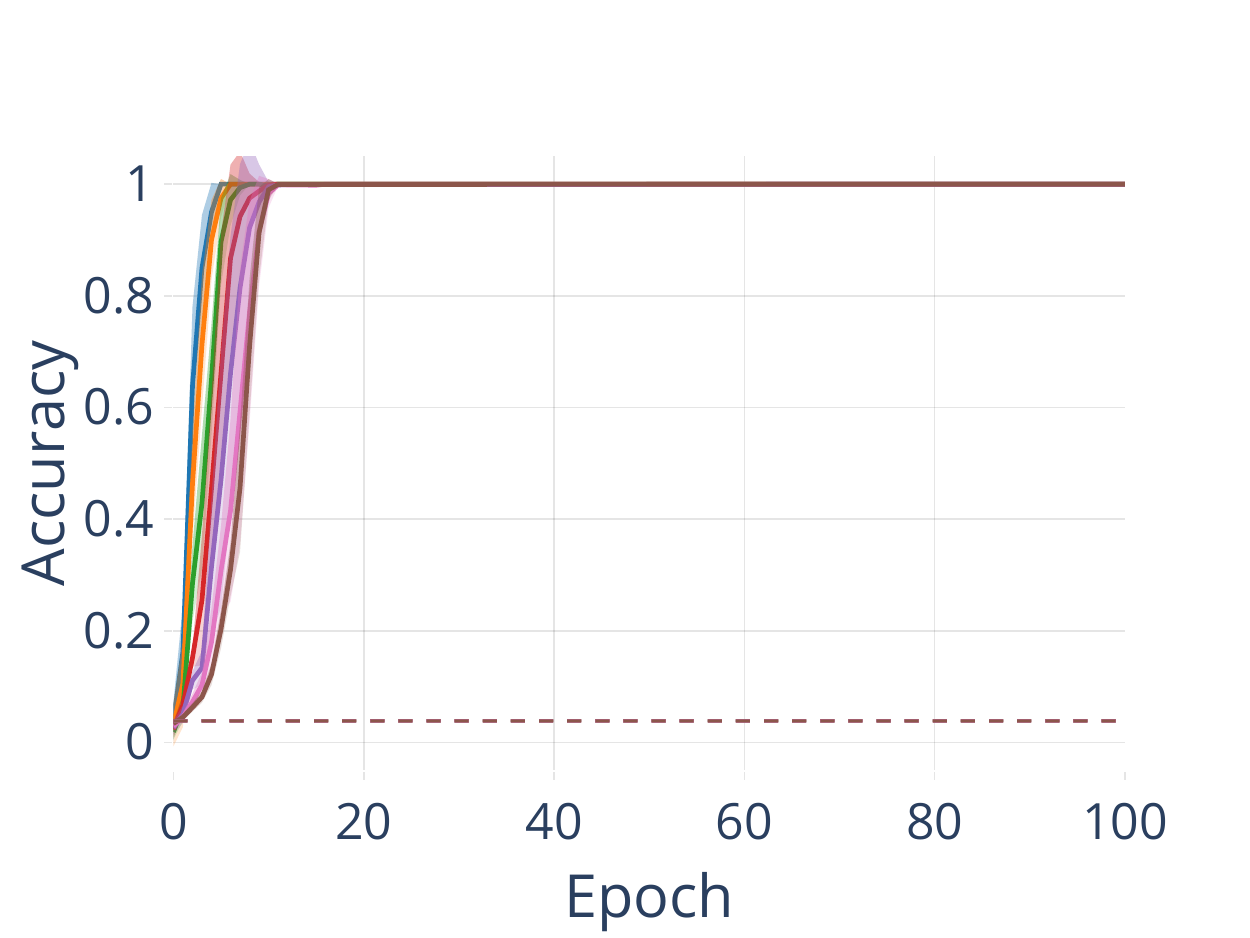}
    }
\caption{\capthead{Accuracy and loss for different string lengths $n$.}{$\ell = 26$}
}
\label{fig:string_length_a26_all}
\end{figure}

We show results for $\ell = 2$ and $\ell = 26$.
In Figures~\ref{fig:string_length_a2_all} and~\ref{fig:string_length_a26_all} we show the effect of string length on the memorisation dynamics.
Shorter strings are memorised faster than longer ones.

\begin{figure}[H]
    \centering
    \subfloat[Pythia-1B, Loss]{
        \includegraphics[width=\thirdWidth]{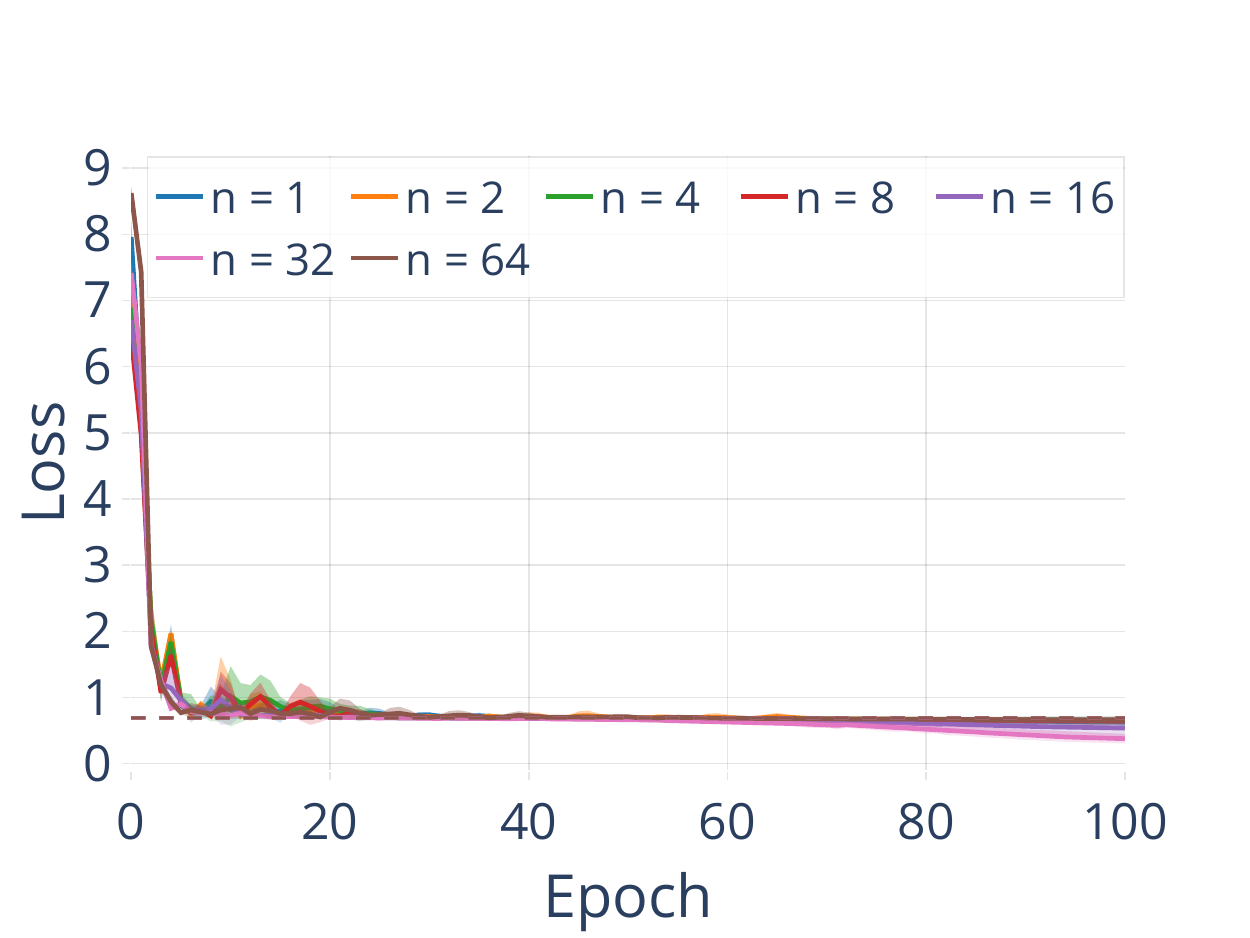}
    }
    \subfloat[Phi-2.7B, Loss]{
        \includegraphics[width=\thirdWidth]{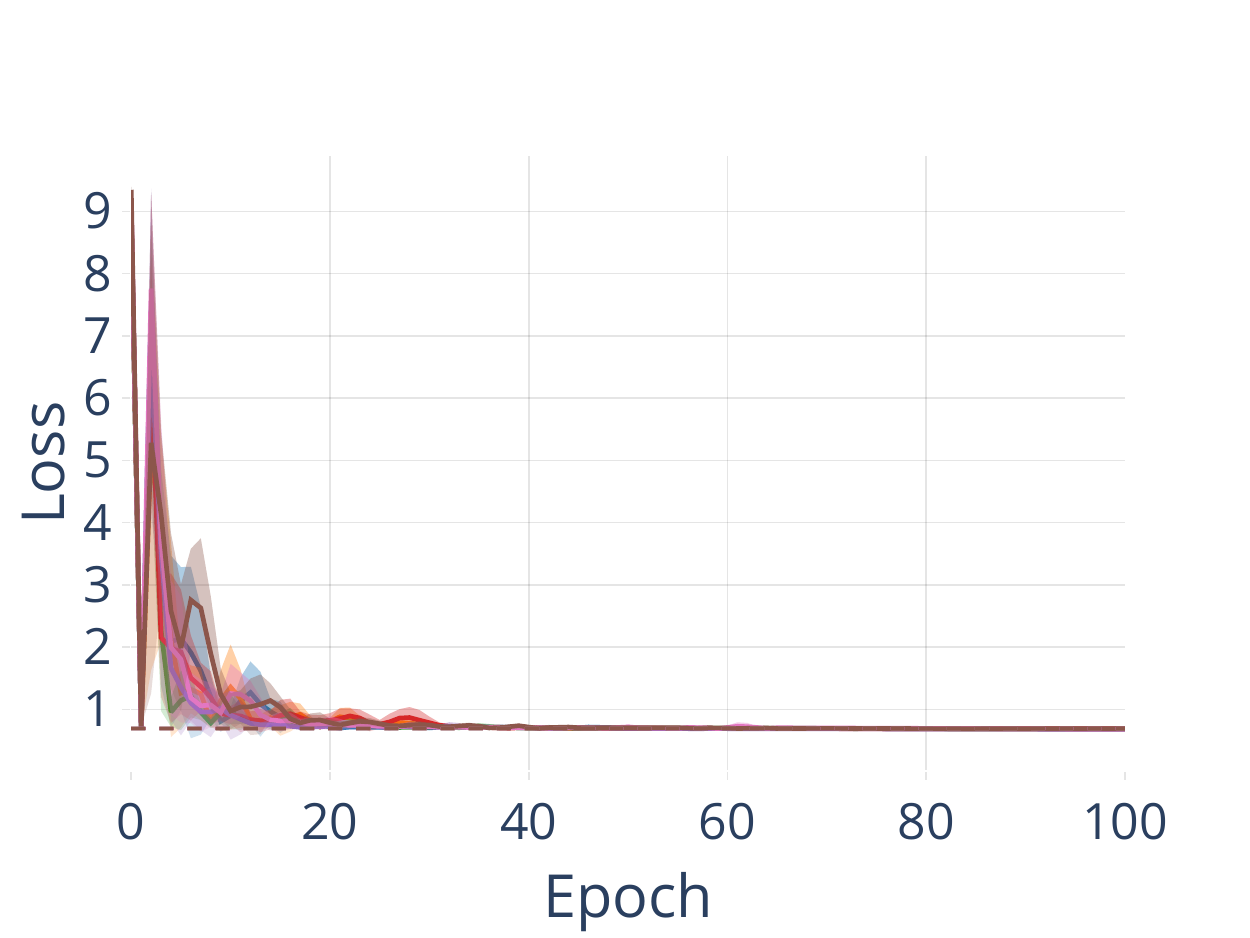}
    }
    \subfloat[Llama2-13B, Loss]{
        \includegraphics[width=\thirdWidth]{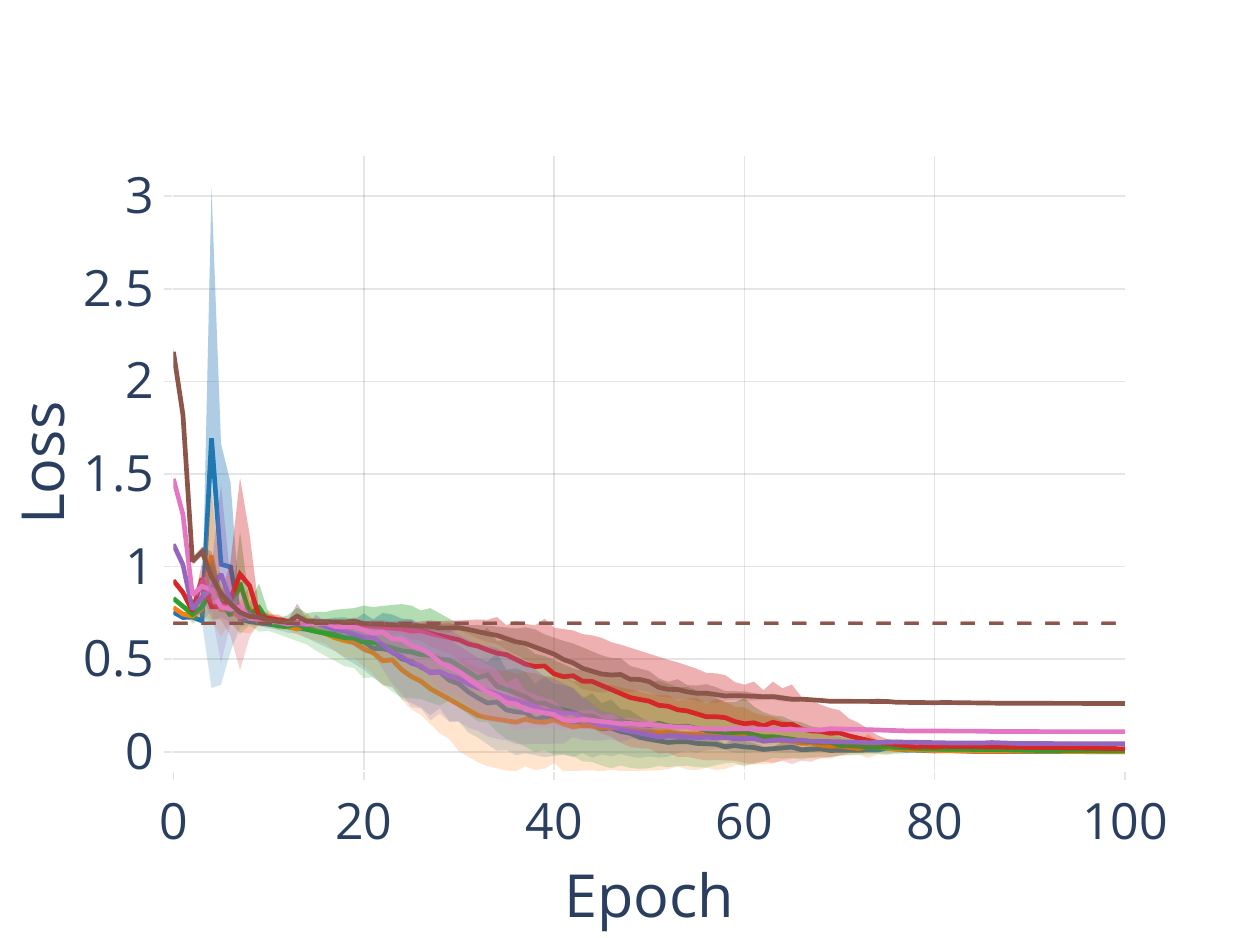}
    }
    \\
    \subfloat[Pythia-1B, Accuracy]{
        \includegraphics[width=\thirdWidth]{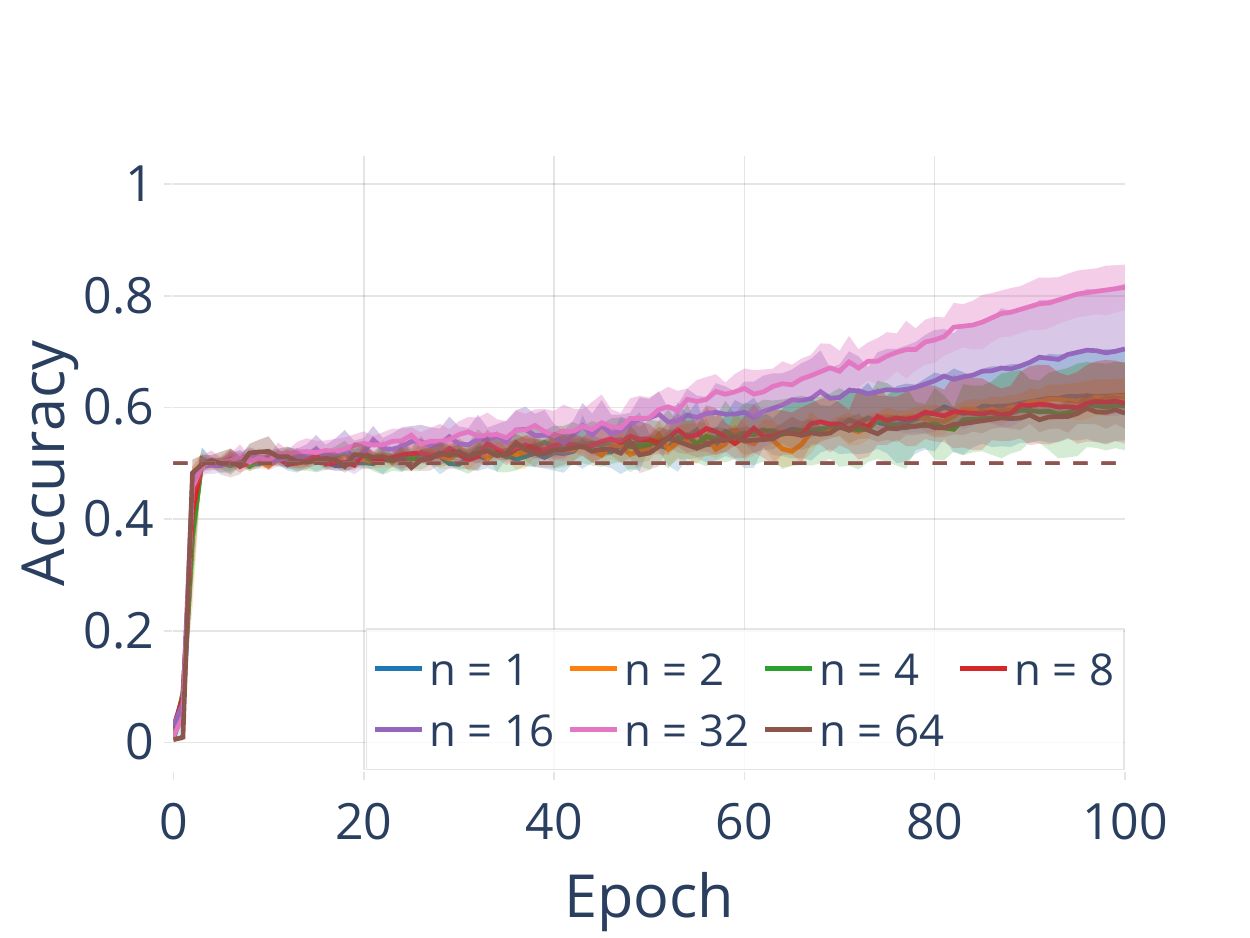}
    }
    \subfloat[Phi-2.7B, Accuracy]{
        \includegraphics[width=\thirdWidth]{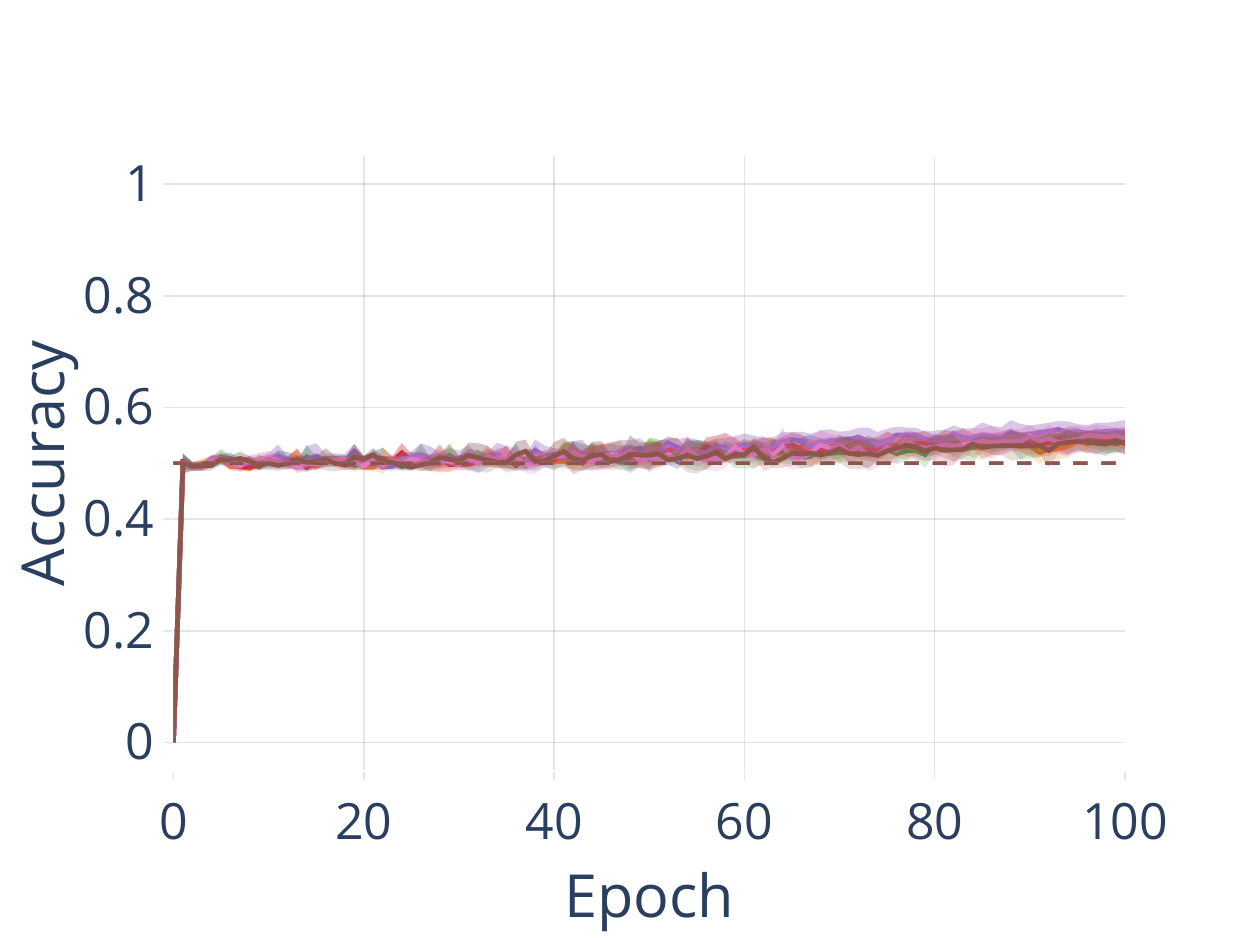}
    }
    \subfloat[Llama2-13B, Accuracy]{
        \includegraphics[width=\thirdWidth]{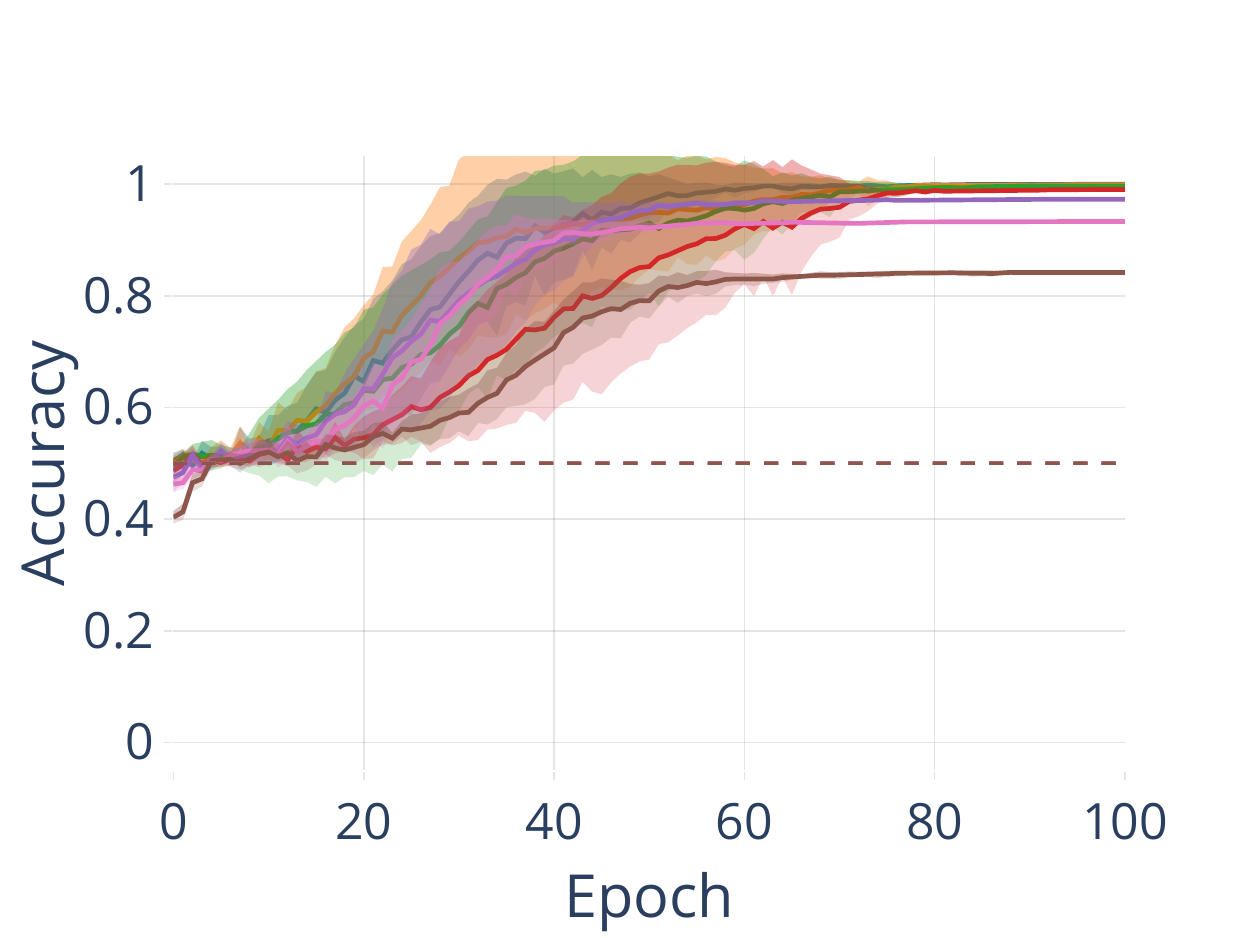}
    }
\caption{\capthead{Accuracy and loss for different partitions of the same string.}{$n = 1024, \ell = 2$}
}
\label{fig:partitions_a2_all}
\end{figure}

\begin{figure}[H]
    \centering
    \subfloat[Pythia-1B, Loss]{
        \includegraphics[width=\thirdWidth]{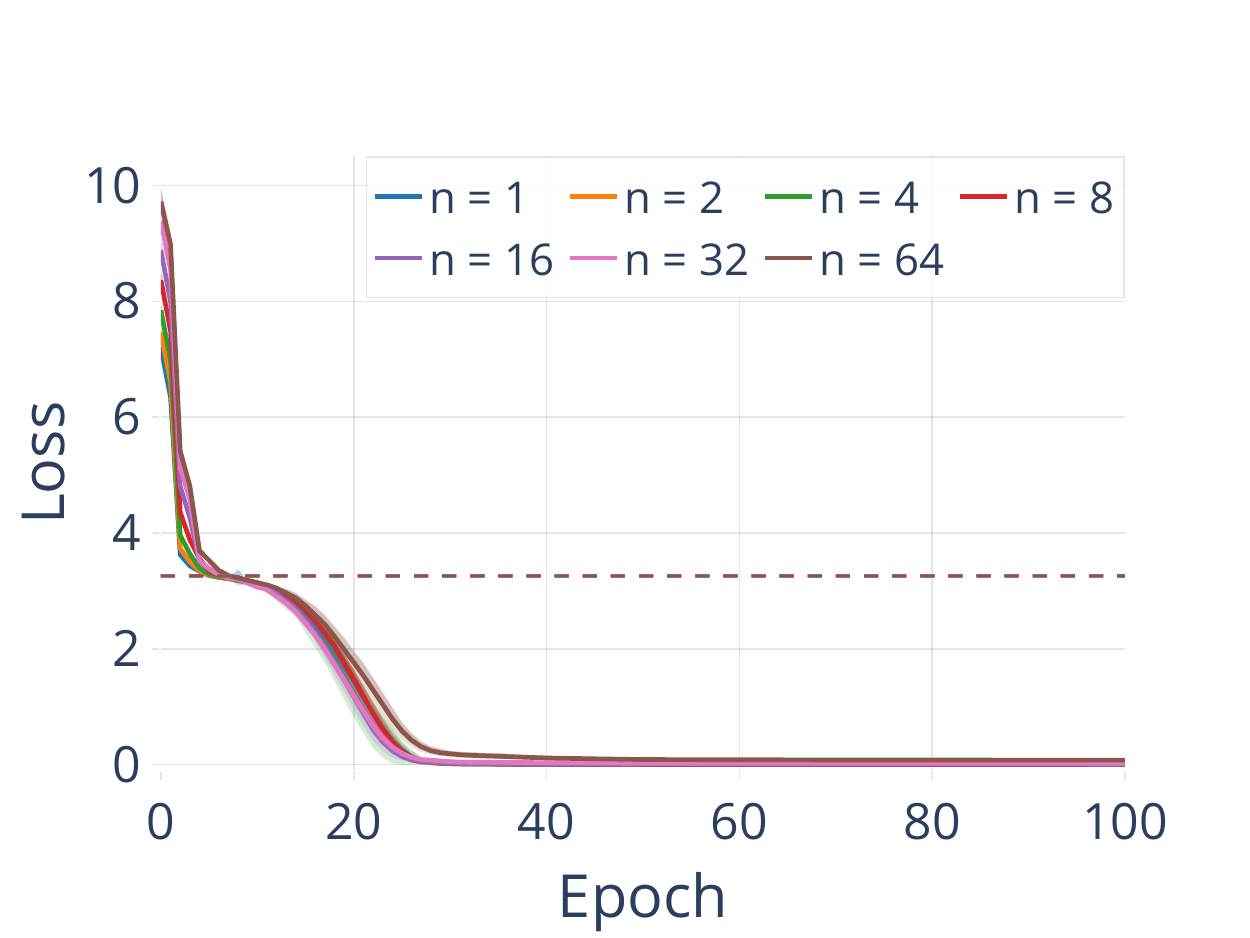}
    }
    \subfloat[Phi-2.7B, Loss]{
        \includegraphics[width=\thirdWidth]{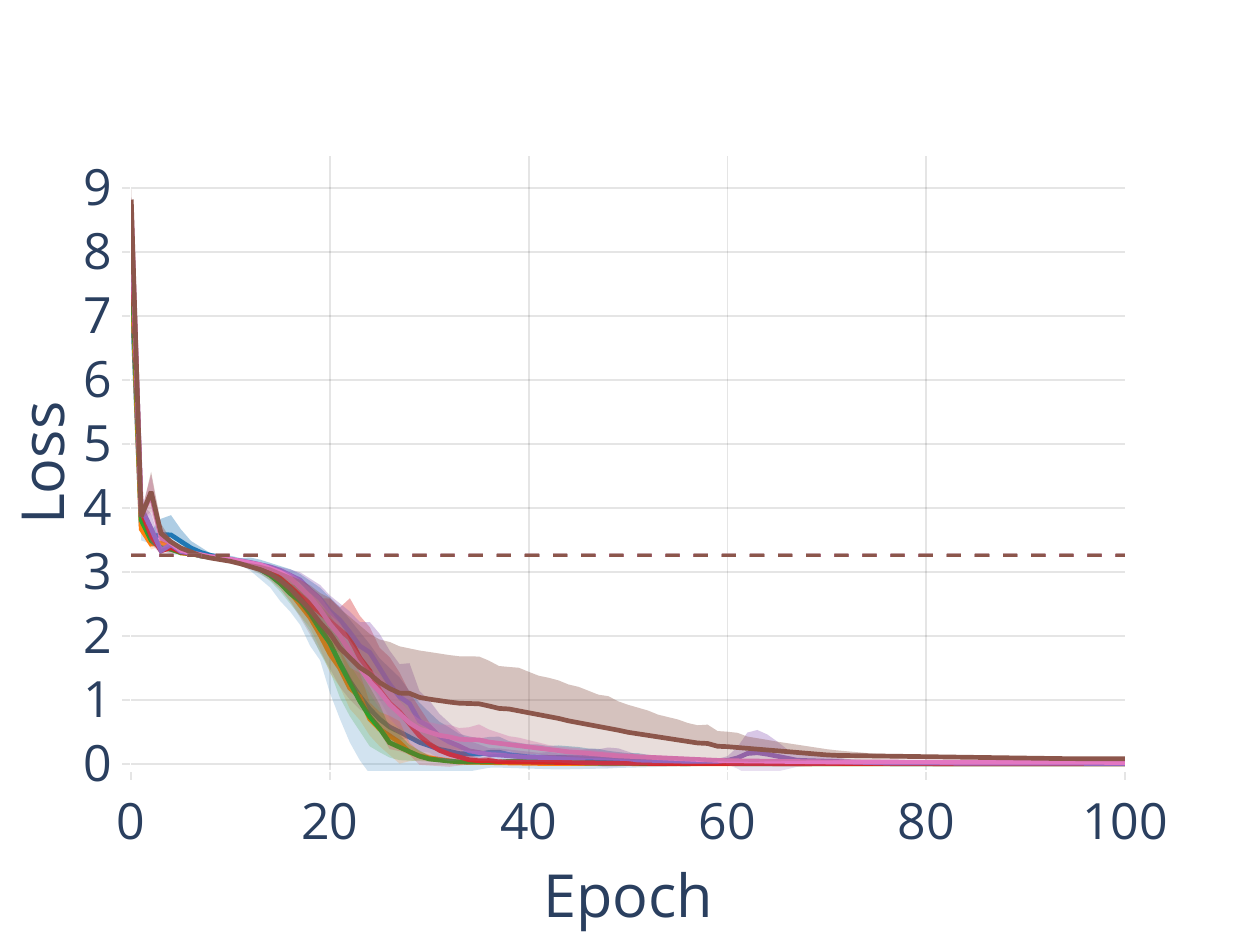}
    }
    \subfloat[Llama2-13B, Loss]{
        \includegraphics[width=\thirdWidth]{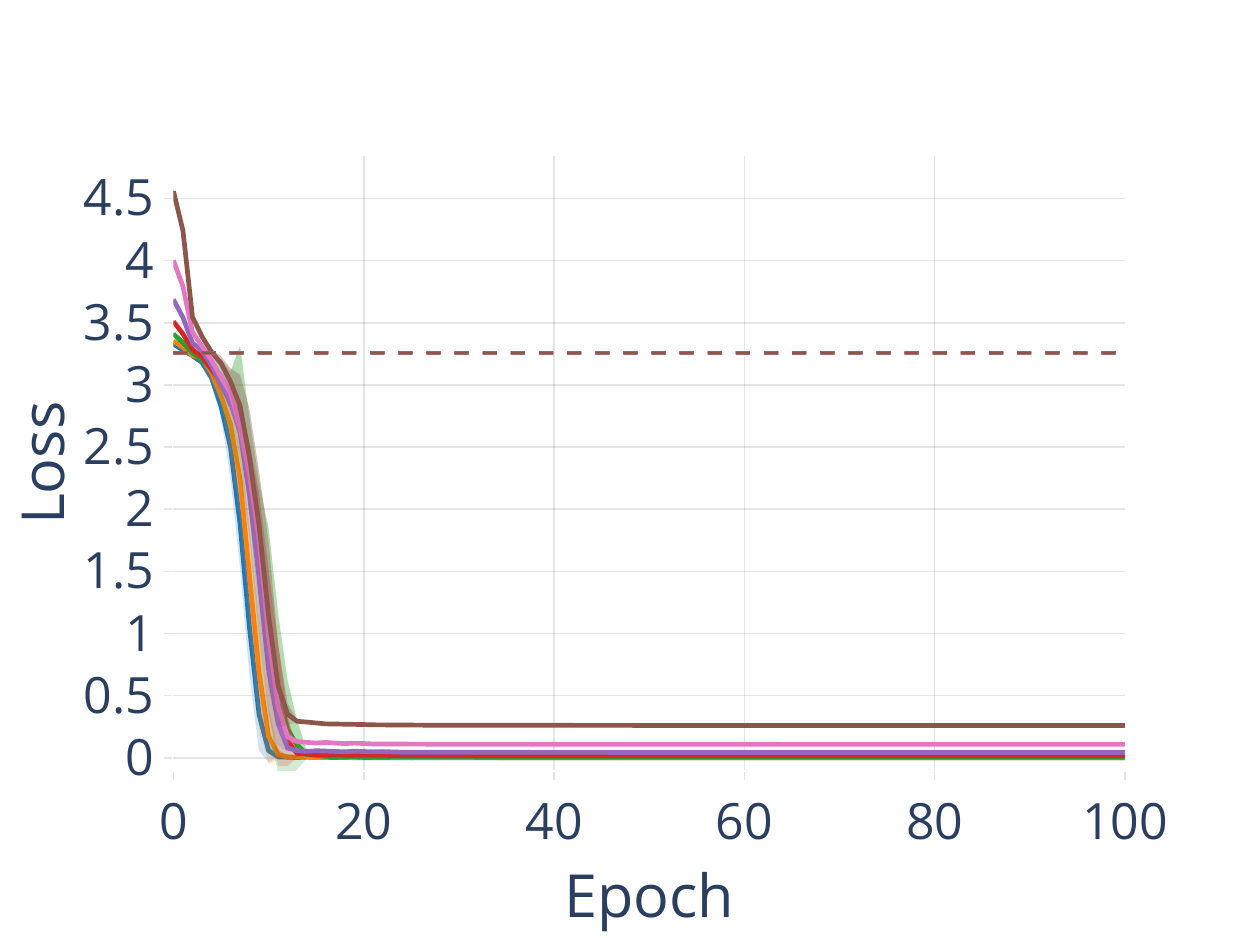}
    }
    \\
    \subfloat[Pythia-1B, Accuracy]{
        \includegraphics[width=\thirdWidth]{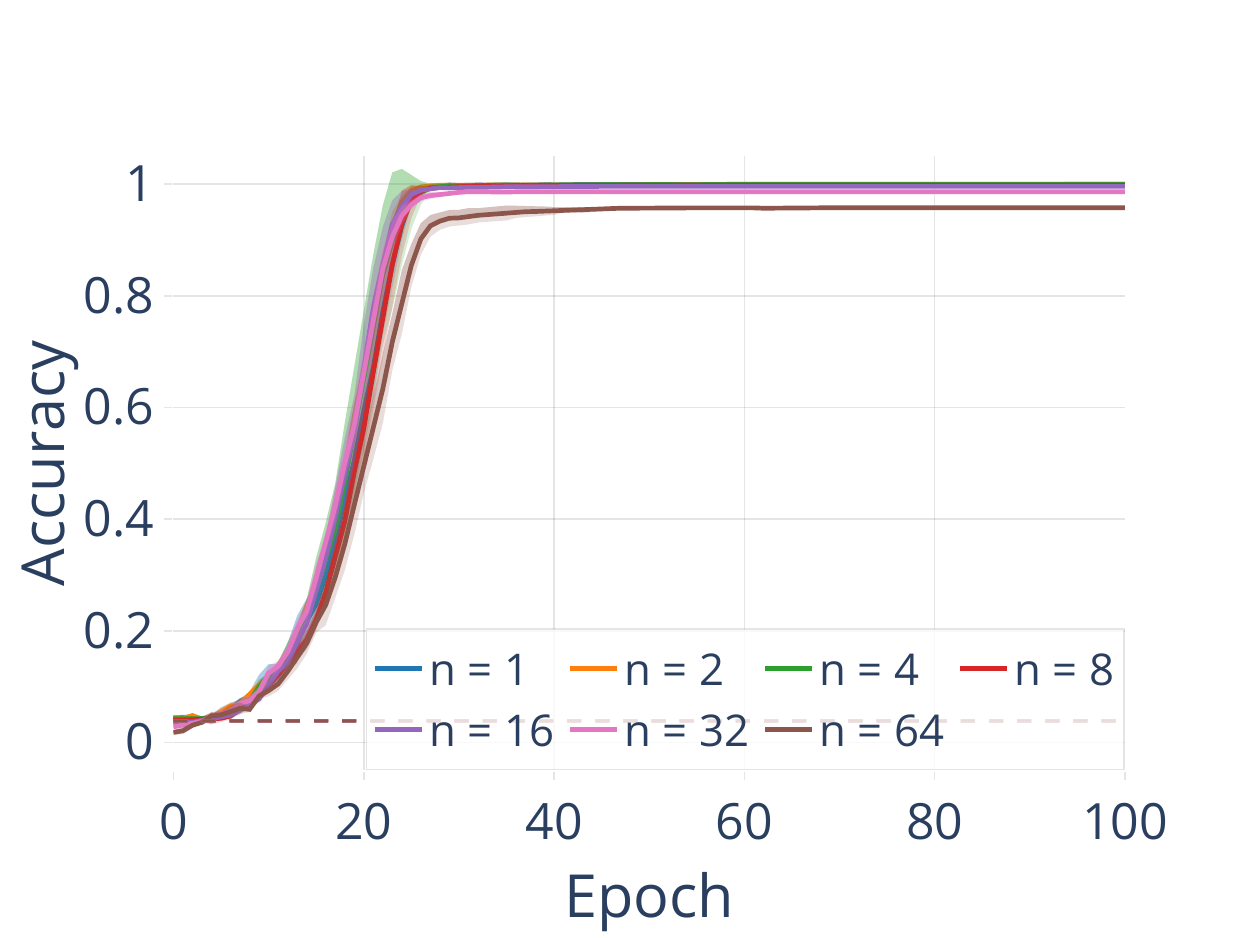}
    }
    \subfloat[Phi-2.7B, Accuracy]{
        \includegraphics[width=\thirdWidth]{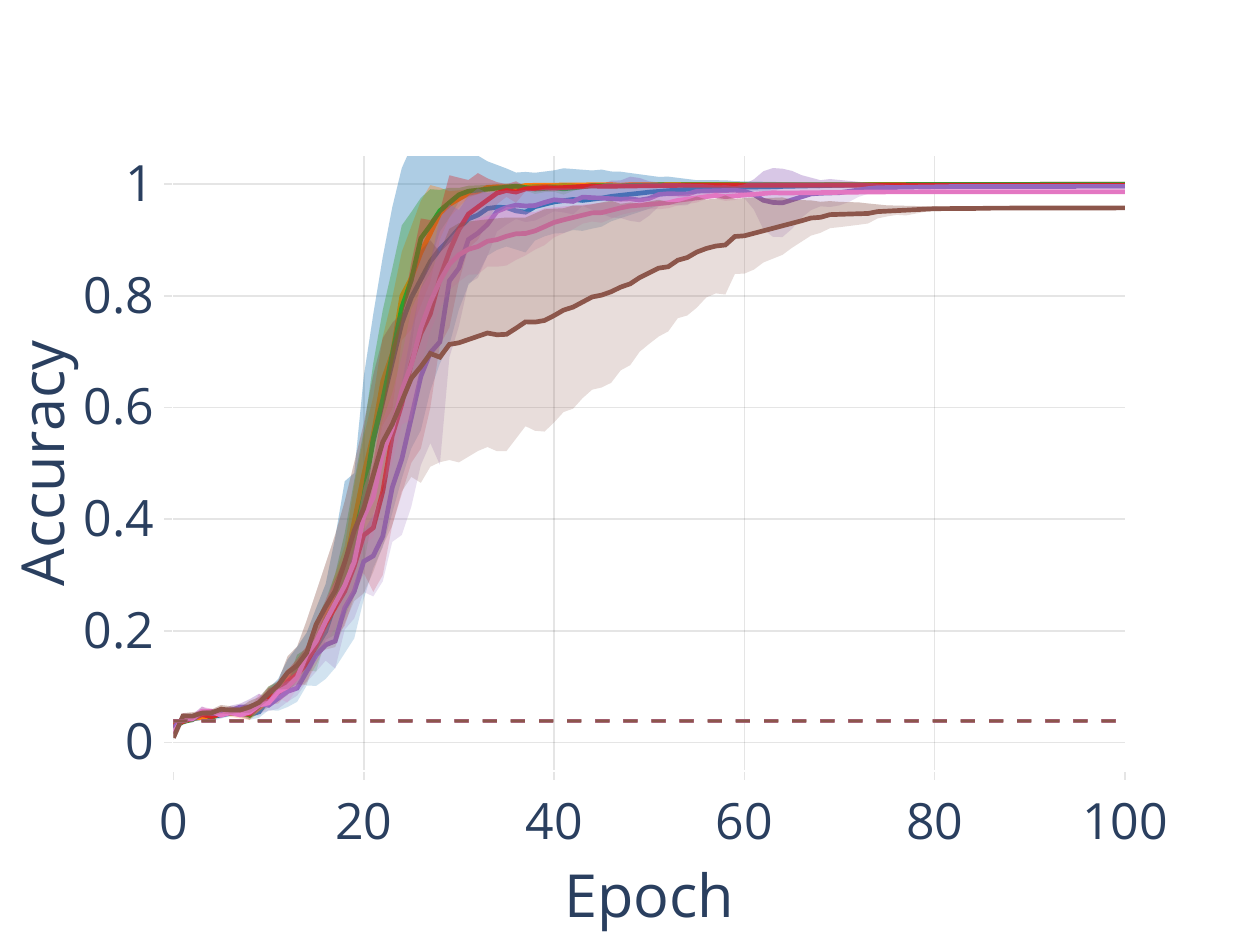}
    }
    \subfloat[Llama2-13B, Accuracy]{
        \includegraphics[width=\thirdWidth]{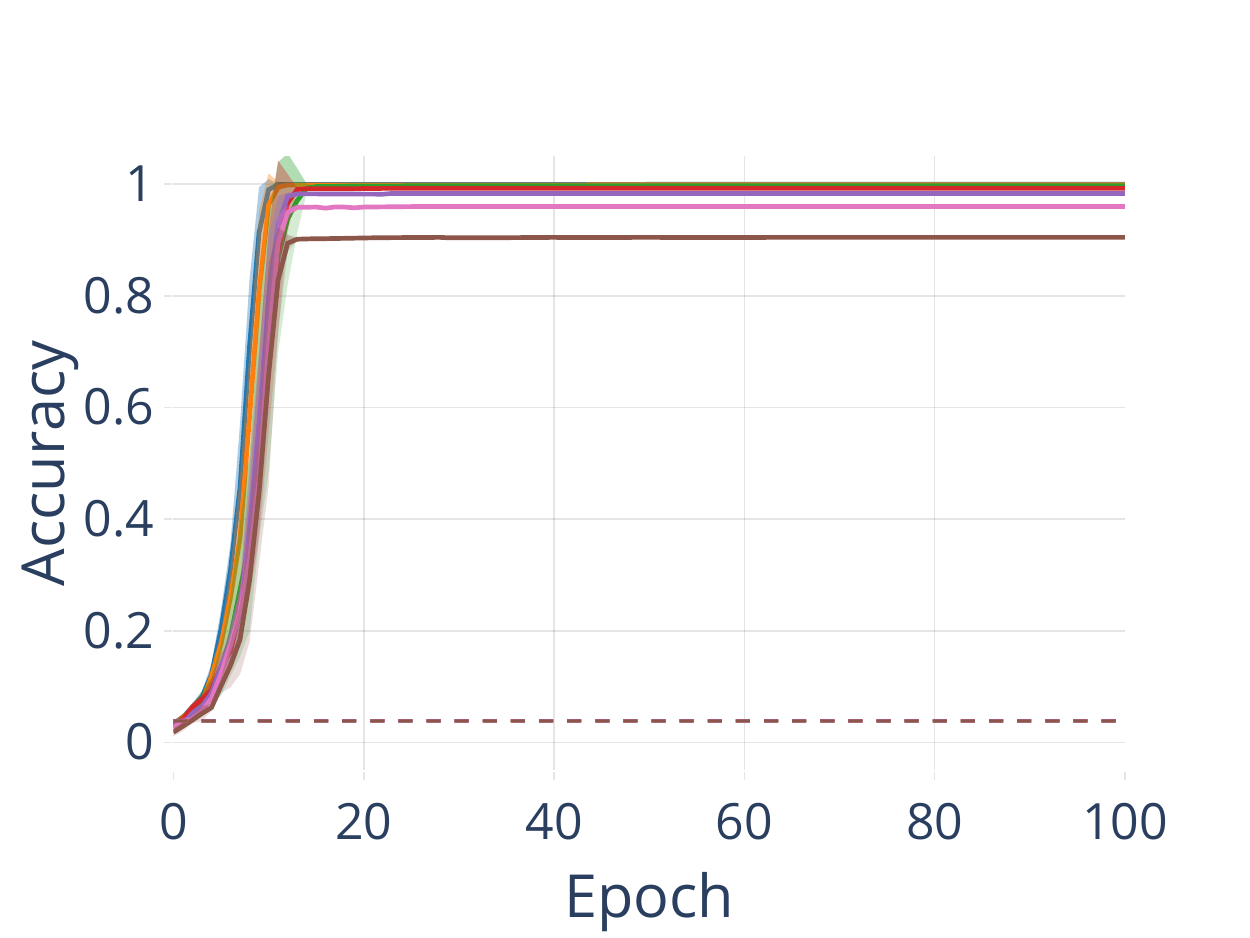}
    }
\caption{\capthead{Accuracy and loss for different partitions of the same string.}{$n = 1024, \ell = 26$}
}
\label{fig:partitions_a26_all}
\end{figure}

In Figures~\ref{fig:partitions_a2_all} and~\ref{fig:partitions_a26_all} we show results for partitioning the same $n = 1024$ token string into $k \in \{1, 2, 4, 8, 16, 32, 64\}$ pieces and memorising them in a batch.
Whether the string is memorised in one long piece or as multiple shorter fragments does barely affect memorisation speed.

\begin{figure}[H]
    \centering
    \subfloat[Pythia-1B, Loss]{
        \includegraphics[width=\thirdWidth]{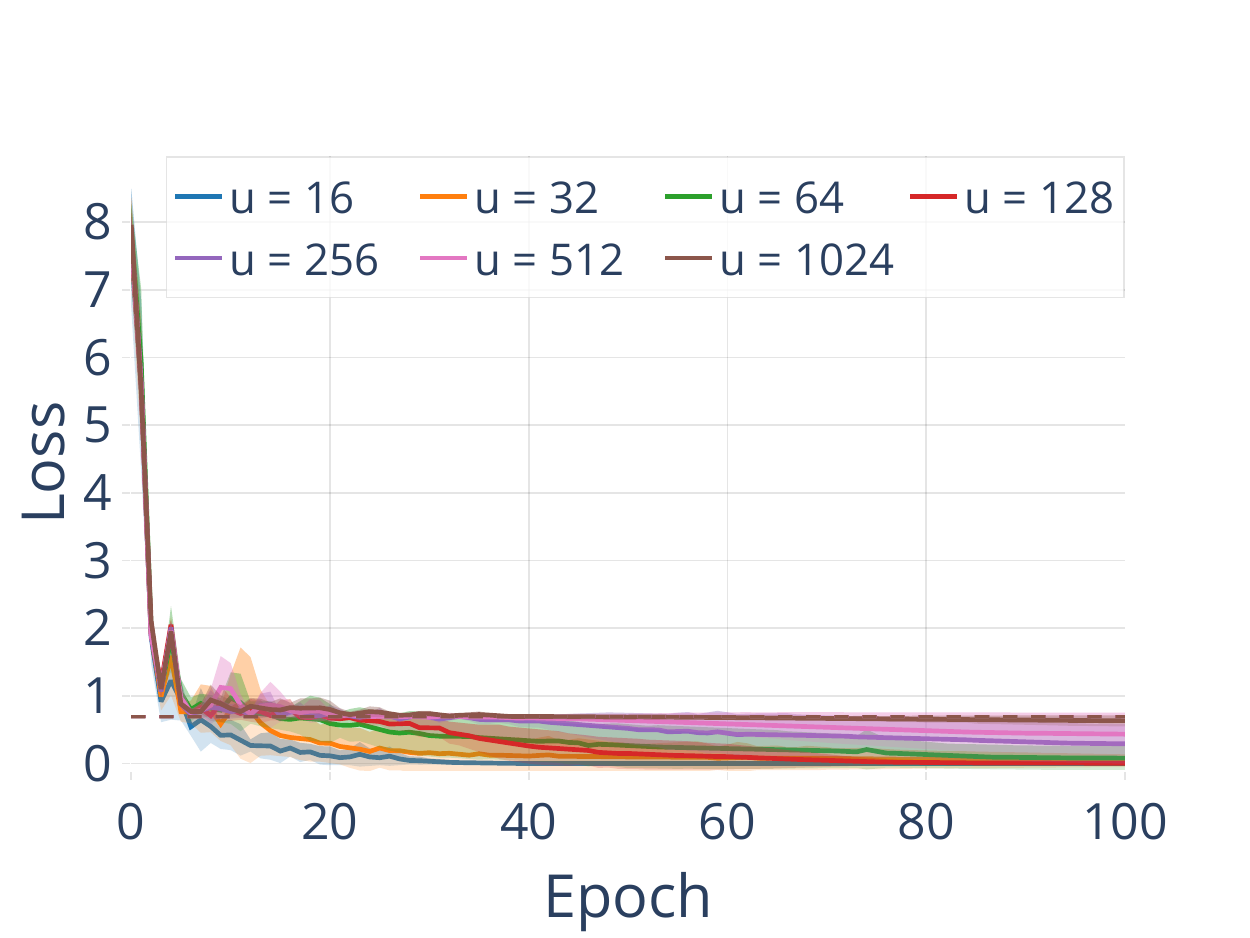}
    }
    \subfloat[Phi-2.7B, Loss]{
        \includegraphics[width=\thirdWidth]{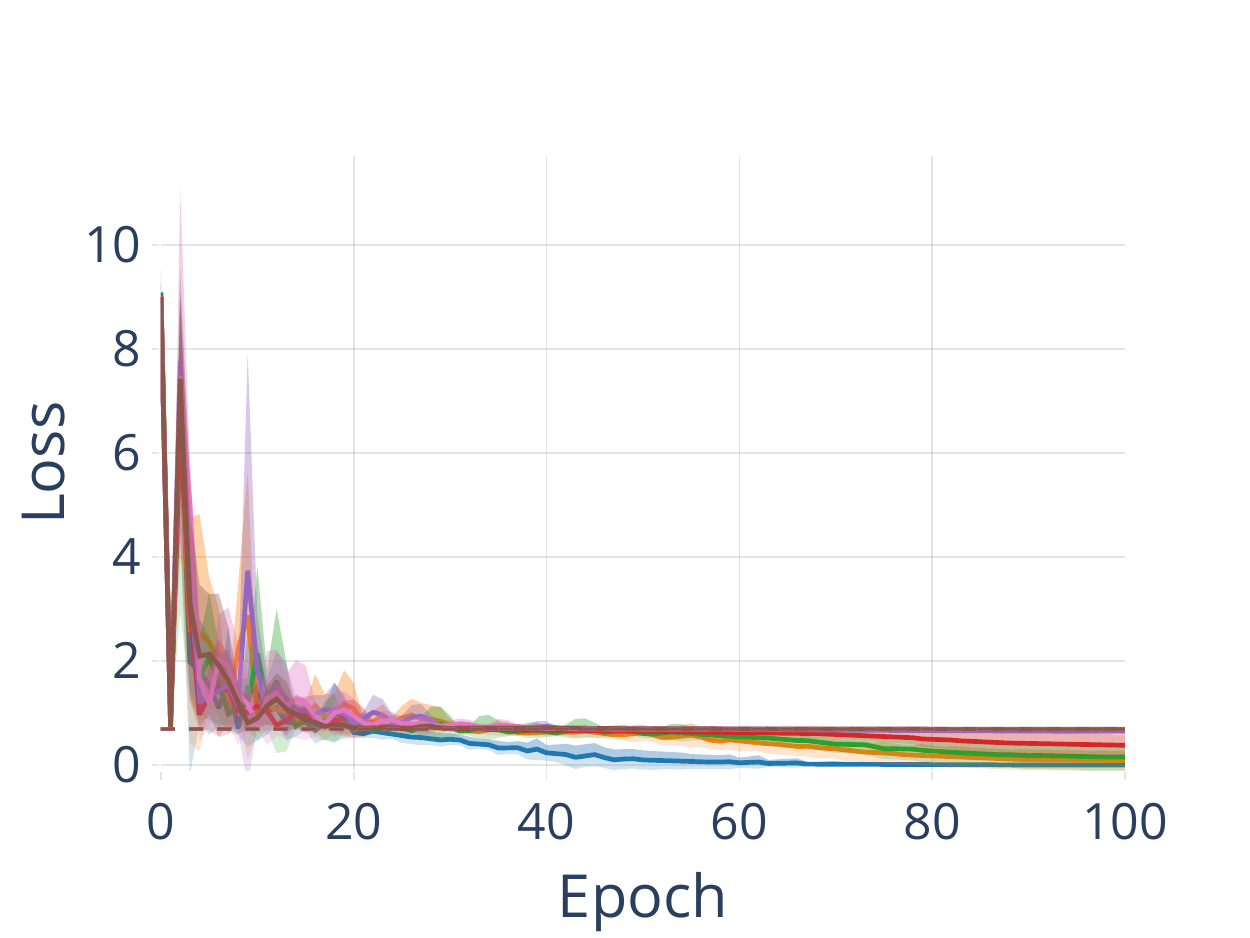}
    }
    \subfloat[Llama2-13B, Loss]{
        \includegraphics[width=\thirdWidth]{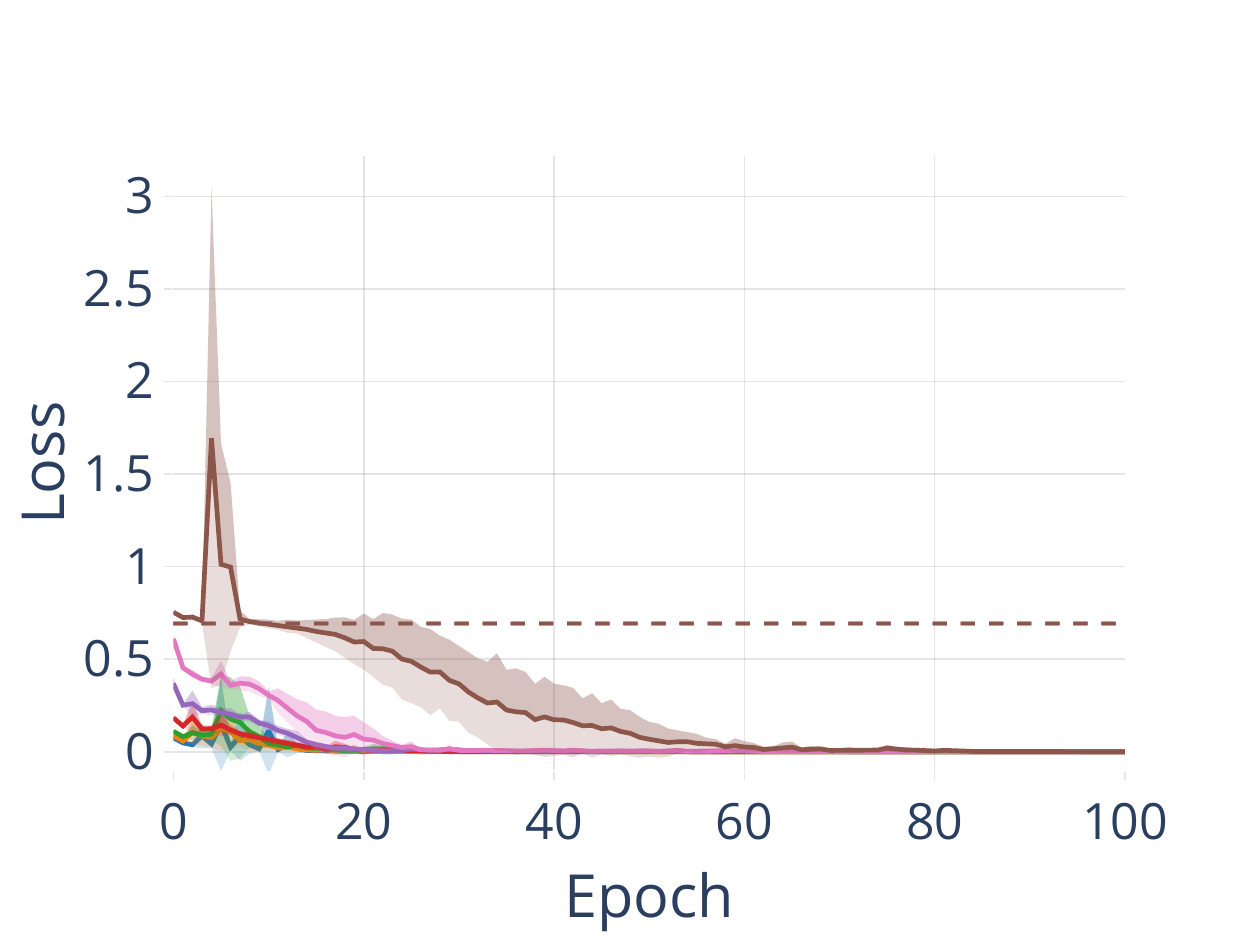}
    }
    \\
    \subfloat[Pythia-1B, Accuracy]{
        \includegraphics[width=\thirdWidth]{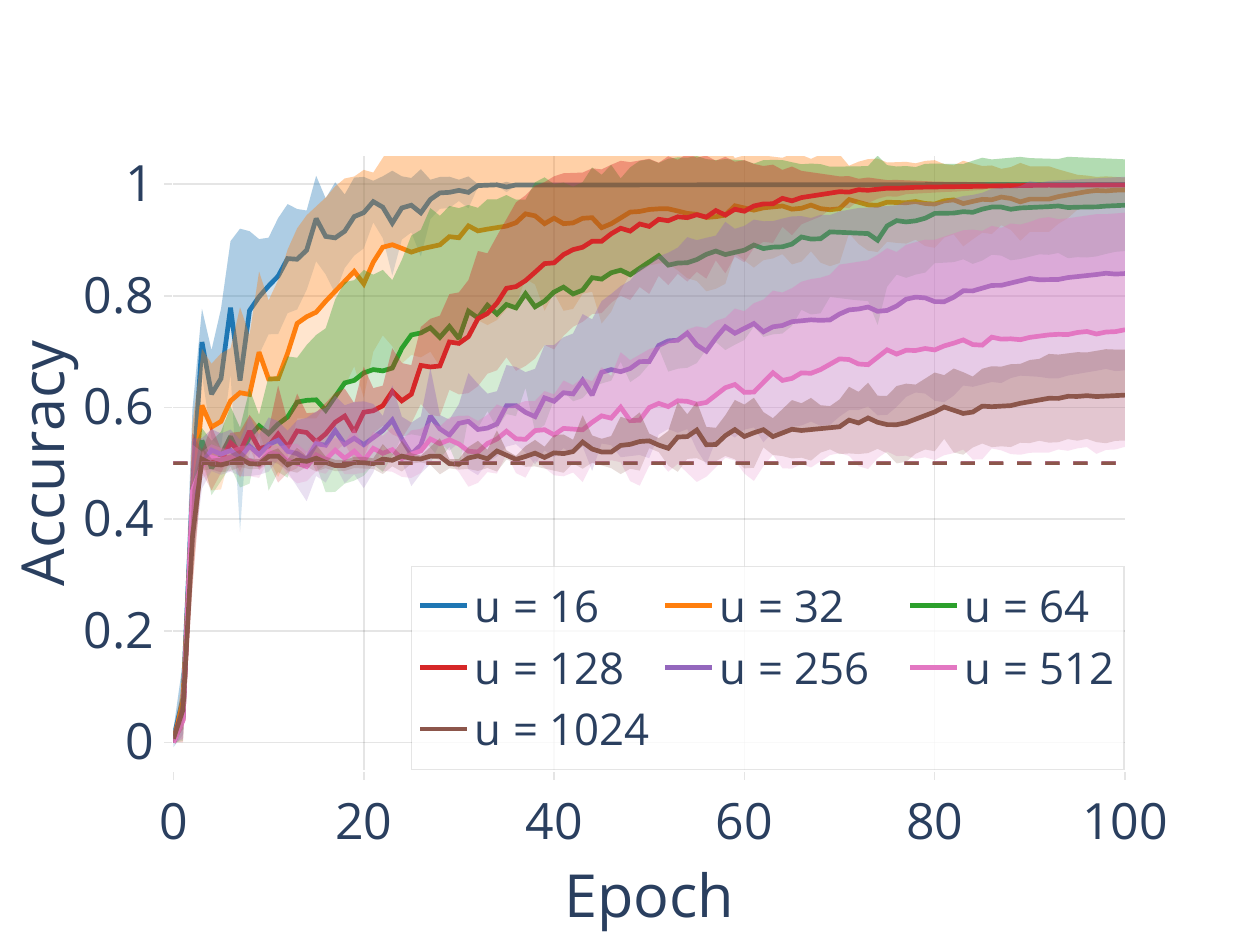}
    }
    \subfloat[Phi-2.7B, Accuracy]{
        \includegraphics[width=\thirdWidth]{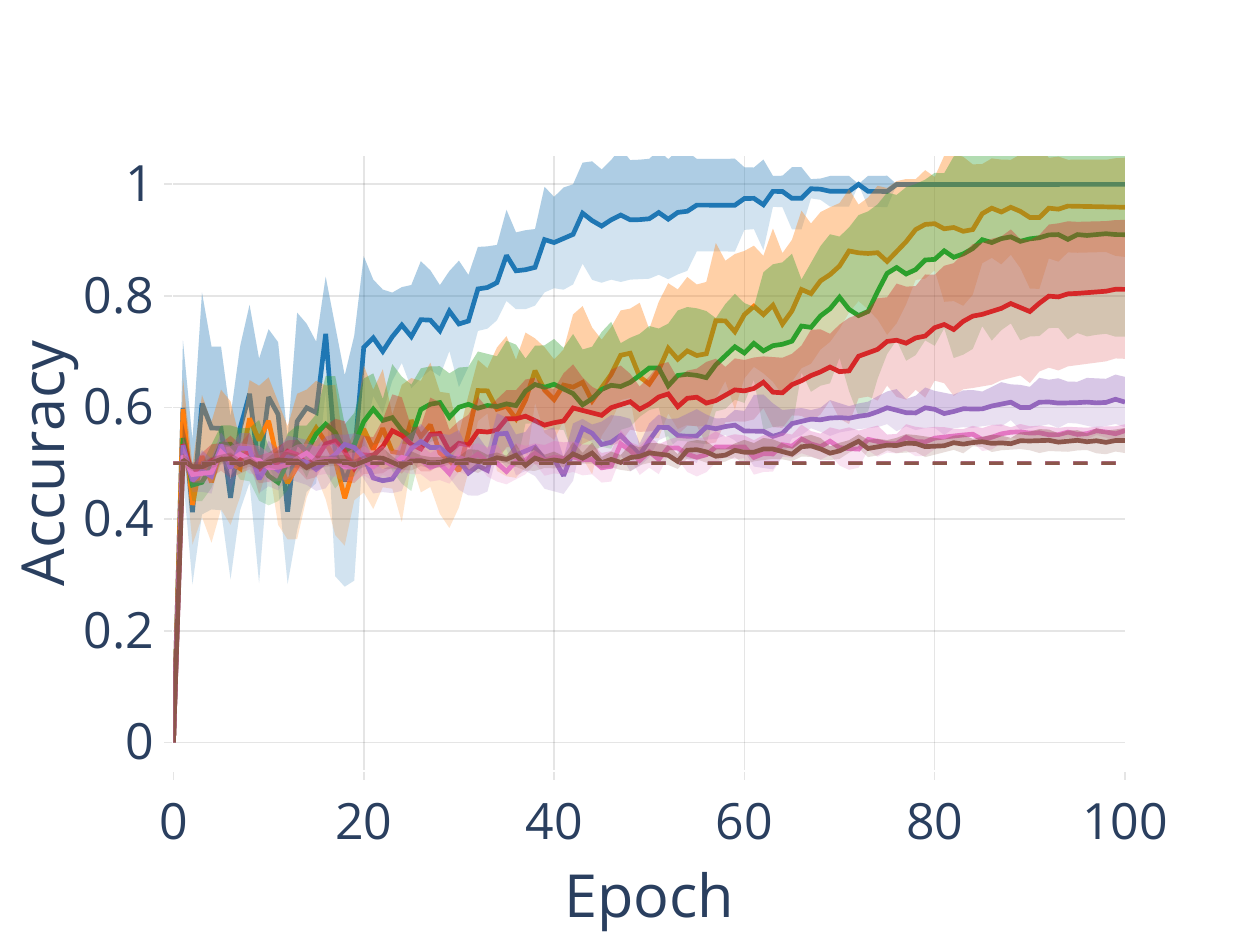}
    }
    \subfloat[Llama2-13B, Accuracy]{
        \includegraphics[width=\thirdWidth]{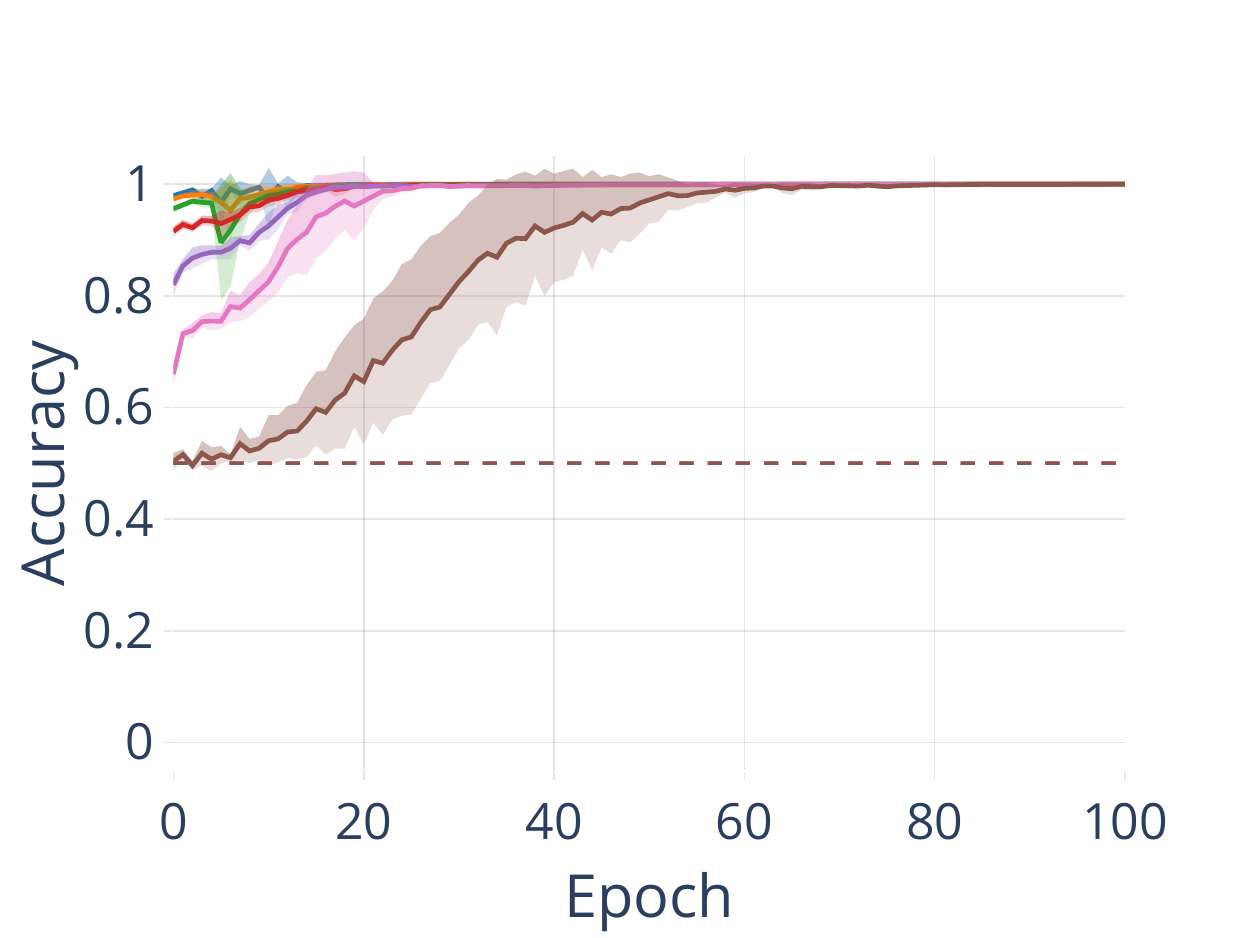}
    }
\caption{\capthead{Accuracy and loss for different sizes of unique substrings $u$.}{$n = 1024, \ell = 2$}
We sample $u$ tokens independently and then repeat the resulting substring $n/u$ times to create the $n = 1024$ token string.
Repetitions of the same random string do not increase memorisation speed.
Additionally, the accuracy at epoch 0 before any training is higher, the smaller $u$, indicating that the models use in-context learning to predict subsequent occurrences of the substrings without having memorised them.
}
\label{fig:unique_substrings_a2_all}
\end{figure}

\begin{figure}[H]
    \centering
    \subfloat[Pythia-1B, Loss]{
        \includegraphics[width=\thirdWidth]{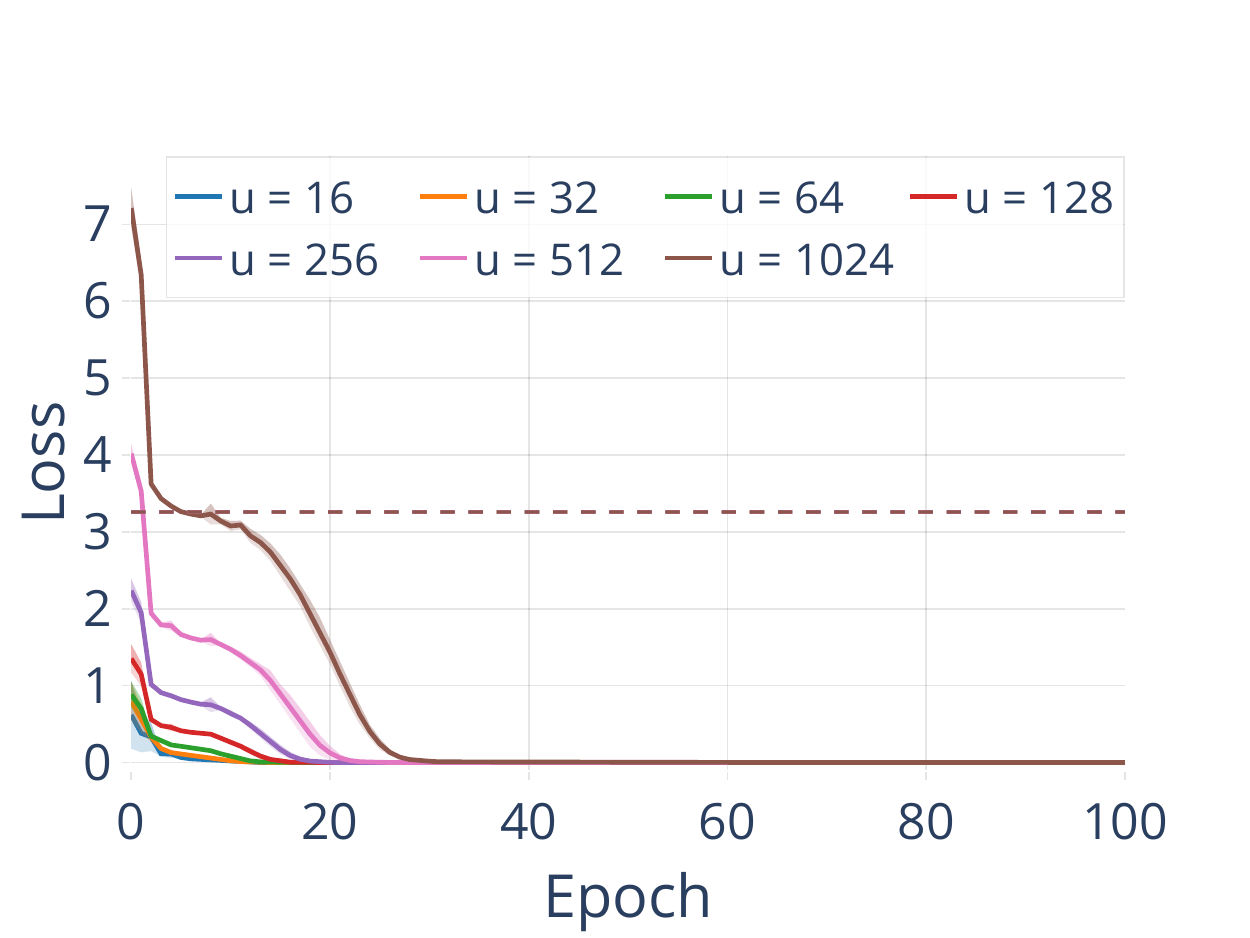}
    }
    \subfloat[Phi-2.7B, Loss]{
        \includegraphics[width=\thirdWidth]{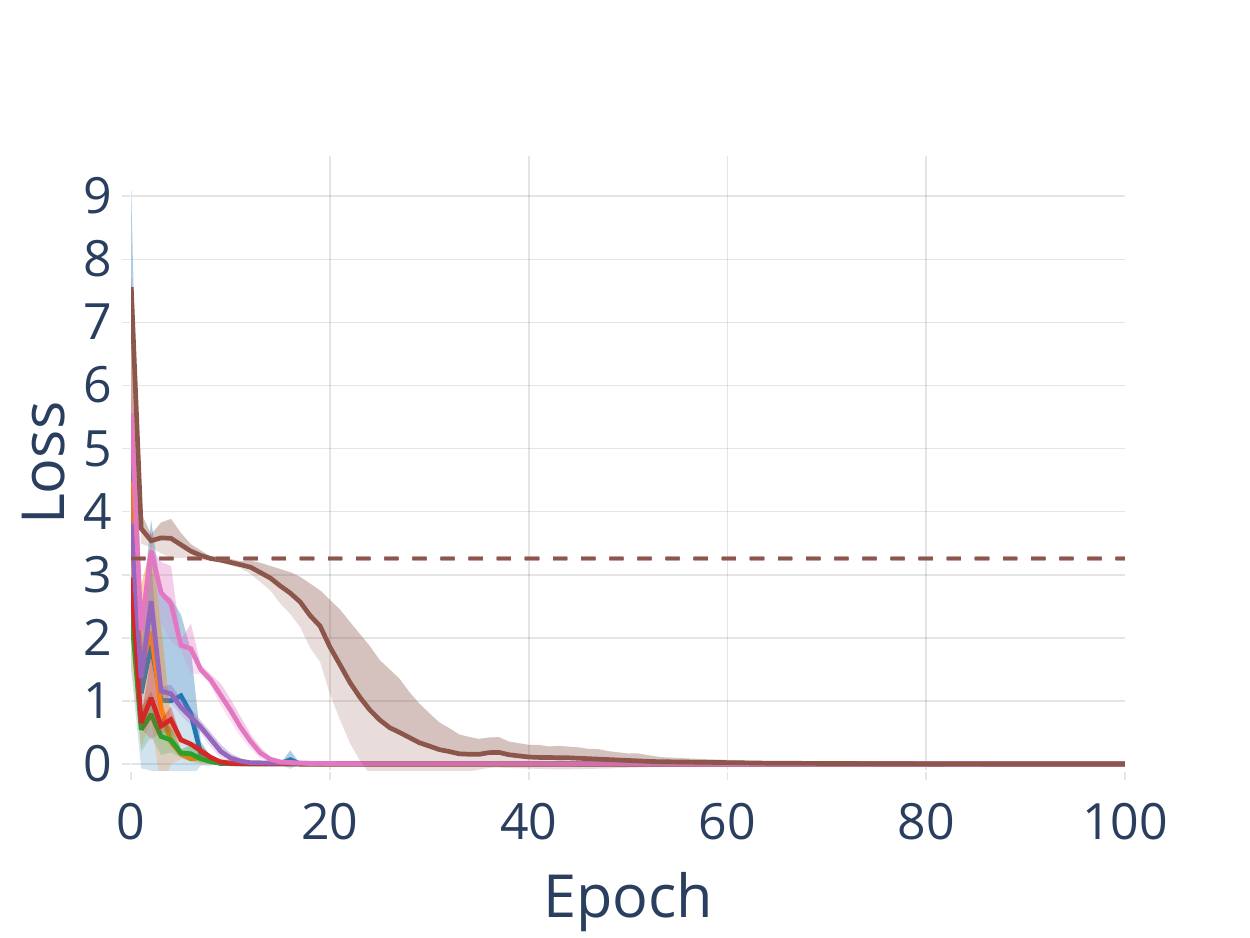}
    }
    \subfloat[Llama2-13B, Loss]{
        \includegraphics[width=\thirdWidth]{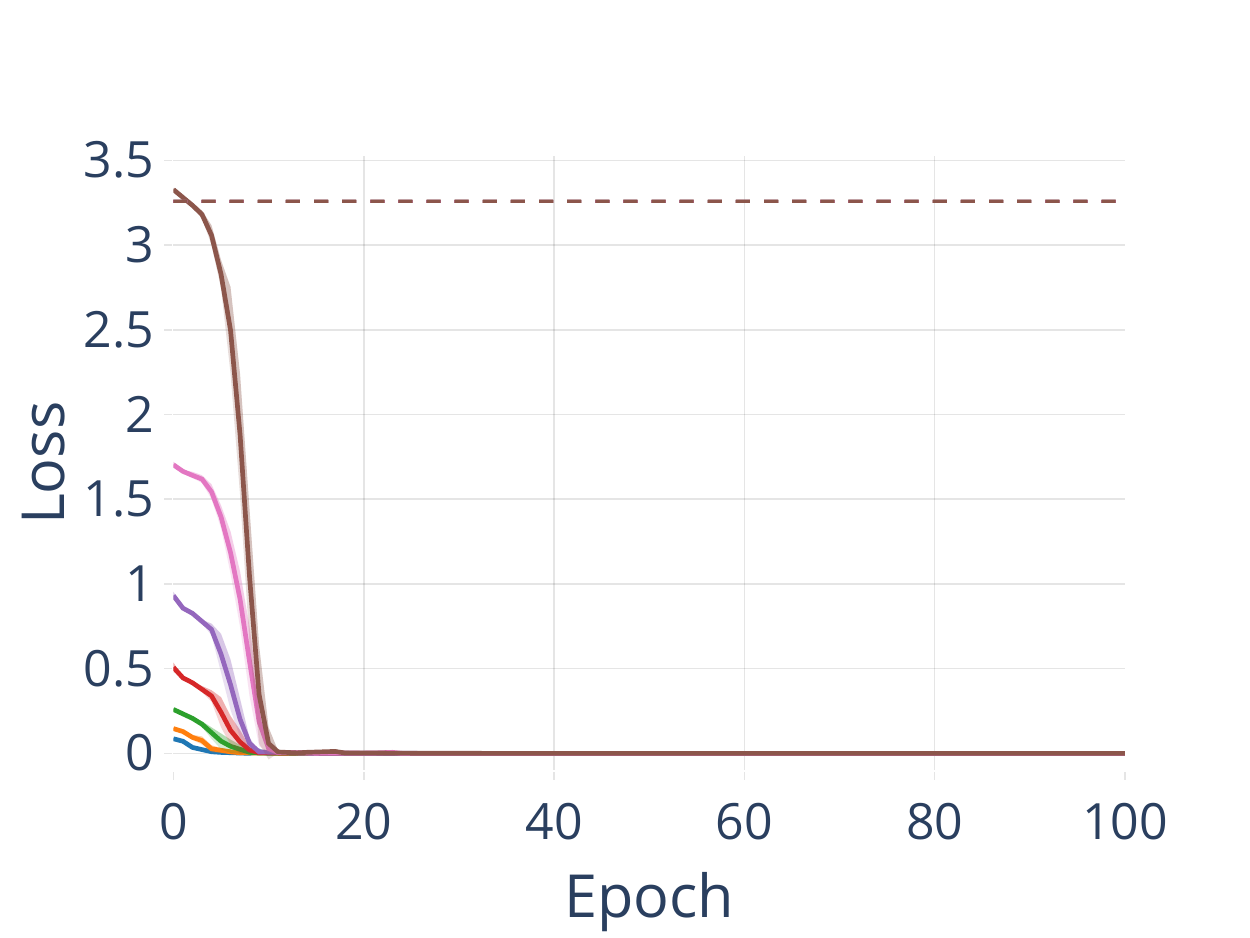}
    }
    \\
    \subfloat[Pythia-1B, Accuracy]{
        \includegraphics[width=\thirdWidth]{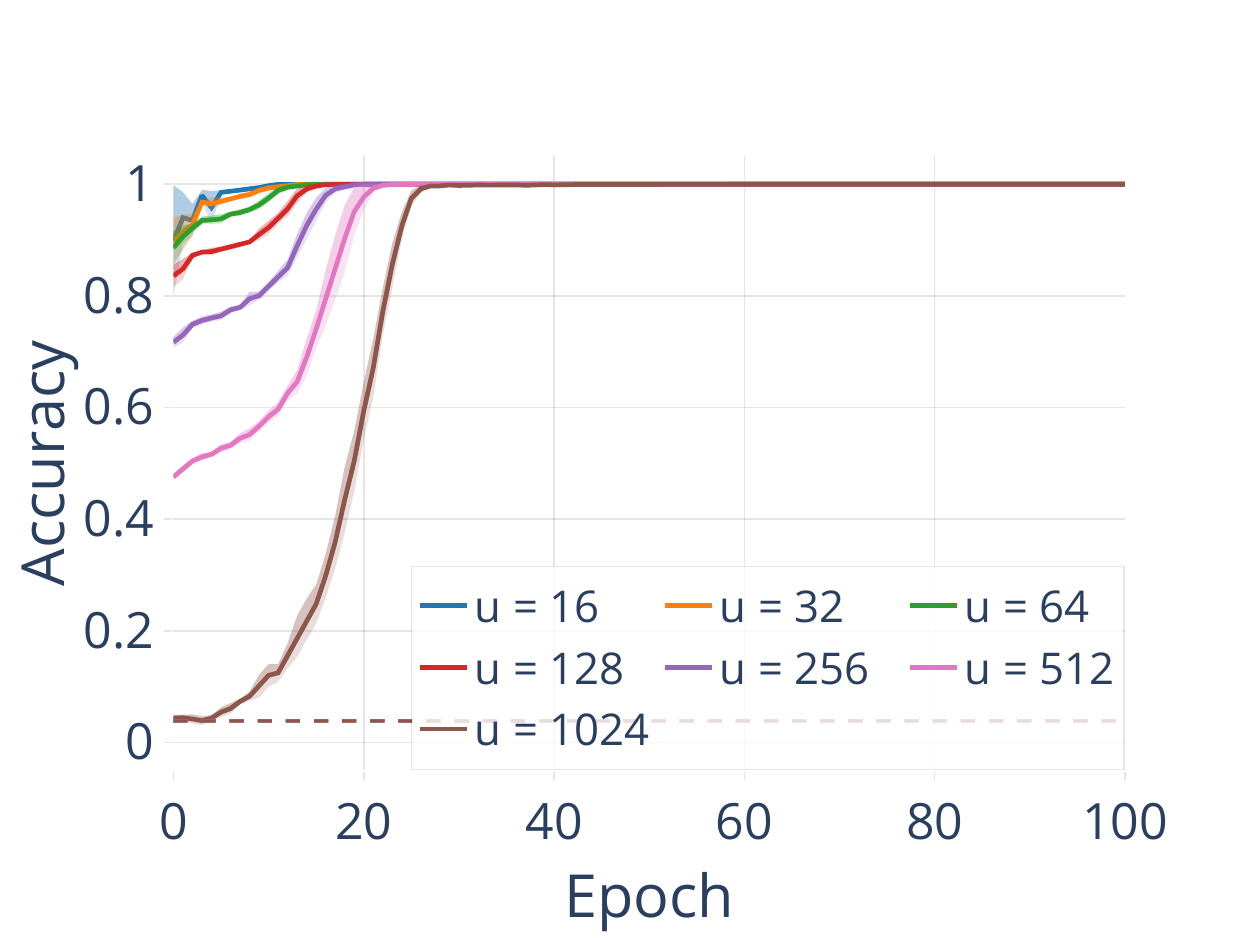}
    }
    \subfloat[Phi-2.7B, Accuracy]{
        \includegraphics[width=\thirdWidth]{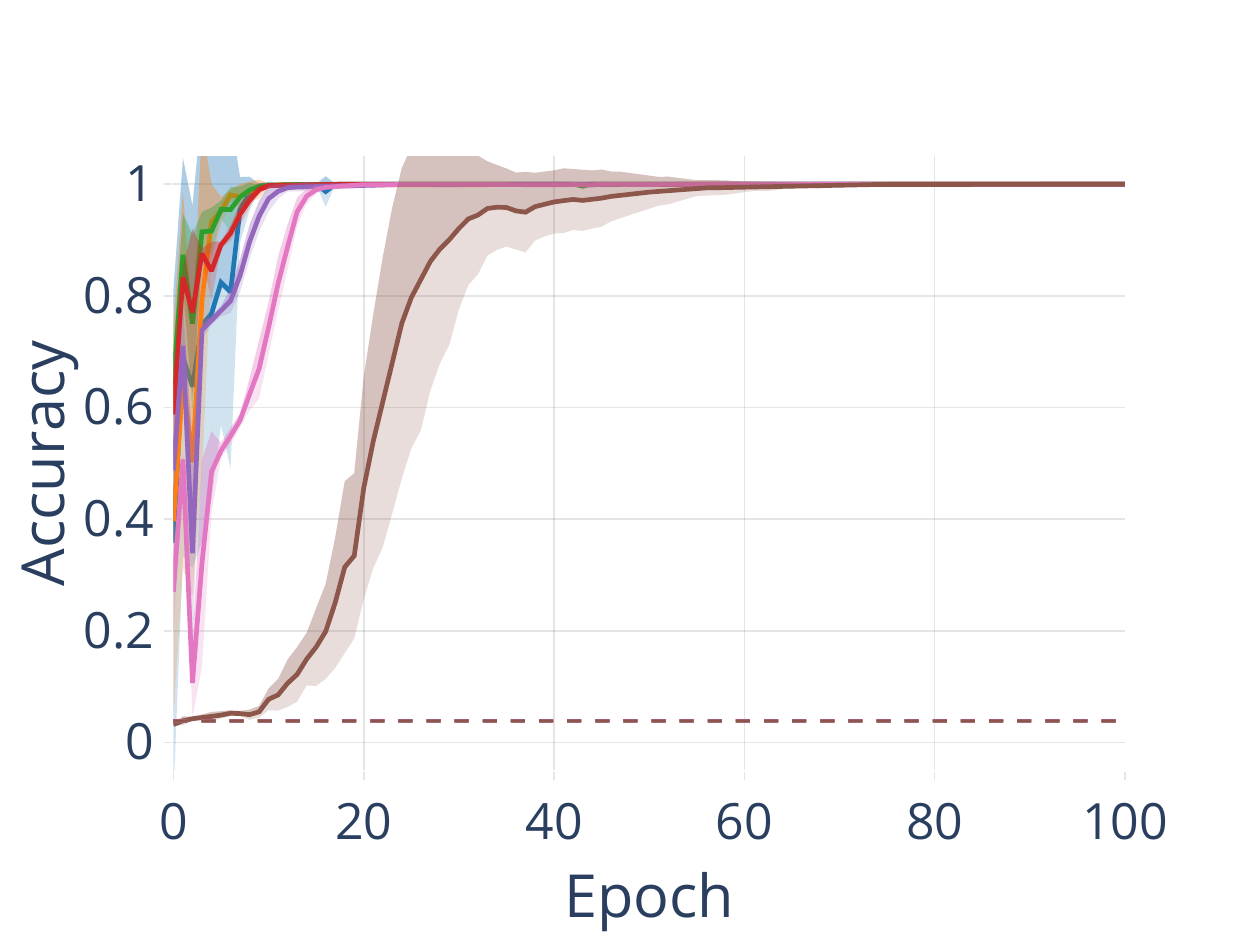}
    }
    \subfloat[Llama2-13B, Accuracy]{
        \includegraphics[width=\thirdWidth]{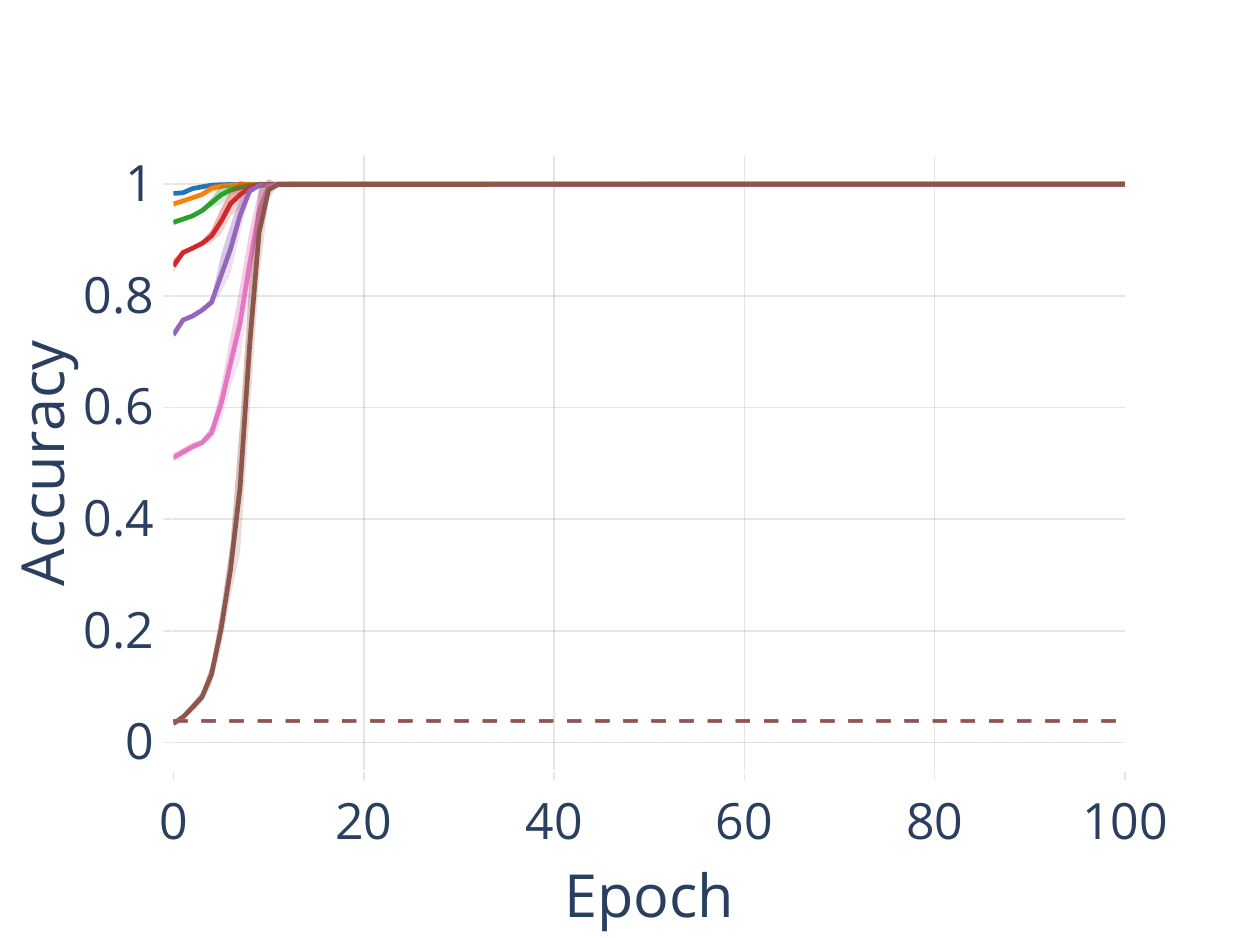}
    }
\caption{\capthead{Accuracy and loss for different sizes of unique substrings $u$.}{$n = 1024, \ell = 26$}
We sample $u$ tokens independently and then repeat the resulting substring $n/u$ times to create the $n = 1024$ token string.
Repetitions of the same random string do not increase memorisation speed.
Additionally, the accuracy at epoch 0 before any training is higher, the smaller $u$, indicating that the models use in-context learning to predict subsequent occurrences of the substrings without having memorised them.
}
\label{fig:unique_substrings_a26_all}
\end{figure}

In Figures~\ref{fig:unique_substrings_a2_all} and~\ref{fig:unique_substrings_a26_all} we show results for sampling a shorter substring of length $u \in \{16, 32, 64, 128, 256, 512, 1024\}$ and then repeating it $n / u$ times to create the full $n = 1024$ token string.
The results show that the memorisation speed strongly depends on $u$ and not on $n$, indicating that what matters is the fraction of unique tokens resp. independently sampled tokens in the string.
Repetitions of the same random string do not increase memorisation speed.
Furthermore, the accuracy plots show that the initial accuracy of the models at epoch 0, before they are trained, are at values $1 - (u / n)$, which means that for a small $u$, resp. many repetitions of the unique substring, the model can predict most of the string correctly.
This can only happen because the models uses in-context learning to predict tokens in subsequent occurrences of the substrings.

\subsection{Results on conditional probability strings}
\label{app:conditional_probability_strings}

We know that the entropy of a string affects how hard it is for models to memorize it.
To obtain a better understanding about how entropy interacts with the memorability of a string, we test how conditional entropy affects memorability.
In order to do so, we construct strings with the same unconditional entropy, but different levels of conditional entropy.

By the n-conditional entropy $H_n(s)$ of a string $s$, we refer to the entropy $H_n(s) = H(s_i | s_{i-n}, \dots, s_{i-1})$, i.e. the entropy over tokens in $s$, that the preceding n tokens, i.e. the preceding n-gram is known.
We are interested in knowing whether at the same level of unconditional entropy $H(s)$, i.e. 0-conditional entropy, strings with different levels of n-conditional entropy $H_n(s)$ differ in their memorability.

\textbf{Data construction:}

Privileged continuation tokens:
We create string $s$ with alphabet $A$ with a certain level of n-conditional entropy by assigning each possible n-gram $g$ over $A$ a certain \emph{privileged continuation token} $t_g$.
E.g. for $A = \{a, b\}$, there are the 2-grams $aa, ab, ba, bb$, and each of them would have a privileged continuation, e.g. $b$ for $aa$, $a$ for $ab$, etc.

Constructing strings with different levels of conditional entropy:
To sample string $s$, we first sample $n$ tokens from $A$ uniformly at random.
To sample the next token $s_i$, we get its preceding n-gram $g = s_{i-n}, \dots, s_{i-1}$, look up its privileged token $t_g$ and then sample a token from $A$ with $k \times$ \emph{relative probability} $p_k = k * p_u$ for $t_g$, and uniform probability $p_u$ for all other tokens $t \in A \setminus \{t_g\}$.
I.e. we are $k$ times more likely to sample the privileged token $t_g$ as a continuation to $g$ than the other tokens in $A$.
We obtain $p_k$ as $p_k = \frac{k}{|A| -1 + k}$ and $p_u = \frac{1 - p_k}{|A| - 1}$.
Increasing the relative probability $p_k$ lowers the conditional entropy $H_n(s)$ of string $s$.

Ensuring the same level of unconditional entropy:
To ensure that strings with different $p_k$ have the same unconditional entropy $H(s)$, we ensure that each token $t \in A$ appears the same number of times as a privileged continuation token.
I.e. for 1-grams, where there are $|A|$ combinations (single tokens from $A$), each $t \in A$ appears once as the privileged token of a 1-gram.
For 2-grams, with $|A|^2$ possible combinations, each token appears $A$ times as privileged token, etc.
E.g. for 2-grams over $A = \{a, b\}$ a privileged token mapping $aa \rightarrow b, ab \rightarrow b, ba \rightarrow a, bb \rightarrow a$ would be valid, whereas the mapping $aa \rightarrow b, ab \rightarrow b, ba \rightarrow b, bb \rightarrow b$ would be not.
Making each token appear the same number of times as privileged continuation ensures that the overall probability of each $t \in A$ is the same, and thus the unconditional entropy of the strings is the same.

As usual, we train models for 100 epochs to memorize strings with alphabets of different sizes (i.e. entropy levels) and record their memorization dynamics.
We also compute the empirical unconditional and conditional entropy of the sampled strings.

\begin{figure}[H]
    \centering
    \subfloat[Pythia-1B, $\ell = 2$]{
        \includegraphics[width=\thirdWidth]{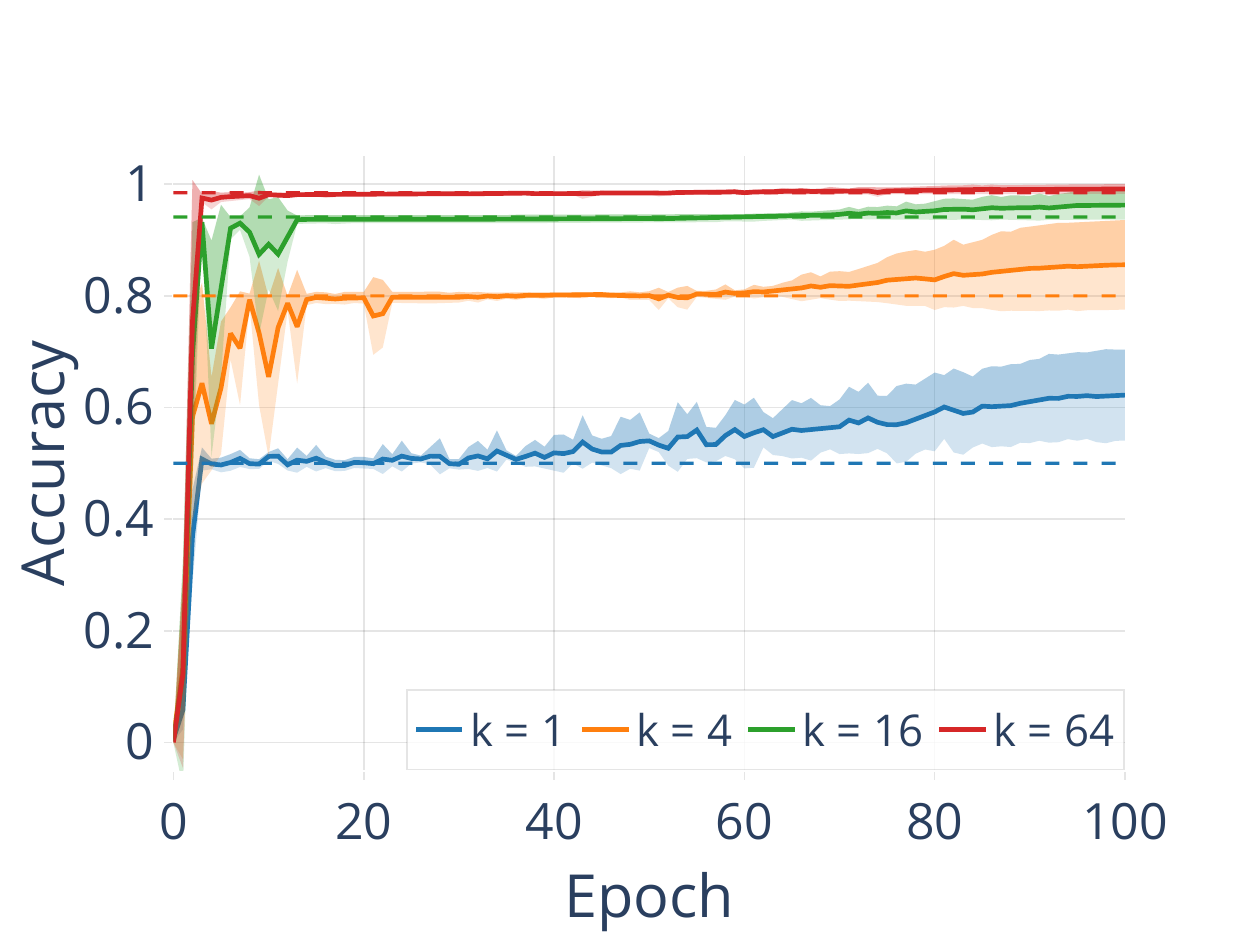}
    }
    \subfloat[Phi-2.7B, $\ell = 2$]{
        \includegraphics[width=\thirdWidth]{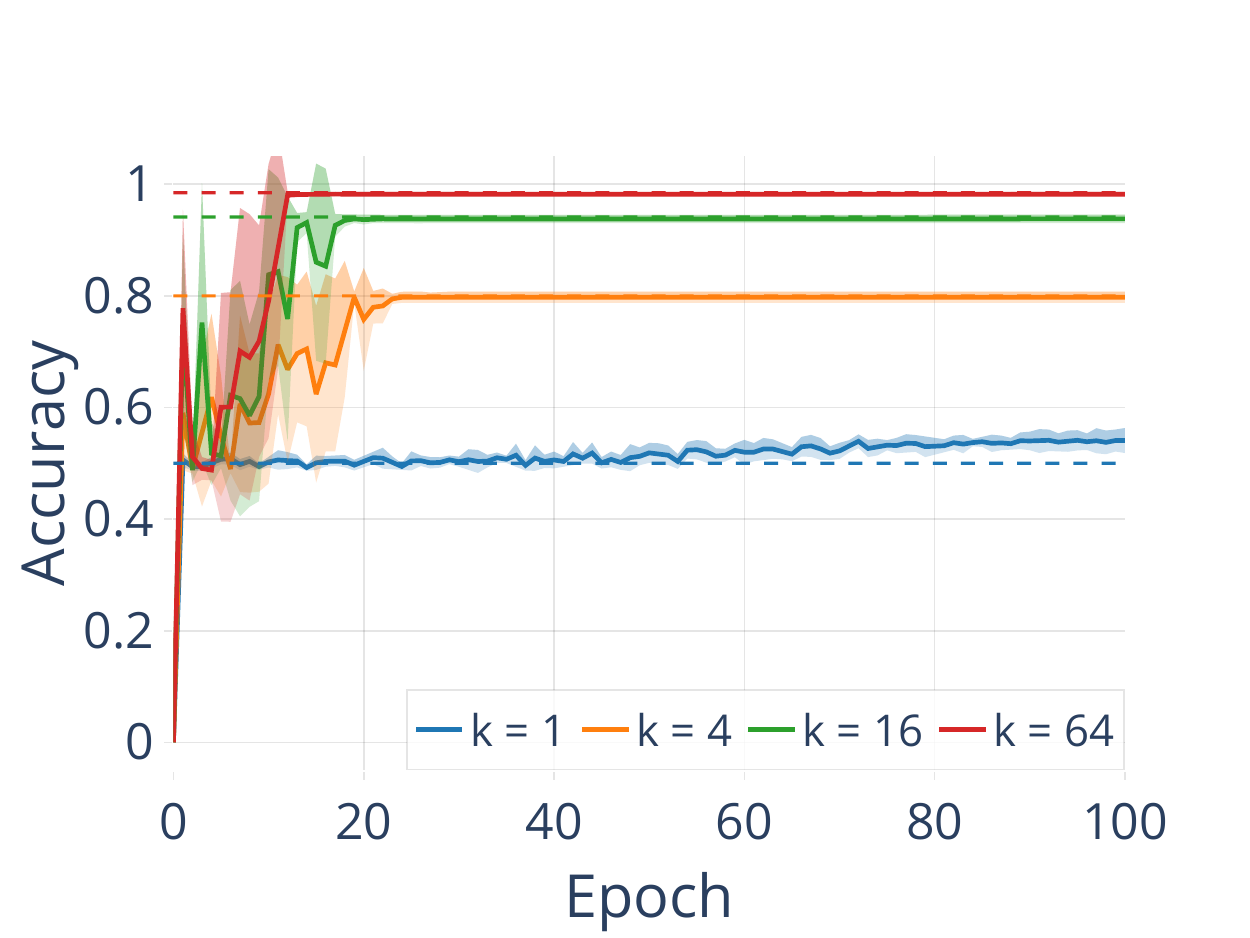}
    }
    \subfloat[Llama2-13B, $\ell = 2$]{
        \includegraphics[width=\thirdWidth]{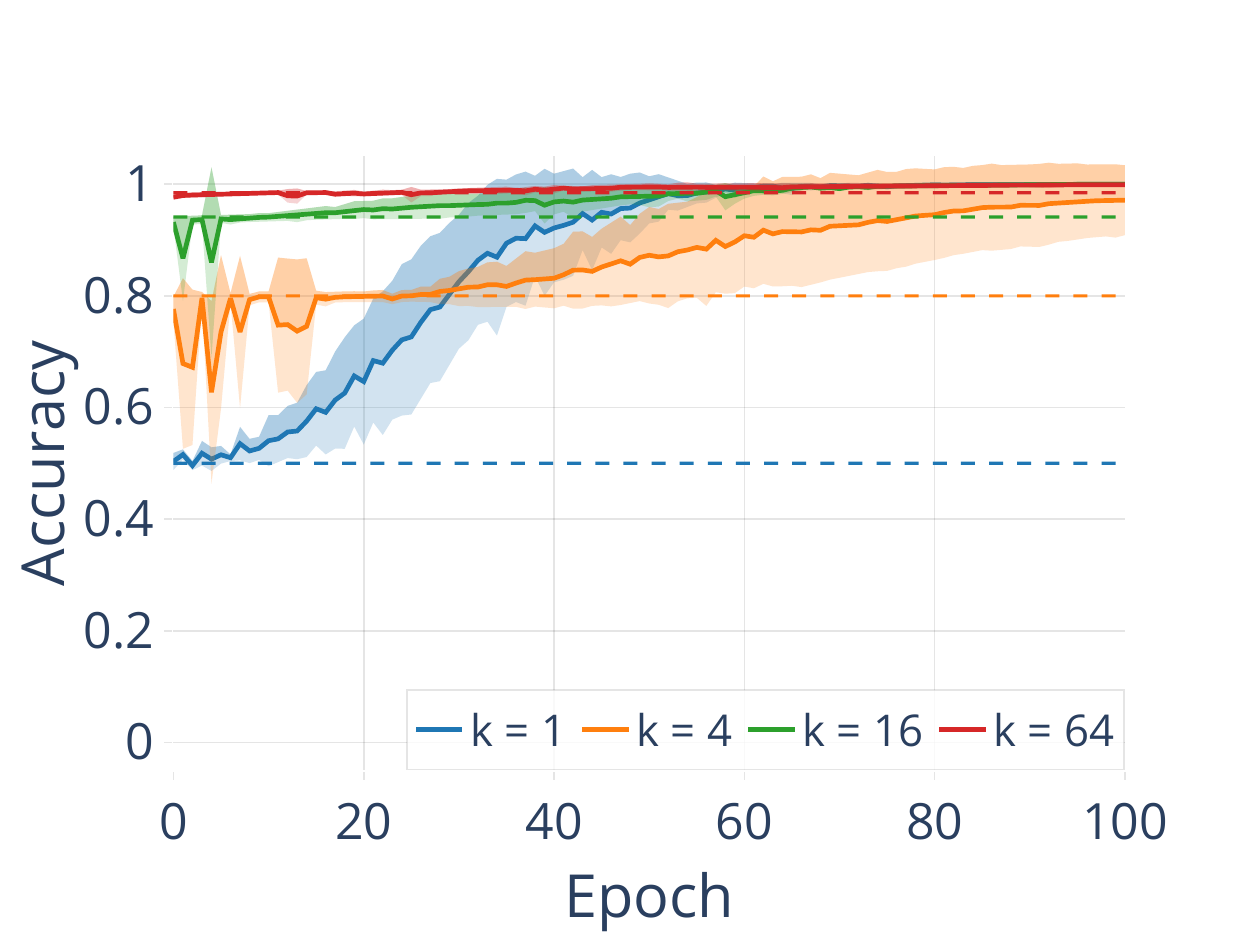}
    }
    \\
    \subfloat[Pythia-1B, $\ell = 7$]{
        \includegraphics[width=\thirdWidth]{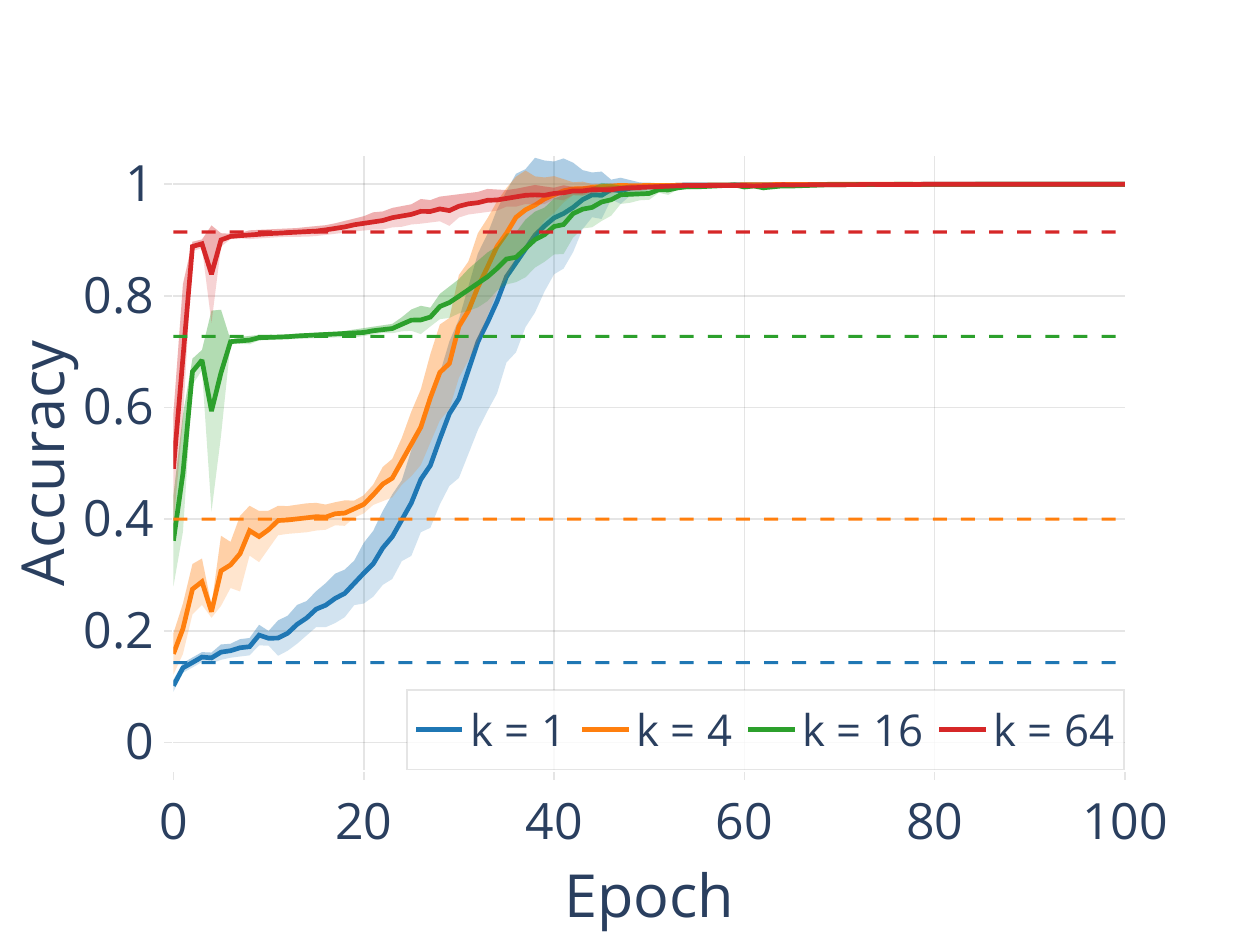}
    }
    \subfloat[Phi-2.7B, $\ell = 7$]{
        \includegraphics[width=\thirdWidth]{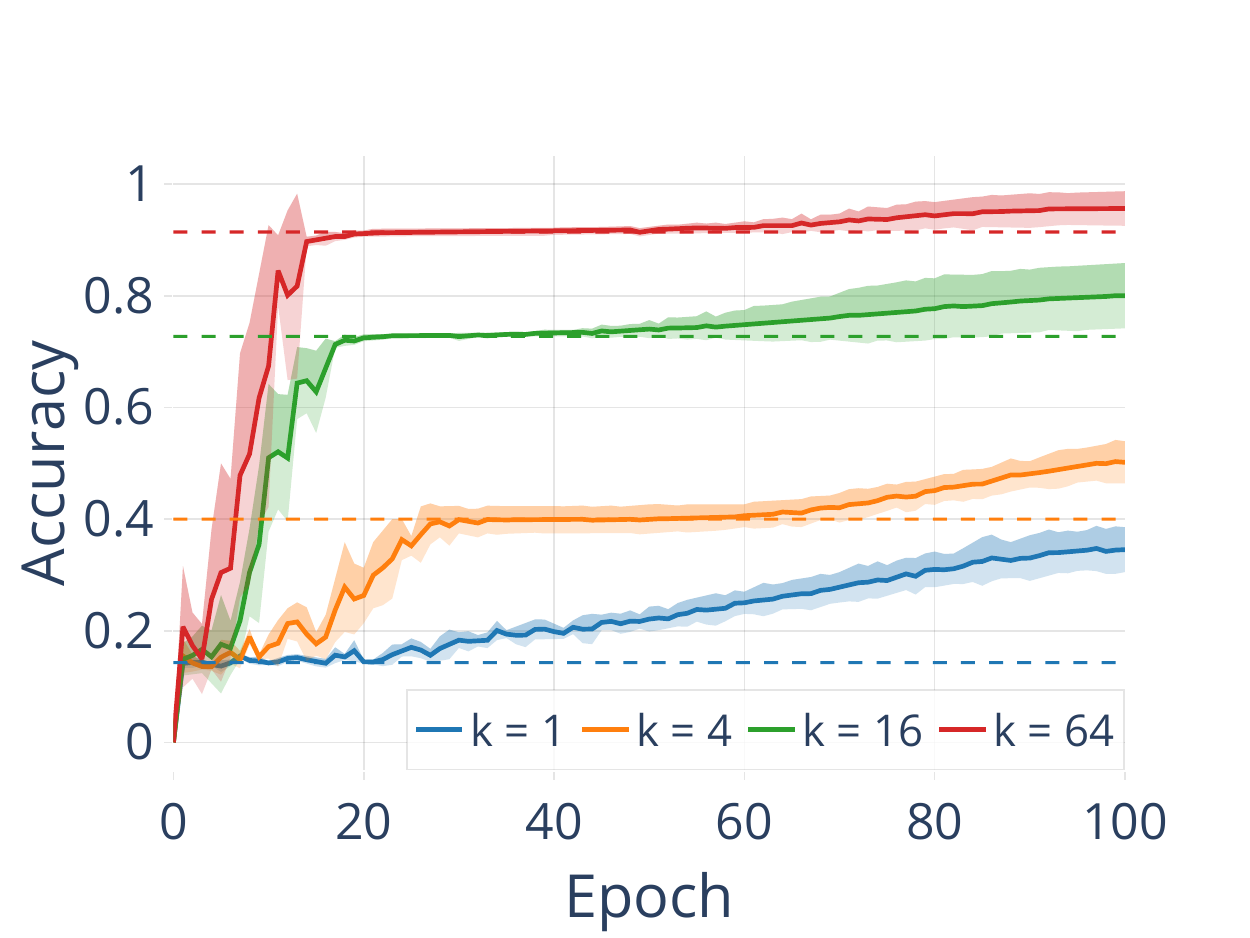}
    }
    \subfloat[Llama2-13B, $\ell = 7$]{
        \includegraphics[width=\thirdWidth]{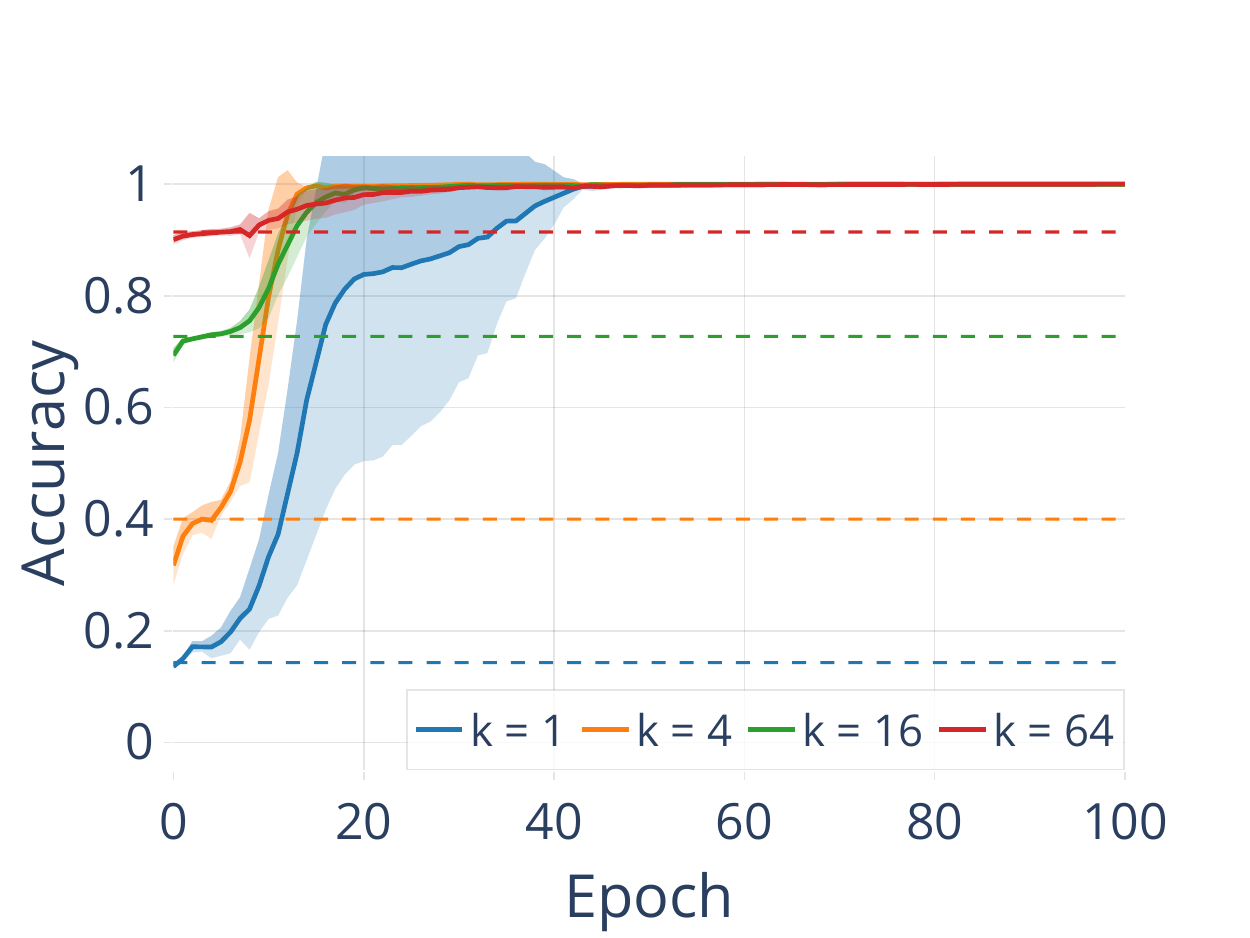}
    }
    \\
    \subfloat[Pythia-1B, $\ell = 26$]{
        \includegraphics[width=\thirdWidth]{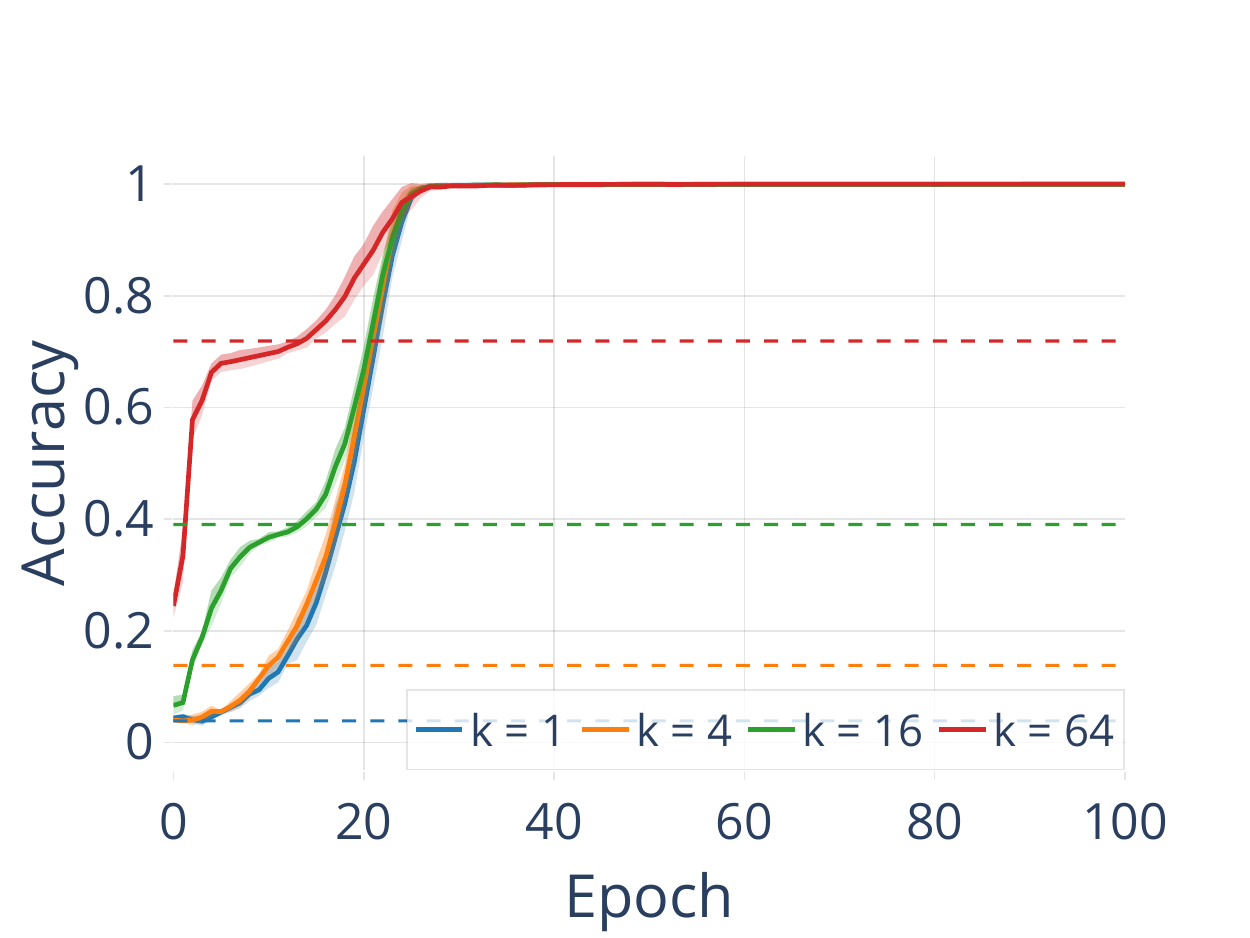}
    }
    \subfloat[Phi-2.7B, $\ell = 26$]{
        \includegraphics[width=\thirdWidth]{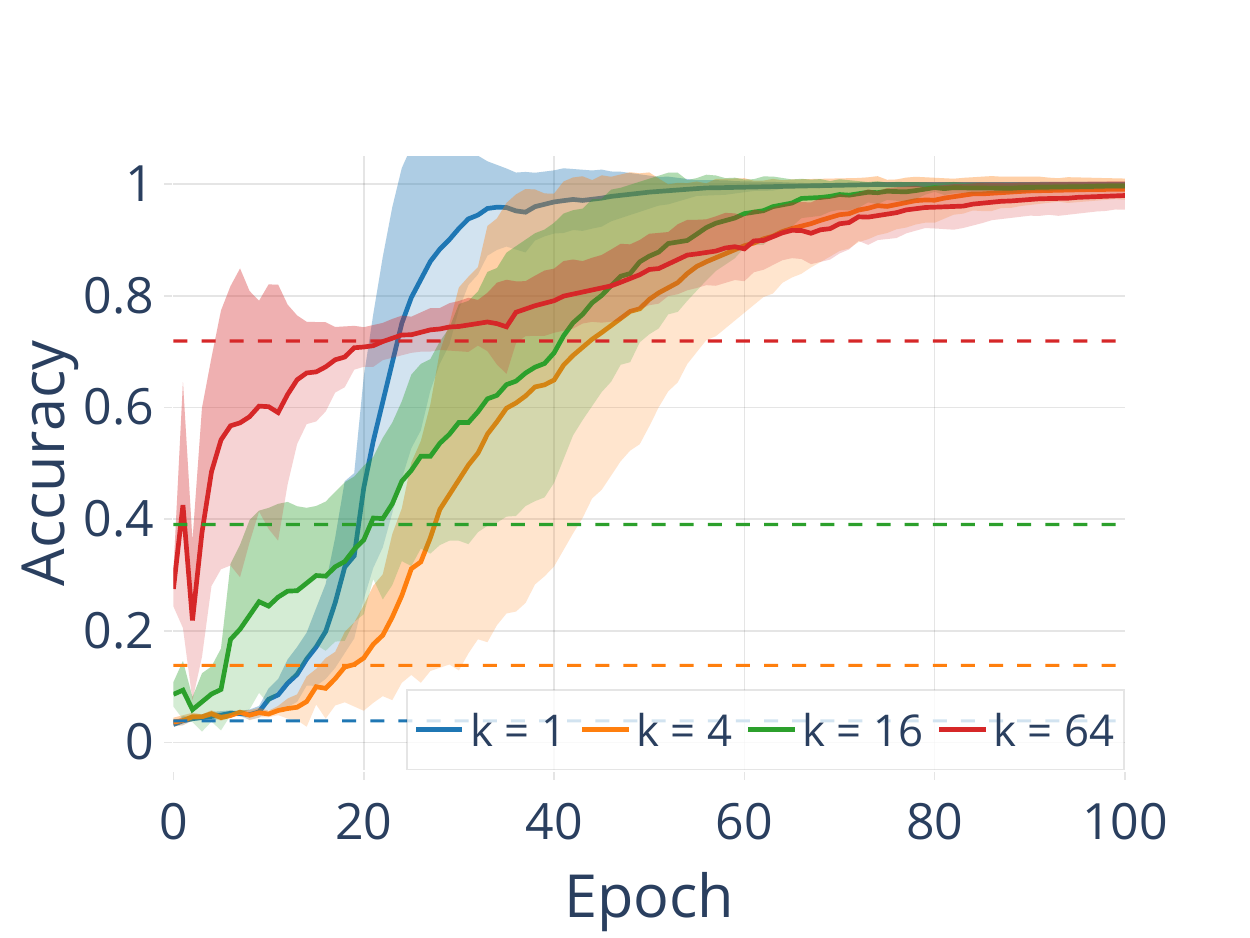}
    }
    \subfloat[Llama2-13B, $\ell = 26$]{
        \includegraphics[width=\thirdWidth]{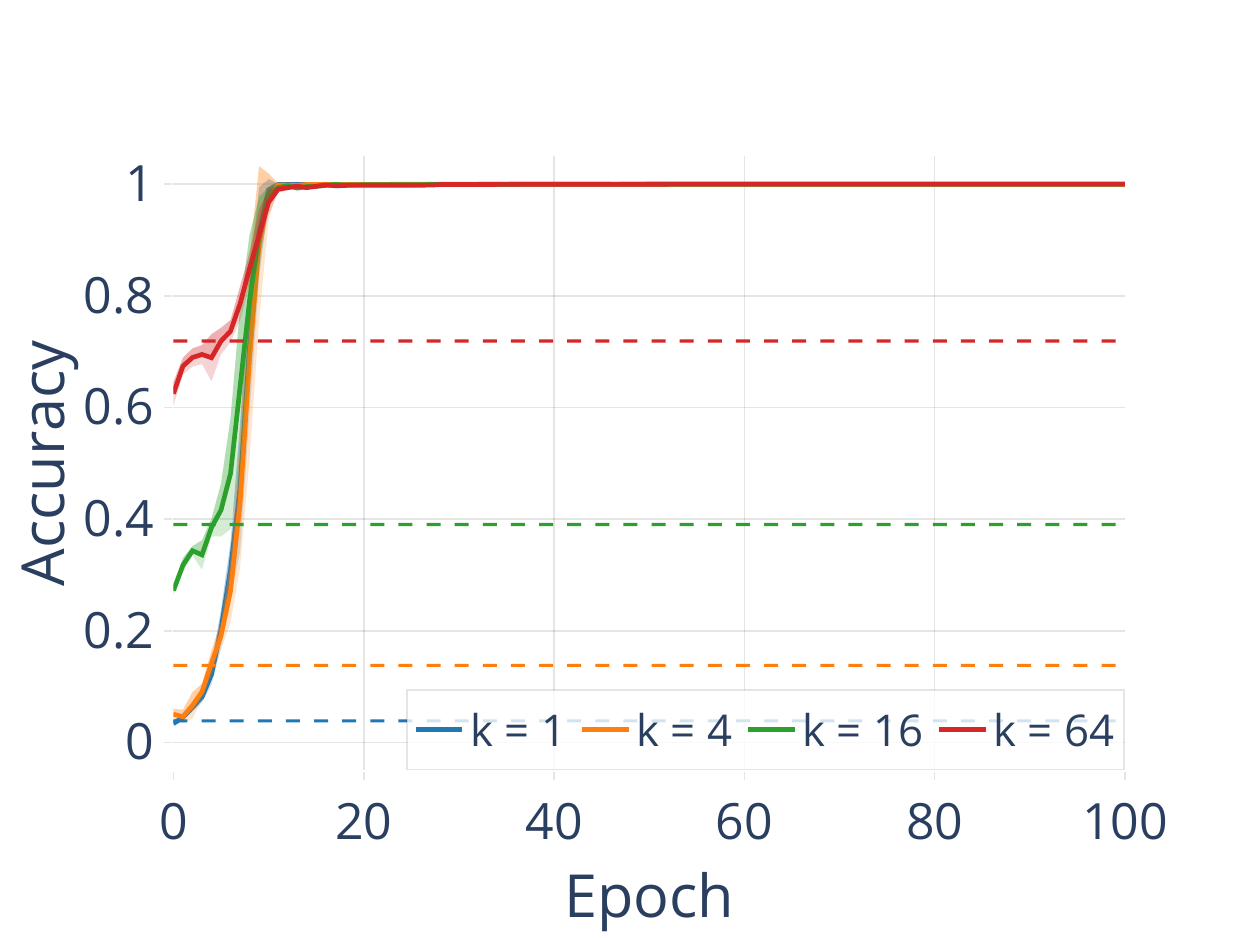}
    }
\caption{\capthead{Accuracy for conditional probability strings with different $\ell$.}{$n = 1024$}
    While the conditional entropy affects the accuracy models achieve during the \GuessPhase, \ie~models learn the conditional probability distribution of the string, the length of the \MemPhase is not affected by the conditional entropy.
}
\label{fig:conditional_rel-prob}
\end{figure}

We fix $n = 1$ and train and evaluate models on strings with different relative probabilities $p_k$, i.e. where the priviledged continuation tokens are $k$ times as likely to appear after their n-grams than the ramining tokens from $A$.
We use $k \in \{1, 4, 16, 64\}$.
For $k = 1$ the unconditional entropy $H(s)$ is the same as the conditional entropy $H_1(s) = H(s)$.

Figure~\ref{fig:conditional_rel-prob} shows the memorisation dynamics for different models and $\ell$.
Conditional entropy affects the accuracy models achieve during the \GuessPhase, which means that models are able to learn the conditional probabilities of the strings.
However, the length of the \MemPhase is not affected by the conditional entropy, since all strings are fully memorized at roughly the same epoch, with no consistent relationship between relative probability and full memorization epoch.
The effect of unconditional entropy, by comparison, is much stronger.

\subsection{Additional results on in-context learning}
\label{app:dynamics_in_context}

We saw in Section~\ref{sec:phases} that all models exhibit two phases of memorization, except the Llama2 models, which skip the \GuessPhase and start directly with the \MemPhase.
To better understand why this is happening, we test how well the different models are able to learn the distribution of the random strings $P_A$ via in-context learning.

\begin{figure}[H]
    \centering
    \subfloat[Pythia-1B]{
        \includegraphics[width=\thirdWidth]{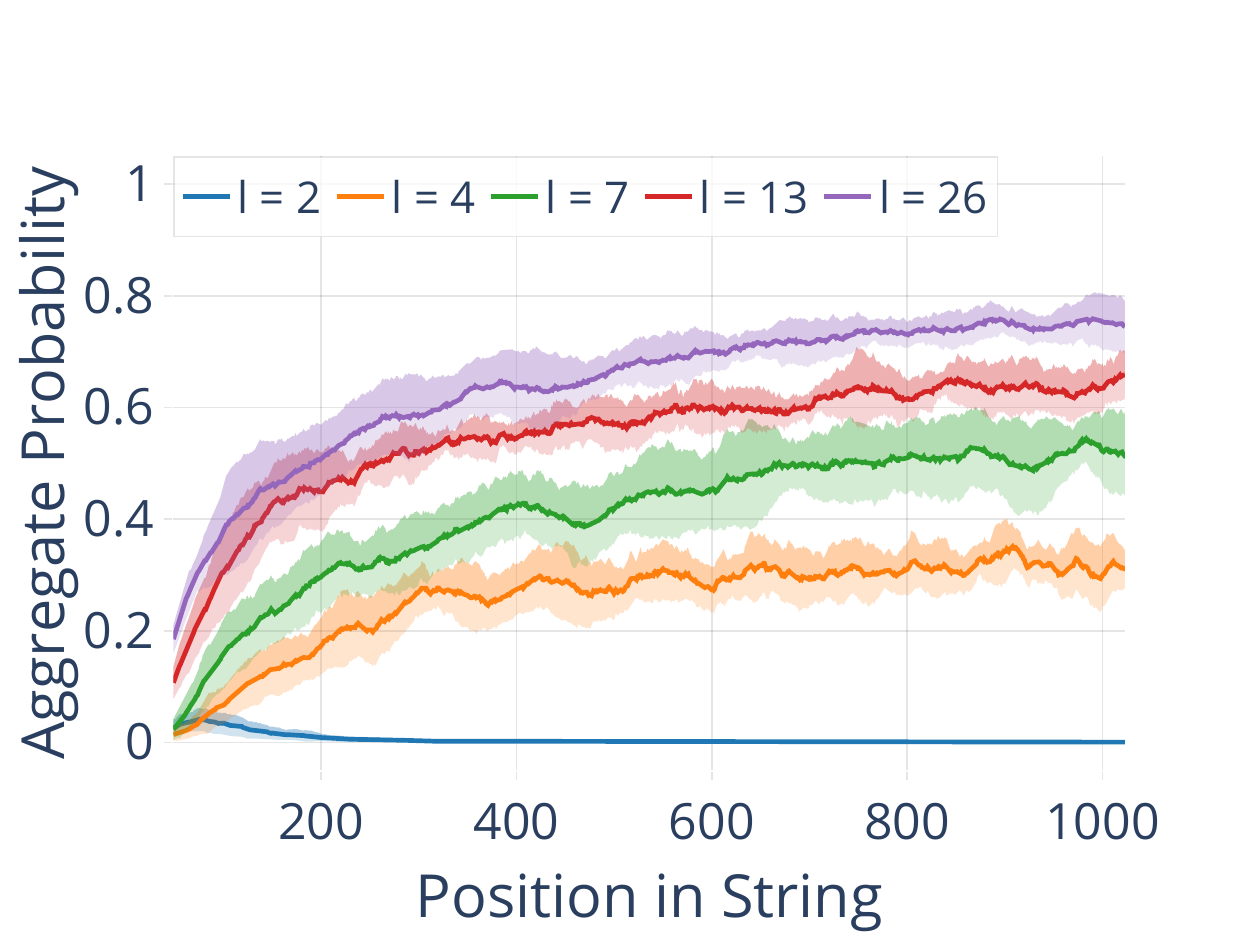}
    }
    \subfloat[Phi-2.7B]{
        \includegraphics[width=\thirdWidth]{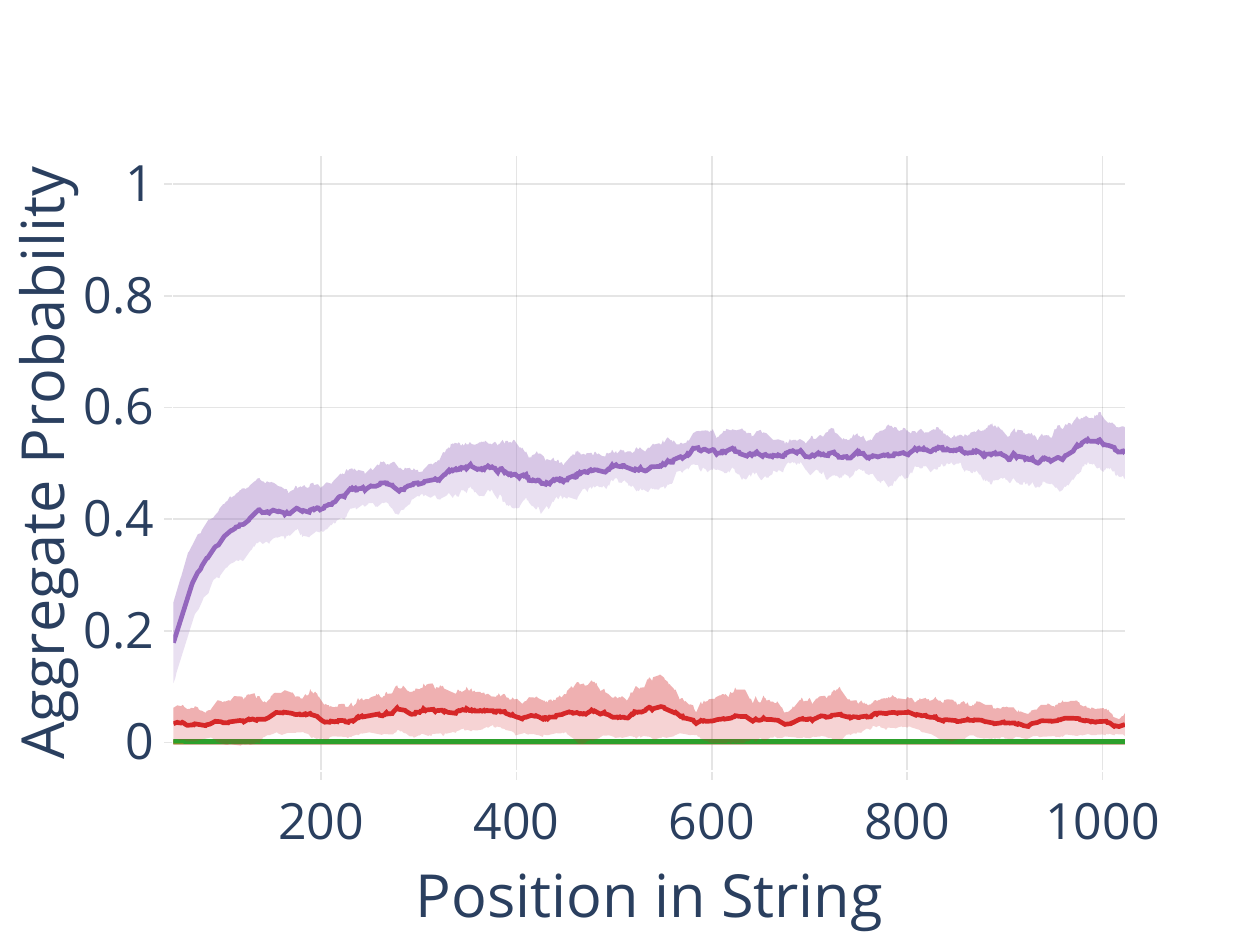}
    }
    \subfloat[Llama2-13B]{
        \includegraphics[width=\thirdWidth]{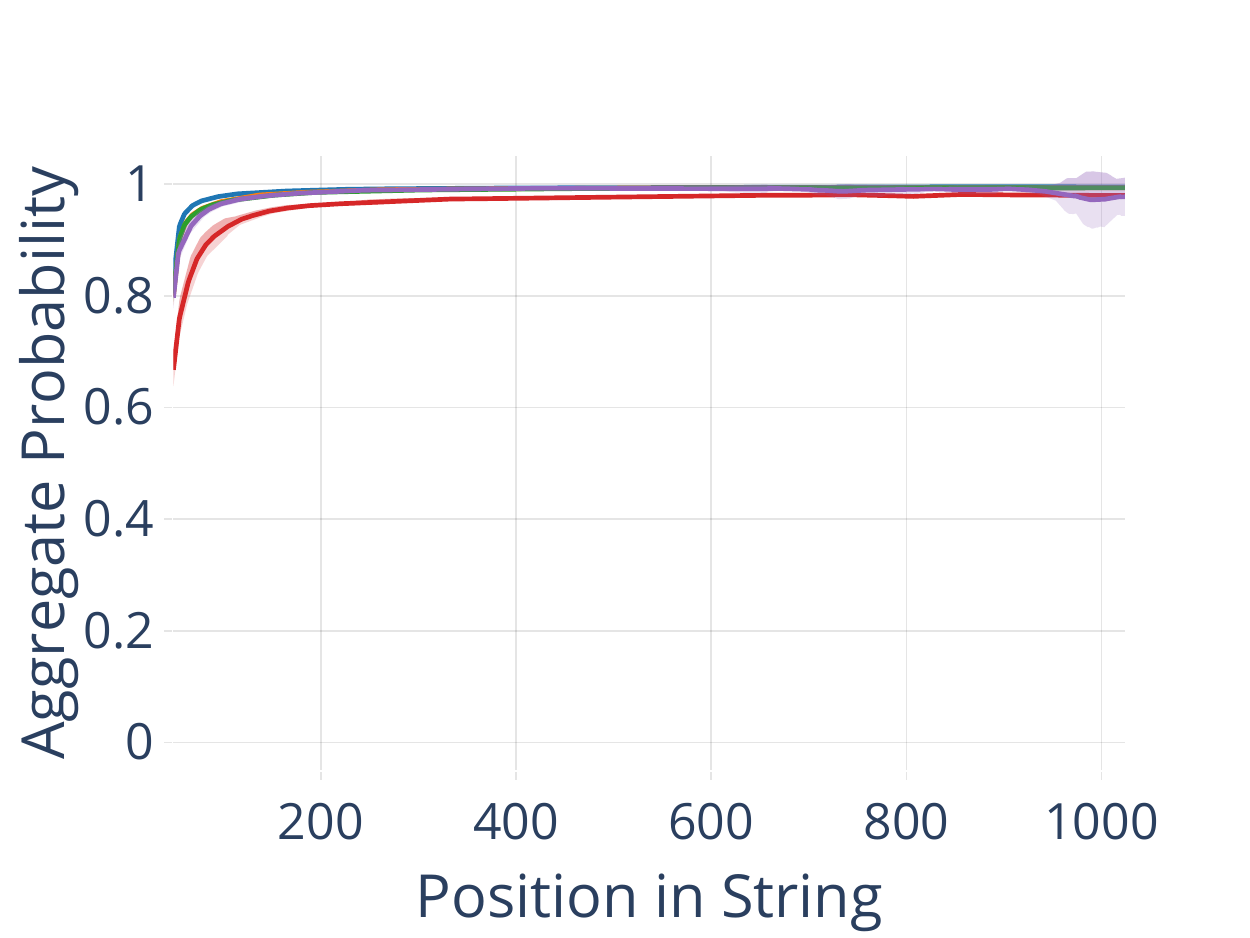}
    }
\caption{\capthead{Aggregate probability over $A$ at different string positions, before training.}{$n = 1024$}
    The aggregate probability over the tokens in the alphabet shows how well models are able to infer $P_A$ from the prefix of the string at a given positions.
    Models differ in their ability to learn the distribution via in-context learning.
    Llama-2 models are particularly good, assigning almost all probability mass to tokens in the alphabet with just 100 tokens of context.
    Other models do not exhibit the same in-context learning abilities.
}
\label{fig:dynamics_in_context}
\end{figure}

In Figure~\ref{fig:dynamics_in_context} we show the aggregate probability over all tokens in the alphabet $A$ that models assign at each position in the string.
We use a sliding window of size 50 to smooth the curves.
The figure shows models at epoch 0, \ie~before they have started to memorize the string.
Without any training, models can only infer the string distribution, \ie~detect that the string only contains tokens from $A$ and not other tokens from $V$, via in-context learning.

Indeed, we see that Llama2 models exhibit strong in-context learning abilities and quickly assign all probability mass to tokens within the alphabet.
The other models are not able to infer the distribution nearly as well.
Note that the differences in in-context learning ability observed in Figure~\ref{fig:dynamics_in_context} correspond to the differences in the initial loss in Figure~\ref{fig:accuracy_loss_alphabet_size}.
Thus, the \GuessPhase appears to be a stage that all models go through, but sufficiently strong in-context learning abilities allow models to effectively shorten it to zero.

\section{Results on Memorization Order}
\label{app:memorization_order}

We aim to characterize the \MemPhase more closely and ask:
\emph{Is there a specific ordering in which tokens are memorised or are the positions of the correctly recollected tokens random?}

\subsection{Visualizing Memorisation Order}
\label{app:memorization_order_results}

\begin{figure}[H]
    \centering
    \subfloat[Pythia-1B, $\ell=2$]{
        \includegraphics[width=\smallThirdWidth]{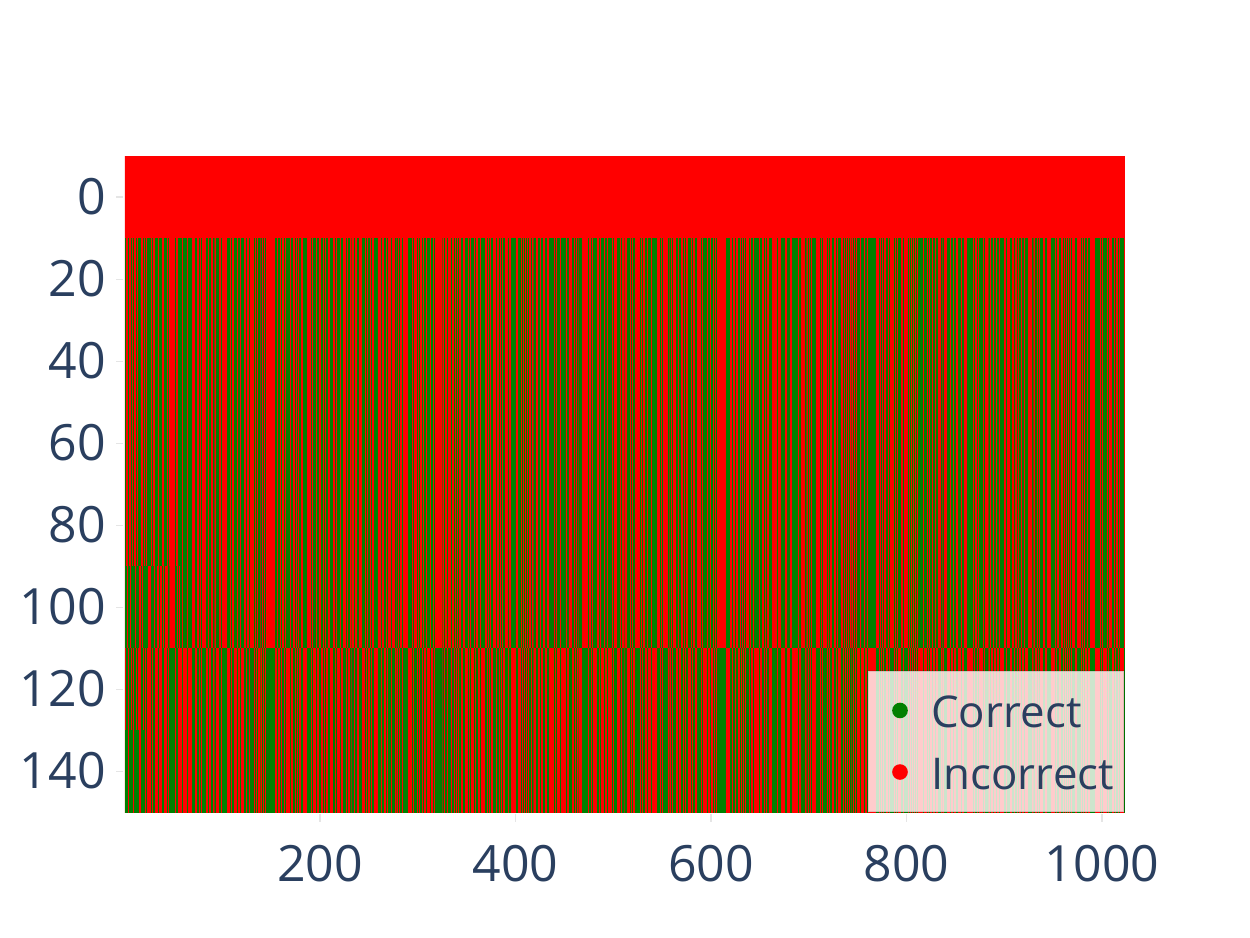}
    }
    \subfloat[Phi-2.7B, $\ell=2$]{
        \includegraphics[width=\smallThirdWidth]{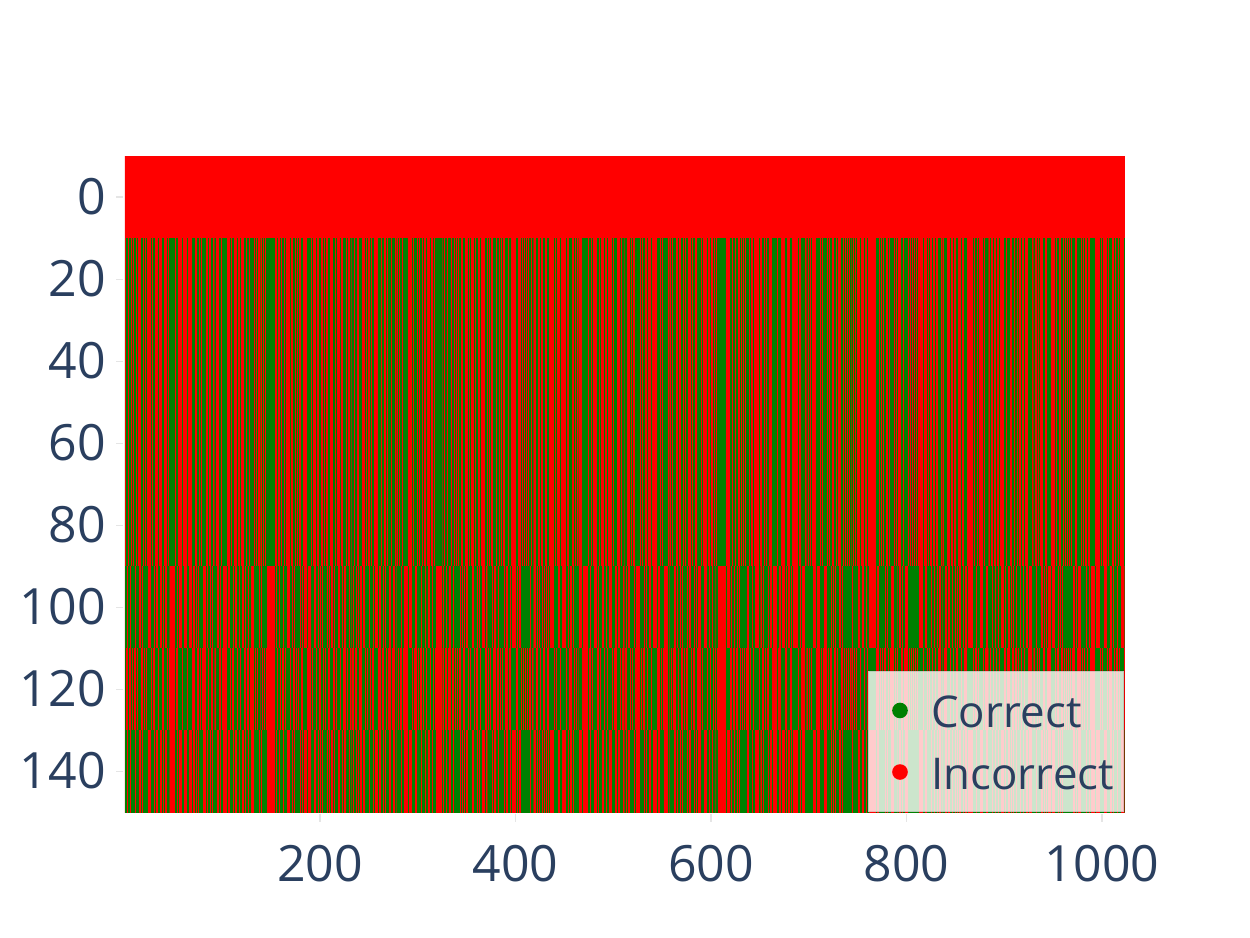}
    }
    \subfloat[Llama2-13B, $\ell=2$]{
        \includegraphics[width=\smallThirdWidth]{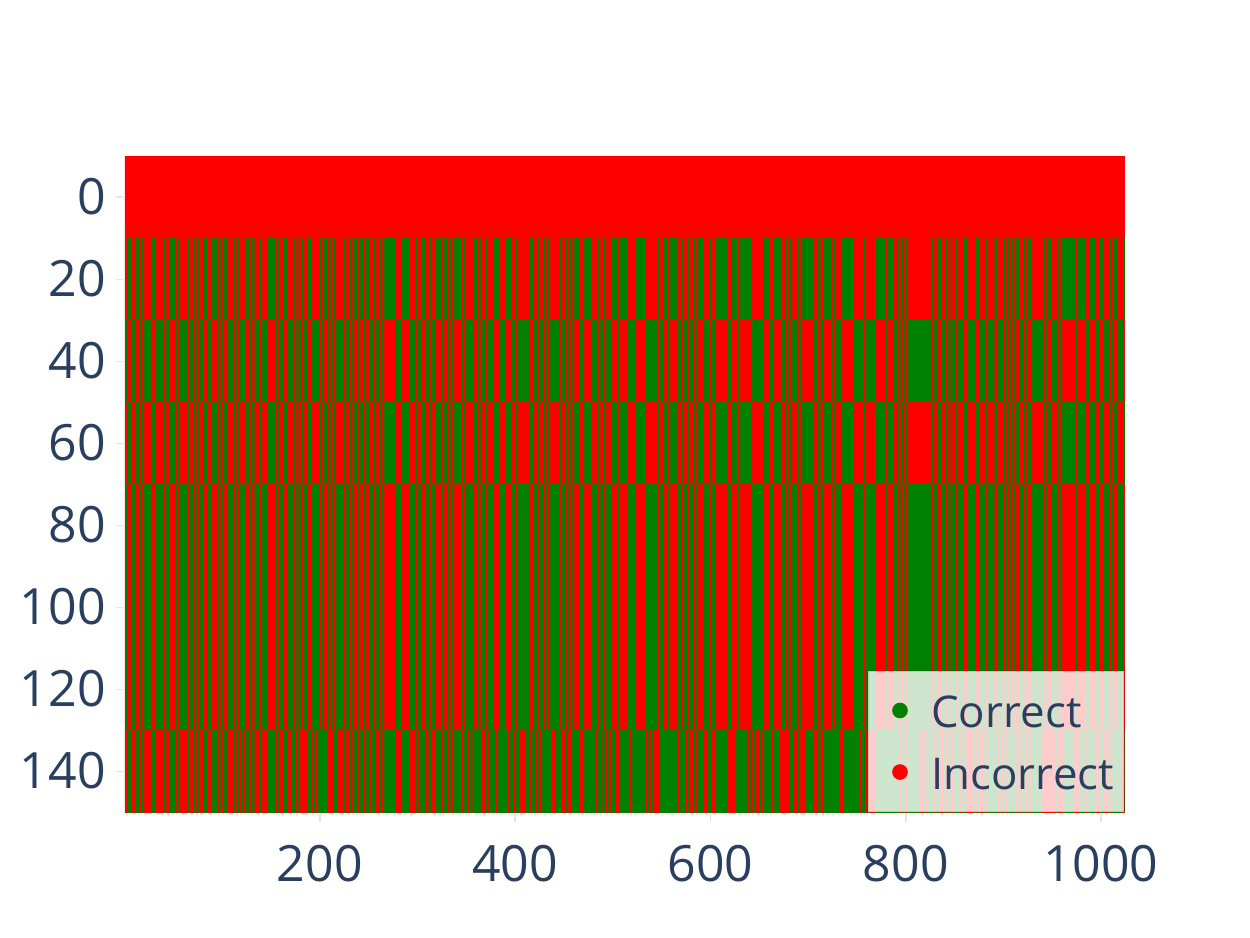}
    }
    \\
    \subfloat[Pythia-1B, $\ell=4$]{
        \includegraphics[width=\smallThirdWidth]{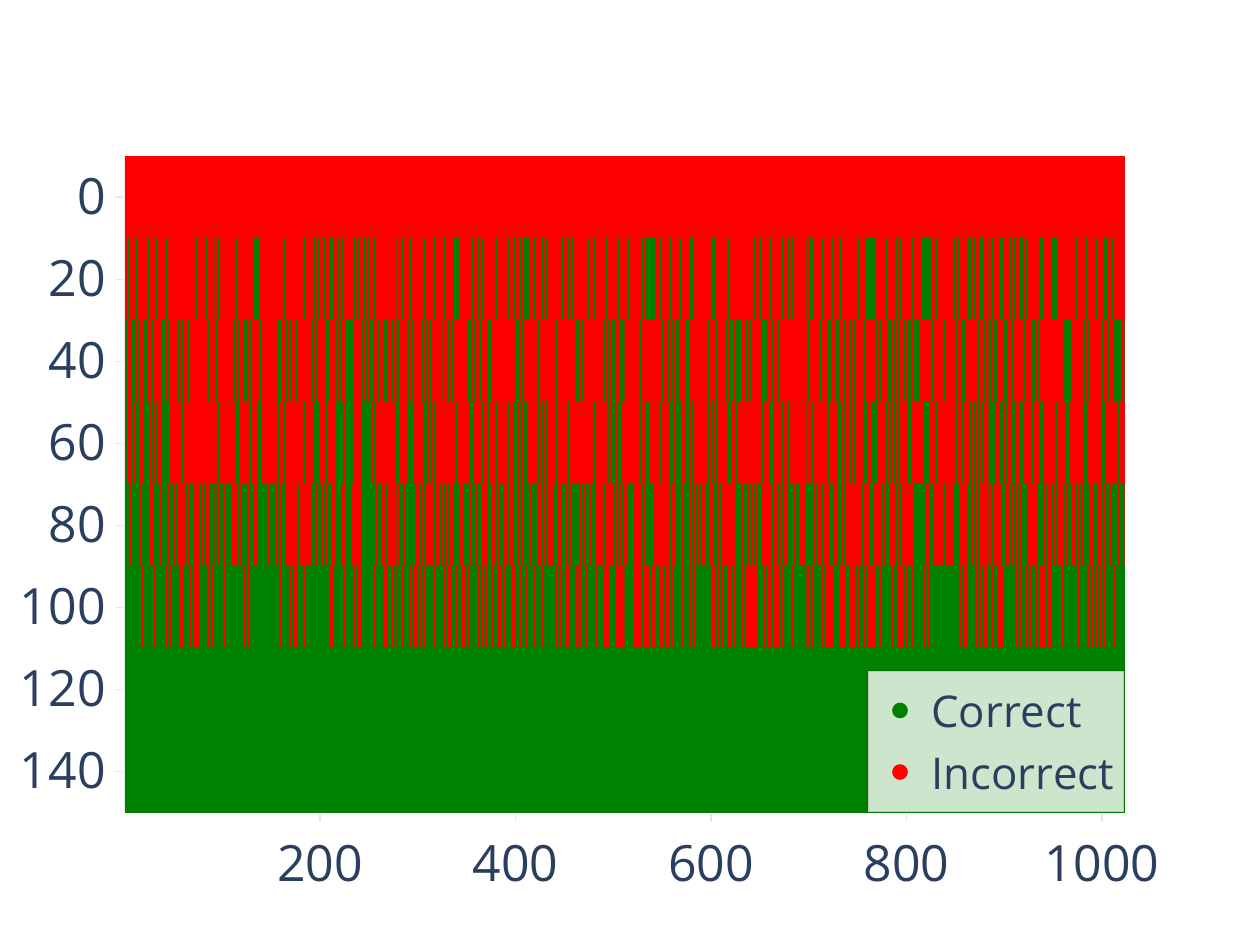}
    }
    \subfloat[Phi-2.7B, $\ell=4$]{
        \includegraphics[width=\smallThirdWidth]{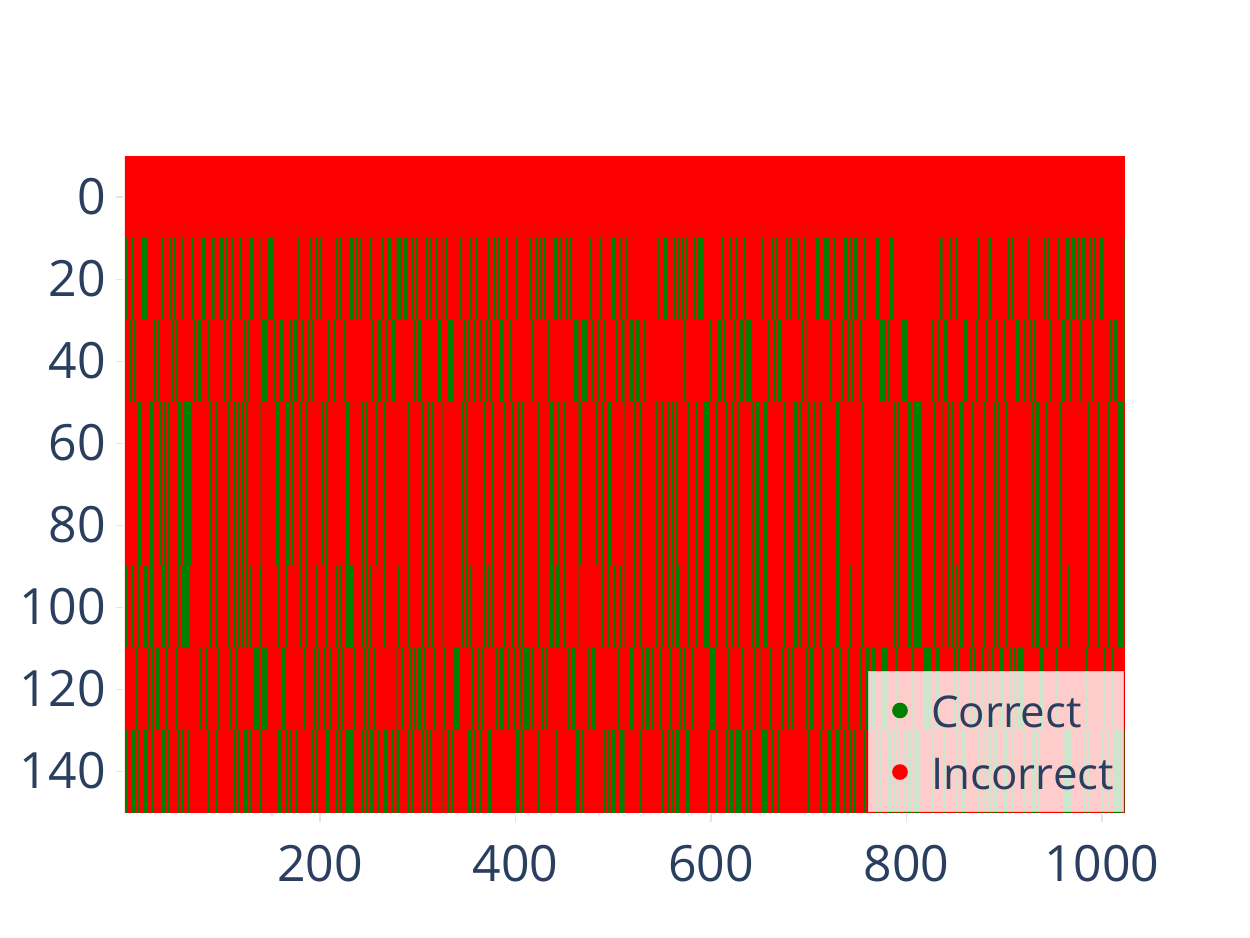}
    }
    \subfloat[Llama2-13B, $\ell=4$]{
        \includegraphics[width=\smallThirdWidth]{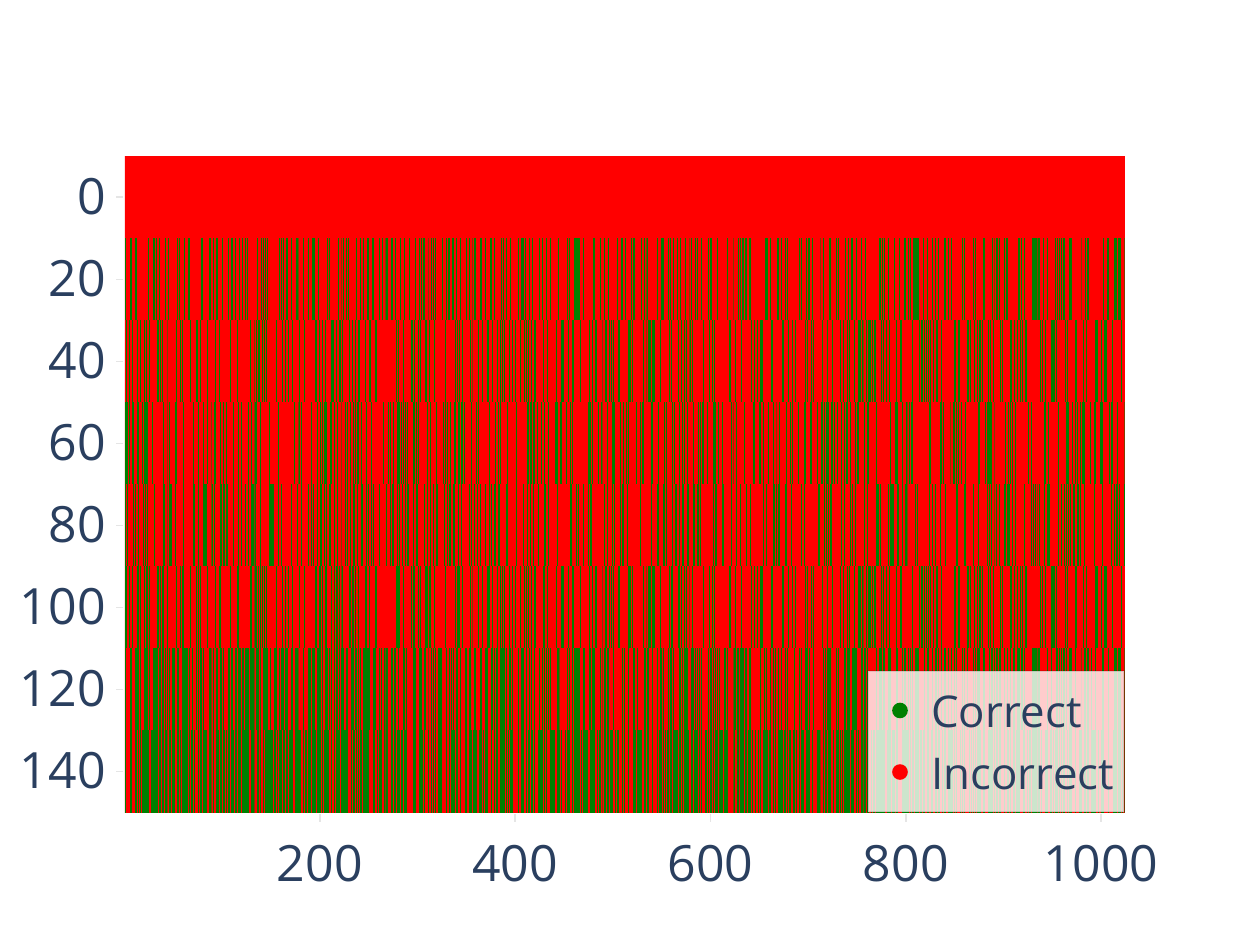}
    }
    \\
    \subfloat[Pythia-1B, $\ell=7$]{
        \includegraphics[width=\smallThirdWidth]{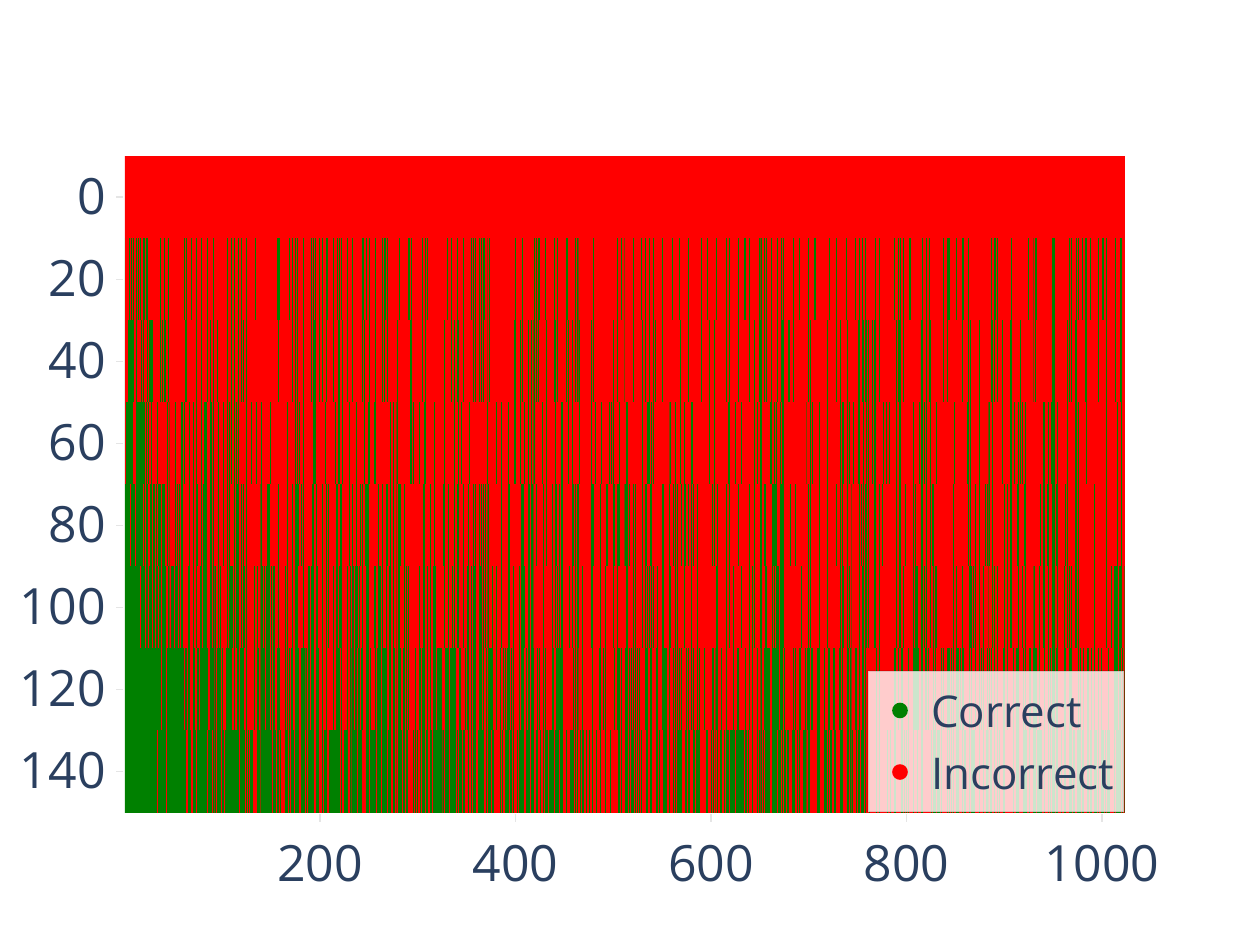}
    }
    \subfloat[Phi-2.7B, $\ell=7$]{
        \includegraphics[width=\smallThirdWidth]{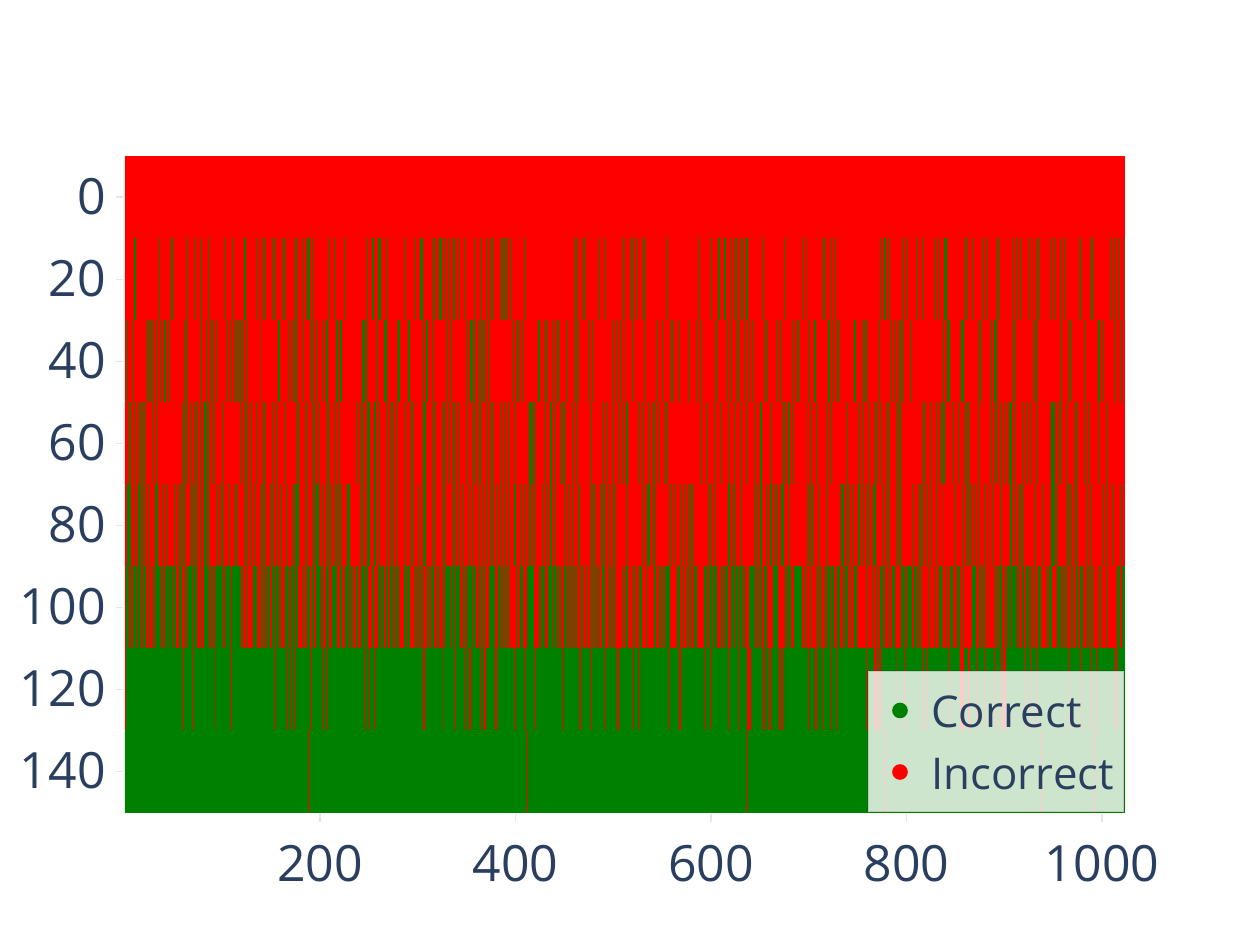}
    }
    \subfloat[Llama2-13B, $\ell=7$]{
        \includegraphics[width=\smallThirdWidth]{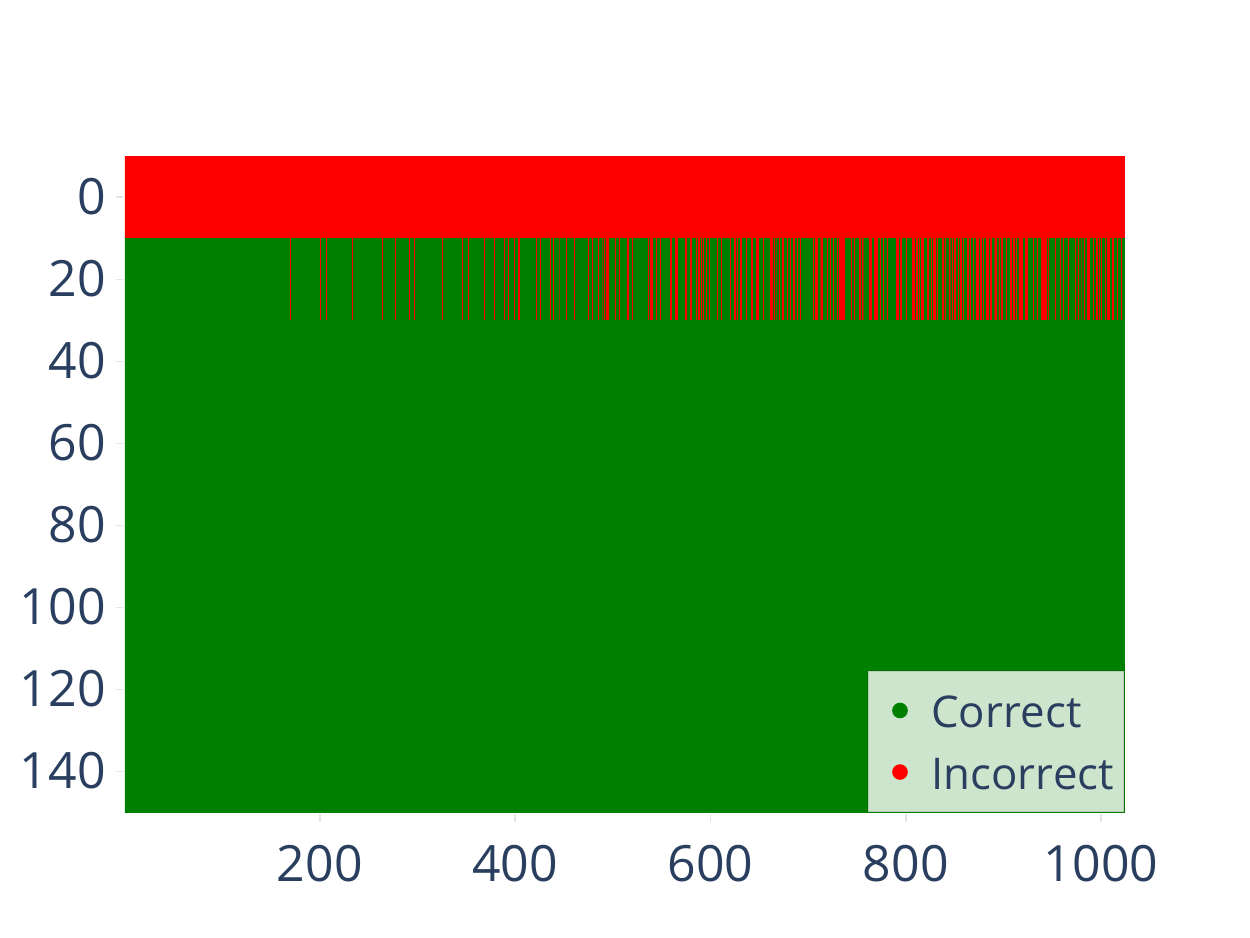}
    }
    \\
    \subfloat[Pythia-1B, $\ell=13$]{
        \includegraphics[width=\smallThirdWidth]{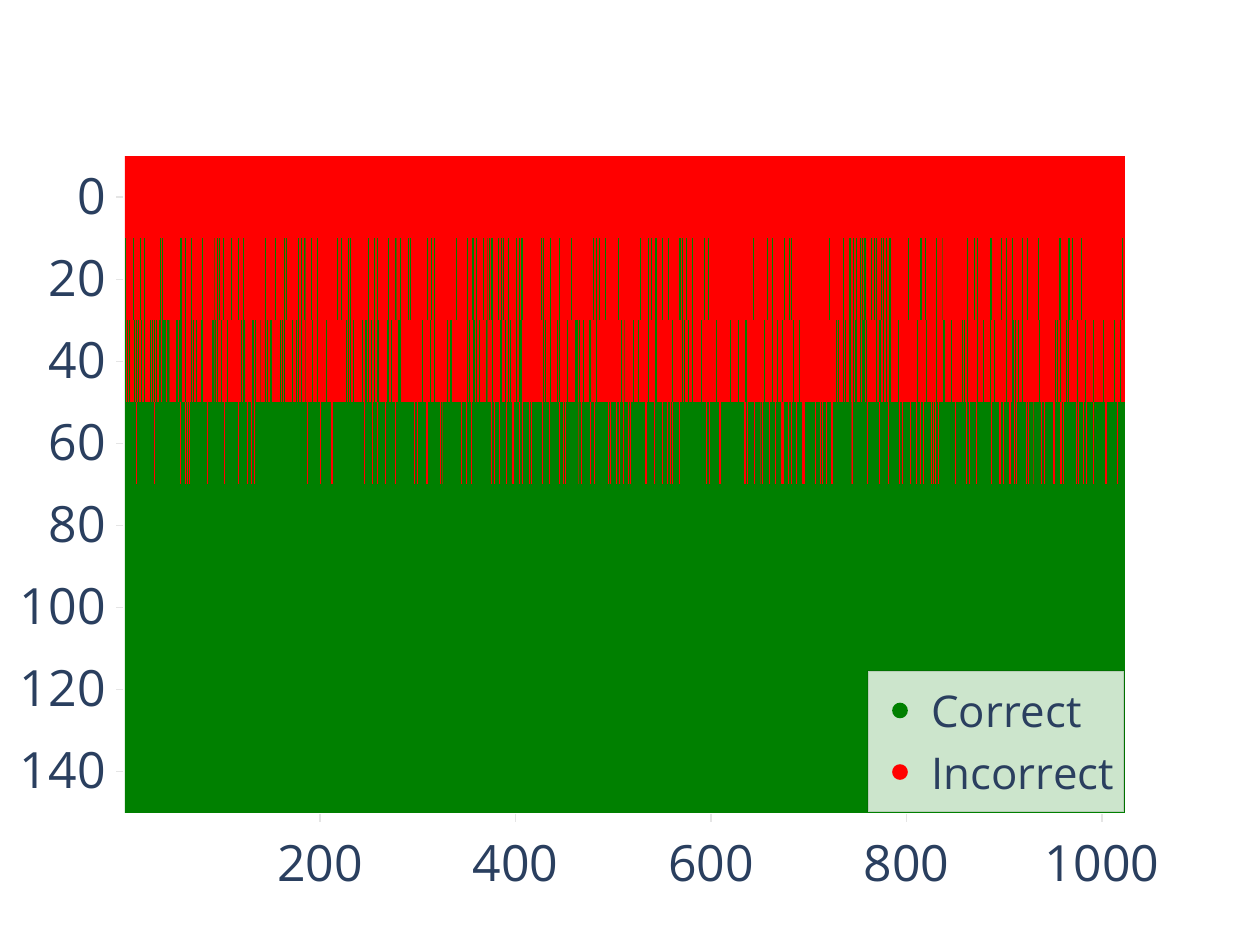}
    }
    \subfloat[Phi-2.7B, $\ell=13$]{
        \includegraphics[width=\smallThirdWidth]{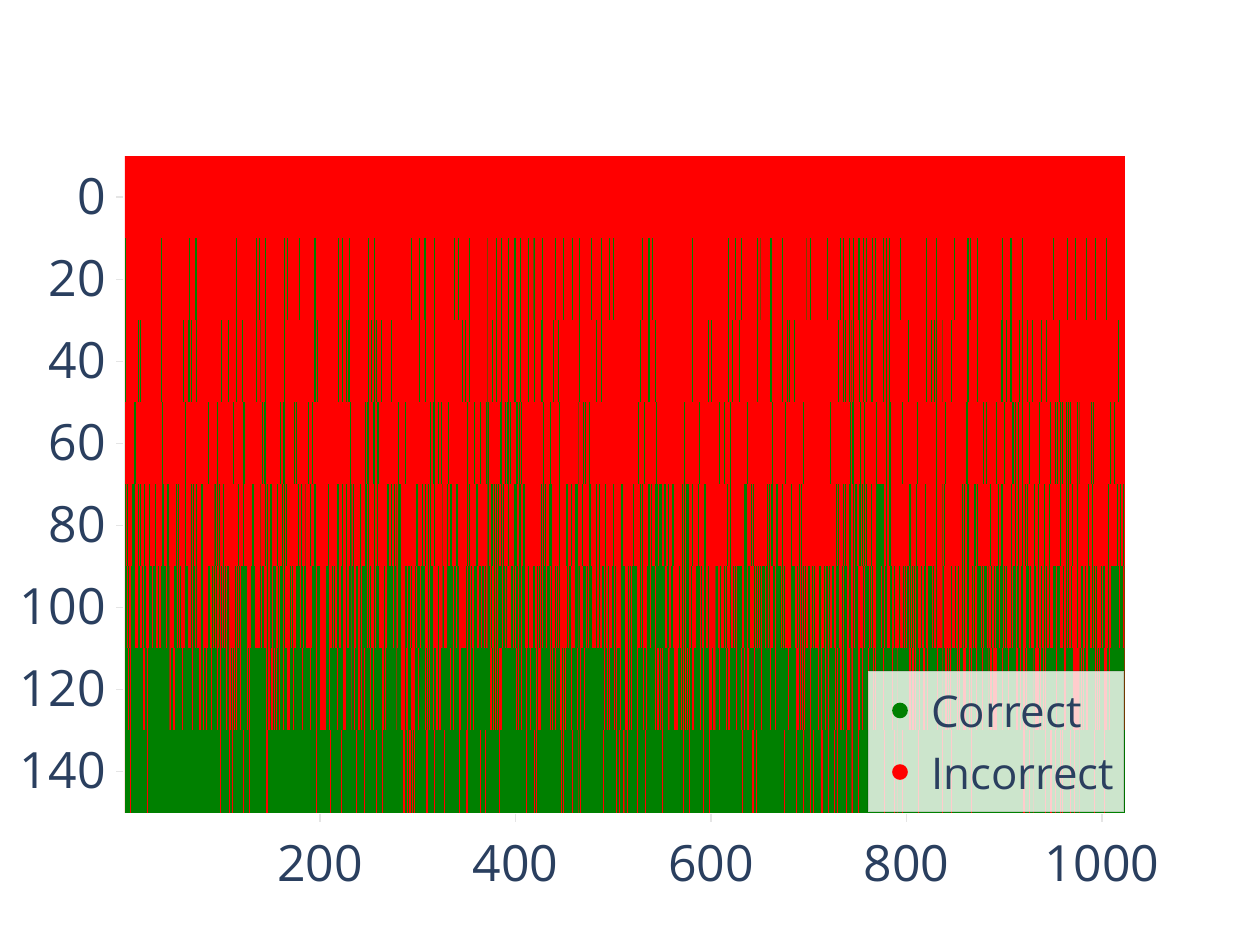}
    }
    \subfloat[Llama2-13B, $\ell=13$]{
        \includegraphics[width=\smallThirdWidth]{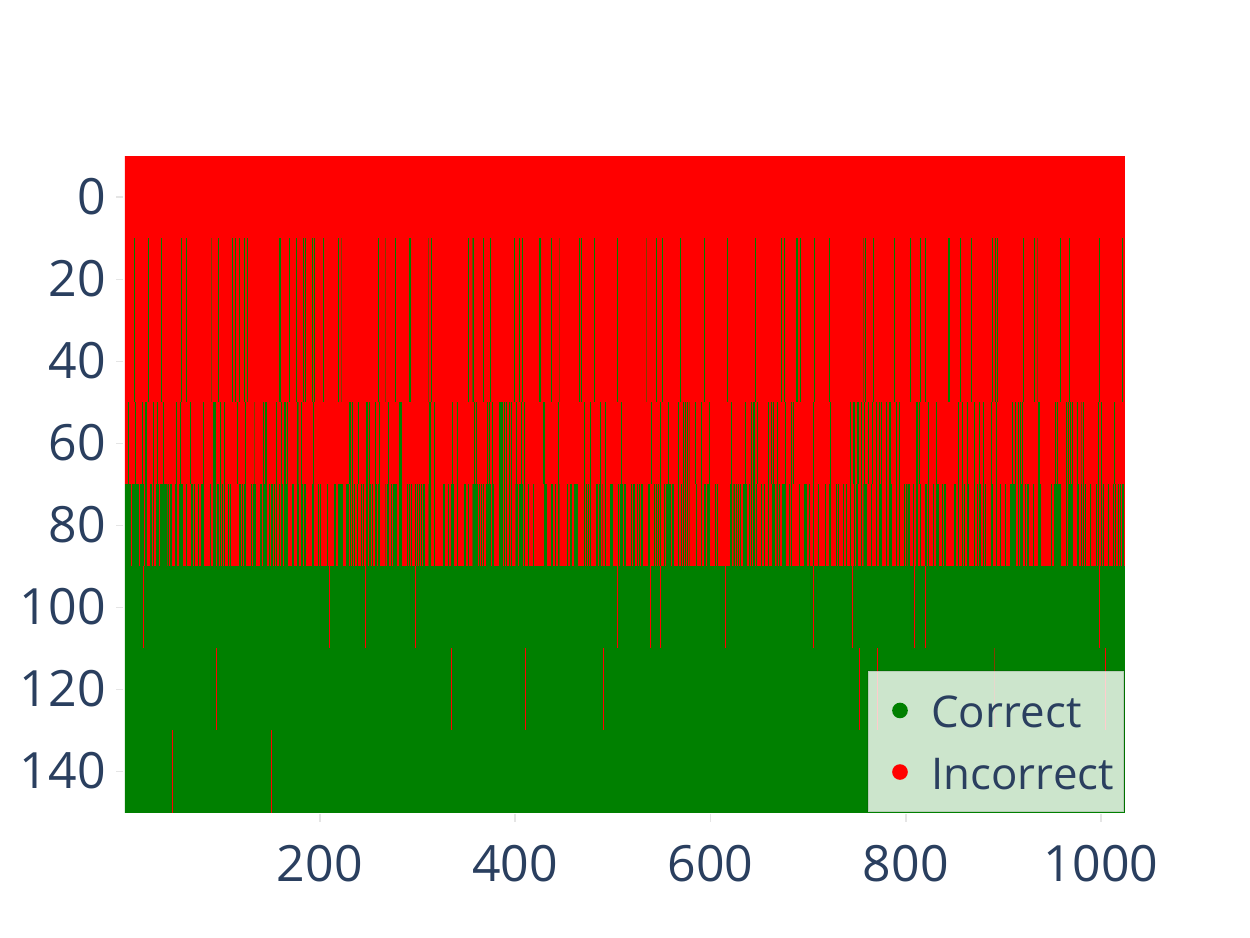}
    }
    \\
    \subfloat[Pythia-1B, $\ell=26$]{
        \includegraphics[width=\smallThirdWidth]{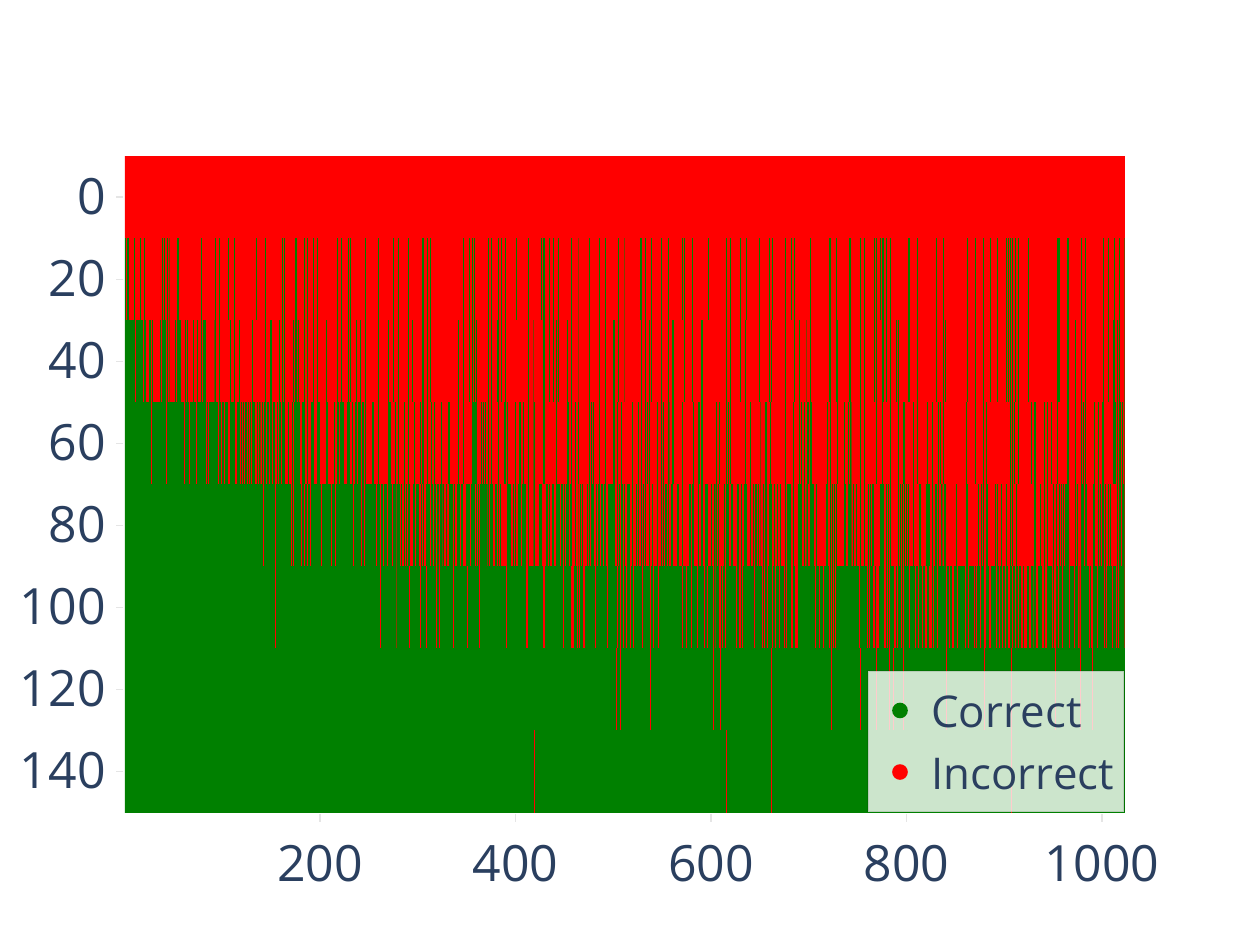}
    }
    \subfloat[Phi-2.7B, $\ell=26$]{
        \includegraphics[width=\smallThirdWidth]{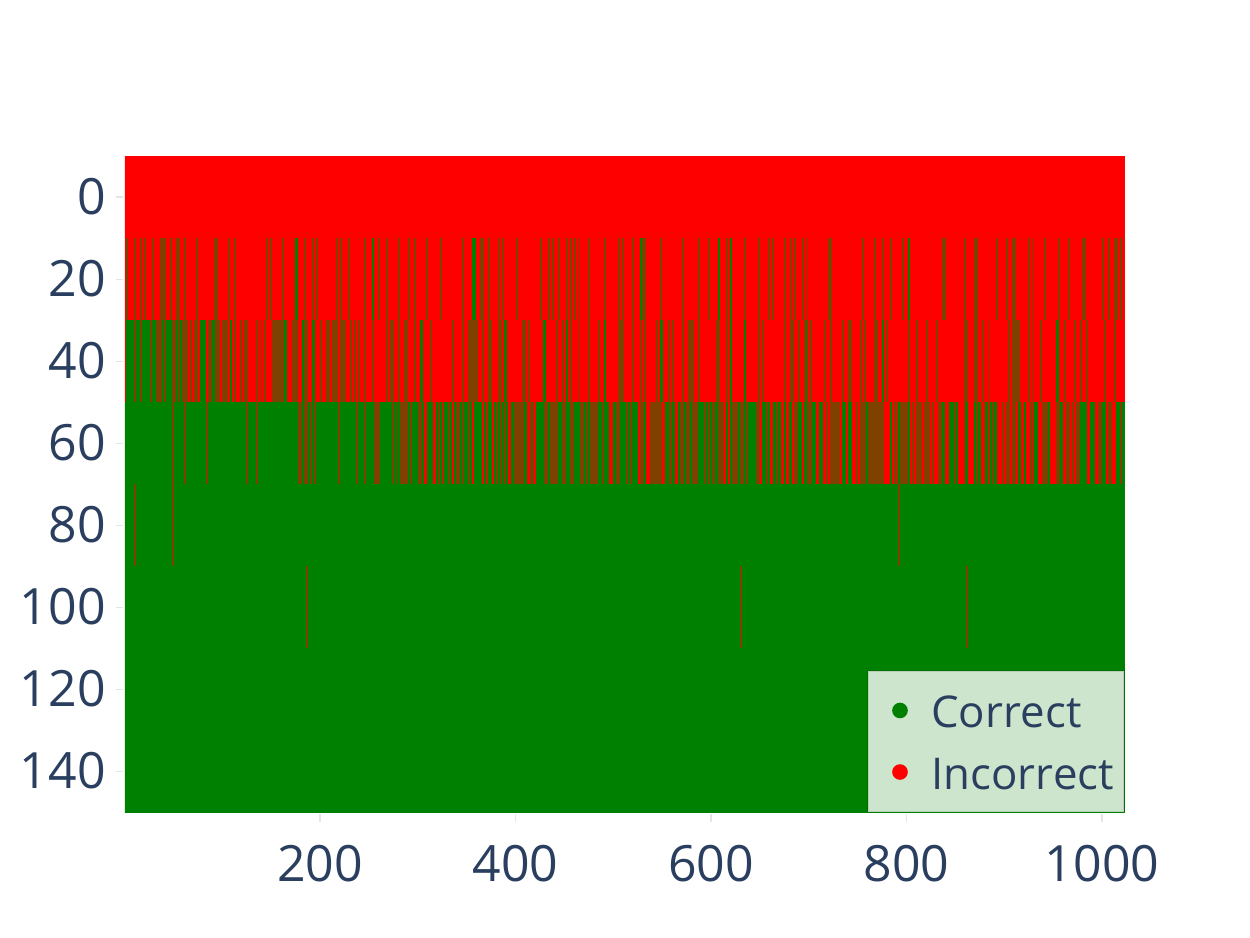}
    }
    \subfloat[Llama2-13B, $\ell=26$]{
        \includegraphics[width=\smallThirdWidth]{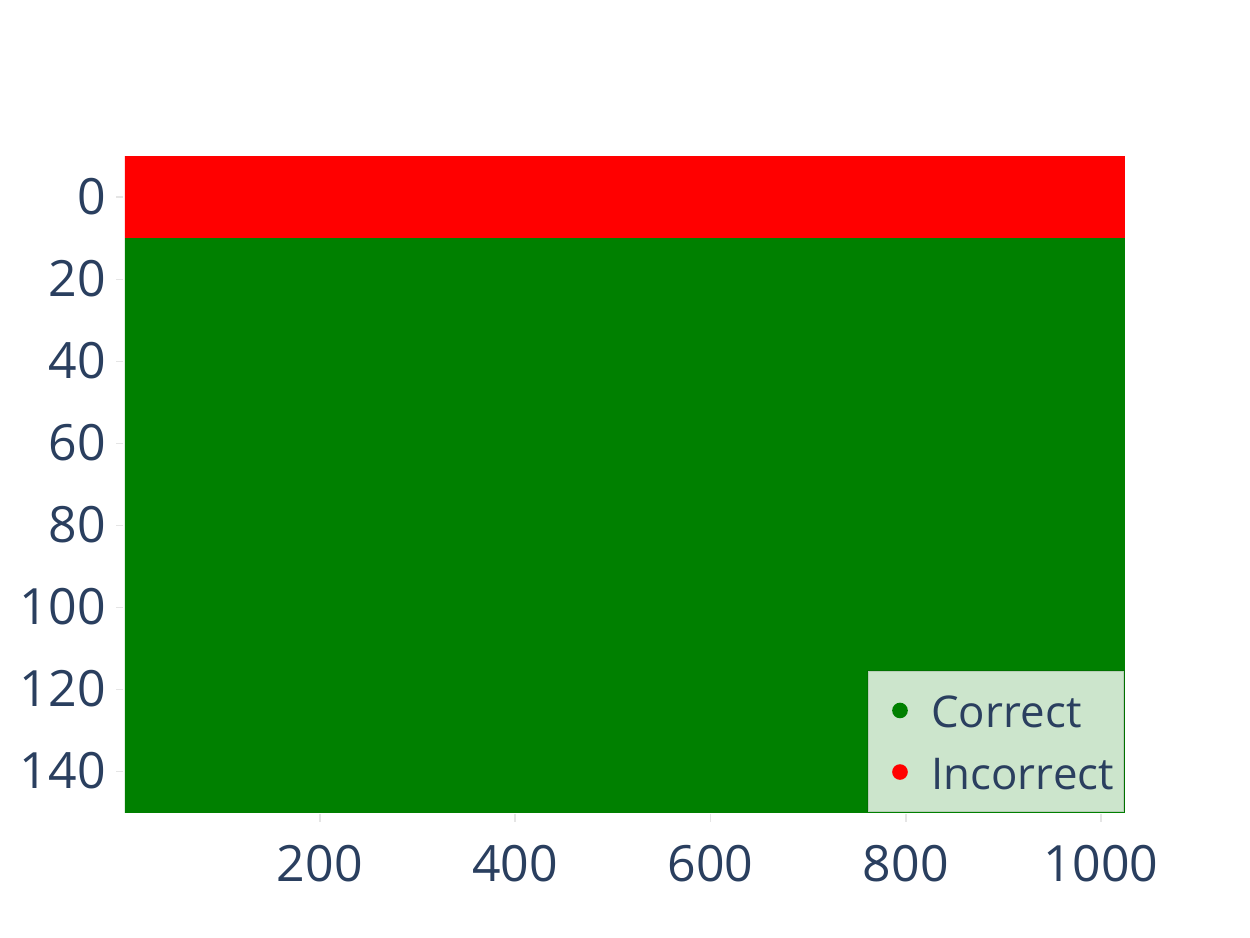}
    }
\caption{\capthead{Memorization order for different pretrained models for different $\ell$.}{$n = 1024$}
Memorisation across strings happens in essentially random order, and is unstable until the \MemPhase starts to converge.
Llama2-13B models tend to memorise tokens at the beginning of the string slightly earlier, which may be because --- in contrast to the other models --- they use a beginning of string (BOS) token that could serve as a reference.
Whether BOS tokens indeed affect memorisation order needs more careful exploration, however.
}
\label{fig:mem_order_all}
\end{figure}

Figure~\ref{fig:mem_order_all} shows which tokens in a string have been memorised correctly during the early epochs of training.
There is no discernible order to the memorisation.
Some tokens in the middle or at the end of the string are memorised before earlier tokens and vice versa.
Additionally, memorisation is not stable in the initial parts of training, until the \MemPhase has made some progress;
previously memorised tokens are often forgotten (\ie,~predicted incorrectly)  at later epochs, until memorisation starts to converge, around epoch $20$.

\subsection{Quantifying randomness in memorisation order using rank correlation}

To quantify whether memorisation order is indeed random, \ie~does not depend on the position of a token in the string, we compute the rank correlation between the tokens’ positions and the epochs at which they are memorized
As memorisation epoch, we use both the initial memorisation epoch (\ie~the epoch when the token is first predicted correctly), as well as the stable memorisation epoch (\ie~the first epoch at and after which the token is not predicted incorrectly anymore).

\begin{table}[ht]
\centering
\small{
\begin{tabular}{l|ccccc|ccccc}
 & \multicolumn{5}{c|}{Initial Memorisation} & \multicolumn{5}{c}{Stable Memorisation} \\
Model & $l = 2$ & $l = 4$ & $l = 7$ & $l = 13$ & $l = 26$
      & $l = 2$ & $l = 4$ & $l = 7$ & $l = 13$ & $l = 26$ \\
\hline
Pythia-1B & -0.052 & -0.079 & -0.058 & -0.026 & -0.002
          & 0.081 & 0.105 & -0.005 & 0.060 & 0.095 \\
Phi-2.7B & -0.007 & -0.000 & -0.017 & -0.019 & -0.021
          & 0.018 & 0.072 & 0.026 & 0.055 & -0.020 \\
Llama2-13B & -0.017 & -0.029 & 0.031 & 0.045 & 0.095
          & 0.109 & 0.138 & 0.253 & 0.166 & 0.171 \\
\end{tabular}
}
\vspace{0.5em}
\caption{\capthead{Spearman rank correlation between token position in the string and the epoch at which tokens are memorised, for \emph{pretrained models}.}{$n = 1024$}
Correlation is very low in all cases, except for Llama2-13B models, where the stable memorisation correlation is slightly higher, presumably because they use a BOS token that could serve as a reference.
}
\label{tab:rank_correlation}
\end{table}

\begin{table}[ht]
\centering
\small{
\begin{tabular}{l|ccccc|ccccc}
 & \multicolumn{5}{c|}{Initial Memorisation} & \multicolumn{5}{c}{Stable Memorisation} \\
Model & $l = 2$ & $l = 4$ & $l = 7$ & $l = 13$ & $l = 26$
      & $l = 2$ & $l = 4$ & $l = 7$ & $l = 13$ & $l = 26$ \\
\hline
Pythia-1B & 0.031 & 0.009 & 0.026 & 0.028 & 0.243
          & 0.230 & 0.206 & 0.434 & 0.336 & 0.657 \\
Phi-2.7B & -0.001 & 0.018 & 0.014 & 0.003 & 0.108
          & 0.187 & 0.132 & 0.268 & 0.236 & 0.572 \\
Llama2-13B & -0.058 & 0.002 & 0.066 & 0.029 & 0.076
          & 0.144 & 0.124 & 0.309 & 0.102 & 0.157 \\
\end{tabular}
}
\vspace{0.5em}
\caption{\capthead{Spearman rank correlation between token position in the string and the epoch at which tokens are memorised, for \emph{untrained models}.}{$n = 1024$}
Correlation is low for initial memorisation and low to medium for stable memorisation.
}
\label{tab:rank_correlation_untrained}
\end{table}

We report results for Spearman rank correlation for pretrained Pythia-1B, Phi-2.7B and Llama2-13B models, for different alphabet sizes in Table~\ref{tab:rank_correlation}.
Across the board we observe a very low correlation between a token’s position in the string and the epoch at which it is memorised (both for initial and stable memorisation).
Correlation values in almost all cases are between -0.1 and 0.1.

We also report results for untrained models in Table~\ref{tab:rank_correlation_untrained}.
Rank correlation between token position and initial memorisation epoch is similarly low as for pretrained models.
For stable memorisation it is also low in most cases, but reaching medium correlation values for larger $\ell$, for Pythia-1B and Phi-2.7B models.
Overall, the results suggest that memorisation order is mostly random, and does not depend on the position of a token in the string.

\subsection{Quantifying randomness in memorisation order using the discrepancy score}

In order to quantify whether the memorised positions are uniformly distributed we compute the \emph{discrepancy score}; this score is motivated by the notion of discrepancy in statistics~\citep{niederreiter1992random} and is defined as follows: for any epoch, we count the number of correct recollections of our model on the \emph{input string} and then we pick uniformly at random the same number of positions on 
\emph{random string} of the same length.
Utilizing a fixed window of 20 tokens, we randomly sample $50$ substrings from each of the two strings. For each of these sampled substrings, we calculate difference in the number of correct recollections between the tested and target substrings.
The average of these differences provides the \emph{discrepancy score}.

Formally, the discrepancy score is defined as follows:
\[
discrepancy(M, s) = \frac{1}{|PS|} \sum_{i \in PS} \Delta_{i, k}(M, s, c^{(rand)})
\]
where
\[
    \Delta_{i, k}(M, s, c^{(rand)}) = \frac{1}{k} \sum_{j = 1}^{k} \delta \left( pred(M | s_{[1, i + j - 1]}) = s_{i + j}) - c_{i + j}^{(rand)} \right)
\]
and $PS$ is a set of randomly selected positions in the string $s$, $c^{(rand)}$ is a binary string where $N_{correct} = \sum_i^n \delta(pred(M | s_{[1, i - 1]}) = s_i)$ (\ie~as many positions as there are correct predictions by $M$) randomly chosen positions are $1$, and else $0$.
$pred(M | s_{[1, i - 1]}) = \argmax_{t \in V} P_M(t | s_{[1, i - 1]})$ is the model's prediction at position $i$.
In our analysis, we sample $|PS| = 50$ positions in the string $s$ and use a window size of $k = 20$.

\textbf{Intuition on discrepancy}:
The intuition behind the discrepancy score is to measure how different the set of positions correctly predicted by the model is from randomly picking the same number positions as there are correct predictions.
If the difference is close to zero, then selecting the same number of string positions based on whether they are correctly predicted or randomly picked is equivalent.

\begin{figure}[H]
    \centering
    \subfloat[Pythia-1B]{
        \includegraphics[width=\thirdWidth]{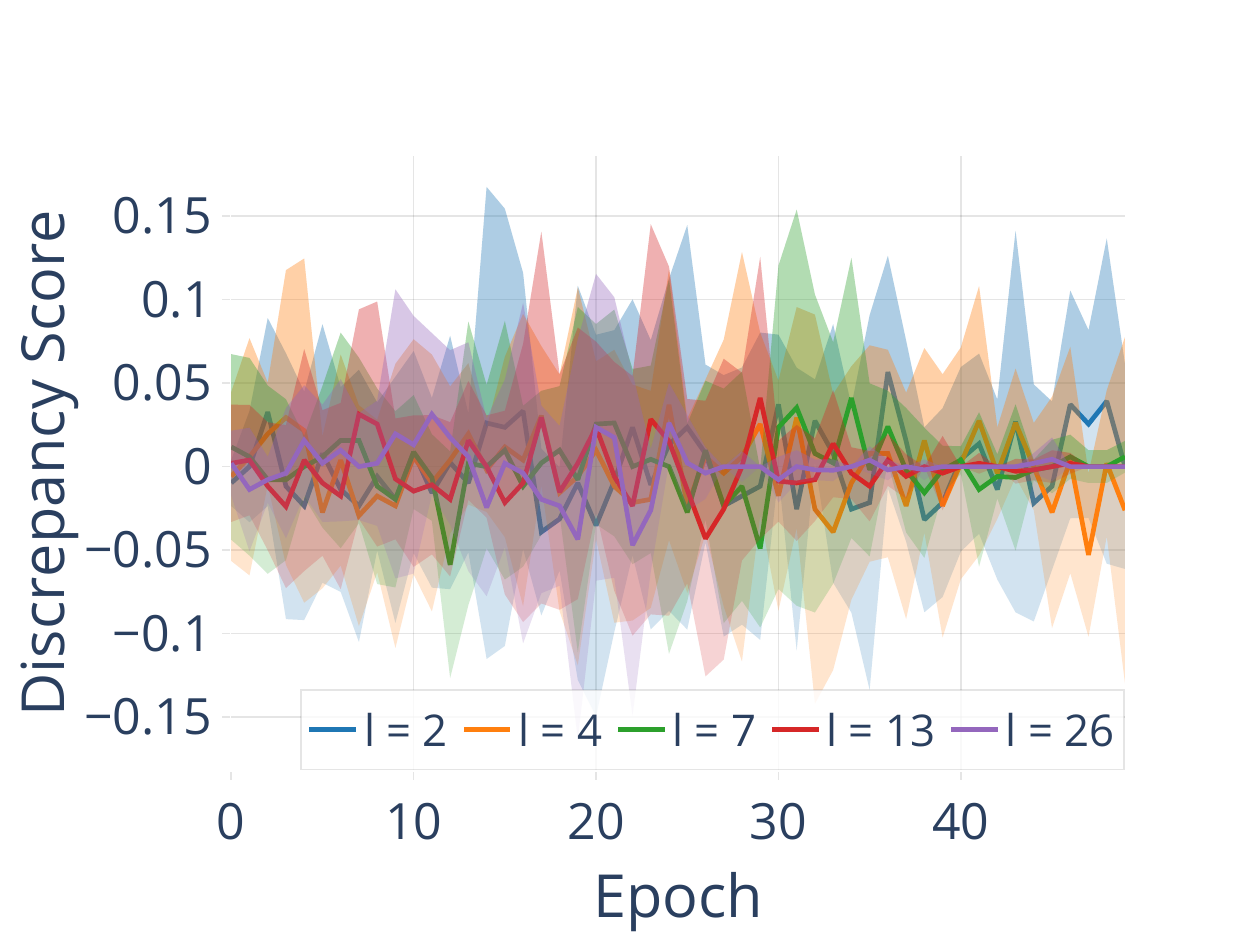}
    }
    \subfloat[Phi-2.7B]{
        \includegraphics[width=\thirdWidth]{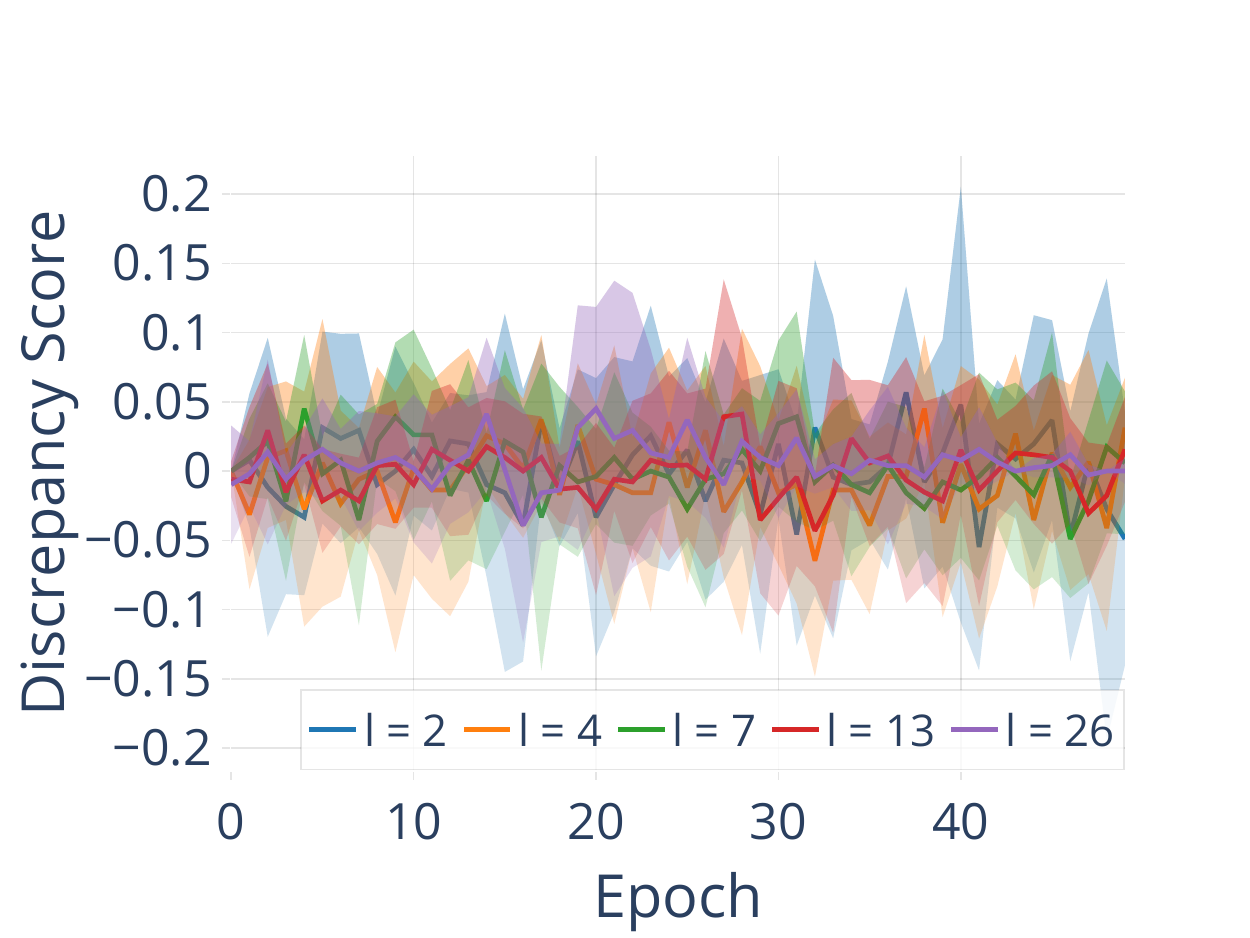}
    }
    \subfloat[Llama2-13B]{
        \includegraphics[width=\thirdWidth]{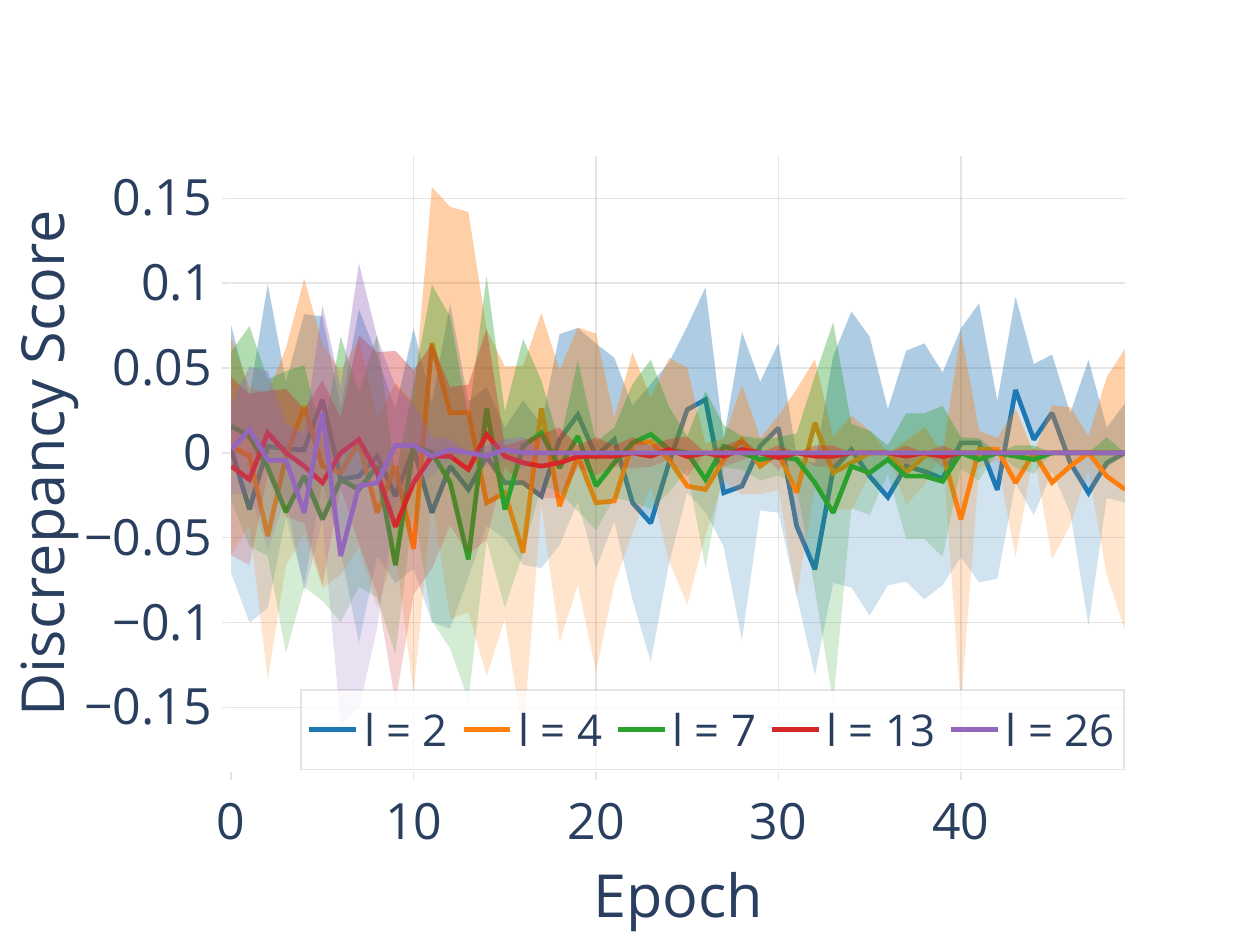}
    }
\caption{\capthead{Discrepancy scores for different pretrained models.}{$n = 1024$}
Discrepancy scores for all models and strings are low, indicating random memorisation order.
}
\label{fig:discrepancy_all}
\end{figure}

\begin{figure}[H]
    \centering
    \subfloat[Pythia-1B]{
        \includegraphics[width=\thirdWidth]{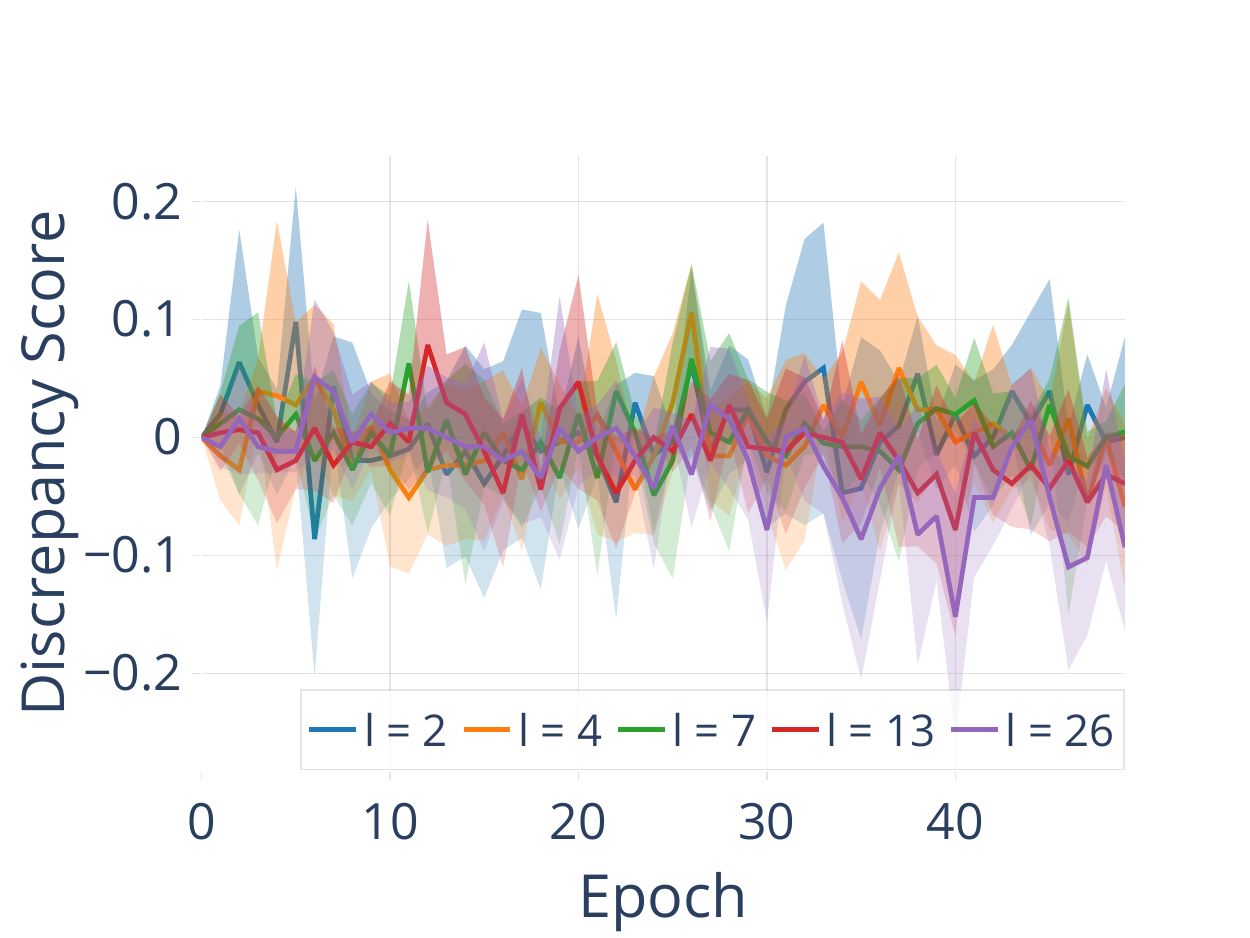}
    }
    \subfloat[Phi-2.7B]{
        \includegraphics[width=\thirdWidth]{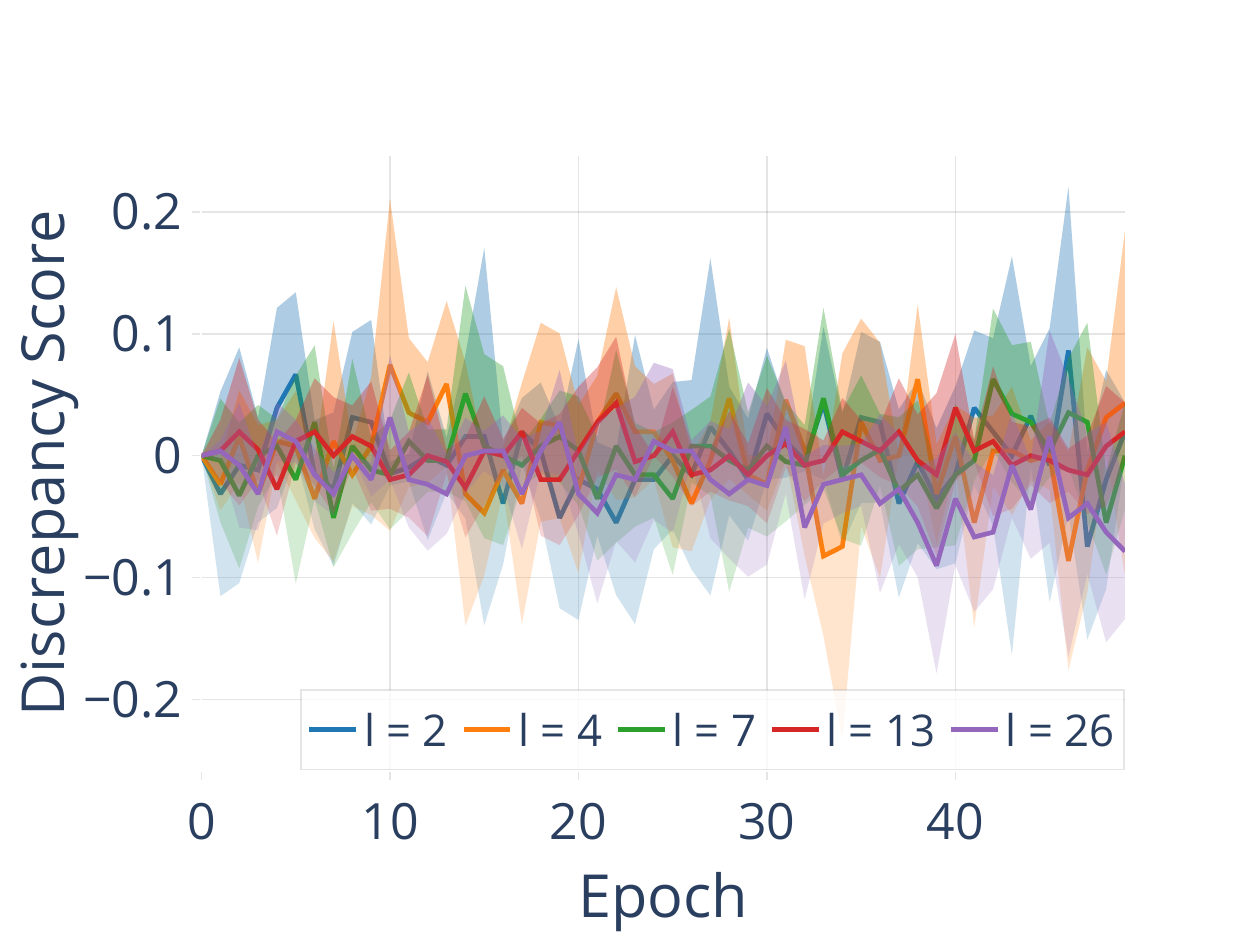}
    }
    \subfloat[Llama2-13B]{
        \includegraphics[width=\thirdWidth]{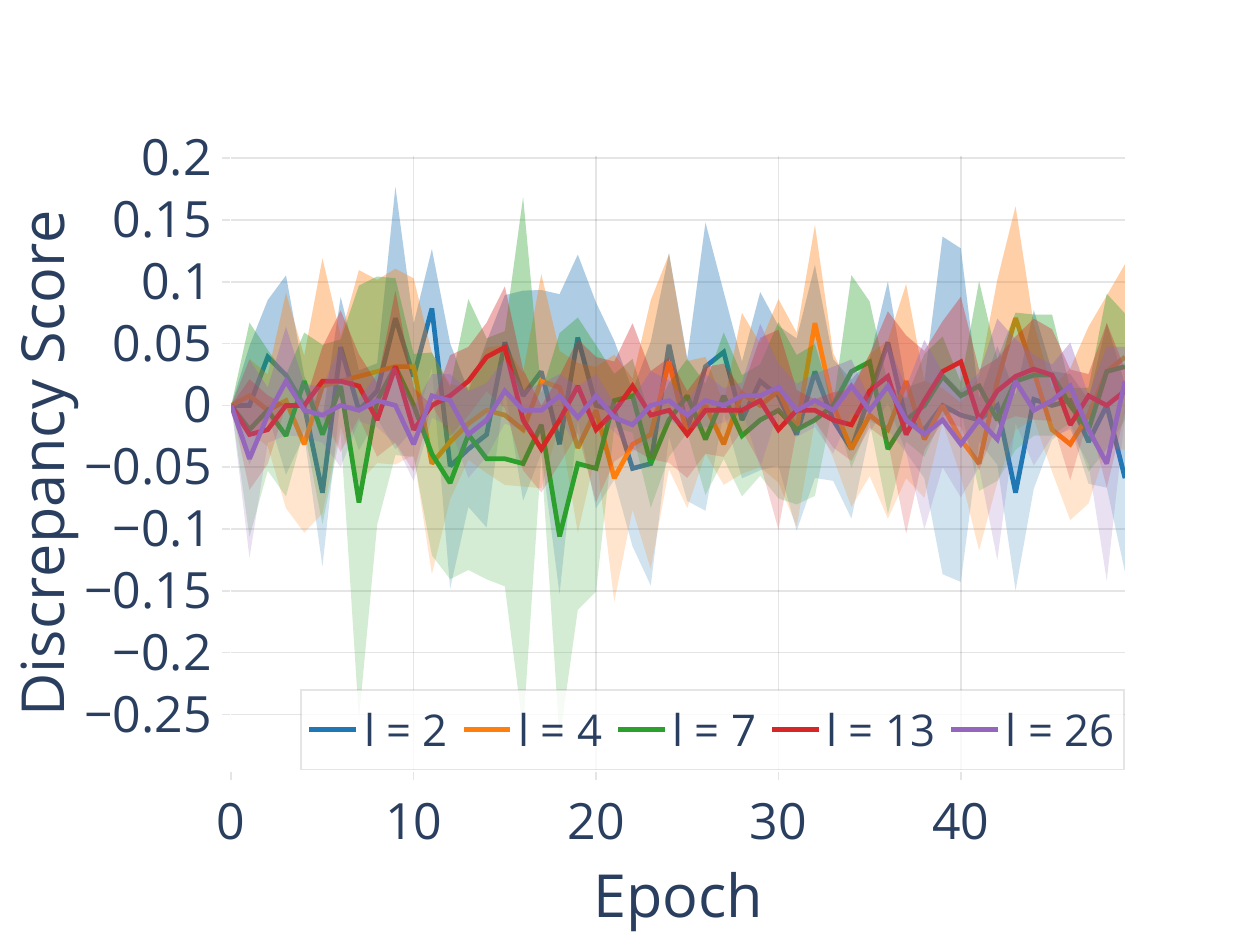}
    }
\caption{\capthead{Discrepancy scores for different untrained models.}{$n = 1024$}
Discrepancy scores for all models and strings are low, indicating random memorisation order.
}
\label{fig:discrepancy_all_untrained}
\end{figure}

\textbf{Results:}
The low discrepancy scores for pretrained models observed in Figure \ref{fig:discrepancy_all} suggests that for all the models evaluated, the memorised positions are random. 
Thus, we conclude that memorisation happens at the granularity of individual tokens and not entire strings.
The discrepancy scores for untrained models in Figure \ref{fig:discrepancy_all_untrained} are also low, though slightly higher than for pretrained models, indicating that memorisation order is largely random for untrained models as well.

\section{Memorisation dynamics under real-world conditions}
\label{app:real_world_validation}

To validate our observations in practical settings, we train models to memorise random strings under conditions that closely resemble real-world settings.
We present random strings to the model in the context of natural language data, in two different ways: 1) by adding additional natural language sequences to the training batches, and 2) by presenting single natural language sequences inside which random strings appear as substrings.
We use the wikitext~\cite{merity2016pointer} dataset as a source of natural language training data.

In both of the cases, the random string to natural data ratio varies.
To minimise its loss, the model still has to memorise the random string, but it simultaneously also needs to become better at modeling the wikitext data.
We study both pretrained and untrained (to mimic pretraining from scratch) models, over alphabet sizes $l = 2, 7, 26$.

\textbf{Results summary:}
We investigate the memorisation dynamics of random strings as part of larger natural language training batches in Appendix~\ref{app:rw_val_batches}, and of random strings as substrings of larger natural language strings in Appendix~\ref{app:rw_val_strings}.
We make the same observations about memorisation dynamics as with single strings.
When memorising random strings in the context of other natural data, models still exhibit the two phases during the memorisation process, and lower entropy strings are also harder to memorise.
Memorisation becomes slower, however, the more natural data there is, relative to the size of the random string.
Overall, our results on the dynamics of memorisation are robust under more practical training schemes, since the observations made on random strings in isolation match those made when random strings are embedded in natural language data.

\subsection{Embedding random strings within batches of natural language}
\label{app:rw_val_batches}

To mimic typical pretraining setups, we train models on batches of size up to $BS = 1, 4, 16, 64$, with sequence length $n = 1024$.
The same $n = 1024$ token random string appears repeatedly, once in each batch, \ie~as one of the batch elements.
Batch size $BS = 1$ is equivalent to our previous experiments.
The other sequences in each batch come from wikitext and change at every step without repetition.

\begin{figure}[H]
    \centering
    \subfloat[Pythia-1B, $\ell = 2$]{
        \includegraphics[width=\smallThirdWidth]{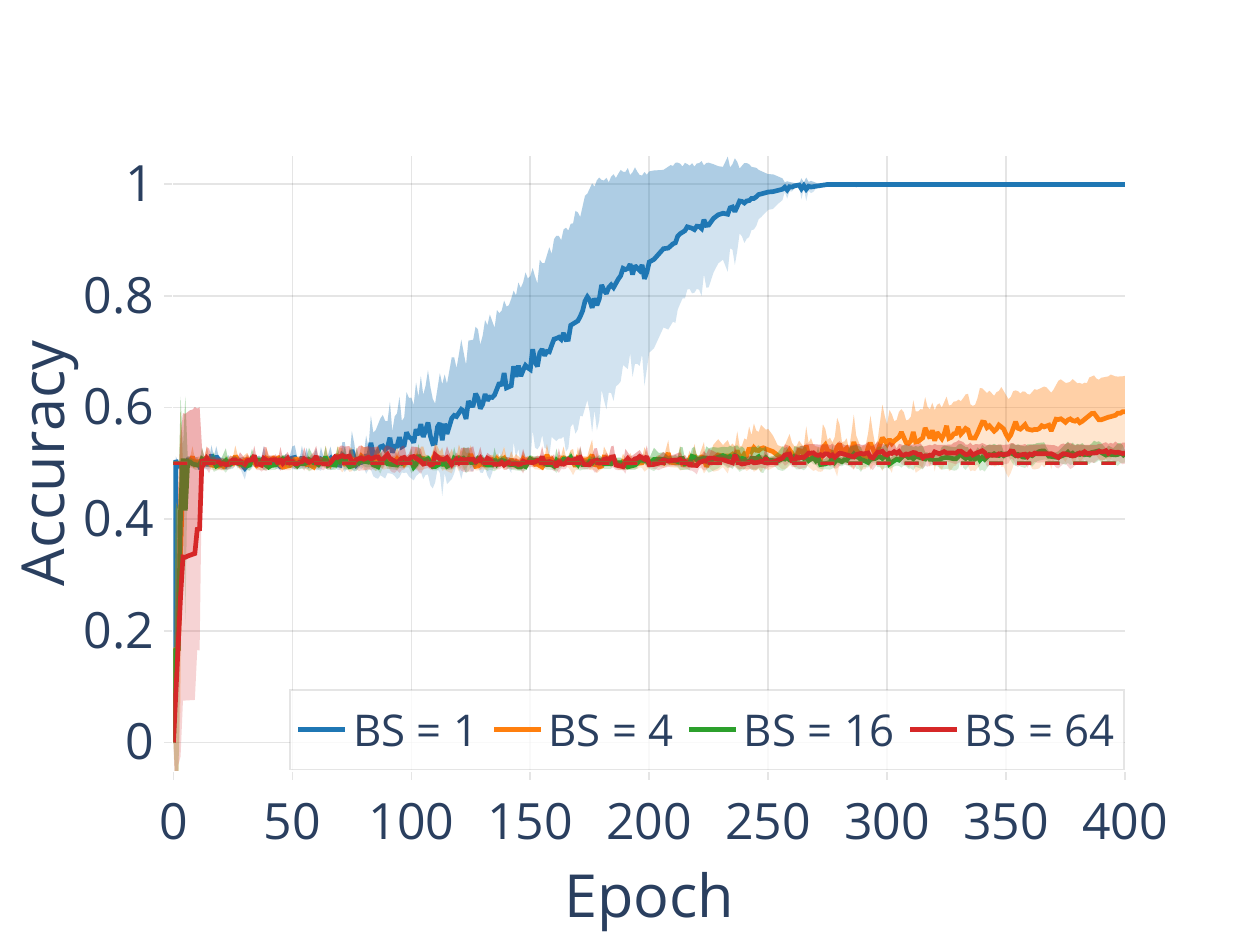}
    }
    \subfloat[Pythia-1B, $\ell = 7$]{
        \includegraphics[width=\smallThirdWidth]{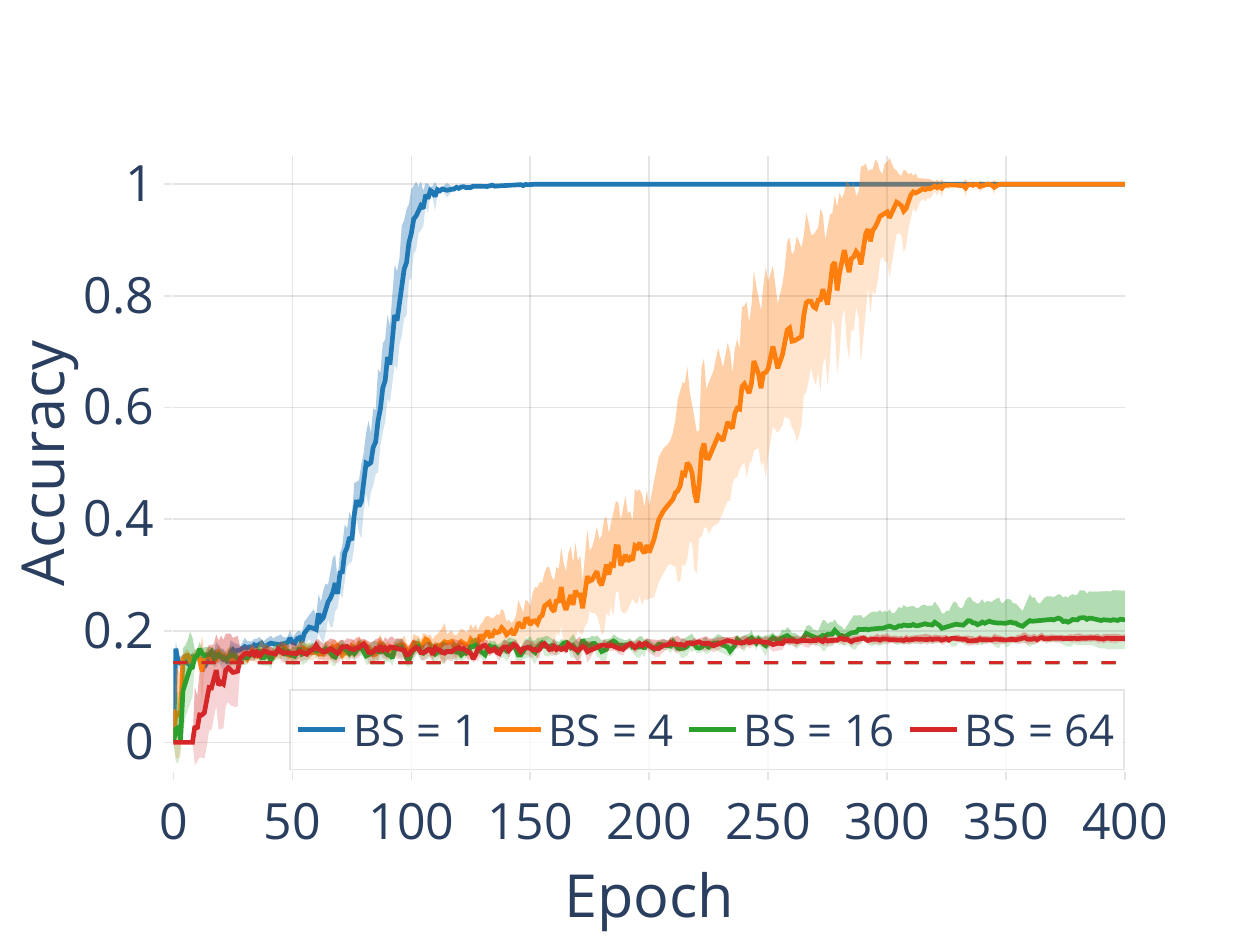}
    }
    \subfloat[Pythia-1B, $\ell = 26$]{
        \includegraphics[width=\smallThirdWidth]{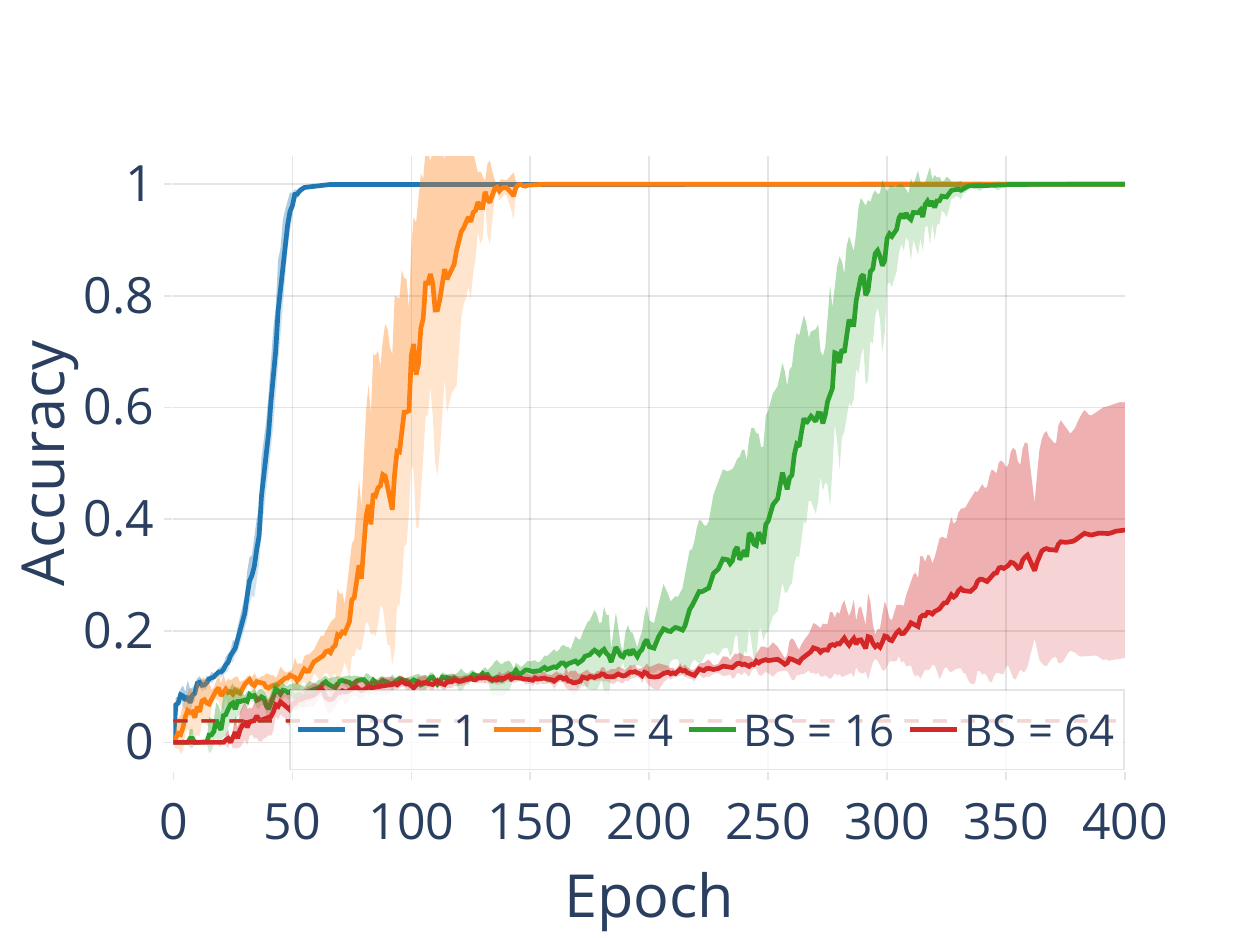}
    }
    \\
    \subfloat[Phi-2.7B, $\ell = 2$]{
        \includegraphics[width=\smallThirdWidth]{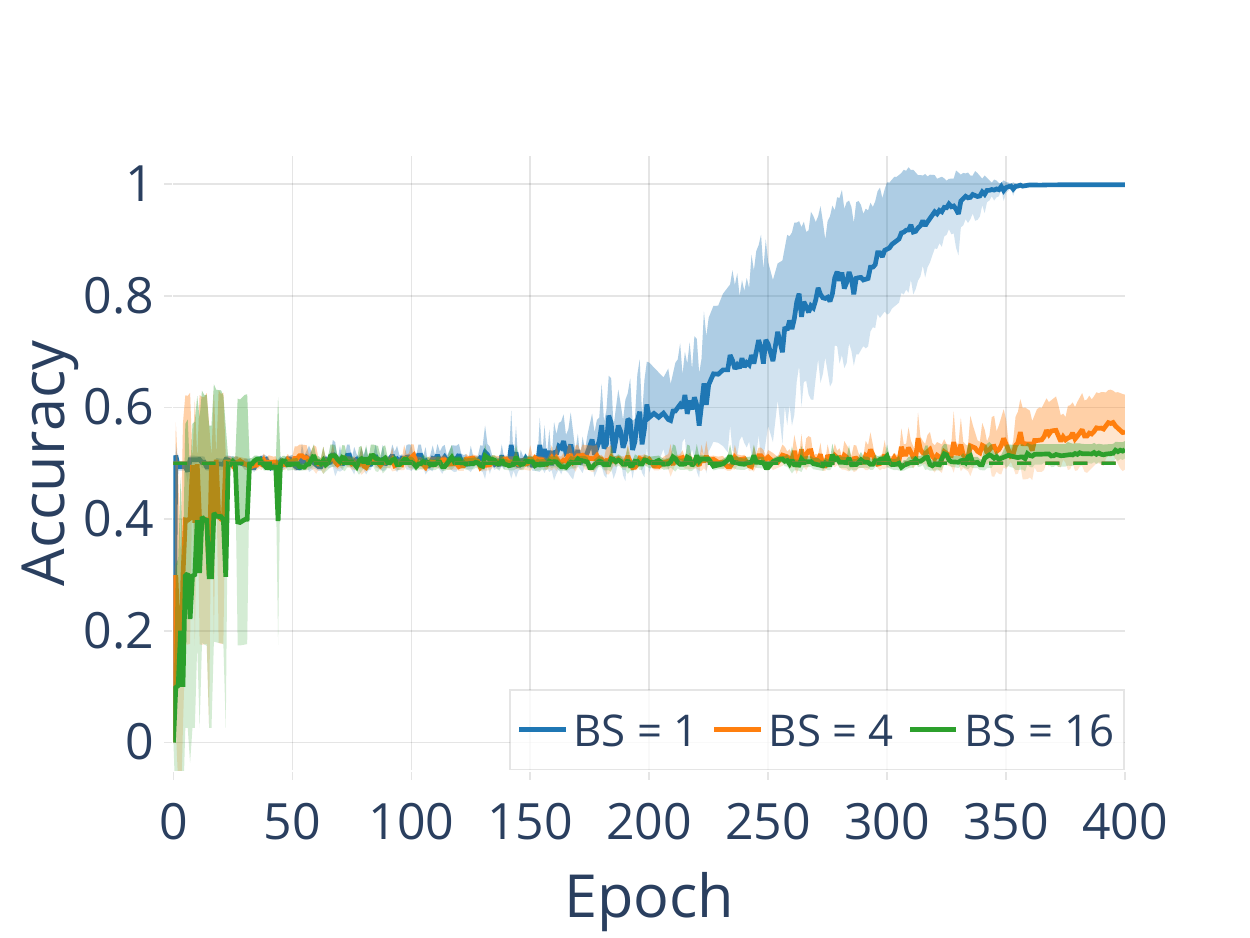}
    }
    \subfloat[Phi-2.7B, $\ell = 7$]{
        \includegraphics[width=\smallThirdWidth]{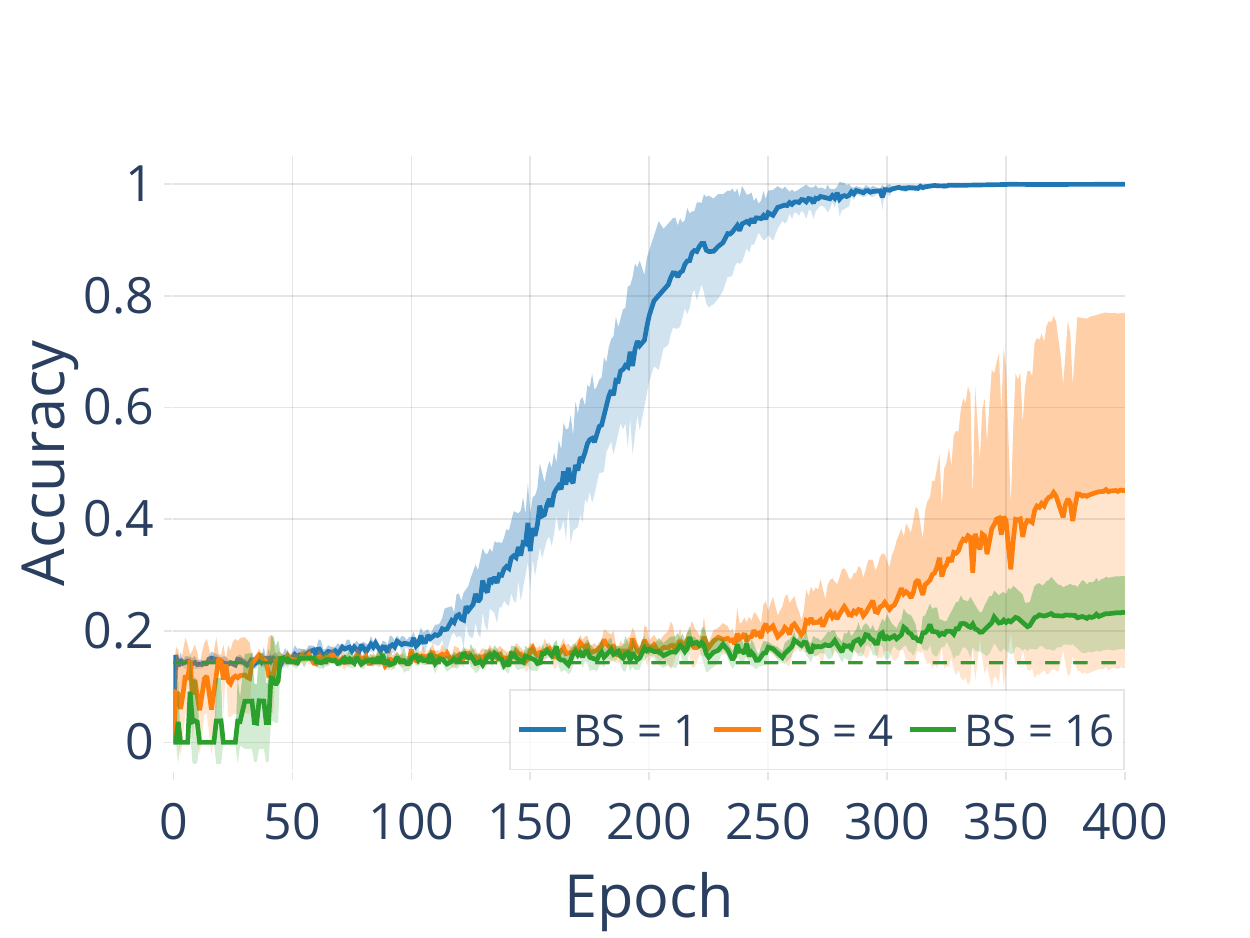}
    }
    \subfloat[Phi-2.7B, $\ell = 26$]{
        \includegraphics[width=\smallThirdWidth]{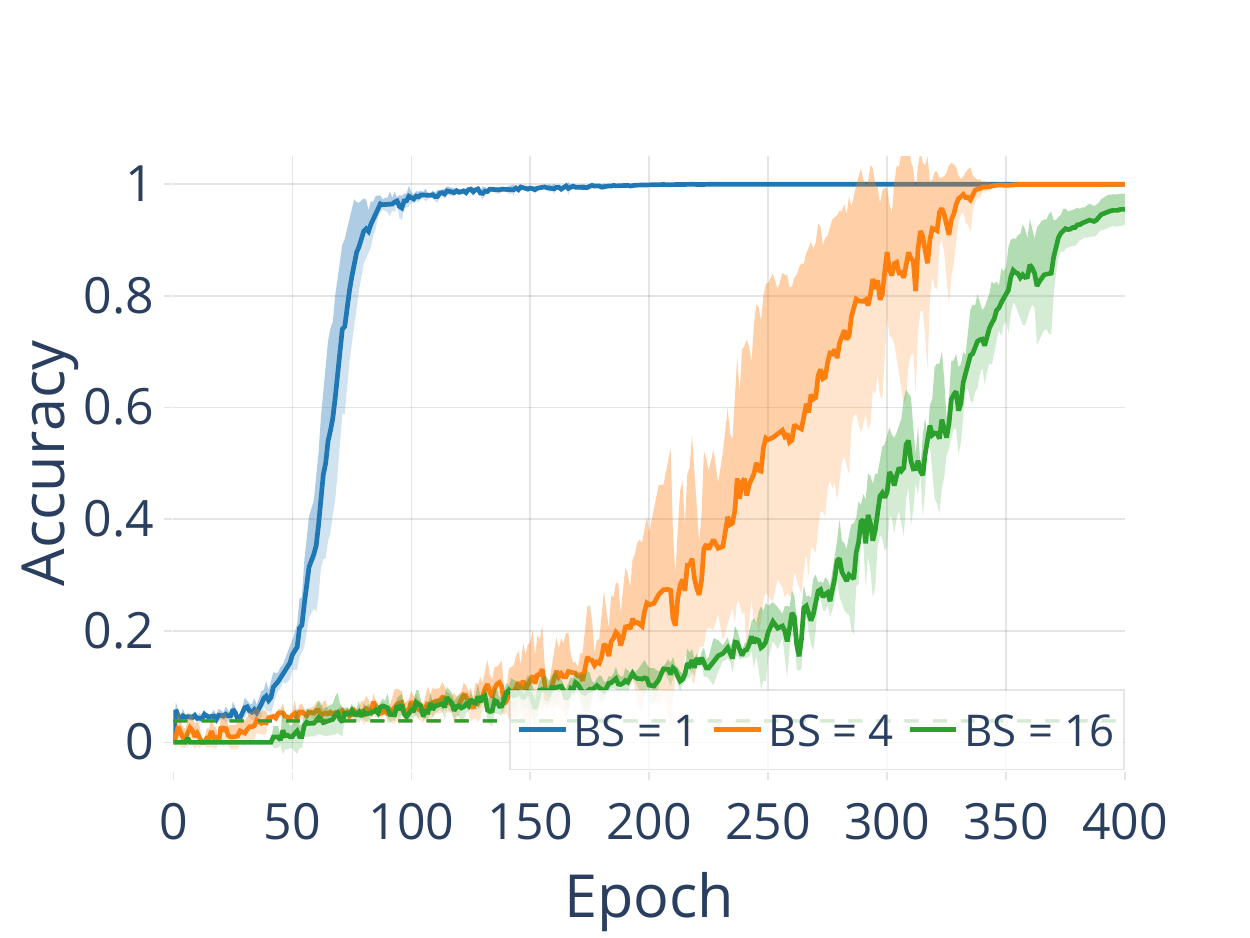}
    }
    \\
    \subfloat[Llama2-13B, $\ell = 2$]{
        \includegraphics[width=\smallThirdWidth]{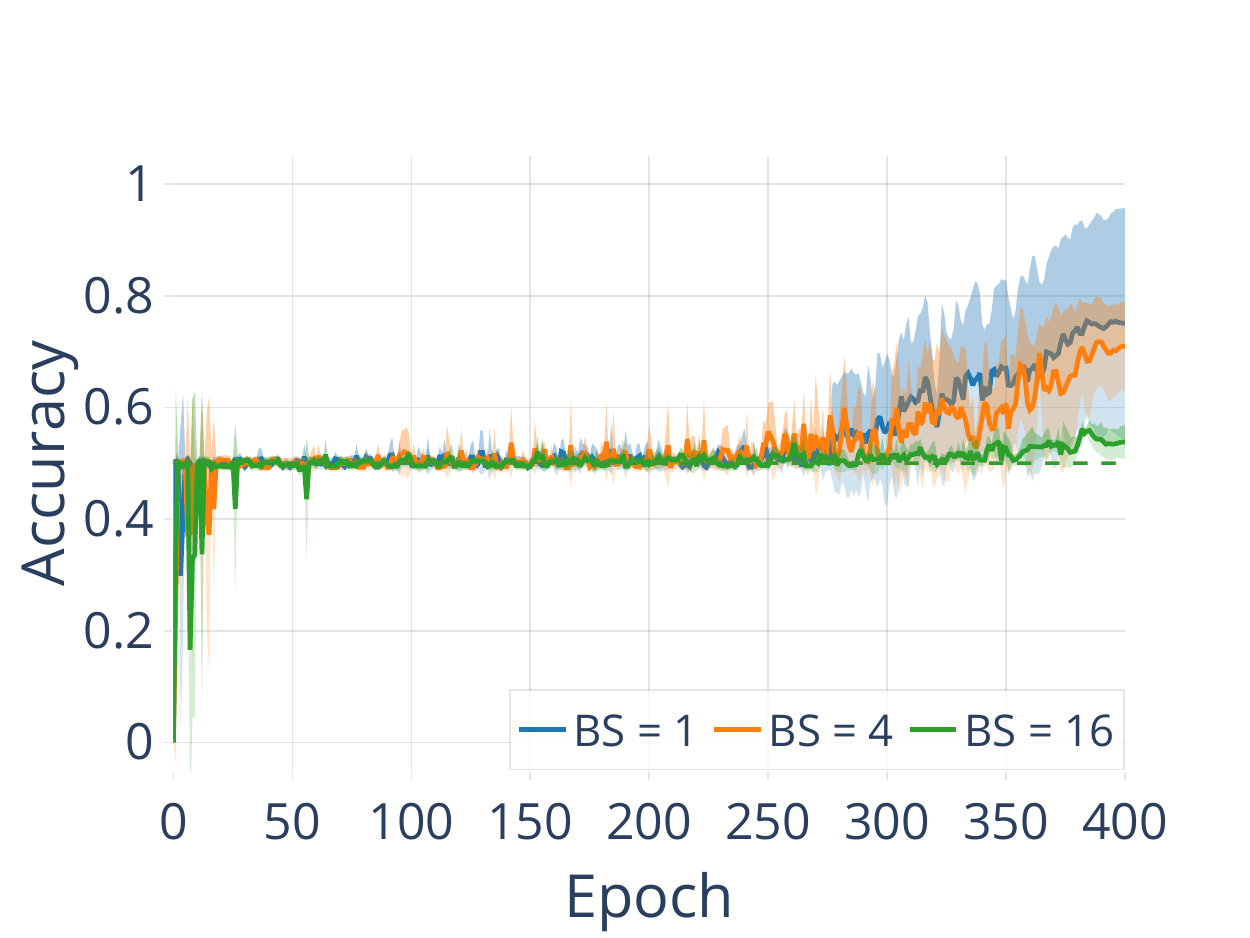}
    }
    \subfloat[Llama2-13B, $\ell = 7$]{
        \includegraphics[width=\smallThirdWidth]{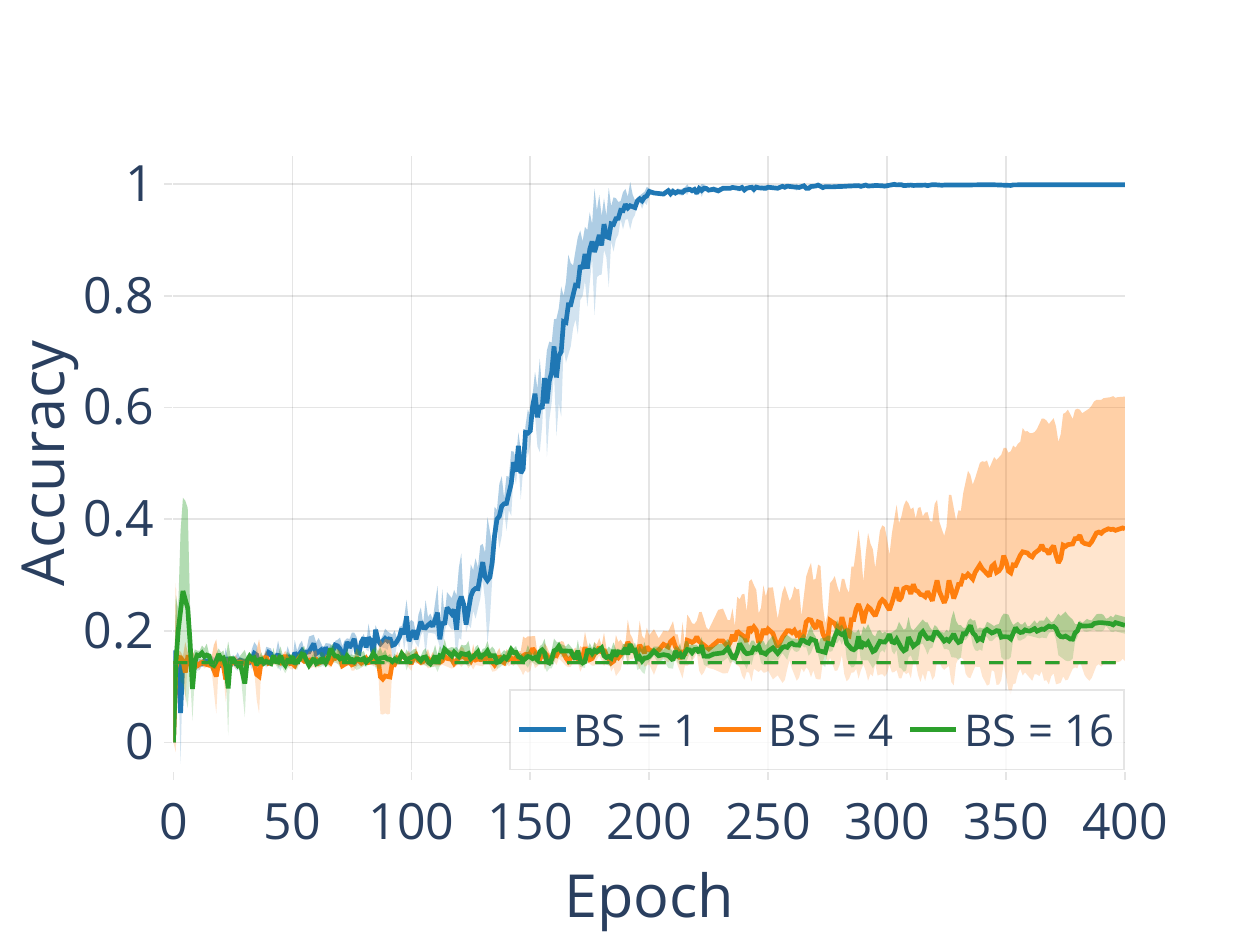}
    }
    \subfloat[Llama2-13B, $\ell = 26$]{
        \includegraphics[width=\smallThirdWidth]{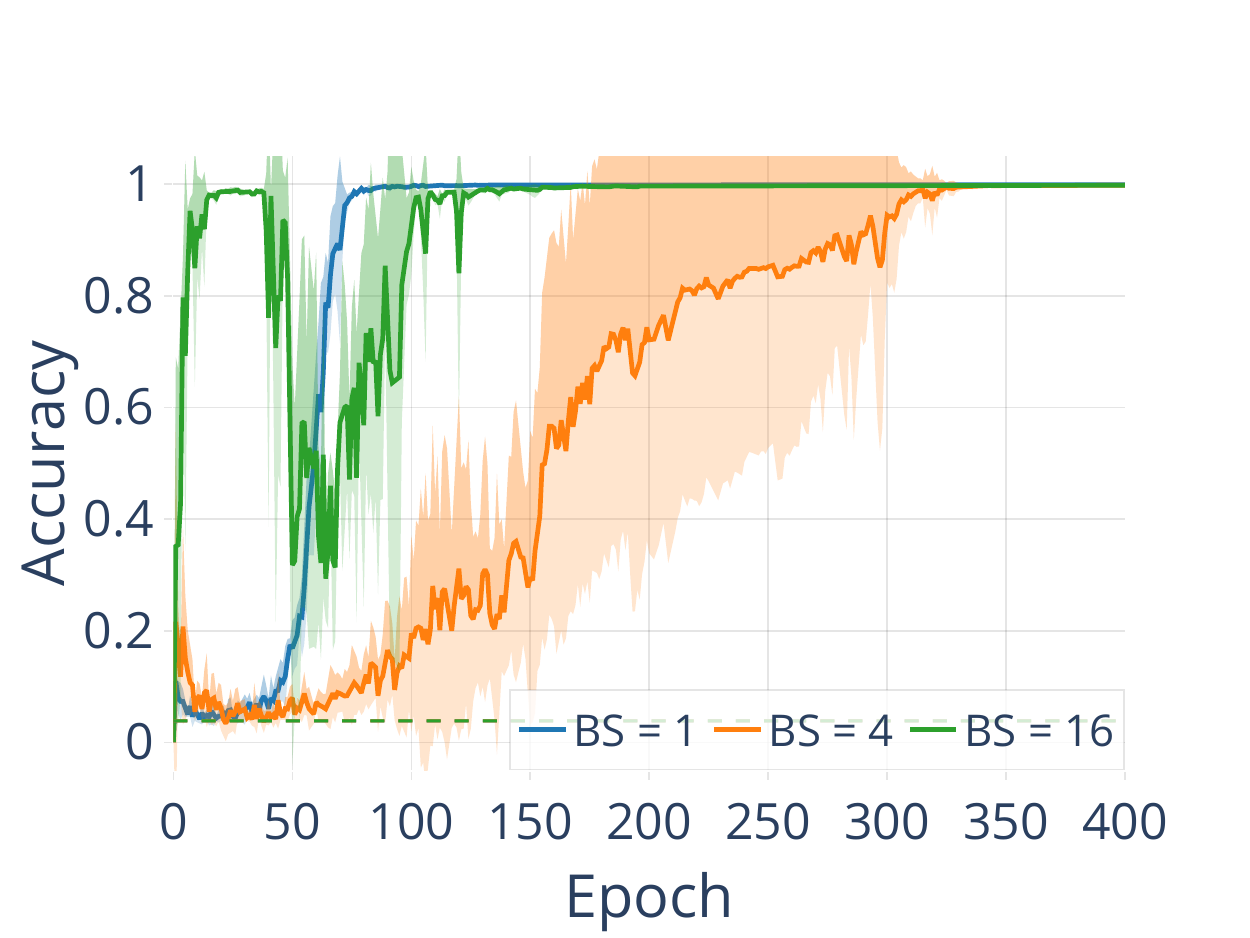}
    }
\caption{\capthead{Accuracy for \emph{untrained} models, when embedding random strings inside of batches of natural language data, for multiple $\ell$ and batch sizes $b$.}{$n = 1024$}
}
\label{fig:rw_val_bs_accuracy_untrained}
\end{figure}

\begin{figure}[H]
    \centering
    \subfloat[Pythia-1B, $\ell = 2$]{
        \includegraphics[width=\smallThirdWidth]{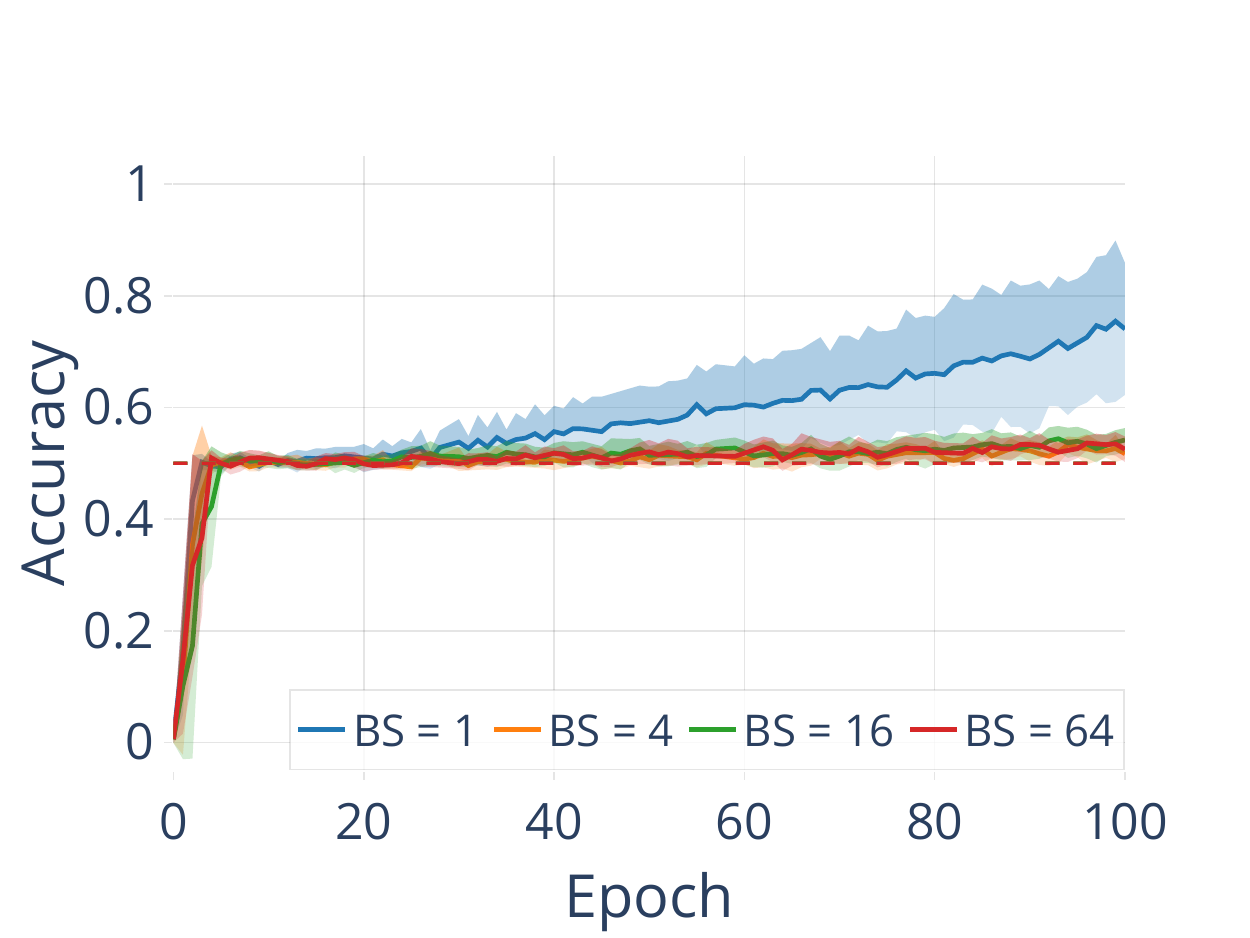}
    }
    \subfloat[Pythia-1B, $\ell = 7$]{
        \includegraphics[width=\smallThirdWidth]{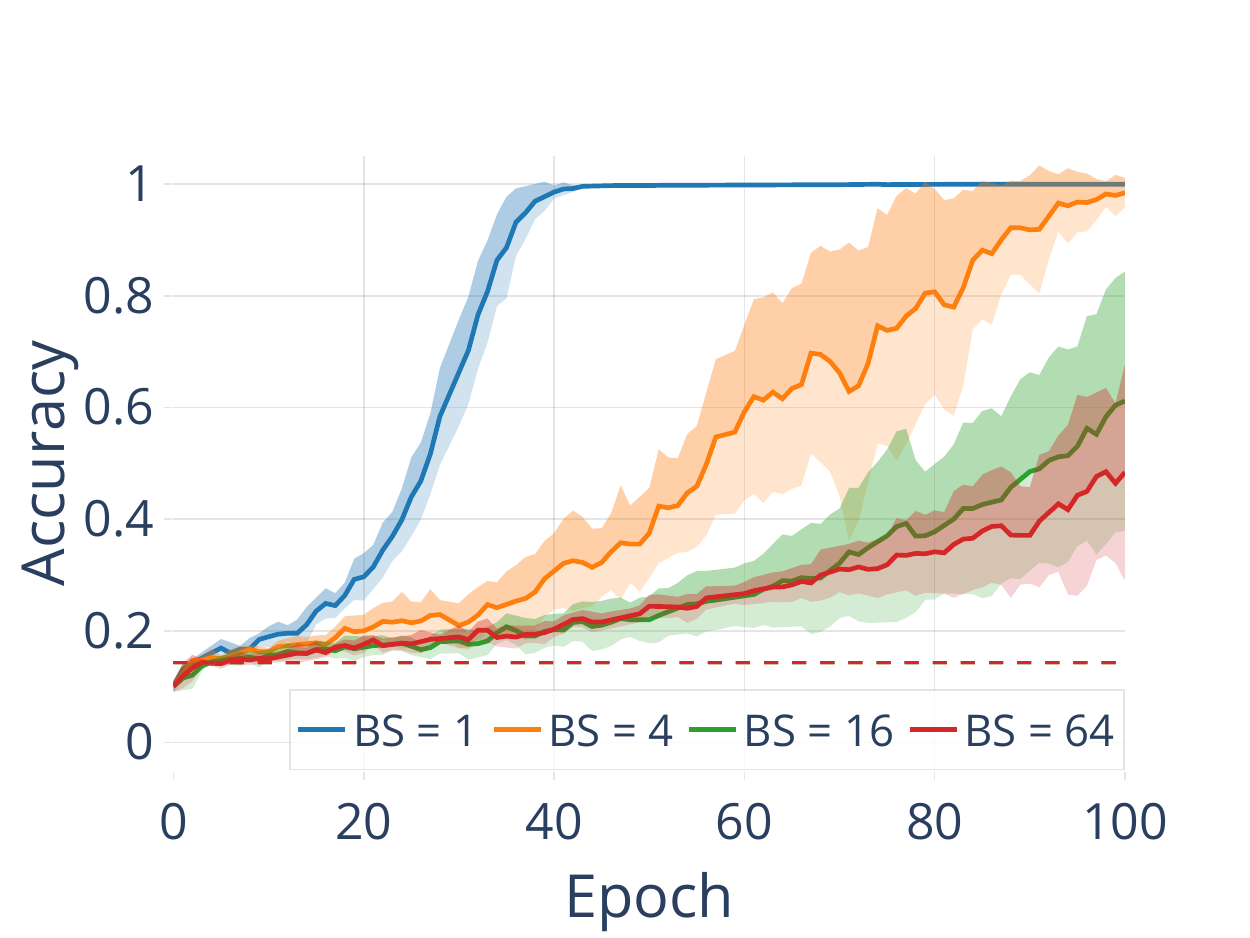}
    }
    \subfloat[Pythia-1B, $\ell = 26$]{
        \includegraphics[width=\smallThirdWidth]{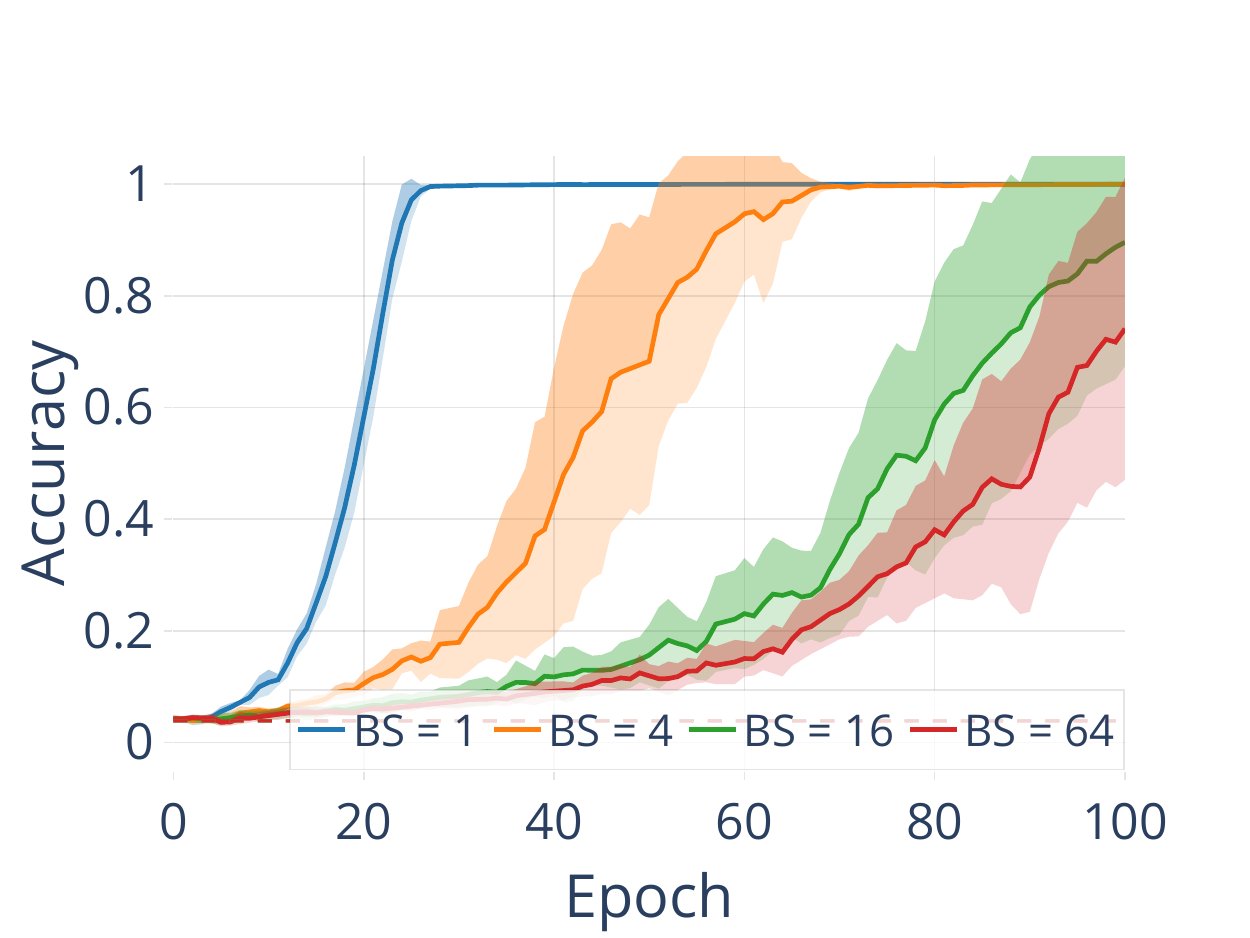}
    }
    \\
    \subfloat[Phi-2.7B, $\ell = 2$]{
        \includegraphics[width=\smallThirdWidth]{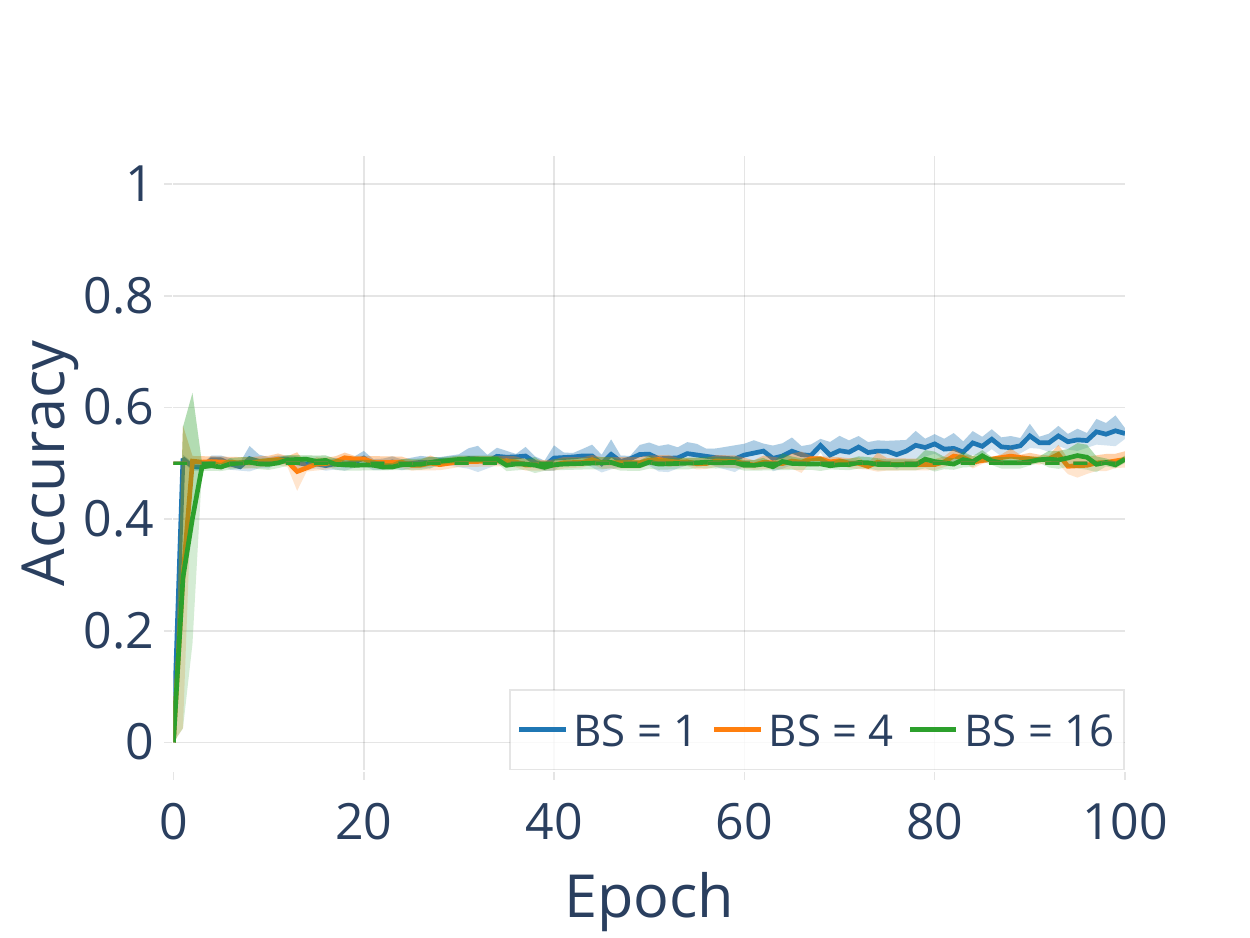}
    }
    \subfloat[Phi-2.7B, $\ell = 7$]{
        \includegraphics[width=\smallThirdWidth]{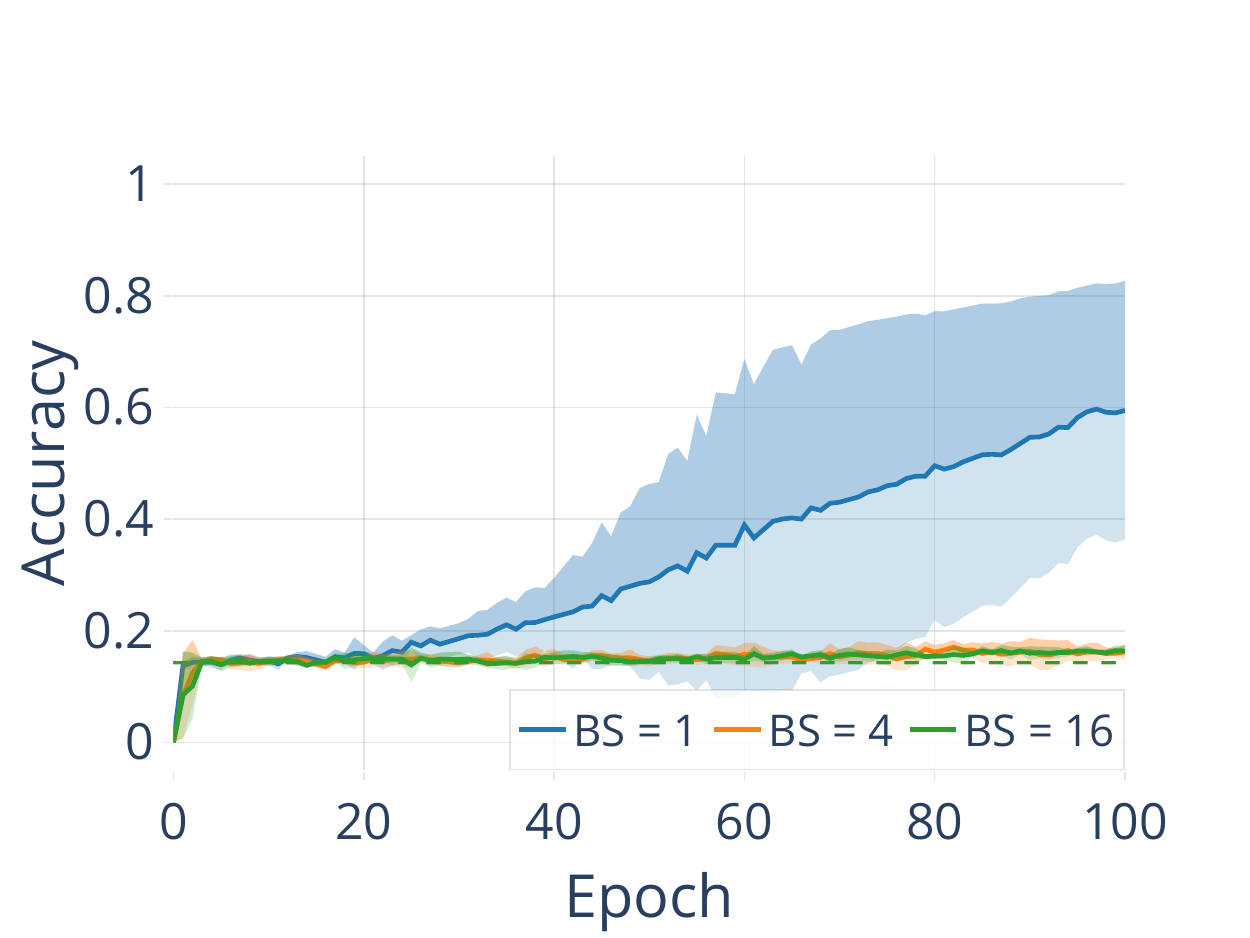}
    }
    \subfloat[Phi-2.7B, $\ell = 26$]{
        \includegraphics[width=\smallThirdWidth]{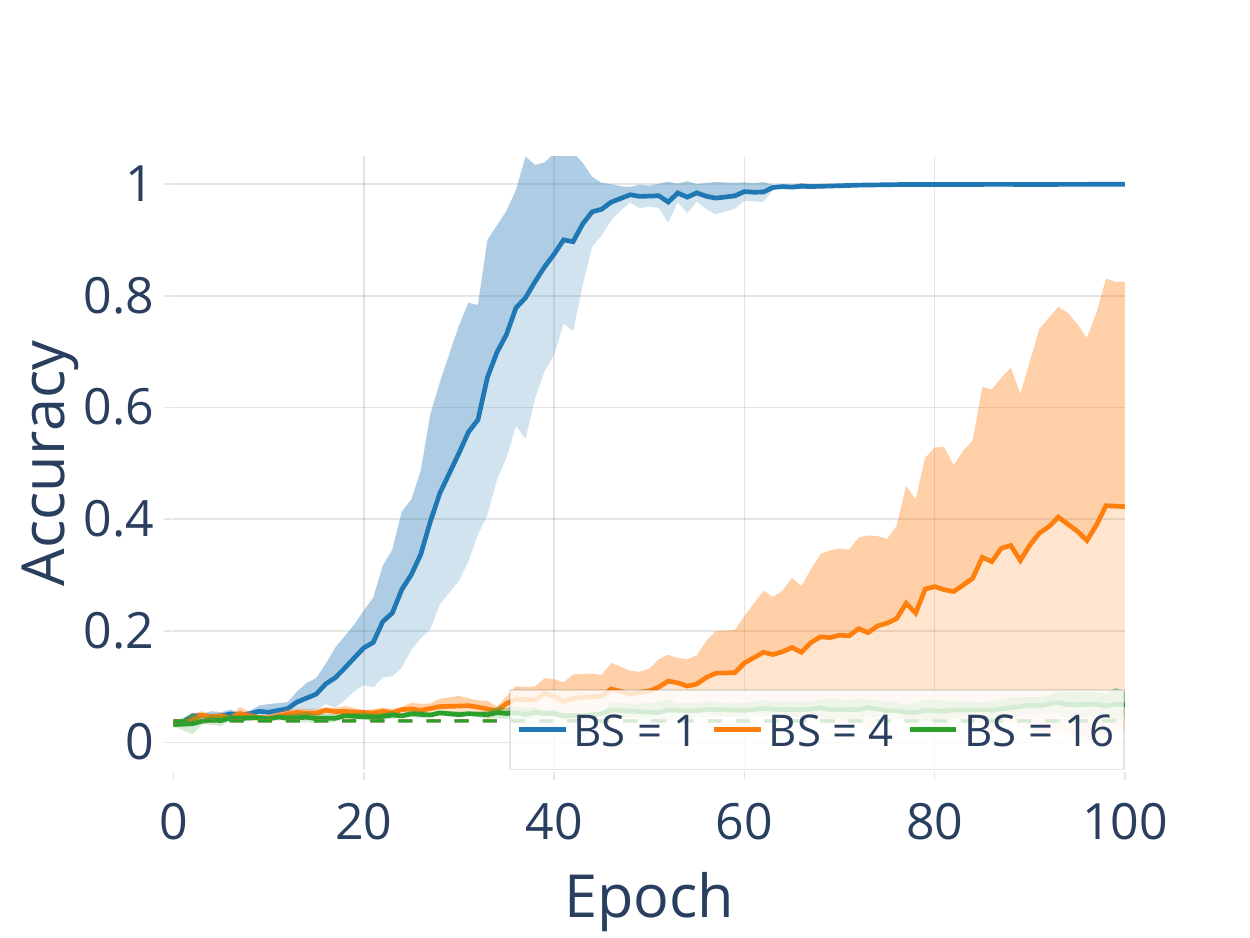}
    }
    \\
    \subfloat[Llama2-13B, $\ell = 2$]{
        \includegraphics[width=\smallThirdWidth]{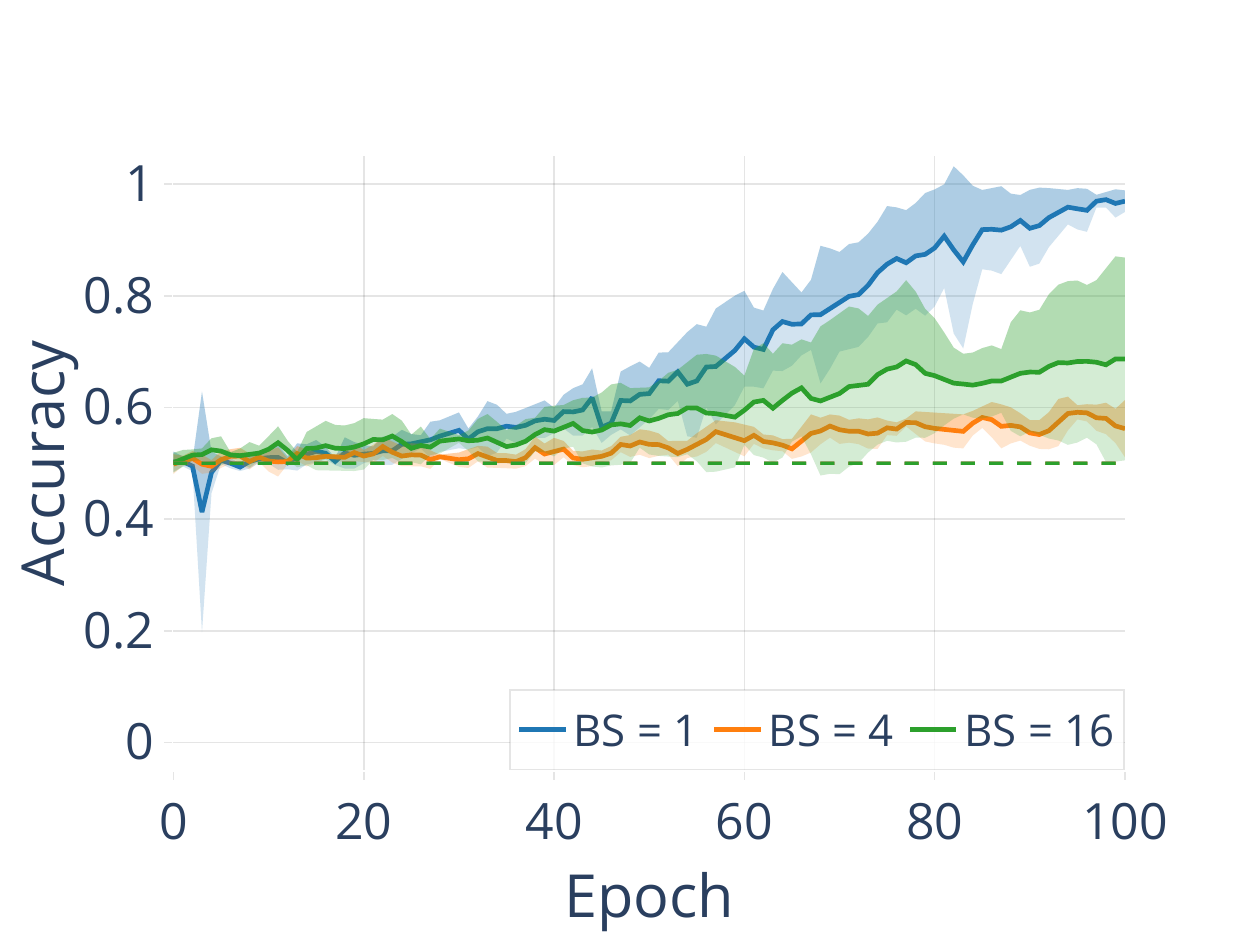}
    }
    \subfloat[Llama2-13B, $\ell = 7$]{
        \includegraphics[width=\smallThirdWidth]{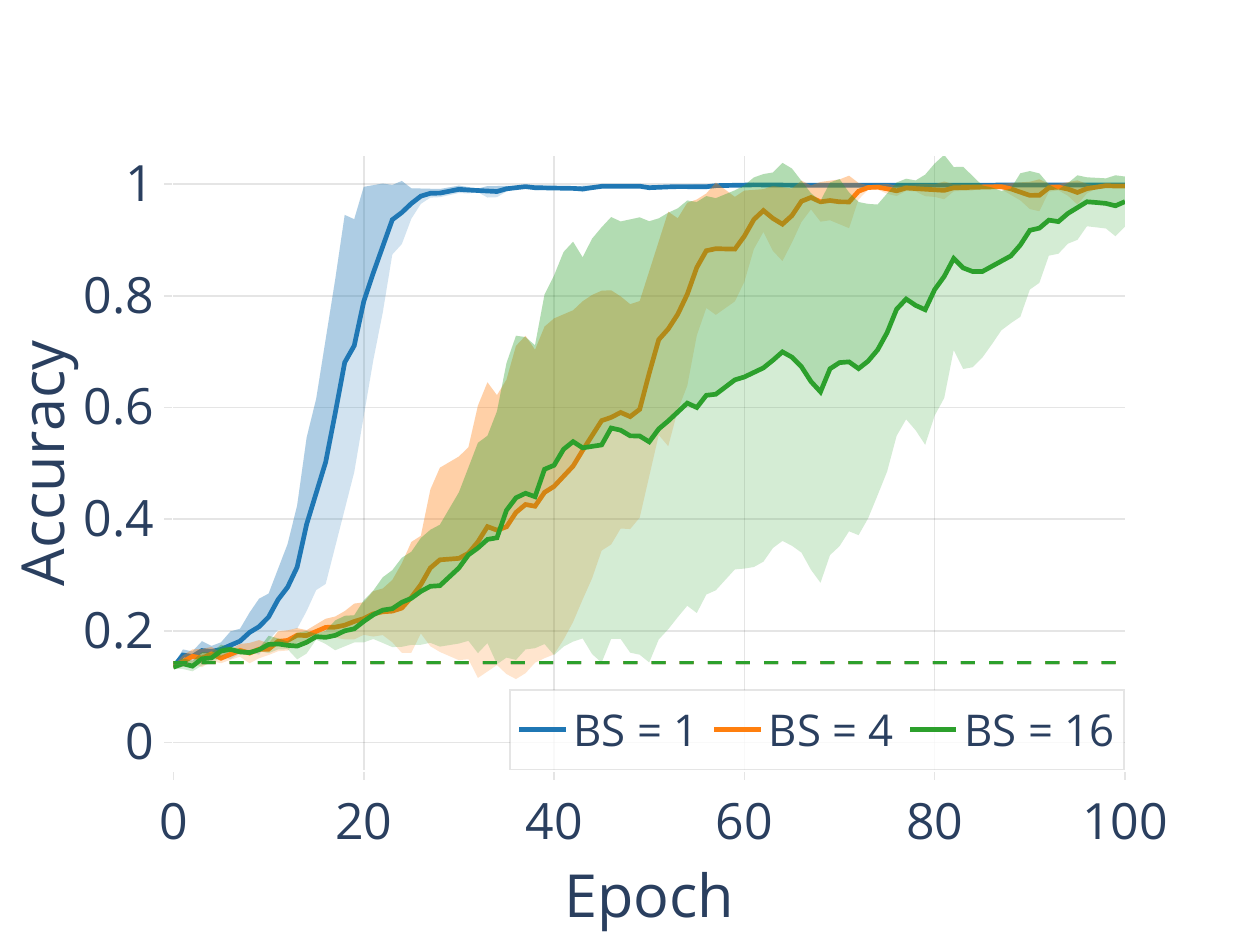}
    }
    \subfloat[Llama2-13B, $\ell = 26$]{
        \includegraphics[width=\smallThirdWidth]{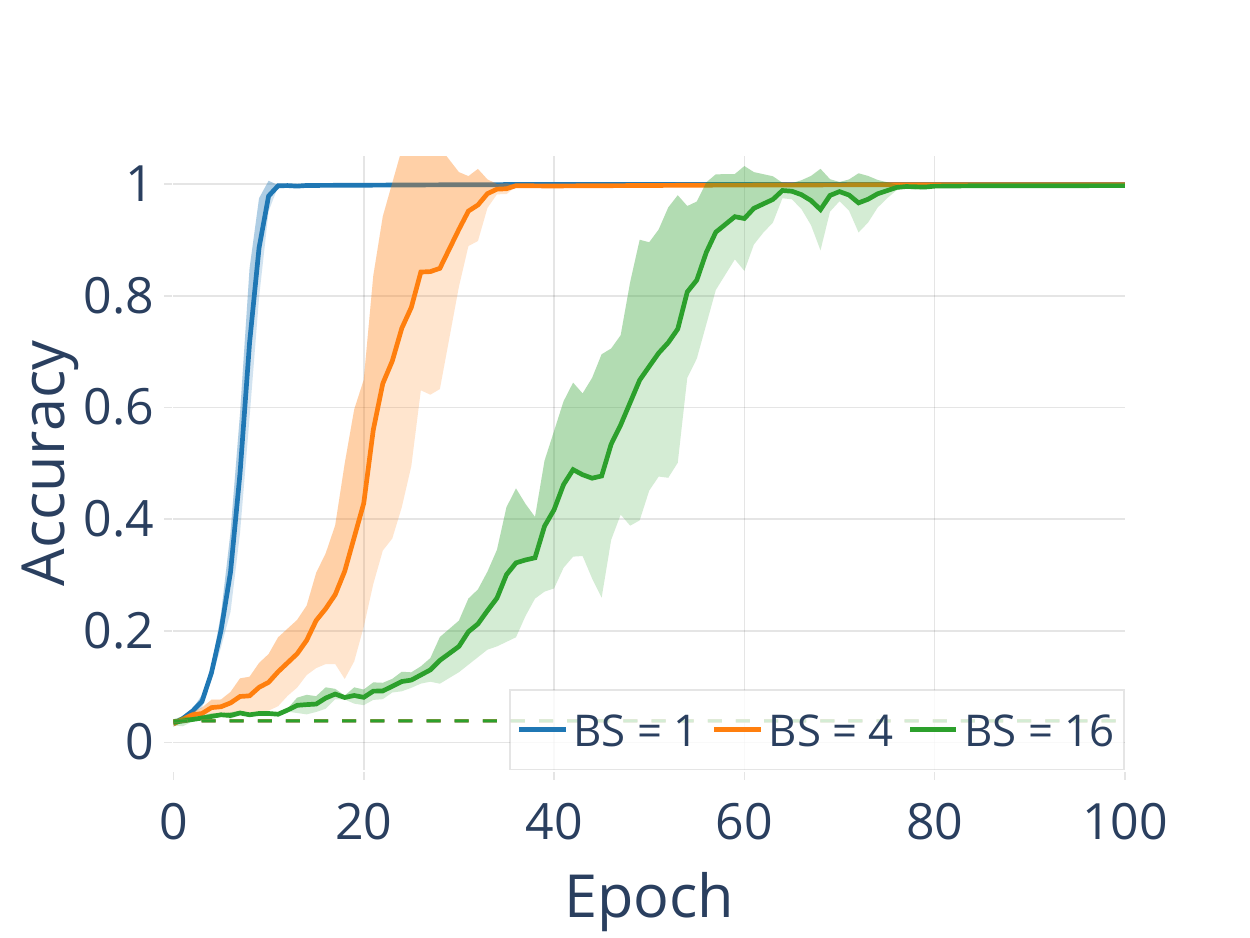}
    }
\caption{\capthead{Accuracy for \emph{pretrained} models, when embedding random strings inside of batches of natural language data, for multiple $\ell$ and batch sizes $b$.}{$n = 1024$}
}
\label{fig:rw_val_bs_accuracy_pretrained}
\end{figure}

Figure~\ref{fig:rw_val_bs_accuracy_untrained} shows the accuracy of untrained models on batched training, and Figure~\ref{fig:rw_val_bs_accuracy_pretrained} shows the accuracy of pretrained models.
In all cases, models initially converge to the random guess level during the \GuessPhase, and then start memorising the random string during the \MemPhase, at least for higher-entropy strings.
Higher entropy strings are consistently easier to memorise.
These observations match our previous results where we train on random strings in isolation.

\subsection{Embedding random strings within longer natural language strings}
\label{app:rw_val_strings}

To simulate cases where random strings appear as substrings inside other strings (e.g. as with typical sensitive data such as email addresses, phone numbers, SSH-keys, etc.), we train models on single natural language strings of lengths $CS = 256, 512, 1024, 2048$.
We use a different string from wikitext at each step, but always insert the same 256 token random substring at a random position.
For context size $CS = 256$ the entire string is the random string, as in our previous experiments.

\begin{figure}[H]
    \centering
    \subfloat[Pythia-1B, $\ell = 2$]{
        \includegraphics[width=\smallThirdWidth]{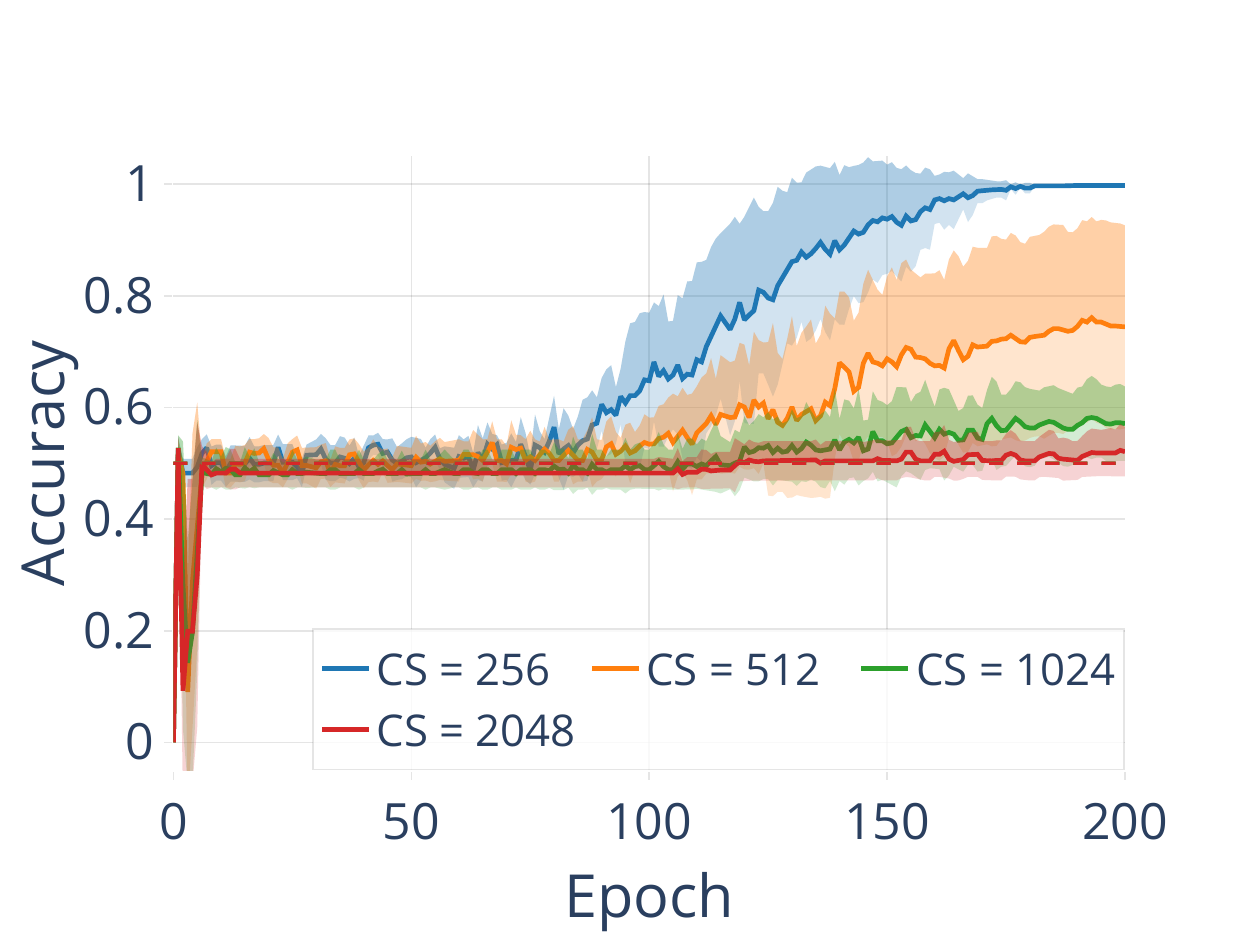}
    }
    \subfloat[Pythia-1B, $\ell = 7$]{
        \includegraphics[width=\smallThirdWidth]{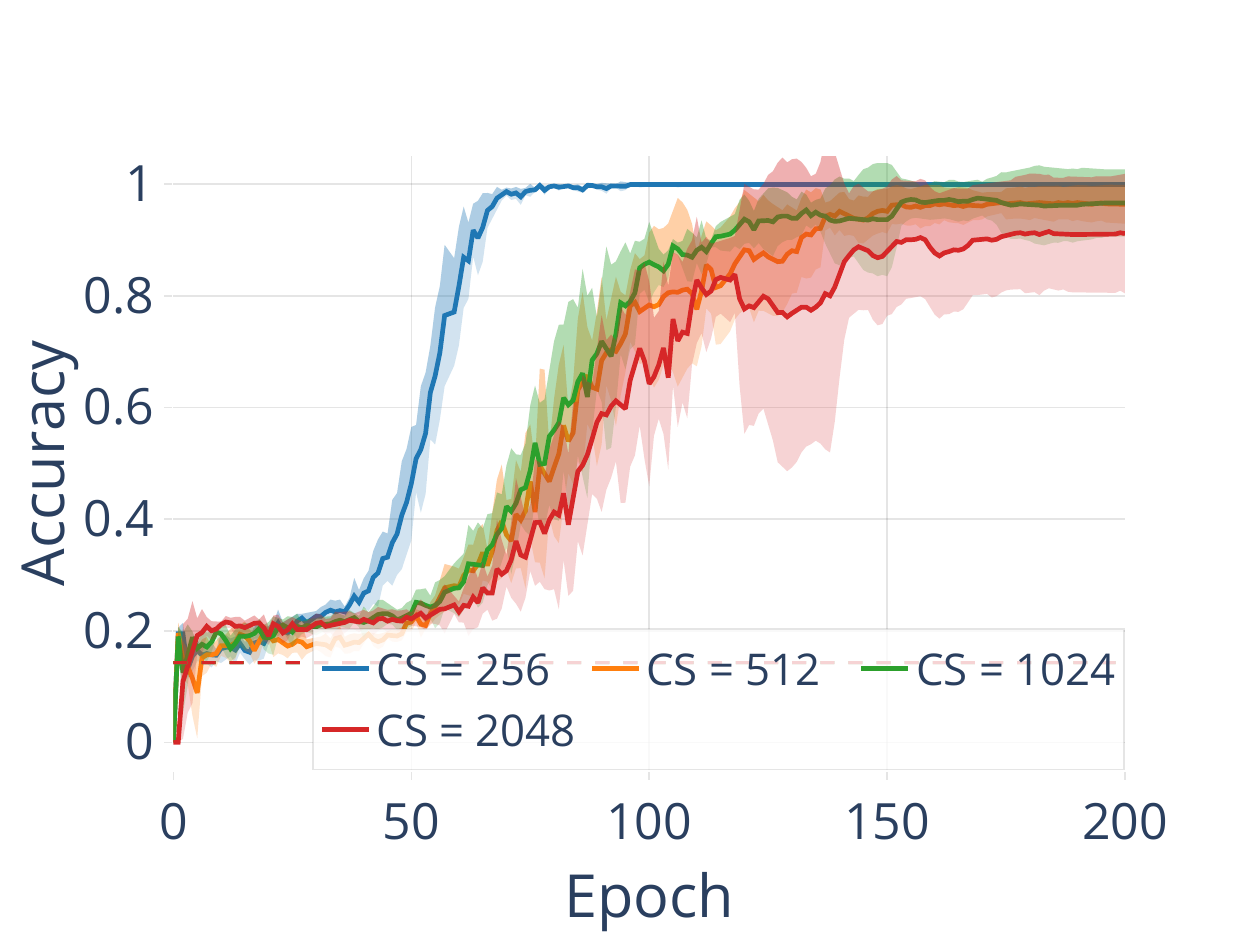}
    }
    \subfloat[Pythia-1B, $\ell = 26$]{
        \includegraphics[width=\smallThirdWidth]{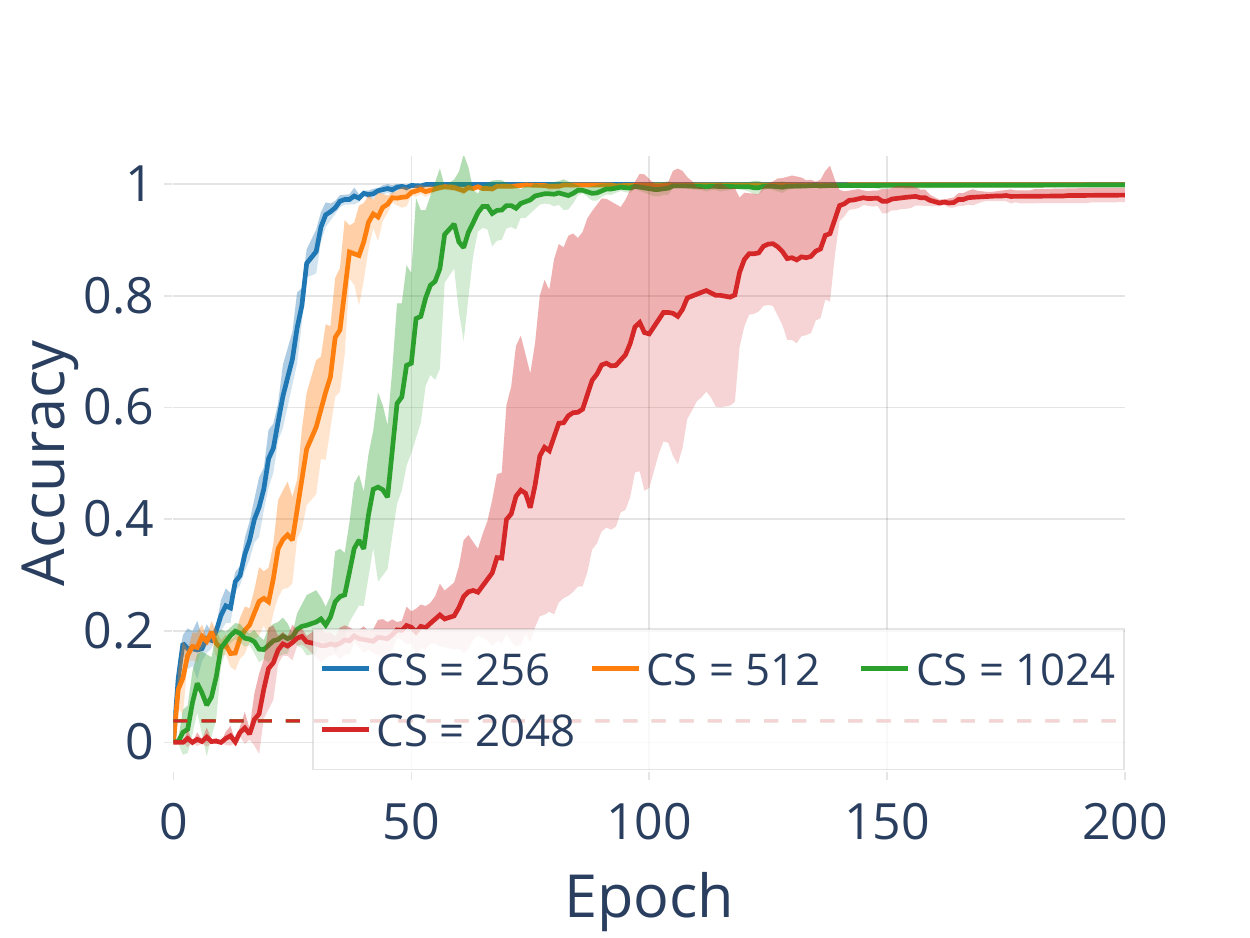}
    }
    \\
    \subfloat[Phi-2.7B, $\ell = 2$]{
        \includegraphics[width=\smallThirdWidth]{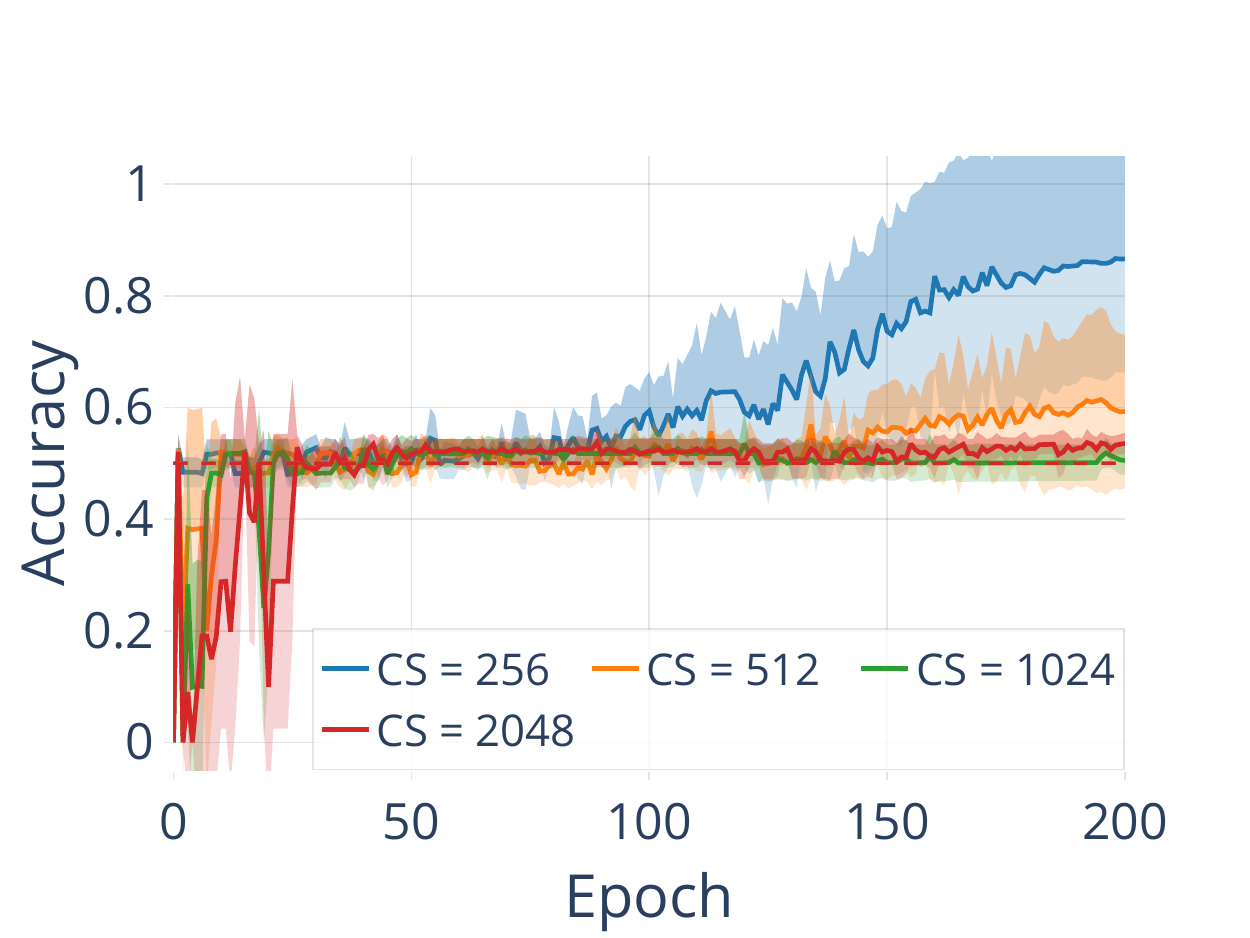}
    }
    \subfloat[Phi-2.7B, $\ell = 7$]{
        \includegraphics[width=\smallThirdWidth]{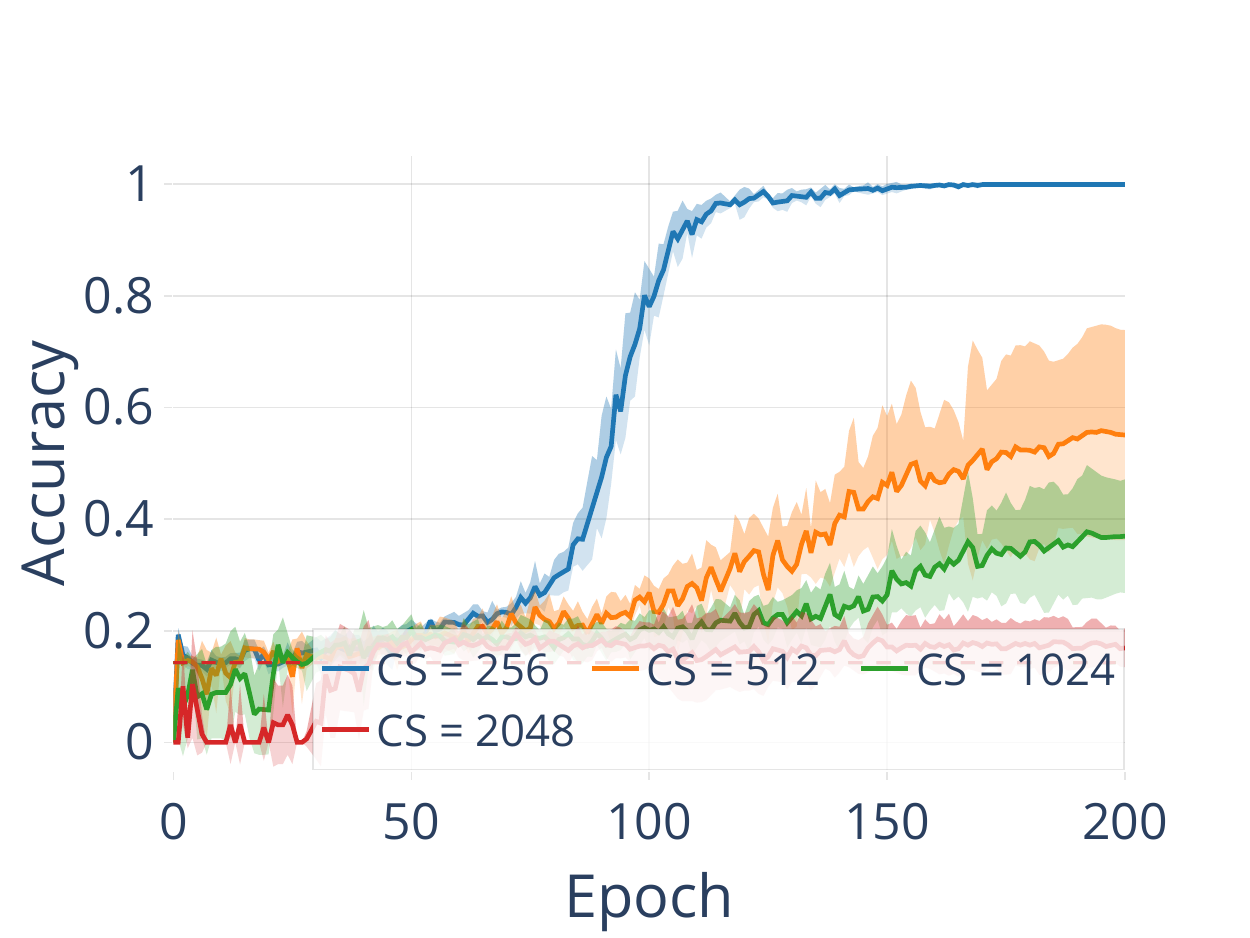}
    }
    \subfloat[Phi-2.7B, $\ell = 26$]{
        \includegraphics[width=\smallThirdWidth]{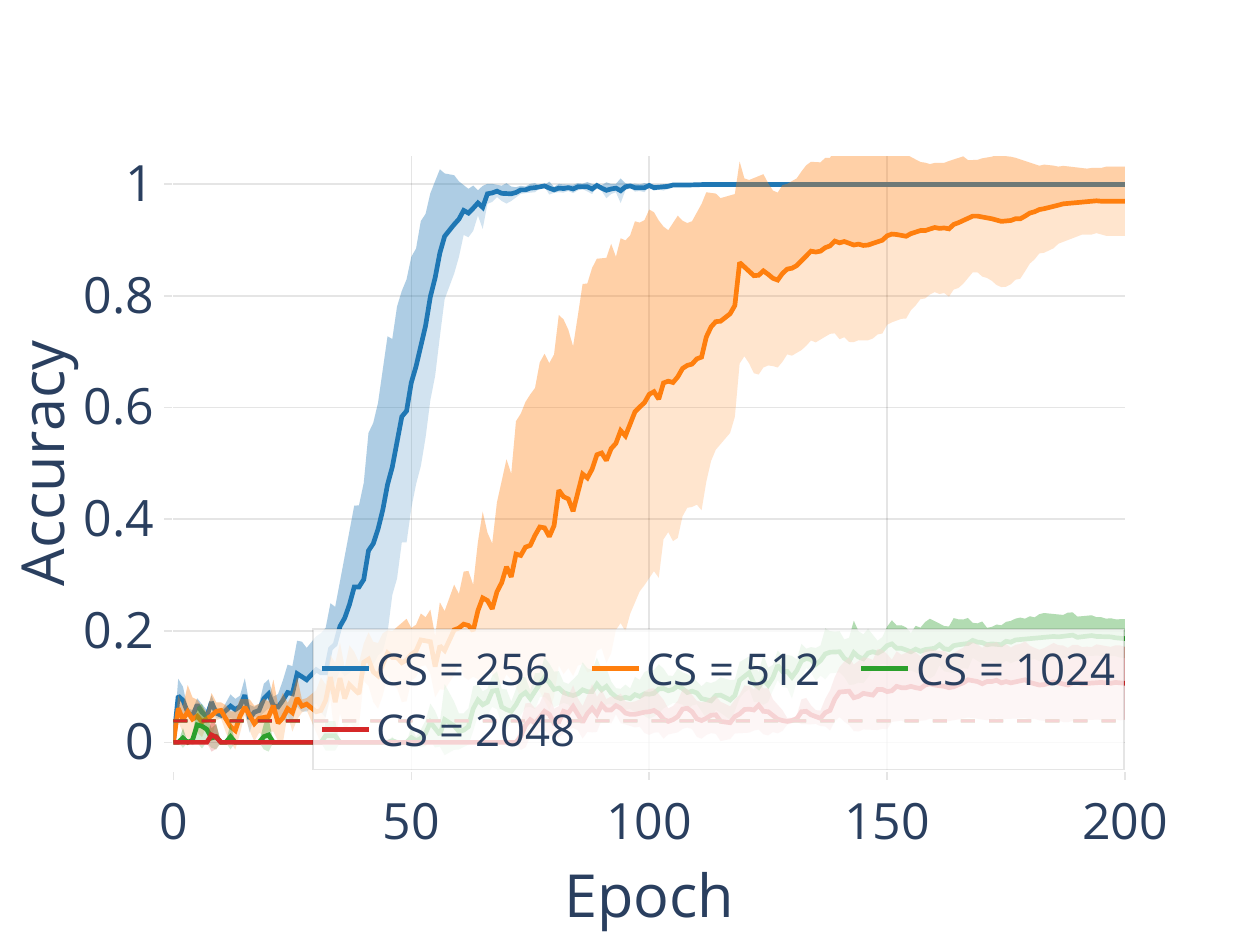}
    }
    \\
    \subfloat[Llama2-13B, $\ell = 2$]{
        \includegraphics[width=\smallThirdWidth]{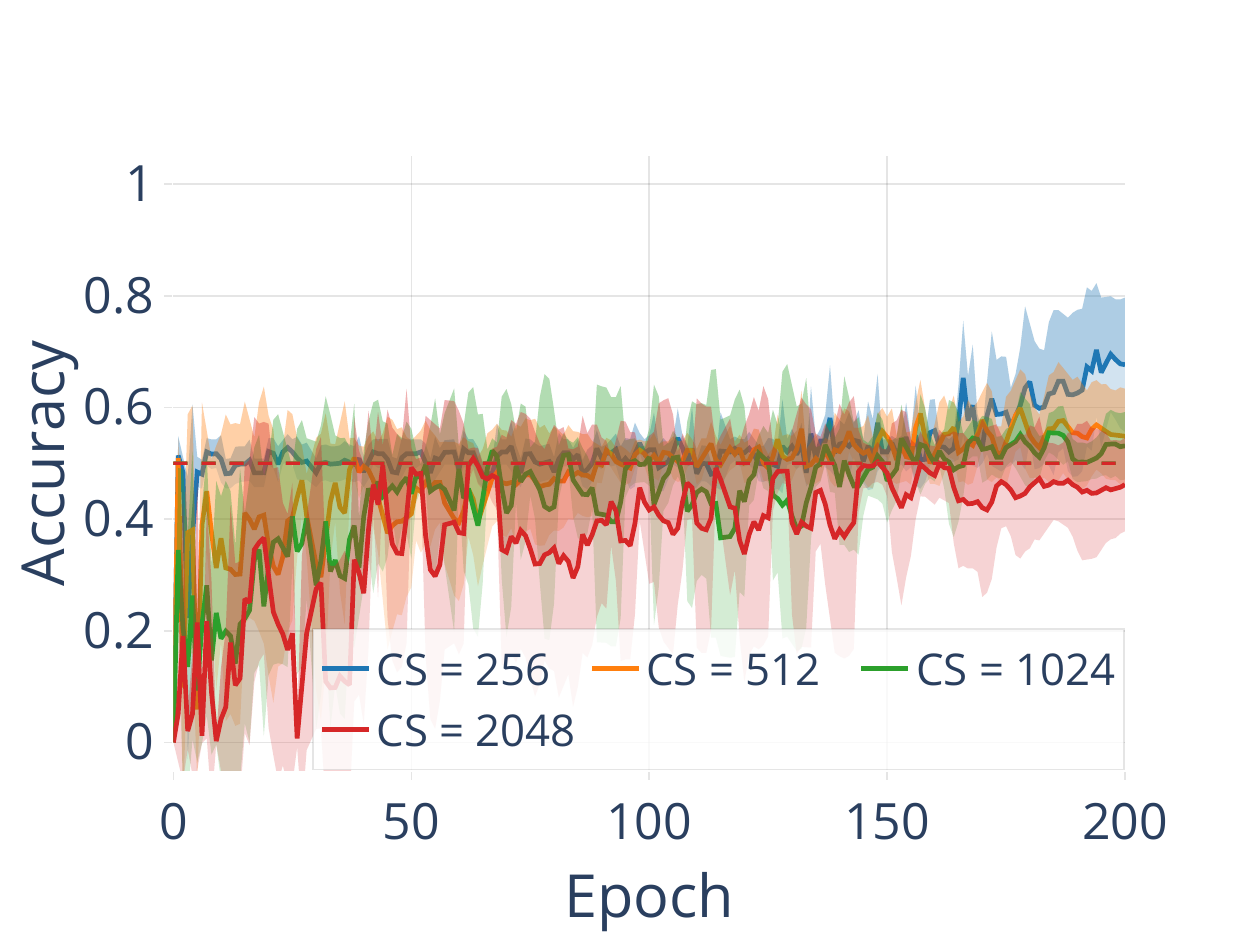}
    }
    \subfloat[Llama2-13B, $\ell = 7$]{
        \includegraphics[width=\smallThirdWidth]{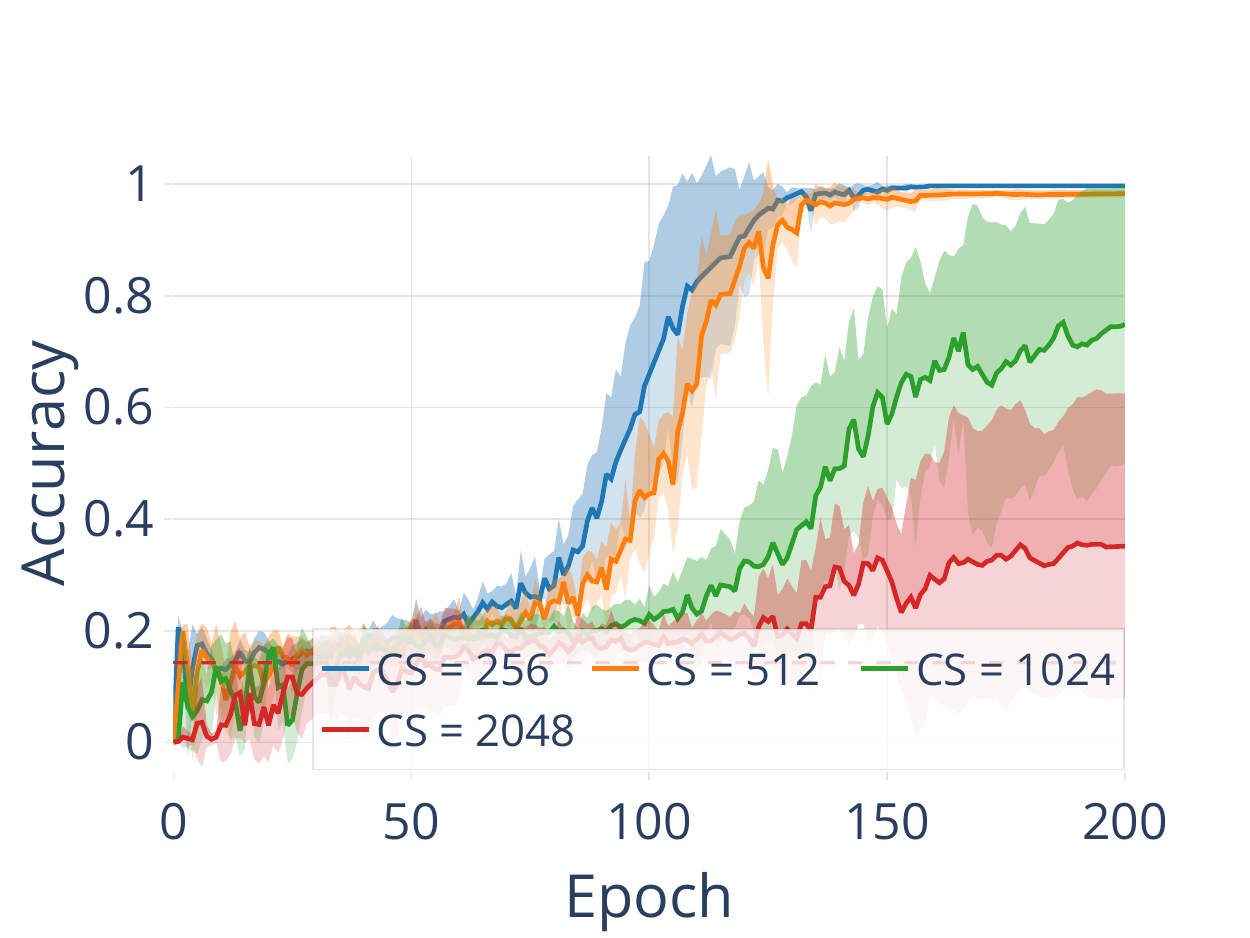}
    }
    \subfloat[Llama2-13B, $\ell = 26$]{
        \includegraphics[width=\smallThirdWidth]{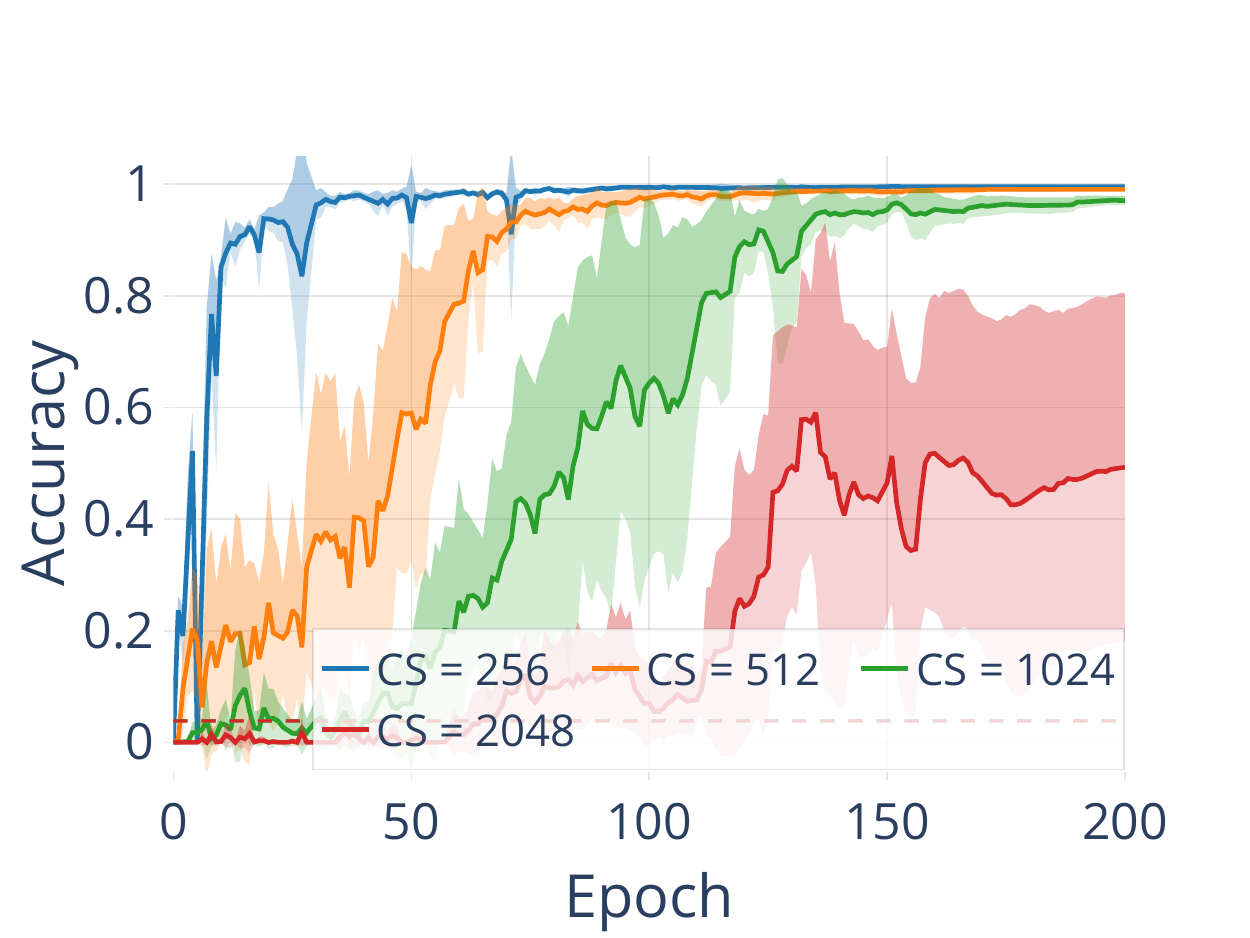}
    }
\caption{\capthead{Accuracy for \emph{untrained} models, when embedding random strings inside of longer strings of natural language data, for multiple $\ell$ and context sizes $c$.}{$n = 256$}
}
\label{fig:rw_val_cs_accuracy_untrained}
\end{figure}

\begin{figure}[H]
    \centering
    \subfloat[Pythia-1B, $\ell = 2$]{
        \includegraphics[width=\smallThirdWidth]{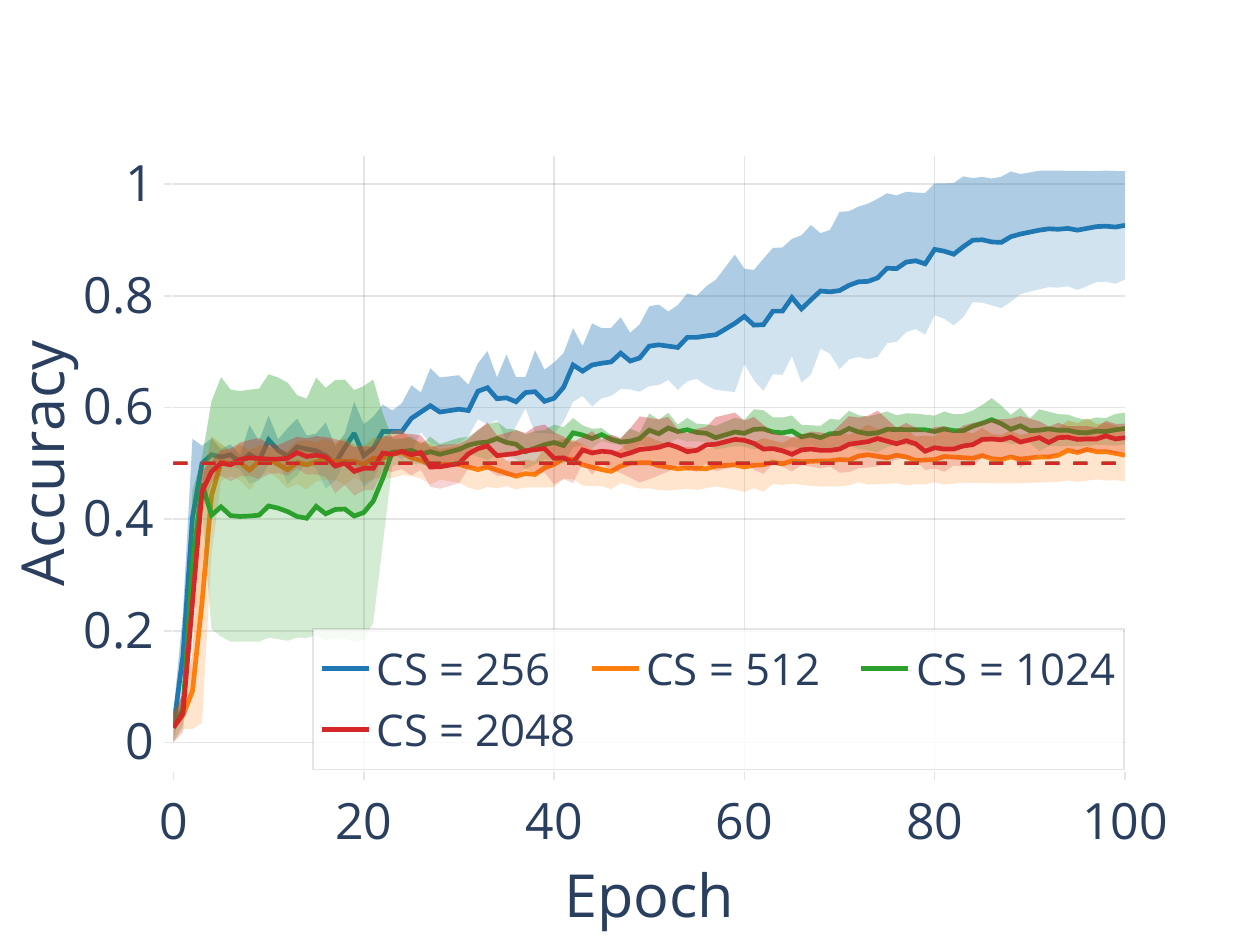}
    }
    \subfloat[Pythia-1B, $\ell = 7$]{
        \includegraphics[width=\smallThirdWidth]{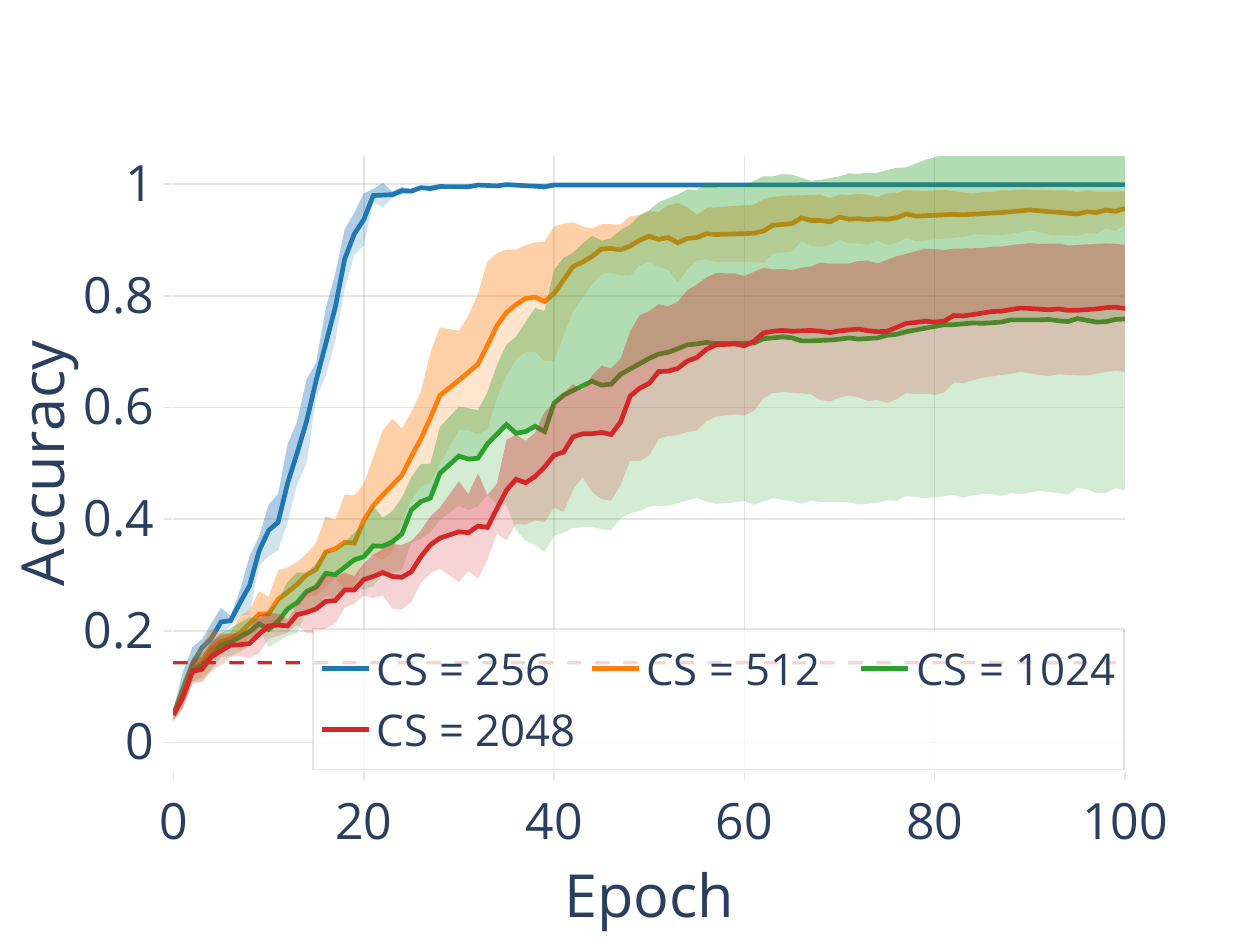}
    }
    \subfloat[Pythia-1B, $\ell = 26$]{
        \includegraphics[width=\smallThirdWidth]{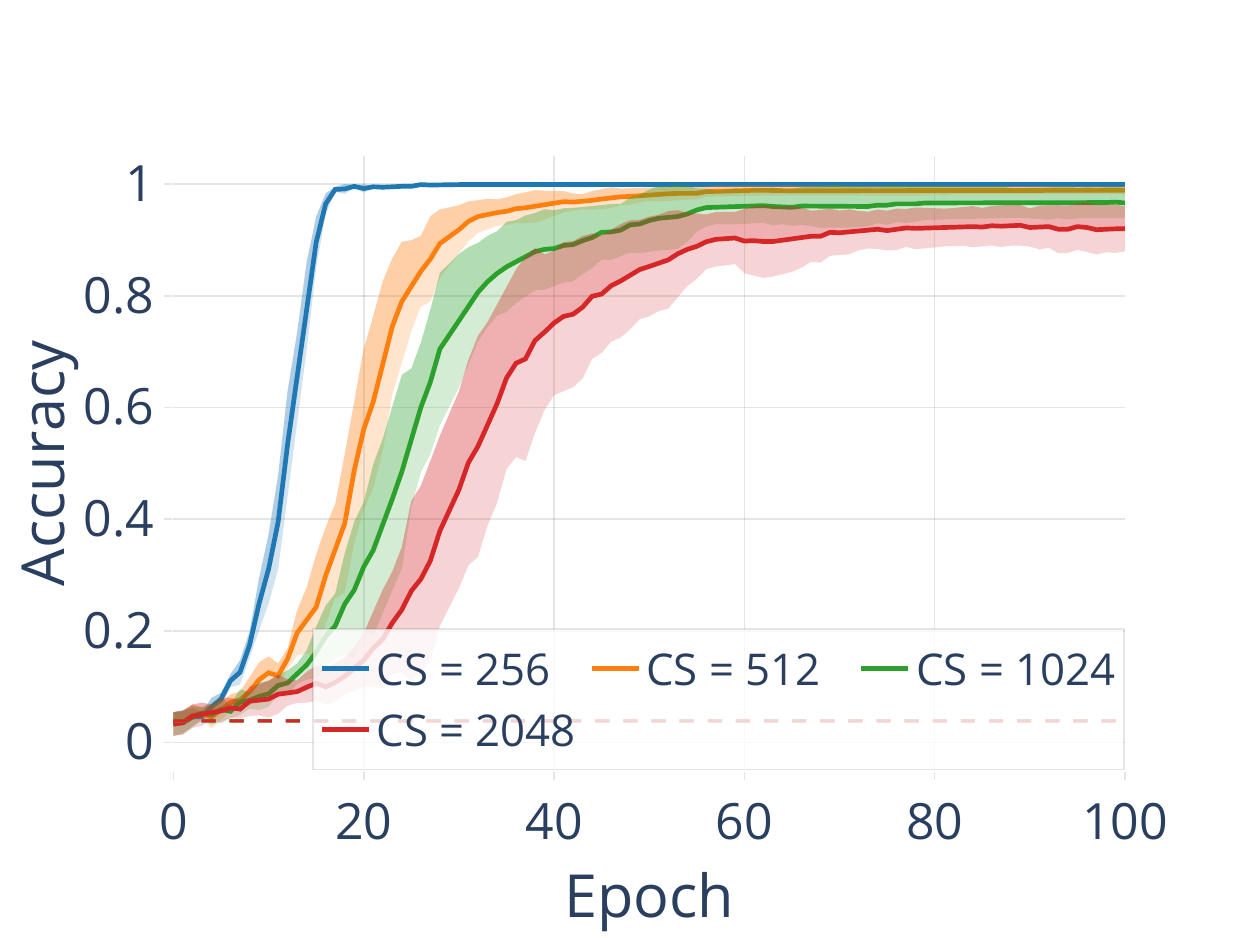}
    }
    \\
    \subfloat[Phi-2.7B, $\ell = 2$]{
        \includegraphics[width=\smallThirdWidth]{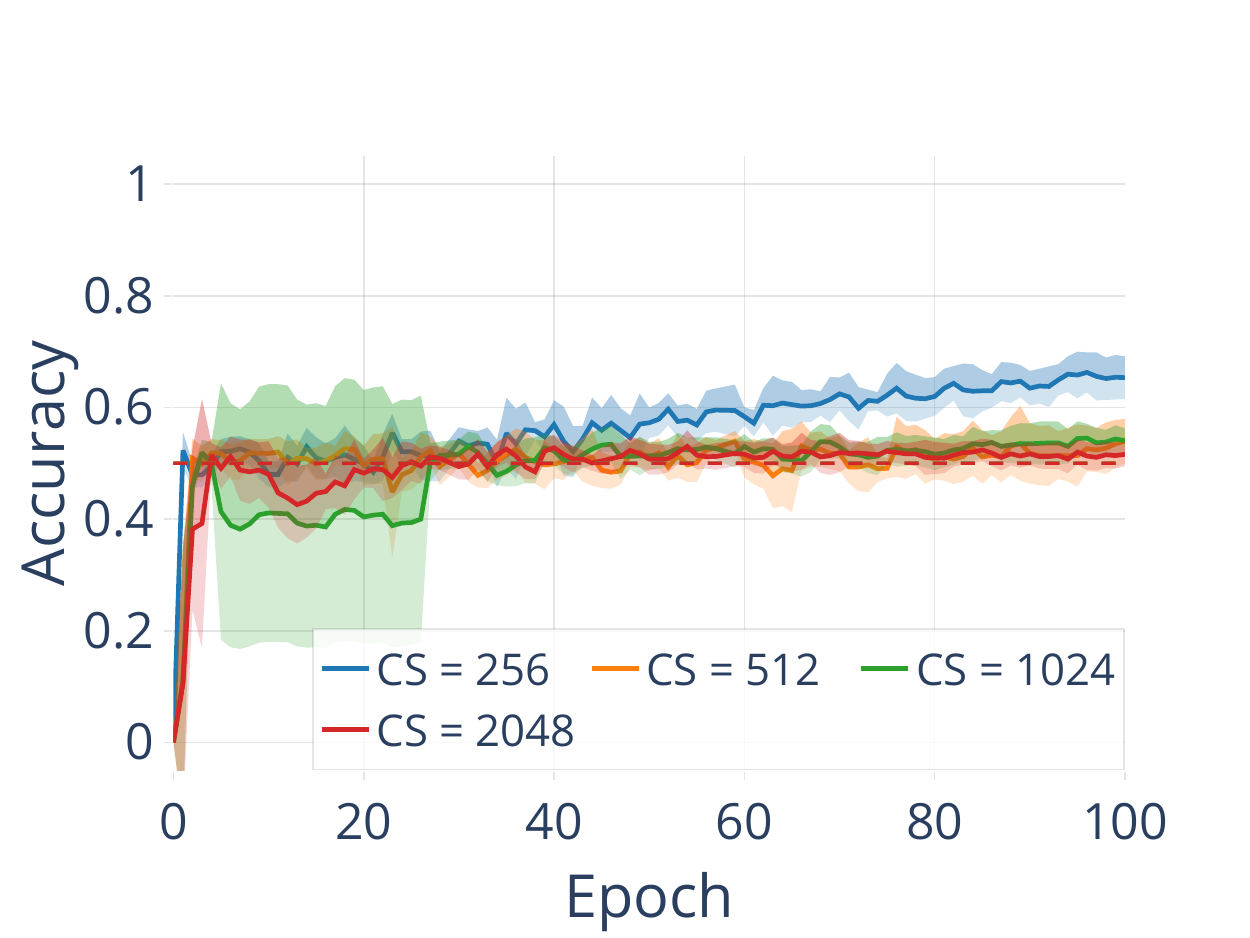}
    }
    \subfloat[Phi-2.7B, $\ell = 7$]{
        \includegraphics[width=\smallThirdWidth]{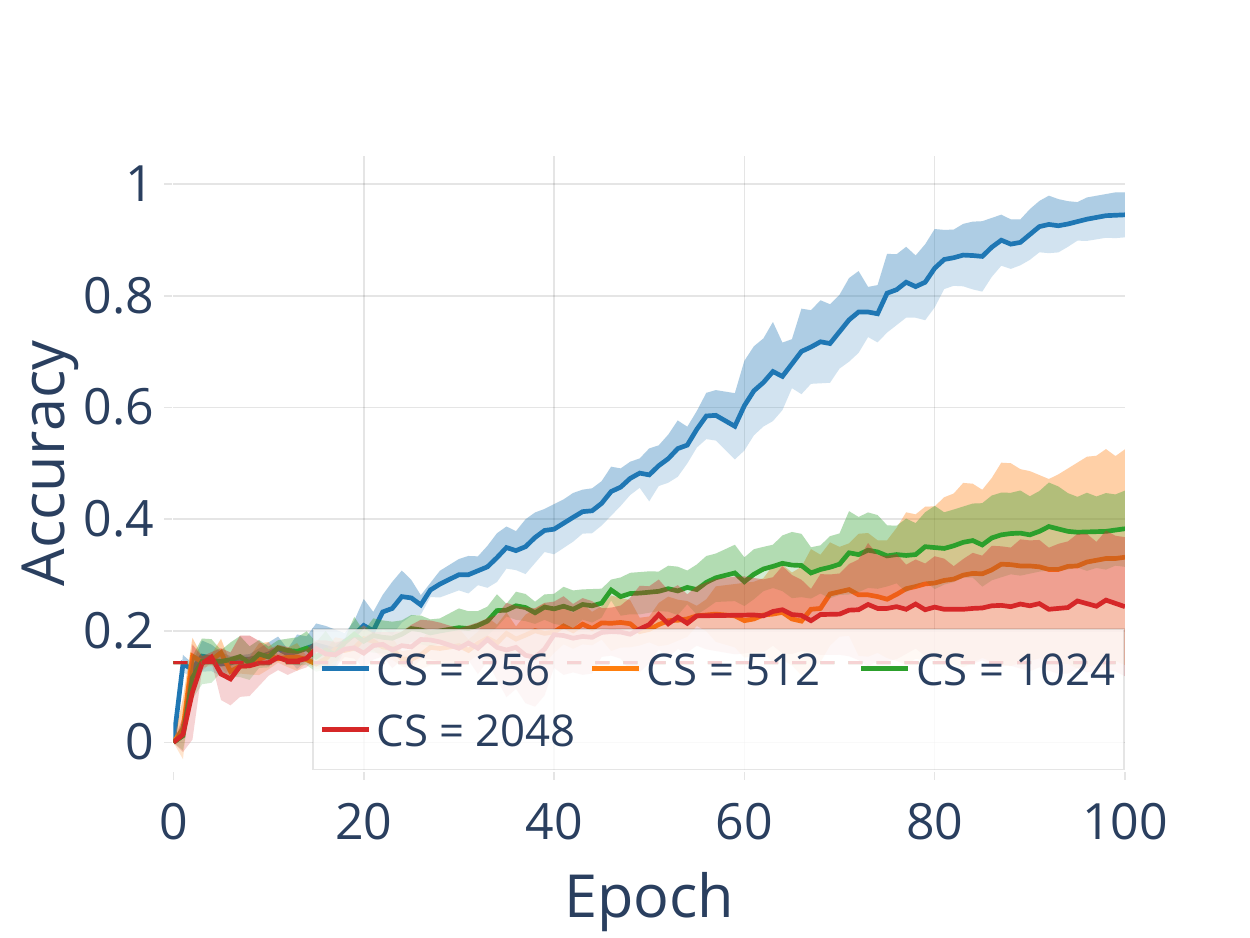}
    }
    \subfloat[Phi-2.7B, $\ell = 26$]{
        \includegraphics[width=\smallThirdWidth]{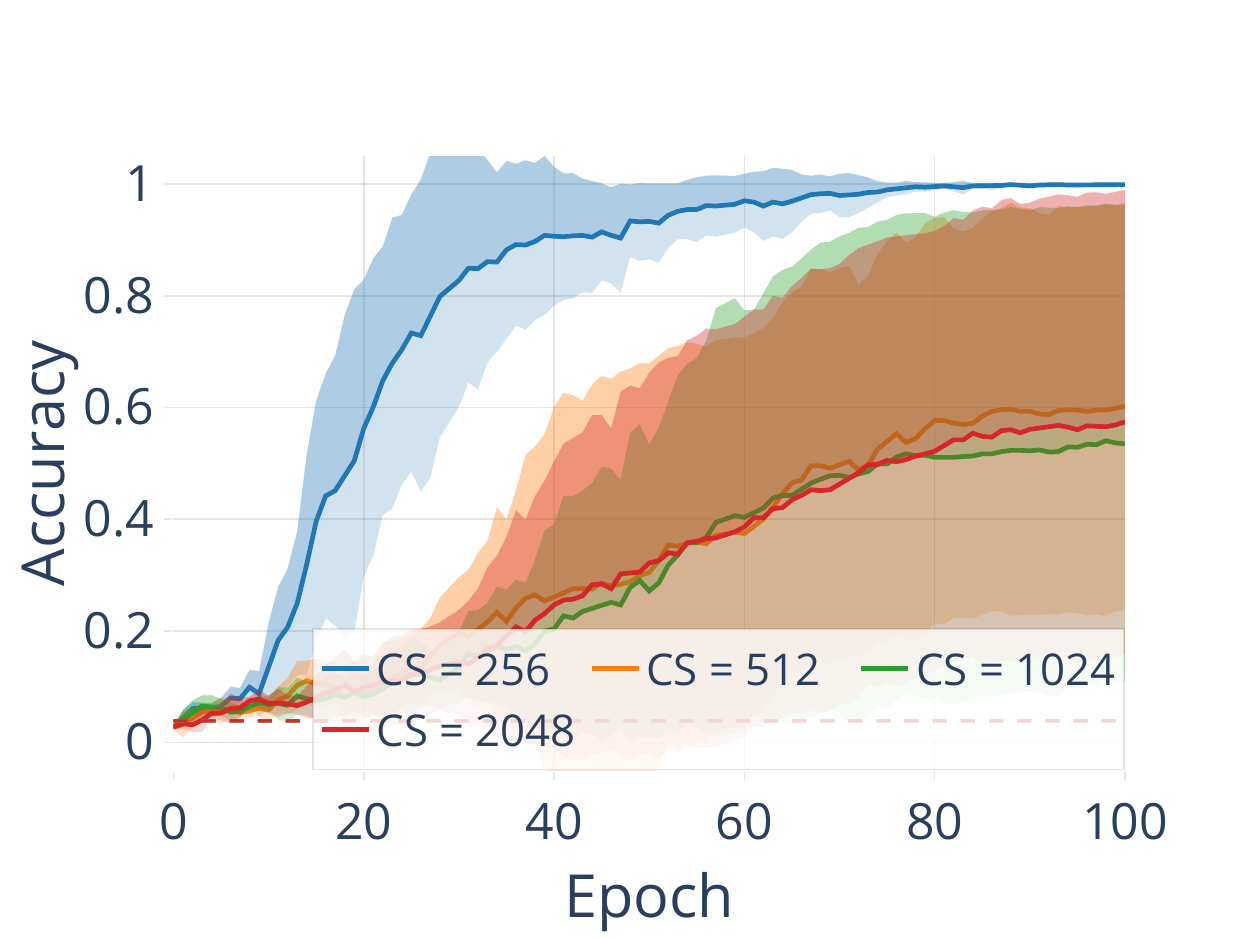}
    }
    \\
    \subfloat[Llama2-13B, $\ell = 2$]{
        \includegraphics[width=\smallThirdWidth]{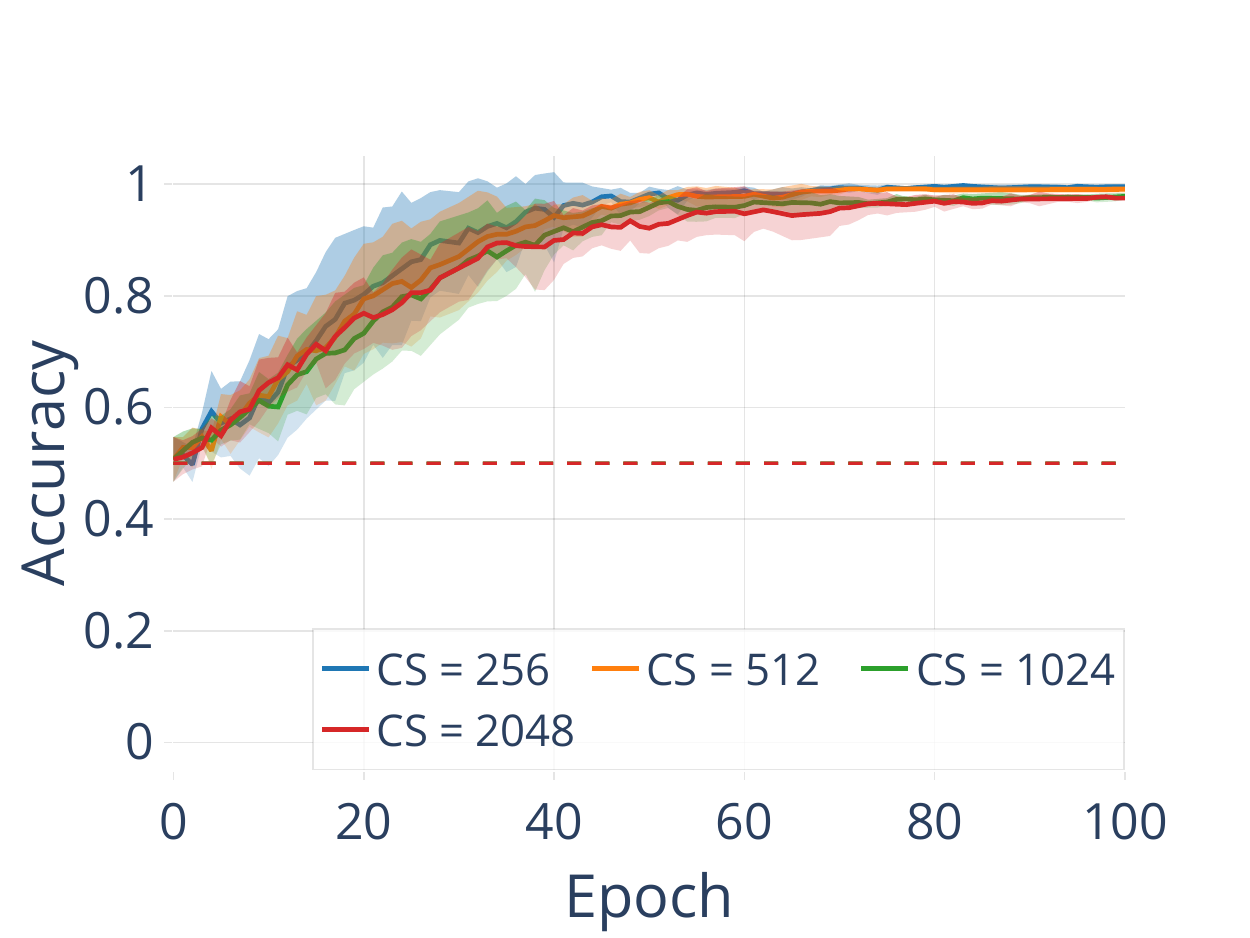}
    }
    \subfloat[Llama2-13B, $\ell = 7$]{
        \includegraphics[width=\smallThirdWidth]{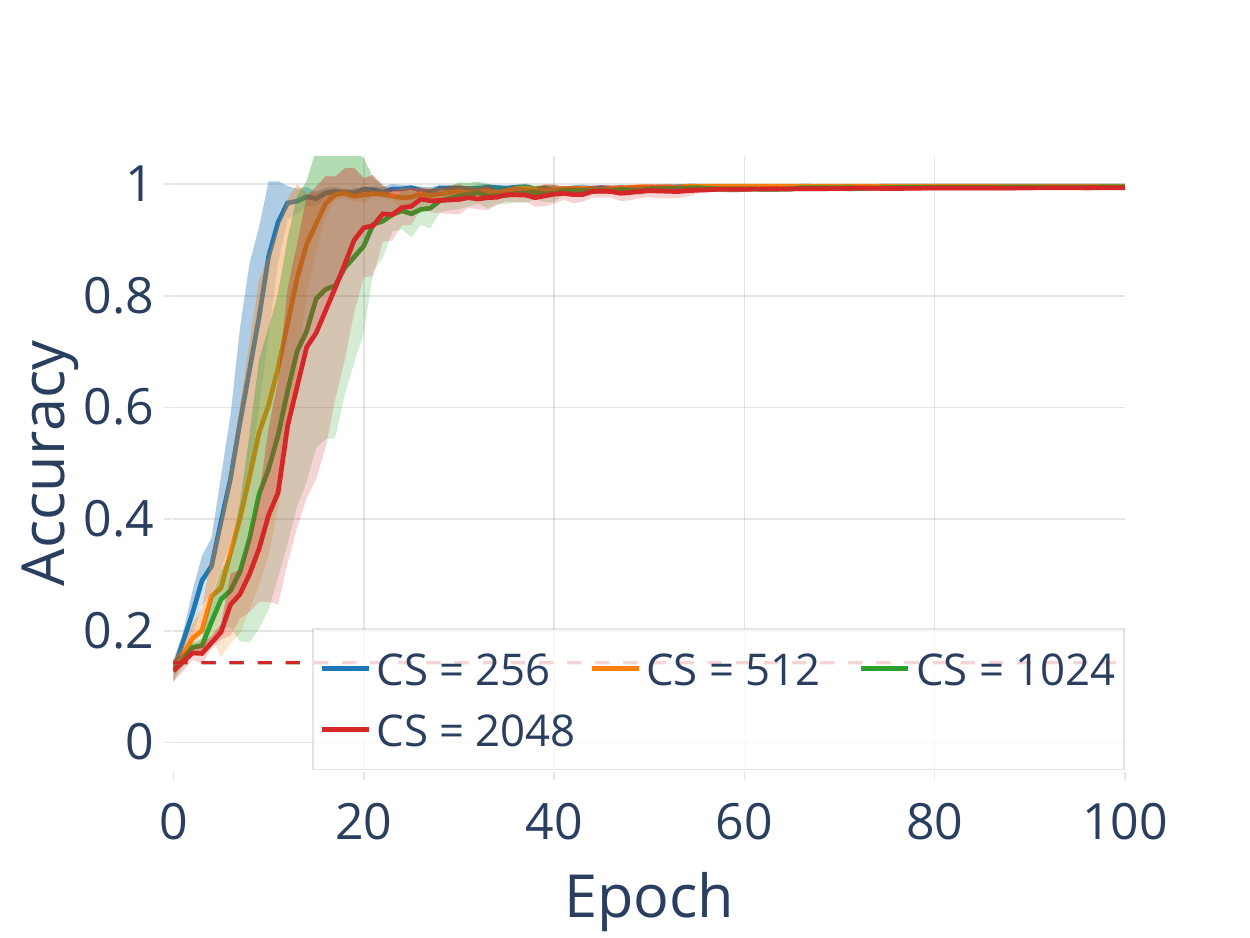}
    }
    \subfloat[Llama2-13B, $\ell = 26$]{
        \includegraphics[width=\smallThirdWidth]{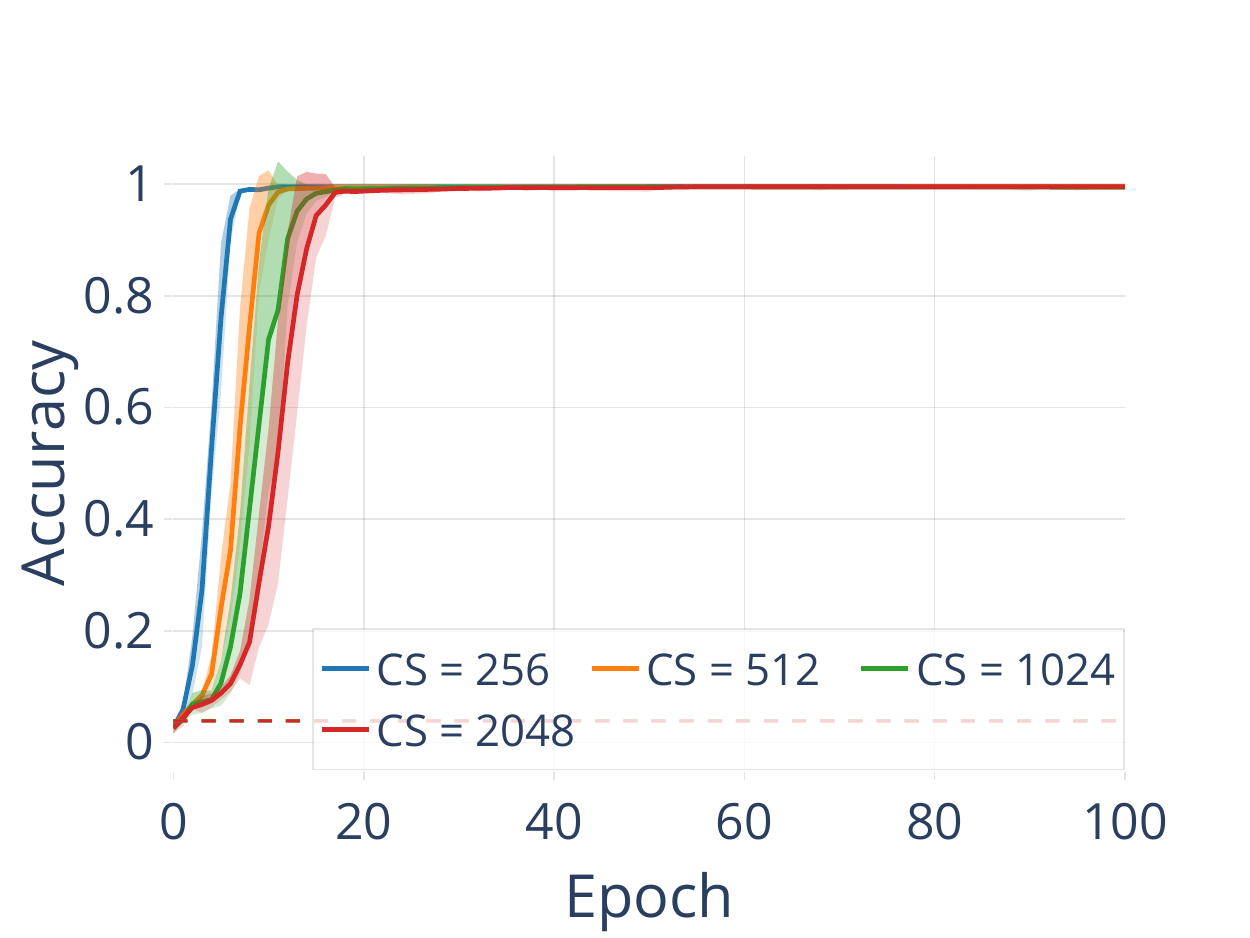}
    }
\caption{\capthead{Accuracy for \emph{pretrained} models, when embedding random strings inside of longer strings of natural language data, for multiple $\ell$ and context sizes $c$.}{$n = 256$}
}
\label{fig:rw_val_cs_accuracy_pretrained}
\end{figure}

Figure~\ref{fig:rw_val_cs_accuracy_untrained} shows the accuracy of untrained models on embedded random strings, and Figure~\ref{fig:rw_val_cs_accuracy_pretrained} shows the accuracy of pretrained models.
As with batched training, the models exhibit both the \GuessPhase and the \MemPhase.
Again, the difficulty of memorisation depends on the entropy of the random string, with higher entropy strings being easier to memorise.
These findings are are also consistent with our previous results where we train on random strings in isolation.

\subsection{Impact of random string memorisation on natural language performance}
\label{app:mem_perf_impact}

To better understand the implications of our experimental setup, we estimate the performance impact of memorising random strings on natural language modelling.
In particular, we measure how memorising a single random string impacts the model’s loss on the wikitext testset~\cite{merity2016pointer}.
We perform these measurements for the same models trained on batches of natural language data with single random sequences injected as described in Appendix~\ref{app:rw_val_batches}, and on strings of natural data with embedded random substrings as in Appendix~\ref{app:rw_val_strings}.
We focus on pretrained models here to understand how their previously acquired language modelling capabilities are affected by memorising random strings.

\begin{figure}[H]
    \centering
    \subfloat[Pythia-1B, $\ell = 2$]{
        \includegraphics[width=\smallThirdWidth]{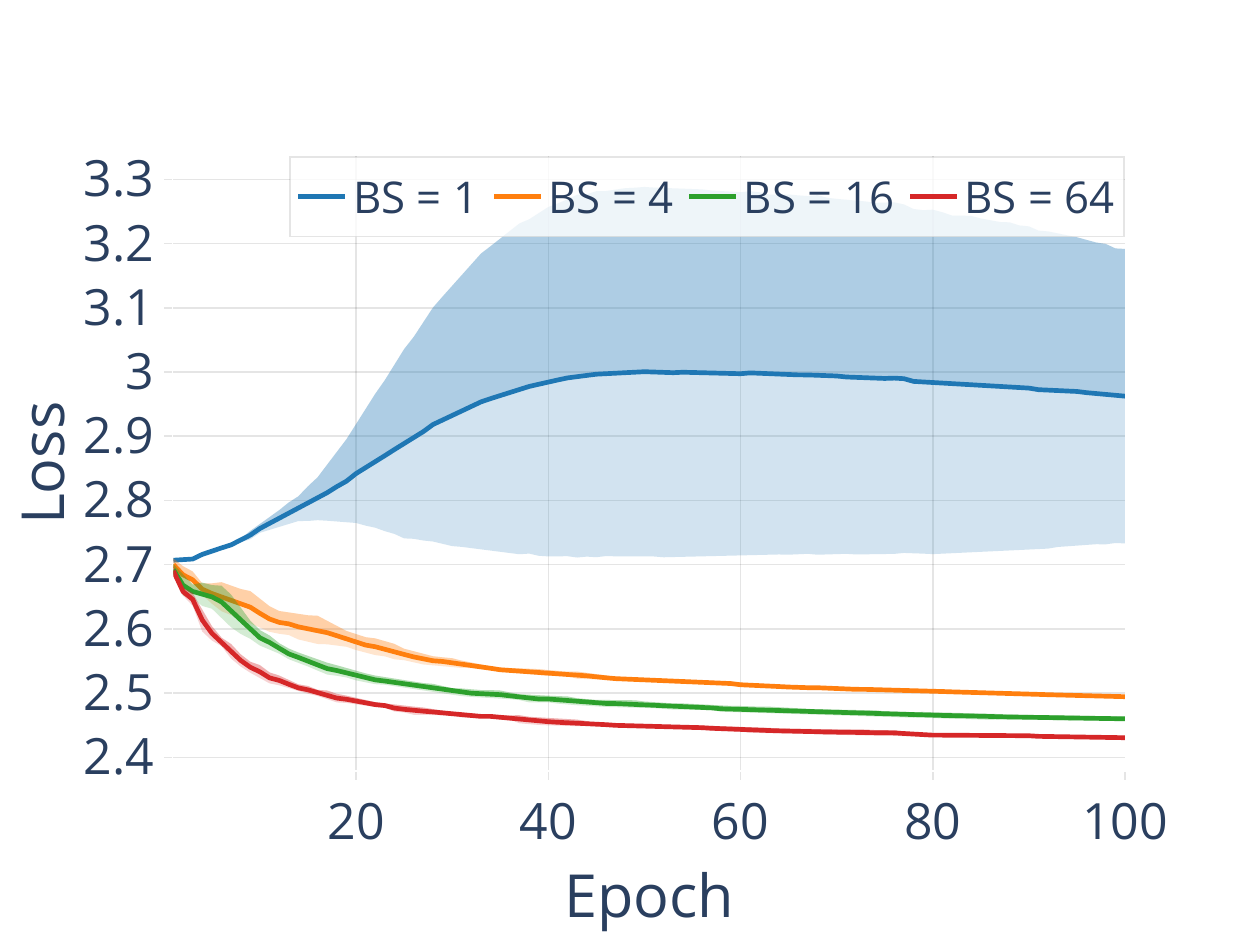}
    }
    \subfloat[Pythia-1B, $\ell = 7$]{
        \includegraphics[width=\smallThirdWidth]{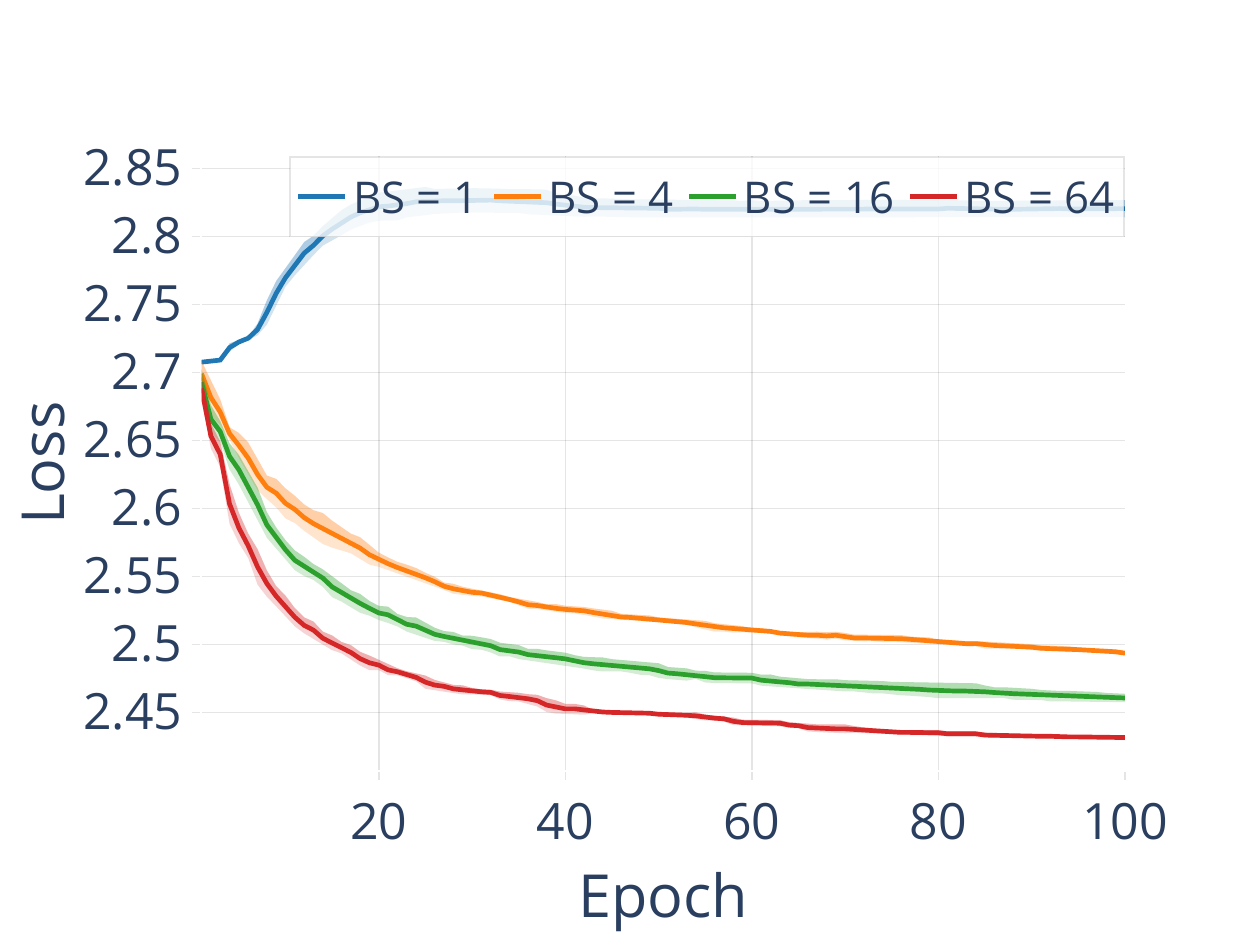}
    }
    \subfloat[Pythia-1B, $\ell = 26$]{
        \includegraphics[width=\smallThirdWidth]{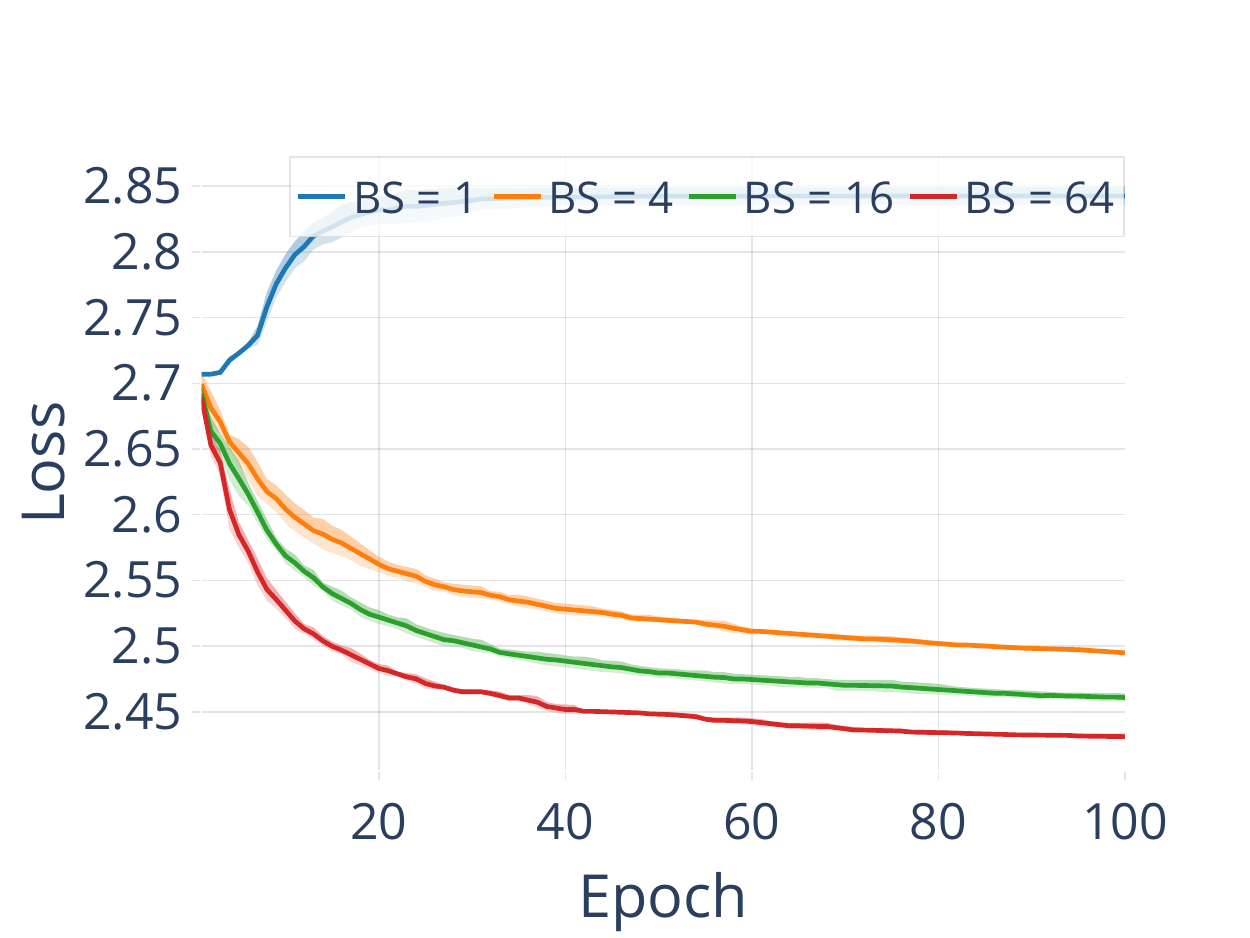}
    }
    \\
    \subfloat[Phi-2.7B, $\ell = 2$]{
        \includegraphics[width=\smallThirdWidth]{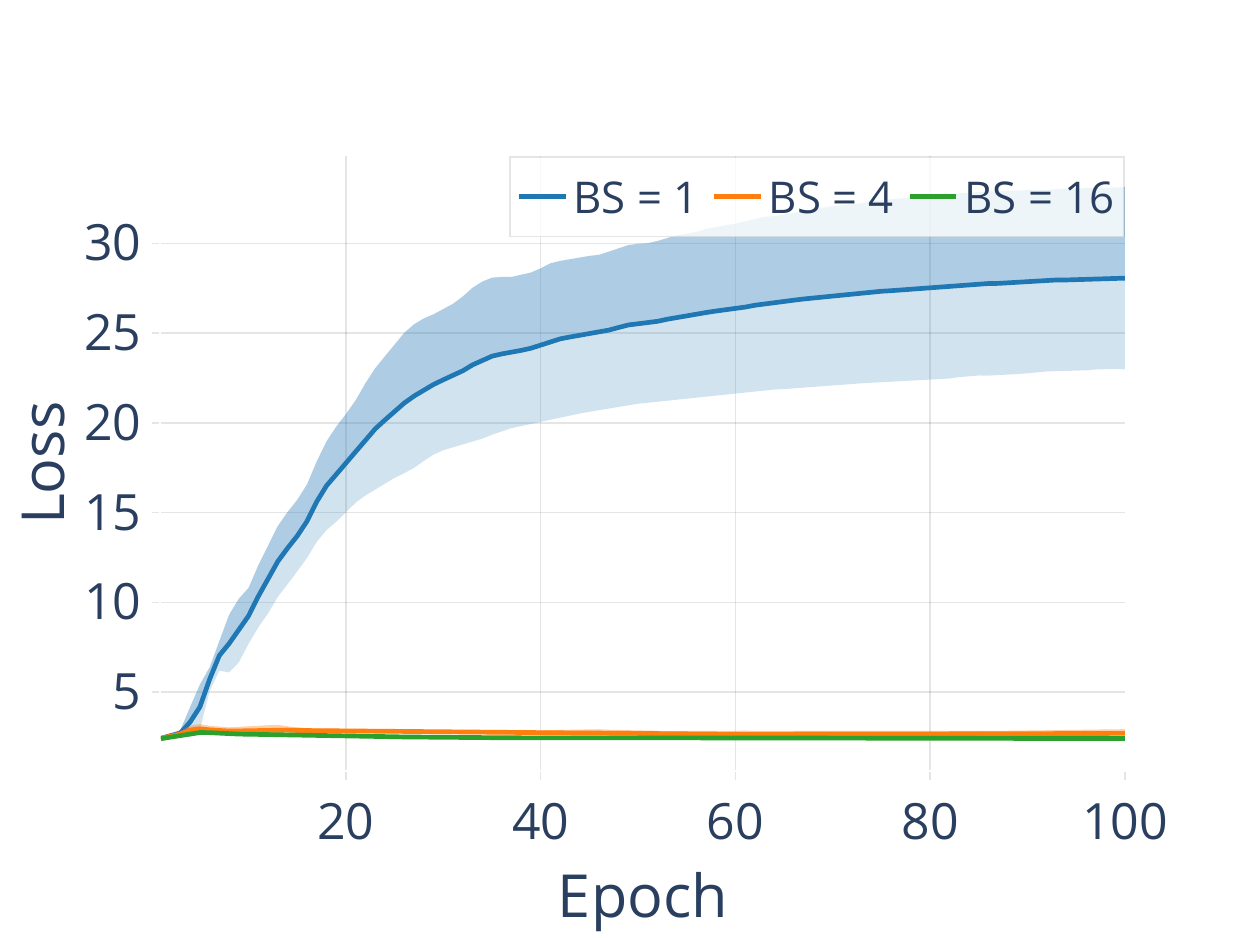}
    }
    \subfloat[Phi-2.7B, $\ell = 7$]{
        \includegraphics[width=\smallThirdWidth]{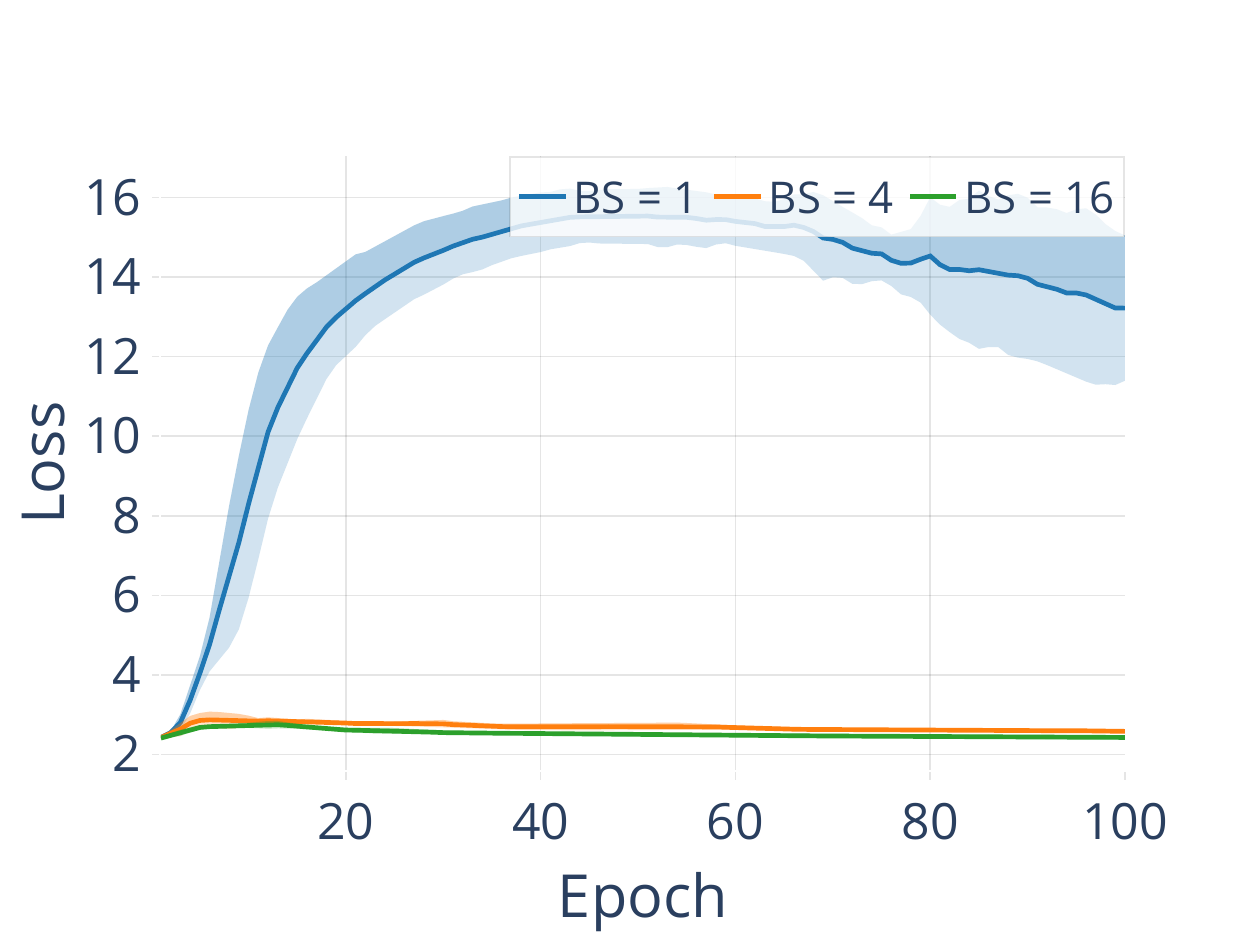}
    }
    \subfloat[Phi-2.7B, $\ell = 26$]{
        \includegraphics[width=\smallThirdWidth]{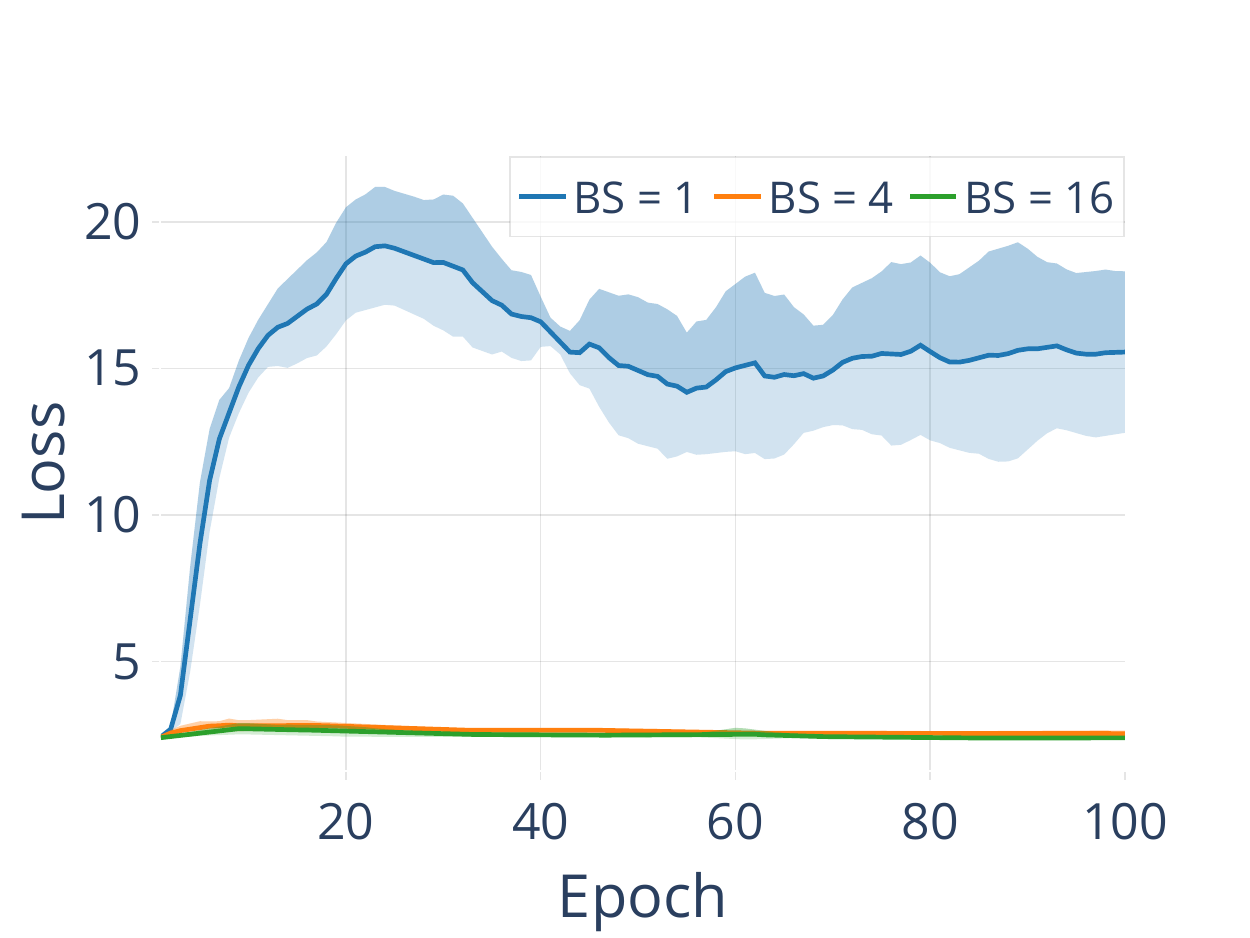}
    }
    \\
    \subfloat[Llama2-13B, $\ell = 2$]{
        \includegraphics[width=\smallThirdWidth]{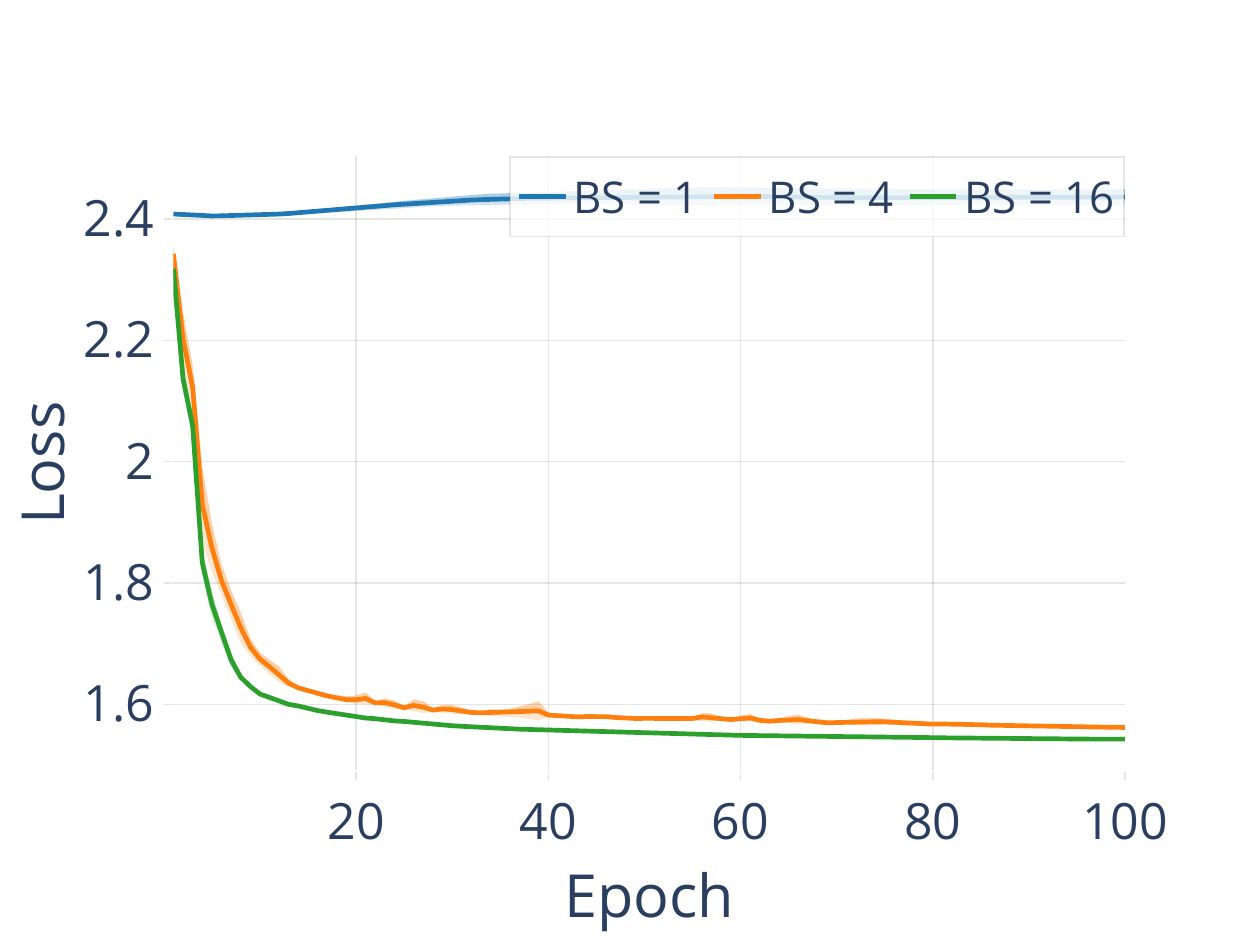}
    }
    \subfloat[Llama2-13B, $\ell = 7$]{
        \includegraphics[width=\smallThirdWidth]{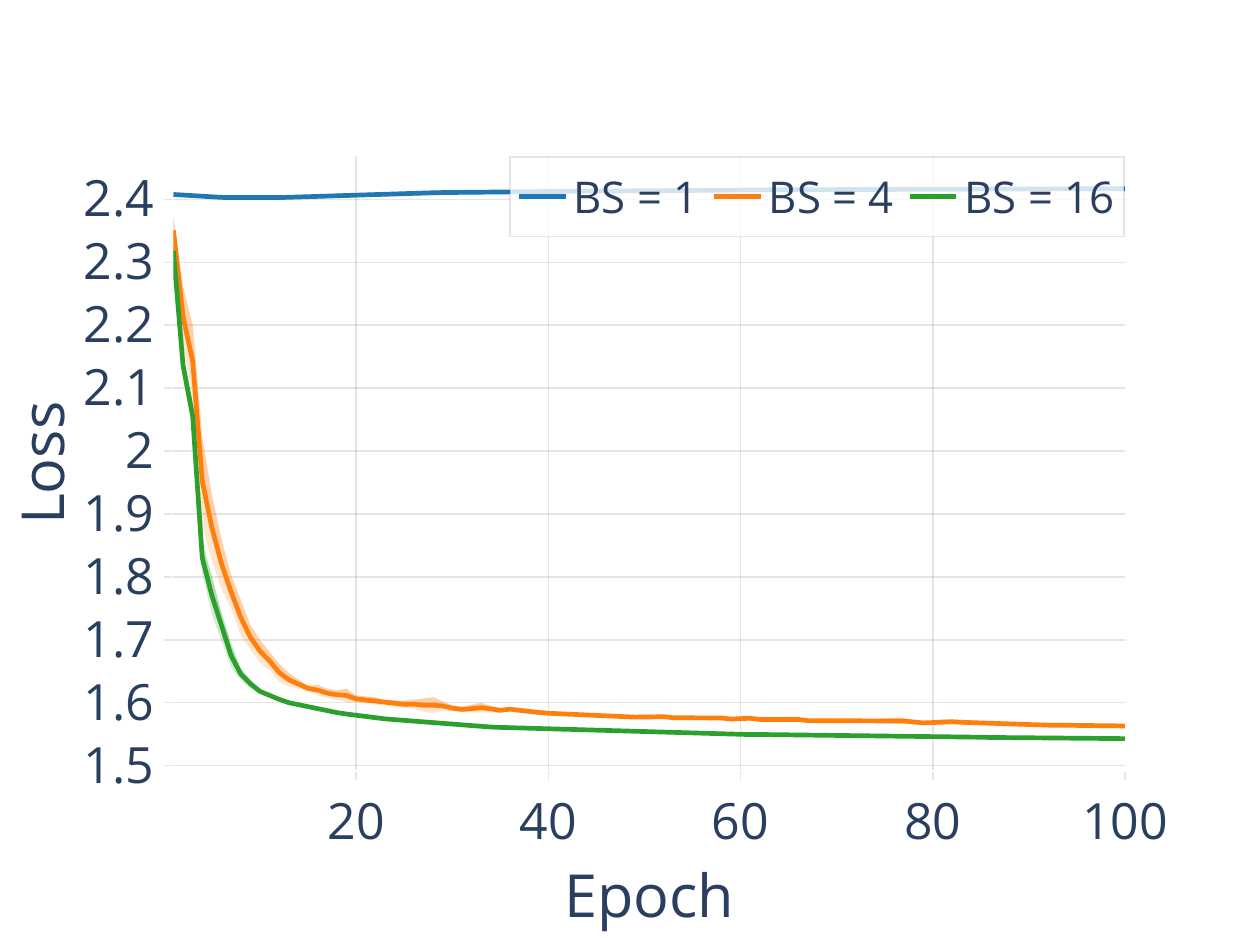}
    }
    \subfloat[Llama2-13B, $\ell = 26$]{
        \includegraphics[width=\smallThirdWidth]{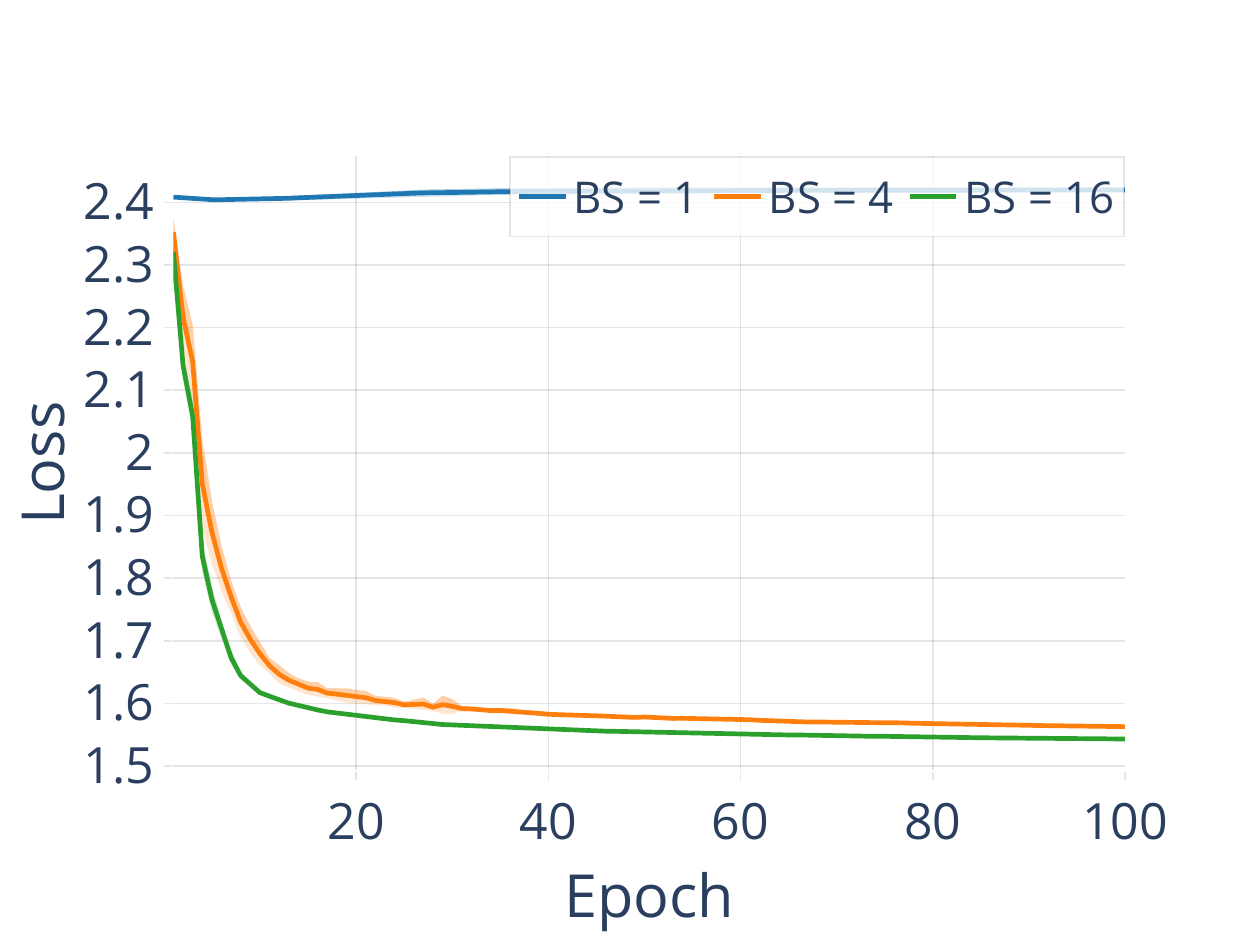}
    }
\caption{\capthead{Loss on a natural-language testset during memorisation, when embedding random strings inside of batches of natural language data, for multiple $\ell$ and batch sizes.}{$n = 1024$}
}
\label{fig:perf_impact_bs}
\end{figure}

\begin{figure}[H]
    \centering
    \subfloat[Pythia-1B, $\ell = 2$]{
        \includegraphics[width=\smallThirdWidth]{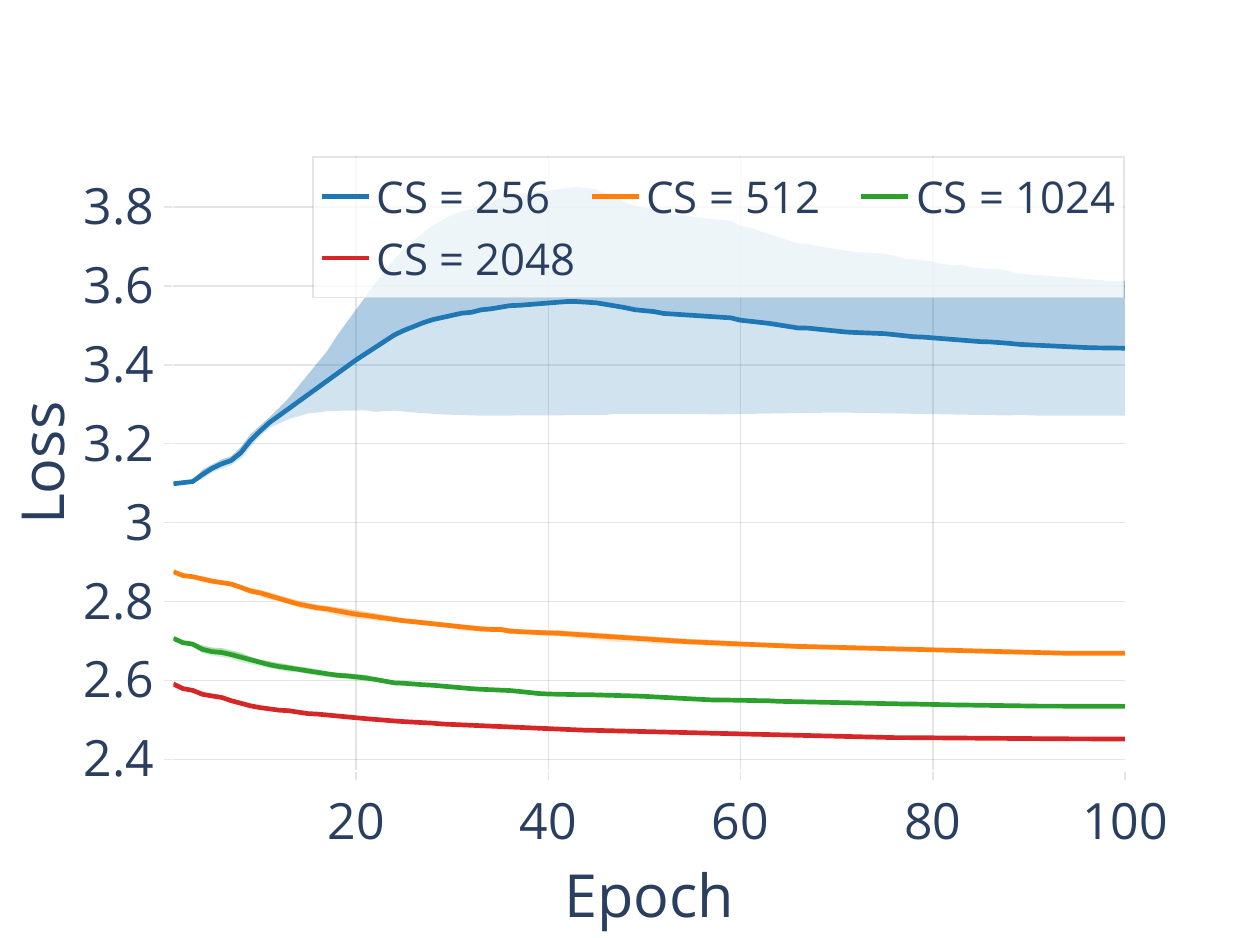}
    }
    \subfloat[Pythia-1B, $\ell = 7$]{
        \includegraphics[width=\smallThirdWidth]{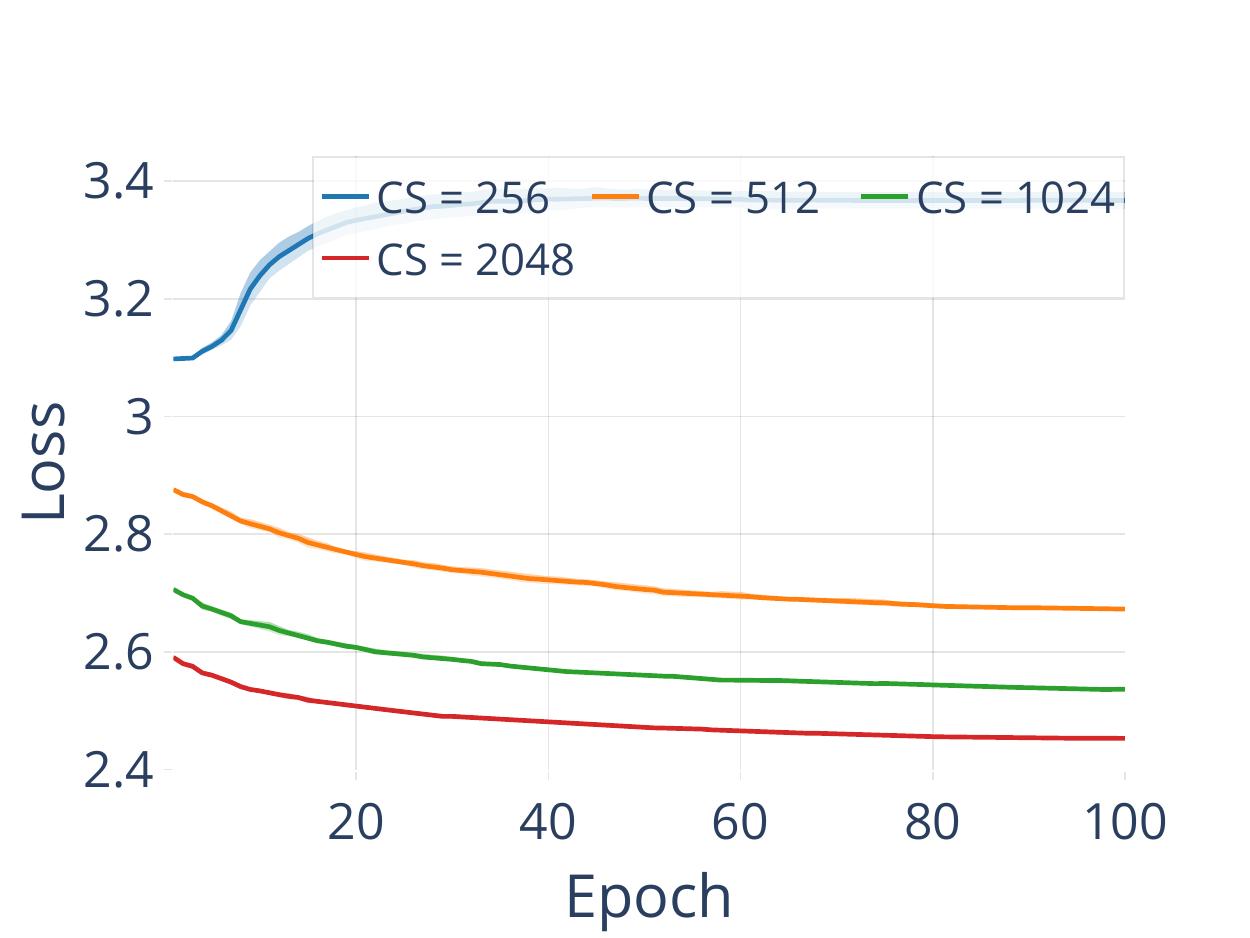}
    }
    \subfloat[Pythia-1B, $\ell = 26$]{
        \includegraphics[width=\smallThirdWidth]{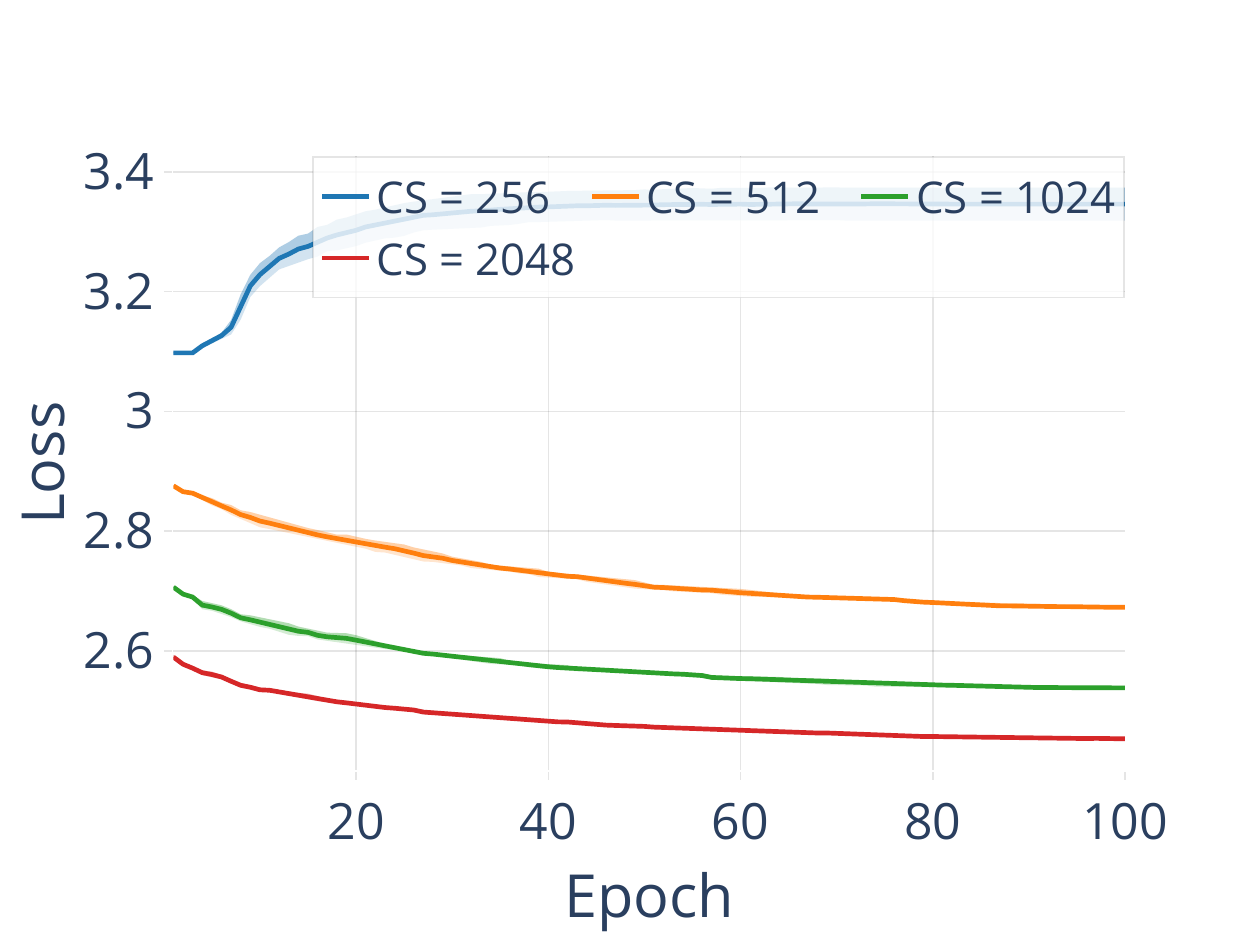}
    }
    \\
    \subfloat[Phi-2.7B, $\ell = 2$]{
        \includegraphics[width=\smallThirdWidth]{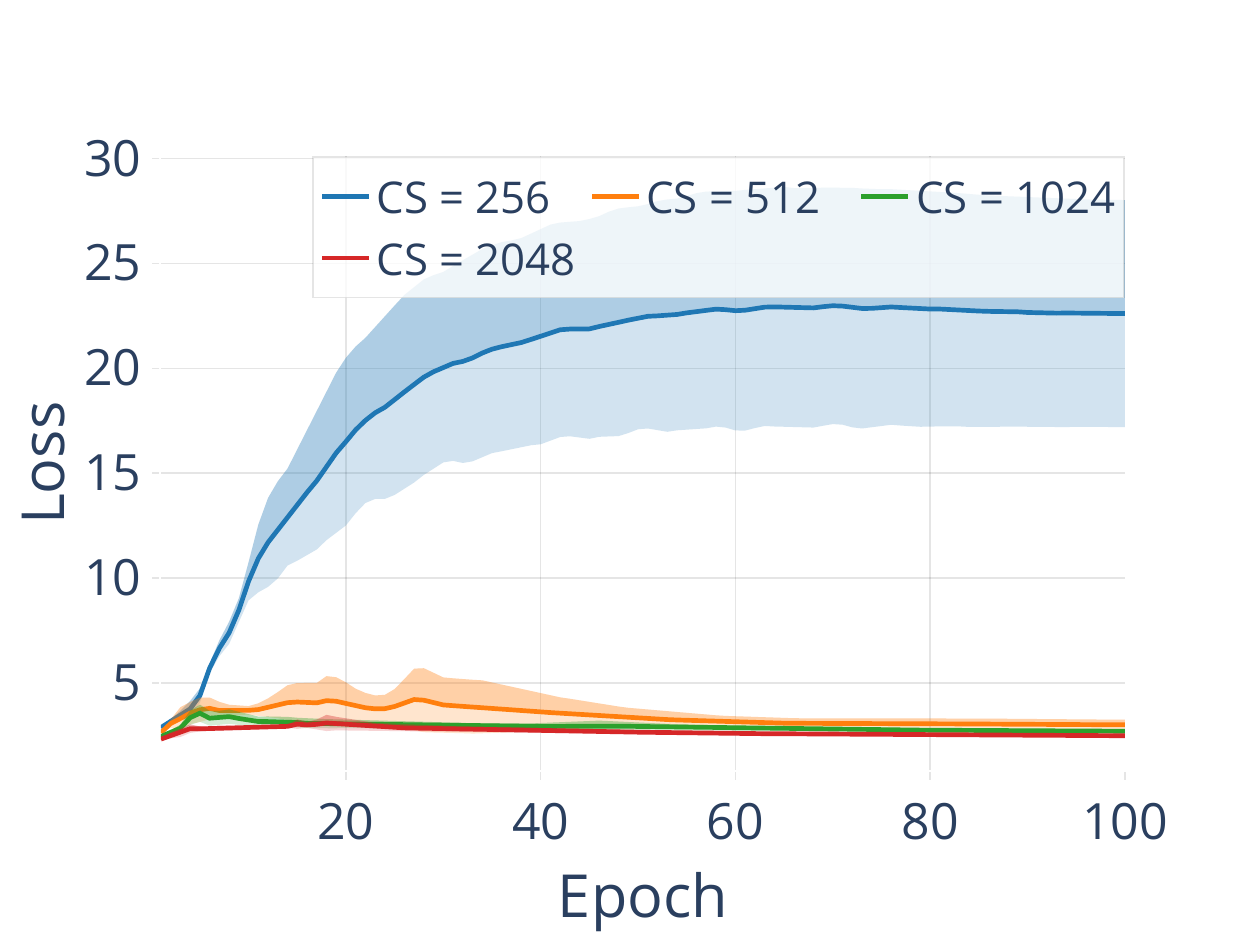}
    }
    \subfloat[Phi-2.7B, $\ell = 7$]{
        \includegraphics[width=\smallThirdWidth]{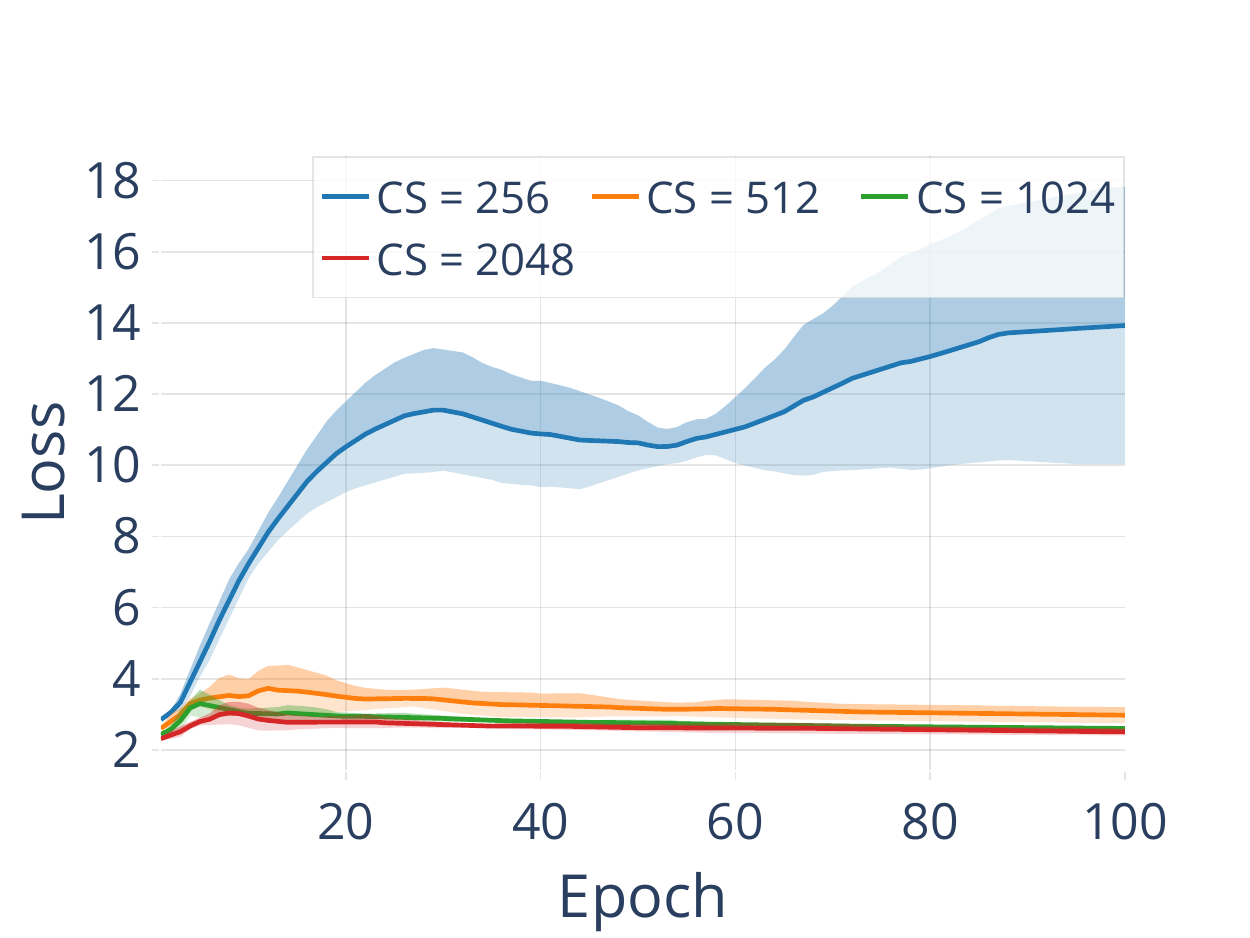}
    }
    \subfloat[Phi-2.7B, $\ell = 26$]{
        \includegraphics[width=\smallThirdWidth]{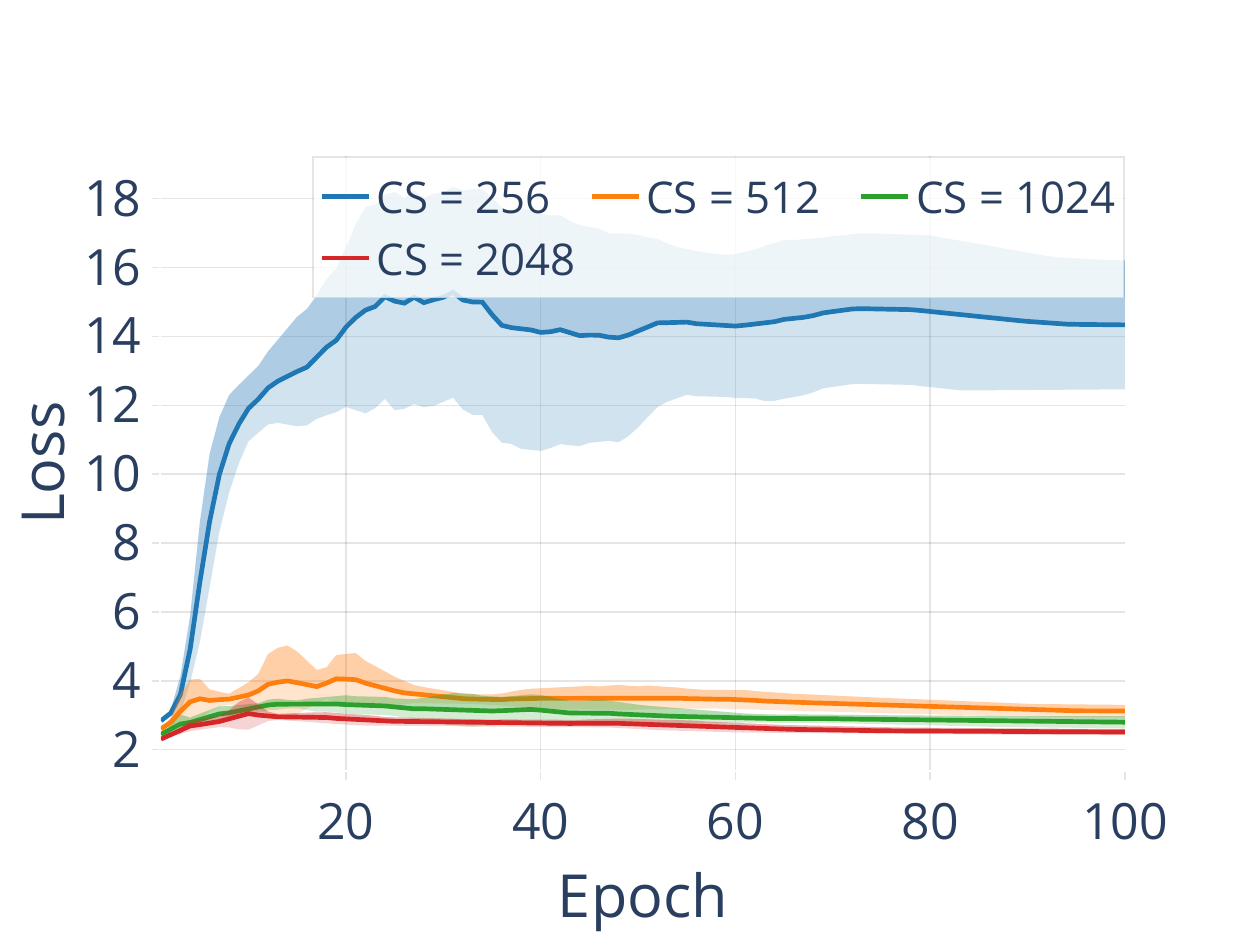}
    }
    \\
    \subfloat[Llama2-13B, $\ell = 2$]{
        \includegraphics[width=\smallThirdWidth]{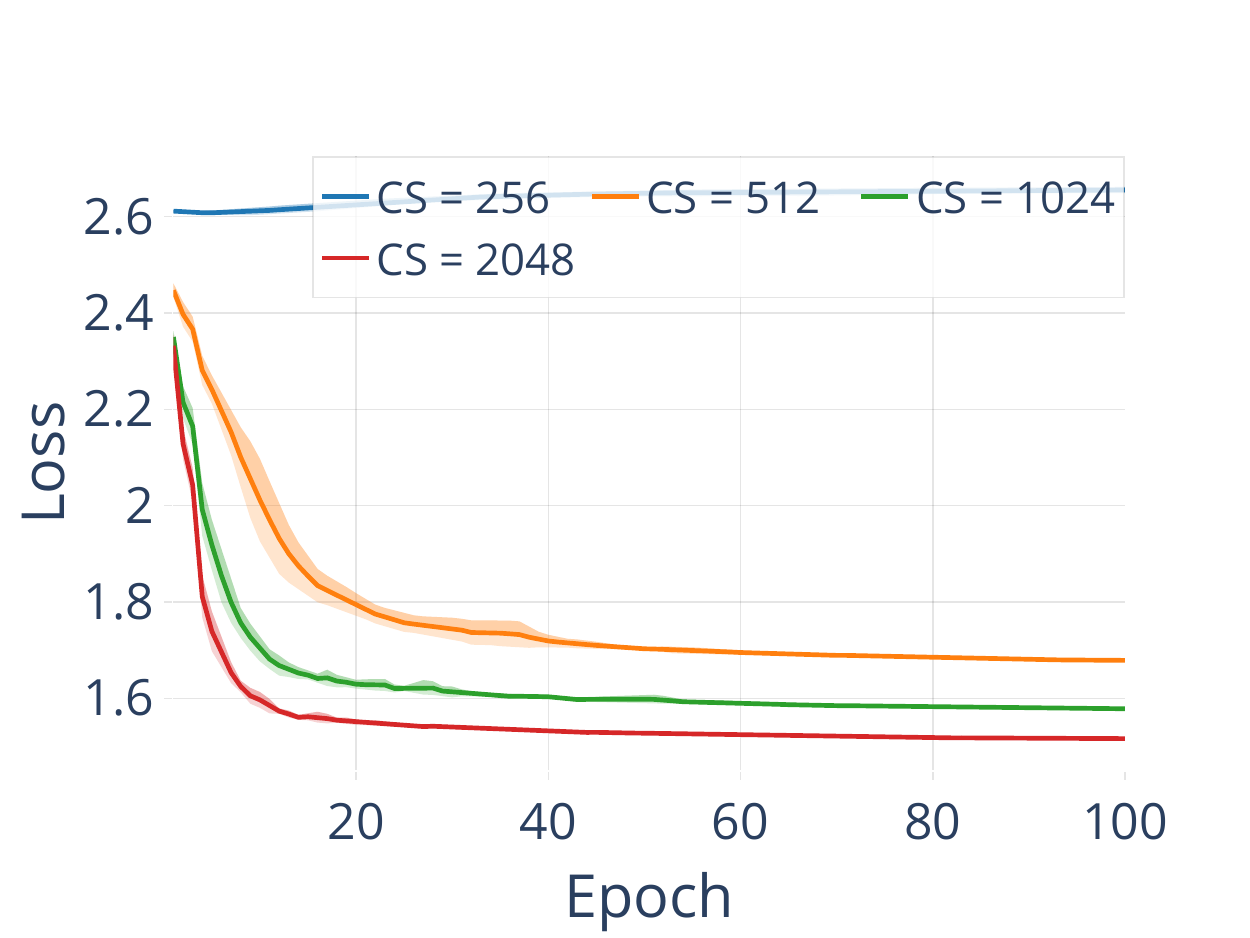}
    }
    \subfloat[Llama2-13B, $\ell = 7$]{
        \includegraphics[width=\smallThirdWidth]{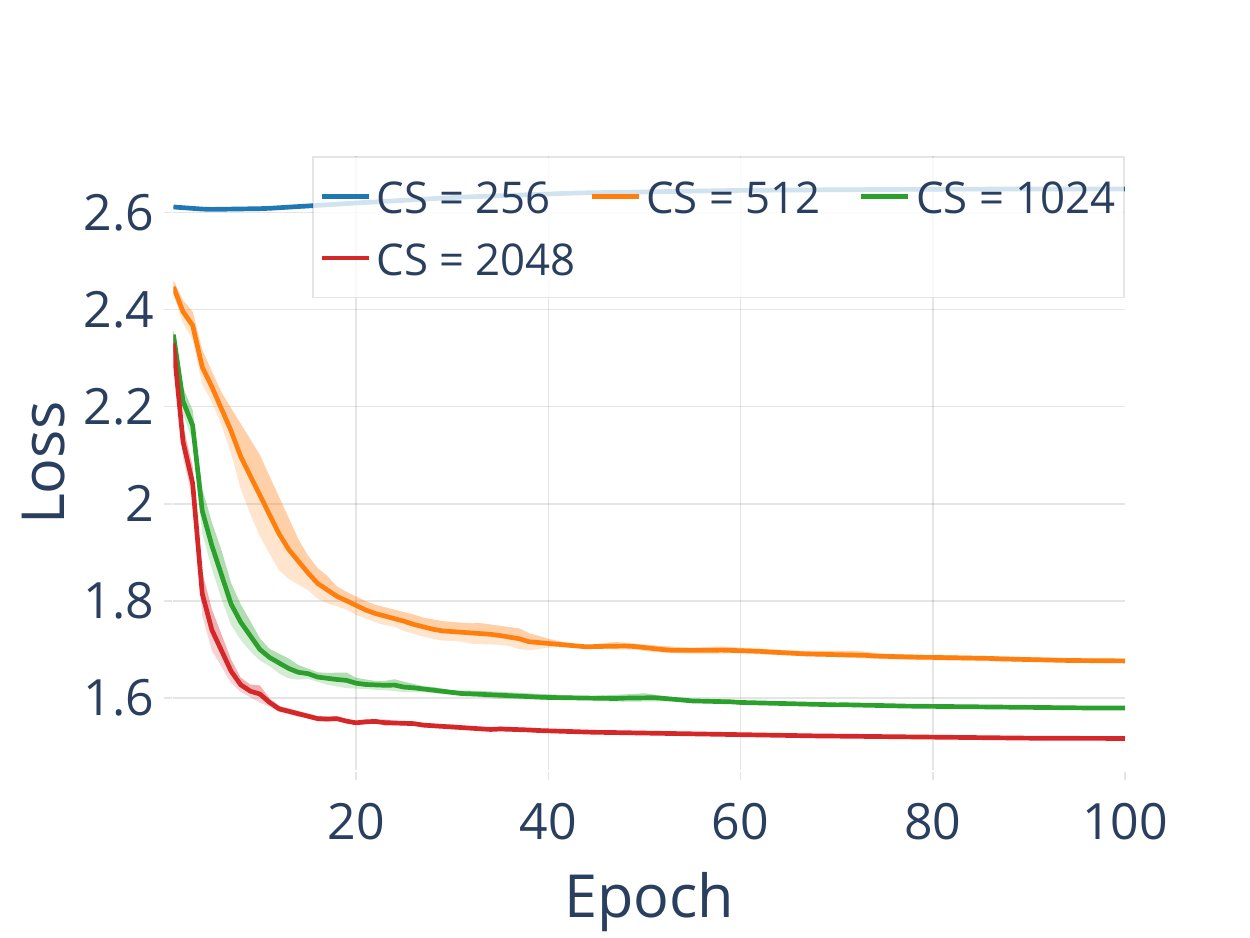}
    }
    \subfloat[Llama2-13B, $\ell = 26$]{
        \includegraphics[width=\smallThirdWidth]{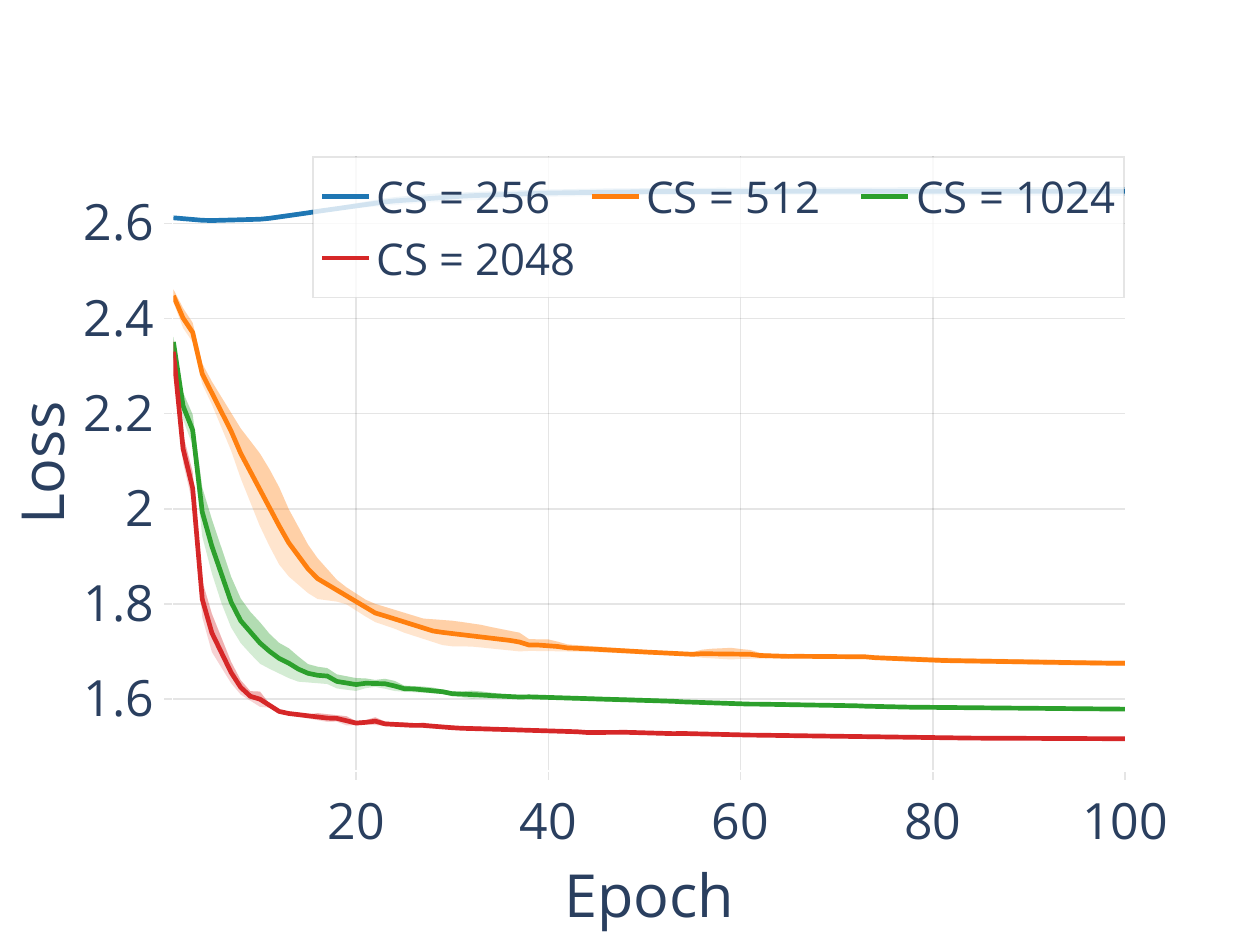}
    }
\caption{\capthead{Loss on a natural-language testset during memorisation, when embedding random strings inside of longer strings of natural language data, for multiple $\ell$ and context sizes.}{$n = 256$}
}
\label{fig:perf_impact_cs}
\end{figure}

\textbf{Memorising random strings in isolation}:
The blue curves in Figure~\ref{fig:perf_impact_bs} ($BS = 1$) and Figure~\ref{fig:perf_impact_cs} ($CS = 256$) show the loss of Pythia-1B, Phi-2.7B and Llama2-13B models on the wikitext testset while memorising strings of length $n = 1024$, resp. $n = 256$ with different alphabet sizes ($l = 2, 7, 26$) for 100 epochs.
The results show that memorising a single random string does indeed increase the loss on the testset.
For Pythia-1B and Llama2-13B models, the loss increase is quite moderate, from $\sim 2.7$ to $\sim 2.9$ for Pythia-1B, and from $\sim 2.4$ to less than $\sim 2.5$ for Llama2-13B.
The loss increases quite significantly from $\sim 2.4$ to $15 - 30$ (depending on the alphabet size) for the Phi-2.7B model.

\textbf{Memorising random strings in the context of natural language}:
The other curves in Figure~\ref{fig:perf_impact_bs} ($BS > 1$) and Figure~\ref{fig:perf_impact_cs} ($CS > 256$) show the loss on the wikitext testset when the model is trained on batches of (non-repeated) text of size $BS$ from the wikitext dataset where one of the elements is the (repeated) random string, and on strings from wikitext with random substrings of length $256$, respectively.
When we show the models wikitext data in addition to the random string, their loss on the testset actually decreases.
For Pythia-1B from $\sim 2.7$ to $\sim 2.45$, for Phi-2.7B from $\sim 2.4$ to $\sim 2.35$, and for Llama2-13B from $\sim 2.4$ to $\sim 1.5$.
At the same time we still observe the same memorisation dynamics, as discussed in Appendices~\ref{app:rw_val_batches} and~\ref{app:rw_val_strings}.

In summary, our results show that memorising a single random string in isolation can increase the model’s loss on natural language text.
However, when embedding random strings into larger natural language contexts, models memorise random strings with the same dynamics as for strings in isolation, while increasing natural language performance.
Therefore, it is likely that the memorisation dynamics we observe also occur in real-world training settings.

\section{Additional details on prefix mappings}
\label{app:mechanics}

\subsection{Additional details on the experimental setup for testing local prefix accuracy}
\label{app:local_prefix_setup}

We test whether a token $s_i$ in string $s$ sampled from distribution $P_A$ can be correctly recalled with a local prefix $s_{[i-k, i-1]}$ of length $k$ for the different replacement strategies {\random} and {\constant}.

To test recall for the \random~strategy, we sample 10 replacements $r_j$ of length $i - k - 1$ for the global context $s_{[1, i - k - 1]}$ from $P_A$.
Then we compute how many times the model correctly predicts token $s_i$ as the token with the highest probability, given the input $r_j \circ s_{[i-k, i-1]}$, \ie~when we randomize all tokens in the input according to $P_A$, other than the local prefix.
If a \emph{plurality} of predictions among the 10 samples match $s_i$, \ie~if $s_i$ is the most frequently predicted token, then we say that the local prefix of size $k$ can correctly recall $s_i$.

For the \constant~strategy we follow a similar process, but instead of sampling all tokens in each $r_j$ randomly from $P_A$, we only sample a single token for each $r_j$, that we use at all positions in $r_j$.

Since finding local prefixes with multiple samples for different lengths and positions in the string requires a large number of inference calls, for the 1024 tokens strings, we randomly subsample 256 positions and compute the performance of their local prefixes of different lengths.

\subsection{Additional results on local prefixes}
\label{app:local_prefix_results}

\begin{figure}[H]
    \centering
    \subfloat[Pythia-1B, $\ell = 2$]{
        \includegraphics[width=\thirdWidth]{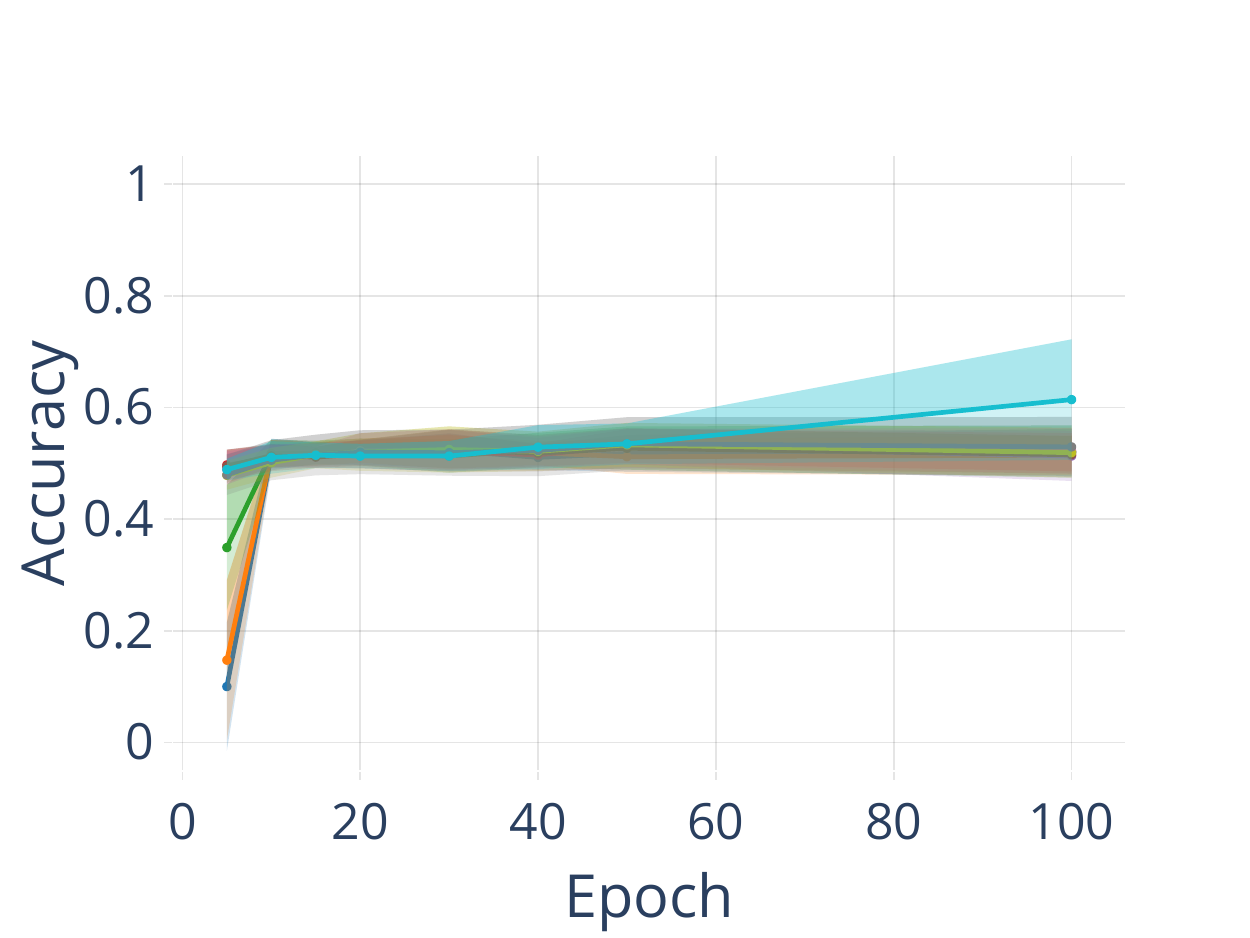}
    }
    \subfloat[Phi-2.7B, $\ell = 2$]{
        \includegraphics[width=\thirdWidth]{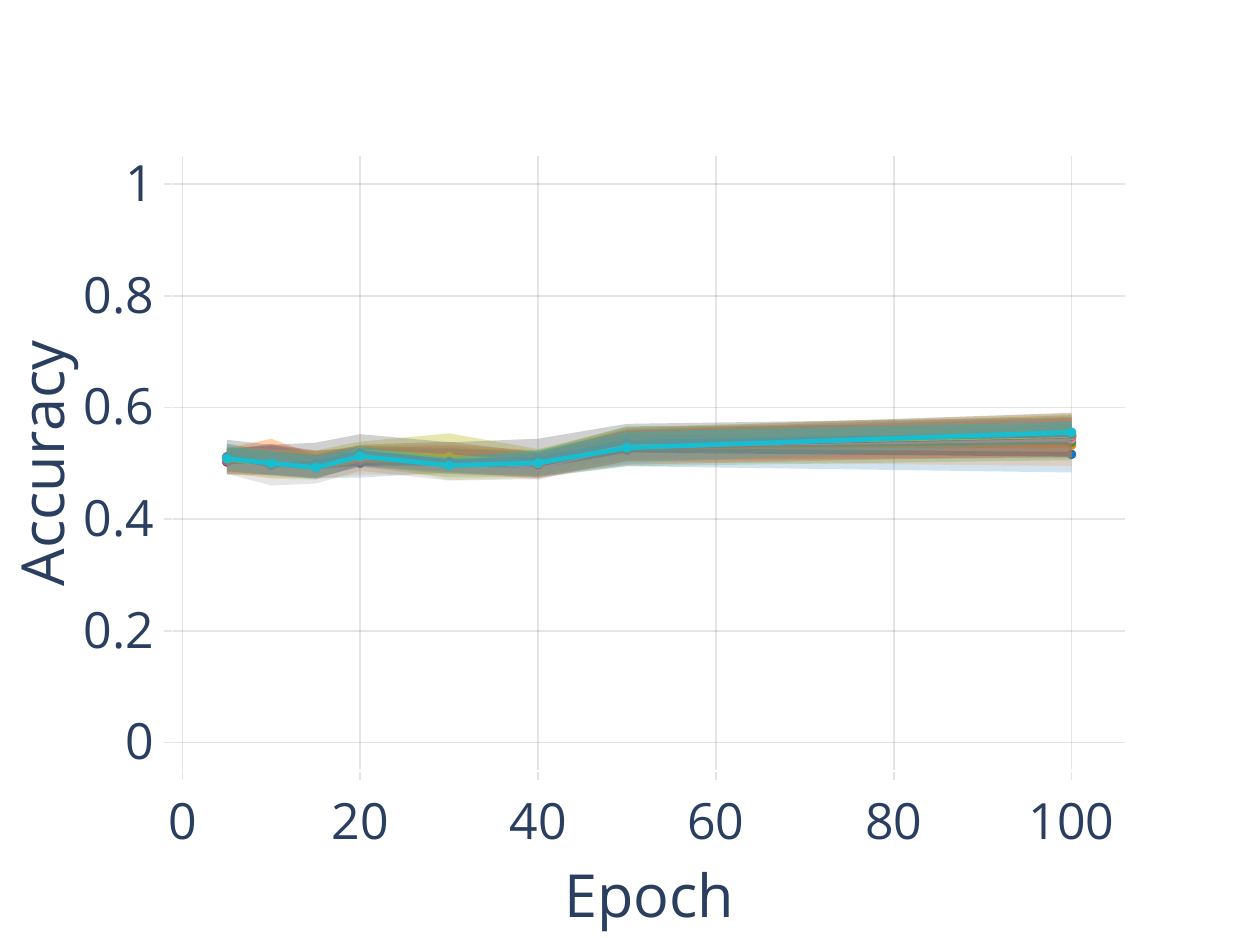}
    }
    \subfloat[Llama2-13B, $\ell = 2$]{
        \includegraphics[width=\thirdWidth]{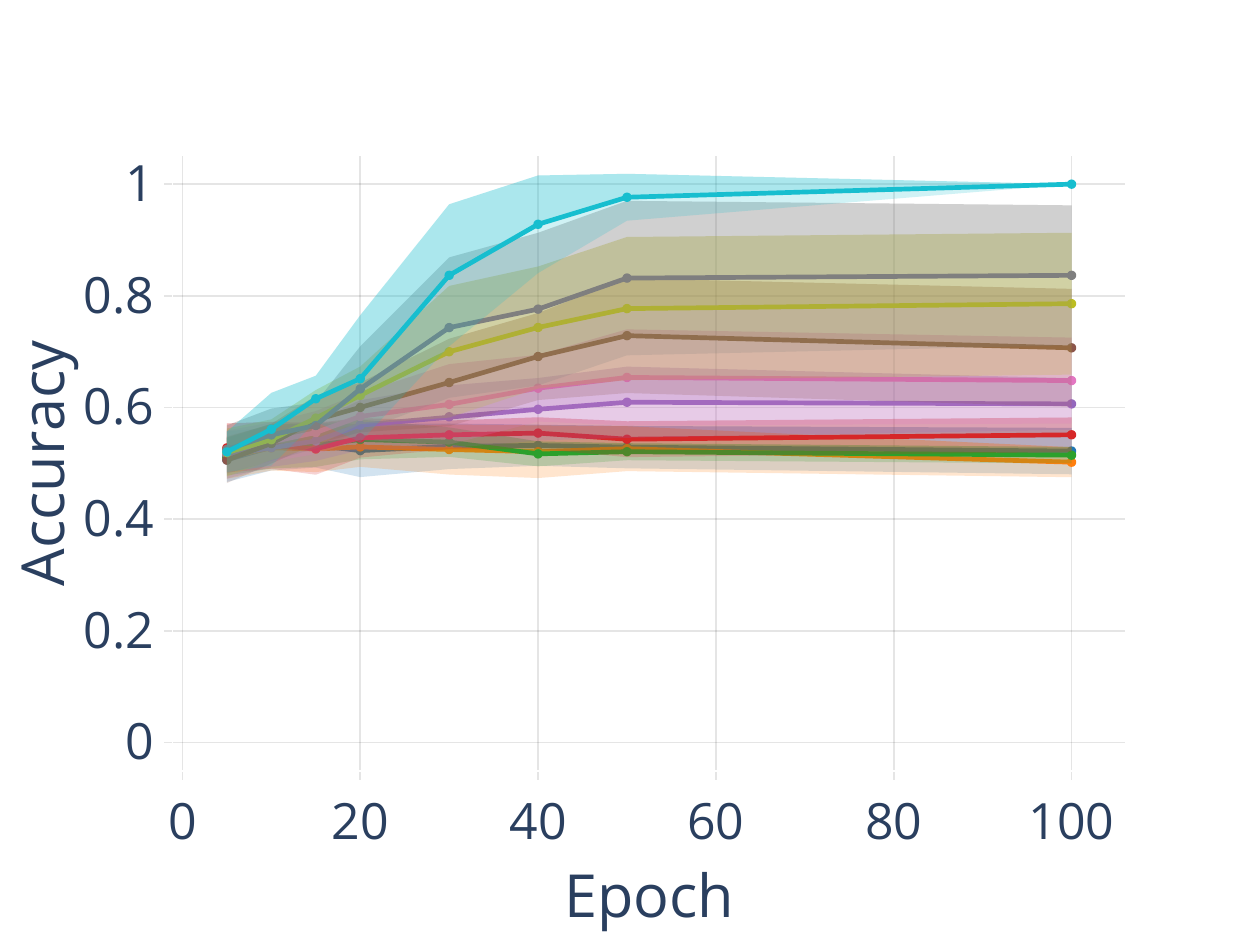}
    }
    \\
    \subfloat[Pythia-1B, $\ell = 4$]{
        \includegraphics[width=\thirdWidth]{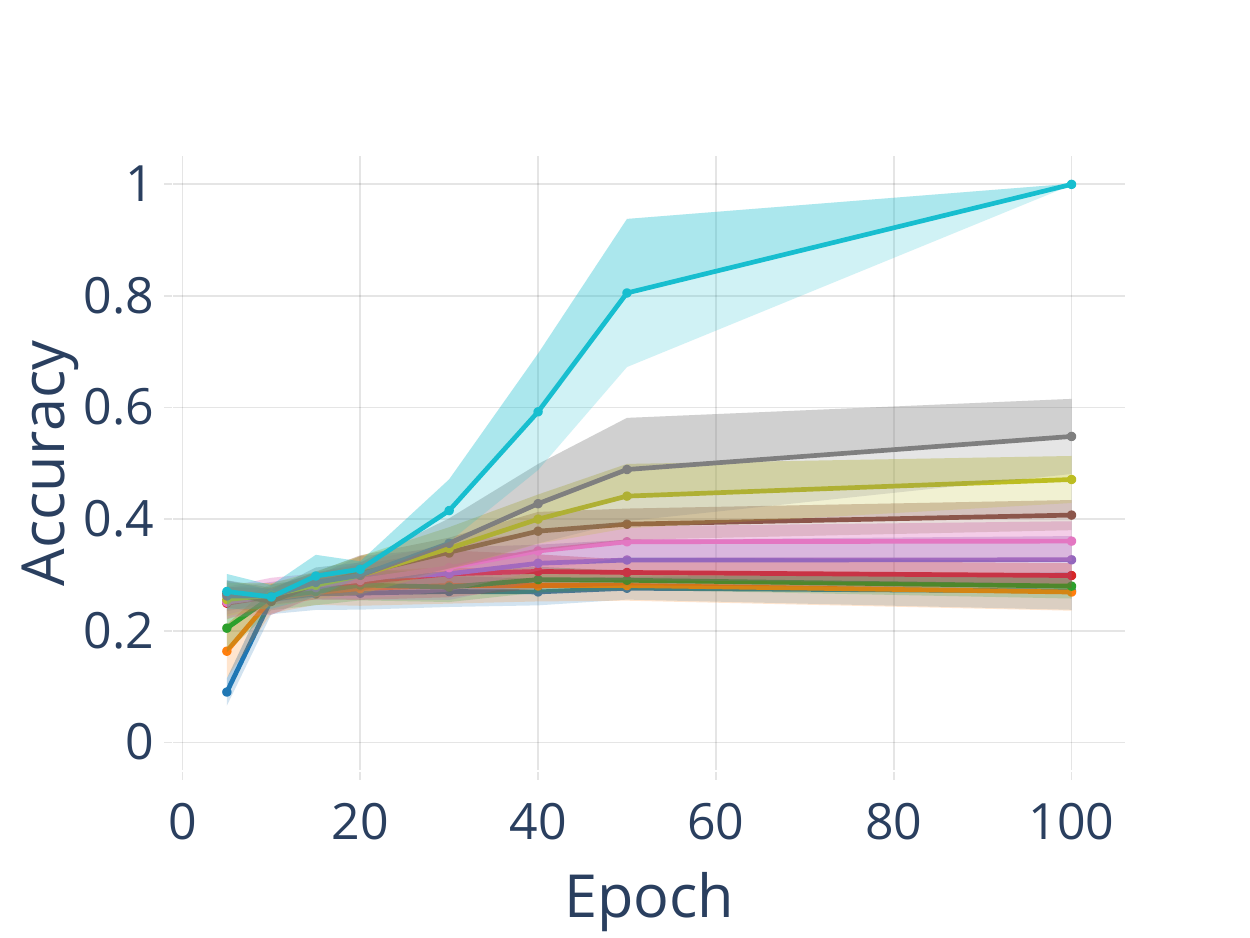}
    }
    \subfloat[Phi-2.7B, $\ell = 4$]{
        \includegraphics[width=\thirdWidth]{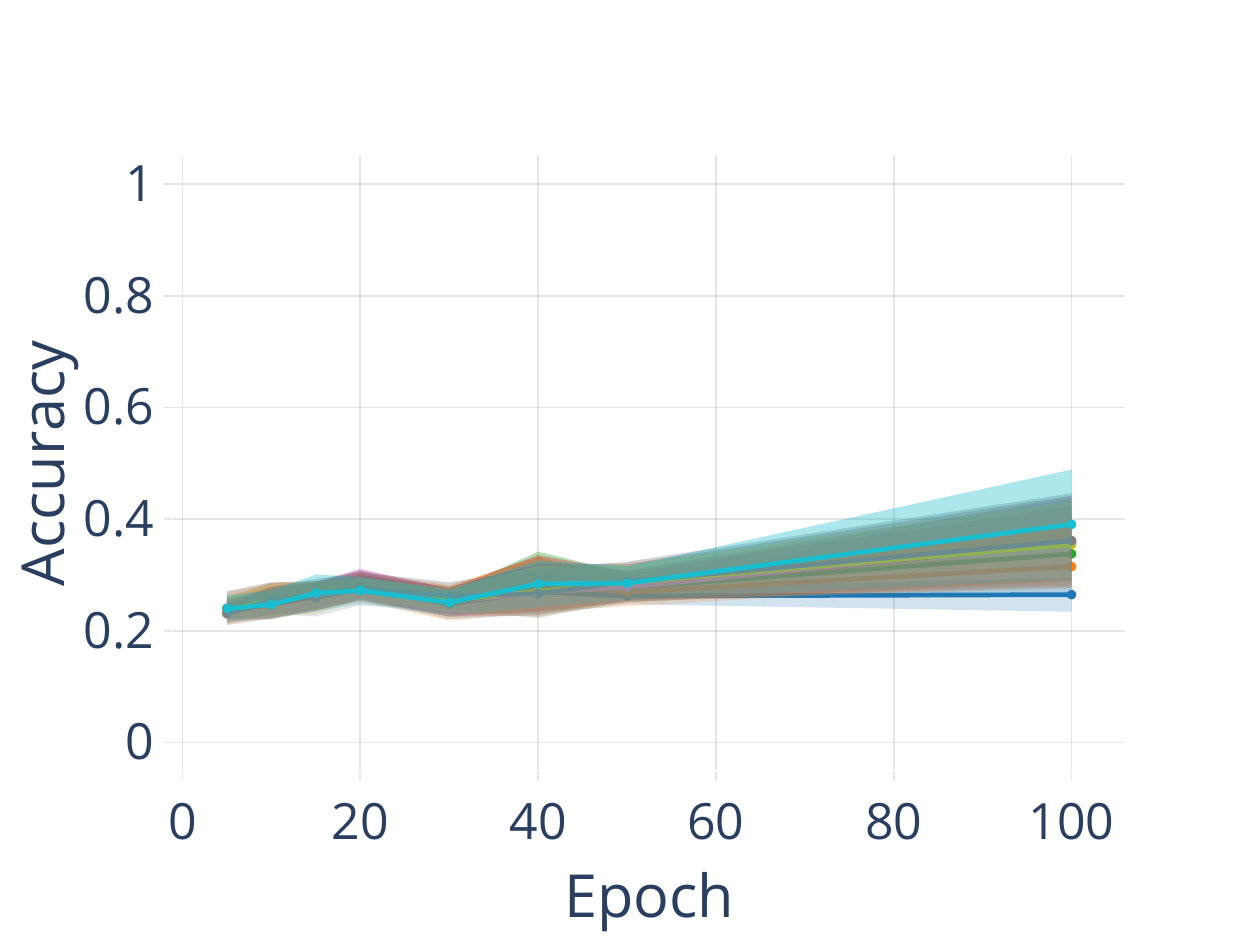}
    }
    \subfloat[Llama2-13B, $\ell = 4$]{
        \includegraphics[width=\thirdWidth]{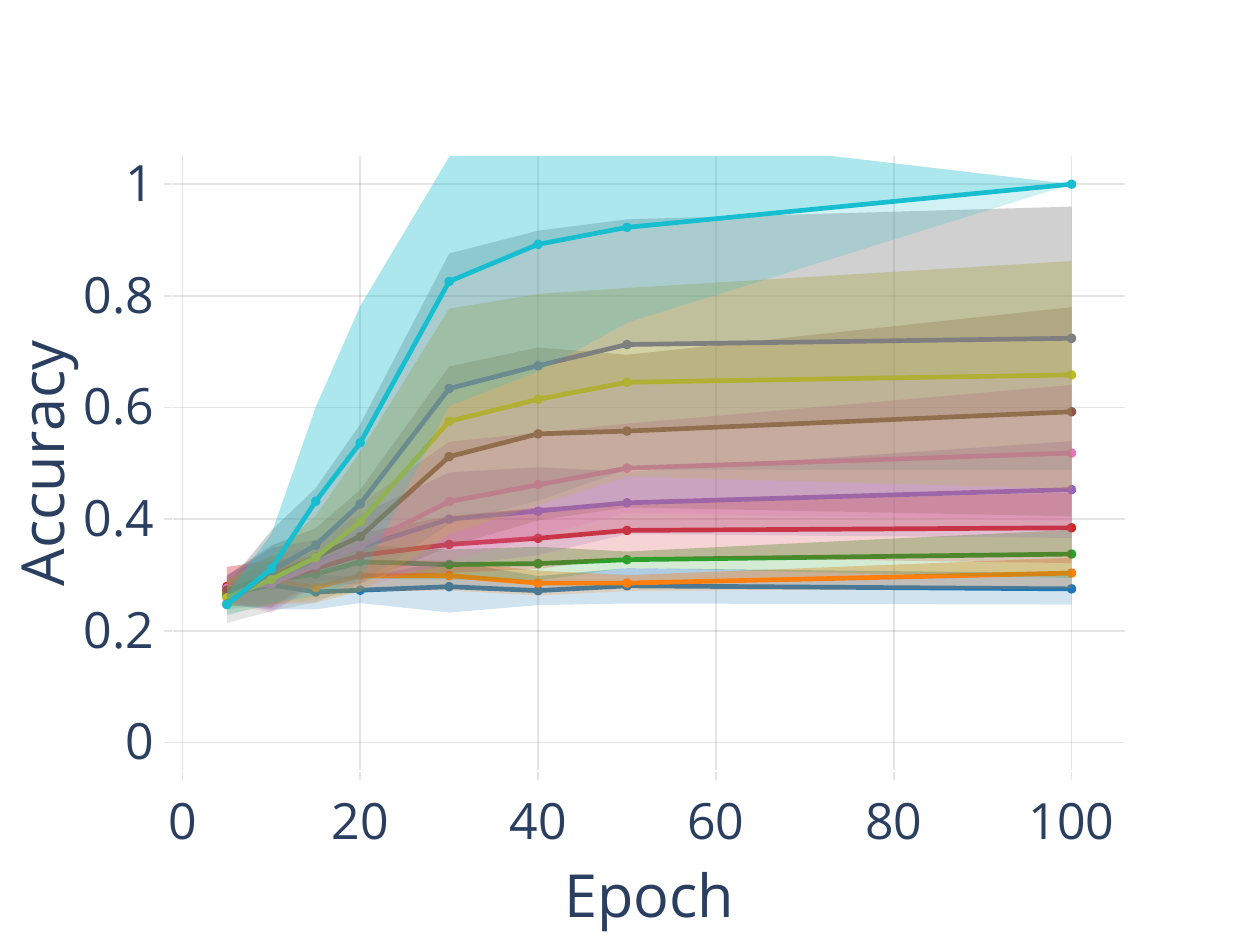}
    }
    \\
    \subfloat[Pythia-1B, $\ell = 7$]{
        \includegraphics[width=\thirdWidth]{figures/prefix_mappings/prefix_lengths/pretrained/epochs_alphabet-size_a-7_pythia-1b.pdf}
    }
    \subfloat[Phi-2.7B, $\ell = 7$]{
        \includegraphics[width=\thirdWidth]{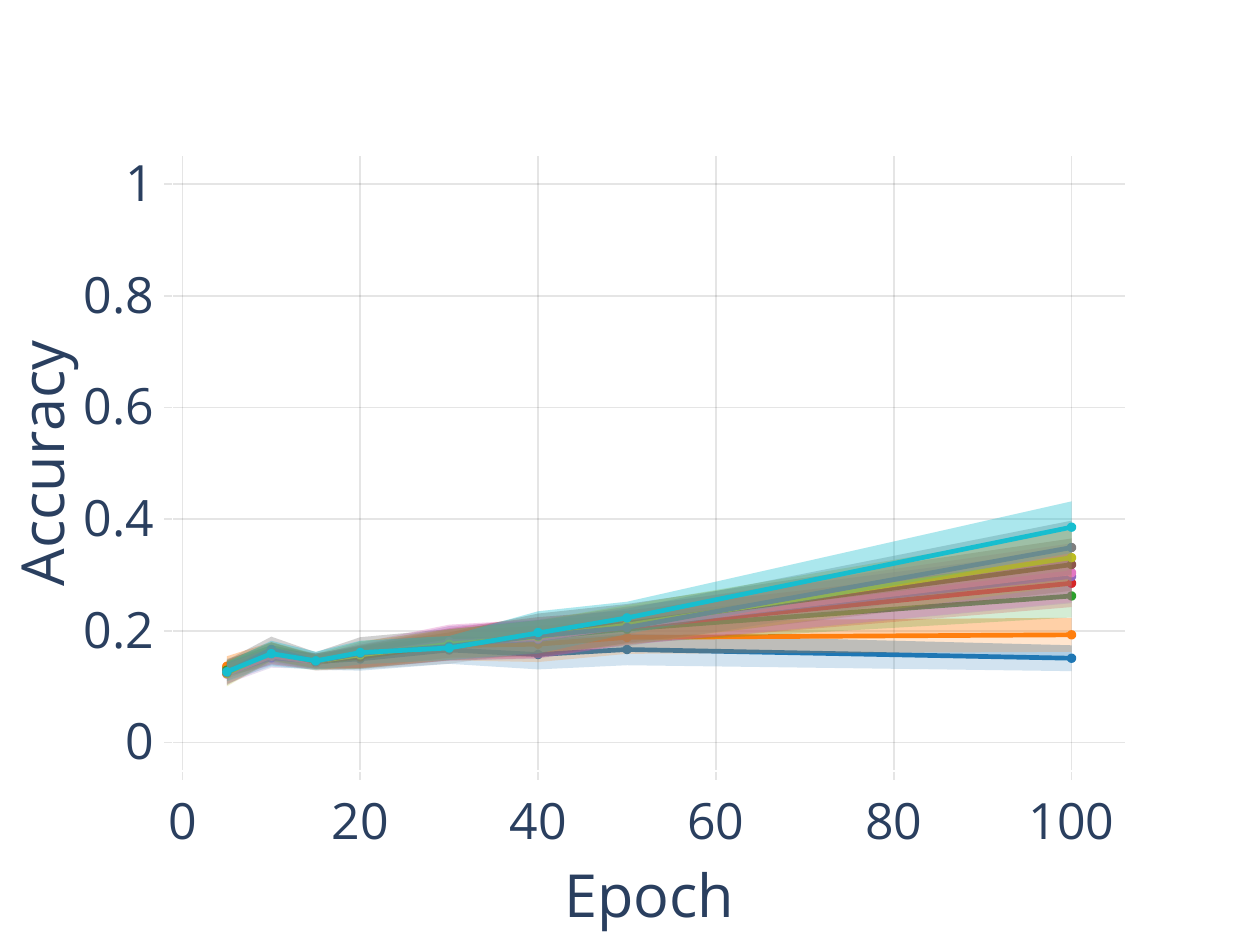}
    }
    \subfloat[Llama2-13B, $\ell = 7$]{
        \includegraphics[width=\thirdWidth]{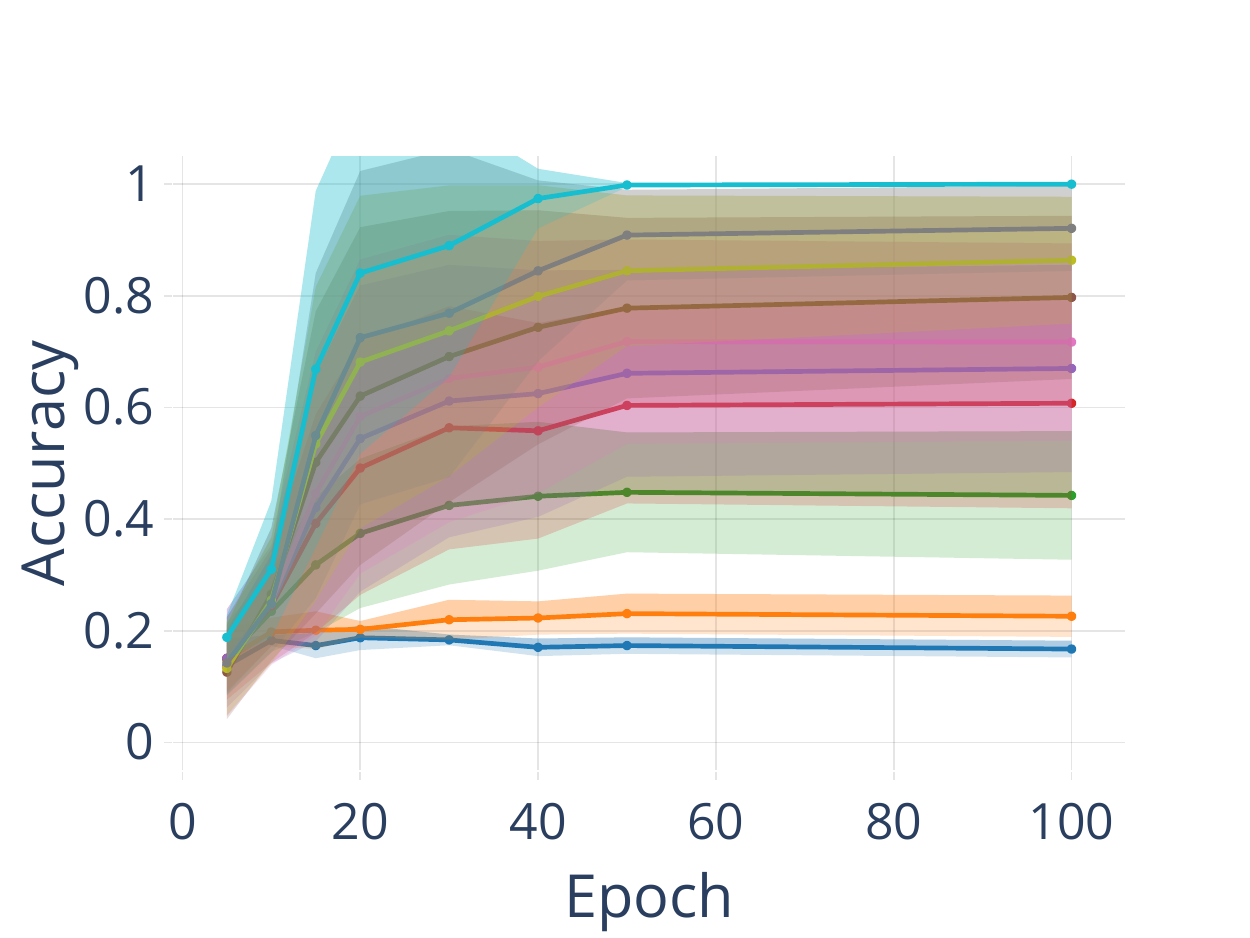}
    }
    \\
    \subfloat[Pythia-1B, $\ell = 13$]{
        \includegraphics[width=\thirdWidth]{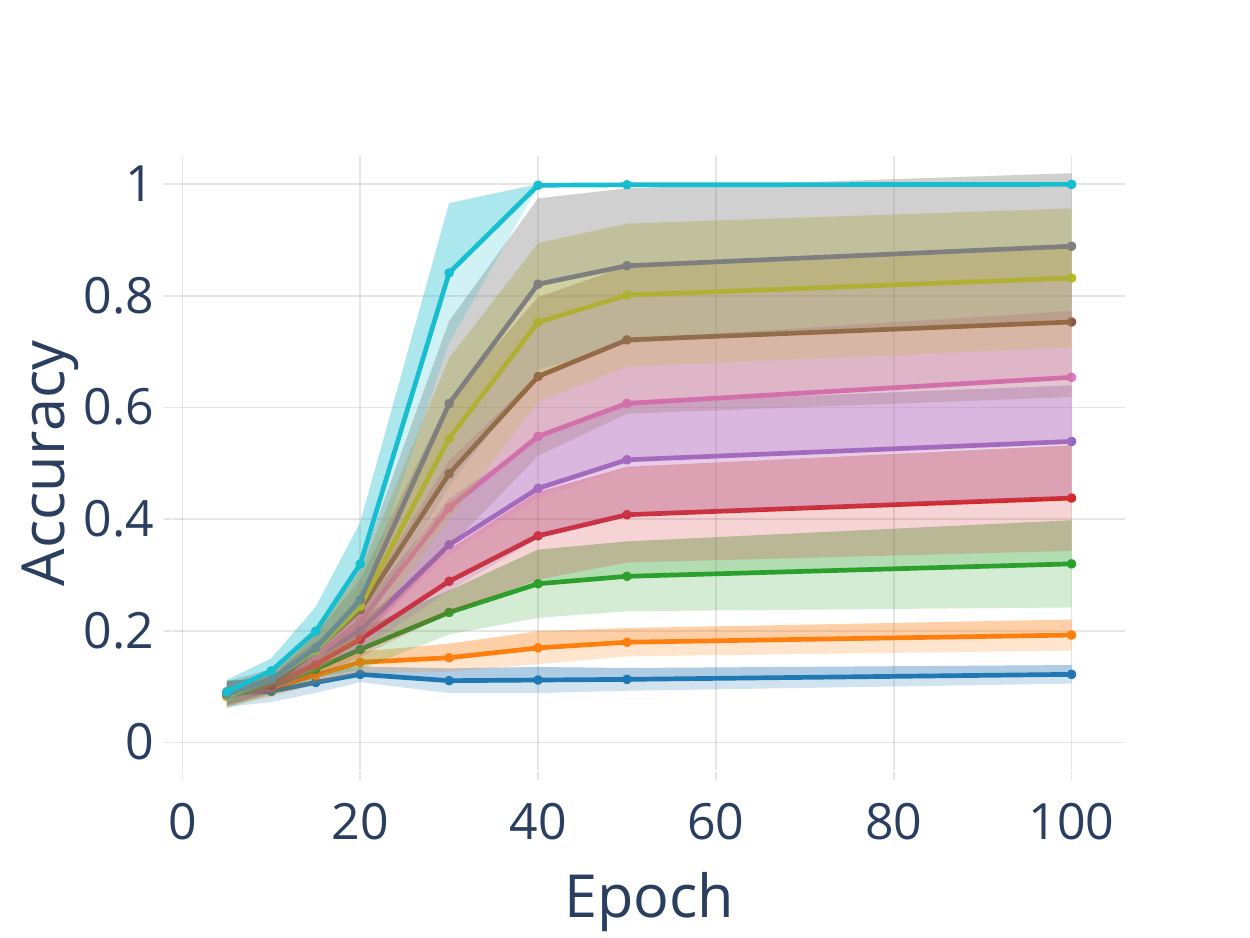}
    }
    \subfloat[Phi-2.7B, $\ell = 13$]{
        \includegraphics[width=\thirdWidth]{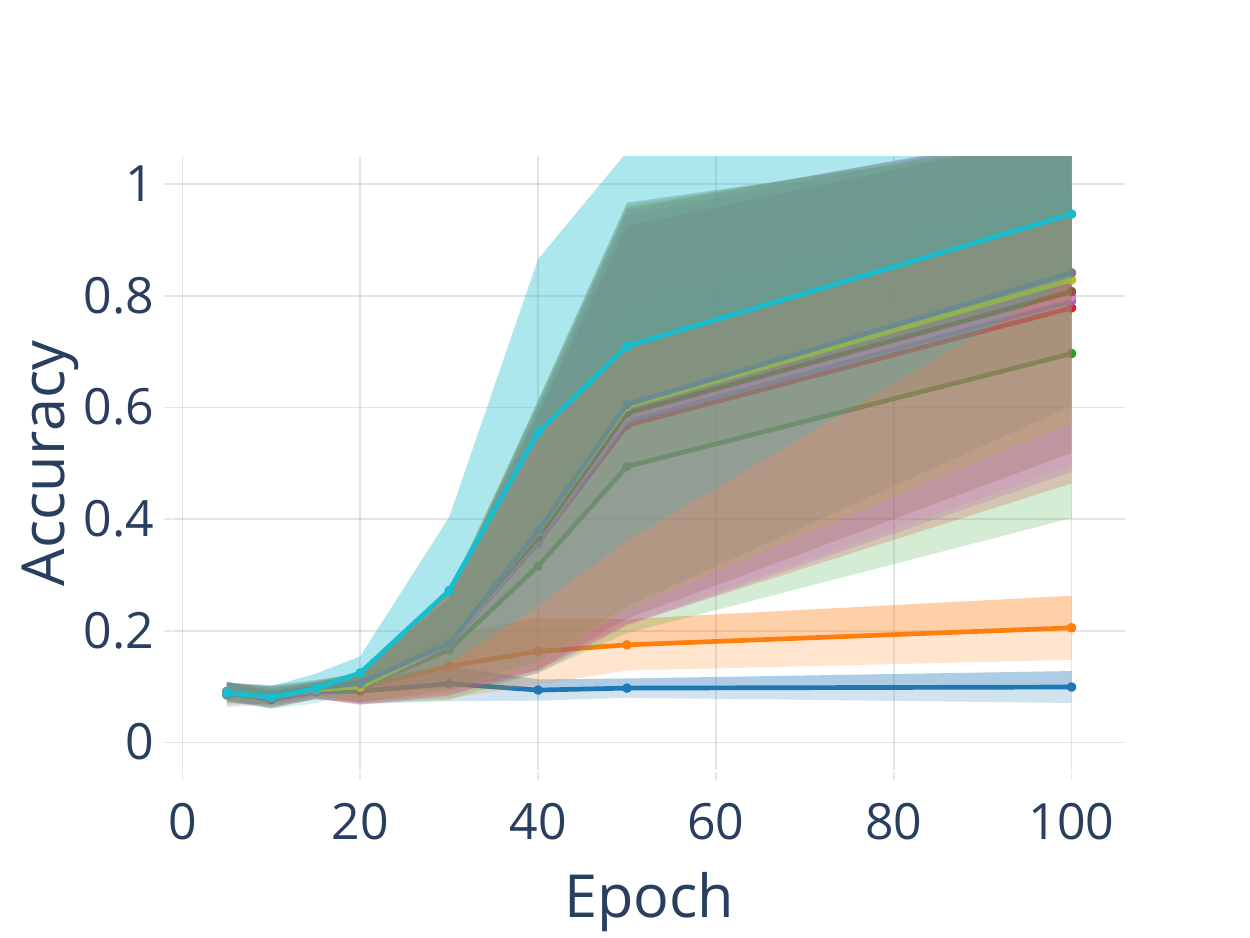}
    }
    \subfloat[Llama2-13B, $\ell = 13$]{
        \includegraphics[width=\thirdWidth]{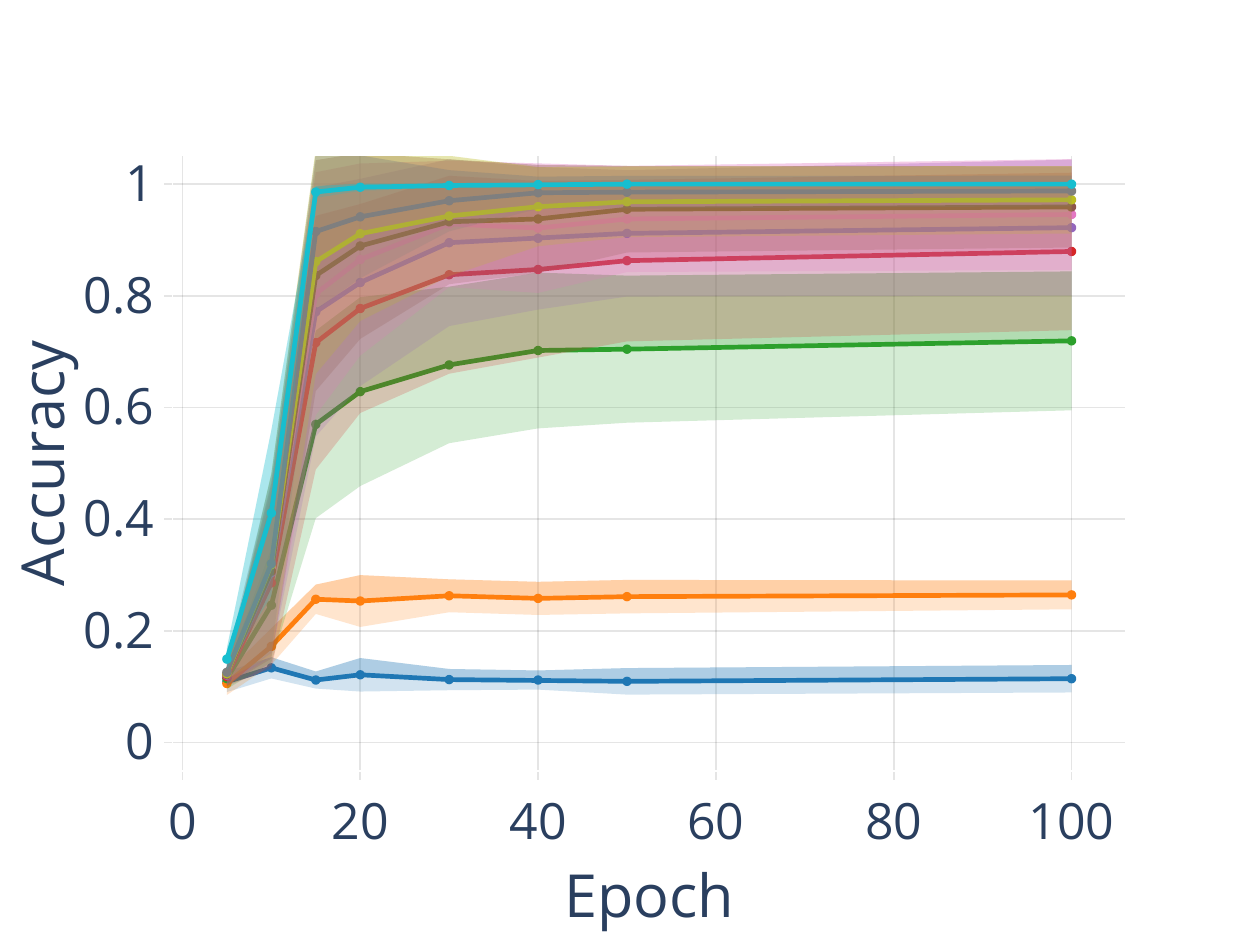}
    }
    \\
    \subfloat[Pythia-1B, $\ell = 26$]{
        \includegraphics[width=\thirdWidth]{figures/prefix_mappings/prefix_lengths/pretrained/epochs_alphabet-size_a-26_pythia-1b.pdf}
    }
    \subfloat[Phi-2.7B, $\ell = 26$]{
        \includegraphics[width=\thirdWidth]{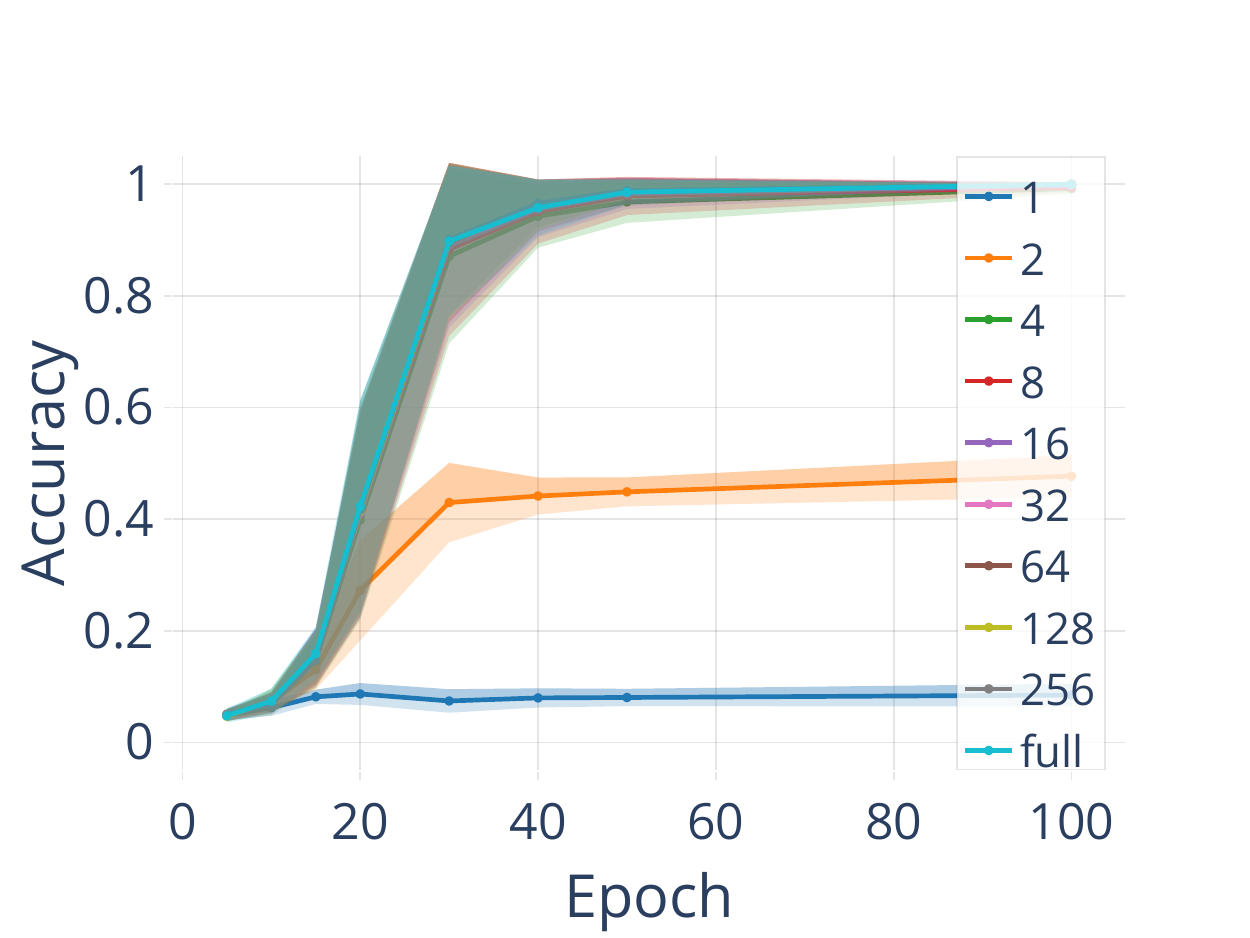}
    }
    \subfloat[Llama2-13B, $\ell = 26$]{
        \includegraphics[width=\thirdWidth]{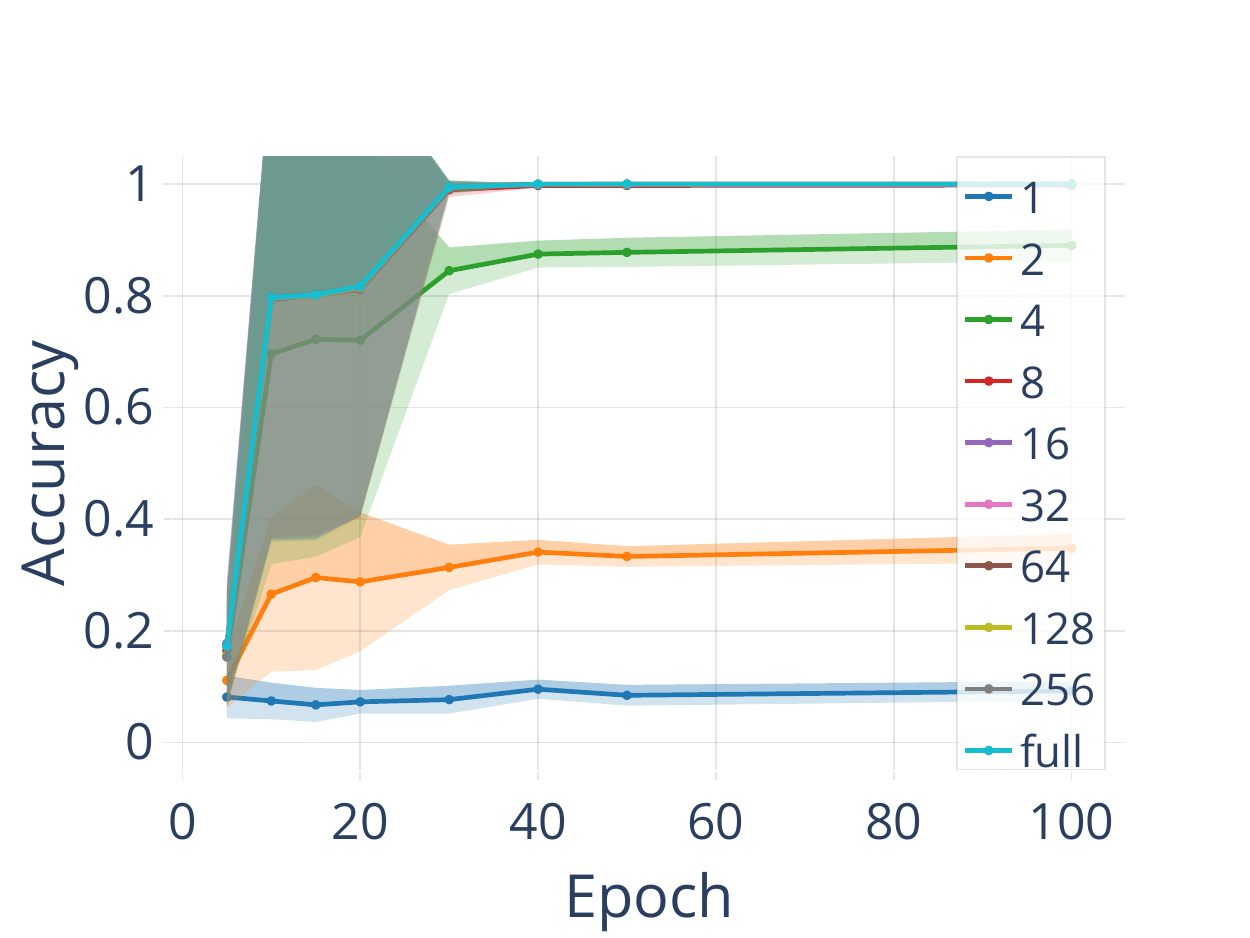}
    }
    \\
\caption{\capthead{Recollection accuracy for \emph{pretrained} models for different prefix lengths during training for different $\ell$'s}{$n = 1024$}
    In most cases, local prefixes, \ie~a small number of tokens immediately preceding the token to predict, much fewer than the full string's length of $n = 1024$ perform well.
}
\label{fig:epoch_prefix_len_all}
\end{figure}

\begin{figure}[H]
    \centering
    \subfloat[Pythia-1B, $\ell = 2$]{
        \includegraphics[width=\thirdWidth]{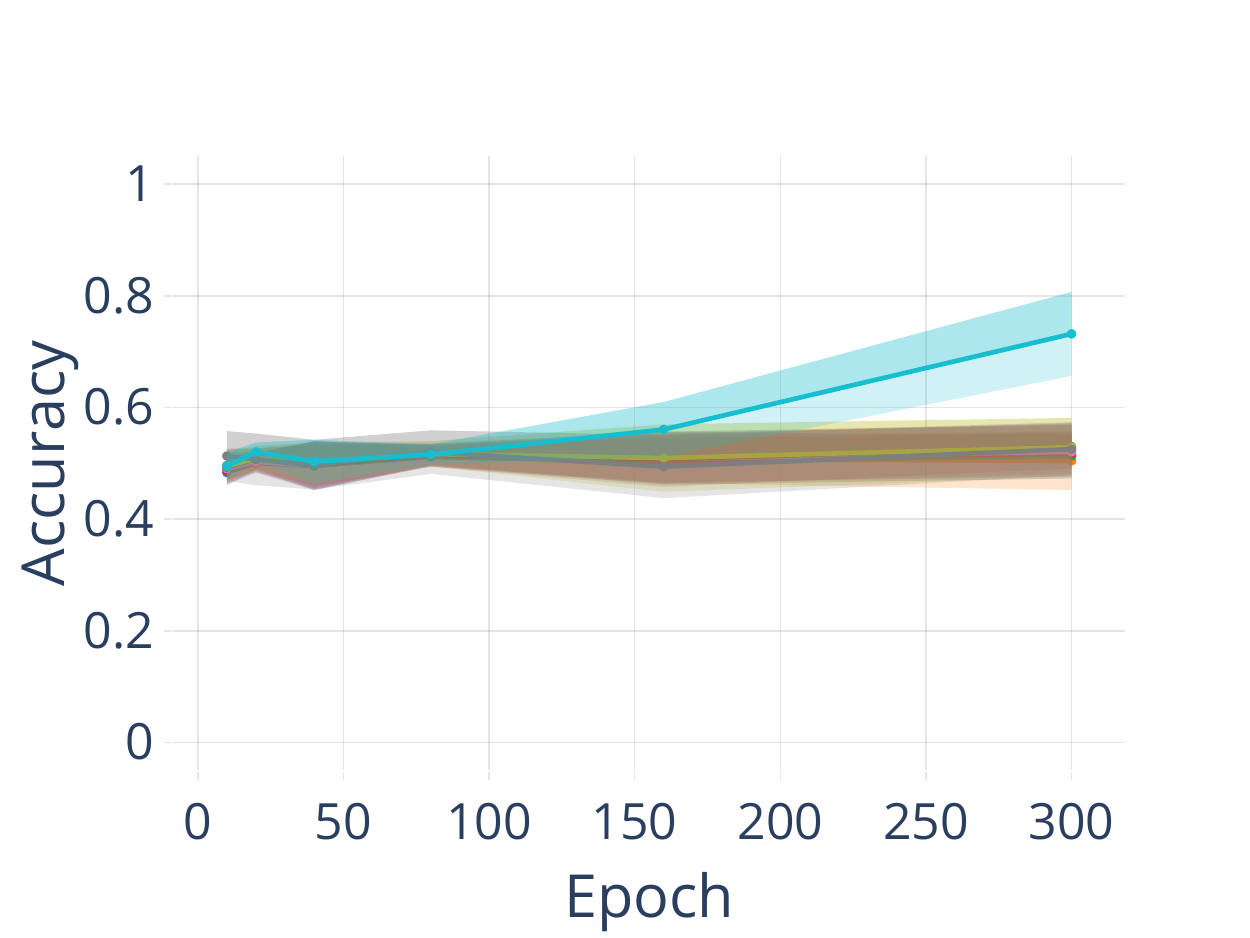}
    }
    \subfloat[Phi-2.7B, $\ell = 2$]{
        \includegraphics[width=\thirdWidth]{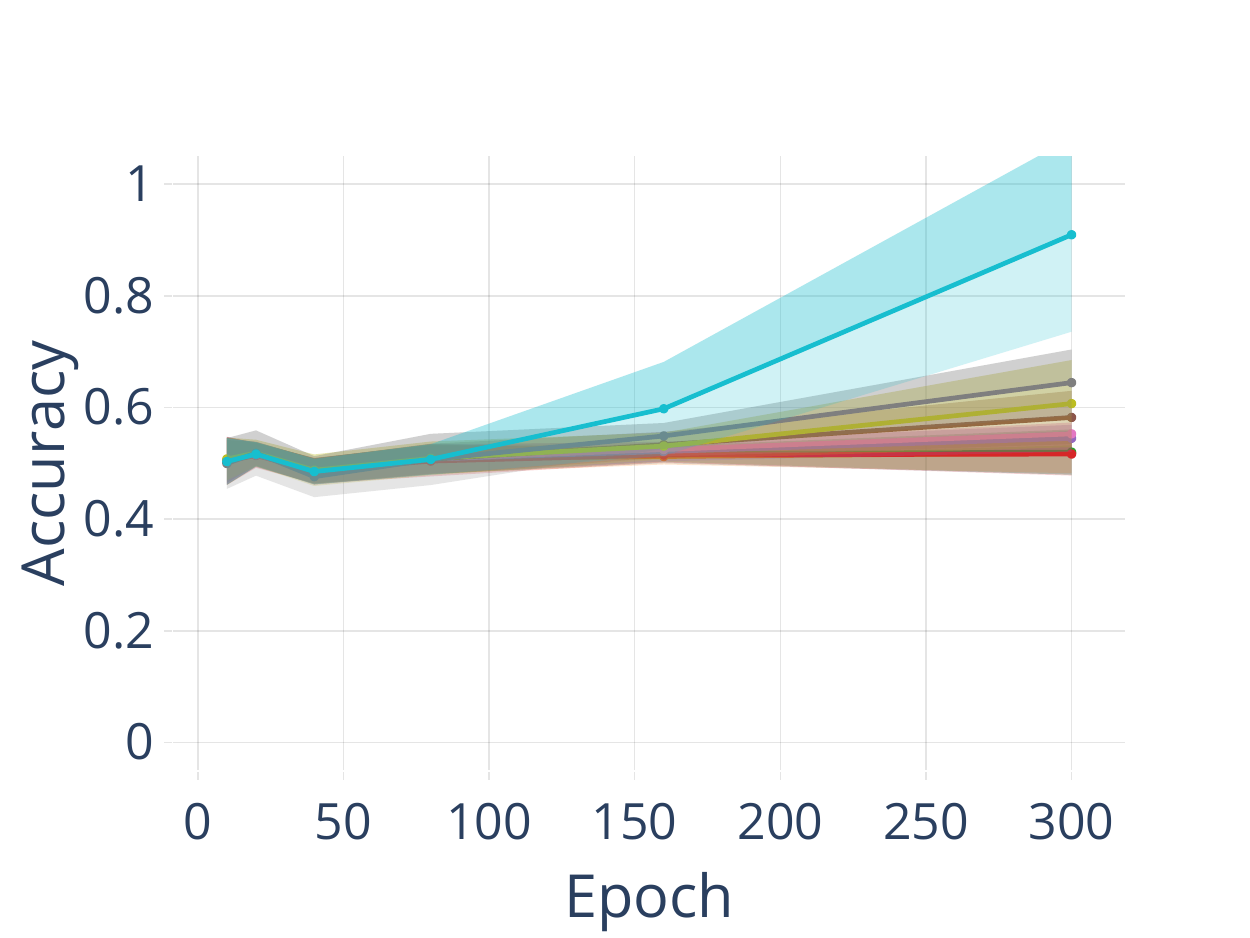}
    }
    \subfloat[Llama2-13B, $\ell = 2$]{
        \includegraphics[width=\thirdWidth]{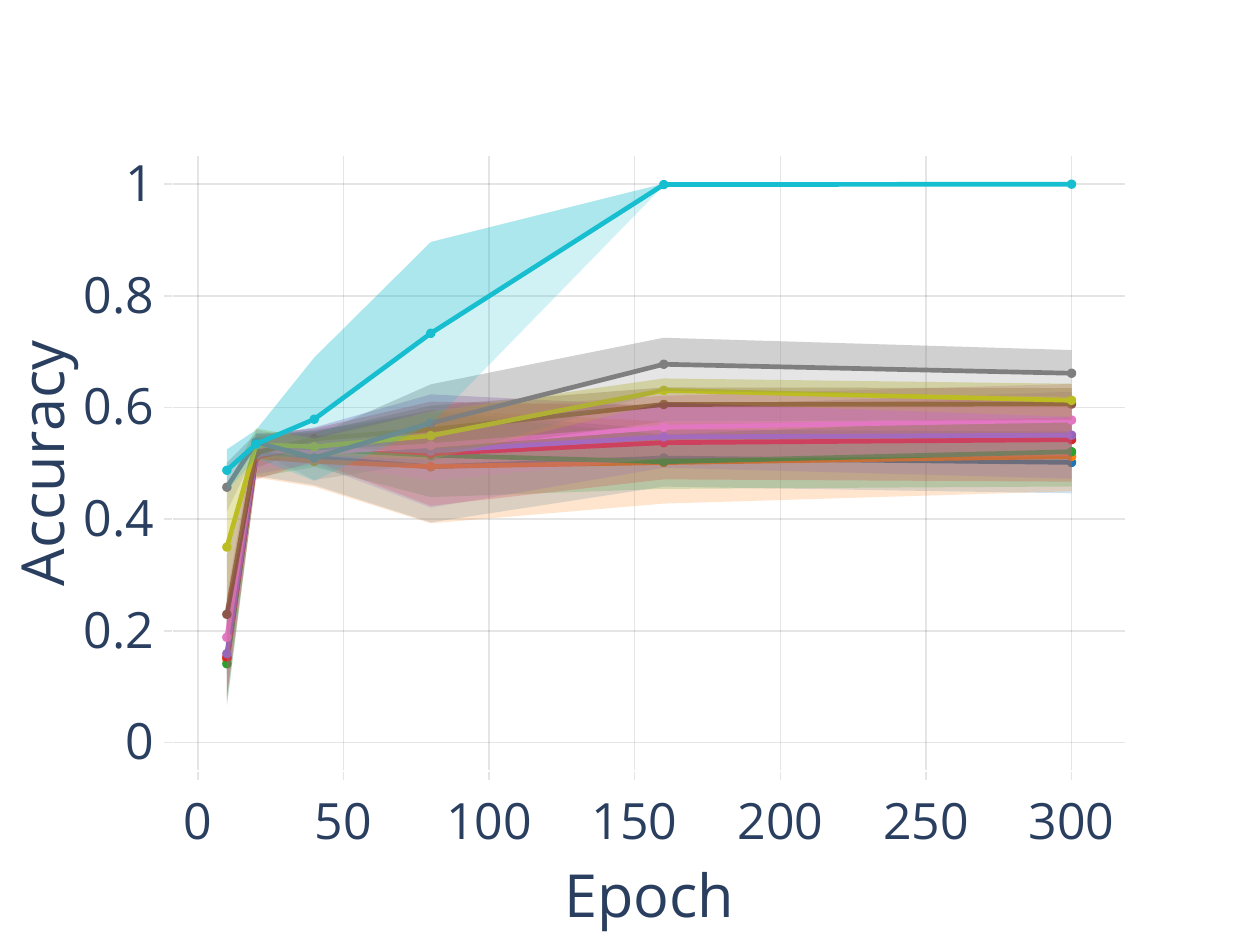}
    }
    \\
    \subfloat[Pythia-1B, $\ell = 7$]{
        \includegraphics[width=\thirdWidth]{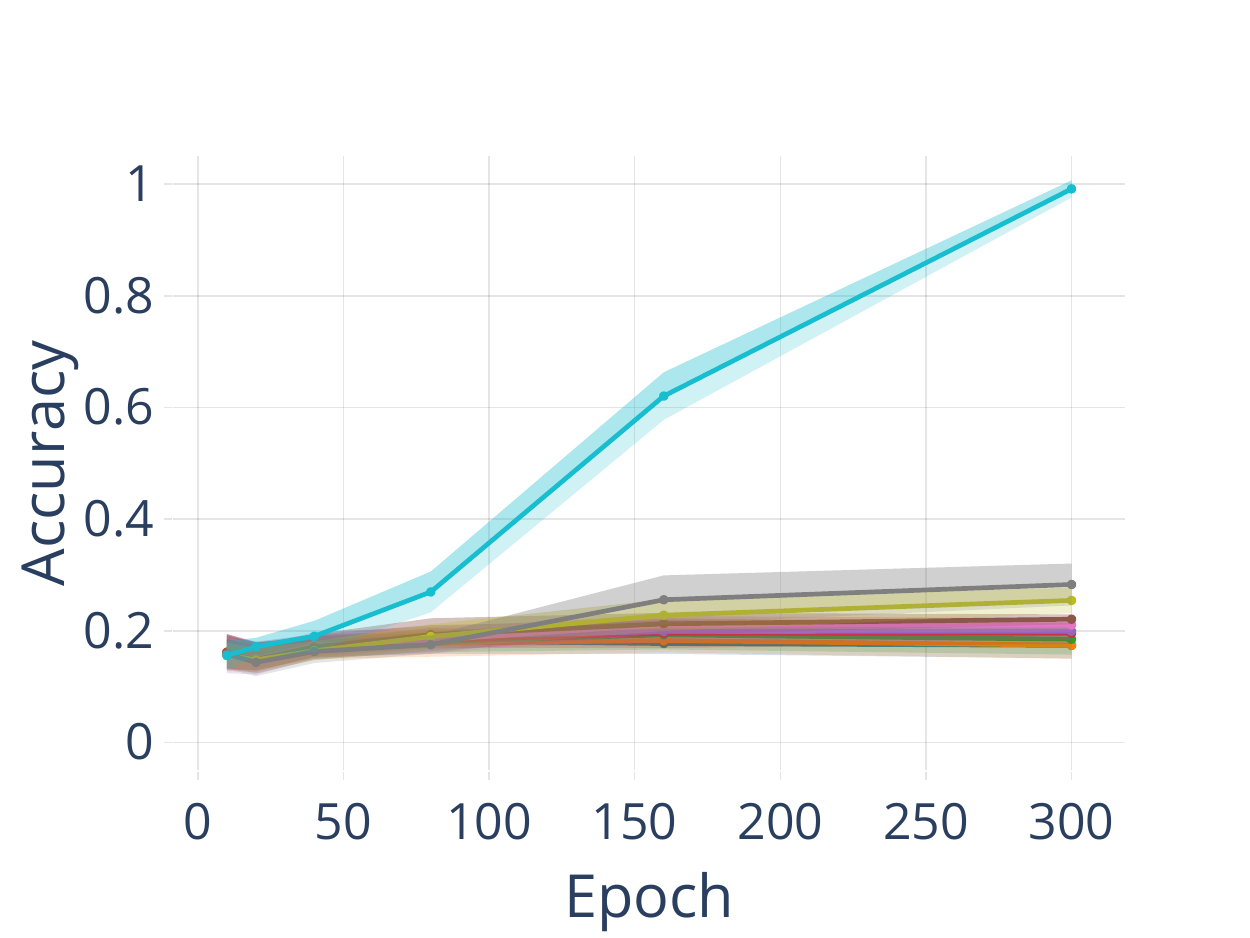}
    }
    \subfloat[Phi-2.7B, $\ell = 7$]{
        \includegraphics[width=\thirdWidth]{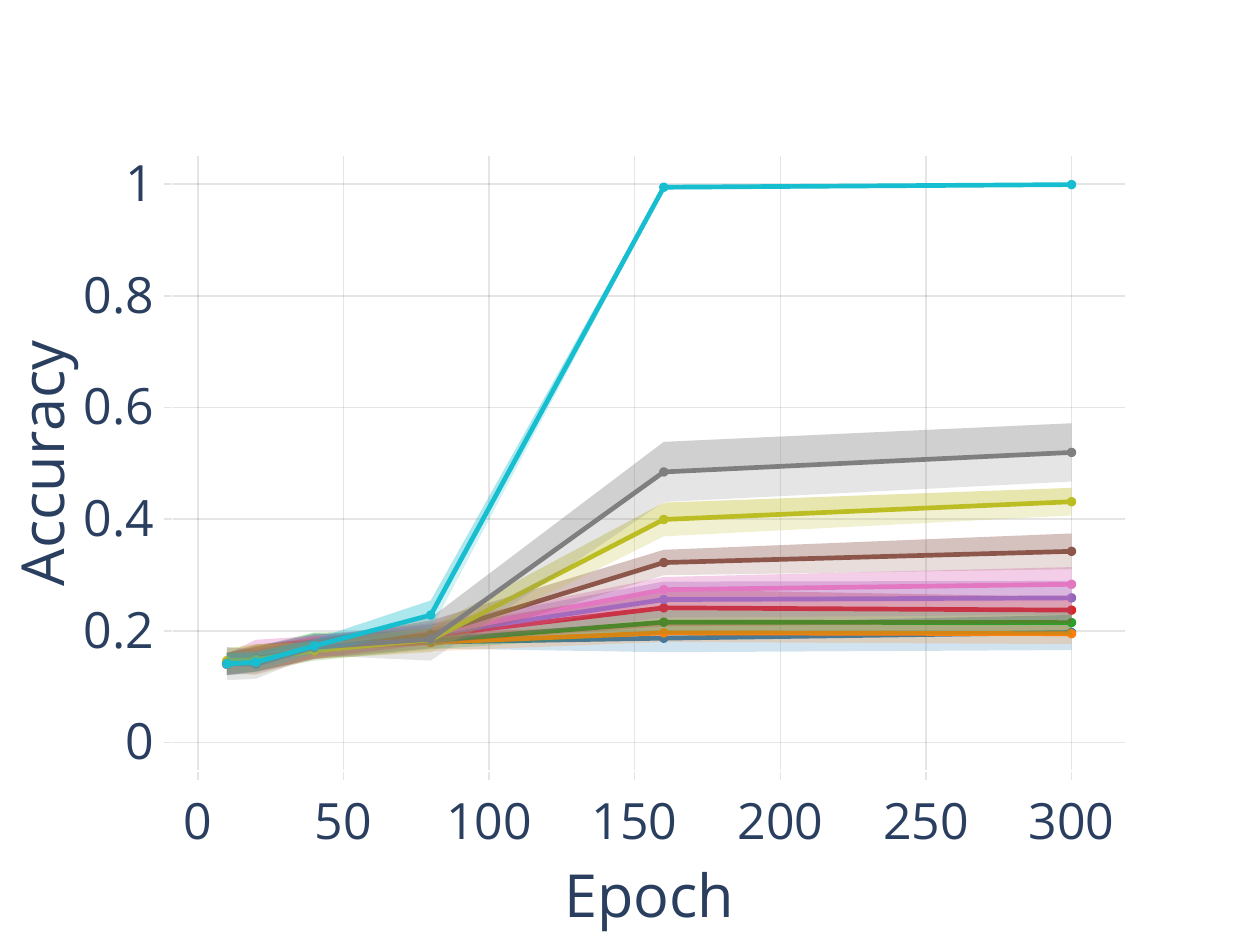}
    }
    \subfloat[Llama2-13B, $\ell = 7$]{
        \includegraphics[width=\thirdWidth]{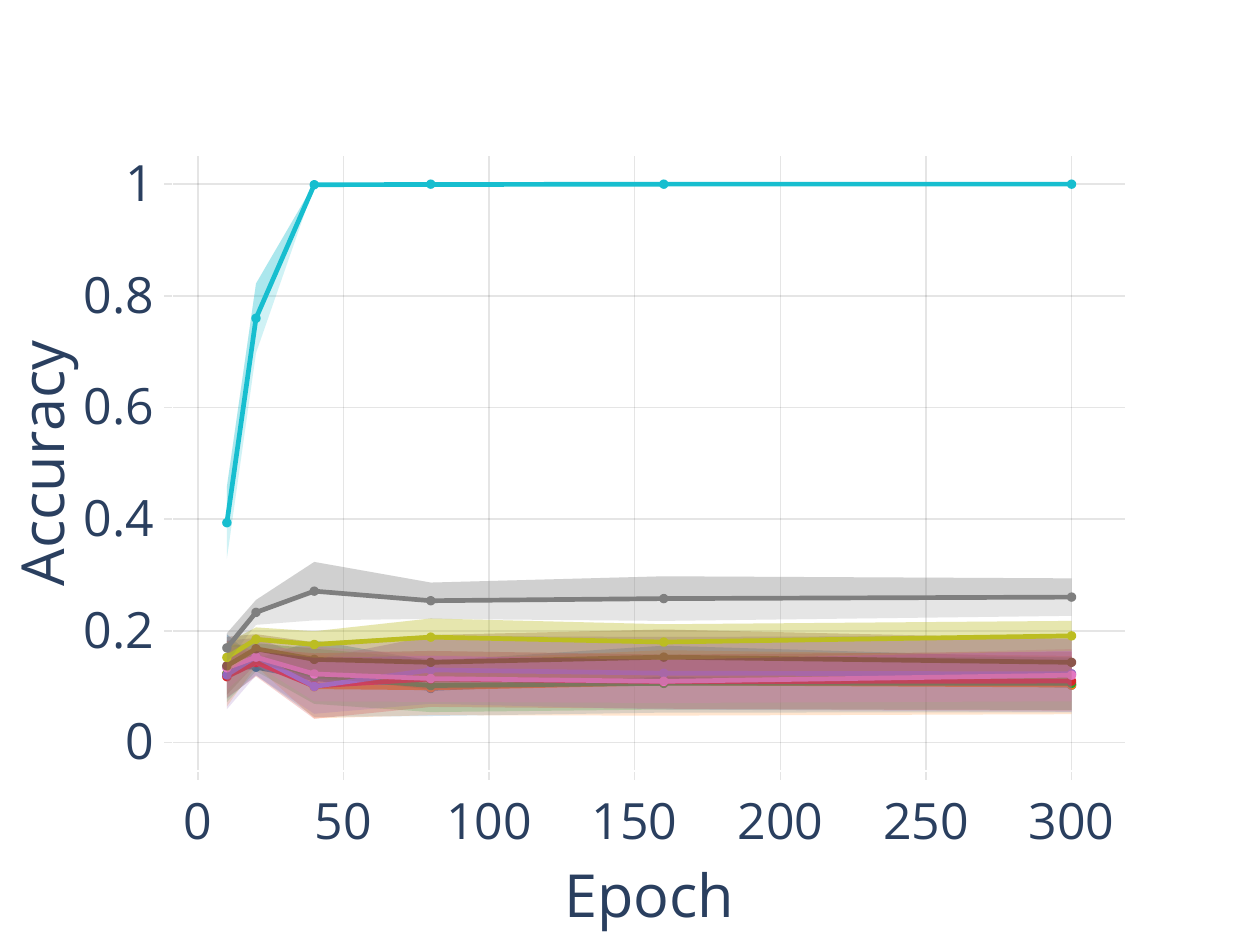}
    }
    \\
    \subfloat[Pythia-1B, $\ell = 26$]{
        \includegraphics[width=\thirdWidth]{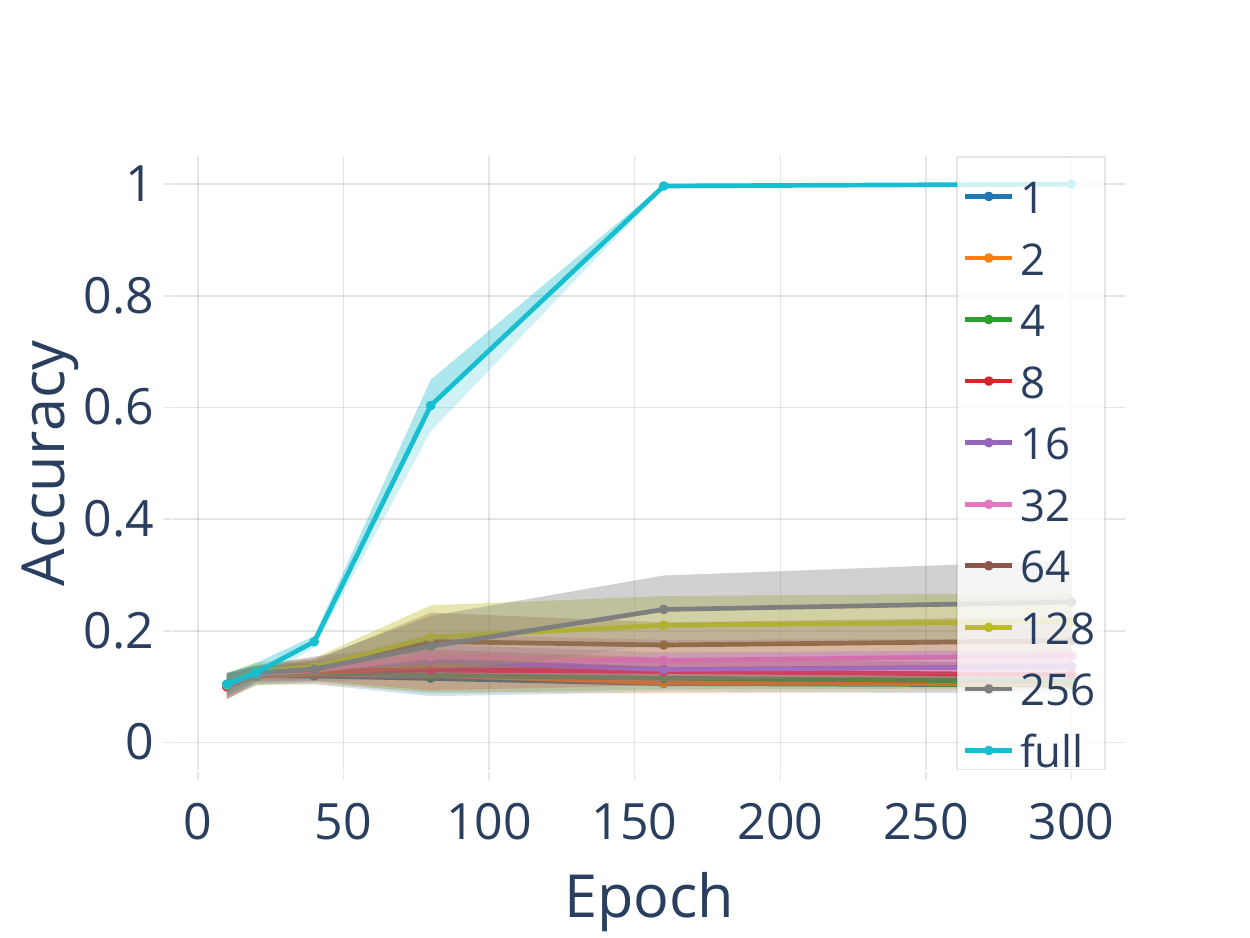}
    }
    \subfloat[Phi-2.7B, $\ell = 26$]{
        \includegraphics[width=\thirdWidth]{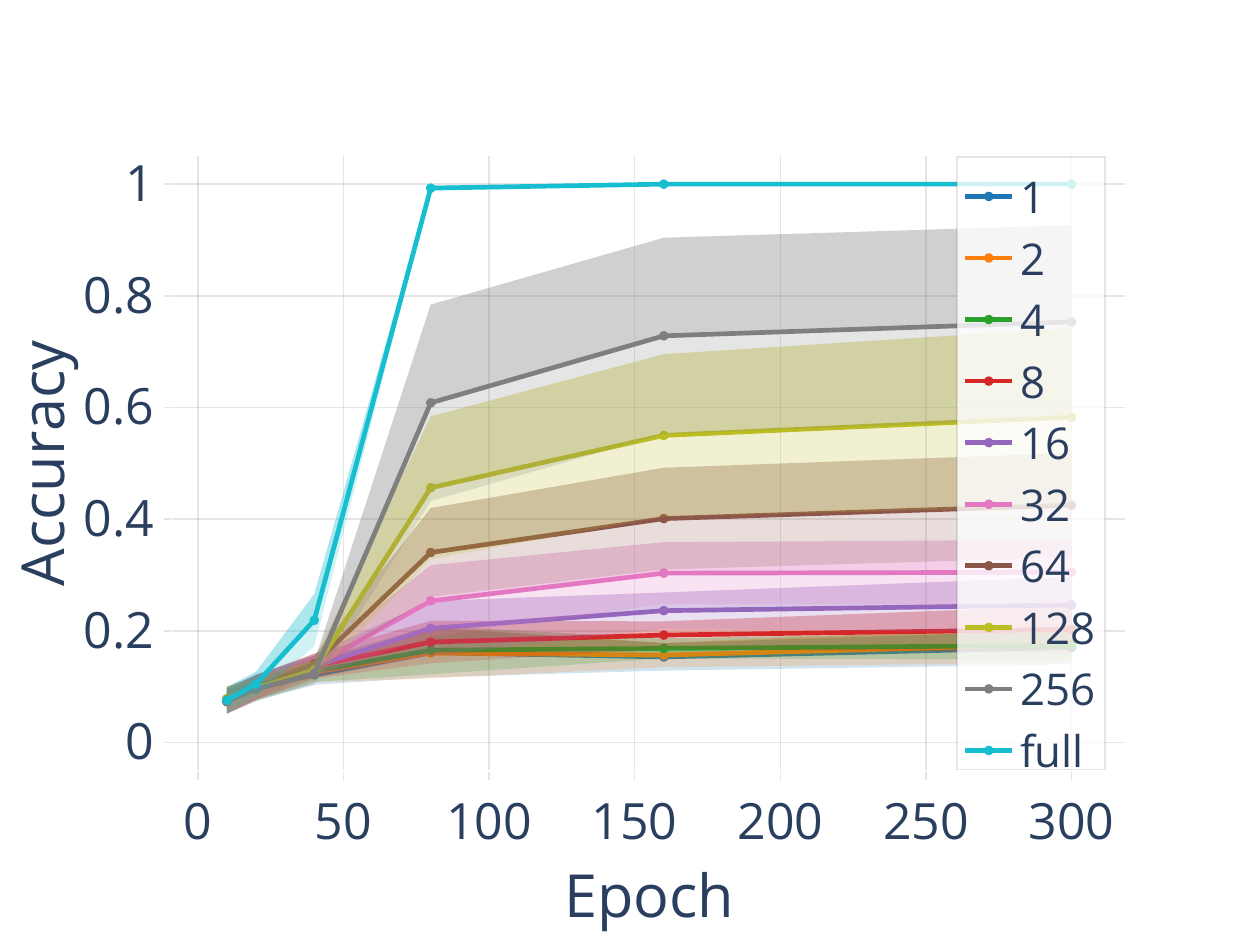}
    }
    \subfloat[Llama2-13B, $\ell = 26$]{
        \includegraphics[width=\thirdWidth]{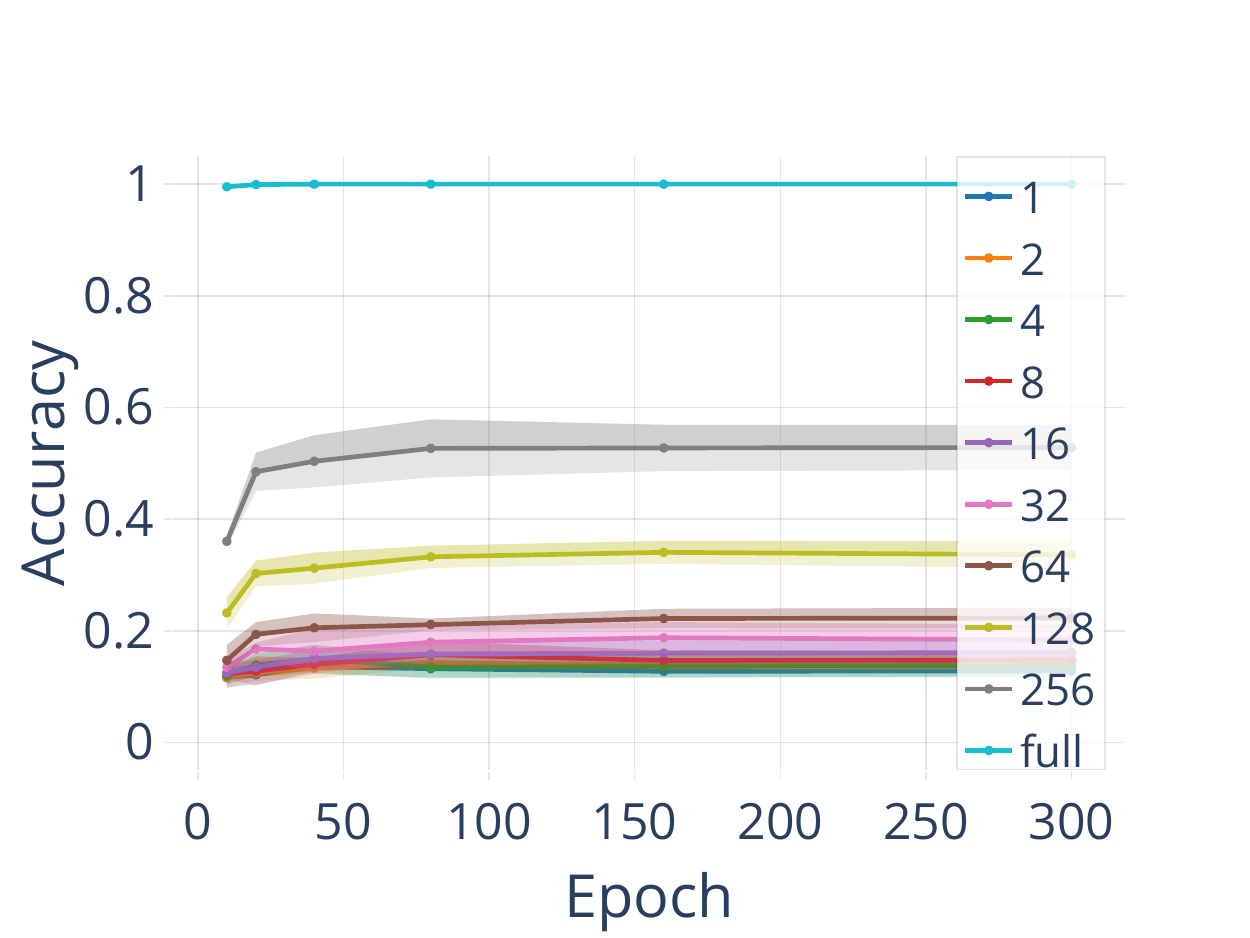}
    }
    \\
\caption{\capthead{Recollection accuracy for \emph{untrained} models for different prefix lengths during training for different $\ell$'s}{$n = 1024$}
    For untrained models, local prefixes perform significantly worse than for pretrained models, which might mean that untrained models either rely on more tokens from the context to make predictions, or that they use a similar number of tokens as pretrained models which are distributed throughout the context.
}
\label{fig:epoch_prefix_len_untrained_all}
\end{figure}

Figure~\ref{fig:epoch_prefix_len_all} shows the recollection accuracy of prefixes with different length for different models and $\ell$, for pretrained models.
In Figure~\ref{fig:epoch_prefix_len_untrained_all} we show the same for untrained models.
For untrained models, prefixes shorter than the full prefix recollect significantly fewer tokens correctly than for pretrained models, in most cases not performing much better than random guessing.
This could mean that untrained models either rely on more tokens from the context to make predictions, or that they use a similar number of tokens as pretrained models, but which are not necessarily immediately preceding the token to predict, but more spread out over the context.
In the latter case this would mean that during pretraining, models acquire a bias that makes them pay more attention to tokens immediately preceding the token to predict.

\subsection{Additional results on global context}
\label{app:global_context_results}

\begin{figure}[H]
    \centering
    \subfloat[Pythia-1B]{
        \includegraphics[width=\thirdWidth]{figures/prefix_mappings/replacement_strategy/replacement-strategy_pythia-1b.pdf}
    }
    \subfloat[Phi-2.7B]{
        \includegraphics[width=\thirdWidth]{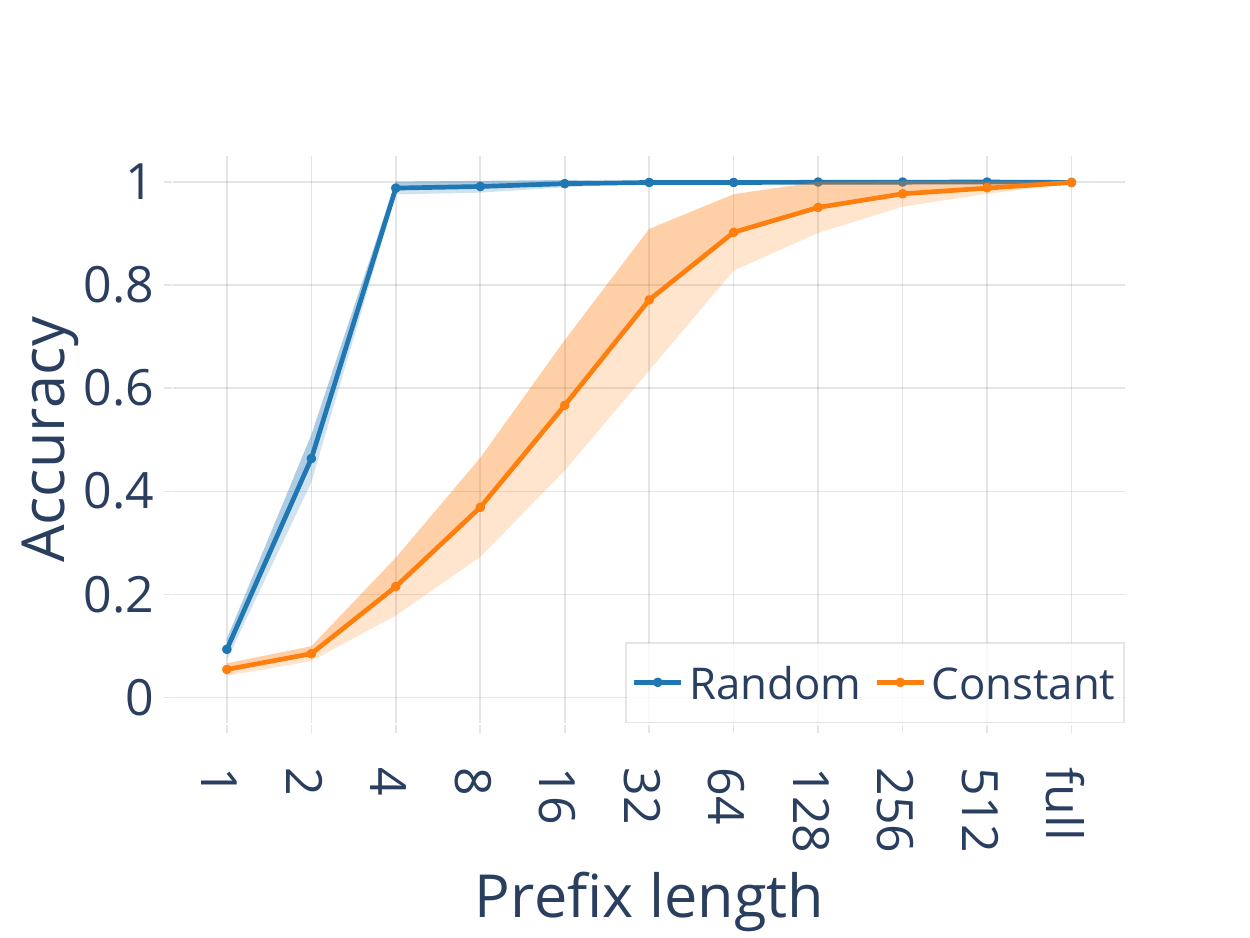}
    }
    \subfloat[Llama2-13B]{
        \includegraphics[width=\thirdWidth]{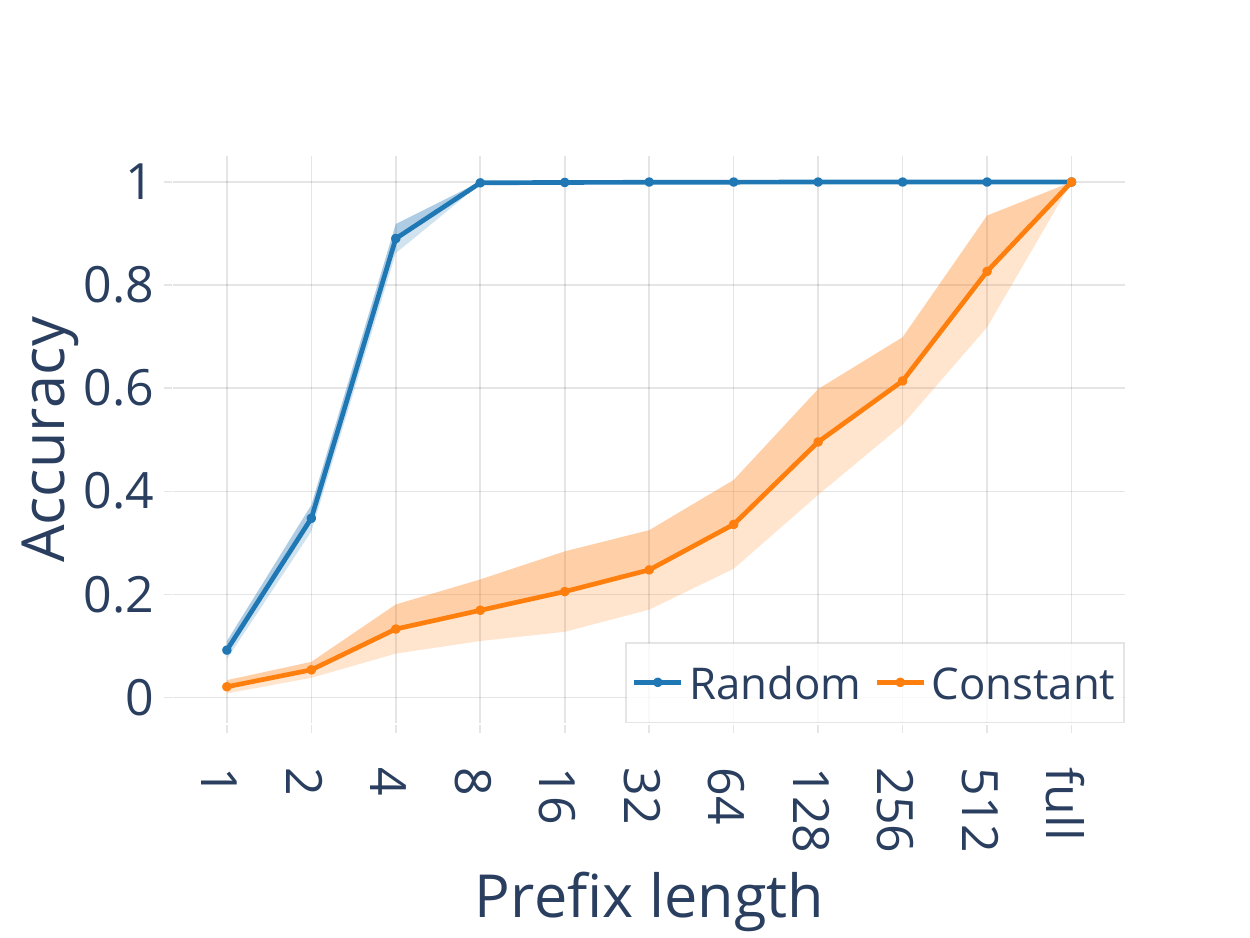}
    }
\caption{\capthead{Effect of replacement strategies for the global context on recollection accuracy.}{$n = 1024, \ell = 26$}
}
\label{fig:replacement_strategy_all}
\end{figure}

\begin{figure}[H]
    \centering
    \subfloat[Pythia-1B]{
        \includegraphics[width=\thirdWidth]{figures/prefix_mappings/size_change/size-change_pythia-1b.pdf}
    }
    \subfloat[Phi-2.7B]{
        \includegraphics[width=\thirdWidth]{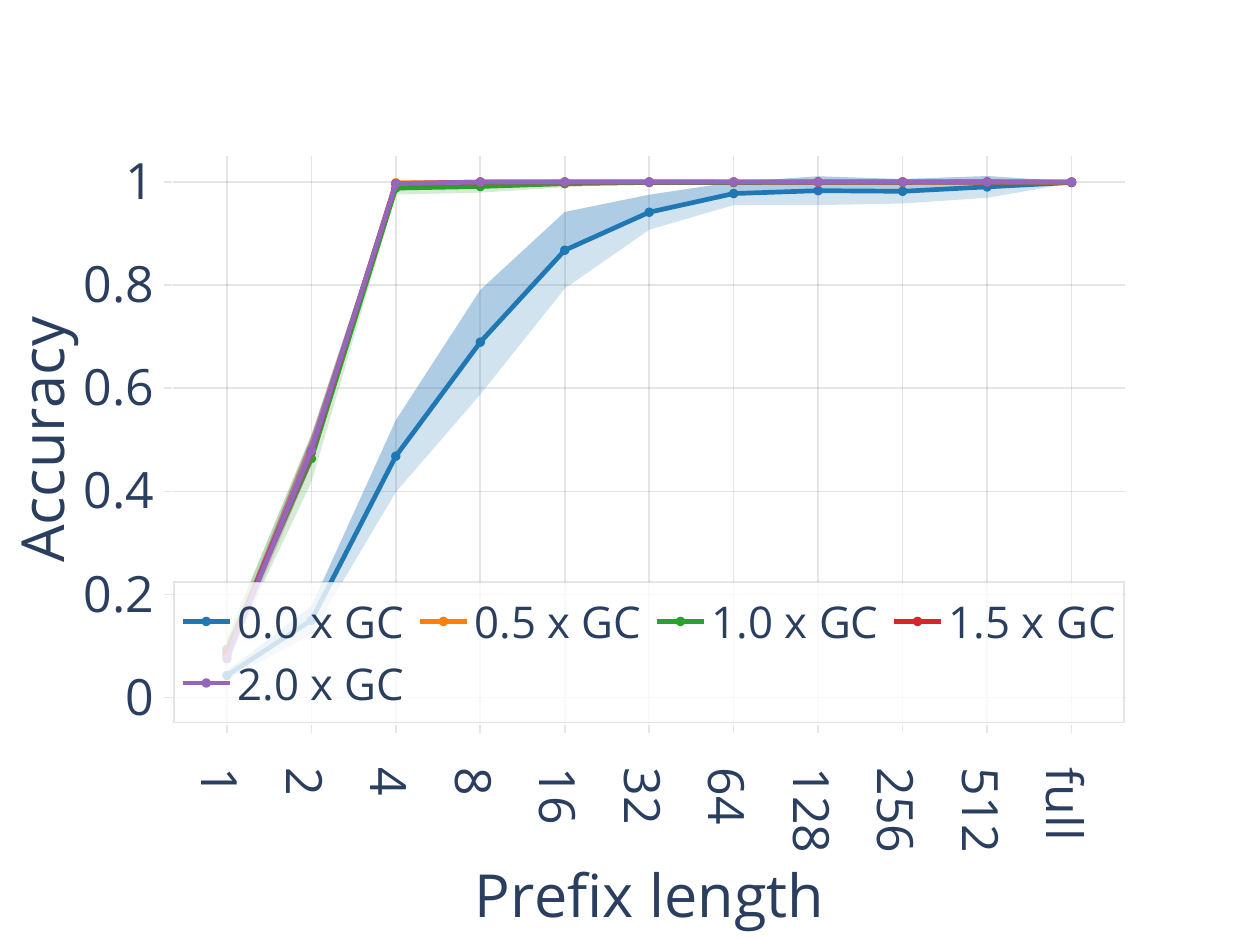}
    }
    \subfloat[Llama2-13B]{
        \includegraphics[width=\thirdWidth]{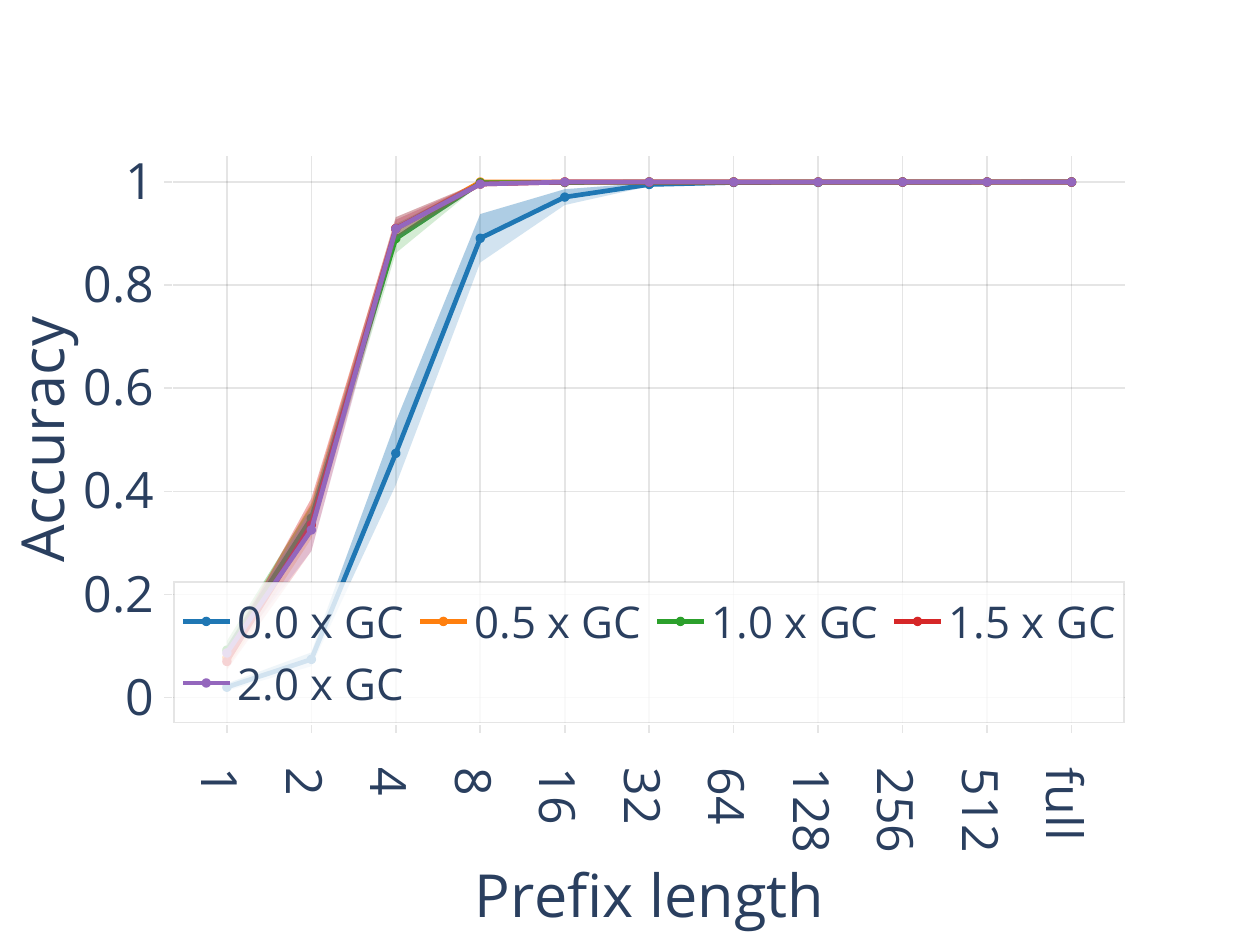}
    }
\caption{\capthead{Effect of changing the size of the global on recollection accuracy.}{$n = 1024, \ell = 26$}
}
\label{fig:size_change_all}
\end{figure}

In Figure~\ref{fig:replacement_strategy_all} we show the effect of the replacement strategy ({\random} and {\constant}) on recollection accuracy.
In Figure~\ref{fig:size_change_all} we show the effect of changing the size of the global context.

\subsection{Results for models with absolute position embeddings}
\label{app:abs_pos_prefix_mappings}

The main model families discussed in the paper (Pythia, Phi, Llama2), as well as many other modern architectures, use \emph{relative position encodings}, and in particular rotary embeddings~\citep{su2024roformer}.
Some older architectures, however, use \emph{absolute position encodings}, such as GPT-2~\citep{radford2019language}, GPT-3~\citep{brown2020language} and OPT~\cite{zhang2022opt}.
To account for this difference, we also study models from these families, in particular the 140M parameter GPT-2 and the OPT-350M model.

\begin{figure}[H]
    \centering
    \subfloat[GPT2-124M, Accuracy]{
        \includegraphics[width=\smallThirdWidth]{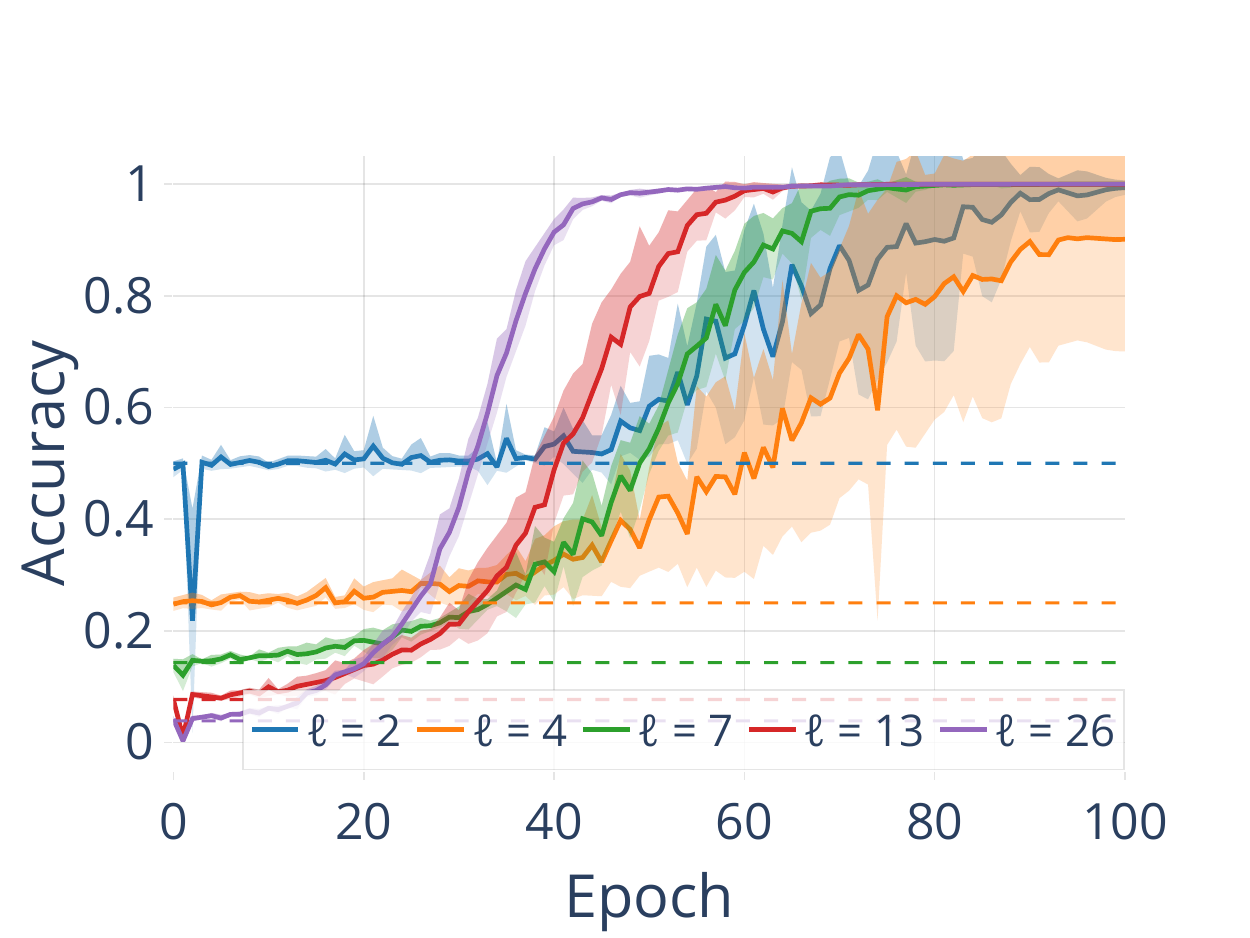}
    }
    \subfloat[OPT-350M, Accuracy]{
        \includegraphics[width=\smallThirdWidth]{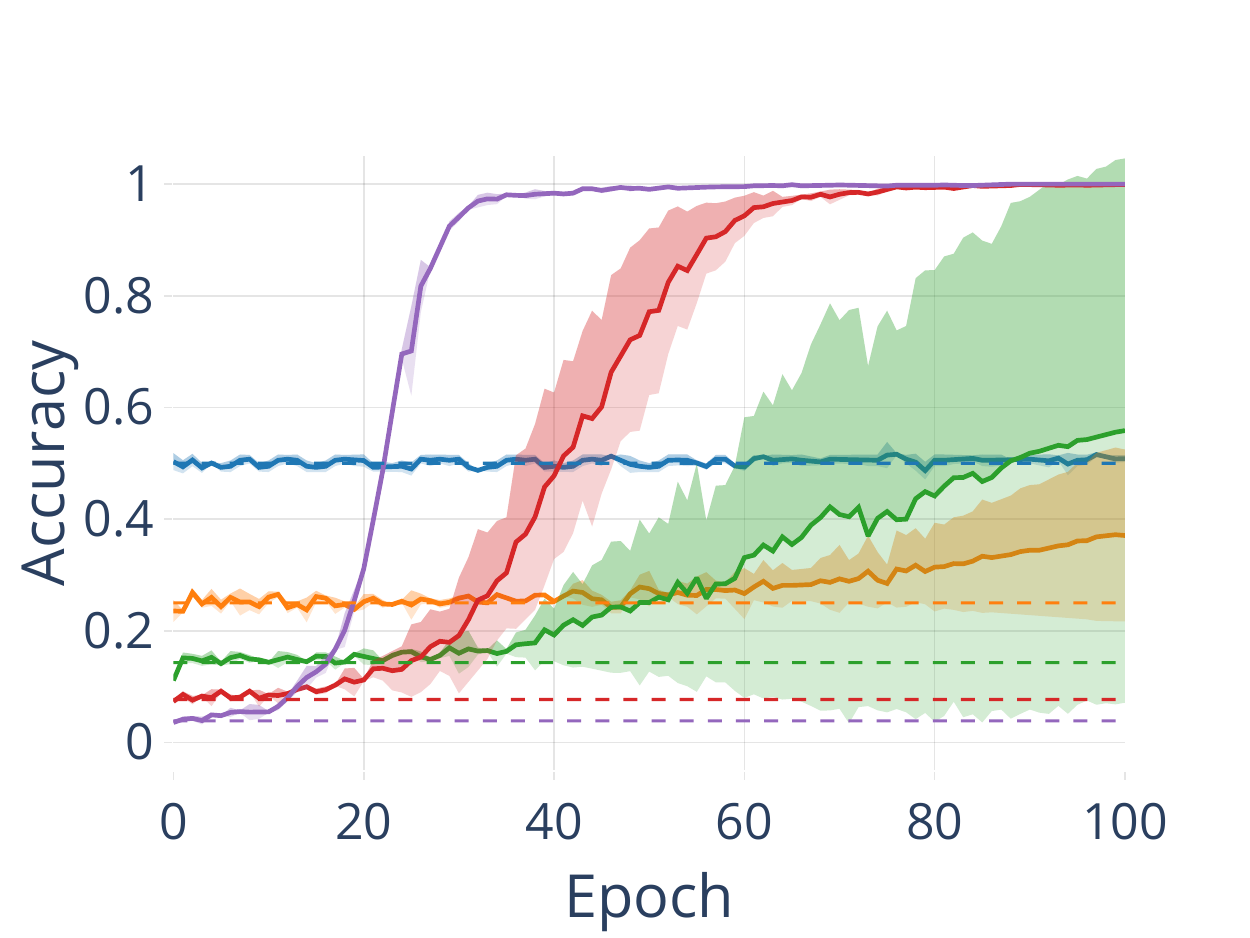}
    }
\caption{\capthead{Memorisation dynamics for different $\ell$, for models with absolute position encodings}{$n = 1024$}
We observe the same memorisation dynamics for models with absolute position encodings as for models with relative position encodings.
}
\label{fig:abs_pos_dynamics_alphabet_size}
\end{figure}

Figure~\ref{fig:abs_pos_dynamics_alphabet_size} shows that these models behave similarly to their more modern counterparts in terms of memorisation behaviour (phases of memorisation, higher entropy strings being easier to memorise, etc.)
However, similar to~\citet{sinha2022curious} we find in our analysis on the role of local prefixes and global context, that these models tend to over-rely on position information, which can impact their memorisation behaviour.

\begin{figure}[H]
    \centering
    \subfloat[GPT2-124M, $\ell = 2$]{
        \includegraphics[width=\smallThirdWidth]{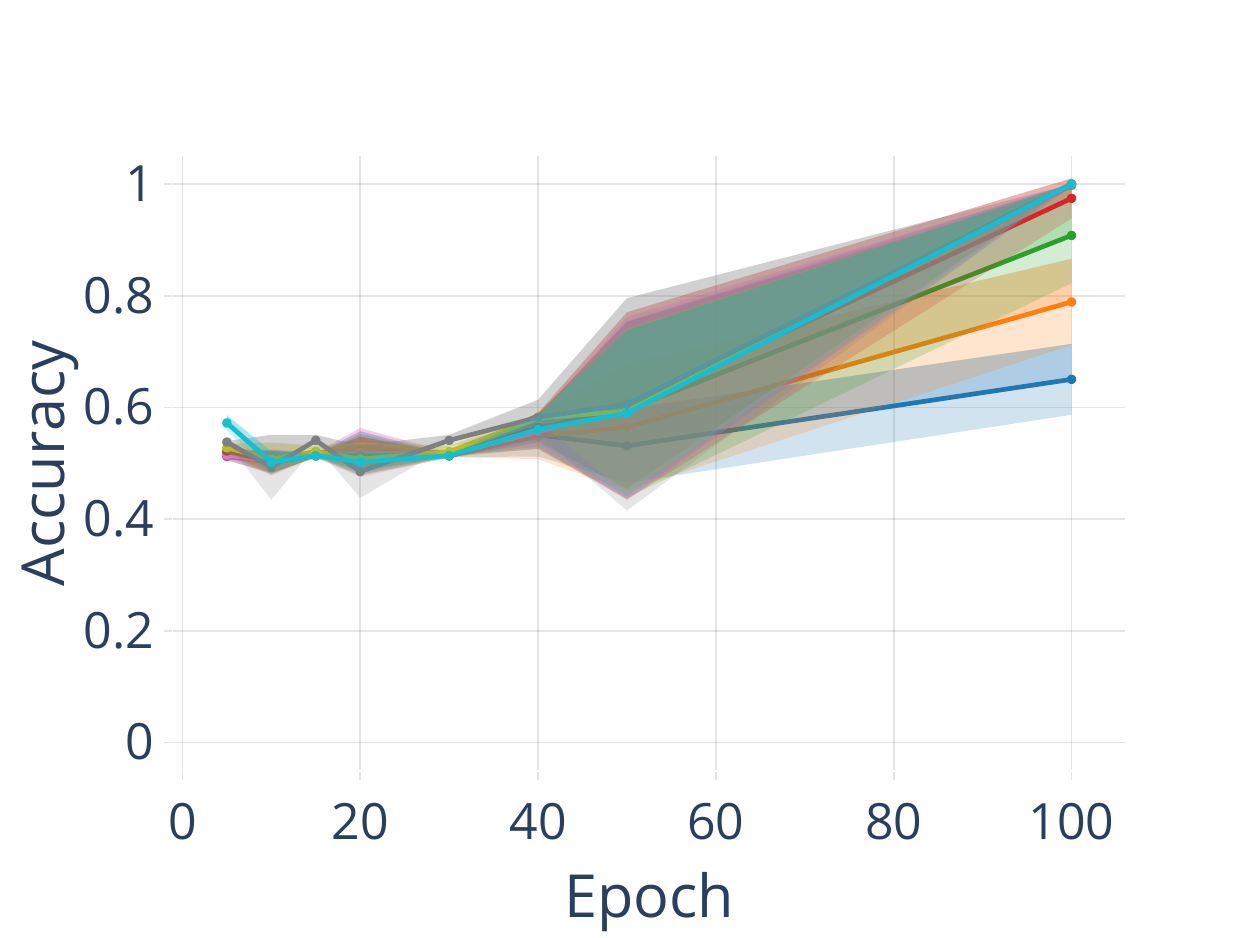}
    }
    \subfloat[OPT-350M, $\ell = 2$]{
        \includegraphics[width=\smallThirdWidth]{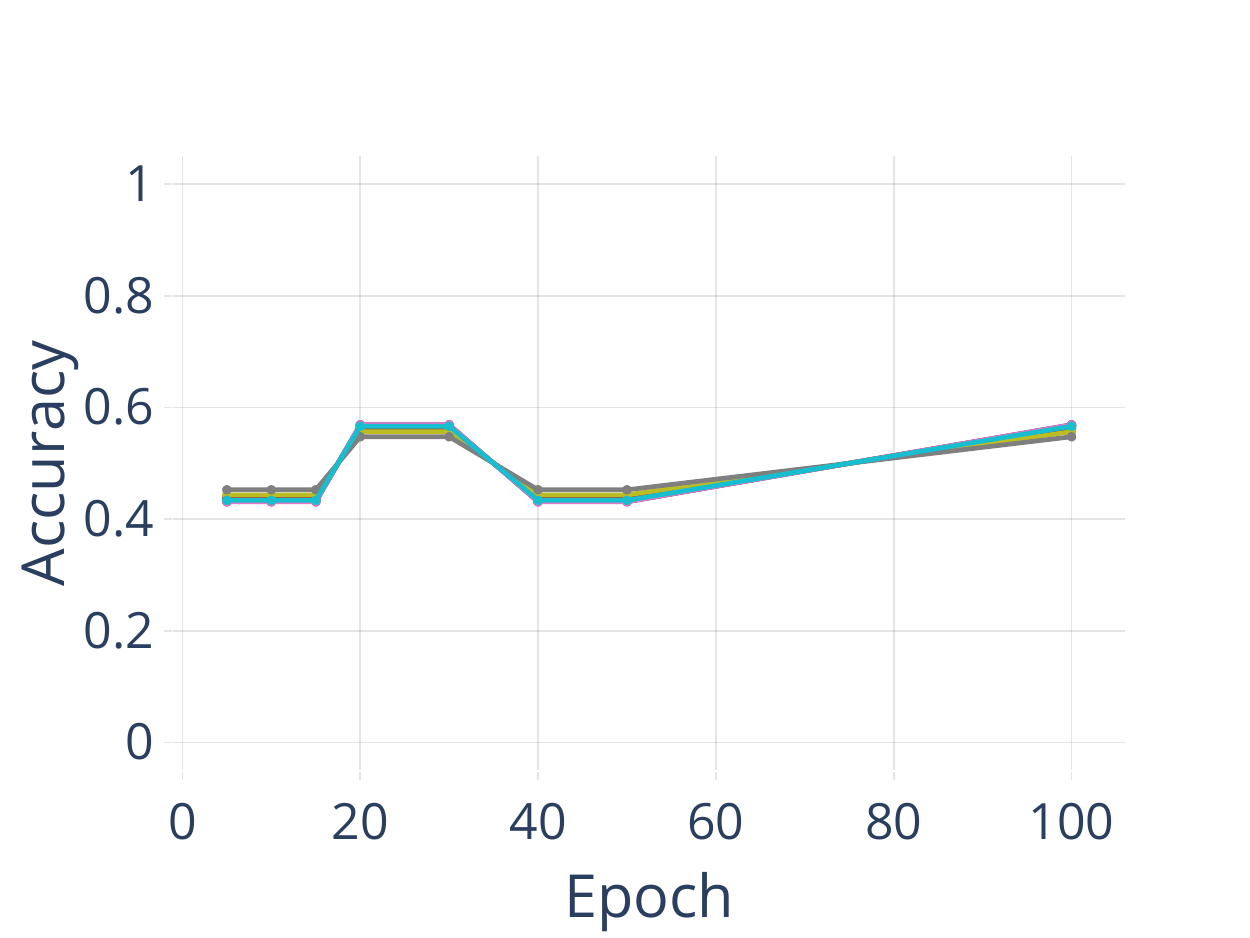}
    }
    \\
    \subfloat[GPT2-124M, $\ell = 26$]{
        \includegraphics[width=\smallThirdWidth]{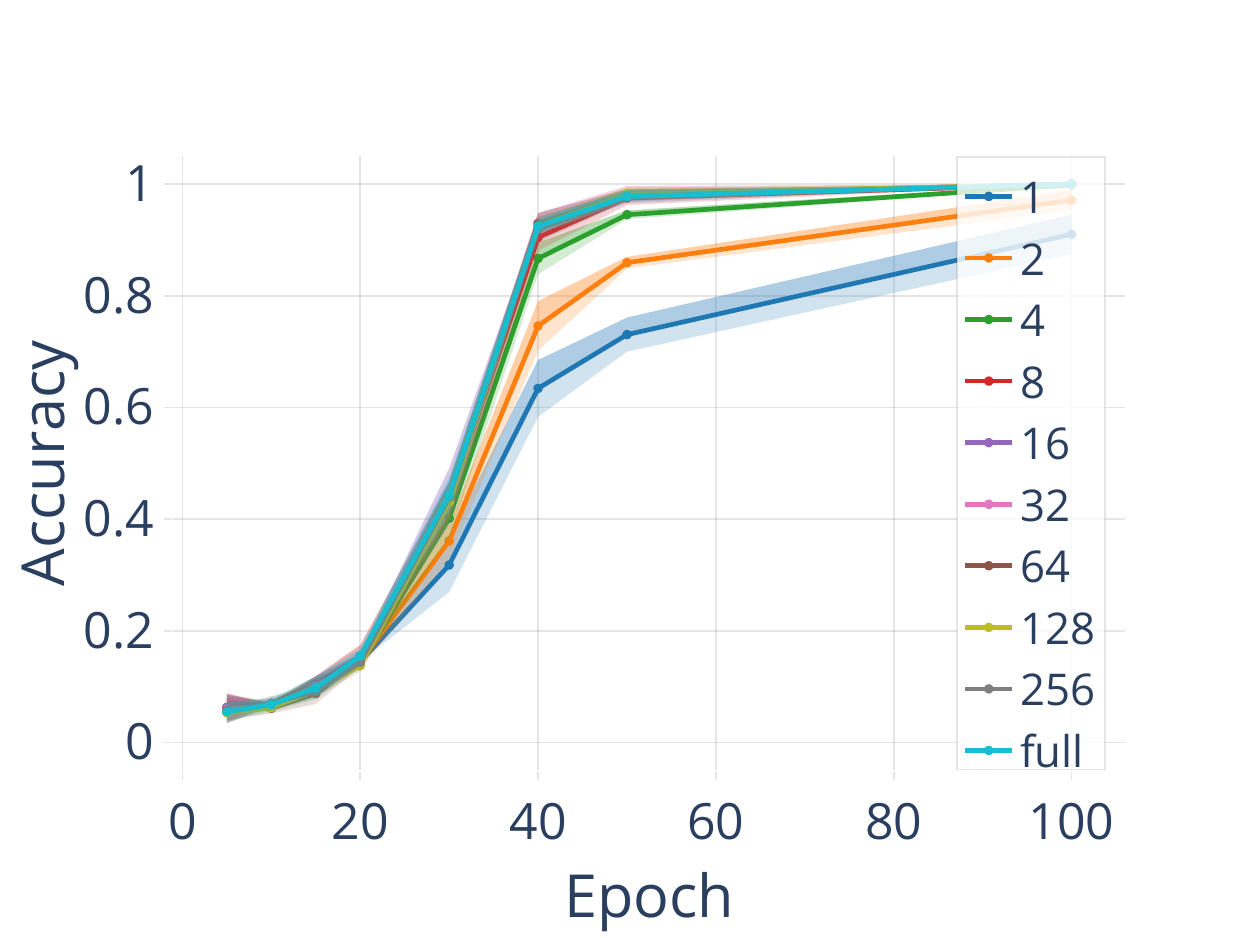}
    }
    \subfloat[OPT-350M, $\ell = 26$]{
        \includegraphics[width=\smallThirdWidth]{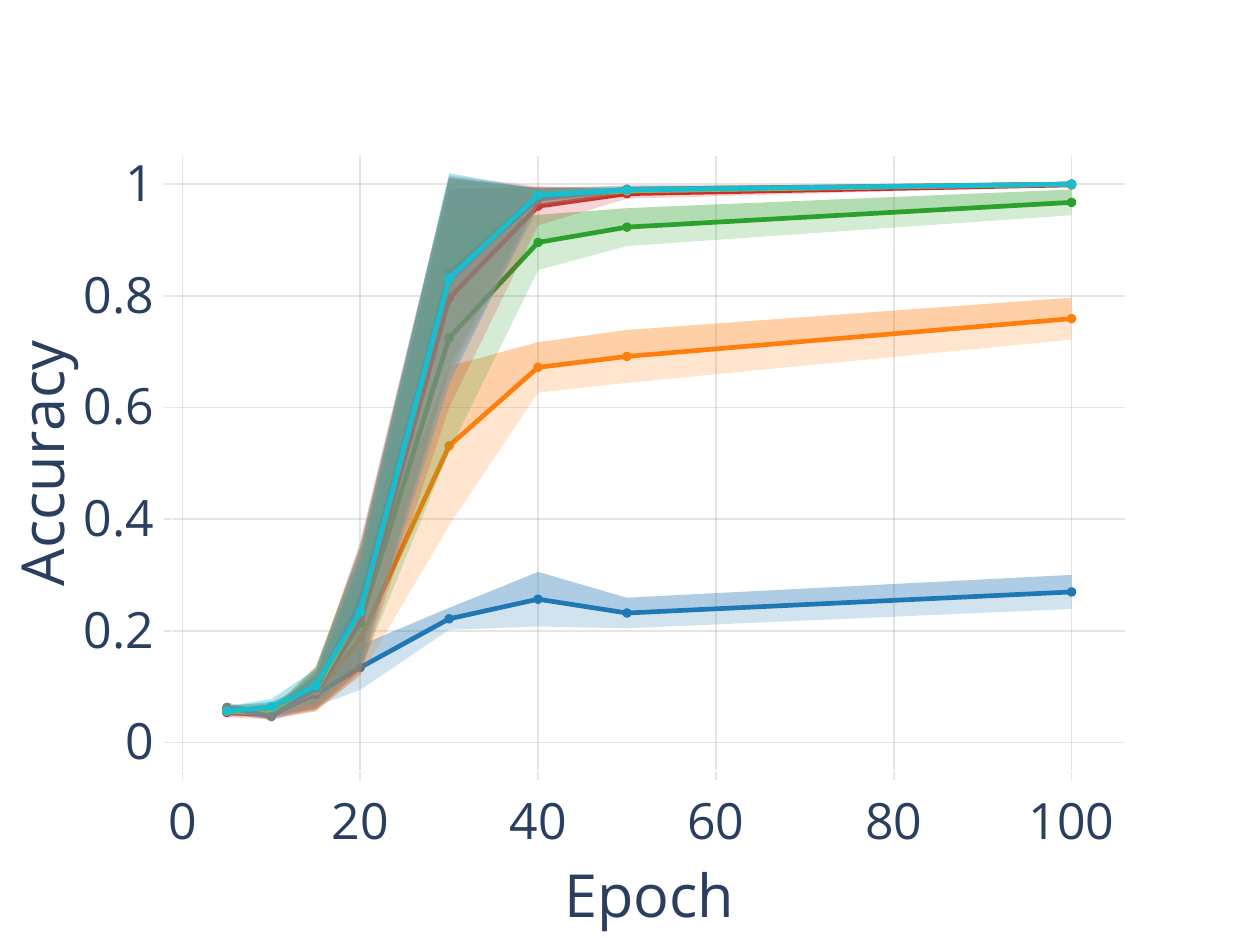}
    }
\caption{\textbf{[Recollection accuracy for different prefix lengths during training for different $\ell$'s]}
Short prefixes perform considerably better for models with absolute position encodings, than for ones with relative position encodings.
}
\label{fig:abs_pos_models_epoch_prefix_len}
\end{figure}

\begin{figure}[H]
    \centering
    \subfloat[GPT2-124M]{
        \includegraphics[width=\thirdWidth]{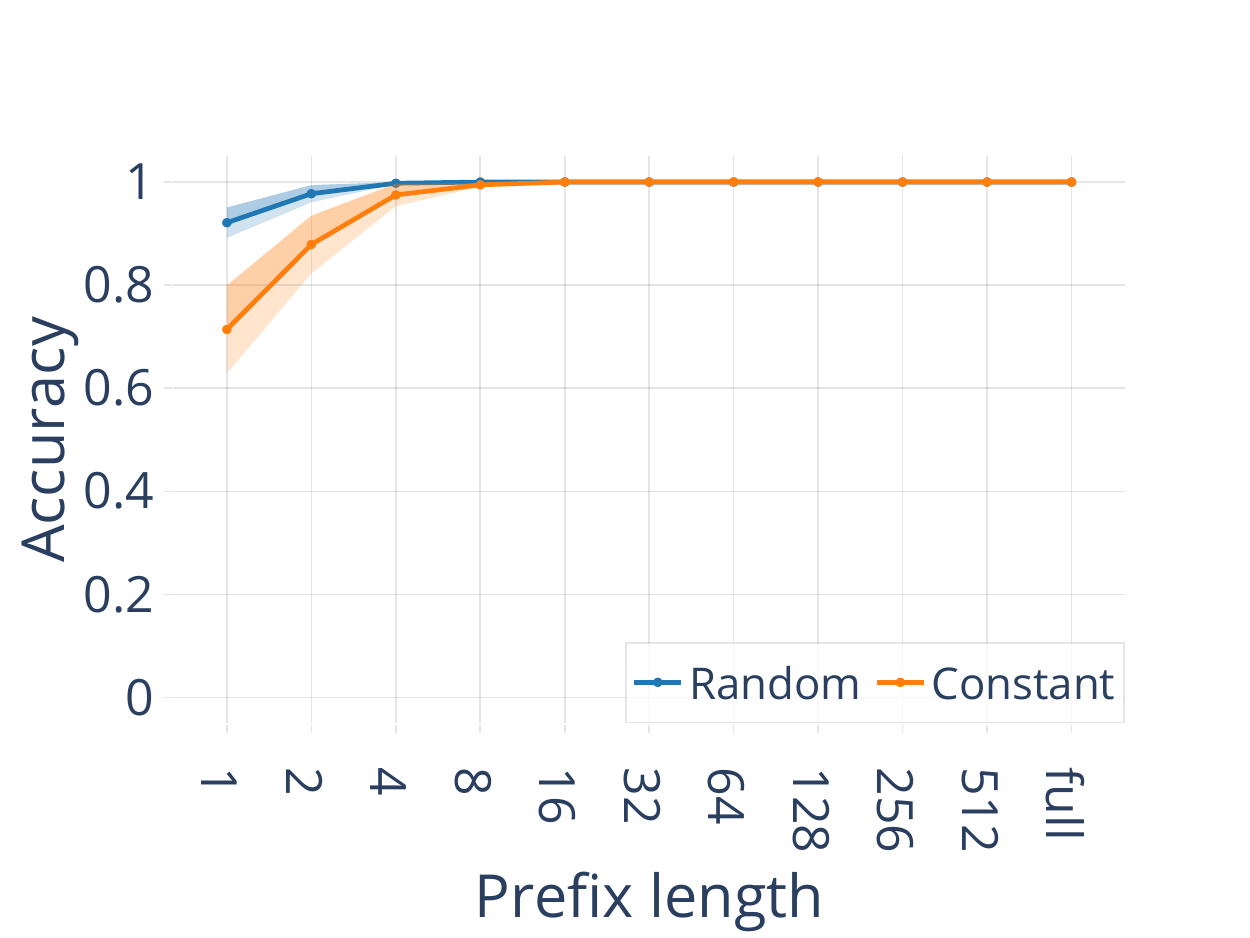}
    }
    \subfloat[OPT-350M]{
        \includegraphics[width=\thirdWidth]{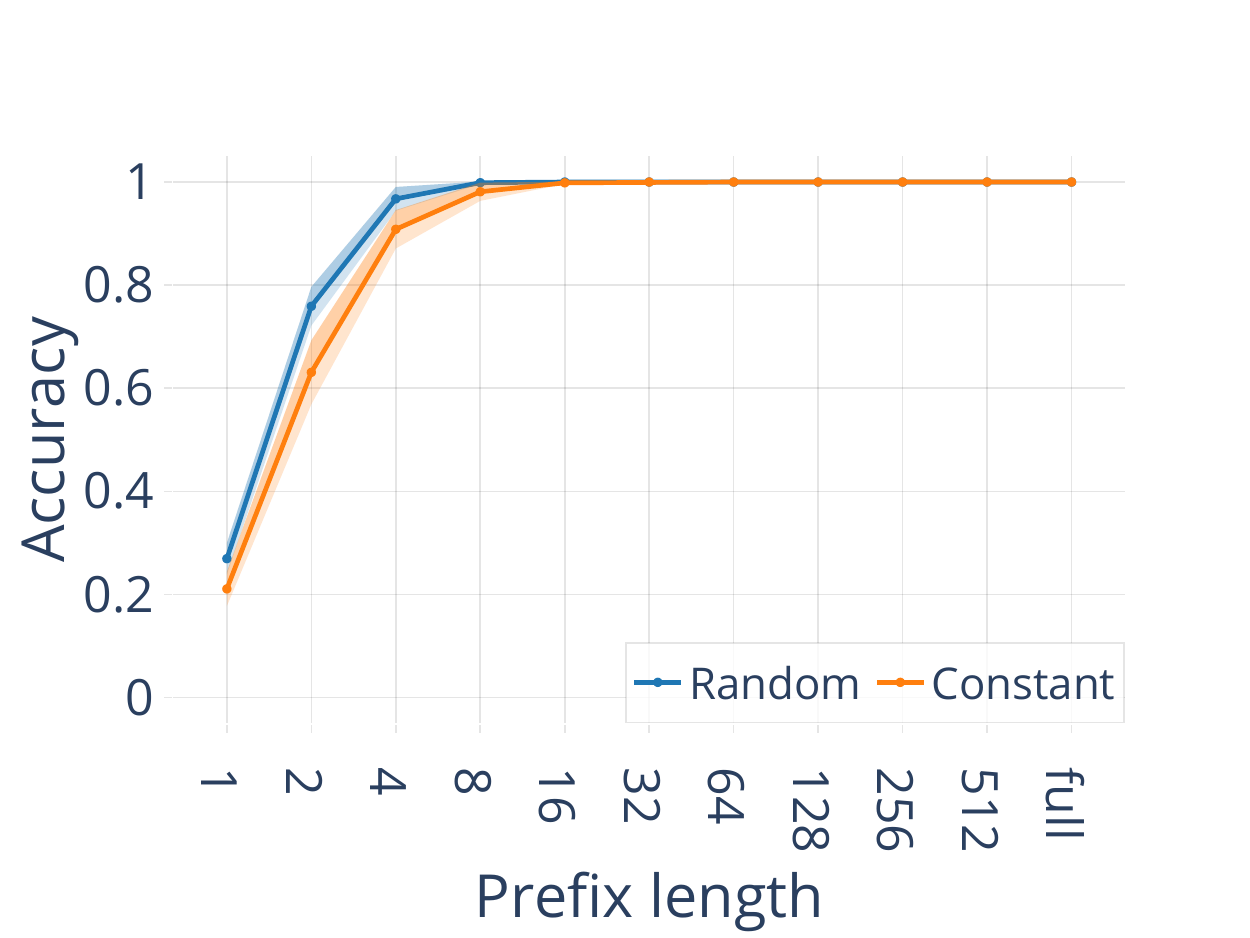}
    }
\caption{\capthead{Effect of replacement strategies for the global context on recollection accuracy.}{$n = 1024, \ell = 26$}
Models that use global position encodings are barely affected by changes in the global context distribution.
}
\label{fig:replacement_strategy_abs_pos}
\end{figure}

\begin{figure}[H]
    \centering
    \subfloat[GPT2-124M, $n = 512$]{
        \includegraphics[width=\thirdWidth]{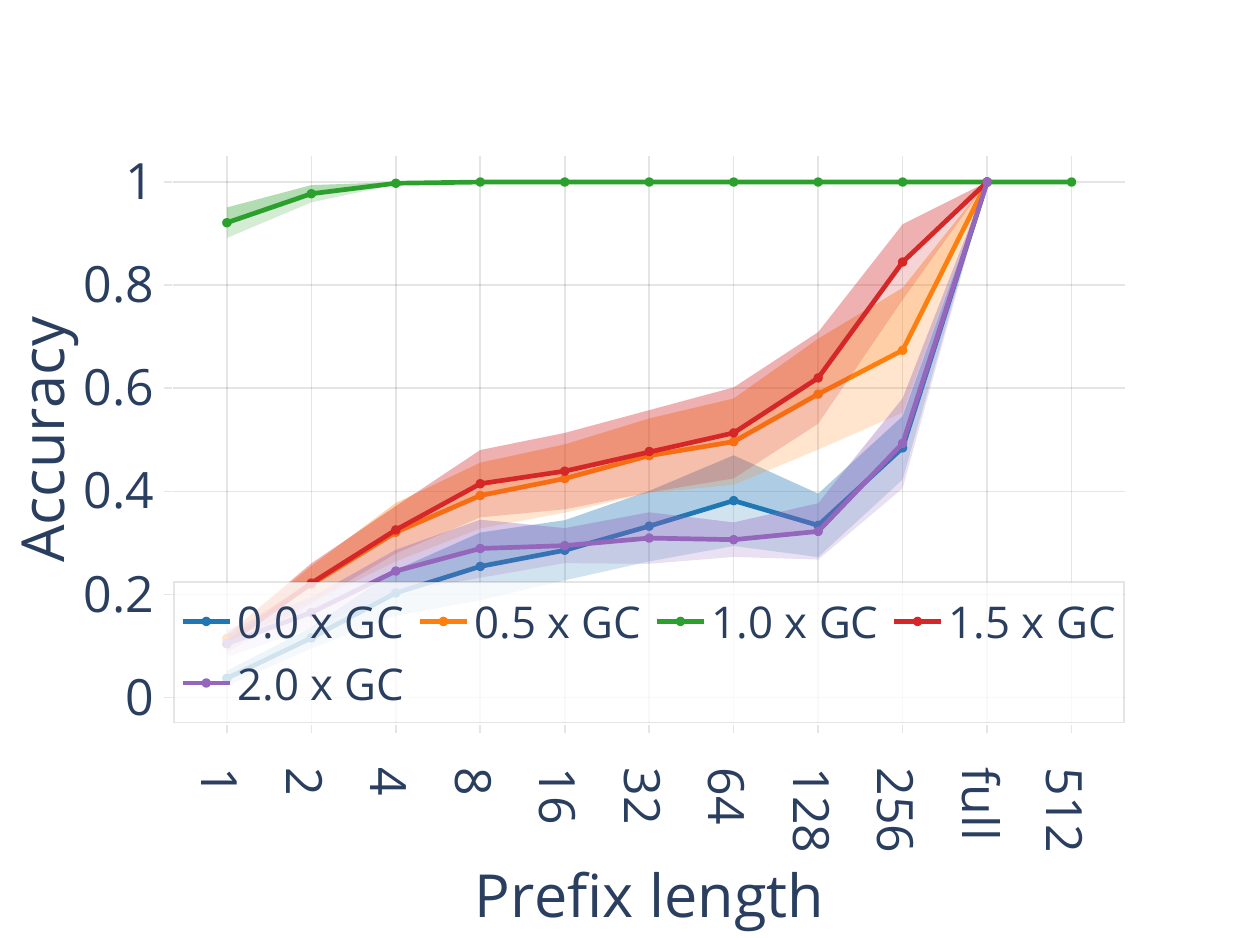}
    }
    \subfloat[OPT-350M, $n = 1024$]{
        \includegraphics[width=\thirdWidth]{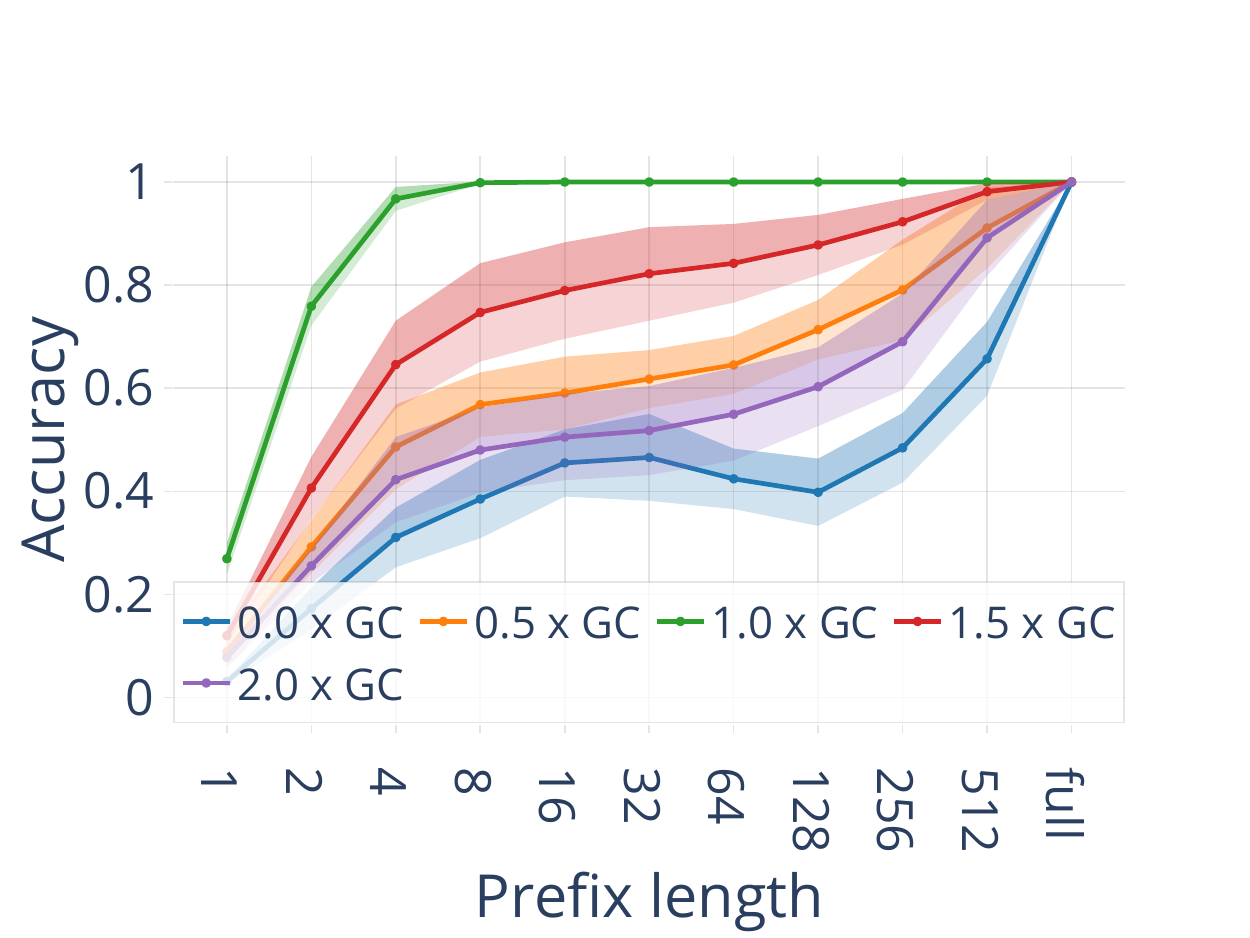}
    }
\caption{\capthead{Effect of changing the size of the global context on recollection accuracy.}{$\ell = 26$}
In contrast to models using relative position encodings, models with absolute position encodings are heavily affected by changes to the length of the global context, \ie~to the token position.
We use $512$ token strings for GPT2 here, since its context window is only $1024$ tokens long, so we would not be able to double the context size with a $1024$ base string.
}
\label{fig:size_change_abs_pos}
\end{figure}

Figure~\ref{fig:abs_pos_models_epoch_prefix_len} shows that for GPT-2 and OPT, short prefixes perform considerably better than for models with relative position encodings.
In Figure~\ref{fig:replacement_strategy_abs_pos} we show results for changing the replacement strategy for the global context, and in Figure~\ref{fig:size_change_abs_pos} we show results for changing the size of the global context.
In contrast to models using relative position encodings, models with absolute position encodings are not affected by changes in the global context distribution, but heavily affected by changes in the size of the global context, \ie~in the position of the token in the string.
These results indicate, that absolute position encoding models rely, at least partially, on position information for memorisation.
Exploring this connection further is an interesting avenue for future work.

\section{Additional Results on Repeated Memorization}
\label{app:repeated_memorization}

\begin{figure}[H]
    \centering
    \subfloat[Pythia-1B, $\ell = 2$, 32 strings]{
        \includegraphics[width=\thirdWidth]{figures/repeated_memorization/pretrained/accuracy_a-2_x32_sequential_pythia-1b.pdf}
    }
    \subfloat[Phi-2.7B, $\ell = 2$, 32 strings]{
        \includegraphics[width=\thirdWidth]{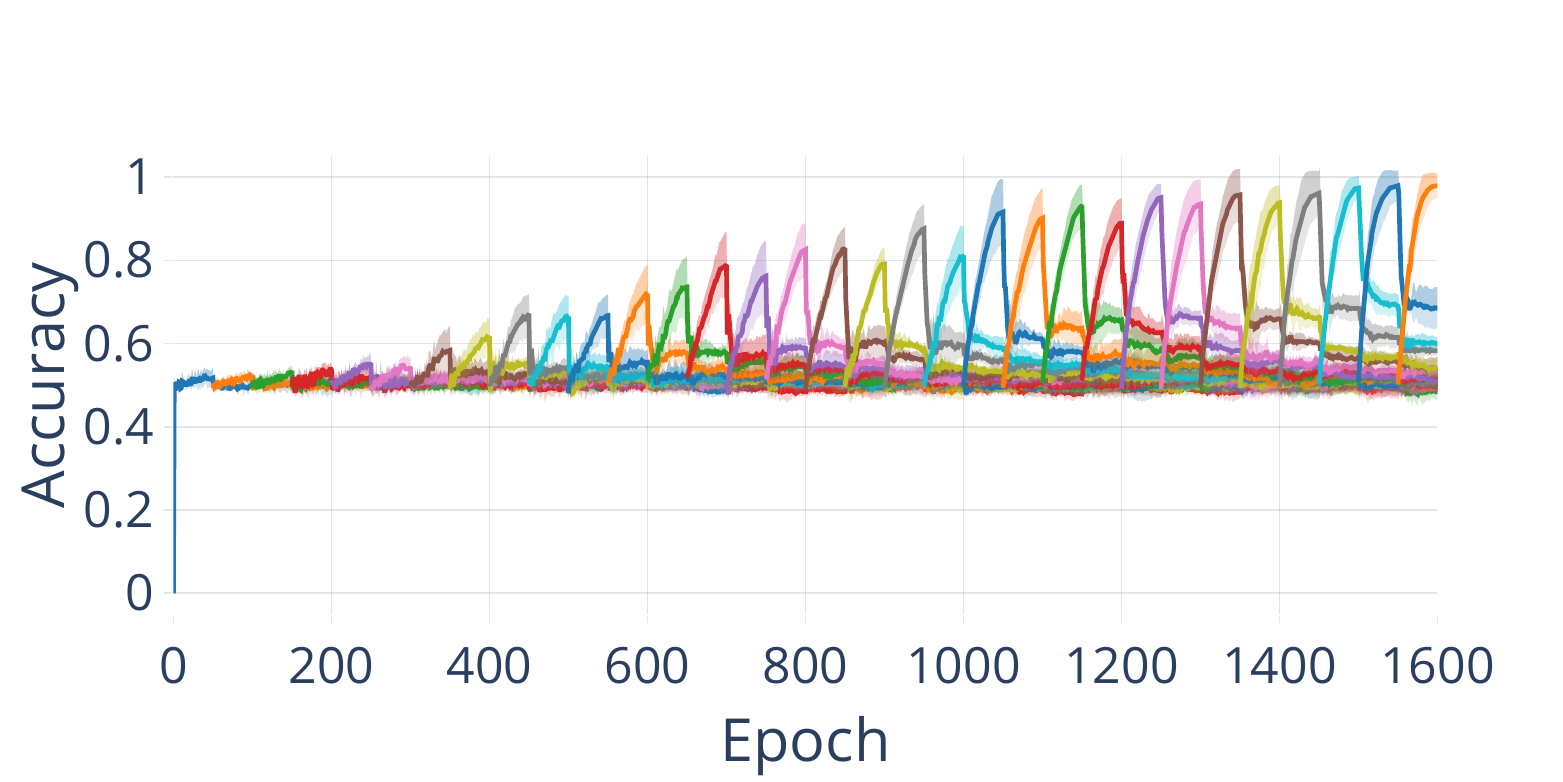}
    }
    \subfloat[Llama2-13B, $\ell = 2$, 16 strings]{
        \includegraphics[width=\thirdWidth]{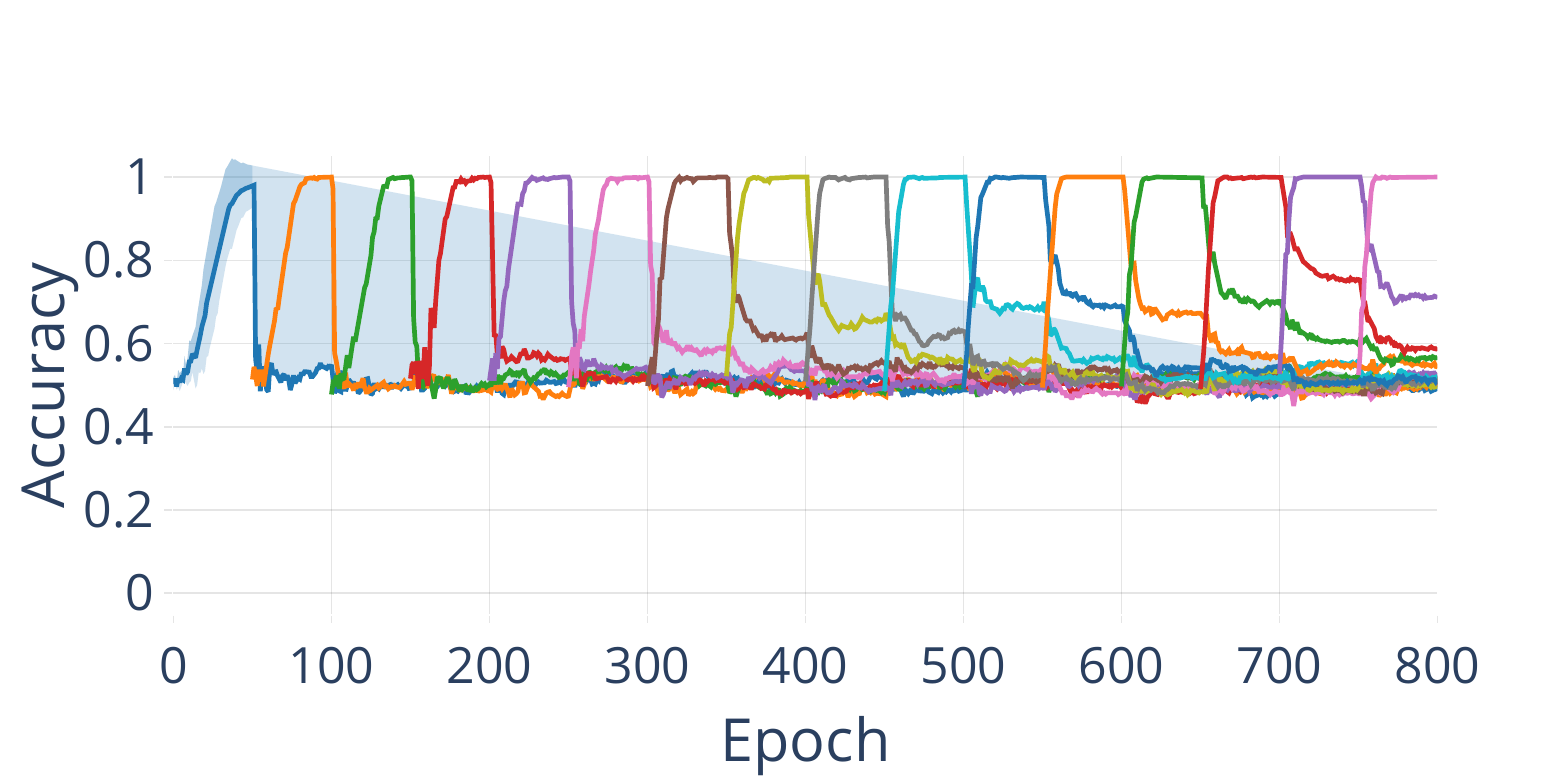}
    }
    \\
    \subfloat[Pythia-1B, $\ell = 26$, 16 strings]{
        \includegraphics[width=\thirdWidth]{figures/repeated_memorization/pretrained/accuracy_a-26_x16_sequential_pythia-1b.pdf}
    }
    \subfloat[Phi-2.7B, $\ell = 26$, 16 strings]{
        \includegraphics[width=\thirdWidth]{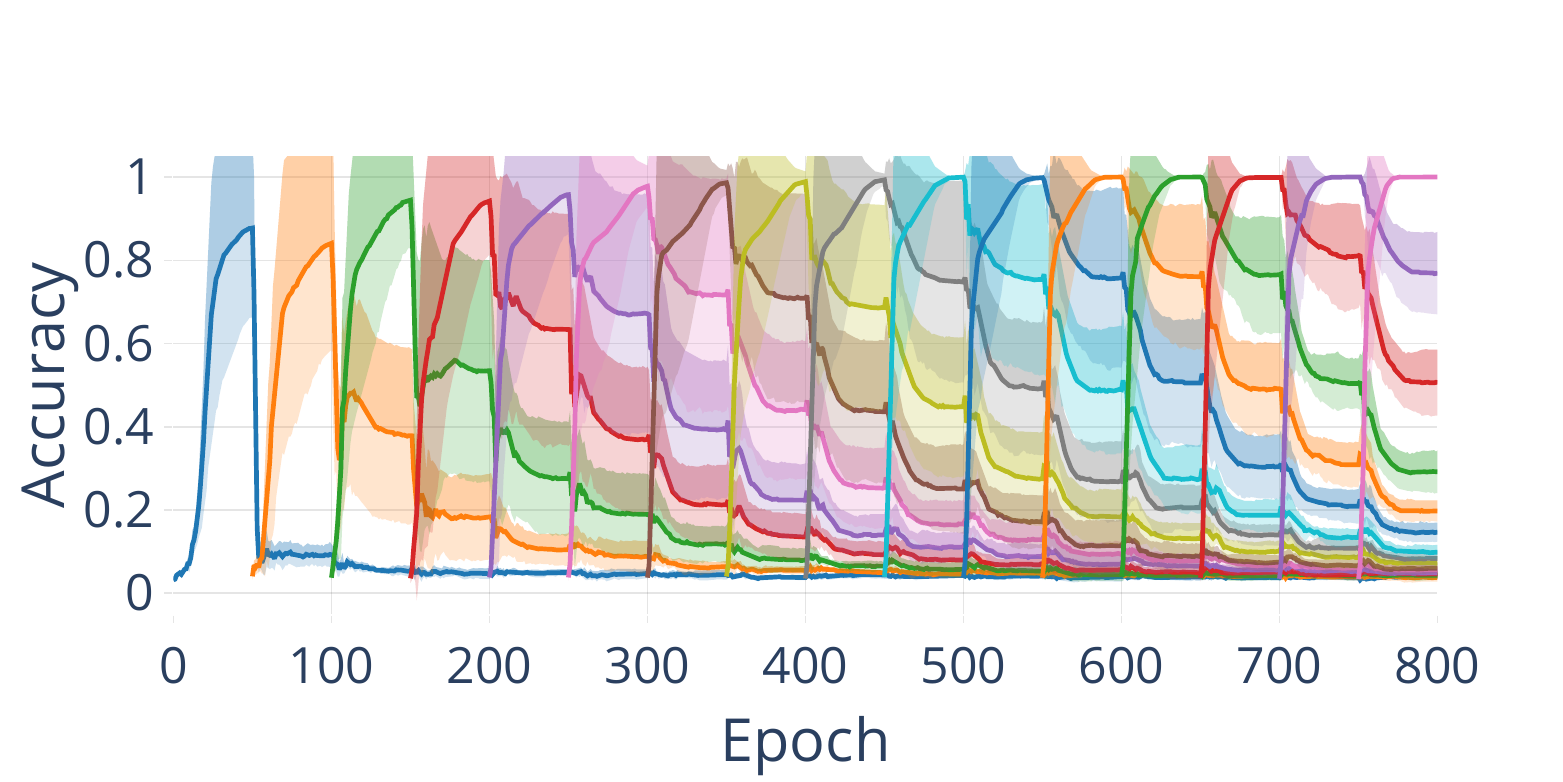}
    }
    \subfloat[Llama2-13B, $\ell = 26$, 16 strings]{
        \includegraphics[width=\thirdWidth]{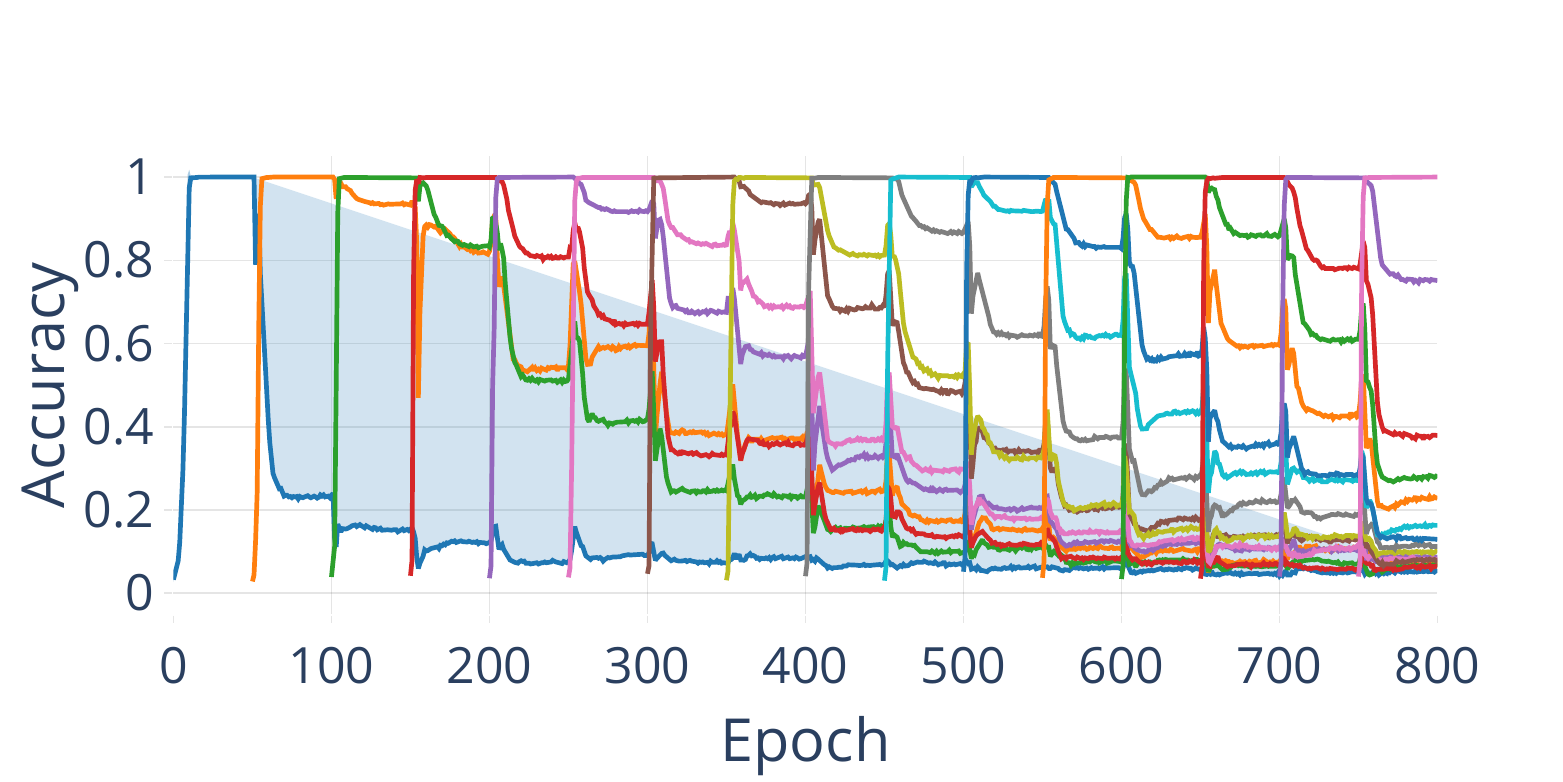}
    }
\caption{\capthead{Accuracy on repeated memorization for different pretrained models.}{$n = 1024$}
As models memorise additional strings, they forget previous ones.
However, forgetting slows down as more strings are memorised.
}
\label{fig:repeated_mem_accuracy_seq_all_pretrained}
\end{figure}

\begin{figure}[H]
    \centering
    \subfloat[Pythia-1B, $\ell = 2$]{
        \includegraphics[width=\thirdWidth]{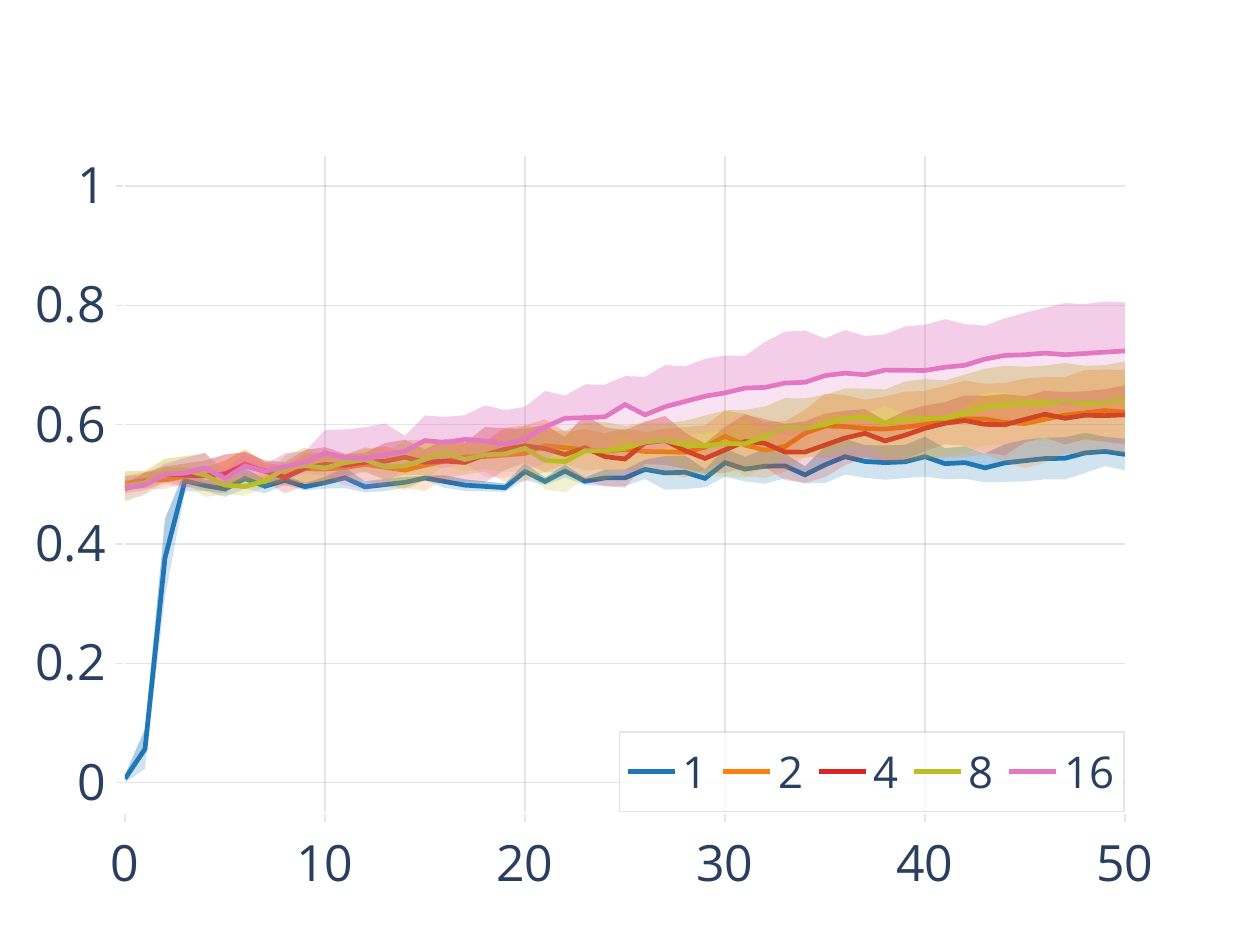}
    }
    \subfloat[Phi-2.7B, $\ell = 2$]{
        \includegraphics[width=\thirdWidth]{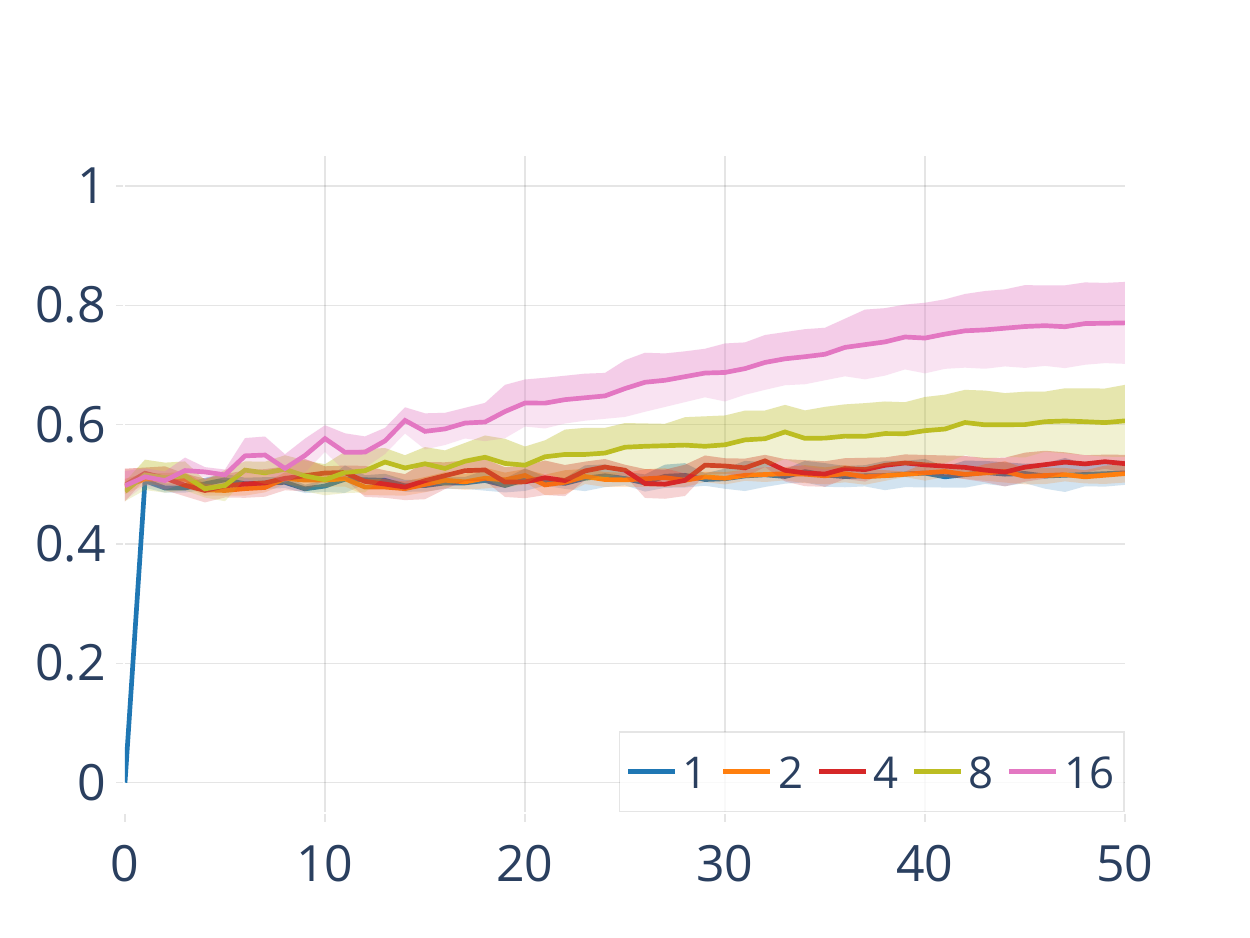}
    }
    \subfloat[Llama2-13B, $\ell = 2$]{
        \includegraphics[width=\thirdWidth]{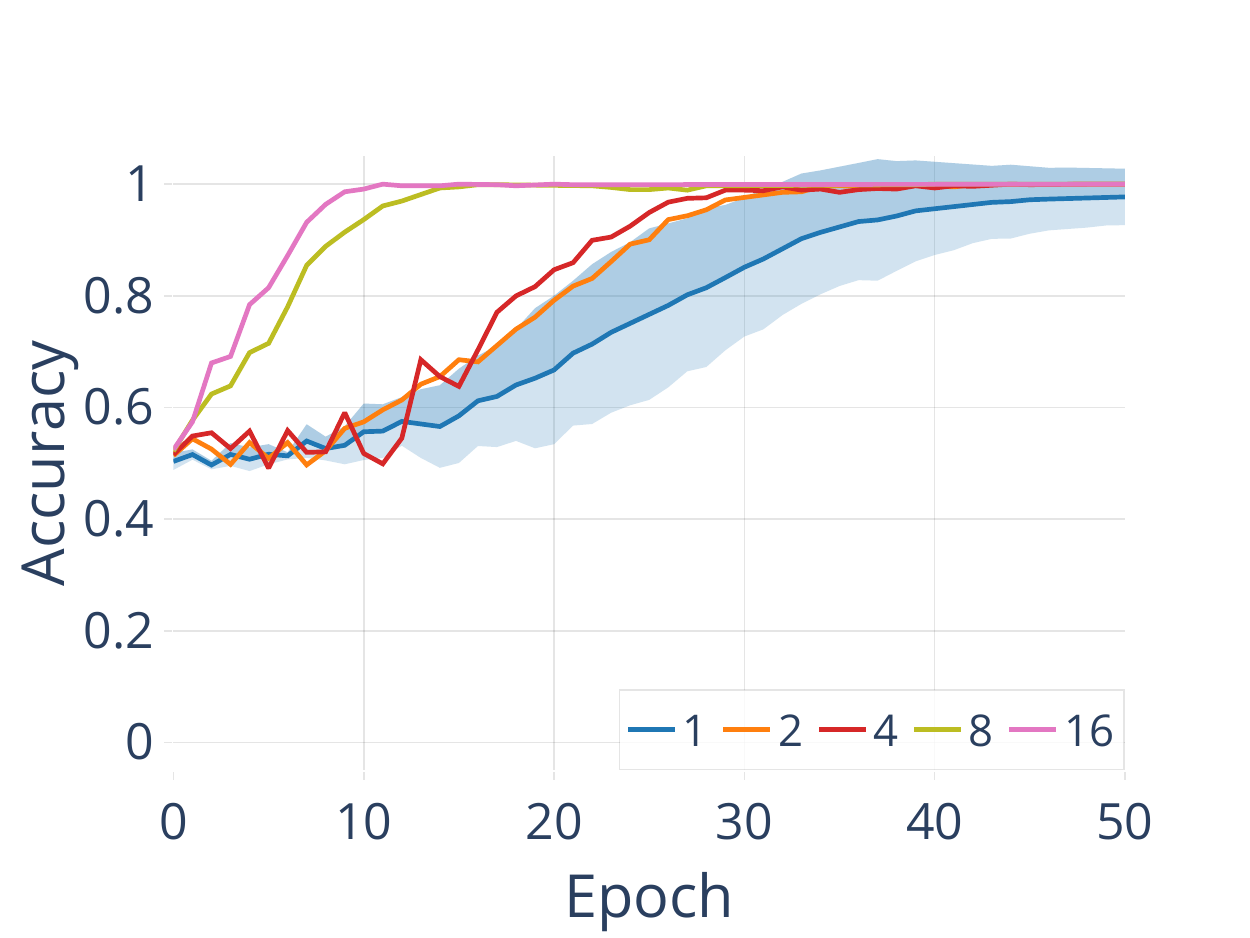}
    }
    \\
    \subfloat[Pythia-1B, $\ell = 26$]{
        \includegraphics[width=\thirdWidth]{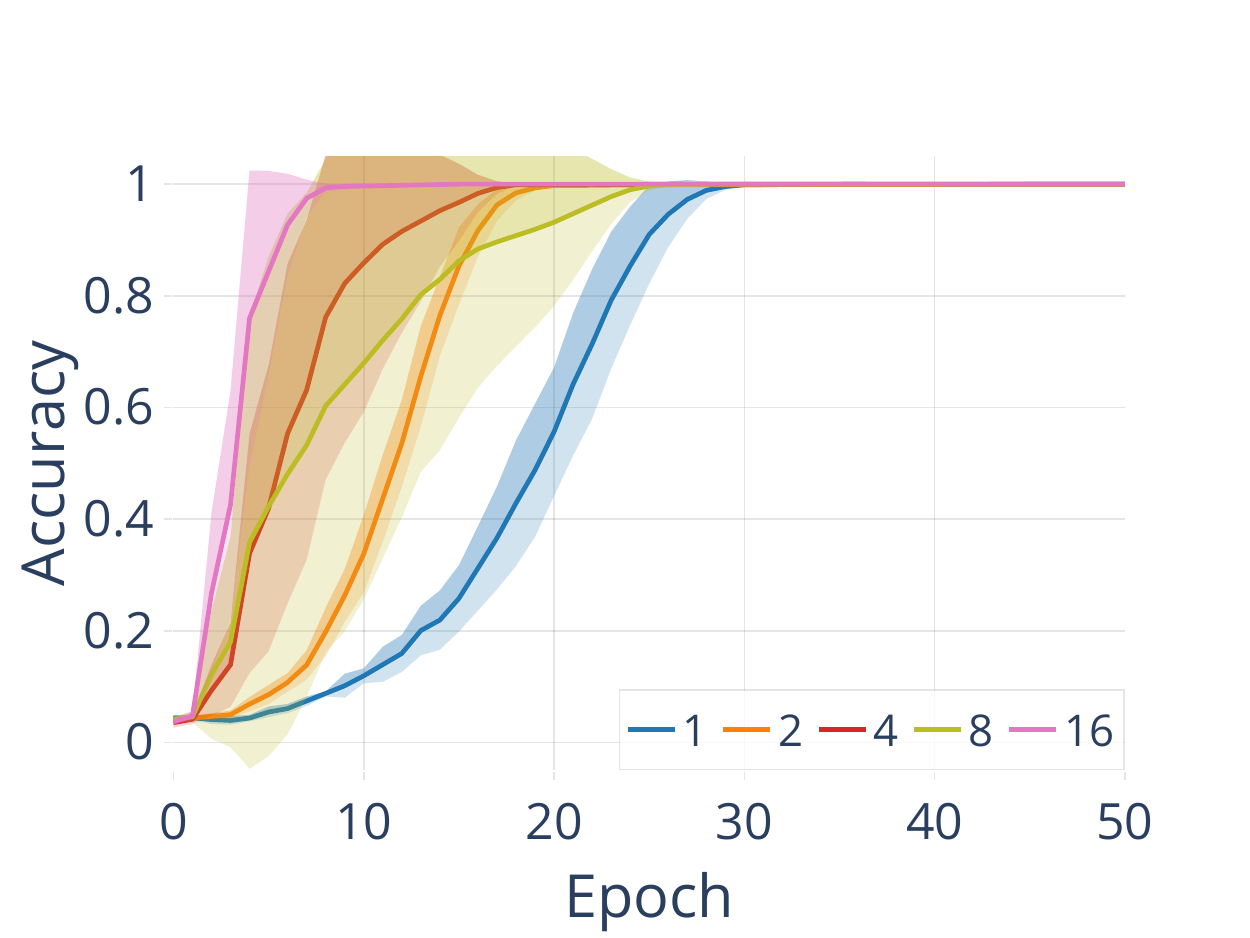}
    }
    \subfloat[Phi-2.7B, $\ell = 26$]{
        \includegraphics[width=\thirdWidth]{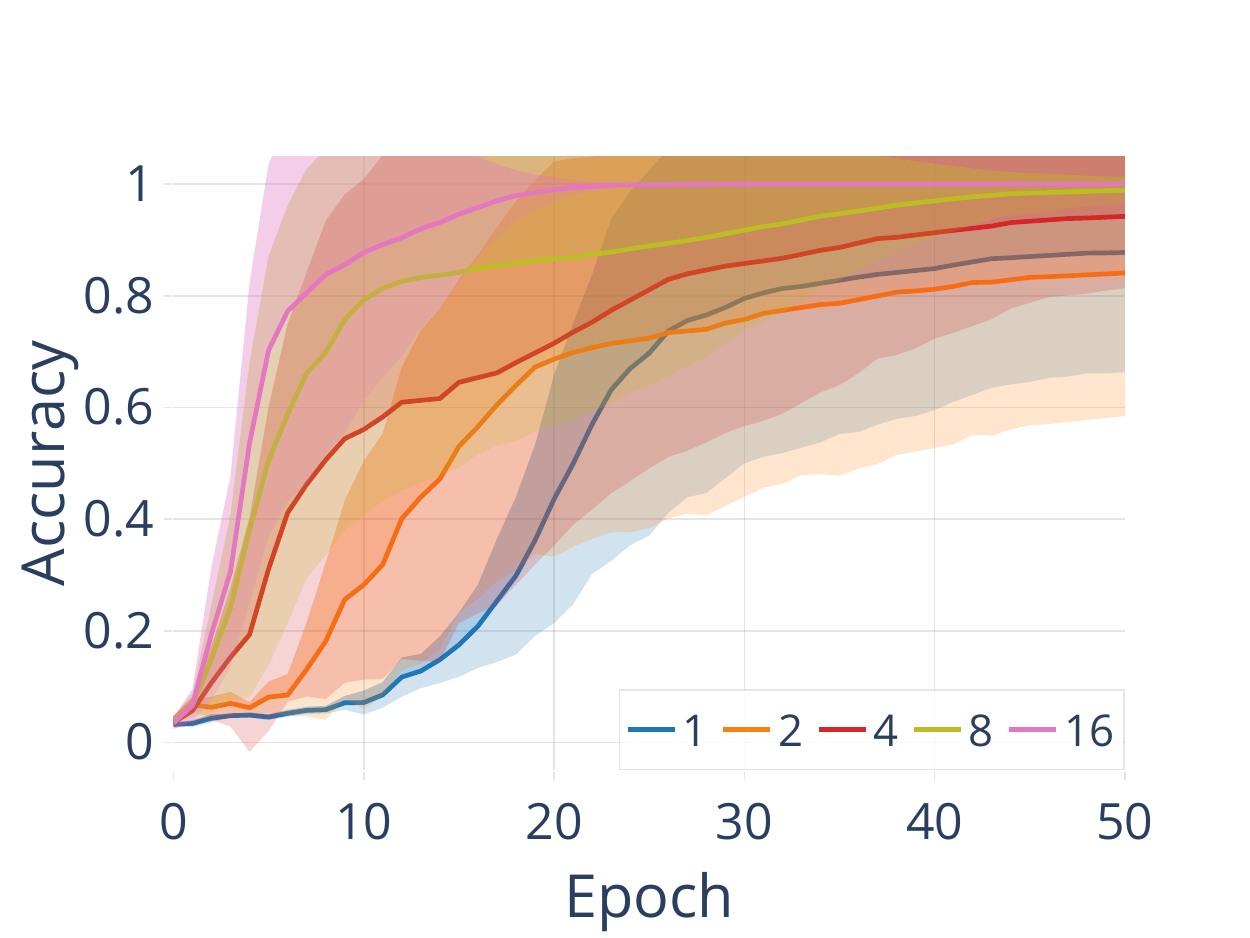}
    }
    \subfloat[Llama2-13B, $\ell = 26$]{
        \includegraphics[width=\thirdWidth]{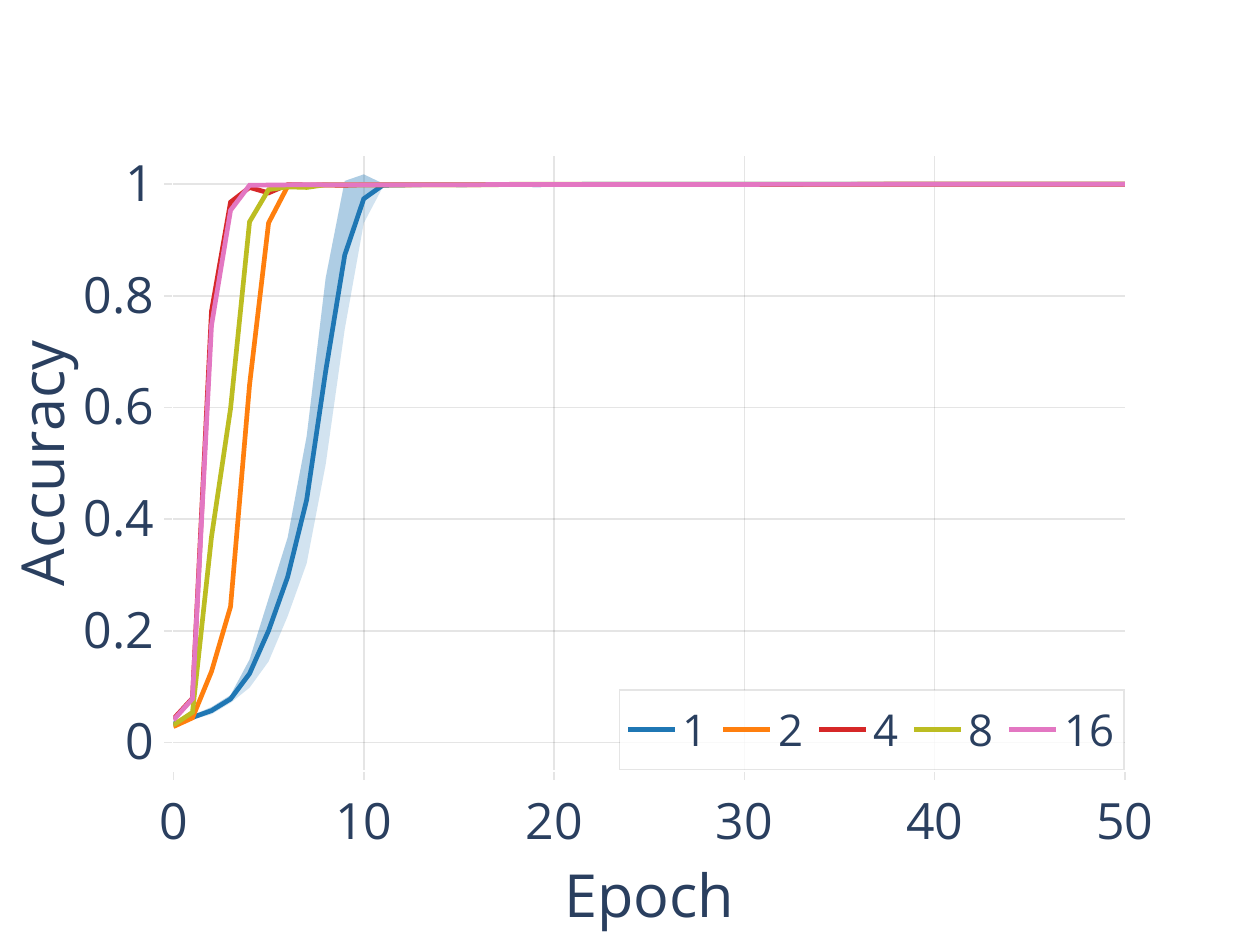}
    }
\caption{\capthead{Accuracy comparison for the initial 50 epochs for different pretrained models.}{$n = 1024$}
During repeated memorisation, models become faster at memorising new random strings.
}
\label{fig:repeated_mem_accuracy_para_all_pretrained}
\end{figure}

\begin{figure}[H]
    \centering
    \subfloat[Pythia-1B, $\ell = 2$, 32 strings]{
        \includegraphics[width=\thirdWidth]{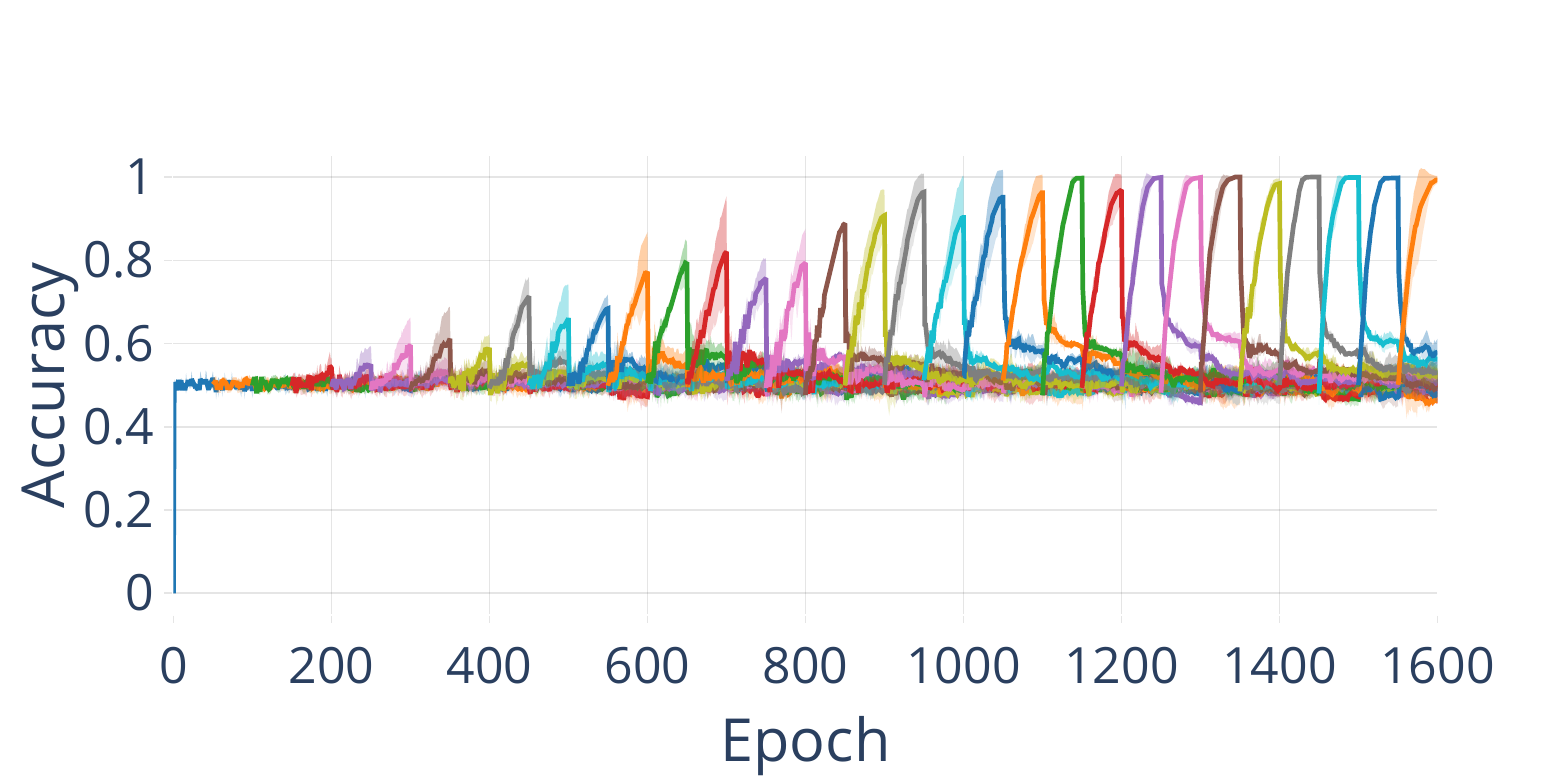}
    }
    \subfloat[Phi-2.7B, $\ell = 2$, 32 strings]{
        \includegraphics[width=\thirdWidth]{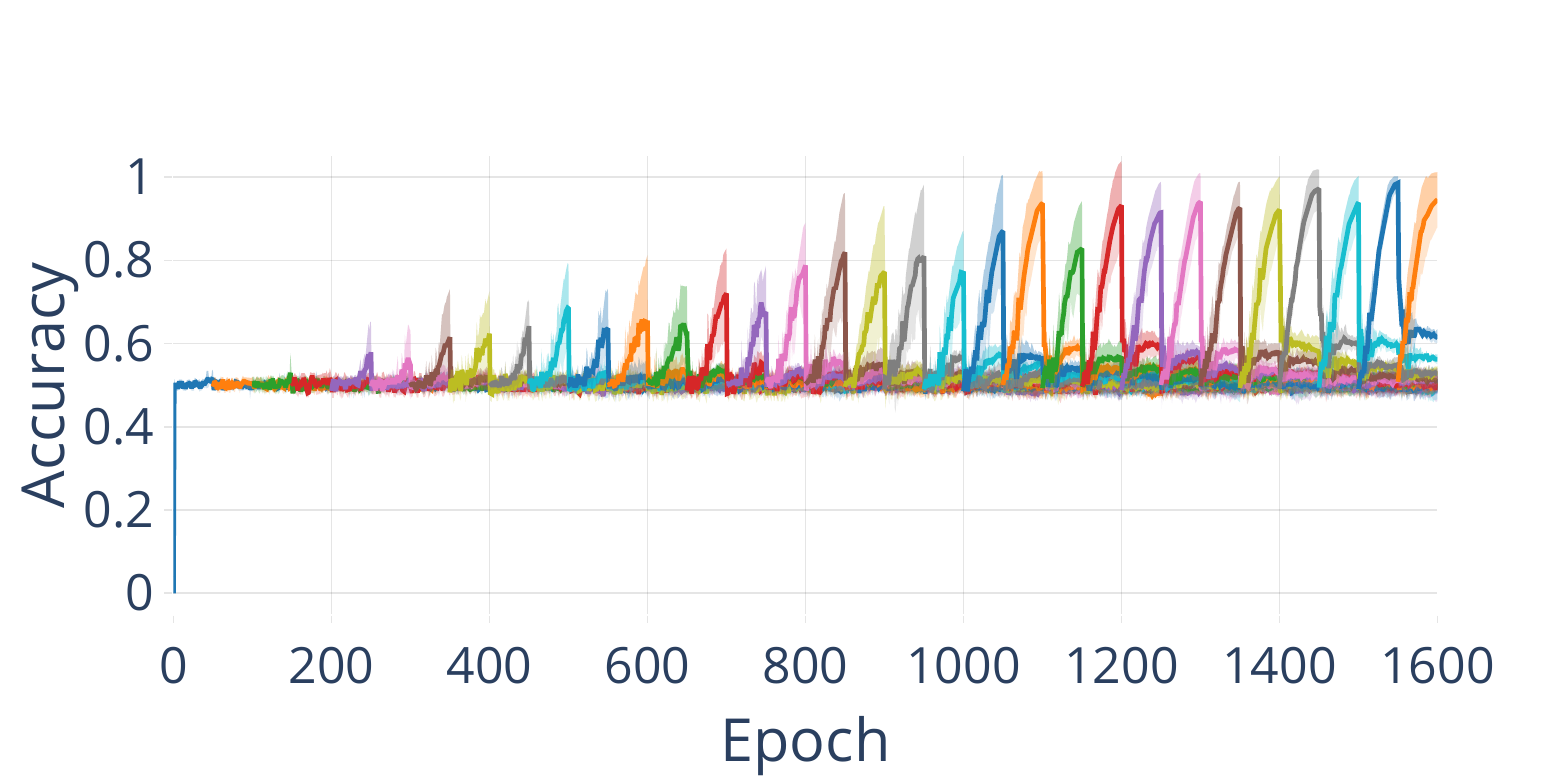}
    }
    \subfloat[Llama2-13B, $\ell = 2$, 16 strings]{
        \includegraphics[width=\thirdWidth]{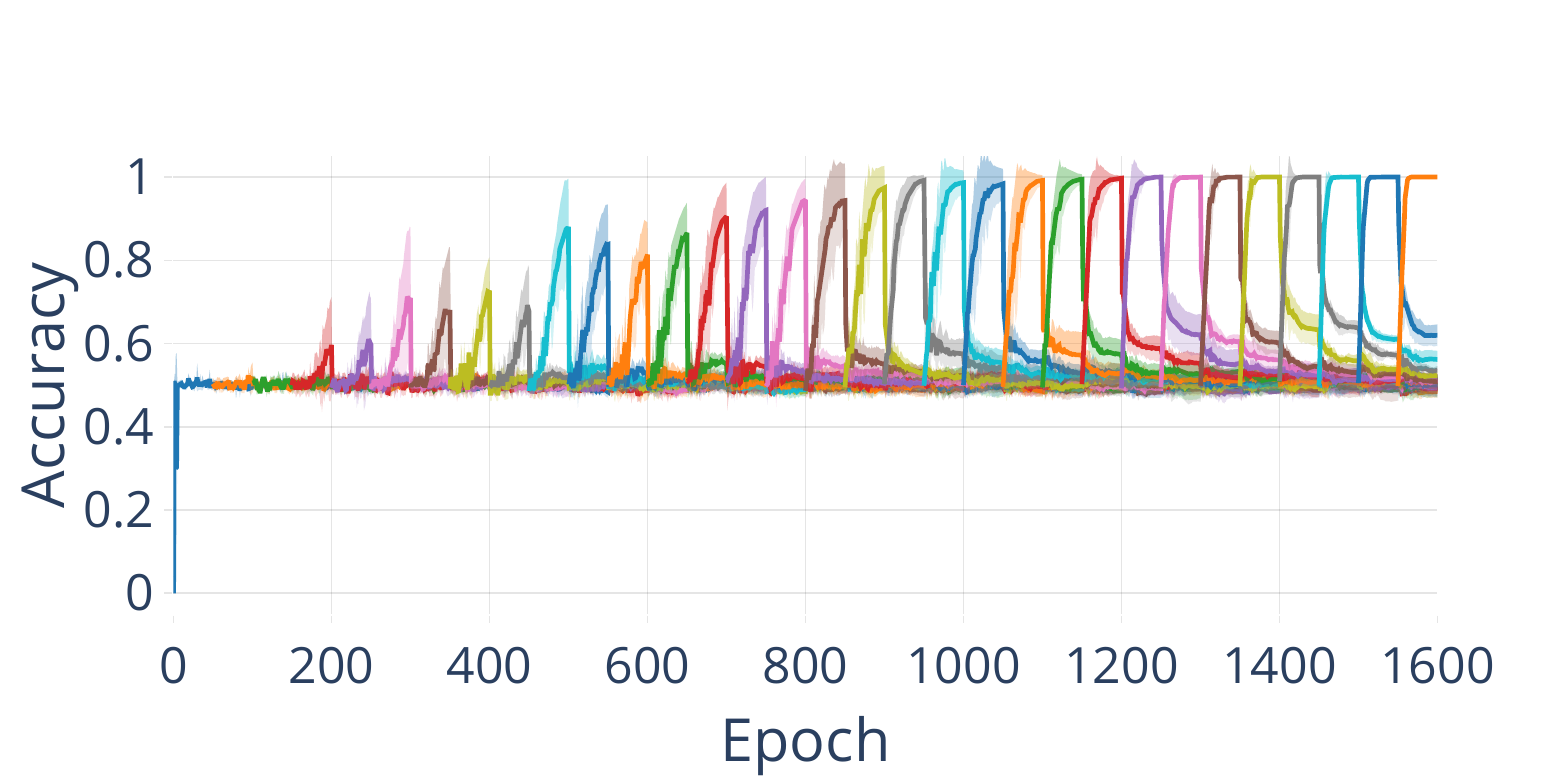}
    }
    \\
    \subfloat[Pythia-1B, $\ell = 26$, 16 strings]{
        \includegraphics[width=\thirdWidth]{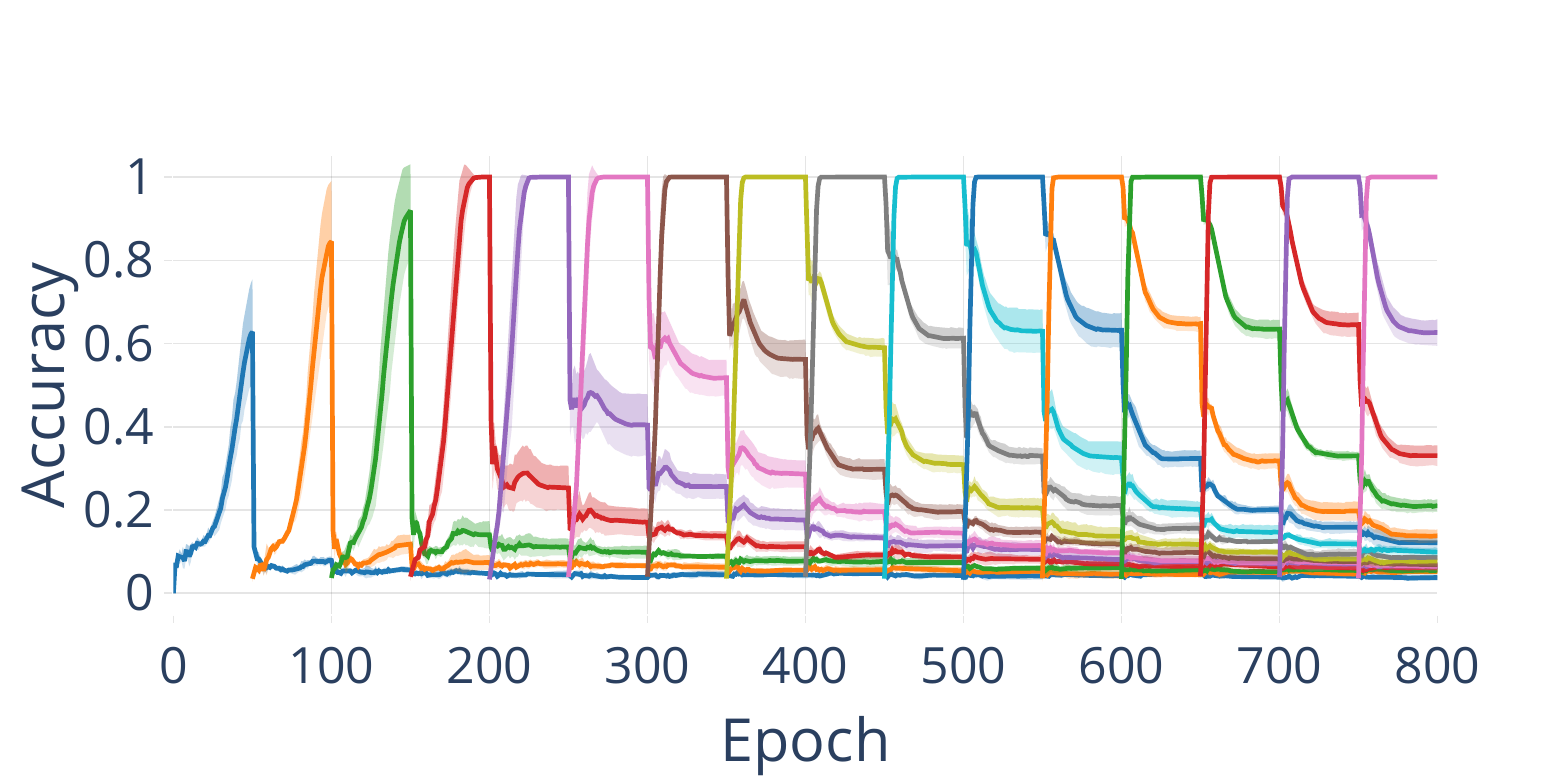}
    }
    \subfloat[Phi-2.7B, $\ell = 26$, 16 strings]{
        \includegraphics[width=\thirdWidth]{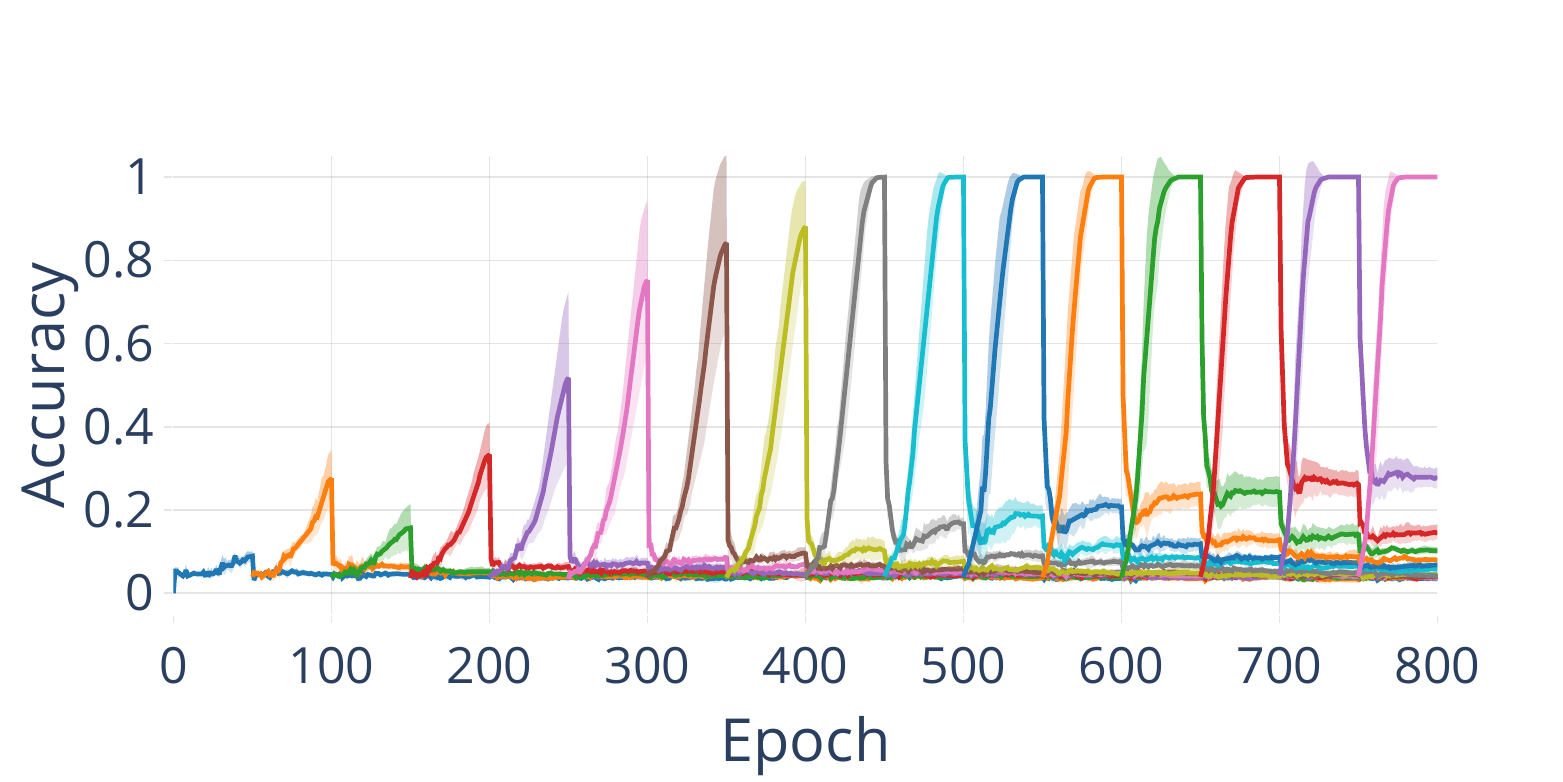}
    }
    \subfloat[Llama2-13B, $\ell = 26$, 16 strings]{
        \includegraphics[width=\thirdWidth]{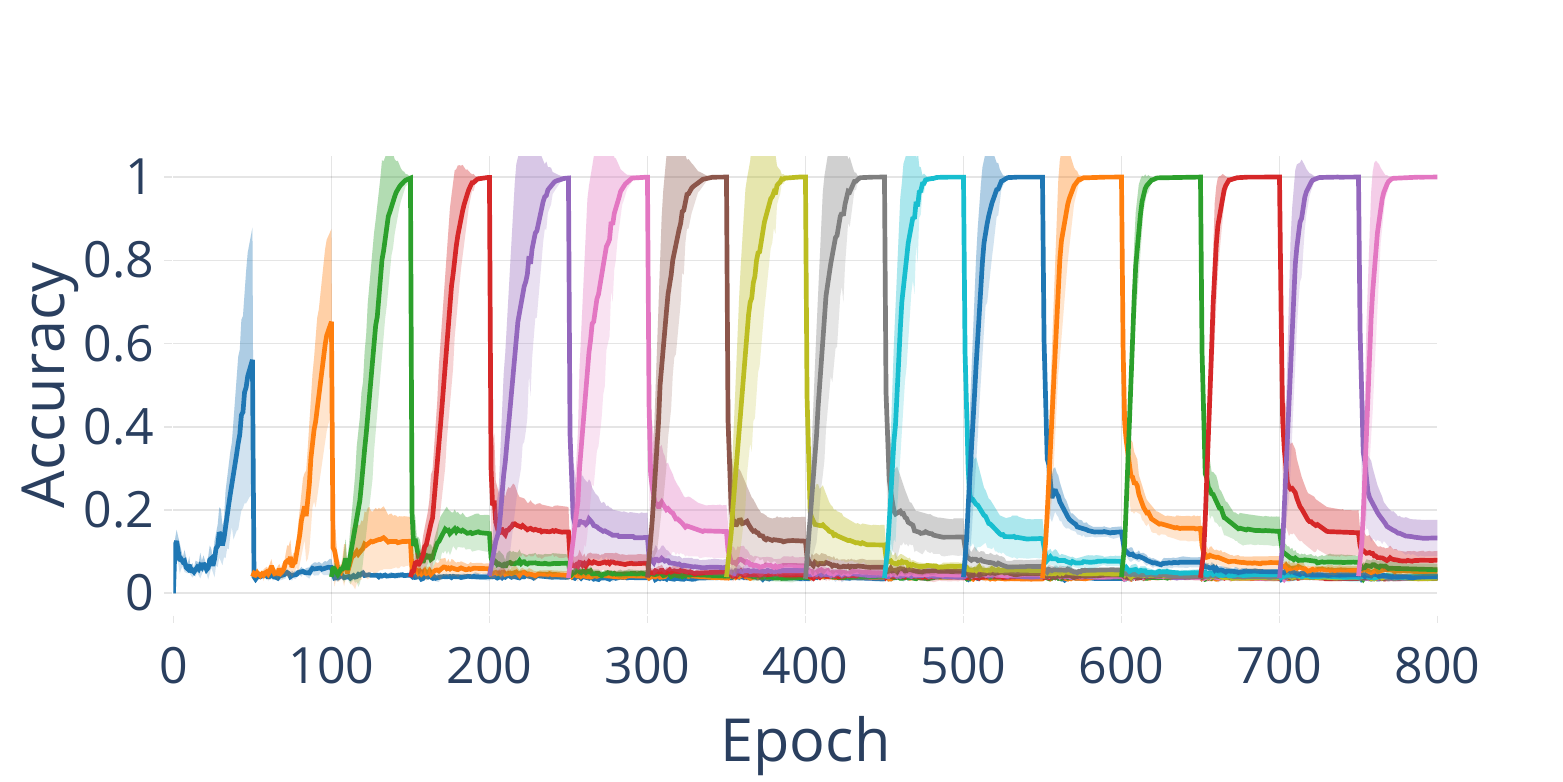}
    }
\caption{\capthead{Accuracy on repeated memorization for different untrained models.}{$n = 1024$}
As models memorise additional strings, they forget previous ones.
}
\label{fig:repeated_mem_accuracy_seq_all_untrained}
\end{figure}

\begin{figure}[H]
    \centering
    \subfloat[Pythia-1B, $\ell = 2$]{
        \includegraphics[width=\thirdWidth]{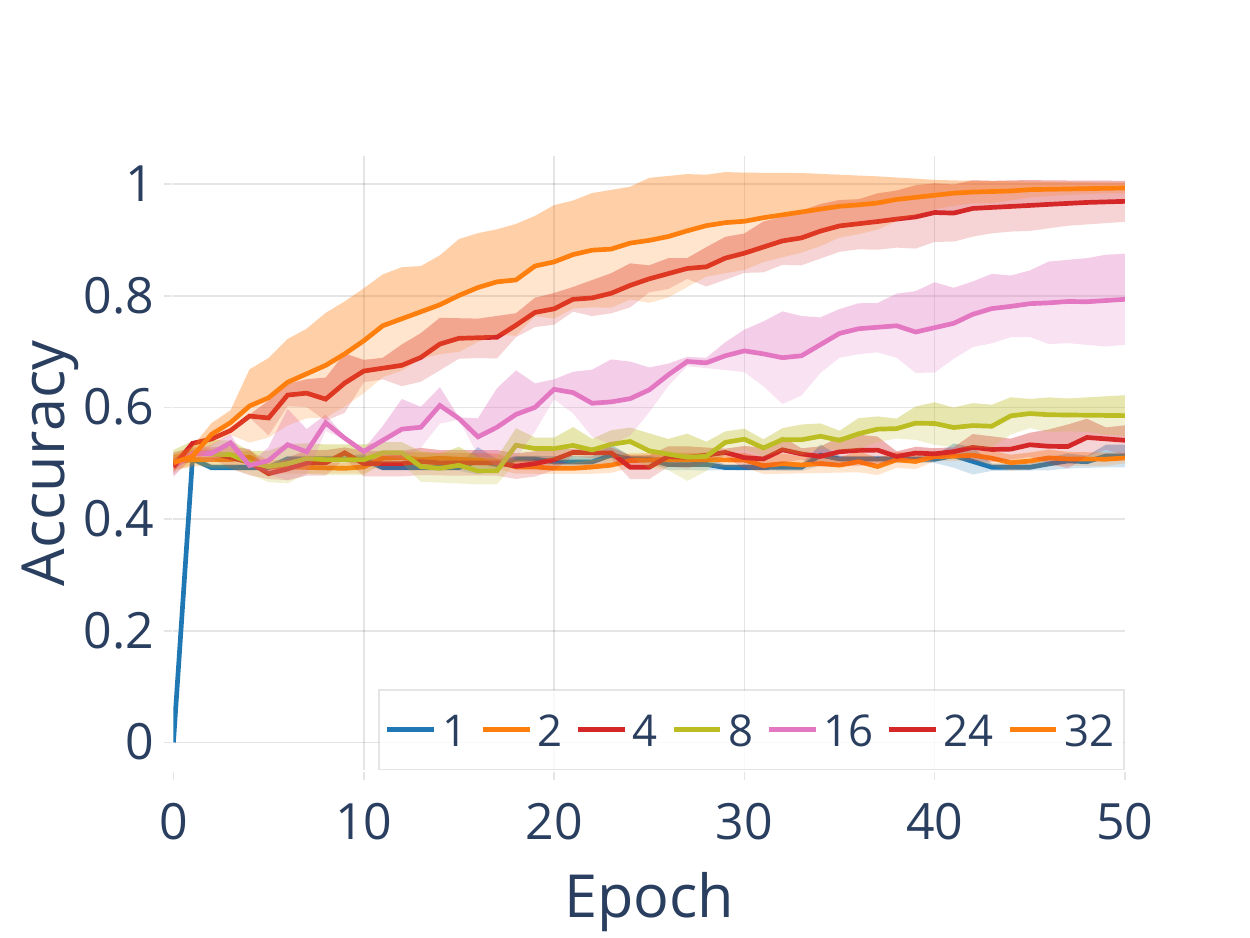}
    }
    \subfloat[Phi-2.7B, $\ell = 2$]{
        \includegraphics[width=\thirdWidth]{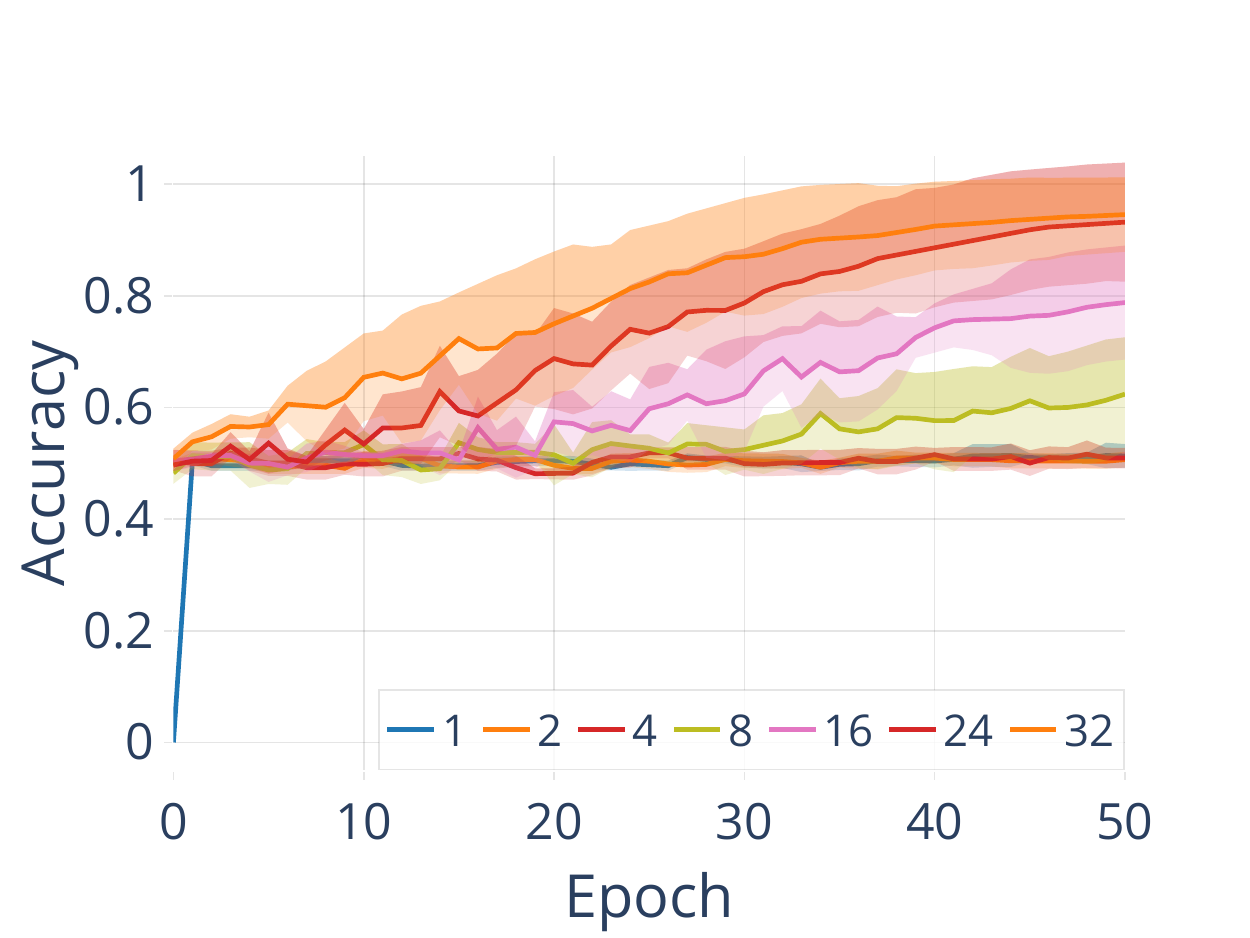}
    }
    \subfloat[Llama2-13B, $\ell = 2$]{
        \includegraphics[width=\thirdWidth]{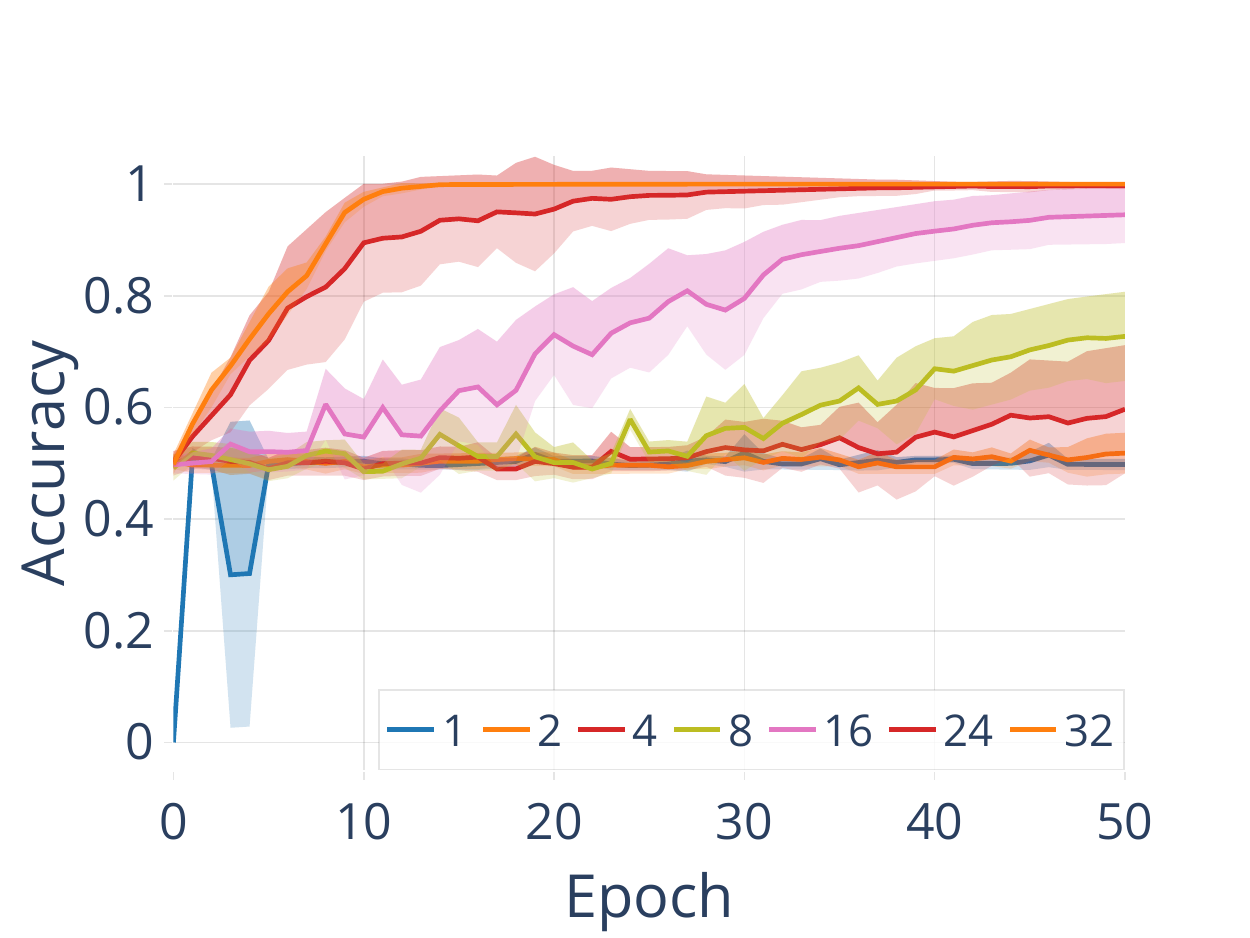}
    }
    \\
    \subfloat[Pythia-1B, $\ell = 26$]{
        \includegraphics[width=\thirdWidth]{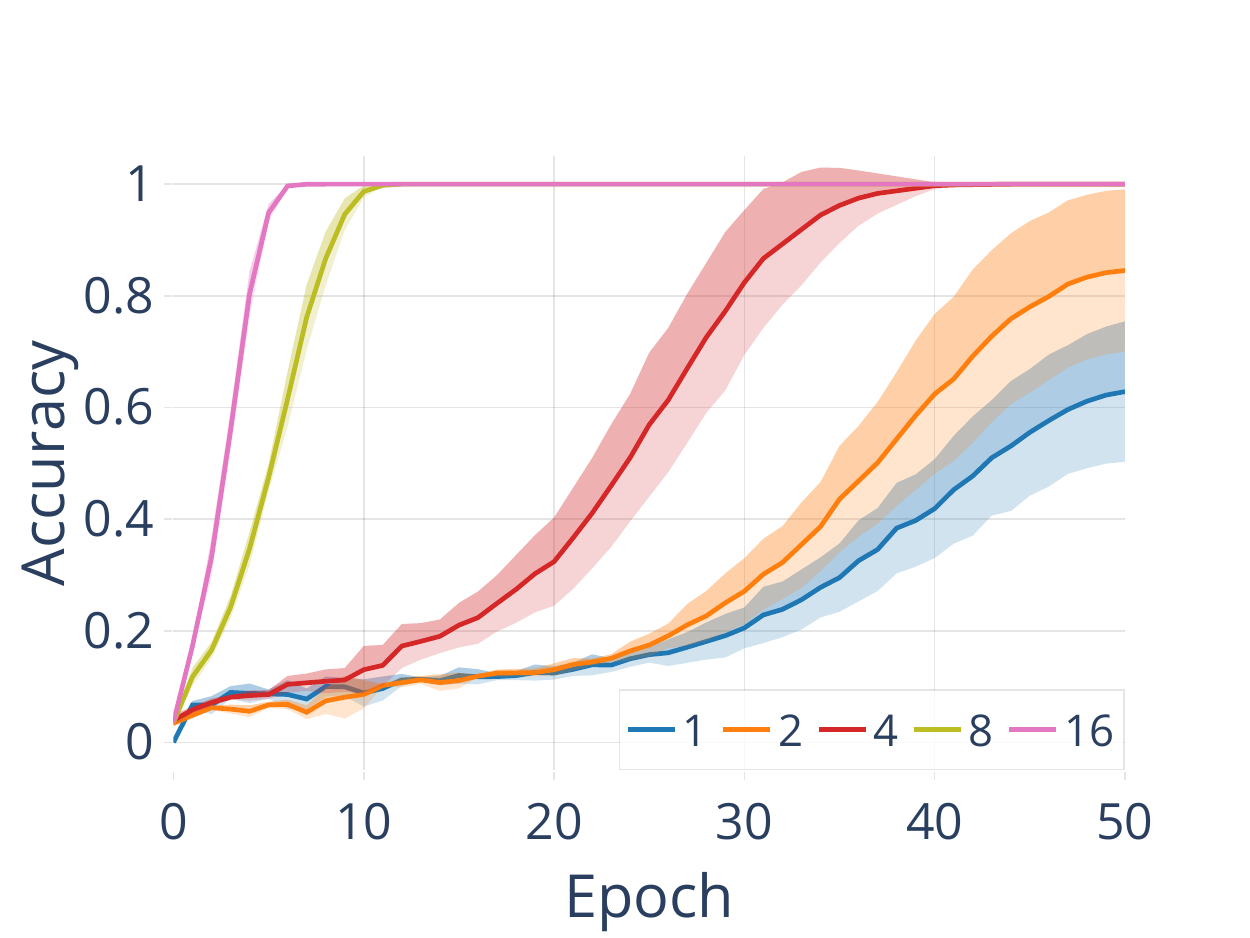}
    }
    \subfloat[Phi-2.7B, $\ell = 26$]{
        \includegraphics[width=\thirdWidth]{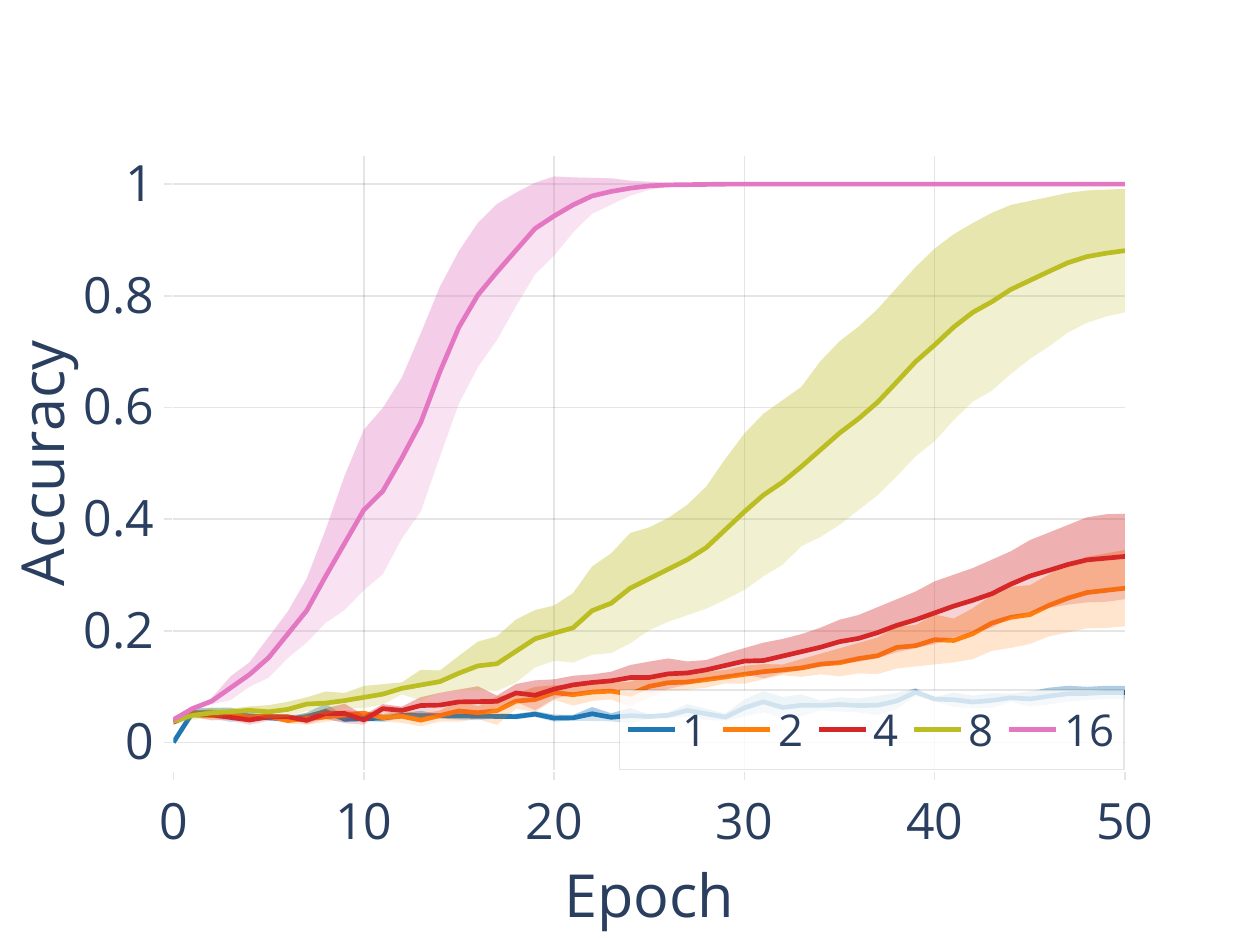}
    }
    \subfloat[Llama2-13B, $\ell = 26$]{
        \includegraphics[width=\thirdWidth]{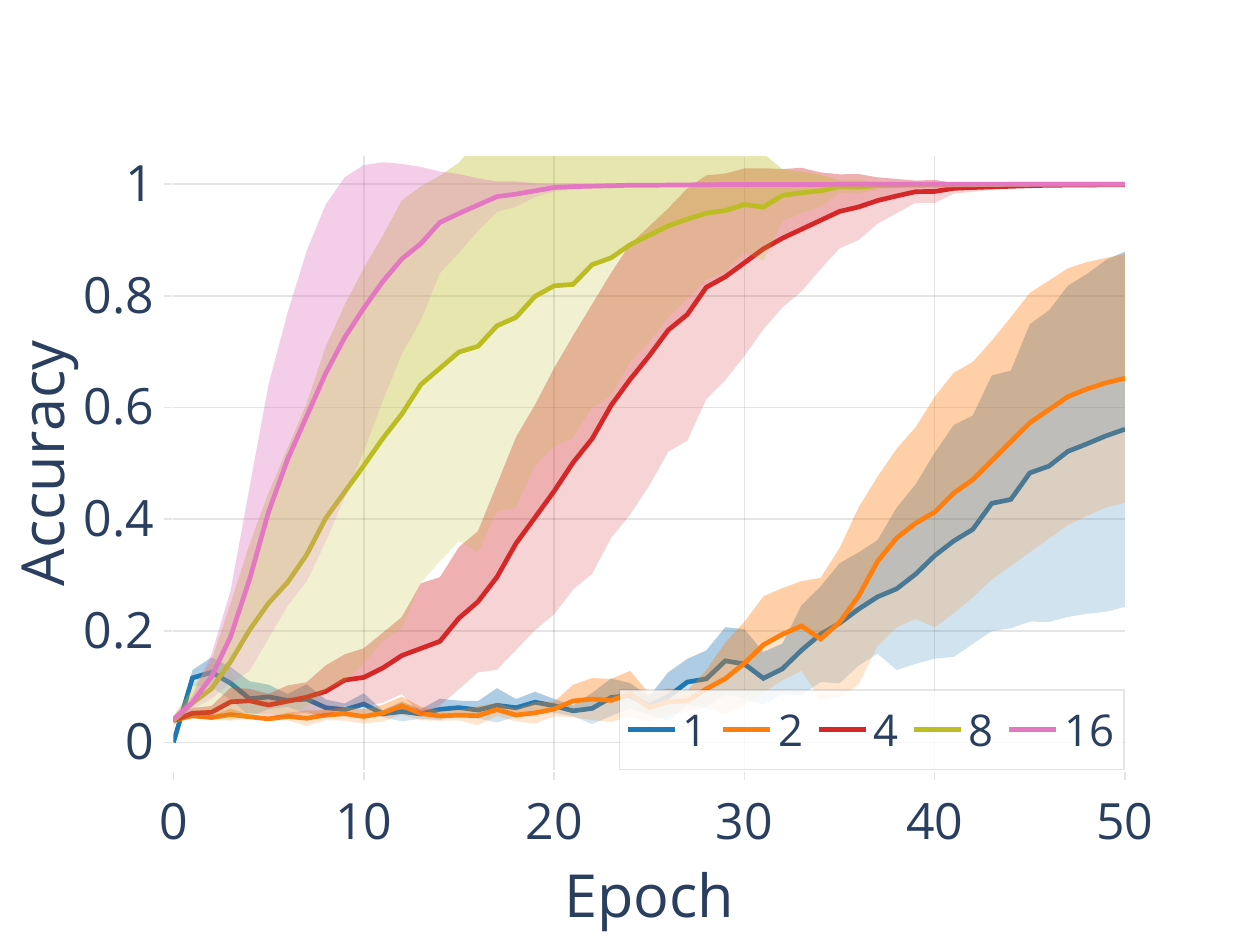}
    }
\caption{\capthead{Accuracy comparison for the initial 50 epochs for different untrained models.}{$n = 1024$}
During repeated memorisation, models become faster at memorising new random strings.
}
\label{fig:repeated_mem_accuracy_para_all_untrained}
\end{figure}

Figure~\ref{fig:repeated_mem_accuracy_seq_all_pretrained} shows results for different pretrained models for iteratively memorising a series of random strings.
Figure~\ref{fig:repeated_mem_accuracy_para_all_pretrained} shows the accuracy during the initial $50$ epochs of memorisation for the respective strings, for pretrained models.
Figures~\ref{fig:repeated_mem_accuracy_seq_all_untrained} and~\ref{fig:repeated_mem_accuracy_para_all_untrained} show the same results for untrained models.
In all cases, models become faster at memorising new strings with repetition, and forgetting of old strings slows down.

\section{Existing memorization measures can severely underestimate the degree of memorization}
\label{app:string_measure_underestimation}

\begin{figure}[H]
  \centering
    \subfloat[Individual token accuracy]{
        \includegraphics[width=\thirdWidth]{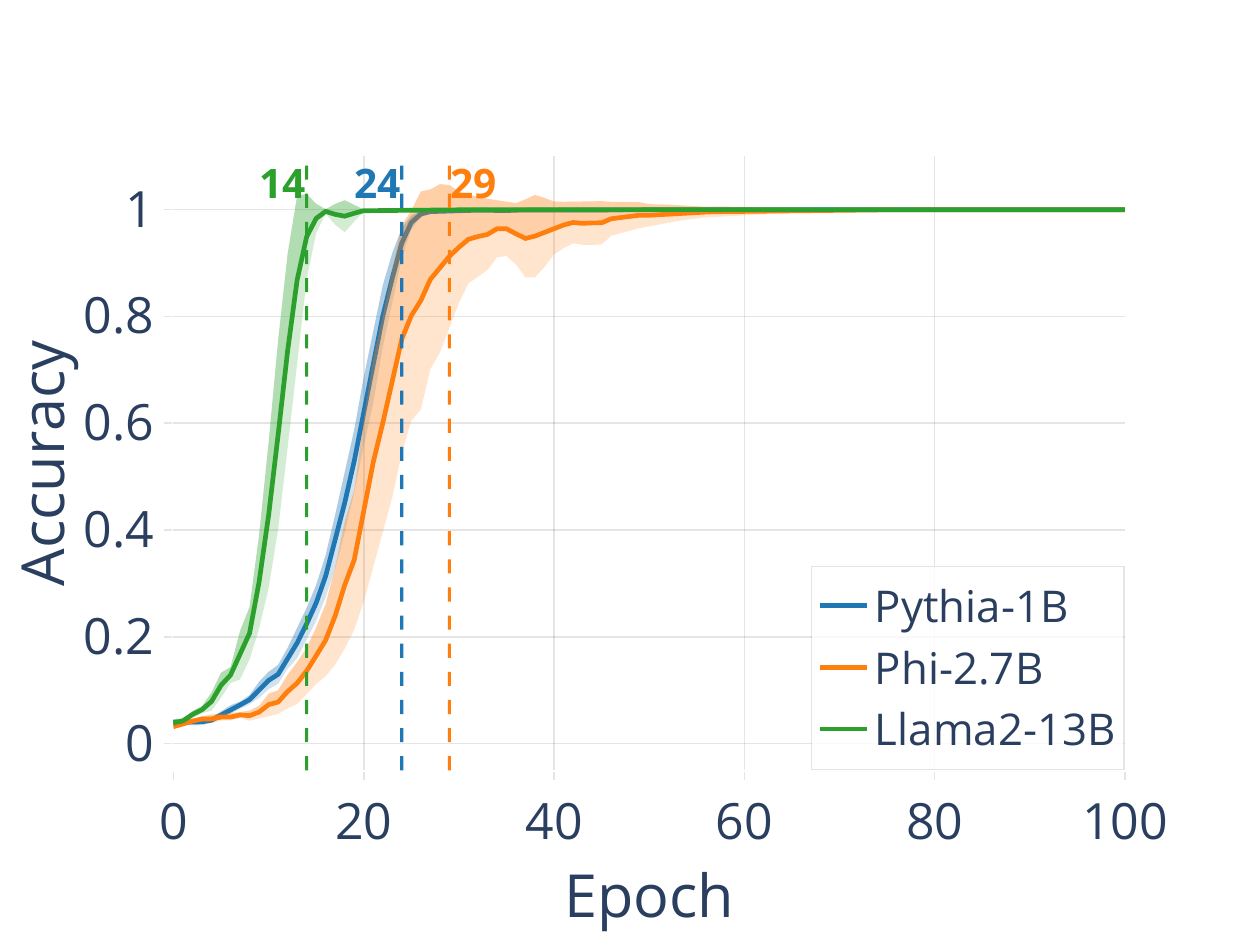}
        \label{:fig:mem_measure_accuracy}
    }
    \subfloat[50 contiguous tokens recall]{
        \includegraphics[width=\thirdWidth]{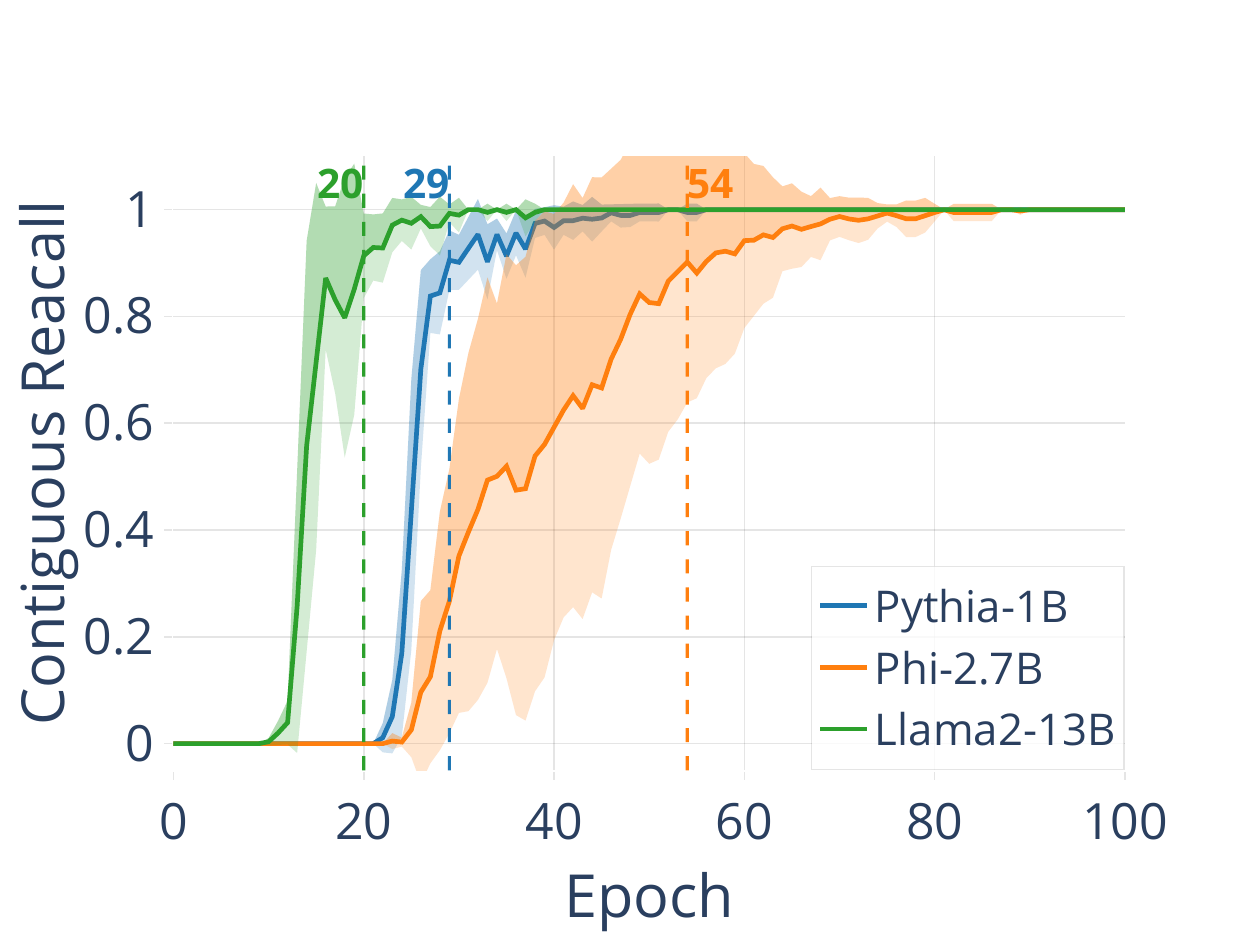}
        \label{fig:mem_measure_contig_recall}
    }
    \subfloat[Accuracy vs. contiguous recall]{
        \includegraphics[width=\thirdWidth]{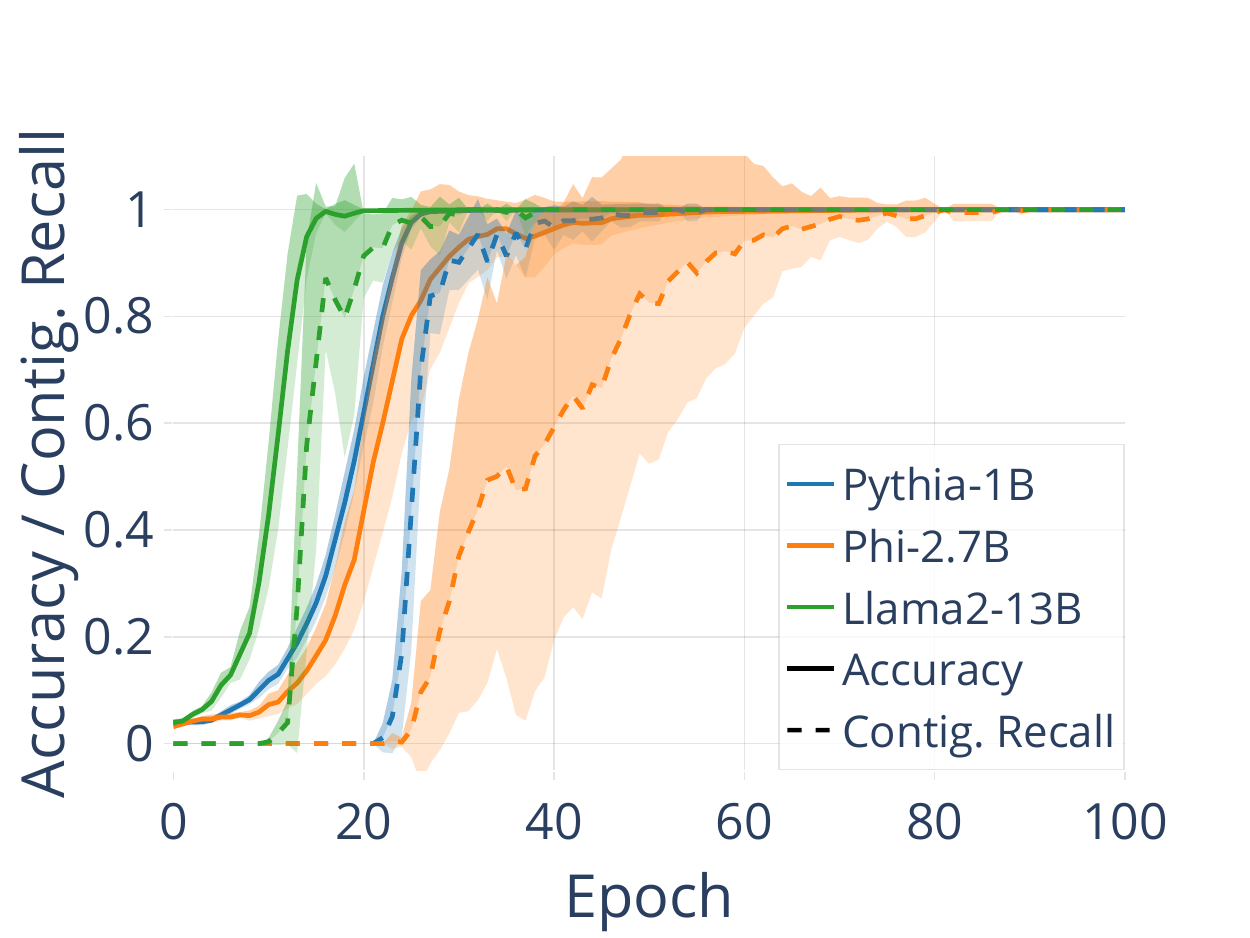}
    }

  \caption{\capthead{Applying string-based memorisation measures to random strings.}{$n=1024, \ell = 26$}
      String-based memorisation metrics severely underestimate the degree of memorisation in random strings.
      The number of substrings detect as memorised with a strict 50 correct adjacent token requirement in (b) is much lower than the prediction accuracy of the model over individual string positions in (a).
      Dashed vertical lines indicate the positions at which the respective metric first crosses the $90\%$ mark.
  }
  \label{fig:string_mem_measure}
\end{figure}

We apply the popular memorization metric from~\citet{carlini2022quantifying} in our random token string setting.
The measure detects a string $s$ as memorised by model $\mathcal{M}$ if $\mathcal{M}$ produces $s$ (with greedy decoding) when prompted with string prefix $p$, and $[ p || s]$ is contained in $\mathcal{M}$'s training data.

We apply this measure to $n = 1024$ random token strings at each training epoch of models and measure the fraction of substrings that it detects as memorised.
Analogously to~\citet{carlini2022quantifying}, we set the string length $|s|$ to 50 tokens.
Then, we detect for every contiguous position $i$ in the 1024 token string whether, when prompted with the entire preceding string, $\mathcal{M}$ predicts all of the next 50 tokens correctly, i.e. with the highest probability.
If it does then we consider tokens $[i:i + 50]$ to be memorized, otherwise not.

Figure~\ref{fig:mem_measure_contig_recall} shows what fraction of the positions in the 1024 token string the measure recollects as memorised at each epoch.
For the Pythia-1B model, recall remains essentially zero up until epoch 25, when the model already accurately predicts more than $90\%$ of the tokens (Figure~\ref{:fig:mem_measure_accuracy}).
Recall is low, because even when the string is largely memorised, models still make a small number of mispredictions, which are randomly scattered. %
Any misprediction, however, results in many substrings not being detected as memorised.
The requirement of 50 contiguous correctly predicted tokens is too strict in this case, and the contiguous string-based memorisation metric largely fails to detect memorisation.
Therefore, we argue that memorisation metrics should operate at the token- rather than at the (sub-)string level.

\end{document}